\documentclass{article}
\newif\ifemail
\newif\ifchecklist
\newif\ifappendix
\newif\ifbanner

\usepackage[final,nonatbib]{neurips_2022}\emailfalse\checklistfalse\appendixtrue\bannertrue  %

\ifbanner\else\makeatletter\renewcommand{\@noticestring}{}\makeatother\fi

\newcommand{\confnotice}{}

\usepackage[utf8]{inputenc} %
\usepackage[T1]{fontenc}    %
\usepackage{setspace}       %
\usepackage{hyperref}       %
\usepackage{url}            %
\usepackage{booktabs}       %
\usepackage{amsfonts}       %
\usepackage{nicefrac}       %
\usepackage{microtype}      %
\usepackage{color}
\usepackage{graphicx}
\usepackage{tikz}
\usepackage{amsmath,amssymb}
\usepackage{mathtools}
\usepackage[normalem]{ulem}  %
\usepackage{tabu}
\usepackage{multirow}
\usepackage{pgfplots,pgfplotstable}
\usepgfplotslibrary{fillbetween}
\usepackage{calc}
\usepackage{pbox}
\usepackage{algorithm}
\usepackage[noend]{algpseudocode}
\usepackage{caption}
\usepackage{afterpage}

\pgfplotsset{compat=1.14}	 %
\pgfplotsset{compat/show suggested version=false}
\graphicspath{{./figures/}}
\definecolor{olive}{rgb}{0.5, 0.5, 0.0}
\definecolor{maroon}{rgb}{0.69, 0.19, 0.38}
\definecolor{celestialblue}{rgb}{0.29, 0.59, 0.82}
\definecolor{darkgreen}{rgb}{0.0, 0.6, 0.0}
\definecolor{grey}{rgb}{0.5,0.5,0.5}
\definecolor{darkblue}{rgb}{0.19, 0.19, 0.62}
\definecolor{silver}{rgb}{0.7,0.7,0.7}
\definecolor{darkcyan}{rgb}{0.0, 0.55, 0.55}

\def\clap#1{\hbox to 0pt{\hss #1\hss}}%

\newcommand{\real}{\mathbb{R}}

\newcommand{\low}[1]{\raisebox{0pt}[0pt][0pt]{#1}}

\newcommand{\xx}{\boldsymbol{x}}

\newcommand{\xxh}{\boldsymbol{\hat x}}

\newcommand\undefcolumntype[1]{\expandafter\let\csname NC@find@#1\endcsname\relax}
\newcommand\forcenewcolumntype[1]{\undefcolumntype{#1}\newcolumntype{#1}}

\newcommand{\s}{\hphantom{0}}

\newcommand{\plusplus}{\raisebox{0.15ex}[0pt][0pt]{\scalebox{0.9}{++}}}
\newcommand{\ddpm}{DDPM}
\newcommand{\ddpmpp}{DDPM\plusplus}
\newcommand{\ncsnpp}{NCSN\plusplus}

\newcommand{\boldzero}{\mathbf{0}}
\newcommand{\boldi}{\mathbf{I}}
\newcommand{\sigmamax}{\sigma_{\text{max}}}
\newcommand{\odetime}{t}
\newcommand{\scale}{s}
\newcommand{\diff}{\mathrm{d}}
\newcommand{\signal}{\boldsymbol{y}}

\newcommand{\noise}{\boldsymbol{n}}

\newcommand{\wproc}{\omega_\odetime}
\newcommand{\churn}{\beta}
\newcommand{\dd}{\boldsymbol{d}}

\newcommand{\ddp}{{\boldsymbol{d}\mathrlap{\hspace*{.075em}'}}}

\newcommand{\pdata}{p_\text{data}}
\newcommand{\pdatahat}{\hat{p}_\text{data}}
\newcommand{\lte}{\boldsymbol{\tau}}
\newcommand{\gte}{\boldsymbol{e}}
\newcommand{\ptrain}{p_\text{train}}

\newcommand{\nablaxx}{\nabla_{\hspace{-0.5mm}\xx}}
\newcommand{\Deltaxx}{{\Delta_{\xx}}}

\DeclareMathOperator{\argmin}{arg\,min}

\DeclareMathOperator{\Var}{Var}
\DeclareMathOperator{\Cov}{Cov}
\DeclareMathOperator{\score}{score}

\newcommand{\stext}[1]{\text{\raisebox{0pt}[0pt][0pt]{#1}}}
\newcommand{\smin}{\sigma_\stext{min}}
\newcommand{\smax}{\sigma_\stext{max}}
\newcommand{\sdyn}{\sigma_\stext{d}}

\newcommand{\bmin}{\beta_\text{min}}
\newcommand{\bmax}{\beta_\text{max}}
\newcommand{\bdyn}{\beta_\text{d}}
\newcommand{\sdata}{\sigma_\text{data}}
\newcommand{\cskip}{c_\text{skip}}
\newcommand{\cout}{c_\text{out}}
\newcommand{\cin}{c_\text{in}}
\newcommand{\cnoise}{c_\text{noise}}

\newcommand{\origT}[1]{u_{#1}}
\newcommand{\origTsup}[2]{u_{#1}^{#2}}
\newcommand{\origN}{M}

\newcommand{\Schurn}{S_\text{churn}}
\newcommand{\Stmin}{S_\text{tmin}}
\newcommand{\Stmax}{S_\text{tmax}}
\newcommand{\Snoise}{S_\text{noise}}
\newcommand{\StminStmax}{S_\text{tmin,tmax}}

\newcommand{\confhdr}[1]{\makebox[1em]{\textsc{#1}}}

\newcommand{\vparagraph}[1]{\vspace*{-1mm}\paragraph{#1}}

\newcommand{\pd}{q}
\newcommand{\condu}{\kappa}
\newcommand{\freq}{\boldsymbol{\nu}}
\newcommand{\fp}{r}
\newcommand{\ff}{\boldsymbol{f}}
\newcommand{\gb}{\boldsymbol{g}}
\newcommand{\Fargs}{\scalebox{0.85}{$\cin(\sigma) (\signal {+} \noise); \cnoise(\sigma)$}}

\pgfplotsset{xtick style={draw=none}}
\pgfplotsset{ytick style={draw=none}}
\pgfplotsset{major grid style={gray!40}}
\pgfplotsset{every axis plot/.style={thick, mark size=1.5pt}}
\pgfplotsset{legend image code/.code={\draw[mark repeat=2, mark phase=2] plot coordinates {(0cm, 0cm) (0.2cm, 0cm) (0.4cm, 0cm)};}} %

\definecolor{C0}{rgb}{0.121569, 0.466667, 0.705882}
\definecolor{C1}{rgb}{1.000000, 0.498039, 0.054902}
\definecolor{C2}{rgb}{0.172549, 0.627451, 0.172549}
\definecolor{C3}{rgb}{0.839216, 0.152941, 0.156863}
\definecolor{C4}{rgb}{0.580392, 0.403922, 0.741176}
\definecolor{C5}{rgb}{0.549020, 0.337255, 0.294118}
\definecolor{C6}{rgb}{0.890196, 0.466667, 0.760784}
\definecolor{C7}{rgb}{0.498039, 0.498039, 0.498039}
\definecolor{C8}{rgb}{0.737255, 0.741176, 0.133333}
\definecolor{C9}{rgb}{0.090196, 0.745098, 0.811765}

\newcommand{\fillbetween}[3][]{\addplot+[name path=A, draw=none, mark=none, forget plot] #2; \addplot+[name path=B, draw=none, mark=none, forget plot] #3; \addplot[#1] fill between[of=A and B]}

\newcommand{\cs}{}
\newcommand{\hh}{0mm}
\newcommand{\hhh}{0mm}

\newcommand{\vv}{0mm}

\newcommand{\vlabel}[3]{\makebox[0mm][l]{\rotatebox{90}{\makebox[#2][c]{#3}}}\hspace{#1}}
\newcommand{\hrlabel}[1]{\hfill\makebox[0mm]{#1}\hfill}
\newcommand{\vrlabel}[1]{\vfill\makebox[0mm]{#1}\vfill}

\newcommand{\tickYtopD}[1]{\raisebox{-1.5ex}[0ex][0ex]{#1}}
\newcommand{\tickFID}{\tickYtopD{FID}}
\newcommand{\tickLoss}{\tickYtopD{loss}}
\newcommand{\tickTau}{\tickYtopD{$\lVert\lte\rVert$}}
\newcommand{\tickNFE}[1]{\smash{NFE$=$}{$#1$}\hspace*{1em}}

\newcommand{\tickSchurnB}[1]{\smash{$\Schurn{=}$}{$#1$}\hspace*{2em}}
\newcommand{\tickRho}[1]{$\mathllap{\smash{\rho{=}}}{#1}$}
\newcommand{\tickSigma}[1]{$\mathllap{\smash{\sigma{=}}}{#1}$}

\newcommand{\atphantom}{\vphantom{${}^2$}}
\newcommand{\AProcedure}[2]{\Procedure{\smash{#1}}{\smash{#2}}}
\newcommand{\AComment}[1]{\Comment{\smash{#1}}}
\newcommand{\AState}[1]{\State{\smash{#1}}}
\newcommand{\AFor}[1]{\For{\smash{#1}}}
\newcommand{\AIf}[1]{\If{\smash{#1}}}
\newcommand{\DState}[1]{\State{#1 \vphantom{$\displaystyle\Bigg)$}}}

\newcommand{\genOdePlotNfeCifaruDdpmppRepro}{%
(8, 244.79)
(12, 145.549)
(16, 92.4244)
(24, 56.5574)
(32, 46.3142)
(48, 38.0137)
(64, 31.1644)
(96, 20.3582)
(128, 12.148)
(192, 4.4652)
(256, 2.9402)
(384, 3.1828)
(512, 3.0842)
(768, 2.858)
(1024, 2.8546)
}

\newcommand{\genOdePlotNfeCifaruDdpmppEmu}{%
(8, 156.316)
(12, 74.4001)
(16, 43.5768)
(24, 21.9297)
(32, 13.99)
(48, 7.8499)
(64, 5.6903)
(96, 4.1566)
(128, 3.6748)
(192, 3.153)
(256, 3.0037)
(384, 2.8784)
(512, 2.8504)
(768, 2.7936)
(1024, 2.8185)
}

\newcommand{\genOdePlotNfeCifaruDdpmppDHeun}{%
(7, 286.817)
(9, 278.413)
(11, 156.389)
(13, 131.049)
(15, 70.5949)
(19, 16.5407)
(23, 9.5407)
(27, 7.0445)
(31, 5.8043)
(35, 4.8917)
(39, 4.412)
(47, 3.8164)
(55, 3.4925)
(63, 3.2728)
(79, 3.1378)
(95, 3.0531)
(127, 2.9955)
(191, 2.9862)
(255, 2.9365)
(383, 2.9427)
(511, 2.8835)
(767, 2.9415)
(1023, 2.9355)
}

\newcommand{\genOdePlotNfeCifaruDdpmppSched}{%
(7, 105.812)
(9, 45.9258)
(11, 18.4101)
(13, 9.6882)
(15, 6.5231)
(19, 4.1687)
(23, 3.519)
(27, 3.1906)
(31, 3.1026)
(35, 3.0139)
(39, 3.0129)
(47, 2.9913)
(55, 2.9415)
(63, 2.946)
(79, 2.959)
(95, 2.9596)
(127, 2.9854)
(191, 3.0103)
(255, 2.9333)
(383, 2.9701)
(511, 2.9325)
(767, 2.985)
(1023, 2.9644)
(1535, 2.9772)
(2047, 2.9684)
}

\newcommand{\genOdePlotNfeCifaruDdpmppBbox}{%
(21, 434.322)
(27, 423.151)
(51.0612, 147.627)
(87, 16.467)
(97.4388, 5.6185)
(114.75, 2.9403)
(141.26, 2.9745)
(159.918, 2.9407)
(224.066, 2.9702)
(525.934, 2.9448)
(1109.43, 2.9587)
}

\newcommand{\genOdePlotNfeCifaruNcsnppRepro}{%
(8, 456.981)
(12, 457.102)
(16, 457.398)
(24, 457.399)
(32, 455.984)
(48, 449.205)
(64, 432.536)
(96, 373.228)
(128, 326.381)
(192, 245.84)
(256, 161.473)
(384, 66.668)
(512, 33.894)
(768, 15.3301)
(1024, 10.4541)
(1536, 7.4372)
(2048, 6.5582)
(3072, 5.9151)
(4096, 5.8765)
(6144, 5.6732)
(8192, 5.4525)
}

\newcommand{\genOdePlotNfeCifaruNcsnppEmu}{%
(8, 457.325)
(12, 457.728)
(16, 457.889)
(24, 458.013)
(32, 456.729)
(48, 450.088)
(64, 433.201)
(96, 373.315)
(128, 325.876)
(192, 241.479)
(256, 156.102)
(384, 62.9697)
(512, 31.8794)
(768, 14.767)
(1024, 9.9014)
(1536, 7.0364)
(2048, 6.1636)
(3072, 5.4476)
(4096, 5.1215)
(6144, 4.9831)
(8192, 4.781)
}

\newcommand{\genOdePlotNfeCifaruNcsnppDHeun}{%
(7, 460.925)
(9, 453.522)
(11, 393.896)
(13, 310.143)
(15, 302.281)
(19, 305.288)
(23, 334.292)
(27, 302.039)
(31, 233.539)
(35, 164.495)
(39, 116.967)
(47, 66.6204)
(55, 42.6327)
(63, 29.5286)
(79, 15.2948)
(95, 9.7384)
(127, 5.6221)
(191, 4.2859)
(255, 4.2273)
(383, 4.3703)
(511, 4.4843)
(767, 4.5162)
(1023, 4.5522)
(1535, 4.5386)
(2047, 4.6152)
(3071, 4.6034)
(4095, 4.5927)
(6143, 4.6197)
(8191, 4.5827)
}

\newcommand{\genOdePlotNfeCifaruNcsnppSched}{%
(7, 143.823)
(9, 38.6784)
(11, 13.4623)
(13, 7.7137)
(15, 5.6188)
(19, 4.3772)
(23, 4.0435)
(27, 3.8237)
(31, 3.8236)
(35, 3.7742)
(39, 3.7945)
(47, 3.7868)
(55, 3.7343)
(63, 3.8021)
(79, 3.7876)
(95, 3.7879)
(127, 3.8128)
(191, 3.7998)
(255, 3.8232)
(383, 3.7629)
(511, 3.8202)
(767, 3.7767)
(1023, 3.76)
(1535, 3.82)
(2047, 3.7658)
(3071, 3.7654)
(4095, 3.838)
(6143, 3.8228)
(8191, 3.8304)
}

\newcommand{\genOdePlotNfeCifaruNcsnppBbox}{%
(21, 423.935)
(32.8469, 213.618)
(62.4184, 12.0016)
(93.1224, 3.74)
(100.883, 3.6874)
(133.745, 3.7152)
(156.245, 3.8563)
(325.01, 3.8077)
(845.281, 3.8421)
(1877.83, 3.821)
(3480.02, 3.7652)
(5527.56, 3.7869)
(8716.48, 3.8441)
}

\newcommand{\genOdePlotNfeImgcDhariwalRepro}{%
(8, 18.8955)
(10, 13.556)
(20, 6.7867)
(25, 5.723)
(40, 4.3655)
(50, 3.9373)
(100, 3.1669)
(125, 3.041)
(200, 2.953)
(250, 2.9108)
(500, 2.8486)
(1000, 2.9062)
}

\newcommand{\genOdePlotNfeImgcDhariwalEmu}{%
(8, 14.7243)
(12, 8.3666)
(16, 6.264)
(24, 4.6733)
(32, 4.1784)
(48, 3.637)
(64, 3.3889)
(96, 3.1523)
(128, 3.0834)
(192, 2.9613)
(256, 2.8951)
(384, 2.7995)
(512, 2.7583)
(768, 2.735)
(992, 2.7333)
}

\newcommand{\genOdePlotNfeImgcDhariwalDHeun}{%
(7, 87.1633)
(9, 33.6137)
(11, 12.3924)
(13, 6.6292)
(15, 4.7799)
(19, 3.5569)
(23, 3.1176)
(27, 2.9263)
(31, 2.8375)
(35, 2.8157)
(39, 2.7784)
(47, 2.7296)
(55, 2.7168)
(63, 2.7305)
(79, 2.6611)
(95, 2.6714)
(127, 2.6489)
(191, 2.6944)
(255, 2.6363)
(383, 2.6425)
(511, 2.7042)
(767, 2.6387)
(1023, 2.6936)
(1535, 2.6846)
(1983, 2.6495)
}

\newcommand{\genOdePlotNfeImgcDhariwalBbox}{%
(21, 399.389)
(27, 352.735)
(39, 216.422)
(62.9617, 15.5028)
(98.1658, 3.3586)
(130.661, 2.6565)
(168.857, 2.7373)
(442.198, 2.6904)
}

\newcommand{\genOdePlotNfeCifaruDdpmppReproMarks}{%
(256, 2.9402)
}

\newcommand{\genOdePlotNfeCifaruDdpmppEmuMarks}{%
(512, 2.8504)
}

\newcommand{\genOdePlotNfeCifaruDdpmppDHeunMarks}{%
(255, 2.9365)
}

\newcommand{\genOdePlotNfeCifaruDdpmppSchedMarks}{%
(35, 3.0139)
}

\newcommand{\genOdePlotNfeCifaruDdpmppBboxMarks}{%
(114.75, 2.9403)
}

\newcommand{\genOdePlotNfeCifaruNcsnppReproMarks}{%
(8192, 5.4525)
}

\newcommand{\genOdePlotNfeCifaruNcsnppEmuMarks}{%
(8192, 4.781)
}

\newcommand{\genOdePlotNfeCifaruNcsnppDHeunMarks}{%
(191, 4.2859)
}

\newcommand{\genOdePlotNfeCifaruNcsnppSchedMarks}{%
(27, 3.8237)
}

\newcommand{\genOdePlotNfeCifaruNcsnppBboxMarks}{%
(93.1224, 3.74)
}

\newcommand{\genOdePlotNfeImgcDhariwalReproMarks}{%
(250, 2.9108)
}

\newcommand{\genOdePlotNfeImgcDhariwalEmuMarks}{%
(384, 2.7995)
}

\newcommand{\genOdePlotNfeImgcDhariwalDHeunMarks}{%
(79, 2.6611)
}

\newcommand{\genOdePlotNfeImgcDhariwalBboxMarks}{%
(130.661, 2.6565)
}

\newcommand{\genSdePlotNfeCifaruDdpmppOde}{%
(9, 45.9258)
(11, 18.4101)
(13, 9.6882)
(15, 6.5231)
(19, 4.1687)
(23, 3.519)
(27, 3.1906)
(31, 3.1026)
(35, 3.0139)
(39, 3.0129)
(47, 2.9913)
(55, 2.9415)
(63, 2.946)
(79, 2.959)
(95, 2.9596)
(127, 2.9854)
(191, 3.0103)
(255, 2.9333)
(383, 2.9701)
(511, 2.9325)
(767, 2.985)
(1023, 2.9644)
(1535, 2.9772)
(2047, 2.9684)
}

\newcommand{\genSdePlotNfeCifaruDdpmppPlain}{%
(11, 34.3163)
(15, 11.267)
(23, 4.5922)
(31, 3.4442)
(47, 2.8946)
(63, 2.7706)
(95, 2.7103)
(127, 2.7233)
(191, 2.7305)
(255, 2.6895)
(383, 2.7328)
(511, 2.6941)
(767, 2.7261)
(1023, 2.7342)
(1535, 2.7096)
(2047, 2.7251)
}

\newcommand{\genSdePlotNfeCifaruDdpmppLambda}{%
(11, 35.213)
(15, 12.4434)
(23, 5.3812)
(31, 3.9916)
(47, 3.1366)
(63, 2.9038)
(95, 2.7276)
(127, 2.6063)
(191, 2.6029)
(255, 2.587)
(383, 2.562)
(511, 2.5598)
(767, 2.5573)
(1023, 2.5428)
(1535, 2.5413)
(2047, 2.5568)
}

\newcommand{\genSdePlotNfeCifaruDdpmppRange}{%
(11, 29.636)
(15, 10.0158)
(23, 4.5162)
(31, 3.4312)
(47, 2.8718)
(63, 2.718)
(95, 2.5805)
(127, 2.5286)
(191, 2.5655)
(255, 2.5192)
(383, 2.5961)
(511, 2.5486)
(767, 2.5857)
(1023, 2.5758)
(1535, 2.5679)
(2047, 2.5844)
}

\newcommand{\genSdePlotNfeCifaruDdpmppOurs}{%
(11, 29.8952)
(15, 10.2092)
(23, 4.8545)
(31, 3.7743)
(47, 3.0833)
(63, 2.8412)
(95, 2.6328)
(127, 2.4749)
(191, 2.3995)
(255, 2.3741)
(383, 2.3117)
(511, 2.2726)
(767, 2.3005)
(1023, 2.2932)
(1535, 2.3028)
(2047, 2.3034)
}

\newcommand{\genSdePlotNfeCifaruDdpmppRepro}{%
(48, 64.6301)
(64, 45.9351)
(96, 28.8456)
(128, 20.6623)
(192, 12.7601)
(256, 8.5455)
(384, 4.6581)
(512, 3.2031)
(768, 2.5832)
(1024, 2.5476)
(1536, 2.5624)
(2048, 2.6223)
}

\newcommand{\genSdePlotNfeCifaruDdpmppGotta}{%
(49, 72.29)
(147, 2.95)
(179, 2.59)
(274, 2.74)
(329, 2.7)
}

\newcommand{\genSdePlotNfeCifaruNcsnppOde}{%
(9, 38.6784)
(11, 13.4623)
(13, 7.7137)
(15, 5.6188)
(19, 4.3772)
(23, 4.0435)
(27, 3.8237)
(31, 3.8236)
(35, 3.7742)
(39, 3.7945)
(47, 3.7868)
(55, 3.7343)
(63, 3.8021)
(79, 3.7876)
(95, 3.7879)
(127, 3.8128)
(191, 3.7998)
(255, 3.8232)
(383, 3.7629)
(511, 3.8202)
(767, 3.7767)
(1023, 3.76)
(1535, 3.82)
(2047, 3.7658)
(3071, 3.7654)
(4095, 3.838)
(6143, 3.8228)
(8191, 3.8304)
}

\newcommand{\genSdePlotNfeCifaruNcsnppPlain}{%
(11, 30.3815)
(15, 8.8985)
(23, 4.6457)
(31, 3.9546)
(47, 3.5684)
(63, 3.4131)
(95, 3.2524)
(127, 3.1952)
(191, 3.1232)
(255, 3.0728)
(383, 3.0335)
(511, 3.0205)
(767, 2.9749)
(1023, 3.0412)
(1535, 3.0107)
(2047, 3.0221)
}

\newcommand{\genSdePlotNfeCifaruNcsnppLambda}{%
(11, 32.7695)
(15, 10.2872)
(23, 5.304)
(31, 4.3205)
(47, 3.6864)
(63, 3.4235)
(95, 3.1305)
(127, 2.9636)
(191, 2.783)
(255, 2.6659)
(383, 2.6073)
(511, 2.5666)
(767, 2.58)
(1023, 2.5301)
(1535, 2.5492)
(2047, 2.5121)
}

\newcommand{\genSdePlotNfeCifaruNcsnppRange}{%
(11, 23.7817)
(15, 8.4649)
(23, 4.4522)
(31, 3.7596)
(47, 3.4882)
(63, 3.2788)
(95, 3.1165)
(127, 2.9937)
(191, 2.8386)
(255, 2.9053)
(383, 2.8801)
(511, 2.8592)
(767, 2.9012)
(1023, 2.894)
(1535, 2.8662)
(2047, 2.927)
}

\newcommand{\genSdePlotNfeCifaruNcsnppOurs}{%
(11, 23.943)
(15, 8.9425)
(23, 4.7304)
(31, 3.9484)
(47, 3.5945)
(63, 3.3567)
(95, 3.0572)
(127, 2.8967)
(191, 2.6481)
(255, 2.4722)
(383, 2.3381)
(511, 2.3012)
(767, 2.2721)
(1023, 2.2598)
(1535, 2.2543)
(2047, 2.2268)
}

\newcommand{\genSdePlotNfeCifaruNcsnppRepro}{%
(192, 35.2965)
(256, 11.5479)
(384, 4.7541)
(512, 3.4214)
(768, 2.6818)
(1024, 2.4767)
(1536, 2.4639)
(2048, 2.4584)
}

\newcommand{\genSdePlotNfeCifaruNcsnppGotta}{%
(170, 8.85)
(271, 3.23)
(490, 2.87)
(738, 2.91)
}

\newcommand{\genSdePlotNfeImgcDhariwalOde}{%
(7, 87.1633)
(9, 33.6137)
(11, 12.3924)
(13, 6.6292)
(15, 4.7799)
(19, 3.5569)
(23, 3.1176)
(27, 2.9263)
(31, 2.8375)
(35, 2.8157)
(39, 2.7784)
(47, 2.7296)
(55, 2.7168)
(63, 2.7305)
(79, 2.6611)
(95, 2.6714)
(127, 2.6489)
(191, 2.6944)
(255, 2.6363)
(383, 2.6425)
(511, 2.7042)
(767, 2.6387)
(1023, 2.6936)
(1535, 2.6846)
(1983, 2.6495)
}

\newcommand{\genSdePlotNfeImgcDhariwalPlain}{%
(11, 24.8074)
(15, 8.0839)
(23, 3.9706)
(31, 3.1185)
(47, 2.594)
(63, 2.4342)
(95, 2.2404)
(127, 2.1007)
(191, 2.0078)
(255, 2.021)
(383, 1.8853)
(511, 1.873)
(767, 1.9183)
(1023, 1.8922)
(1535, 1.878)
(2047, 1.857)
}

\newcommand{\genSdePlotNfeImgcDhariwalLambda}{%
(11, 25.5929)
(15, 8.9137)
(23, 4.296)
(31, 3.3093)
(47, 2.7224)
(63, 2.5358)
(95, 2.2563)
(127, 2.0952)
(191, 1.9364)
(255, 1.87)
(383, 1.7006)
(511, 1.7412)
(767, 1.6421)
(1023, 1.6329)
(1535, 1.6536)
(2047, 1.6547)
}

\newcommand{\genSdePlotNfeImgcDhariwalRange}{%
(11, 25.1298)
(15, 8.3482)
(23, 3.8893)
(31, 3.1269)
(47, 2.6321)
(63, 2.3461)
(95, 2.1514)
(127, 2.047)
(191, 1.9158)
(255, 1.8837)
(383, 1.8457)
(511, 1.8398)
(767, 1.8356)
(1023, 1.8434)
(1535, 1.8466)
(2047, 1.861)
}

\newcommand{\genSdePlotNfeImgcDhariwalOurs}{%
(11, 25.8699)
(15, 8.9358)
(23, 4.3017)
(31, 3.4041)
(47, 2.7234)
(63, 2.4435)
(95, 2.2166)
(127, 1.9937)
(191, 1.812)
(255, 1.7135)
(383, 1.6508)
(511, 1.548)
(767, 1.5712)
(1023, 1.5462)
(1535, 1.5587)
(2047, 1.5724)
}

\newcommand{\genSdePlotNfeImgcDhariwalRepro}{%
(8, 34.8578)
(12, 22.1581)
(16, 14.9458)
(24, 8.4171)
(32, 5.9203)
(48, 3.969)
(64, 3.2057)
(96, 2.5822)
(128, 2.3708)
(192, 2.2135)
(256, 2.0846)
(384, 2.0547)
(512, 2.0111)
(768, 2.0305)
(1000, 2.0582)
}

\newcommand{\genSdePlotNfeCifaruDdpmppOdeMarks}{%
(511, 2.9325)
}

\newcommand{\genSdePlotNfeCifaruDdpmppPlainMarks}{%
(255, 2.6895)
}

\newcommand{\genSdePlotNfeCifaruDdpmppLambdaMarks}{%
(1535, 2.5413)
}

\newcommand{\genSdePlotNfeCifaruDdpmppRangeMarks}{%
(255, 2.5192)
}

\newcommand{\genSdePlotNfeCifaruDdpmppOursMarks}{%
(511, 2.2726)
}

\newcommand{\genSdePlotNfeCifaruDdpmppReproMarks}{%
(1024, 2.5476)
}

\newcommand{\genSdePlotNfeCifaruDdpmppGottaMarks}{%
(179, 2.59)
}

\newcommand{\genSdePlotNfeCifaruNcsnppOdeMarks}{%
(55, 3.7343)
}

\newcommand{\genSdePlotNfeCifaruNcsnppPlainMarks}{%
(767, 2.9749)
}

\newcommand{\genSdePlotNfeCifaruNcsnppLambdaMarks}{%
(2047, 2.5121)
}

\newcommand{\genSdePlotNfeCifaruNcsnppRangeMarks}{%
(191, 2.8386)
}

\newcommand{\genSdePlotNfeCifaruNcsnppOursMarks}{%
(2047, 2.2268)
}

\newcommand{\genSdePlotNfeCifaruNcsnppReproMarks}{%
(2048, 2.4584)
}

\newcommand{\genSdePlotNfeCifaruNcsnppGottaMarks}{%
(490, 2.87)
}

\newcommand{\genSdePlotNfeImgcDhariwalOdeMarks}{%
(255, 2.6363)
}

\newcommand{\genSdePlotNfeImgcDhariwalPlainMarks}{%
(2047, 1.857)
}

\newcommand{\genSdePlotNfeImgcDhariwalLambdaMarks}{%
(1023, 1.6329)
}

\newcommand{\genSdePlotNfeImgcDhariwalRangeMarks}{%
(767, 1.8356)
}

\newcommand{\genSdePlotNfeImgcDhariwalOursMarks}{%
(1023, 1.5462)
}

\newcommand{\genSdePlotNfeImgcDhariwalReproMarks}{%
(512, 2.0111)
}

\newcommand{\genSdePlotNfeCifaruDdpmppOdeLo}{%
(9, 45.9258)
(11, 18.4101)
(13, 9.6882)
(15, 6.5231)
(19, 4.1687)
(23, 3.519)
(27, 3.1906)
(31, 3.1026)
(35, 3.0139)
(39, 3.0129)
(47, 2.9913)
(55, 2.9415)
(63, 2.946)
(79, 2.959)
(95, 2.9596)
(127, 2.9854)
(191, 3.0103)
(255, 2.9333)
(383, 2.9701)
(511, 2.9325)
(767, 2.985)
(1023, 2.9644)
(1535, 2.9772)
(2047, 2.9684)
}

\newcommand{\genSdePlotNfeCifaruDdpmppOdeHi}{%
(9, 46.1394)
(11, 18.6101)
(13, 9.6962)
(15, 6.5782)
(19, 4.2034)
(23, 3.5901)
(27, 3.2397)
(31, 3.1855)
(35, 3.0608)
(39, 3.0479)
(47, 3.067)
(55, 2.9746)
(63, 2.9979)
(79, 2.9953)
(95, 3.0074)
(127, 3.0013)
(191, 3.0245)
(255, 2.9871)
(383, 3.0016)
(511, 2.9889)
(767, 3.0067)
(1023, 3.0278)
(1535, 2.9982)
(2047, 3.0386)
}

\newcommand{\genSdePlotNfeCifaruDdpmppPlainLo}{%
(11, 34.3163)
(15, 11.267)
(23, 4.5922)
(31, 3.4442)
(47, 2.8946)
(63, 2.7706)
(95, 2.7103)
(127, 2.7233)
(191, 2.7305)
(255, 2.6895)
(383, 2.7328)
(511, 2.6941)
(767, 2.7261)
(1023, 2.7342)
(1535, 2.7096)
(2047, 2.7251)
}

\newcommand{\genSdePlotNfeCifaruDdpmppPlainHi}{%
(11, 34.3942)
(15, 11.357)
(23, 4.6719)
(31, 3.4697)
(47, 2.9392)
(63, 2.8586)
(95, 2.7352)
(127, 2.7594)
(191, 2.7736)
(255, 2.8015)
(383, 2.783)
(511, 2.7561)
(767, 2.7858)
(1023, 2.7905)
(1535, 2.7447)
(2047, 2.7937)
}

\newcommand{\genSdePlotNfeCifaruDdpmppLambdaLo}{%
(11, 35.213)
(15, 12.4434)
(23, 5.3812)
(31, 3.9916)
(47, 3.1366)
(63, 2.9038)
(95, 2.7276)
(127, 2.6063)
(191, 2.6029)
(255, 2.587)
(383, 2.562)
(511, 2.5598)
(767, 2.5573)
(1023, 2.5428)
(1535, 2.5413)
(2047, 2.5568)
}

\newcommand{\genSdePlotNfeCifaruDdpmppLambdaHi}{%
(11, 35.4265)
(15, 12.7321)
(23, 5.5091)
(31, 4.0365)
(47, 3.2144)
(63, 2.9318)
(95, 2.7427)
(127, 2.68)
(191, 2.6421)
(255, 2.6288)
(383, 2.5953)
(511, 2.5705)
(767, 2.6107)
(1023, 2.5984)
(1535, 2.674)
(2047, 2.647)
}

\newcommand{\genSdePlotNfeCifaruDdpmppRangeLo}{%
(11, 29.636)
(15, 10.0158)
(23, 4.5162)
(31, 3.4312)
(47, 2.8718)
(63, 2.718)
(95, 2.5805)
(127, 2.5286)
(191, 2.5655)
(255, 2.5192)
(383, 2.5961)
(511, 2.5486)
(767, 2.5857)
(1023, 2.5758)
(1535, 2.5679)
(2047, 2.5844)
}

\newcommand{\genSdePlotNfeCifaruDdpmppRangeHi}{%
(11, 29.7279)
(15, 10.0674)
(23, 4.5472)
(31, 3.4816)
(47, 2.9125)
(63, 2.7686)
(95, 2.6225)
(127, 2.5409)
(191, 2.5767)
(255, 2.5789)
(383, 2.6214)
(511, 2.5992)
(767, 2.6051)
(1023, 2.6178)
(1535, 2.584)
(2047, 2.6298)
}

\newcommand{\genSdePlotNfeCifaruDdpmppOursLo}{%
(11, 29.8952)
(15, 10.2092)
(23, 4.8545)
(31, 3.7743)
(47, 3.0833)
(63, 2.8412)
(95, 2.6328)
(127, 2.4749)
(191, 2.3995)
(255, 2.3741)
(383, 2.3117)
(511, 2.2726)
(767, 2.3005)
(1023, 2.2932)
(1535, 2.3028)
(2047, 2.3034)
}

\newcommand{\genSdePlotNfeCifaruDdpmppOursHi}{%
(11, 30.141)
(15, 10.3523)
(23, 4.9885)
(31, 3.8925)
(47, 3.1472)
(63, 2.8859)
(95, 2.725)
(127, 2.5628)
(191, 2.4538)
(255, 2.4269)
(383, 2.3756)
(511, 2.3612)
(767, 2.3539)
(1023, 2.3654)
(1535, 2.3575)
(2047, 2.3276)
}

\newcommand{\genSdePlotNfeCifaruDdpmppReproLo}{%
(48, 64.6301)
(64, 45.9351)
(96, 28.8456)
(128, 20.6623)
(192, 12.7601)
(256, 8.5455)
(384, 4.6581)
(512, 3.2031)
(768, 2.5832)
(1024, 2.5476)
(1536, 2.5624)
(2048, 2.6223)
}

\newcommand{\genSdePlotNfeCifaruDdpmppReproHi}{%
(48, 65.3)
(64, 46.1548)
(96, 29.1209)
(128, 20.9689)
(192, 12.8105)
(256, 8.725)
(384, 4.6827)
(512, 3.2234)
(768, 2.6179)
(1024, 2.6023)
(1536, 2.6)
(2048, 2.6877)
}

\newcommand{\genSdePlotNfeCifaruNcsnppOdeLo}{%
(9, 38.6784)
(11, 13.4623)
(13, 7.7137)
(15, 5.6188)
(19, 4.3772)
(23, 4.0435)
(27, 3.8237)
(31, 3.8236)
(35, 3.7742)
(39, 3.7945)
(47, 3.7868)
(55, 3.7343)
(63, 3.8021)
(79, 3.7876)
(95, 3.7879)
(127, 3.8128)
(191, 3.7998)
(255, 3.8232)
(383, 3.7629)
(511, 3.8202)
(767, 3.7767)
(1023, 3.76)
(1535, 3.82)
(2047, 3.7658)
(3071, 3.7654)
(4095, 3.838)
(6143, 3.8228)
(8191, 3.8304)
}

\newcommand{\genSdePlotNfeCifaruNcsnppOdeHi}{%
(9, 39.1242)
(11, 13.7924)
(13, 7.9249)
(15, 5.6968)
(19, 4.5408)
(23, 4.1931)
(27, 3.9219)
(31, 3.8657)
(35, 3.8513)
(39, 3.9182)
(47, 3.9113)
(55, 3.8719)
(63, 3.8922)
(79, 3.89)
(95, 3.8571)
(127, 3.8558)
(191, 3.8568)
(255, 3.9084)
(383, 3.8351)
(511, 3.869)
(767, 3.907)
(1023, 3.841)
(1535, 3.8756)
(2047, 3.8592)
(3071, 3.8381)
(4095, 3.8622)
(6143, 3.945)
(8191, 3.8877)
}

\newcommand{\genSdePlotNfeCifaruNcsnppPlainLo}{%
(11, 30.3815)
(15, 8.8985)
(23, 4.6457)
(31, 3.9546)
(47, 3.5684)
(63, 3.4131)
(95, 3.2524)
(127, 3.1952)
(191, 3.1232)
(255, 3.0728)
(383, 3.0335)
(511, 3.0205)
(767, 2.9749)
(1023, 3.0412)
(1535, 3.0107)
(2047, 3.0221)
}

\newcommand{\genSdePlotNfeCifaruNcsnppPlainHi}{%
(11, 30.5468)
(15, 8.9745)
(23, 4.7496)
(31, 3.9939)
(47, 3.6297)
(63, 3.5118)
(95, 3.3302)
(127, 3.2247)
(191, 3.2226)
(255, 3.1263)
(383, 3.1003)
(511, 3.1472)
(767, 3.0674)
(1023, 3.0774)
(1535, 3.0264)
(2047, 3.0611)
}

\newcommand{\genSdePlotNfeCifaruNcsnppLambdaLo}{%
(11, 32.7695)
(15, 10.2872)
(23, 5.304)
(31, 4.3205)
(47, 3.6864)
(63, 3.4235)
(95, 3.1305)
(127, 2.9636)
(191, 2.783)
(255, 2.6659)
(383, 2.6073)
(511, 2.5666)
(767, 2.58)
(1023, 2.5301)
(1535, 2.5492)
(2047, 2.5121)
}

\newcommand{\genSdePlotNfeCifaruNcsnppLambdaHi}{%
(11, 33.2863)
(15, 10.3841)
(23, 5.4555)
(31, 4.4028)
(47, 3.8728)
(63, 3.4832)
(95, 3.2345)
(127, 3.0517)
(191, 2.8217)
(255, 2.7224)
(383, 2.6507)
(511, 2.623)
(767, 2.5985)
(1023, 2.5899)
(1535, 2.5927)
(2047, 2.5616)
}

\newcommand{\genSdePlotNfeCifaruNcsnppRangeLo}{%
(11, 23.7817)
(15, 8.4649)
(23, 4.4522)
(31, 3.7596)
(47, 3.4882)
(63, 3.2788)
(95, 3.1165)
(127, 2.9937)
(191, 2.8386)
(255, 2.9053)
(383, 2.8801)
(511, 2.8592)
(767, 2.9012)
(1023, 2.894)
(1535, 2.8662)
(2047, 2.927)
}

\newcommand{\genSdePlotNfeCifaruNcsnppRangeHi}{%
(11, 24.0347)
(15, 8.6031)
(23, 4.4876)
(31, 3.8578)
(47, 3.514)
(63, 3.3528)
(95, 3.1465)
(127, 3.0181)
(191, 2.9139)
(255, 2.9212)
(383, 2.9624)
(511, 2.9598)
(767, 2.9507)
(1023, 2.9761)
(1535, 2.9389)
(2047, 2.9505)
}

\newcommand{\genSdePlotNfeCifaruNcsnppOursLo}{%
(11, 23.943)
(15, 8.9425)
(23, 4.7304)
(31, 3.9484)
(47, 3.5945)
(63, 3.3567)
(95, 3.0572)
(127, 2.8967)
(191, 2.6481)
(255, 2.4722)
(383, 2.3381)
(511, 2.3012)
(767, 2.2721)
(1023, 2.2598)
(1535, 2.2543)
(2047, 2.2268)
}

\newcommand{\genSdePlotNfeCifaruNcsnppOursHi}{%
(11, 24.1449)
(15, 9.0319)
(23, 4.8576)
(31, 4.0423)
(47, 3.6727)
(63, 3.4363)
(95, 3.1703)
(127, 2.9665)
(191, 2.7689)
(255, 2.5682)
(383, 2.4002)
(511, 2.3887)
(767, 2.3407)
(1023, 2.2954)
(1535, 2.3218)
(2047, 2.299)
}

\newcommand{\genSdePlotNfeCifaruNcsnppReproLo}{%
(192, 35.2965)
(256, 11.5479)
(384, 4.7541)
(512, 3.4214)
(768, 2.6818)
(1024, 2.4767)
(1536, 2.4639)
(2048, 2.4584)
}

\newcommand{\genSdePlotNfeCifaruNcsnppReproHi}{%
(192, 35.6545)
(256, 11.634)
(384, 4.8486)
(512, 3.4541)
(768, 2.7131)
(1024, 2.5085)
(1536, 2.481)
(2048, 2.4947)
}

\newcommand{\genSdePlotNfeImgcDhariwalOdeLo}{%
(7, 87.1633)
(9, 33.6137)
(11, 12.3924)
(13, 6.6292)
(15, 4.7799)
(19, 3.5569)
(23, 3.1176)
(27, 2.9263)
(31, 2.8375)
(35, 2.8157)
(39, 2.7784)
(47, 2.7296)
(55, 2.7168)
(63, 2.7305)
(79, 2.6611)
(95, 2.6714)
(127, 2.6489)
(191, 2.6944)
(255, 2.6363)
(383, 2.6425)
(511, 2.7042)
(767, 2.6387)
(1023, 2.6936)
(1535, 2.6846)
(1983, 2.6495)
}

\newcommand{\genSdePlotNfeImgcDhariwalOdeHi}{%
(7, 87.5603)
(9, 33.9041)
(11, 12.6988)
(13, 6.826)
(15, 4.8841)
(19, 3.6318)
(23, 3.2247)
(27, 3.0712)
(31, 2.9244)
(35, 2.8939)
(39, 2.8488)
(47, 2.9231)
(55, 2.8145)
(63, 2.8041)
(79, 2.7861)
(95, 2.7469)
(127, 2.768)
(191, 2.7719)
(255, 2.715)
(383, 2.7099)
(511, 2.7391)
(767, 2.7408)
(1023, 2.8023)
(1535, 2.7319)
(1983, 2.7487)
}

\newcommand{\genSdePlotNfeImgcDhariwalPlainLo}{%
(11, 24.8074)
(15, 8.0839)
(23, 3.9706)
(31, 3.1185)
(47, 2.594)
(63, 2.4342)
(95, 2.2404)
(127, 2.1007)
(191, 2.0078)
(255, 2.021)
(383, 1.8853)
(511, 1.873)
(767, 1.9183)
(1023, 1.8922)
(1535, 1.878)
(2047, 1.857)
}

\newcommand{\genSdePlotNfeImgcDhariwalPlainHi}{%
(11, 25.1521)
(15, 8.2637)
(23, 4.0157)
(31, 3.1763)
(47, 2.678)
(63, 2.5219)
(95, 2.2833)
(127, 2.155)
(191, 2.1076)
(255, 2.0753)
(383, 1.9149)
(511, 1.934)
(767, 1.9376)
(1023, 1.9454)
(1535, 1.8992)
(2047, 1.9112)
}

\newcommand{\genSdePlotNfeImgcDhariwalLambdaLo}{%
(11, 25.5929)
(15, 8.9137)
(23, 4.296)
(31, 3.3093)
(47, 2.7224)
(63, 2.5358)
(95, 2.2563)
(127, 2.0952)
(191, 1.9364)
(255, 1.87)
(383, 1.7006)
(511, 1.7412)
(767, 1.6421)
(1023, 1.6329)
(1535, 1.6536)
(2047, 1.6547)
}

\newcommand{\genSdePlotNfeImgcDhariwalLambdaHi}{%
(11, 25.7204)
(15, 8.9492)
(23, 4.3764)
(31, 3.431)
(47, 2.9023)
(63, 2.607)
(95, 2.3062)
(127, 2.1488)
(191, 1.9696)
(255, 1.8868)
(383, 1.7646)
(511, 1.7706)
(767, 1.6983)
(1023, 1.6653)
(1535, 1.6882)
(2047, 1.7022)
}

\newcommand{\genSdePlotNfeImgcDhariwalRangeLo}{%
(11, 25.1298)
(15, 8.3482)
(23, 3.8893)
(31, 3.1269)
(47, 2.6321)
(63, 2.3461)
(95, 2.1514)
(127, 2.047)
(191, 1.9158)
(255, 1.8837)
(383, 1.8457)
(511, 1.8398)
(767, 1.8356)
(1023, 1.8434)
(1535, 1.8466)
(2047, 1.861)
}

\newcommand{\genSdePlotNfeImgcDhariwalRangeHi}{%
(11, 25.2363)
(15, 8.3919)
(23, 4.0581)
(31, 3.1759)
(47, 2.6592)
(63, 2.415)
(95, 2.2362)
(127, 2.0672)
(191, 1.961)
(255, 1.9379)
(383, 1.8914)
(511, 1.8791)
(767, 1.8414)
(1023, 1.8832)
(1535, 1.8515)
(2047, 1.8957)
}

\newcommand{\genSdePlotNfeImgcDhariwalOursLo}{%
(11, 25.8699)
(15, 8.9358)
(23, 4.3017)
(31, 3.4041)
(47, 2.7234)
(63, 2.4435)
(95, 2.2166)
(127, 1.9937)
(191, 1.812)
(255, 1.7135)
(383, 1.6508)
(511, 1.548)
(767, 1.5712)
(1023, 1.5462)
(1535, 1.5587)
(2047, 1.5724)
}

\newcommand{\genSdePlotNfeImgcDhariwalOursHi}{%
(11, 26.1821)
(15, 9.1335)
(23, 4.4424)
(31, 3.4586)
(47, 2.8592)
(63, 2.5609)
(95, 2.2905)
(127, 2.1065)
(191, 1.9119)
(255, 1.7729)
(383, 1.6989)
(511, 1.6532)
(767, 1.6601)
(1023, 1.6095)
(1535, 1.5962)
(2047, 1.6334)
}

\newcommand{\genSdePlotNfeImgcDhariwalReproLo}{%
(8, 34.8578)
(12, 22.1581)
(16, 14.9458)
(24, 8.4171)
(32, 5.9203)
(48, 3.969)
(64, 3.2057)
(96, 2.5822)
(128, 2.3708)
(192, 2.2135)
(256, 2.0846)
(384, 2.0547)
(512, 2.0111)
(768, 2.0305)
(1000, 2.0582)
}

\newcommand{\genSdePlotNfeImgcDhariwalReproHi}{%
(8, 35.0521)
(12, 22.2618)
(16, 15.0954)
(24, 8.4356)
(32, 6.039)
(48, 4.0164)
(64, 3.2336)
(96, 2.6267)
(128, 2.3882)
(192, 2.2484)
(256, 2.1525)
(384, 2.0905)
(512, 2.0803)
(768, 2.0971)
(1000, 2.0836)
}

\newcommand{\genTrainingTable}{%
\multicolumn{1}{l|}{}                                                                    & \multicolumn{4}{c|}{CIFAR-10~\cite{Krizhevsky2009cifar} at 32$\times$32}  & \multicolumn{2}{c|}{FFHQ~\cite{Karras2018stylegan} 64$\times$64}  & \multicolumn{2}{c|}{AFHQv2~\cite{Choi2020afhq} 64$\times$64}  \\
\multicolumn{1}{l|}{}                                                                    & \multicolumn{2}{c|}{Conditional}  & \multicolumn{2}{c|}{Unconditional}    & \multicolumn{2}{c|}{Unconditional}                                & \multicolumn{2}{c|}{Unconditional}                            \\
\tabucline{-}
{\bf Training configuration}                                                             & VP         & VE                   & VP          & VE                      & VP         & VE                                                   & VP         & VE                                               \\
\tabucline{-}
\confhdr{a} \low{Baseline~\cite{Song2021sde}}\hfill\low{\scriptsize{($^*$pre-trained)}}  & 2.48       & 3.11                 & 3.01\smast  & 3.77\smast              & 3.39       & 25.95                                                & 2.58       & 18.52                                            \\
\confhdr{b} \low{+ Adjust hyperparameters}                                               & 2.18       & 2.48                 & 2.51        & 2.94                    & 3.13       & 22.53                                                & 2.43       & 23.12                                            \\
\confhdr{c} \low{+ Redistribute capacity}                                                & 2.08       & 2.52                 & 2.31        & 2.83                    & 2.78       & 41.62                                                & 2.54       & 15.04                                            \\
\confhdr{d} \low{+ Our preconditioning}                                                  & 2.09       & 2.64                 & 2.29        & 3.10                    & 2.94       & \s3.39                                               & 2.79       & \s3.81                                           \\
\confhdr{e} \low{+ Our loss function}                                                    & 1.88       & 1.86                 & 2.05        & 1.99                    & 2.60       & \s2.81                                               & 2.29       & \s2.28                                           \\
\confhdr{f} \low{+ Non-leaky augmentation}                                               & {\bf1.79}  & {\bf1.79}            & {\bf1.97}   & {\bf1.98}               & {\bf2.39}  & {\bf\s2.53}                                          & {\bf1.96}  & {\bf\s2.16}                                      \\
\tabucline{-}
NFE                                                                                      & 35         & 35                   & 35          & 35                      & 79         & 79                                                   & 79         & 79                                               \\
}

\newcommand{\genTrainingPlotLossCifaruDdpmppPrec}{%
(0.001, 0.935887)
(0.00104618, 0.932065)
(0.0010945, 0.925746)
(0.00114505, 0.920826)
(0.00119793, 0.91716)
(0.00125325, 0.91021)
(0.00131113, 0.904185)
(0.00137169, 0.897268)
(0.00143504, 0.89002)
(0.00150131, 0.88419)
(0.00157065, 0.876304)
(0.00164318, 0.870736)
(0.00171907, 0.864375)
(0.00179846, 0.855685)
(0.00188152, 0.845838)
(0.00196842, 0.836854)
(0.00205933, 0.829903)
(0.00215443, 0.819906)
(0.00225393, 0.812892)
(0.00235803, 0.801377)
(0.00246693, 0.792468)
(0.00258086, 0.785023)
(0.00270005, 0.776177)
(0.00282475, 0.764449)
(0.00295521, 0.755364)
(0.00309169, 0.746797)
(0.00323448, 0.738941)
(0.00338386, 0.727761)
(0.00354013, 0.717063)
(0.00370363, 0.70708)
(0.00387468, 0.69822)
(0.00405362, 0.68778)
(0.00424083, 0.68041)
(0.00443669, 0.669461)
(0.00464159, 0.659355)
(0.00485595, 0.648078)
(0.00508022, 0.641452)
(0.00531484, 0.633582)
(0.0055603, 0.624919)
(0.00581709, 0.614211)
(0.00608574, 0.60869)
(0.00636681, 0.597473)
(0.00666085, 0.589852)
(0.00696847, 0.582495)
(0.00729029, 0.573531)
(0.00762699, 0.566812)
(0.00797923, 0.558219)
(0.00834773, 0.550781)
(0.00873326, 0.546285)
(0.00913659, 0.537274)
(0.00955855, 0.530516)
(0.01, 0.523281)
(0.0104618, 0.515301)
(0.010945, 0.506895)
(0.0114505, 0.501477)
(0.0119793, 0.494155)
(0.0125325, 0.485231)
(0.0131113, 0.478172)
(0.0137169, 0.473135)
(0.0143504, 0.465707)
(0.0150131, 0.458195)
(0.0157065, 0.451803)
(0.0164318, 0.445469)
(0.0171907, 0.437426)
(0.0179846, 0.430112)
(0.0188152, 0.422169)
(0.0196842, 0.41911)
(0.0205933, 0.409621)
(0.0215443, 0.403375)
(0.0225393, 0.395353)
(0.0235803, 0.390269)
(0.0246693, 0.381483)
(0.0258086, 0.373198)
(0.0270005, 0.366038)
(0.0282475, 0.358493)
(0.0295521, 0.354065)
(0.0309169, 0.344192)
(0.0323448, 0.338743)
(0.0338386, 0.332262)
(0.0354013, 0.325112)
(0.0370363, 0.318813)
(0.0387468, 0.309584)
(0.0405362, 0.30595)
(0.0424083, 0.296519)
(0.0443669, 0.290659)
(0.0464159, 0.284524)
(0.0485595, 0.276492)
(0.0508022, 0.269897)
(0.0531484, 0.26307)
(0.055603, 0.256682)
(0.0581709, 0.251942)
(0.0608574, 0.245454)
(0.0636681, 0.240354)
(0.0666085, 0.234278)
(0.0696847, 0.226034)
(0.0729029, 0.221832)
(0.0762699, 0.215726)
(0.0797923, 0.20959)
(0.0834773, 0.205326)
(0.0873326, 0.197943)
(0.0913659, 0.193842)
(0.0955855, 0.188872)
(0.1, 0.183719)
(0.104618, 0.17898)
(0.10945, 0.174022)
(0.114505, 0.169462)
(0.119793, 0.166319)
(0.125325, 0.160835)
(0.131113, 0.157256)
(0.137169, 0.152763)
(0.143504, 0.148323)
(0.150131, 0.144441)
(0.157065, 0.141149)
(0.164318, 0.136746)
(0.171907, 0.134268)
(0.179846, 0.131022)
(0.188152, 0.127626)
(0.196842, 0.124851)
(0.205933, 0.121814)
(0.215443, 0.11997)
(0.225393, 0.117522)
(0.235803, 0.11439)
(0.246693, 0.112883)
(0.258086, 0.110774)
(0.270005, 0.108501)
(0.282475, 0.107486)
(0.295521, 0.105305)
(0.309169, 0.10385)
(0.323448, 0.102992)
(0.338386, 0.101692)
(0.354013, 0.100656)
(0.370363, 0.09989)
(0.387468, 0.0988965)
(0.405362, 0.0984758)
(0.424083, 0.0991853)
(0.443669, 0.0977531)
(0.464159, 0.099352)
(0.485595, 0.0985395)
(0.508022, 0.0994221)
(0.531484, 0.100006)
(0.55603, 0.100567)
(0.581709, 0.101206)
(0.608574, 0.102515)
(0.636681, 0.104733)
(0.666085, 0.106243)
(0.696847, 0.107652)
(0.729029, 0.110097)
(0.762699, 0.112422)
(0.797923, 0.115168)
(0.834773, 0.117068)
(0.873326, 0.12029)
(0.913659, 0.125115)
(0.955855, 0.128211)
(1, 0.132666)
(1.04618, 0.135977)
(1.0945, 0.140414)
(1.14505, 0.145856)
(1.19793, 0.149765)
(1.25325, 0.156488)
(1.31113, 0.161838)
(1.37169, 0.168818)
(1.43504, 0.174431)
(1.50131, 0.180221)
(1.57065, 0.188314)
(1.64318, 0.194418)
(1.71907, 0.203214)
(1.79846, 0.211329)
(1.88152, 0.220457)
(1.96842, 0.229329)
(2.05933, 0.236927)
(2.15443, 0.247964)
(2.25393, 0.258296)
(2.35803, 0.267435)
(2.46693, 0.276526)
(2.58086, 0.287577)
(2.70005, 0.295673)
(2.82475, 0.307909)
(2.95521, 0.319063)
(3.09169, 0.330699)
(3.23448, 0.344923)
(3.38386, 0.352291)
(3.54013, 0.36743)
(3.70363, 0.375873)
(3.87468, 0.388822)
(4.05362, 0.402824)
(4.24083, 0.412)
(4.43669, 0.428658)
(4.64159, 0.440337)
(4.85595, 0.452347)
(5.08022, 0.470259)
(5.31484, 0.480306)
(5.5603, 0.493761)
(5.81709, 0.506865)
(6.08574, 0.521668)
(6.36681, 0.533678)
(6.66085, 0.546686)
(6.96847, 0.56425)
(7.29029, 0.575038)
(7.62699, 0.589285)
(7.97923, 0.602651)
(8.34773, 0.613603)
(8.73326, 0.630561)
(9.13659, 0.640862)
(9.55855, 0.65557)
(10, 0.667005)
(10.4618, 0.679101)
(10.945, 0.701901)
(11.4505, 0.710492)
(11.9793, 0.723285)
(12.5325, 0.732753)
(13.1113, 0.749752)
(13.7169, 0.75579)
(14.3504, 0.773313)
(15.0131, 0.783338)
(15.7065, 0.793835)
(16.4318, 0.806996)
(17.1907, 0.813807)
(17.9846, 0.831776)
(18.8152, 0.83726)
(19.6842, 0.844589)
(20.5933, 0.854221)
(21.5443, 0.871325)
(22.5393, 0.870565)
(23.5803, 0.882587)
(24.6693, 0.890907)
(25.8086, 0.906591)
(27.0005, 0.894146)
(28.2475, 0.913723)
(29.5521, 0.921919)
(30.9169, 0.916764)
(32.3448, 0.924548)
(33.8386, 0.925978)
(35.4013, 0.932524)
(37.0363, 0.939586)
(38.7468, 0.940625)
(40.5362, 0.942137)
(42.4083, 0.959107)
(44.3669, 0.960813)
(46.4159, 0.957803)
(48.5595, 0.96508)
(50.8022, 0.964065)
(53.1484, 0.974022)
(55.603, 0.970108)
(58.1709, 0.976704)
(60.8574, 0.972617)
(63.6681, 0.969775)
(66.6085, 0.977529)
(69.6847, 0.978901)
(72.9029, 0.984227)
(76.2699, 0.979895)
(79.7923, 0.97624)
(83.4773, 0.980807)
(87.3326, 0.986343)
(91.3659, 0.980545)
(95.5855, 0.989526)
(100, 0.985461)
}

\newcommand{\genTrainingPlotLossCifaruDdpmppPrecLo}{%
(0.001, 0.860963)
(0.00104618, 0.858543)
(0.0010945, 0.845707)
(0.00114505, 0.842838)
(0.00119793, 0.839565)
(0.00125325, 0.830039)
(0.00131113, 0.822195)
(0.00137169, 0.81103)
(0.00143504, 0.803812)
(0.00150131, 0.797616)
(0.00157065, 0.788406)
(0.00164318, 0.780272)
(0.00171907, 0.77525)
(0.00179846, 0.764186)
(0.00188152, 0.751319)
(0.00196842, 0.739264)
(0.00205933, 0.73112)
(0.00215443, 0.719639)
(0.00225393, 0.711752)
(0.00235803, 0.697164)
(0.00246693, 0.686754)
(0.00258086, 0.679911)
(0.00270005, 0.668307)
(0.00282475, 0.654908)
(0.00295521, 0.645515)
(0.00309169, 0.63591)
(0.00323448, 0.627157)
(0.00338386, 0.613594)
(0.00354013, 0.601776)
(0.00370363, 0.59131)
(0.00387468, 0.580265)
(0.00405362, 0.569804)
(0.00424083, 0.56079)
(0.00443669, 0.548067)
(0.00464159, 0.538567)
(0.00485595, 0.526046)
(0.00508022, 0.520119)
(0.00531484, 0.510717)
(0.0055603, 0.501106)
(0.00581709, 0.491683)
(0.00608574, 0.485805)
(0.00636681, 0.473887)
(0.00666085, 0.46655)
(0.00696847, 0.46142)
(0.00729029, 0.453199)
(0.00762699, 0.444672)
(0.00797923, 0.437072)
(0.00834773, 0.428805)
(0.00873326, 0.42575)
(0.00913659, 0.417643)
(0.00955855, 0.412234)
(0.01, 0.406314)
(0.0104618, 0.399166)
(0.010945, 0.390131)
(0.0114505, 0.387851)
(0.0119793, 0.380699)
(0.0125325, 0.37259)
(0.0131113, 0.367837)
(0.0137169, 0.361917)
(0.0143504, 0.355066)
(0.0150131, 0.349388)
(0.0157065, 0.343404)
(0.0164318, 0.338361)
(0.0171907, 0.331641)
(0.0179846, 0.326842)
(0.0188152, 0.318965)
(0.0196842, 0.316386)
(0.0205933, 0.308805)
(0.0215443, 0.303556)
(0.0225393, 0.29596)
(0.0235803, 0.292452)
(0.0246693, 0.284714)
(0.0258086, 0.277728)
(0.0270005, 0.271014)
(0.0282475, 0.264378)
(0.0295521, 0.261145)
(0.0309169, 0.25178)
(0.0323448, 0.249609)
(0.0338386, 0.242271)
(0.0354013, 0.238077)
(0.0370363, 0.233101)
(0.0387468, 0.224797)
(0.0405362, 0.221588)
(0.0424083, 0.214428)
(0.0443669, 0.20935)
(0.0464159, 0.203408)
(0.0485595, 0.197974)
(0.0508022, 0.191665)
(0.0531484, 0.186496)
(0.055603, 0.18191)
(0.0581709, 0.177778)
(0.0608574, 0.173777)
(0.0636681, 0.169394)
(0.0666085, 0.163985)
(0.0696847, 0.157481)
(0.0729029, 0.153877)
(0.0762699, 0.148367)
(0.0797923, 0.145147)
(0.0834773, 0.141566)
(0.0873326, 0.136572)
(0.0913659, 0.133264)
(0.0955855, 0.129062)
(0.1, 0.125553)
(0.104618, 0.121738)
(0.10945, 0.117191)
(0.114505, 0.114575)
(0.119793, 0.112611)
(0.125325, 0.107392)
(0.131113, 0.105456)
(0.137169, 0.102424)
(0.143504, 0.0988032)
(0.150131, 0.0960962)
(0.157065, 0.0938123)
(0.164318, 0.0912976)
(0.171907, 0.0896395)
(0.179846, 0.087196)
(0.188152, 0.083868)
(0.196842, 0.0832687)
(0.205933, 0.080391)
(0.215443, 0.0788032)
(0.225393, 0.0771939)
(0.235803, 0.0751865)
(0.246693, 0.0741128)
(0.258086, 0.0722784)
(0.270005, 0.0706153)
(0.282475, 0.0701218)
(0.295521, 0.0687147)
(0.309169, 0.0675181)
(0.323448, 0.0672656)
(0.338386, 0.0660777)
(0.354013, 0.0652172)
(0.370363, 0.0646541)
(0.387468, 0.0642519)
(0.405362, 0.0633545)
(0.424083, 0.0641563)
(0.443669, 0.0625805)
(0.464159, 0.0645788)
(0.485595, 0.0636543)
(0.508022, 0.0639448)
(0.531484, 0.0640935)
(0.55603, 0.0646568)
(0.581709, 0.0654116)
(0.608574, 0.0659652)
(0.636681, 0.0675971)
(0.666085, 0.0681069)
(0.696847, 0.0691611)
(0.729029, 0.0702033)
(0.762699, 0.0723479)
(0.797923, 0.0741149)
(0.834773, 0.0754795)
(0.873326, 0.0770533)
(0.913659, 0.0808099)
(0.955855, 0.0826683)
(1, 0.0859054)
(1.04618, 0.0880763)
(1.0945, 0.0903436)
(1.14505, 0.093258)
(1.19793, 0.0956562)
(1.25325, 0.100825)
(1.31113, 0.103999)
(1.37169, 0.107727)
(1.43504, 0.111342)
(1.50131, 0.115364)
(1.57065, 0.120393)
(1.64318, 0.124575)
(1.71907, 0.129791)
(1.79846, 0.134628)
(1.88152, 0.141135)
(1.96842, 0.146444)
(2.05933, 0.150266)
(2.15443, 0.156852)
(2.25393, 0.164006)
(2.35803, 0.169046)
(2.46693, 0.175235)
(2.58086, 0.182481)
(2.70005, 0.187031)
(2.82475, 0.194171)
(2.95521, 0.200297)
(3.09169, 0.207026)
(3.23448, 0.214104)
(3.38386, 0.221813)
(3.54013, 0.228838)
(3.70363, 0.232394)
(3.87468, 0.241053)
(4.05362, 0.250834)
(4.24083, 0.25382)
(4.43669, 0.262944)
(4.64159, 0.27065)
(4.85595, 0.27676)
(5.08022, 0.286096)
(5.31484, 0.2902)
(5.5603, 0.299134)
(5.81709, 0.306472)
(6.08574, 0.312826)
(6.36681, 0.317888)
(6.66085, 0.32321)
(6.96847, 0.335237)
(7.29029, 0.33756)
(7.62699, 0.345808)
(7.97923, 0.349048)
(8.34773, 0.351882)
(8.73326, 0.364111)
(9.13659, 0.367729)
(9.55855, 0.373058)
(10, 0.375722)
(10.4618, 0.384369)
(10.945, 0.3943)
(11.4505, 0.389751)
(11.9793, 0.402036)
(12.5325, 0.40694)
(13.1113, 0.410911)
(13.7169, 0.410829)
(14.3504, 0.417889)
(15.0131, 0.417461)
(15.7065, 0.421738)
(16.4318, 0.422018)
(17.1907, 0.42975)
(17.9846, 0.436207)
(18.8152, 0.427889)
(19.6842, 0.427389)
(20.5933, 0.437841)
(21.5443, 0.440476)
(22.5393, 0.445442)
(23.5803, 0.449171)
(24.6693, 0.445609)
(25.8086, 0.450731)
(27.0005, 0.444383)
(28.2475, 0.450109)
(29.5521, 0.465272)
(30.9169, 0.44408)
(32.3448, 0.454251)
(33.8386, 0.452606)
(35.4013, 0.452455)
(37.0363, 0.459272)
(38.7468, 0.456334)
(40.5362, 0.452688)
(42.4083, 0.465406)
(44.3669, 0.463029)
(46.4159, 0.463499)
(48.5595, 0.458084)
(50.8022, 0.464771)
(53.1484, 0.471699)
(55.603, 0.458364)
(58.1709, 0.464626)
(60.8574, 0.465796)
(63.6681, 0.463799)
(66.6085, 0.471296)
(69.6847, 0.463654)
(72.9029, 0.46799)
(76.2699, 0.458165)
(79.7923, 0.459793)
(83.4773, 0.462094)
(87.3326, 0.464545)
(91.3659, 0.468129)
(95.5855, 0.466365)
(100, 0.471701)
}

\newcommand{\genTrainingPlotLossCifaruDdpmppPrecHi}{%
(0.001, 1.01081)
(0.00104618, 1.00559)
(0.0010945, 1.00578)
(0.00114505, 0.998814)
(0.00119793, 0.994756)
(0.00125325, 0.990382)
(0.00131113, 0.986174)
(0.00137169, 0.983506)
(0.00143504, 0.976229)
(0.00150131, 0.970764)
(0.00157065, 0.964201)
(0.00164318, 0.9612)
(0.00171907, 0.9535)
(0.00179846, 0.947185)
(0.00188152, 0.940357)
(0.00196842, 0.934443)
(0.00205933, 0.928686)
(0.00215443, 0.920173)
(0.00225393, 0.914031)
(0.00235803, 0.905591)
(0.00246693, 0.898182)
(0.00258086, 0.890136)
(0.00270005, 0.884047)
(0.00282475, 0.87399)
(0.00295521, 0.865213)
(0.00309169, 0.857685)
(0.00323448, 0.850725)
(0.00338386, 0.841927)
(0.00354013, 0.832351)
(0.00370363, 0.82285)
(0.00387468, 0.816176)
(0.00405362, 0.805755)
(0.00424083, 0.800029)
(0.00443669, 0.790855)
(0.00464159, 0.780144)
(0.00485595, 0.77011)
(0.00508022, 0.762786)
(0.00531484, 0.756448)
(0.0055603, 0.748732)
(0.00581709, 0.736739)
(0.00608574, 0.731576)
(0.00636681, 0.721059)
(0.00666085, 0.713154)
(0.00696847, 0.70357)
(0.00729029, 0.693863)
(0.00762699, 0.688953)
(0.00797923, 0.679366)
(0.00834773, 0.672756)
(0.00873326, 0.666819)
(0.00913659, 0.656905)
(0.00955855, 0.648799)
(0.01, 0.640249)
(0.0104618, 0.631436)
(0.010945, 0.623659)
(0.0114505, 0.615103)
(0.0119793, 0.60761)
(0.0125325, 0.597872)
(0.0131113, 0.588507)
(0.0137169, 0.584353)
(0.0143504, 0.576348)
(0.0150131, 0.567003)
(0.0157065, 0.560201)
(0.0164318, 0.552577)
(0.0171907, 0.543211)
(0.0179846, 0.533381)
(0.0188152, 0.525373)
(0.0196842, 0.521834)
(0.0205933, 0.510437)
(0.0215443, 0.503193)
(0.0225393, 0.494745)
(0.0235803, 0.488085)
(0.0246693, 0.478253)
(0.0258086, 0.468668)
(0.0270005, 0.461062)
(0.0282475, 0.452607)
(0.0295521, 0.446984)
(0.0309169, 0.436604)
(0.0323448, 0.427877)
(0.0338386, 0.422253)
(0.0354013, 0.412147)
(0.0370363, 0.404525)
(0.0387468, 0.394371)
(0.0405362, 0.390311)
(0.0424083, 0.378609)
(0.0443669, 0.371969)
(0.0464159, 0.36564)
(0.0485595, 0.35501)
(0.0508022, 0.348128)
(0.0531484, 0.339644)
(0.055603, 0.331454)
(0.0581709, 0.326106)
(0.0608574, 0.317131)
(0.0636681, 0.311314)
(0.0666085, 0.30457)
(0.0696847, 0.294586)
(0.0729029, 0.289787)
(0.0762699, 0.283085)
(0.0797923, 0.274034)
(0.0834773, 0.269087)
(0.0873326, 0.259315)
(0.0913659, 0.254419)
(0.0955855, 0.248683)
(0.1, 0.241885)
(0.104618, 0.236223)
(0.10945, 0.230852)
(0.114505, 0.224348)
(0.119793, 0.220027)
(0.125325, 0.214277)
(0.131113, 0.209056)
(0.137169, 0.203103)
(0.143504, 0.197842)
(0.150131, 0.192785)
(0.157065, 0.188486)
(0.164318, 0.182195)
(0.171907, 0.178897)
(0.179846, 0.174848)
(0.188152, 0.171384)
(0.196842, 0.166434)
(0.205933, 0.163236)
(0.215443, 0.161137)
(0.225393, 0.157849)
(0.235803, 0.153594)
(0.246693, 0.151653)
(0.258086, 0.149271)
(0.270005, 0.146387)
(0.282475, 0.144849)
(0.295521, 0.141896)
(0.309169, 0.140182)
(0.323448, 0.138718)
(0.338386, 0.137305)
(0.354013, 0.136094)
(0.370363, 0.135126)
(0.387468, 0.133541)
(0.405362, 0.133597)
(0.424083, 0.134214)
(0.443669, 0.132926)
(0.464159, 0.134125)
(0.485595, 0.133425)
(0.508022, 0.134899)
(0.531484, 0.135918)
(0.55603, 0.136478)
(0.581709, 0.137)
(0.608574, 0.139065)
(0.636681, 0.141869)
(0.666085, 0.14438)
(0.696847, 0.146142)
(0.729029, 0.149991)
(0.762699, 0.152496)
(0.797923, 0.15622)
(0.834773, 0.158656)
(0.873326, 0.163527)
(0.913659, 0.16942)
(0.955855, 0.173753)
(1, 0.179427)
(1.04618, 0.183878)
(1.0945, 0.190485)
(1.14505, 0.198454)
(1.19793, 0.203874)
(1.25325, 0.212151)
(1.31113, 0.219677)
(1.37169, 0.229909)
(1.43504, 0.237521)
(1.50131, 0.245078)
(1.57065, 0.256235)
(1.64318, 0.264261)
(1.71907, 0.276637)
(1.79846, 0.288031)
(1.88152, 0.299779)
(1.96842, 0.312214)
(2.05933, 0.323588)
(2.15443, 0.339075)
(2.25393, 0.352587)
(2.35803, 0.365825)
(2.46693, 0.377816)
(2.58086, 0.392673)
(2.70005, 0.404316)
(2.82475, 0.421647)
(2.95521, 0.43783)
(3.09169, 0.454372)
(3.23448, 0.475743)
(3.38386, 0.482769)
(3.54013, 0.506021)
(3.70363, 0.519352)
(3.87468, 0.536592)
(4.05362, 0.554814)
(4.24083, 0.570181)
(4.43669, 0.594373)
(4.64159, 0.610025)
(4.85595, 0.627935)
(5.08022, 0.654422)
(5.31484, 0.670411)
(5.5603, 0.688389)
(5.81709, 0.707258)
(6.08574, 0.730509)
(6.36681, 0.749468)
(6.66085, 0.770162)
(6.96847, 0.793262)
(7.29029, 0.812516)
(7.62699, 0.832762)
(7.97923, 0.856253)
(8.34773, 0.875323)
(8.73326, 0.897012)
(9.13659, 0.913994)
(9.55855, 0.938082)
(10, 0.958289)
(10.4618, 0.973833)
(10.945, 1.0095)
(11.4505, 1.03123)
(11.9793, 1.04453)
(12.5325, 1.05857)
(13.1113, 1.08859)
(13.7169, 1.10075)
(14.3504, 1.12874)
(15.0131, 1.14922)
(15.7065, 1.16593)
(16.4318, 1.19197)
(17.1907, 1.19786)
(17.9846, 1.22735)
(18.8152, 1.24663)
(19.6842, 1.26179)
(20.5933, 1.2706)
(21.5443, 1.30217)
(22.5393, 1.29569)
(23.5803, 1.316)
(24.6693, 1.3362)
(25.8086, 1.36245)
(27.0005, 1.34391)
(28.2475, 1.37734)
(29.5521, 1.37857)
(30.9169, 1.38945)
(32.3448, 1.39484)
(33.8386, 1.39935)
(35.4013, 1.41259)
(37.0363, 1.4199)
(38.7468, 1.42492)
(40.5362, 1.43159)
(42.4083, 1.45281)
(44.3669, 1.4586)
(46.4159, 1.45211)
(48.5595, 1.47208)
(50.8022, 1.46336)
(53.1484, 1.47635)
(55.603, 1.48185)
(58.1709, 1.48878)
(60.8574, 1.47944)
(63.6681, 1.47575)
(66.6085, 1.48376)
(69.6847, 1.49415)
(72.9029, 1.50046)
(76.2699, 1.50162)
(79.7923, 1.49269)
(83.4773, 1.49952)
(87.3326, 1.50814)
(91.3659, 1.49296)
(95.5855, 1.51269)
(100, 1.49922)
}

\newcommand{\genTrainingPlotLossCifaruDdpmppInit}{%
(0.001, 0.999902)
(0.00104618, 0.999665)
(0.0010945, 1.00001)
(0.00114505, 0.999921)
(0.00119793, 0.99991)
(0.00125325, 1.00031)
(0.00131113, 1.00002)
(0.00137169, 1.00025)
(0.00143504, 1.00023)
(0.00150131, 1.00023)
(0.00157065, 0.999917)
(0.00164318, 1.00055)
(0.00171907, 1.00004)
(0.00179846, 1.00003)
(0.00188152, 1.00027)
(0.00196842, 1.00037)
(0.00205933, 0.999833)
(0.00215443, 1.0002)
(0.00225393, 1.00037)
(0.00235803, 0.999879)
(0.00246693, 0.999901)
(0.00258086, 0.999962)
(0.00270005, 1.00004)
(0.00282475, 1.00013)
(0.00295521, 0.999571)
(0.00309169, 1.00016)
(0.00323448, 0.999954)
(0.00338386, 0.999795)
(0.00354013, 0.999806)
(0.00370363, 0.999891)
(0.00387468, 1.00031)
(0.00405362, 1.00025)
(0.00424083, 0.999905)
(0.00443669, 1.00027)
(0.00464159, 1.00007)
(0.00485595, 1.00029)
(0.00508022, 0.999637)
(0.00531484, 1.00008)
(0.0055603, 0.999495)
(0.00581709, 0.999956)
(0.00608574, 0.999792)
(0.00636681, 1.00045)
(0.00666085, 0.999996)
(0.00696847, 1.00013)
(0.00729029, 0.999943)
(0.00762699, 1.00003)
(0.00797923, 1.00002)
(0.00834773, 0.999552)
(0.00873326, 0.999957)
(0.00913659, 0.999714)
(0.00955855, 1.00014)
(0.01, 1.00019)
(0.0104618, 1.00014)
(0.010945, 0.999886)
(0.0114505, 1.00021)
(0.0119793, 0.999426)
(0.0125325, 0.999809)
(0.0131113, 0.999472)
(0.0137169, 0.999826)
(0.0143504, 0.999925)
(0.0150131, 1.00021)
(0.0157065, 0.999793)
(0.0164318, 0.99978)
(0.0171907, 0.999525)
(0.0179846, 0.999742)
(0.0188152, 1.00007)
(0.0196842, 0.999932)
(0.0205933, 1.00039)
(0.0215443, 0.999562)
(0.0225393, 1.00044)
(0.0235803, 1.00035)
(0.0246693, 1.00026)
(0.0258086, 0.999608)
(0.0270005, 0.999703)
(0.0282475, 1.00014)
(0.0295521, 1.00024)
(0.0309169, 0.999908)
(0.0323448, 0.999579)
(0.0338386, 1.00033)
(0.0354013, 1.00009)
(0.0370363, 1.00029)
(0.0387468, 1.00041)
(0.0405362, 0.999964)
(0.0424083, 0.999932)
(0.0443669, 1.00018)
(0.0464159, 0.999965)
(0.0485595, 0.999567)
(0.0508022, 1.00058)
(0.0531484, 0.999985)
(0.055603, 1.00049)
(0.0581709, 0.999873)
(0.0608574, 1.00012)
(0.0636681, 1.00001)
(0.0666085, 1.00062)
(0.0696847, 1.00029)
(0.0729029, 0.999931)
(0.0762699, 1.0007)
(0.0797923, 1.00083)
(0.0834773, 1.00037)
(0.0873326, 1.00056)
(0.0913659, 1.00033)
(0.0955855, 1.00058)
(0.1, 1.00118)
(0.104618, 1.00054)
(0.10945, 1.00129)
(0.114505, 1.00076)
(0.119793, 1.00087)
(0.125325, 1.00116)
(0.131113, 1.00166)
(0.137169, 1.00239)
(0.143504, 1.00066)
(0.150131, 1.00113)
(0.157065, 1.00163)
(0.164318, 1.00282)
(0.171907, 1.00299)
(0.179846, 1.00271)
(0.188152, 1.00157)
(0.196842, 1.00412)
(0.205933, 1.00302)
(0.215443, 1.0035)
(0.225393, 1.00424)
(0.235803, 1.00542)
(0.246693, 1.0055)
(0.258086, 1.00438)
(0.270005, 1.00492)
(0.282475, 1.00549)
(0.295521, 1.00366)
(0.309169, 1.00862)
(0.323448, 1.00554)
(0.338386, 1.00841)
(0.354013, 1.00817)
(0.370363, 1.00897)
(0.387468, 1.0106)
(0.405362, 1.00631)
(0.424083, 1.01187)
(0.443669, 1.00975)
(0.464159, 1.01075)
(0.485595, 1.01028)
(0.508022, 1.01544)
(0.531484, 1.00772)
(0.55603, 1.0117)
(0.581709, 1.01345)
(0.608574, 1.01341)
(0.636681, 1.01335)
(0.666085, 1.00975)
(0.696847, 1.0189)
(0.729029, 1.01811)
(0.762699, 1.01184)
(0.797923, 1.0184)
(0.834773, 1.01839)
(0.873326, 1.02092)
(0.913659, 1.01299)
(0.955855, 1.01759)
(1, 1.01592)
(1.04618, 1.02106)
(1.0945, 1.02215)
(1.14505, 1.01622)
(1.19793, 1.02544)
(1.25325, 1.02609)
(1.31113, 1.01949)
(1.37169, 1.0172)
(1.43504, 1.01536)
(1.50131, 1.01929)
(1.57065, 1.01544)
(1.64318, 1.021)
(1.71907, 1.01341)
(1.79846, 1.02588)
(1.88152, 1.02105)
(1.96842, 1.02194)
(2.05933, 1.02422)
(2.15443, 1.02643)
(2.25393, 1.0271)
(2.35803, 1.02079)
(2.46693, 1.02446)
(2.58086, 1.01764)
(2.70005, 1.01667)
(2.82475, 1.01301)
(2.95521, 1.02928)
(3.09169, 1.02102)
(3.23448, 1.02256)
(3.38386, 1.01791)
(3.54013, 1.01667)
(3.70363, 1.03152)
(3.87468, 1.02304)
(4.05362, 1.01996)
(4.24083, 1.032)
(4.43669, 1.02429)
(4.64159, 1.02395)
(4.85595, 1.02883)
(5.08022, 1.02259)
(5.31484, 1.02608)
(5.5603, 1.01199)
(5.81709, 1.0342)
(6.08574, 1.01962)
(6.36681, 1.01678)
(6.66085, 1.02497)
(6.96847, 1.02654)
(7.29029, 1.02341)
(7.62699, 1.025)
(7.97923, 1.02388)
(8.34773, 1.02471)
(8.73326, 1.02475)
(9.13659, 1.02478)
(9.55855, 1.015)
(10, 1.02096)
(10.4618, 1.01782)
(10.945, 1.01965)
(11.4505, 1.01427)
(11.9793, 1.02923)
(12.5325, 1.02717)
(13.1113, 1.03589)
(13.7169, 1.02867)
(14.3504, 1.01925)
(15.0131, 1.02094)
(15.7065, 1.0199)
(16.4318, 1.02022)
(17.1907, 1.01546)
(17.9846, 1.02399)
(18.8152, 1.01778)
(19.6842, 1.02076)
(20.5933, 1.02358)
(21.5443, 1.02777)
(22.5393, 1.02566)
(23.5803, 1.02274)
(24.6693, 1.02827)
(25.8086, 1.02535)
(27.0005, 1.02197)
(28.2475, 1.02859)
(29.5521, 1.0256)
(30.9169, 1.02695)
(32.3448, 1.02027)
(33.8386, 1.01807)
(35.4013, 1.02752)
(37.0363, 1.01826)
(38.7468, 1.02698)
(40.5362, 1.02441)
(42.4083, 1.03646)
(44.3669, 1.0199)
(46.4159, 1.02612)
(48.5595, 1.02387)
(50.8022, 1.03019)
(53.1484, 1.02375)
(55.603, 1.02019)
(58.1709, 1.03182)
(60.8574, 1.0311)
(63.6681, 1.03008)
(66.6085, 1.02222)
(69.6847, 1.01526)
(72.9029, 1.02212)
(76.2699, 1.02902)
(79.7923, 1.02372)
(83.4773, 1.02668)
(87.3326, 1.02385)
(91.3659, 1.0251)
(95.5855, 1.01825)
(100, 1.02166)
}

\newcommand{\genTrainingPlotLossCifaruDdpmppInitLo}{%
(0.001, 0.974522)
(0.00104618, 0.974212)
(0.0010945, 0.974297)
(0.00114505, 0.974562)
(0.00119793, 0.974301)
(0.00125325, 0.974978)
(0.00131113, 0.974527)
(0.00137169, 0.974786)
(0.00143504, 0.974459)
(0.00150131, 0.974623)
(0.00157065, 0.974333)
(0.00164318, 0.975006)
(0.00171907, 0.974544)
(0.00179846, 0.974226)
(0.00188152, 0.974559)
(0.00196842, 0.975104)
(0.00205933, 0.974173)
(0.00215443, 0.974785)
(0.00225393, 0.974948)
(0.00235803, 0.974393)
(0.00246693, 0.974569)
(0.00258086, 0.974598)
(0.00270005, 0.974396)
(0.00282475, 0.974571)
(0.00295521, 0.974334)
(0.00309169, 0.974765)
(0.00323448, 0.974663)
(0.00338386, 0.974401)
(0.00354013, 0.974178)
(0.00370363, 0.973949)
(0.00387468, 0.975)
(0.00405362, 0.975111)
(0.00424083, 0.974231)
(0.00443669, 0.975362)
(0.00464159, 0.974527)
(0.00485595, 0.974585)
(0.00508022, 0.974396)
(0.00531484, 0.974577)
(0.0055603, 0.973955)
(0.00581709, 0.974418)
(0.00608574, 0.973991)
(0.00636681, 0.974638)
(0.00666085, 0.974687)
(0.00696847, 0.974678)
(0.00729029, 0.974149)
(0.00762699, 0.975025)
(0.00797923, 0.974448)
(0.00834773, 0.974243)
(0.00873326, 0.974511)
(0.00913659, 0.9742)
(0.00955855, 0.974642)
(0.01, 0.974609)
(0.0104618, 0.974537)
(0.010945, 0.974208)
(0.0114505, 0.974666)
(0.0119793, 0.97394)
(0.0125325, 0.974368)
(0.0131113, 0.97393)
(0.0137169, 0.974254)
(0.0143504, 0.974489)
(0.0150131, 0.974664)
(0.0157065, 0.974144)
(0.0164318, 0.974244)
(0.0171907, 0.974355)
(0.0179846, 0.97434)
(0.0188152, 0.974766)
(0.0196842, 0.974219)
(0.0205933, 0.974893)
(0.0215443, 0.973974)
(0.0225393, 0.975006)
(0.0235803, 0.974554)
(0.0246693, 0.974708)
(0.0258086, 0.973975)
(0.0270005, 0.974577)
(0.0282475, 0.974447)
(0.0295521, 0.974802)
(0.0309169, 0.974308)
(0.0323448, 0.973971)
(0.0338386, 0.974521)
(0.0354013, 0.974648)
(0.0370363, 0.974698)
(0.0387468, 0.97453)
(0.0405362, 0.973927)
(0.0424083, 0.973963)
(0.0443669, 0.974585)
(0.0464159, 0.97421)
(0.0485595, 0.973774)
(0.0508022, 0.97453)
(0.0531484, 0.97424)
(0.055603, 0.9742)
(0.0581709, 0.97343)
(0.0608574, 0.973483)
(0.0636681, 0.973583)
(0.0666085, 0.973638)
(0.0696847, 0.973169)
(0.0729029, 0.972239)
(0.0762699, 0.972215)
(0.0797923, 0.972491)
(0.0834773, 0.971462)
(0.0873326, 0.971377)
(0.0913659, 0.969873)
(0.0955855, 0.969221)
(0.1, 0.969169)
(0.104618, 0.967707)
(0.10945, 0.966684)
(0.114505, 0.964936)
(0.119793, 0.96314)
(0.125325, 0.962137)
(0.131113, 0.960307)
(0.137169, 0.958633)
(0.143504, 0.953873)
(0.150131, 0.951843)
(0.157065, 0.949043)
(0.164318, 0.946415)
(0.171907, 0.9424)
(0.179846, 0.939284)
(0.188152, 0.933754)
(0.196842, 0.930515)
(0.205933, 0.924872)
(0.215443, 0.91937)
(0.225393, 0.915292)
(0.235803, 0.908752)
(0.246693, 0.901241)
(0.258086, 0.895403)
(0.270005, 0.886637)
(0.282475, 0.879799)
(0.295521, 0.870925)
(0.309169, 0.86424)
(0.323448, 0.855223)
(0.338386, 0.84479)
(0.354013, 0.836703)
(0.370363, 0.828102)
(0.387468, 0.816374)
(0.405362, 0.804755)
(0.424083, 0.796069)
(0.443669, 0.781871)
(0.464159, 0.773847)
(0.485595, 0.763655)
(0.508022, 0.752266)
(0.531484, 0.742277)
(0.55603, 0.728415)
(0.581709, 0.719793)
(0.608574, 0.705613)
(0.636681, 0.693498)
(0.666085, 0.685761)
(0.696847, 0.682853)
(0.729029, 0.67009)
(0.762699, 0.658572)
(0.797923, 0.647253)
(0.834773, 0.648902)
(0.873326, 0.6345)
(0.913659, 0.618957)
(0.955855, 0.617127)
(1, 0.611506)
(1.04618, 0.597841)
(1.0945, 0.598581)
(1.14505, 0.590592)
(1.19793, 0.588026)
(1.25325, 0.585572)
(1.31113, 0.574926)
(1.37169, 0.569419)
(1.43504, 0.562559)
(1.50131, 0.559714)
(1.57065, 0.55785)
(1.64318, 0.557323)
(1.71907, 0.544184)
(1.79846, 0.547519)
(1.88152, 0.541269)
(1.96842, 0.538393)
(2.05933, 0.53968)
(2.15443, 0.539577)
(2.25393, 0.541781)
(2.35803, 0.533457)
(2.46693, 0.538902)
(2.58086, 0.529711)
(2.70005, 0.517297)
(2.82475, 0.525768)
(2.95521, 0.530197)
(3.09169, 0.525799)
(3.23448, 0.529894)
(3.38386, 0.521498)
(3.54013, 0.528519)
(3.70363, 0.521003)
(3.87468, 0.527075)
(4.05362, 0.514238)
(4.24083, 0.536406)
(4.43669, 0.526527)
(4.64159, 0.522346)
(4.85595, 0.519073)
(5.08022, 0.518043)
(5.31484, 0.516199)
(5.5603, 0.514098)
(5.81709, 0.525332)
(6.08574, 0.511275)
(6.36681, 0.510673)
(6.66085, 0.522042)
(6.96847, 0.515369)
(7.29029, 0.514038)
(7.62699, 0.518127)
(7.97923, 0.511706)
(8.34773, 0.514327)
(8.73326, 0.516204)
(9.13659, 0.51295)
(9.55855, 0.518768)
(10, 0.514163)
(10.4618, 0.51444)
(10.945, 0.513347)
(11.4505, 0.50823)
(11.9793, 0.524174)
(12.5325, 0.515312)
(13.1113, 0.521153)
(13.7169, 0.508278)
(14.3504, 0.514868)
(15.0131, 0.507426)
(15.7065, 0.514602)
(16.4318, 0.509998)
(17.1907, 0.511551)
(17.9846, 0.511345)
(18.8152, 0.505852)
(19.6842, 0.517532)
(20.5933, 0.512185)
(21.5443, 0.509252)
(22.5393, 0.517361)
(23.5803, 0.514839)
(24.6693, 0.511748)
(25.8086, 0.513339)
(27.0005, 0.514808)
(28.2475, 0.516573)
(29.5521, 0.510346)
(30.9169, 0.51725)
(32.3448, 0.511529)
(33.8386, 0.512461)
(35.4013, 0.514269)
(37.0363, 0.517763)
(38.7468, 0.512628)
(40.5362, 0.511686)
(42.4083, 0.51728)
(44.3669, 0.50868)
(46.4159, 0.512218)
(48.5595, 0.515457)
(50.8022, 0.517853)
(53.1484, 0.520658)
(55.603, 0.516409)
(58.1709, 0.512062)
(60.8574, 0.511094)
(63.6681, 0.517208)
(66.6085, 0.513246)
(69.6847, 0.503247)
(72.9029, 0.516399)
(76.2699, 0.512259)
(79.7923, 0.512419)
(83.4773, 0.516937)
(87.3326, 0.51038)
(91.3659, 0.507148)
(95.5855, 0.507181)
(100, 0.514895)
}

\newcommand{\genTrainingPlotLossCifaruDdpmppInitHi}{%
(0.001, 1.02528)
(0.00104618, 1.02512)
(0.0010945, 1.02573)
(0.00114505, 1.02528)
(0.00119793, 1.02552)
(0.00125325, 1.02564)
(0.00131113, 1.0255)
(0.00137169, 1.02571)
(0.00143504, 1.026)
(0.00150131, 1.02583)
(0.00157065, 1.0255)
(0.00164318, 1.02609)
(0.00171907, 1.02554)
(0.00179846, 1.02584)
(0.00188152, 1.02599)
(0.00196842, 1.02564)
(0.00205933, 1.02549)
(0.00215443, 1.02561)
(0.00225393, 1.02578)
(0.00235803, 1.02537)
(0.00246693, 1.02523)
(0.00258086, 1.02533)
(0.00270005, 1.02568)
(0.00282475, 1.0257)
(0.00295521, 1.02481)
(0.00309169, 1.02555)
(0.00323448, 1.02524)
(0.00338386, 1.02519)
(0.00354013, 1.02544)
(0.00370363, 1.02583)
(0.00387468, 1.02562)
(0.00405362, 1.02538)
(0.00424083, 1.02558)
(0.00443669, 1.02518)
(0.00464159, 1.02561)
(0.00485595, 1.026)
(0.00508022, 1.02488)
(0.00531484, 1.02559)
(0.0055603, 1.02503)
(0.00581709, 1.02549)
(0.00608574, 1.02559)
(0.00636681, 1.02626)
(0.00666085, 1.0253)
(0.00696847, 1.02559)
(0.00729029, 1.02574)
(0.00762699, 1.02504)
(0.00797923, 1.0256)
(0.00834773, 1.02486)
(0.00873326, 1.0254)
(0.00913659, 1.02523)
(0.00955855, 1.02564)
(0.01, 1.02578)
(0.0104618, 1.02574)
(0.010945, 1.02556)
(0.0114505, 1.02576)
(0.0119793, 1.02491)
(0.0125325, 1.02525)
(0.0131113, 1.02501)
(0.0137169, 1.0254)
(0.0143504, 1.02536)
(0.0150131, 1.02576)
(0.0157065, 1.02544)
(0.0164318, 1.02532)
(0.0171907, 1.0247)
(0.0179846, 1.02514)
(0.0188152, 1.02537)
(0.0196842, 1.02565)
(0.0205933, 1.0259)
(0.0215443, 1.02515)
(0.0225393, 1.02586)
(0.0235803, 1.02614)
(0.0246693, 1.0258)
(0.0258086, 1.02524)
(0.0270005, 1.02483)
(0.0282475, 1.02584)
(0.0295521, 1.02568)
(0.0309169, 1.02551)
(0.0323448, 1.02519)
(0.0338386, 1.02614)
(0.0354013, 1.02554)
(0.0370363, 1.02588)
(0.0387468, 1.02629)
(0.0405362, 1.026)
(0.0424083, 1.0259)
(0.0443669, 1.02578)
(0.0464159, 1.02572)
(0.0485595, 1.02536)
(0.0508022, 1.02662)
(0.0531484, 1.02573)
(0.055603, 1.02677)
(0.0581709, 1.02632)
(0.0608574, 1.02676)
(0.0636681, 1.02644)
(0.0666085, 1.02761)
(0.0696847, 1.0274)
(0.0729029, 1.02762)
(0.0762699, 1.02919)
(0.0797923, 1.02916)
(0.0834773, 1.02928)
(0.0873326, 1.02974)
(0.0913659, 1.0308)
(0.0955855, 1.03194)
(0.1, 1.03319)
(0.104618, 1.03337)
(0.10945, 1.03589)
(0.114505, 1.03659)
(0.119793, 1.03859)
(0.125325, 1.04018)
(0.131113, 1.04301)
(0.137169, 1.04615)
(0.143504, 1.04745)
(0.150131, 1.05042)
(0.157065, 1.05421)
(0.164318, 1.05922)
(0.171907, 1.06358)
(0.179846, 1.06614)
(0.188152, 1.06939)
(0.196842, 1.07772)
(0.205933, 1.08117)
(0.215443, 1.08763)
(0.225393, 1.09318)
(0.235803, 1.1021)
(0.246693, 1.10975)
(0.258086, 1.11337)
(0.270005, 1.12321)
(0.282475, 1.13117)
(0.295521, 1.1364)
(0.309169, 1.15299)
(0.323448, 1.15586)
(0.338386, 1.17202)
(0.354013, 1.17965)
(0.370363, 1.18984)
(0.387468, 1.20482)
(0.405362, 1.20786)
(0.424083, 1.22767)
(0.443669, 1.23762)
(0.464159, 1.24765)
(0.485595, 1.25691)
(0.508022, 1.27861)
(0.531484, 1.27316)
(0.55603, 1.29498)
(0.581709, 1.3071)
(0.608574, 1.32121)
(0.636681, 1.3332)
(0.666085, 1.33373)
(0.696847, 1.35496)
(0.729029, 1.36613)
(0.762699, 1.3651)
(0.797923, 1.38954)
(0.834773, 1.38788)
(0.873326, 1.40734)
(0.913659, 1.40702)
(0.955855, 1.41805)
(1, 1.42034)
(1.04618, 1.44427)
(1.0945, 1.44572)
(1.14505, 1.44185)
(1.19793, 1.46285)
(1.25325, 1.46662)
(1.31113, 1.46405)
(1.37169, 1.46498)
(1.43504, 1.46817)
(1.50131, 1.47887)
(1.57065, 1.47303)
(1.64318, 1.48467)
(1.71907, 1.48264)
(1.79846, 1.50425)
(1.88152, 1.50083)
(1.96842, 1.50548)
(2.05933, 1.50876)
(2.15443, 1.51329)
(2.25393, 1.51241)
(2.35803, 1.50812)
(2.46693, 1.51002)
(2.58086, 1.50558)
(2.70005, 1.51604)
(2.82475, 1.50025)
(2.95521, 1.52836)
(3.09169, 1.51625)
(3.23448, 1.51522)
(3.38386, 1.51433)
(3.54013, 1.50483)
(3.70363, 1.54204)
(3.87468, 1.51901)
(4.05362, 1.52569)
(4.24083, 1.52759)
(4.43669, 1.52205)
(4.64159, 1.52556)
(4.85595, 1.53858)
(5.08022, 1.52713)
(5.31484, 1.53596)
(5.5603, 1.50989)
(5.81709, 1.54307)
(6.08574, 1.52797)
(6.36681, 1.52289)
(6.66085, 1.5279)
(6.96847, 1.53771)
(7.29029, 1.53278)
(7.62699, 1.53188)
(7.97923, 1.53606)
(8.34773, 1.53509)
(8.73326, 1.5333)
(9.13659, 1.53662)
(9.55855, 1.51124)
(10, 1.52777)
(10.4618, 1.5212)
(10.945, 1.52596)
(11.4505, 1.5203)
(11.9793, 1.53429)
(12.5325, 1.53902)
(13.1113, 1.55062)
(13.7169, 1.54905)
(14.3504, 1.52364)
(15.0131, 1.53444)
(15.7065, 1.52521)
(16.4318, 1.53044)
(17.1907, 1.51936)
(17.9846, 1.53664)
(18.8152, 1.5297)
(19.6842, 1.52398)
(20.5933, 1.53498)
(21.5443, 1.54628)
(22.5393, 1.53395)
(23.5803, 1.53064)
(24.6693, 1.54479)
(25.8086, 1.53735)
(27.0005, 1.52913)
(28.2475, 1.54061)
(29.5521, 1.54085)
(30.9169, 1.53665)
(32.3448, 1.52902)
(33.8386, 1.52368)
(35.4013, 1.54077)
(37.0363, 1.51876)
(38.7468, 1.54133)
(40.5362, 1.53714)
(42.4083, 1.55564)
(44.3669, 1.53112)
(46.4159, 1.54002)
(48.5595, 1.53229)
(50.8022, 1.54253)
(53.1484, 1.52683)
(55.603, 1.52397)
(58.1709, 1.55159)
(60.8574, 1.55111)
(63.6681, 1.54295)
(66.6085, 1.5312)
(69.6847, 1.52727)
(72.9029, 1.52783)
(76.2699, 1.54578)
(79.7923, 1.53501)
(83.4773, 1.53643)
(87.3326, 1.53732)
(91.3659, 1.54304)
(95.5855, 1.52932)
(100, 1.52843)
}

\newcommand{\genTrainingPlotLossFfhqDdpmppPrec}{%
(0.001, 0.970795)
(0.00104618, 0.968752)
(0.0010945, 0.965576)
(0.00114505, 0.962998)
(0.00119793, 0.959566)
(0.00125325, 0.956441)
(0.00131113, 0.953069)
(0.00137169, 0.948407)
(0.00143504, 0.94537)
(0.00150131, 0.940572)
(0.00157065, 0.936997)
(0.00164318, 0.930641)
(0.00171907, 0.925823)
(0.00179846, 0.920103)
(0.00188152, 0.915594)
(0.00196842, 0.908833)
(0.00205933, 0.902965)
(0.00215443, 0.89525)
(0.00225393, 0.890239)
(0.00235803, 0.881178)
(0.00246693, 0.875409)
(0.00258086, 0.865675)
(0.00270005, 0.858844)
(0.00282475, 0.8497)
(0.00295521, 0.840683)
(0.00309169, 0.829283)
(0.00323448, 0.822953)
(0.00338386, 0.811423)
(0.00354013, 0.802229)
(0.00370363, 0.791482)
(0.00387468, 0.782155)
(0.00405362, 0.770786)
(0.00424083, 0.759153)
(0.00443669, 0.748987)
(0.00464159, 0.737585)
(0.00485595, 0.725462)
(0.00508022, 0.715264)
(0.00531484, 0.702731)
(0.0055603, 0.692426)
(0.00581709, 0.679979)
(0.00608574, 0.667652)
(0.00636681, 0.655819)
(0.00666085, 0.64463)
(0.00696847, 0.632913)
(0.00729029, 0.61956)
(0.00762699, 0.608979)
(0.00797923, 0.598214)
(0.00834773, 0.585883)
(0.00873326, 0.573913)
(0.00913659, 0.560871)
(0.00955855, 0.549783)
(0.01, 0.538758)
(0.0104618, 0.526861)
(0.010945, 0.516148)
(0.0114505, 0.506247)
(0.0119793, 0.493968)
(0.0125325, 0.483226)
(0.0131113, 0.472298)
(0.0137169, 0.461503)
(0.0143504, 0.452554)
(0.0150131, 0.44093)
(0.0157065, 0.432396)
(0.0164318, 0.421368)
(0.0171907, 0.410221)
(0.0179846, 0.402675)
(0.0188152, 0.392042)
(0.0196842, 0.382937)
(0.0205933, 0.373489)
(0.0215443, 0.364734)
(0.0225393, 0.355883)
(0.0235803, 0.347646)
(0.0246693, 0.337559)
(0.0258086, 0.329549)
(0.0270005, 0.32218)
(0.0282475, 0.312393)
(0.0295521, 0.304784)
(0.0309169, 0.297661)
(0.0323448, 0.289006)
(0.0338386, 0.282597)
(0.0354013, 0.274411)
(0.0370363, 0.267686)
(0.0387468, 0.259396)
(0.0405362, 0.25348)
(0.0424083, 0.245911)
(0.0443669, 0.239827)
(0.0464159, 0.234001)
(0.0485595, 0.226507)
(0.0508022, 0.219798)
(0.0531484, 0.214193)
(0.055603, 0.208496)
(0.0581709, 0.202972)
(0.0608574, 0.196292)
(0.0636681, 0.191749)
(0.0666085, 0.186121)
(0.0696847, 0.180535)
(0.0729029, 0.175417)
(0.0762699, 0.170066)
(0.0797923, 0.16471)
(0.0834773, 0.160282)
(0.0873326, 0.155313)
(0.0913659, 0.150828)
(0.0955855, 0.147246)
(0.1, 0.14297)
(0.104618, 0.138956)
(0.10945, 0.134837)
(0.114505, 0.130579)
(0.119793, 0.126981)
(0.125325, 0.123097)
(0.131113, 0.119125)
(0.137169, 0.115561)
(0.143504, 0.112935)
(0.150131, 0.109877)
(0.157065, 0.10669)
(0.164318, 0.103043)
(0.171907, 0.100965)
(0.179846, 0.0980896)
(0.188152, 0.0962099)
(0.196842, 0.093274)
(0.205933, 0.0911321)
(0.215443, 0.0884553)
(0.225393, 0.0864534)
(0.235803, 0.0848696)
(0.246693, 0.0830058)
(0.258086, 0.0809584)
(0.270005, 0.0796153)
(0.282475, 0.0780575)
(0.295521, 0.0766687)
(0.309169, 0.0757964)
(0.323448, 0.0743122)
(0.338386, 0.0738232)
(0.354013, 0.0726589)
(0.370363, 0.0718314)
(0.387468, 0.0712861)
(0.405362, 0.0707765)
(0.424083, 0.070319)
(0.443669, 0.0705339)
(0.464159, 0.0703762)
(0.485595, 0.0701864)
(0.508022, 0.0703376)
(0.531484, 0.0706093)
(0.55603, 0.0711068)
(0.581709, 0.0718112)
(0.608574, 0.0723601)
(0.636681, 0.0730824)
(0.666085, 0.0739691)
(0.696847, 0.0753415)
(0.729029, 0.0766894)
(0.762699, 0.0780274)
(0.797923, 0.0792918)
(0.834773, 0.0811631)
(0.873326, 0.0836397)
(0.913659, 0.0850509)
(0.955855, 0.0874132)
(1, 0.0900625)
(1.04618, 0.0918922)
(1.0945, 0.0961083)
(1.14505, 0.098674)
(1.19793, 0.101706)
(1.25325, 0.104904)
(1.31113, 0.108873)
(1.37169, 0.111816)
(1.43504, 0.116348)
(1.50131, 0.12112)
(1.57065, 0.125983)
(1.64318, 0.130529)
(1.71907, 0.135299)
(1.79846, 0.140568)
(1.88152, 0.146129)
(1.96842, 0.151655)
(2.05933, 0.157037)
(2.15443, 0.163803)
(2.25393, 0.171394)
(2.35803, 0.178212)
(2.46693, 0.185031)
(2.58086, 0.19181)
(2.70005, 0.200057)
(2.82475, 0.207897)
(2.95521, 0.215612)
(3.09169, 0.225049)
(3.23448, 0.232958)
(3.38386, 0.242523)
(3.54013, 0.251646)
(3.70363, 0.260647)
(3.87468, 0.270665)
(4.05362, 0.281547)
(4.24083, 0.291214)
(4.43669, 0.301455)
(4.64159, 0.312757)
(4.85595, 0.32646)
(5.08022, 0.335054)
(5.31484, 0.347688)
(5.5603, 0.357388)
(5.81709, 0.368747)
(6.08574, 0.38262)
(6.36681, 0.396492)
(6.66085, 0.409004)
(6.96847, 0.421405)
(7.29029, 0.436123)
(7.62699, 0.447357)
(7.97923, 0.460179)
(8.34773, 0.476122)
(8.73326, 0.488167)
(9.13659, 0.504831)
(9.55855, 0.514237)
(10, 0.531)
(10.4618, 0.542619)
(10.945, 0.560161)
(11.4505, 0.571653)
(11.9793, 0.589021)
(12.5325, 0.597188)
(13.1113, 0.613251)
(13.7169, 0.632227)
(14.3504, 0.640431)
(15.0131, 0.661515)
(15.7065, 0.668847)
(16.4318, 0.687715)
(17.1907, 0.697901)
(17.9846, 0.709566)
(18.8152, 0.725939)
(19.6842, 0.740611)
(20.5933, 0.753881)
(21.5443, 0.769262)
(22.5393, 0.779144)
(23.5803, 0.788695)
(24.6693, 0.804998)
(25.8086, 0.811612)
(27.0005, 0.82555)
(28.2475, 0.835138)
(29.5521, 0.849269)
(30.9169, 0.857344)
(32.3448, 0.872048)
(33.8386, 0.871281)
(35.4013, 0.884917)
(37.0363, 0.891349)
(38.7468, 0.900695)
(40.5362, 0.904177)
(42.4083, 0.91581)
(44.3669, 0.924839)
(46.4159, 0.927671)
(48.5595, 0.934876)
(50.8022, 0.942275)
(53.1484, 0.948622)
(55.603, 0.947319)
(58.1709, 0.953846)
(60.8574, 0.961676)
(63.6681, 0.965846)
(66.6085, 0.966509)
(69.6847, 0.971808)
(72.9029, 0.973046)
(76.2699, 0.977349)
(79.7923, 0.984554)
(83.4773, 0.98664)
(87.3326, 0.984066)
(91.3659, 0.98708)
(95.5855, 0.991246)
(100, 0.995532)
}

\newcommand{\genTrainingPlotLossFfhqDdpmppPrecLo}{%
(0.001, 0.944967)
(0.00104618, 0.942666)
(0.0010945, 0.938261)
(0.00114505, 0.935164)
(0.00119793, 0.930897)
(0.00125325, 0.926487)
(0.00131113, 0.921668)
(0.00137169, 0.915385)
(0.00143504, 0.912727)
(0.00150131, 0.905404)
(0.00157065, 0.901733)
(0.00164318, 0.892713)
(0.00171907, 0.886529)
(0.00179846, 0.879363)
(0.00188152, 0.874427)
(0.00196842, 0.865014)
(0.00205933, 0.858259)
(0.00215443, 0.848481)
(0.00225393, 0.8432)
(0.00235803, 0.831137)
(0.00246693, 0.824864)
(0.00258086, 0.812404)
(0.00270005, 0.804757)
(0.00282475, 0.793646)
(0.00295521, 0.78319)
(0.00309169, 0.768826)
(0.00323448, 0.761923)
(0.00338386, 0.748922)
(0.00354013, 0.737394)
(0.00370363, 0.725231)
(0.00387468, 0.715173)
(0.00405362, 0.701839)
(0.00424083, 0.689193)
(0.00443669, 0.677069)
(0.00464159, 0.663879)
(0.00485595, 0.651432)
(0.00508022, 0.641917)
(0.00531484, 0.627338)
(0.0055603, 0.616215)
(0.00581709, 0.602815)
(0.00608574, 0.590574)
(0.00636681, 0.578321)
(0.00666085, 0.565199)
(0.00696847, 0.554355)
(0.00729029, 0.54037)
(0.00762699, 0.529331)
(0.00797923, 0.519139)
(0.00834773, 0.506611)
(0.00873326, 0.494384)
(0.00913659, 0.481809)
(0.00955855, 0.470566)
(0.01, 0.45938)
(0.0104618, 0.448499)
(0.010945, 0.4386)
(0.0114505, 0.427954)
(0.0119793, 0.416005)
(0.0125325, 0.406672)
(0.0131113, 0.396169)
(0.0137169, 0.385221)
(0.0143504, 0.37792)
(0.0150131, 0.366481)
(0.0157065, 0.359115)
(0.0164318, 0.348603)
(0.0171907, 0.338682)
(0.0179846, 0.331623)
(0.0188152, 0.323318)
(0.0196842, 0.313226)
(0.0205933, 0.30558)
(0.0215443, 0.297235)
(0.0225393, 0.288935)
(0.0235803, 0.281646)
(0.0246693, 0.273356)
(0.0258086, 0.265201)
(0.0270005, 0.25928)
(0.0282475, 0.250479)
(0.0295521, 0.244761)
(0.0309169, 0.237677)
(0.0323448, 0.230092)
(0.0338386, 0.224193)
(0.0354013, 0.217062)
(0.0370363, 0.210871)
(0.0387468, 0.204146)
(0.0405362, 0.198425)
(0.0424083, 0.19226)
(0.0443669, 0.187333)
(0.0464159, 0.181855)
(0.0485595, 0.175015)
(0.0508022, 0.169632)
(0.0531484, 0.165697)
(0.055603, 0.160045)
(0.0581709, 0.155603)
(0.0608574, 0.149463)
(0.0636681, 0.145575)
(0.0666085, 0.141459)
(0.0696847, 0.136428)
(0.0729029, 0.132332)
(0.0762699, 0.127432)
(0.0797923, 0.123041)
(0.0834773, 0.119236)
(0.0873326, 0.1147)
(0.0913659, 0.111467)
(0.0955855, 0.108825)
(0.1, 0.105774)
(0.104618, 0.10197)
(0.10945, 0.0983899)
(0.114505, 0.0952386)
(0.119793, 0.0920824)
(0.125325, 0.0887176)
(0.131113, 0.086183)
(0.137169, 0.0832839)
(0.143504, 0.0809073)
(0.150131, 0.0784977)
(0.157065, 0.0760582)
(0.164318, 0.0730193)
(0.171907, 0.0717659)
(0.179846, 0.0697981)
(0.188152, 0.0676588)
(0.196842, 0.0657248)
(0.205933, 0.0637222)
(0.215443, 0.0615426)
(0.225393, 0.0602165)
(0.235803, 0.0588535)
(0.246693, 0.0573615)
(0.258086, 0.0559327)
(0.270005, 0.0550297)
(0.282475, 0.0535684)
(0.295521, 0.0526582)
(0.309169, 0.0518826)
(0.323448, 0.0508576)
(0.338386, 0.0498351)
(0.354013, 0.049477)
(0.370363, 0.0488743)
(0.387468, 0.0479579)
(0.405362, 0.0477066)
(0.424083, 0.0472934)
(0.443669, 0.0474902)
(0.464159, 0.0469776)
(0.485595, 0.0467032)
(0.508022, 0.04697)
(0.531484, 0.0470951)
(0.55603, 0.0469801)
(0.581709, 0.0476397)
(0.608574, 0.0479692)
(0.636681, 0.0488917)
(0.666085, 0.0490404)
(0.696847, 0.0496674)
(0.729029, 0.0506648)
(0.762699, 0.0512485)
(0.797923, 0.0524717)
(0.834773, 0.0534358)
(0.873326, 0.0549236)
(0.913659, 0.0557749)
(0.955855, 0.0574203)
(1, 0.0597375)
(1.04618, 0.0605441)
(1.0945, 0.0629773)
(1.14505, 0.0652794)
(1.19793, 0.0675229)
(1.25325, 0.0696806)
(1.31113, 0.0720196)
(1.37169, 0.0740002)
(1.43504, 0.07679)
(1.50131, 0.0798308)
(1.57065, 0.0835218)
(1.64318, 0.0865339)
(1.71907, 0.0898257)
(1.79846, 0.0930438)
(1.88152, 0.096999)
(1.96842, 0.100965)
(2.05933, 0.104956)
(2.15443, 0.109595)
(2.25393, 0.113986)
(2.35803, 0.119014)
(2.46693, 0.12344)
(2.58086, 0.128883)
(2.70005, 0.134223)
(2.82475, 0.139949)
(2.95521, 0.145392)
(3.09169, 0.151386)
(3.23448, 0.157361)
(3.38386, 0.164034)
(3.54013, 0.171125)
(3.70363, 0.176768)
(3.87468, 0.183766)
(4.05362, 0.191602)
(4.24083, 0.196531)
(4.43669, 0.205435)
(4.64159, 0.213888)
(4.85595, 0.222277)
(5.08022, 0.227591)
(5.31484, 0.239066)
(5.5603, 0.244681)
(5.81709, 0.252556)
(6.08574, 0.261939)
(6.36681, 0.271285)
(6.66085, 0.280291)
(6.96847, 0.28882)
(7.29029, 0.299349)
(7.62699, 0.307005)
(7.97923, 0.314068)
(8.34773, 0.324138)
(8.73326, 0.333182)
(9.13659, 0.344124)
(9.55855, 0.349635)
(10, 0.36042)
(10.4618, 0.369095)
(10.945, 0.379094)
(11.4505, 0.388403)
(11.9793, 0.399599)
(12.5325, 0.402971)
(13.1113, 0.412708)
(13.7169, 0.423287)
(14.3504, 0.428666)
(15.0131, 0.443761)
(15.7065, 0.444005)
(16.4318, 0.455032)
(17.1907, 0.463783)
(17.9846, 0.467949)
(18.8152, 0.477789)
(19.6842, 0.486723)
(20.5933, 0.492925)
(21.5443, 0.502514)
(22.5393, 0.502432)
(23.5803, 0.515529)
(24.6693, 0.51859)
(25.8086, 0.527484)
(27.0005, 0.52849)
(28.2475, 0.53906)
(29.5521, 0.542178)
(30.9169, 0.547851)
(32.3448, 0.554004)
(33.8386, 0.549847)
(35.4013, 0.563041)
(37.0363, 0.565931)
(38.7468, 0.570855)
(40.5362, 0.569164)
(42.4083, 0.576661)
(44.3669, 0.574216)
(46.4159, 0.579538)
(48.5595, 0.587179)
(50.8022, 0.586869)
(53.1484, 0.594918)
(55.603, 0.595741)
(58.1709, 0.596721)
(60.8574, 0.597593)
(63.6681, 0.594803)
(66.6085, 0.603815)
(69.6847, 0.589854)
(72.9029, 0.597954)
(76.2699, 0.610749)
(79.7923, 0.606972)
(83.4773, 0.606935)
(87.3326, 0.605207)
(91.3659, 0.613907)
(95.5855, 0.607689)
(100, 0.616754)
}

\newcommand{\genTrainingPlotLossFfhqDdpmppPrecHi}{%
(0.001, 0.996624)
(0.00104618, 0.994838)
(0.0010945, 0.992891)
(0.00114505, 0.990832)
(0.00119793, 0.988235)
(0.00125325, 0.986396)
(0.00131113, 0.984471)
(0.00137169, 0.981429)
(0.00143504, 0.978014)
(0.00150131, 0.97574)
(0.00157065, 0.972261)
(0.00164318, 0.968569)
(0.00171907, 0.965116)
(0.00179846, 0.960844)
(0.00188152, 0.95676)
(0.00196842, 0.952651)
(0.00205933, 0.947671)
(0.00215443, 0.94202)
(0.00225393, 0.937278)
(0.00235803, 0.931219)
(0.00246693, 0.925953)
(0.00258086, 0.918946)
(0.00270005, 0.912931)
(0.00282475, 0.905755)
(0.00295521, 0.898177)
(0.00309169, 0.889739)
(0.00323448, 0.883983)
(0.00338386, 0.873924)
(0.00354013, 0.867064)
(0.00370363, 0.857732)
(0.00387468, 0.849137)
(0.00405362, 0.839732)
(0.00424083, 0.829112)
(0.00443669, 0.820906)
(0.00464159, 0.811292)
(0.00485595, 0.799493)
(0.00508022, 0.78861)
(0.00531484, 0.778124)
(0.0055603, 0.768637)
(0.00581709, 0.757142)
(0.00608574, 0.74473)
(0.00636681, 0.733317)
(0.00666085, 0.724062)
(0.00696847, 0.71147)
(0.00729029, 0.69875)
(0.00762699, 0.688627)
(0.00797923, 0.67729)
(0.00834773, 0.665154)
(0.00873326, 0.653443)
(0.00913659, 0.639933)
(0.00955855, 0.628999)
(0.01, 0.618136)
(0.0104618, 0.605223)
(0.010945, 0.593695)
(0.0114505, 0.58454)
(0.0119793, 0.571932)
(0.0125325, 0.559779)
(0.0131113, 0.548427)
(0.0137169, 0.537785)
(0.0143504, 0.527187)
(0.0150131, 0.515378)
(0.0157065, 0.505677)
(0.0164318, 0.494133)
(0.0171907, 0.48176)
(0.0179846, 0.473727)
(0.0188152, 0.460766)
(0.0196842, 0.452648)
(0.0205933, 0.441398)
(0.0215443, 0.432232)
(0.0225393, 0.422831)
(0.0235803, 0.413646)
(0.0246693, 0.401762)
(0.0258086, 0.393898)
(0.0270005, 0.385081)
(0.0282475, 0.374307)
(0.0295521, 0.364808)
(0.0309169, 0.357645)
(0.0323448, 0.347919)
(0.0338386, 0.341001)
(0.0354013, 0.33176)
(0.0370363, 0.3245)
(0.0387468, 0.314647)
(0.0405362, 0.308536)
(0.0424083, 0.299563)
(0.0443669, 0.292322)
(0.0464159, 0.286147)
(0.0485595, 0.277999)
(0.0508022, 0.269965)
(0.0531484, 0.262689)
(0.055603, 0.256948)
(0.0581709, 0.250341)
(0.0608574, 0.243121)
(0.0636681, 0.237923)
(0.0666085, 0.230782)
(0.0696847, 0.224642)
(0.0729029, 0.218503)
(0.0762699, 0.212701)
(0.0797923, 0.206379)
(0.0834773, 0.201328)
(0.0873326, 0.195926)
(0.0913659, 0.190189)
(0.0955855, 0.185668)
(0.1, 0.180167)
(0.104618, 0.175942)
(0.10945, 0.171283)
(0.114505, 0.165918)
(0.119793, 0.161879)
(0.125325, 0.157476)
(0.131113, 0.152067)
(0.137169, 0.147838)
(0.143504, 0.144963)
(0.150131, 0.141256)
(0.157065, 0.137321)
(0.164318, 0.133067)
(0.171907, 0.130165)
(0.179846, 0.126381)
(0.188152, 0.124761)
(0.196842, 0.120823)
(0.205933, 0.118542)
(0.215443, 0.115368)
(0.225393, 0.11269)
(0.235803, 0.110886)
(0.246693, 0.10865)
(0.258086, 0.105984)
(0.270005, 0.104201)
(0.282475, 0.102547)
(0.295521, 0.100679)
(0.309169, 0.0997103)
(0.323448, 0.0977668)
(0.338386, 0.0978113)
(0.354013, 0.0958408)
(0.370363, 0.0947884)
(0.387468, 0.0946143)
(0.405362, 0.0938465)
(0.424083, 0.0933446)
(0.443669, 0.0935776)
(0.464159, 0.0937748)
(0.485595, 0.0936695)
(0.508022, 0.0937052)
(0.531484, 0.0941235)
(0.55603, 0.0952336)
(0.581709, 0.0959827)
(0.608574, 0.096751)
(0.636681, 0.0972731)
(0.666085, 0.0988977)
(0.696847, 0.101016)
(0.729029, 0.102714)
(0.762699, 0.104806)
(0.797923, 0.106112)
(0.834773, 0.10889)
(0.873326, 0.112356)
(0.913659, 0.114327)
(0.955855, 0.117406)
(1, 0.120387)
(1.04618, 0.12324)
(1.0945, 0.129239)
(1.14505, 0.132069)
(1.19793, 0.135889)
(1.25325, 0.140128)
(1.31113, 0.145727)
(1.37169, 0.149631)
(1.43504, 0.155905)
(1.50131, 0.162409)
(1.57065, 0.168444)
(1.64318, 0.174524)
(1.71907, 0.180773)
(1.79846, 0.188093)
(1.88152, 0.195259)
(1.96842, 0.202345)
(2.05933, 0.209117)
(2.15443, 0.218011)
(2.25393, 0.228802)
(2.35803, 0.23741)
(2.46693, 0.246622)
(2.58086, 0.254736)
(2.70005, 0.265892)
(2.82475, 0.275845)
(2.95521, 0.285833)
(3.09169, 0.298712)
(3.23448, 0.308554)
(3.38386, 0.321012)
(3.54013, 0.332167)
(3.70363, 0.344526)
(3.87468, 0.357564)
(4.05362, 0.371492)
(4.24083, 0.385897)
(4.43669, 0.397475)
(4.64159, 0.411626)
(4.85595, 0.430643)
(5.08022, 0.442518)
(5.31484, 0.456309)
(5.5603, 0.470094)
(5.81709, 0.484938)
(6.08574, 0.503301)
(6.36681, 0.5217)
(6.66085, 0.537717)
(6.96847, 0.553989)
(7.29029, 0.572897)
(7.62699, 0.587709)
(7.97923, 0.606291)
(8.34773, 0.628106)
(8.73326, 0.643152)
(9.13659, 0.665538)
(9.55855, 0.678838)
(10, 0.70158)
(10.4618, 0.716143)
(10.945, 0.741227)
(11.4505, 0.754902)
(11.9793, 0.778442)
(12.5325, 0.791406)
(13.1113, 0.813794)
(13.7169, 0.841168)
(14.3504, 0.852196)
(15.0131, 0.879269)
(15.7065, 0.893688)
(16.4318, 0.920397)
(17.1907, 0.93202)
(17.9846, 0.951182)
(18.8152, 0.974089)
(19.6842, 0.9945)
(20.5933, 1.01484)
(21.5443, 1.03601)
(22.5393, 1.05586)
(23.5803, 1.06186)
(24.6693, 1.09141)
(25.8086, 1.09574)
(27.0005, 1.12261)
(28.2475, 1.13122)
(29.5521, 1.15636)
(30.9169, 1.16684)
(32.3448, 1.19009)
(33.8386, 1.19272)
(35.4013, 1.20679)
(37.0363, 1.21677)
(38.7468, 1.23054)
(40.5362, 1.23919)
(42.4083, 1.25496)
(44.3669, 1.27546)
(46.4159, 1.2758)
(48.5595, 1.28257)
(50.8022, 1.29768)
(53.1484, 1.30233)
(55.603, 1.2989)
(58.1709, 1.31097)
(60.8574, 1.32576)
(63.6681, 1.33689)
(66.6085, 1.3292)
(69.6847, 1.35376)
(72.9029, 1.34814)
(76.2699, 1.34395)
(79.7923, 1.36214)
(83.4773, 1.36635)
(87.3326, 1.36292)
(91.3659, 1.36025)
(95.5855, 1.3748)
(100, 1.37431)
}

\newcommand{\genTrainingPlotChurnCifaruDdpmppOrig}{%
(0, 2.9325)
(5, 2.6727)
(10, 2.5744)
(15, 2.4327)
(20, 2.3358)
(25, 2.3527)
(30, 2.2738)
(35, 2.3143)
(40, 2.3215)
(45, 2.3613)
(50, 2.325)
(55, 2.3848)
(60, 2.4057)
(65, 2.4339)
(70, 2.4797)
(75, 2.5291)
(80, 2.5801)
(85, 2.5831)
(90, 2.6149)
(95, 2.6859)
(100, 2.7303)
}

\newcommand{\genTrainingPlotChurnCifaruDdpmppOrigLo}{%
(0, 2.9325)
(5, 2.6727)
(10, 2.5744)
(15, 2.4327)
(20, 2.3358)
(25, 2.3527)
(30, 2.2738)
(35, 2.3143)
(40, 2.3215)
(45, 2.3613)
(50, 2.325)
(55, 2.3848)
(60, 2.4057)
(65, 2.4339)
(70, 2.4797)
(75, 2.5291)
(80, 2.5801)
(85, 2.5831)
(90, 2.6149)
(95, 2.6859)
(100, 2.7303)
}

\newcommand{\genTrainingPlotChurnCifaruDdpmppOrigHi}{%
(0, 2.9889)
(5, 2.7267)
(10, 2.5912)
(15, 2.5097)
(20, 2.3967)
(25, 2.3823)
(30, 2.3506)
(35, 2.3393)
(40, 2.4031)
(45, 2.3721)
(50, 2.3996)
(55, 2.4351)
(60, 2.4643)
(65, 2.496)
(70, 2.5358)
(75, 2.5371)
(80, 2.6014)
(85, 2.6438)
(90, 2.6469)
(95, 2.7123)
(100, 2.7678)
}

\newcommand{\genTrainingPlotChurnCifaruNcsnppOrig}{%
(0, 3.7343)
(5, 3.3556)
(10, 3.0772)
(15, 2.913)
(20, 2.7897)
(25, 2.6445)
(30, 2.5774)
(35, 2.5194)
(40, 2.4641)
(45, 2.408)
(50, 2.3798)
(55, 2.3687)
(60, 2.3485)
(65, 2.3592)
(70, 2.3166)
(75, 2.3072)
(80, 2.323)
(85, 2.3316)
(90, 2.3432)
(95, 2.2605)
(100, 2.3154)
}

\newcommand{\genTrainingPlotChurnCifaruNcsnppOrigLo}{%
(0, 3.7343)
(5, 3.3556)
(10, 3.0772)
(15, 2.913)
(20, 2.7897)
(25, 2.6445)
(30, 2.5774)
(35, 2.5194)
(40, 2.4641)
(45, 2.408)
(50, 2.3798)
(55, 2.3687)
(60, 2.3485)
(65, 2.3592)
(70, 2.3166)
(75, 2.3072)
(80, 2.323)
(85, 2.3316)
(90, 2.3432)
(95, 2.2605)
(100, 2.3154)
}

\newcommand{\genTrainingPlotChurnCifaruNcsnppOrigHi}{%
(0, 3.8719)
(5, 3.4672)
(10, 3.2023)
(15, 2.9532)
(20, 2.8392)
(25, 2.7233)
(30, 2.6342)
(35, 2.5613)
(40, 2.5025)
(45, 2.459)
(50, 2.4252)
(55, 2.4524)
(60, 2.3985)
(65, 2.4007)
(70, 2.3444)
(75, 2.3513)
(80, 2.3651)
(85, 2.3536)
(90, 2.349)
(95, 2.3328)
(100, 2.349)
}

\newcommand{\genTrainingPlotChurnCifaruDdpmppAugm}{%
(0, 1.9743)
(5, 2.0188)
(10, 2.0325)
(15, 2.0935)
(20, 2.1371)
(25, 2.2509)
(30, 2.3311)
(35, 2.3992)
(40, 2.5272)
(45, 2.5904)
(50, 2.666)
(55, 2.7388)
(60, 2.9346)
(65, 3.0218)
(70, 3.0659)
(75, 3.1277)
(80, 3.2549)
(85, 3.3654)
(90, 3.4588)
(95, 3.5365)
(100, 3.6108)
}

\newcommand{\genTrainingPlotChurnCifaruDdpmppAugmLo}{%
(0, 1.9743)
(5, 2.0188)
(10, 2.0325)
(15, 2.0935)
(20, 2.1371)
(25, 2.2509)
(30, 2.3311)
(35, 2.3992)
(40, 2.5272)
(45, 2.5904)
(50, 2.666)
(55, 2.7388)
(60, 2.9346)
(65, 3.0218)
(70, 3.0659)
(75, 3.1277)
(80, 3.2549)
(85, 3.3654)
(90, 3.4588)
(95, 3.5365)
(100, 3.6108)
}

\newcommand{\genTrainingPlotChurnCifaruDdpmppAugmHi}{%
(0, 2.0162)
(5, 2.0365)
(10, 2.0677)
(15, 2.1234)
(20, 2.2096)
(25, 2.2731)
(30, 2.3924)
(35, 2.4763)
(40, 2.5847)
(45, 2.6328)
(50, 2.7585)
(55, 2.9861)
(60, 2.9841)
(65, 3.0467)
(70, 3.1497)
(75, 3.2707)
(80, 3.3332)
(85, 3.3941)
(90, 3.5717)
(95, 3.5854)
(100, 3.6769)
}

\newcommand{\genTrainingPlotChurnCifaruNcsnppAugm}{%
(0, 1.9184)
(5, 1.9393)
(10, 1.9765)
(15, 2.0357)
(20, 2.1009)
(25, 2.1663)
(30, 2.254)
(35, 2.3444)
(40, 2.3944)
(45, 2.4817)
(50, 2.6355)
(55, 2.6584)
(60, 2.7598)
(65, 2.8367)
(70, 2.9685)
(75, 3.0071)
(80, 3.0025)
(85, 3.136)
(90, 3.2769)
(95, 3.395)
(100, 3.3584)
}

\newcommand{\genTrainingPlotChurnCifaruNcsnppAugmLo}{%
(0, 1.9184)
(5, 1.9393)
(10, 1.9765)
(15, 2.0357)
(20, 2.1009)
(25, 2.1663)
(30, 2.254)
(35, 2.3444)
(40, 2.3944)
(45, 2.4817)
(50, 2.6355)
(55, 2.6584)
(60, 2.7598)
(65, 2.8367)
(70, 2.9685)
(75, 3.0071)
(80, 3.0025)
(85, 3.136)
(90, 3.2769)
(95, 3.395)
(100, 3.3584)
}

\newcommand{\genTrainingPlotChurnCifaruNcsnppAugmHi}{%
(0, 2.0012)
(5, 1.96)
(10, 2.0154)
(15, 2.0842)
(20, 2.1356)
(25, 2.211)
(30, 2.2723)
(35, 2.3543)
(40, 2.4788)
(45, 2.5467)
(50, 2.6775)
(55, 2.7145)
(60, 2.8333)
(65, 2.8798)
(70, 2.983)
(75, 3.0997)
(80, 3.1875)
(85, 3.2299)
(90, 3.3313)
(95, 3.4694)
(100, 3.4766)
}

\newcommand{\genTrainingPlotImgcOrig}{%
(0, 2.6363)
(5, 2.2398)
(10, 2.0252)
(15, 1.9378)
(20, 1.8534)
(25, 1.7925)
(30, 1.7597)
(35, 1.7399)
(40, 1.6788)
(45, 1.6419)
(50, 1.6459)
(55, 1.6285)
(60, 1.618)
(65, 1.6022)
(70, 1.6166)
(75, 1.5726)
(80, 1.5874)
(85, 1.6096)
(90, 1.5889)
(95, 1.6165)
(100, 1.5827)
}

\newcommand{\genTrainingPlotImgcOrigLo}{%
(0, 2.6363)
(5, 2.2398)
(10, 2.0252)
(15, 1.9378)
(20, 1.8534)
(25, 1.7925)
(30, 1.7597)
(35, 1.7399)
(40, 1.6788)
(45, 1.6419)
(50, 1.6459)
(55, 1.6285)
(60, 1.618)
(65, 1.6022)
(70, 1.6166)
(75, 1.5726)
(80, 1.5874)
(85, 1.6096)
(90, 1.5889)
(95, 1.6165)
(100, 1.5827)
}

\newcommand{\genTrainingPlotImgcOrigHi}{%
(0, 2.715)
(5, 2.3212)
(10, 2.1144)
(15, 2.0056)
(20, 1.9126)
(25, 1.8176)
(30, 1.7838)
(35, 1.7964)
(40, 1.7284)
(45, 1.7034)
(50, 1.6926)
(55, 1.6852)
(60, 1.6732)
(65, 1.6715)
(70, 1.6736)
(75, 1.6274)
(80, 1.6567)
(85, 1.6528)
(90, 1.6198)
(95, 1.6314)
(100, 1.6554)
}

\newcommand{\genTrainingPlotImgcOurs}{%
(0, 2.2246)
(5, 1.8202)
(10, 1.6285)
(15, 1.5386)
(20, 1.478)
(25, 1.436)
(30, 1.4037)
(35, 1.4192)
(40, 1.365)
(45, 1.3639)
(50, 1.3651)
(55, 1.3823)
(60, 1.3693)
(65, 1.3878)
(70, 1.3801)
(75, 1.3977)
(80, 1.4338)
(85, 1.449)
(90, 1.4653)
(95, 1.4788)
(100, 1.4859)
}

\newcommand{\genTrainingPlotImgcOursLo}{%
(0, 2.2246)
(5, 1.8202)
(10, 1.6285)
(15, 1.5386)
(20, 1.478)
(25, 1.436)
(30, 1.4037)
(35, 1.4192)
(40, 1.365)
(45, 1.3639)
(50, 1.3651)
(55, 1.3823)
(60, 1.3693)
(65, 1.3878)
(70, 1.3801)
(75, 1.3977)
(80, 1.4338)
(85, 1.449)
(90, 1.4653)
(95, 1.4788)
(100, 1.4859)
}

\newcommand{\genTrainingPlotImgcOursHi}{%
(0, 2.2246)
(5, 1.8621)
(10, 1.6915)
(15, 1.5913)
(20, 1.5428)
(25, 1.4666)
(30, 1.4549)
(35, 1.4505)
(40, 1.4049)
(45, 1.4137)
(50, 1.4202)
(55, 1.446)
(60, 1.4265)
(65, 1.4225)
(70, 1.4526)
(75, 1.4308)
(80, 1.4598)
(85, 1.4863)
(90, 1.5)
(95, 1.5114)
(100, 1.514)
}

\newcommand{\figDenoising}{%
\renewcommand{\hh}{0.495\linewidth}
\newcommand{\denhruler}{\makebox[\hh]{\scriptsize\hrlabel{$\sigma{=}$0}\hrlabel{0.2}\hrlabel{0.5}\hrlabel{1}\hrlabel{2}\hrlabel{3}\hrlabel{5}\hrlabel{7}\hrlabel{10}\hrlabel{20}\hrlabel{50}}}
\begin{figure}[t]%
\centering%
\footnotesize%
\denhruler\hfill\denhruler\\%
\includegraphics[width=\hh, trim=0 128 0 0, clip]{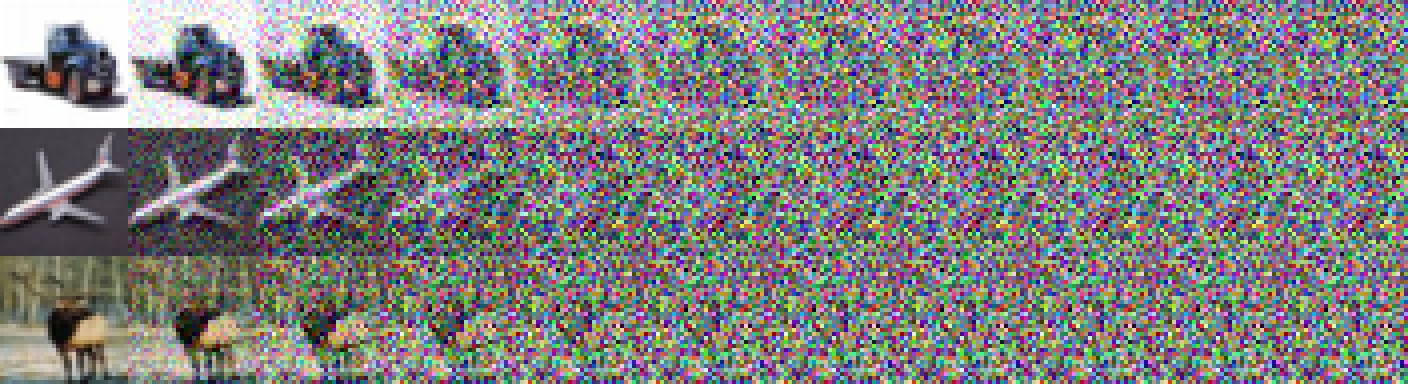}\hfill%
\includegraphics[width=\hh, trim=0 128 0 0, clip]{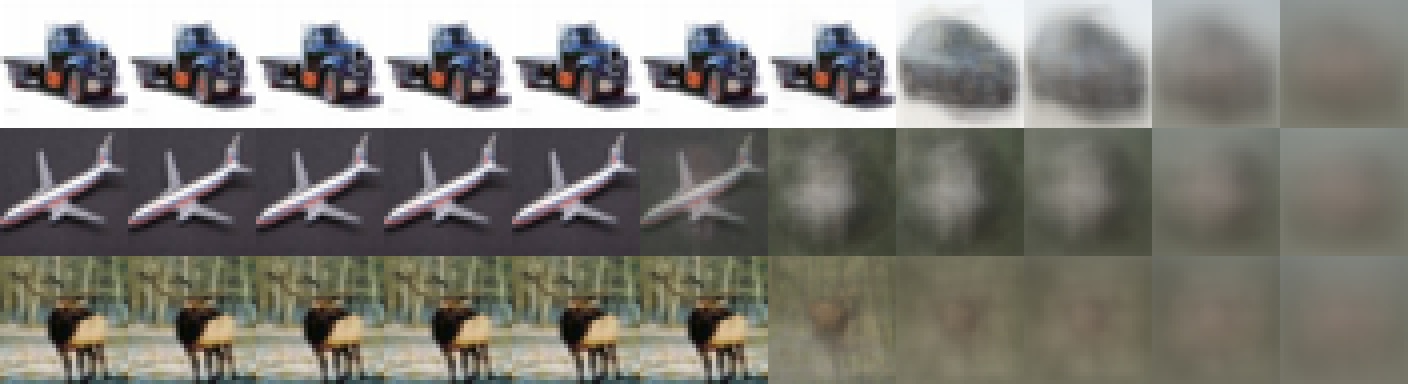}\\%
\makebox[\hh]{(a) Noisy images drawn from $p(\xx; \sigma)$}\hfill%
\makebox[\hh]{(b) Ideal denoiser outputs $D(\xx; \sigma)$}\\%
\caption{\label{fig:denoising}%
Denoising score matching on CIFAR-10.
\textbf{(a)}
Images from the training set corrupted with varying levels of additive Gaussian noise.
High levels of noise lead to oversaturated colors; we normalize the images for cleaner visualization.
\textbf{(b)}
Optimal denoising result from minimizing Eq.~\ref{eq:score} analytically ({see Appendix~\ref{app:scorematching}}). With increasing noise level, the {result} approaches dataset mean.
}%
\vspace*{-2mm}
\end{figure}
}

\newcommand{\eqq}[2]{\pbox[t][5ex]{\linewidth}{{#1}\\{#2}}}
\newcommand{\eqqq}[3]{{\pbox[t][8ex]{\linewidth}{{#1}\\{#2}\\{#3}}}}

\newcommand{\csbox}[1]{\makebox[2.0em][l]{#1}}
\newcommand{\cnbox}[2]{{#1}\hspace*{1em}\scalebox{0.9}{(note: #2)}}
\newcommand{\cpboxa}[1]{\makebox[1.2em][l]{#1}}
\newcommand{\cpboxb}[1]{\makebox[2.0em][l]{#1}}
\newcommand{\cpboxc}[1]{\makebox[1.4em][l]{#1}}
\newcommand{\cpboxd}[1]{\makebox[2em][l]{#1}}

\newcommand{\pphantom}{\vphantom{$= 1 / \sqrt{\sigma^2 + 1} \sdata^2 / \left(\sigma^2 + \sdata^2 \right)$}}
\newcommand{\lphantom}{\vphantom{$= \frac{1}{4} / \sqrt{\sigma^2 + 1} \sdata^2 / \left(\sigma^2 + \sdata^2 \right)$}}

\newcommand{\llphantom}{\lphantom\vphantom{$\underset{i}{\argmin}$}}
\newcommand{\titlerowww}[1]{\makebox[0mm][l]{{\bf #1}}\\}
\newcommand{\titlerowwb}[1]{{\bf #1}\pphantom}

\newcommand{\tabSpecifics}{
\tabulinesep=0.00ex%
\tabulinestyle{0.17mm}%
\begin{table}[t]%
\centering%
\caption{\label{tab:specifics}%
Specific design choices employed by different model families.
$N$ is the number of ODE solver iterations that we wish to execute during sampling.
The corresponding sequence of time steps is $\{t_0, t_1, \dots, t_N\}$, where $t_N = 0$.
If the model was originally trained for specific choices of $N$ and $\{t_i\}$, the originals are denoted by $\origN$ and $\{\origT{j}\}$, respectively.
The denoiser is defined as \smash{$D_\theta(\xx; \sigma) = \cskip(\sigma) \xx + \cout(\sigma) F_\theta \big( \cin(\sigma) \xx; \cnoise(\sigma) \big)$}; $F_\theta$ represents the raw neural network layers.
}
\vspace{1.5mm}%

\resizebox{\textwidth}{!}{%
\begin{tabu}{@{}lllll@{\hspace*{-1mm}}}
\tabucline{-}
& {\bf VP \cite{Song2021sde}}
& {\bf VE \cite{Song2021sde}}
& {\bf iDDPM \cite{Nichol2021a} + DDIM \cite{Song2020ddim}}
& {\bf {Ours (``EDM'')} \vphantom{\"O}}
\\
\hline\vspace*{-3.5mm}\\
\vspace*{.1ex}\titlerowww{Sampling (Section~\protect\ref*{sec:deterministic})}
ODE solver
& {Euler}
& {Euler}
& {Euler}
& {2\textsuperscript{nd} order Heun}\vspace*{-0.2ex}
\vspace*{0.0mm}\\
Time steps\hfill$t_{i<N}$
& \low{$1 + \frac{i}{N-1}(\epsilon_\text{s} - 1)$}
& \low{$\smax^2 \left( \smin^2 / \smax^2 \right)^{\frac{i}{N-1}}$}
& \eqqq
{$\origT{\lfloor j_0 + \frac{\origN - 1 - j_0}{N - 1} i + \frac{1}{2} \rfloor}$\scalebox{0.9}{, where}}
{\scalebox{0.8}{\hspace*{.35em}\raisebox{0mm}[2ex][-0.5ex]{\makebox[2em][l]{$\origT{\origN}$}$= 0$}}}
{\scalebox{0.8}{\hspace*{.35em}\makebox[2em][l]{$\origT{j-1}$}$= \sqrt{\frac{\origTsup{j}{2} + 1}{\max(\bar{\alpha}_{j-1} / \bar{\alpha}_j, C_1)}\!-\!1}$}}
& \eqq
{$\big({\smax}^\frac{1}{\rho} + {}$\vspace*{-.2ex}}
{\hspace*{0.5em}$ \frac{i}{N-1} ( {\smin}^\frac{1}{\rho}\!-\!{\smax}^\frac{1}{\rho} ) \big)^\rho$}
\vspace*{2mm}\\
Schedule\hfill$\sigma(t)$\lphantom
& {$\sqrt{e^{\frac{1}{2}\bdyn t^2 + \bmin t}\!-\!1}$}
& \csbox{$\sqrt{t}$}
& \csbox{$t$}
& \csbox{$t$}
\vspace*{0.5mm}\\
Scaling\hfill$s(t)$\llphantom
& \csbox{$1 \big/ \sqrt{e^{\frac{1}{2}\bdyn t^2 + \bmin t}}$}
& \csbox{$1$}
& \csbox{$1$}
& \csbox{$1$}
\vspace*{-1.0mm}\\
\hline\vspace*{-3.5mm}\\
\titlerowww{Network and preconditioning (Section~\protect\ref*{sec:training})}
Architecture of $F_\theta$\lphantom
& {\ddpmpp{}}
& {\ncsnpp{}}
& {\ddpm{}}
& {(any)}
\vspace*{0.7mm}\\
Skip scaling\hfill$\cskip(\sigma)$\lphantom
& \csbox{$1$}
& \csbox{$1$}
& \csbox{$1$}
& \csbox{$\sdata^2 / \left(\sigma^2 + \sdata^2 \right)$}
\vspace*{0.7mm}\\
Output scaling \hfill$\cout(\sigma)$\lphantom
& \csbox{$-\sigma$}
& \csbox{$\sigma$}
& \csbox{$-\sigma$}
& \csbox{$\sigma \cdot \sdata / \sqrt{\sdata^2 + \sigma^2}$}
\vspace*{0.7mm}\\
Input scaling\hfill$\cin(\sigma)$\lphantom
& \csbox{$1 / \sqrt{\sigma^2 + 1}$}
& \csbox{$1$}
& \csbox{$1 / \sqrt{\sigma^2 + 1}$}
& \csbox{$1 / \sqrt{\sigma^2 + \sdata^2}$}
\vspace*{0.7mm}\\
Noise cond.\hfill$\cnoise(\sigma)$\llphantom
& \csbox{$(\origN - 1) ~\sigma^{-1}(\sigma)$}
& \csbox{$\ln(\frac{1}{2} \sigma)$}
& \csbox{\scalebox{1.0}{$\origN\!-\!1\!-\!\argmin_j|\origT{j} - \sigma|$}}
& \csbox{$\frac{1}{4} \ln(\sigma)$}
\vspace*{-1.0mm}\\
\hline\vspace*{-3.5mm}\\
\titlerowww{Training (Section~\protect\ref*{sec:training})}
Noise distribution\lphantom
& {$\sigma^{-1}(\sigma) \sim \mathcal{U}(\epsilon_\text{t}, 1)$}
& {$\ln(\sigma)\!\sim\!\mathcal{U}(\ln(\smin),$}
& {$\sigma = \origT{j}, \ \  j \sim \mathcal{U}\{0, \origN\!-\!1\}$}
& {$\ln(\sigma) \sim \mathcal{N}(P_\stext{mean}^{}, P_\stext{std}^2)$}
\\
&& {\hspace{4.3em}$\ln(\smax))$} &&
\vspace*{-1.0mm}\\
Loss weighting\hfill$\lambda(\sigma)$\llphantom\lphantom
& {$1 / \sigma^2$}
& {$1 / \sigma^2$}
& \cnbox{$1 / \sigma^2$}{$^*$}
& {$\left( \sigma^2\!+\!\sdata^2 \right) / (\sigma \cdot \sdata)^2$}
\vspace*{-1.5mm}\\
\tabucline{-}\\
\vspace*{-7mm}\\
\titlerowwb{Parameters}
& \cpboxa{\footnotesize{$\bdyn$}}{\footnotesize{$= 19.9,  \bmin=0.1$}}
& \cpboxb{\footnotesize{$\smin$}}{\footnotesize{$= 0.02$}}
& \cpboxc{\footnotesize{$\bar{\alpha}_j$}}{\footnotesize{\smash{$= \sin^2( \frac{\pi}{2} \frac{j}{\origN (C_2 + 1)} )$}}}
& \cpboxd{\footnotesize{$\smin$}}{\footnotesize{$= 0.002, \smax = 80$}}
\\
\pphantom
& \cpboxa{\footnotesize{$\epsilon_\text{s}$}}{\footnotesize{$= 10^{-3}, \epsilon_\text{t} = 10^{-5}$}}
& \cpboxb{\footnotesize{$\smax$}}{\footnotesize{$= 100$}}
& \cpboxc{\footnotesize{$C_1$}}{\footnotesize{$= 0.001$, $C_2=0.008$}}
& \cpboxd{\footnotesize{$\sdata$}}{\footnotesize{$= 0.5, \rho = 7$}}
\\
\pphantom
& \cpboxa{\footnotesize{$\origN$}}{\footnotesize{$= 1000$}}
&
& \cpboxc{\footnotesize{$\origN$}}{\footnotesize{$= 1000, j_0 = 8^\dag$}}
& \cpboxd{\footnotesize{$P_\text{mean}$}}{\footnotesize{$= -1.2$, $P_\text{std} = 1.2$}}
\\
\tabucline{-}
\multicolumn{5}{l@{}}{\hfill%
${}^*$ iDDPM also employs a second loss term $L_\text{vlb}$\hfill%
${}^\dag$ In our tests, $j_0=8$ yielded better FID than $j_0=0$ used by iDDPM\hfill~}
\end{tabu}}%
\end{table}%
}

\newcommand{\figOdePlotNfe}{%
\renewcommand{\hh}{85mm}\renewcommand{\vv}{55mm}\renewcommand{\hhh}{-5mm}%
\begin{figure}[t]
\centering%
\resizebox{\textwidth}{!}{%
\hspace*{-1mm}%
\begin{tikzpicture}%
\begin{axis}[
  width=\hh, height=\vv,
  xmin={8}, xmax={1024}, xmode={log}, xtick={8, 16, 32, 64, 128, 256, 512, 1024}, xticklabels={\tickNFE{8}, $16$, $32$, $64$, $128$, $256$, $512$, $1024$},
  ymin={1.6}, ymax={400}, ymode={log}, ytick={2, 3, 5, 10, 20, 50, 100, 200, 400}, yticklabels={$2$, $3$, $5$, $10$, $20$, $50$, $100$, $200$, \tickFID},
  grid={major}, legend pos={north east}, legend cell align={left}, legend style={font=\small},
]
\addplot[C0]                             coordinates {\genOdePlotNfeCifaruDdpmppRepro};
\addplot[C1]                             coordinates {\genOdePlotNfeCifaruDdpmppEmu};
\addplot[C2]                             coordinates {\genOdePlotNfeCifaruDdpmppDHeun};
\addplot[C3]                             coordinates {\genOdePlotNfeCifaruDdpmppSched};
\addplot[black, dashed]                  coordinates {\genOdePlotNfeCifaruDdpmppBbox};
\addplot[C0, only marks, forget plot]    coordinates {\genOdePlotNfeCifaruDdpmppReproMarks};
\addplot[C1, only marks, forget plot]    coordinates {\genOdePlotNfeCifaruDdpmppEmuMarks};
\addplot[C2, only marks, forget plot]    coordinates {\genOdePlotNfeCifaruDdpmppDHeunMarks};
\addplot[C3, only marks, forget plot]    coordinates {\genOdePlotNfeCifaruDdpmppSchedMarks};
\addplot[black, only marks, forget plot] coordinates {\genOdePlotNfeCifaruDdpmppBboxMarks};
\node at (axis cs:35,3.0) [anchor={north}] {\textcolor{C3}{$35$}};
\end{axis}
\end{tikzpicture}
\hspace*{\hhh}
\begin{tikzpicture}
\begin{axis}[
  width=\hh, height=\vv,
  xmin={8}, xmax={8192}, xmode={log}, xtick={8, 32, 128, 512, 2048, 8192}, xticklabels={$8$, $32$, $128$, $512$, $2048$, $8192$},
  ymin={1.9}, ymax={950}, ymode={log}, ytick={2, 3, 5, 10, 20, 50, 100, 200, 500, 950}, yticklabels={$2$, $3$, $5$, $10$, $20$, $50$, $100$, $200$, $500$, \tickFID},
  grid={major}, legend pos={north east}, legend cell align={left}, legend style={font=\small},
]
\addplot[C0]                             coordinates {\genOdePlotNfeCifaruNcsnppRepro};
\addplot[C1]                             coordinates {\genOdePlotNfeCifaruNcsnppEmu};
\addplot[C2]                             coordinates {\genOdePlotNfeCifaruNcsnppDHeun};
\addplot[C3]                             coordinates {\genOdePlotNfeCifaruNcsnppSched};
\addplot[black, dashed]                  coordinates {\genOdePlotNfeCifaruNcsnppBbox};
\addplot[C0, only marks, forget plot]    coordinates {\genOdePlotNfeCifaruNcsnppReproMarks};
\addplot[C1, only marks, forget plot]    coordinates {\genOdePlotNfeCifaruNcsnppEmuMarks};
\addplot[C2, only marks, forget plot]    coordinates {\genOdePlotNfeCifaruNcsnppDHeunMarks};
\addplot[C3, only marks, forget plot]    coordinates {\genOdePlotNfeCifaruNcsnppSchedMarks};
\addplot[black, only marks, forget plot] coordinates {\genOdePlotNfeCifaruNcsnppBboxMarks};
\node at (axis cs:27,3.8) [anchor={north}] {\textcolor{C3}{$27$}};
\end{axis}
\end{tikzpicture}
\hspace*{\hhh}
\begin{tikzpicture}
\begin{axis}[
  width=\hh, height=\vv,
  xmin={8}, xmax={1024}, xmode={log}, xtick={8, 16, 32, 64, 128, 256, 512, 1024}, xticklabels={$8$, $16$, $32$, $64$, $128$, $256$, $512$, $1024$},
  ymin={1.9}, ymax={28}, ymode={log}, ytick={2, 3, 5, 10, 20, 28}, yticklabels={$2$, $3$, $5$, $10$, $20$, \tickFID},
  grid={major}, legend pos={north east}, legend cell align={left}, legend style={font=\small},
]
\addplot[C0]                                coordinates {\genOdePlotNfeImgcDhariwalRepro};
\addplot[C1]                                coordinates {\genOdePlotNfeImgcDhariwalEmu};
\addplot[C2]                                coordinates {\genOdePlotNfeImgcDhariwalDHeun};
\addplot[C3, dash pattern={on 6pt off 6pt}, forget plot] coordinates {\genOdePlotNfeImgcDhariwalDHeun};
\addlegendimage{C3}
\addplot[black, dashed]                     coordinates {\genOdePlotNfeImgcDhariwalBbox};
\addplot[C0, only marks, forget plot]       coordinates {\genOdePlotNfeImgcDhariwalReproMarks};
\addplot[C1, only marks, forget plot]       coordinates {\genOdePlotNfeImgcDhariwalEmuMarks};
\addplot[C3, only marks, forget plot]       coordinates {\genOdePlotNfeImgcDhariwalDHeunMarks};
\addplot[black, only marks, forget plot]    coordinates {\genOdePlotNfeImgcDhariwalBboxMarks};
\node at (axis cs:79,2.6) [anchor={north}] {\textcolor{C3}{$79$}};
\legend{
  {Original sampler},
  {Our reimplementation},
  {+ Heun \& our $\{t_i\}$},
  {+ Our $\sigma(t)$ \& $s(t)$},
  {Black-box RK45},
}
\end{axis}
\end{tikzpicture}
}\\\hfill%
\makebox[0.33\linewidth]{\footnotesize (a) Uncond. CIFAR-10, VP ODE}\hfill%
\makebox[0.33\linewidth]{\hspace*{-.30em}\footnotesize (b) Uncond. CIFAR-10, VE ODE}\hfill%
\makebox[0.33\linewidth]{\hspace*{-.15em}\footnotesize (c) Class-cond. ImageNet-64, DDIM}\hfill%
\caption{\label{fig:OdePlotNfe}%
Comparison of deterministic sampling methods using three pre-trained models. %
For each curve, the dot indicates the lowest NFE whose FID is within 3\% of the lowest observed FID.
}
\vspace*{-4mm}
\end{figure}
}

\newcommand{\algHeun}{%
\begin{algorithm}[t]
\footnotesize
\captionof{algorithm}[heun]{\atphantom\ \ Deterministic sampling using Heun's 2\textsuperscript{nd} order method with arbitrary $\sigma(t)$ and $s(t)$.}
\begin{spacing}{1.1}
\begin{algorithmic}[1]
  \AProcedure{HeunSampler}{$D_\theta(\xx;\sigma), ~\sigma(t), ~s(t), ~t_{i \in \{0, \dots, N\}}$}
    \AState{{\bf sample} $\xx_0 \sim \mathcal{N} \big( \boldzero, ~\sigma^2(\odetime_0) ~s^2(\odetime_0) ~\boldi \big)$}
        \AComment{Generate initial sample at $\odetime_0$}
    \AFor{$i \in \{0, \dots, N-1\}$}
        \AComment{Solve Eq.~\ref{eq:odescale} over $N$ time steps}
      \DState{$\displaystyle\dd_i \gets \bigg(\frac{\dot \sigma(\odetime_i)}{\sigma(\odetime_i)} + \frac{\dot \scale(\odetime_i)}{\scale(\odetime_i)}\bigg)\xx_i - \frac{\dot\sigma(\odetime_i)\scale(\odetime_i)}{\sigma(\odetime_i)} D_\theta\bigg(\frac{\xx_i}{\scale(\odetime_i)}; \sigma(\odetime_i)\bigg)$}
          \AComment{Evaluate $\diff\xx / \diff\odetime$ at $\odetime_i$}
      \AState{$\xx_{i+1} \gets \xx_i + (\odetime_{i+1} - \odetime_{i}) \dd_i$}
          \AComment{Take Euler step from $\odetime_i$ to $\odetime_{i+1}$}
      \vspace{1mm}
      \AIf{$\sigma(t_{i+1}) \ne 0$}
          \AComment{Apply 2\textsuperscript{nd} order correction unless $\sigma$ goes to zero}
        \DState{$\ddp_i \gets$ \scalebox{0.9}{$\displaystyle\bigg(\frac{\dot \sigma(\odetime_{i+1})}{\sigma(\odetime_{i+1})} + \frac{\dot \scale(\odetime_{i+1})}{\scale(\odetime_{i+1})}\bigg) \xx_{i+1}  - \frac{\dot\sigma(\odetime_{i+1})\scale(\odetime_{i+1})}{\sigma(\odetime_{i+1})} D_\theta\bigg(\frac{\xx_{i+1}}{\scale(\odetime_{i+1})}; \sigma(\odetime_{i+1})\bigg)$}}
            \AComment{Eval. $\diff\xx / \diff\odetime$ at $\odetime_{i+1}$}
        \AState{$\xx_{i+1} \gets \xx_i + (\odetime_{i+1} - \odetime_{i}) \big( \frac{1}{2}\dd_i + \frac{1}{2}\ddp_i \big)$}
            \AComment{Explicit trapezoidal rule at $\odetime_{i+1}$}
      \EndIf
    \EndFor
    \AState{\textbf{return} $\xx_N$}
        \AComment{Return noise-free sample at $\odetime_N$}
  \EndProcedure
\end{algorithmic}
\end{spacing}
\label{alg:heun}
\end{algorithm}
}

\newcommand{\slab}[1]{{\scalebox{0.7}{#1}}}
\newcommand{\iplot}[1]{%
\begin{tikzpicture}[inner sep=0pt, outer sep=0pt]
 \node {\includegraphics[width=0.333\linewidth, trim=5 2 15 15, clip]{trajectories/#1.pdf}};
 \draw (-2.0, 1.35) node {\slab{$\xx$}};
\end{tikzpicture}}
\newcommand{\dplot}[3]{%
\begin{tikzpicture}[inner sep=0pt, outer sep=0pt]
 \node {\includegraphics[width=0.333\linewidth, trim=5 2 15 15, clip]{trajectories/#1.pdf}};
 \draw[green!50!black] (#3, 0.020)  -- (-0.3, 1.34);
 \draw[green!50!black] (#3, -0.095) -- (-0.3, -0.21);
 \draw (0.77, 0.50) node {\includegraphics[width=0.20\linewidth, trim=5 2 15 15, clip]{trajectories/#2.pdf}};
 \draw (-2.05, 1.35) node {\slab{$\xx$}};
\end{tikzpicture}}
\newcommand{\figTrajectories}{%
\begin{figure}[t]
\centering
\begin{tabular}{@{}c@{}c@{}c@{}}
\iplot{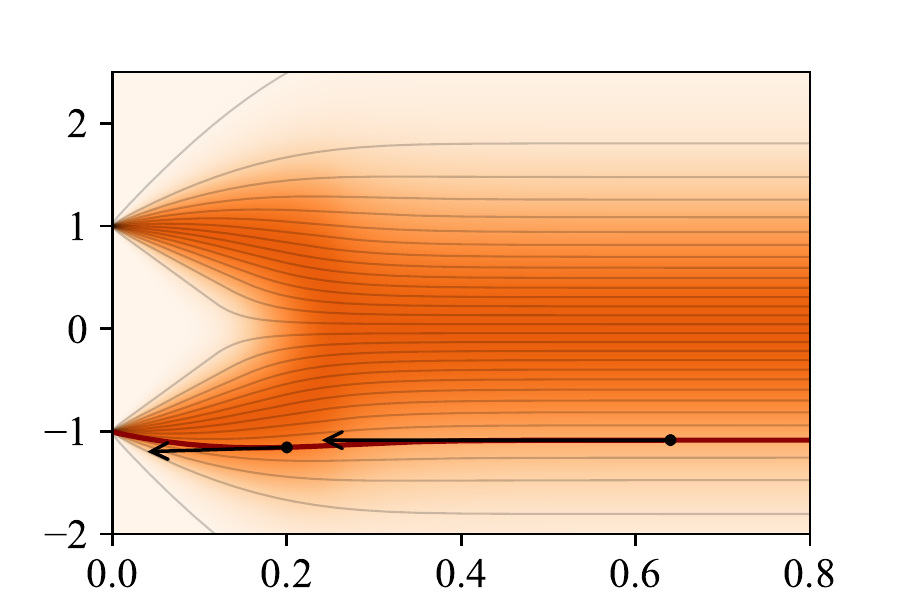} &
\dplot{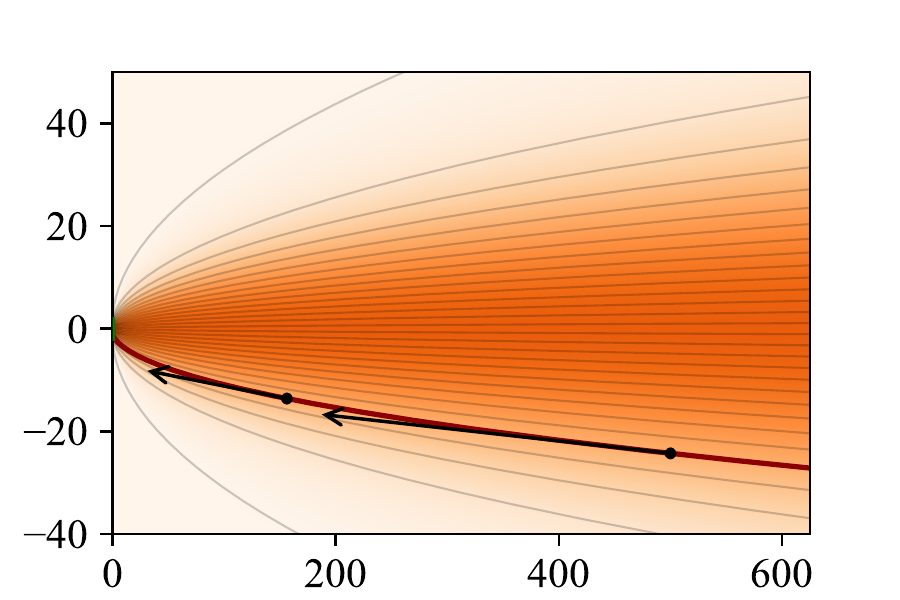}{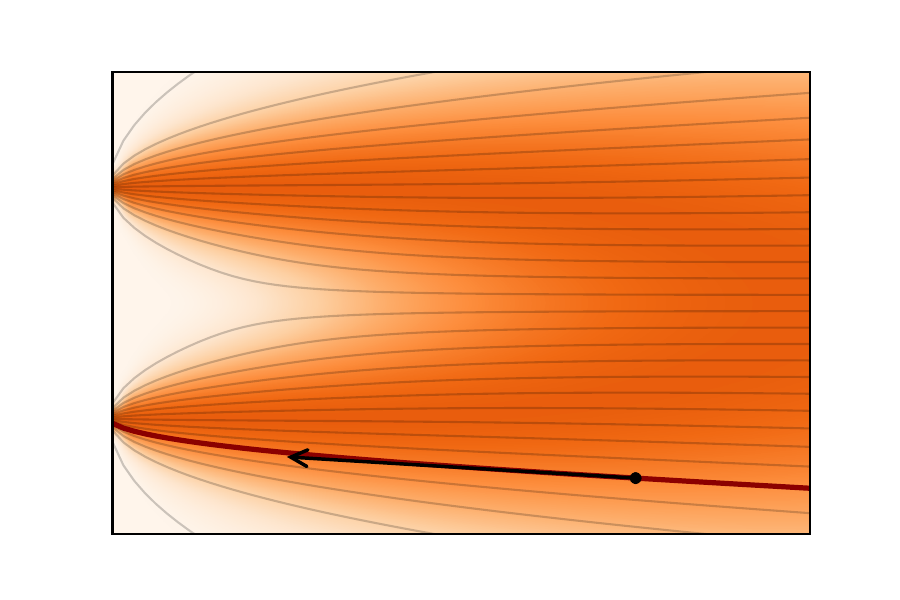}{-1.78} &
\dplot{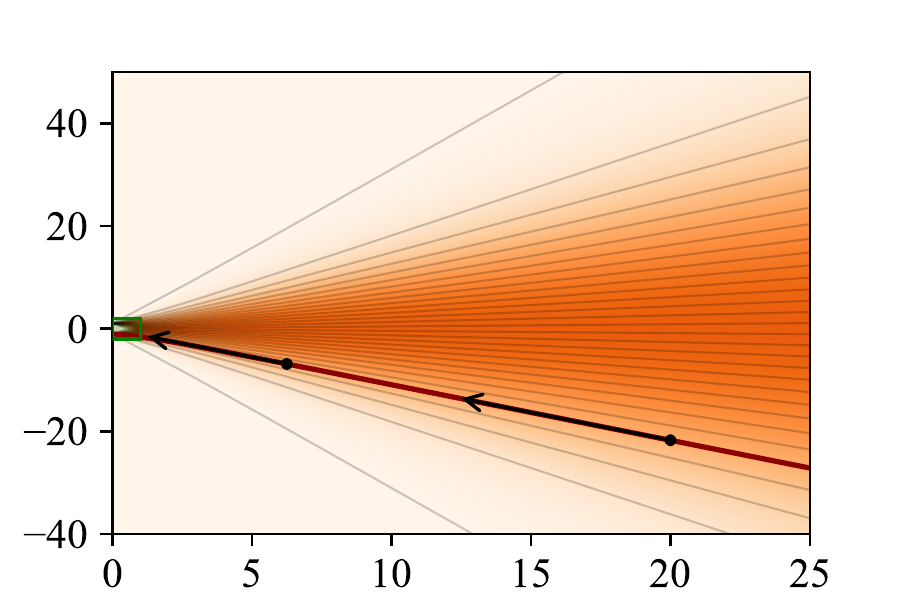}{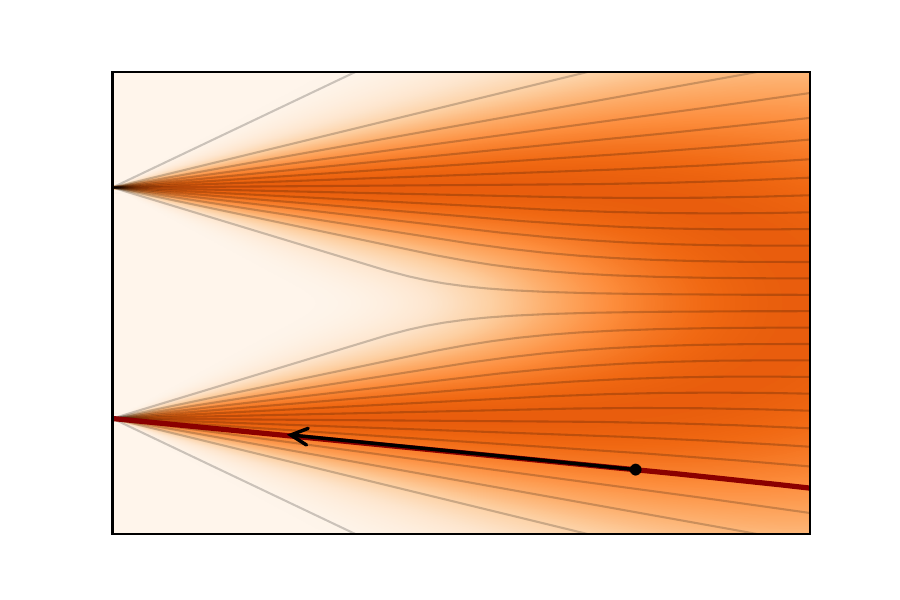}{-1.64} \\
\raisebox{2.63ex}[0pt][0pt]{\hspace*{-11.8em}\slab{$t$$=$}}&%
\raisebox{2.63ex}[0pt][0pt]{\hspace*{-11.3em}\slab{$t$$=$}}&%
\raisebox{2.63ex}[0pt][0pt]{\hspace*{-11.3em}\slab{$t$$=$}}\vspace*{-4mm}\\
{\footnotesize (a) Variance preserving ODE~\cite{Song2021sde}} &
{\footnotesize (b) Variance exploding ODE~\cite{Song2021sde}} &
{\footnotesize (c) DDIM~\cite{Song2020ddim} / Our ODE} \\
\end{tabular}
\caption{\label{fig:trajectories}%
A sketch of ODE curvature in 1D where $\pdata$ is two Dirac peaks at $\xx=\pm1$.
Horizontal $t$ axis is chosen to show $\sigma\in[0,25]$ in each plot, with insets showing $\sigma\in[0,1]$ near the data.
Example local gradients are shown with black arrows.
\textbf{(a)}
Variance preserving ODE of Song et al.~\cite{Song2021sde} has solution trajectories that flatten out to horizontal lines at large $\sigma$.
Local gradients start pointing towards data only at small $\sigma$.
\textbf{(b)}
Variance exploding variant has extreme curvature near data and the solution trajectories are curved everywhere.
\textbf{(c)}
With the schedule used by DDIM~\cite{Song2020ddim} and us, as $\sigma$ increases the solution trajectories approach straight lines that point towards the mean of data.
As $\sigma\to0$, the trajectories become linear and point towards the data manifold.
}%
\end{figure}
}

\newcommand{\hsqrt}[1]{\raisebox{-.0ex}{$\sqrt{\raisebox{0pt}[1.8ex][0.3ex]{$#1$}}$}}
\newcommand{\algStochastic}{%
\begin{algorithm}[t]
\footnotesize
\captionof{algorithm}[stochastic]{\atphantom\ \ Our stochastic sampler with $\sigma(t)=t$ and $s(t)=1$.}
\begin{spacing}{1.1}
\begin{algorithmic}[1]
  \AProcedure{StochasticSampler}{$D_\theta(\xx;\sigma), ~t_{i \in \{0, \dots, N\}}, ~\gamma_{i \in \{0, \dots, N-1\}}, ~\Snoise$}
    \AState{{\bf sample} $\xx_0 \sim \mathcal{N} \big( \boldzero, ~\odetime_0^2 ~\boldi \big)$}
    \AFor{$i \in \{0, \dots, N-1\}$}
        \AComment{$\gamma_i = \begin{cases}\scalebox{0.9}{${\min}\Big( \frac{\Schurn}{N}, \scalebox{0.8}{$\sqrt{2}{-}1$} \Big)$} & \text{if } \scalebox{0.9}{$t_i{\in}[\Stmin{,}\Stmax]$} \\ 0 & \text{otherwise}\end{cases}$}
      \AState{{\bf sample} $\boldsymbol{\epsilon}_i \sim \mathcal{N} \big( \boldzero, ~\Snoise^2 ~\boldi \big)$}
      \AState{$\hat\odetime_i \gets \odetime_i + \gamma_i \odetime_i$}
          \AComment{Select temporarily increased noise level $\hat\odetime_i$}
      \AState{$\xxh_i \gets \xx_i + \hsqrt{\hat\odetime_i^2 - \odetime_i^2} ~\boldsymbol{\epsilon}_i$}
          \AComment{Add new noise to move from $\odetime_i$ to $\hat\odetime_i$}
      \AState{$\dd_i \gets \big( \xxh_i - D_\theta(\xxh_i; \hat\odetime_i) \big) / \hat\odetime_i$}
          \AComment{Evaluate $\diff\xx / \diff\odetime$ at $\hat\odetime_i$}
      \AState{$\xx_{i+1} \gets \xxh_i + (\odetime_{i+1} - \hat\odetime_i) \dd_i$}
          \AComment{Take Euler step from $\hat\odetime_i$ to $\odetime_{i+1}$}
      \AIf{$t_{i+1} \ne 0$}
        \AState{$\ddp_i \gets \big( \xx_{i+1} - D_\theta(\xx_{i+1}; \odetime_{i+1}) \big) / \odetime_{i+1} $}
            \AComment{Apply 2\textsuperscript{nd} order correction}
        \AState{$\xx_{i+1} \gets \xxh_i + (\odetime_{i+1} - \hat\odetime_i) \big( \frac{1}{2}\dd_i + \frac{1}{2}\ddp_i \big)$}
      \EndIf
    \EndFor
    \AState{\textbf{return} $\xx_N$}
  \EndProcedure
\end{algorithmic}
\end{spacing}
\label{alg:stochastic}
\end{algorithm}
}

\newcommand{\figSdePlotNfe}{%
\renewcommand{\hh}{85mm}\renewcommand{\vv}{72mm}\renewcommand{\hhh}{-3mm}%
\begin{figure}[t]
\centering%
\resizebox{\textwidth}{!}{%
\hspace*{-1mm}%
\begin{tikzpicture}%
\begin{axis}[
  width=\hh, height=\vv,
  xmin={16}, xmax={2048}, xmode={log}, xtick={16, 32, 64, 128, 256, 512, 1024, 2048}, xticklabels={\tickNFE{16}, $32$, $64$, $128$, $256$, $512$, $1024$, $2048$},
  ymin={1.61}, ymax={3.35}, ymode={linear}, ytick={1.0, 1.2, 1.4, 1.6, 1.8, 2.0, 2.2, 2.4, 2.6, 2.8, 3.0, 3.2, 3.35}, yticklabels={$1.0$, $1.2$, $1.4$, $1.6$, $1.8$, $2.0$, $2.2$, $2.4$, $2.6$, $2.8$, $3.0$, $3.2$, \tickFID},
  grid={major}, legend pos={south west}, legend cell align={left}, legend columns={2}, legend style={font=\small, /tikz/every even column/.append style={column sep=1.6mm}},
]
\fillbetween[C3, opacity=0.15, forget plot]{coordinates {\genSdePlotNfeCifaruDdpmppOdeLo}}{coordinates {\genSdePlotNfeCifaruDdpmppOdeHi}};
\fillbetween[C1, opacity=0.15, forget plot]{coordinates {\genSdePlotNfeCifaruDdpmppLambdaLo}}{coordinates {\genSdePlotNfeCifaruDdpmppLambdaHi}};
\fillbetween[C2, opacity=0.15, forget plot]{coordinates {\genSdePlotNfeCifaruDdpmppPlainLo}}{coordinates {\genSdePlotNfeCifaruDdpmppPlainHi}};
\fillbetween[C4, opacity=0.15, forget plot]{coordinates {\genSdePlotNfeCifaruDdpmppOursLo}}{coordinates {\genSdePlotNfeCifaruDdpmppOursHi}};
\fillbetween[C0, opacity=0.15, forget plot]{coordinates {\genSdePlotNfeCifaruDdpmppRangeLo}}{coordinates {\genSdePlotNfeCifaruDdpmppRangeHi}};
\fillbetween[black, opacity=0.15, forget plot]{coordinates {\genSdePlotNfeCifaruDdpmppReproLo}}{coordinates {\genSdePlotNfeCifaruDdpmppReproHi}};
\addplot[C3]                             coordinates {\genSdePlotNfeCifaruDdpmppOde};
\addplot[C1]                             coordinates {\genSdePlotNfeCifaruDdpmppLambda};
\addplot[C2]                             coordinates {\genSdePlotNfeCifaruDdpmppPlain};
\addplot[C4]                             coordinates {\genSdePlotNfeCifaruDdpmppOurs};
\addplot[C0]                             coordinates {\genSdePlotNfeCifaruDdpmppRange};
\addplot[black, dashed]                  coordinates {\genSdePlotNfeCifaruDdpmppRepro};
\addplot[black, line width={1.7pt}, line cap={round}, dash pattern={on 0pt off 3pt}] coordinates {\genSdePlotNfeCifaruDdpmppGotta};
\addplot[C3, only marks, forget plot]    coordinates {\genSdePlotNfeCifaruDdpmppOdeMarks};
\addplot[C2, only marks, forget plot]    coordinates {\genSdePlotNfeCifaruDdpmppPlainMarks};
\addplot[C1, only marks, forget plot]    coordinates {\genSdePlotNfeCifaruDdpmppLambdaMarks};
\addplot[C0, only marks, forget plot]    coordinates {\genSdePlotNfeCifaruDdpmppRangeMarks};
\addplot[C4, only marks, forget plot]    coordinates {\genSdePlotNfeCifaruDdpmppOursMarks};
\addplot[black, only marks, forget plot] coordinates {\genSdePlotNfeCifaruDdpmppReproMarks};
\addplot[black, only marks, forget plot] coordinates {\genSdePlotNfeCifaruDdpmppGottaMarks};
\node at (axis cs:511,2.3) [anchor={south}] {\textcolor{C4}{$2.27$}};
\legend{
  {Deterministic},
  {$\StminStmax = [0, \infty]$},
  {$\StminStmax$ + $\Snoise = 1$},
  {Optimal settings},
  {$\Snoise = 1$},
  {Original sampler},
  {\makebox[0mm][l]{Jolicoeur-Martineau~et~al.~\cite{Jolicoeur2021}}},
}
\end{axis}
\end{tikzpicture}
\hspace*{\hhh}
\begin{tikzpicture}
\begin{axis}[
  width=\hh, height=\vv,
  xmin={16}, xmax={2048}, xmode={log}, xtick={16, 32, 64, 128, 256, 512, 1024, 2048}, xticklabels={$16$, $32$, $64$, $128$, $256$, $512$, $1024$, $2048$},
  ymin={1.8}, ymax={4.8}, ymode={linear}, ytick={1.0, 1.5, 2.0, 2.5, 3.0, 3.5, 4.0, 4.5, 4.8}, yticklabels={$1.0$, $1.5$, $2.0$, $2.5$, $3.0$, $3.5$, $4.0$, $4.5$, \tickFID},
  grid={major}, legend pos={north east}, legend cell align={left}, legend style={font=\normalsize},
]
\fillbetween[C3, opacity=0.15, forget plot]{coordinates {\genSdePlotNfeCifaruNcsnppOdeLo}}{coordinates {\genSdePlotNfeCifaruNcsnppOdeHi}};
\fillbetween[C1, opacity=0.15, forget plot]{coordinates {\genSdePlotNfeCifaruNcsnppLambdaLo}}{coordinates {\genSdePlotNfeCifaruNcsnppLambdaHi}};
\fillbetween[C2, opacity=0.15, forget plot]{coordinates {\genSdePlotNfeCifaruNcsnppPlainLo}}{coordinates {\genSdePlotNfeCifaruNcsnppPlainHi}};
\fillbetween[C4, opacity=0.15, forget plot]{coordinates {\genSdePlotNfeCifaruNcsnppOursLo}}{coordinates {\genSdePlotNfeCifaruNcsnppOursHi}};
\fillbetween[C0, opacity=0.15, forget plot]{coordinates {\genSdePlotNfeCifaruNcsnppRangeLo}}{coordinates {\genSdePlotNfeCifaruNcsnppRangeHi}};
\fillbetween[black, opacity=0.15, forget plot]{coordinates {\genSdePlotNfeCifaruNcsnppReproLo}}{coordinates {\genSdePlotNfeCifaruNcsnppReproHi}};
\addplot[C3]                             coordinates {\genSdePlotNfeCifaruNcsnppOde};
\addplot[C1]                             coordinates {\genSdePlotNfeCifaruNcsnppLambda};
\addplot[C2]                             coordinates {\genSdePlotNfeCifaruNcsnppPlain};
\addplot[C4]                             coordinates {\genSdePlotNfeCifaruNcsnppOurs};
\addplot[C0]                             coordinates {\genSdePlotNfeCifaruNcsnppRange};
\addplot[black, dashed]                  coordinates {\genSdePlotNfeCifaruNcsnppRepro};
\addplot[black, line width={1.7pt}, line cap={round}, dash pattern={on 0pt off 3pt}] coordinates {\genSdePlotNfeCifaruNcsnppGotta};
\addplot[C3, only marks, forget plot]    coordinates {\genSdePlotNfeCifaruNcsnppOdeMarks};
\addplot[C2, only marks, forget plot]    coordinates {\genSdePlotNfeCifaruNcsnppPlainMarks};
\addplot[C1, only marks, forget plot]    coordinates {\genSdePlotNfeCifaruNcsnppLambdaMarks};
\addplot[C0, only marks, forget plot]    coordinates {\genSdePlotNfeCifaruNcsnppRangeMarks};
\addplot[C4, only marks, forget plot]    coordinates {\genSdePlotNfeCifaruNcsnppOursMarks};
\addplot[black, only marks, forget plot] coordinates {\genSdePlotNfeCifaruNcsnppReproMarks};
\addplot[black, only marks, forget plot] coordinates {\genSdePlotNfeCifaruNcsnppGottaMarks};
\node at (axis cs:1560,2.22) [anchor={north}] {\textcolor{C4}{$2.23$}};
\end{axis}
\end{tikzpicture}
\hspace*{\hhh}
\begin{tikzpicture}
\begin{axis}[
  width=\hh, height=\vv,
  xmin={16}, xmax={2048}, xmode={log}, xtick={16, 32, 64, 128, 256, 512, 1024, 2048}, xticklabels={$16$, $32$, $64$, $128$, $256$, $512$, $1024$, $2048$},
  ymin={1.3}, ymax={3.2}, ymode={linear}, ytick={1.0, 1.2, 1.4, 1.6, 1.8, 2.0, 2.2, 2.4, 2.6, 2.8, 3.0, 3.2}, yticklabels={$1.0$, $1.2$, $1.4$, $1.6$, $1.8$, $2.0$, $2.2$, $2.4$, $2.6$, $2.8$, $3.0$, \tickFID},
  grid={major}, legend pos={north east}, legend cell align={left}, legend style={font=\normalsize},
]
\fillbetween[C3, opacity=0.15, forget plot]{coordinates {\genSdePlotNfeImgcDhariwalOdeLo}}{coordinates {\genSdePlotNfeImgcDhariwalOdeHi}};
\fillbetween[C1, opacity=0.15, forget plot]{coordinates {\genSdePlotNfeImgcDhariwalLambdaLo}}{coordinates {\genSdePlotNfeImgcDhariwalLambdaHi}};
\fillbetween[C2, opacity=0.15, forget plot]{coordinates {\genSdePlotNfeImgcDhariwalPlainLo}}{coordinates {\genSdePlotNfeImgcDhariwalPlainHi}};
\fillbetween[C4, opacity=0.15, forget plot]{coordinates {\genSdePlotNfeImgcDhariwalOursLo}}{coordinates {\genSdePlotNfeImgcDhariwalOursHi}};
\fillbetween[C0, opacity=0.15, forget plot]{coordinates {\genSdePlotNfeImgcDhariwalRangeLo}}{coordinates {\genSdePlotNfeImgcDhariwalRangeHi}};
\fillbetween[black, opacity=0.15, forget plot]{coordinates {\genSdePlotNfeImgcDhariwalReproLo}}{coordinates {\genSdePlotNfeImgcDhariwalReproHi}};
\addplot[C3]                             coordinates {\genSdePlotNfeImgcDhariwalOde};
\addplot[C1]                             coordinates {\genSdePlotNfeImgcDhariwalLambda};
\addplot[C2]                             coordinates {\genSdePlotNfeImgcDhariwalPlain};
\addplot[C4]                             coordinates {\genSdePlotNfeImgcDhariwalOurs};
\addplot[C0]                             coordinates {\genSdePlotNfeImgcDhariwalRange};
\addplot[black, dashed]                  coordinates {\genSdePlotNfeImgcDhariwalRepro};
\addplot[C3, only marks, forget plot]    coordinates {\genSdePlotNfeImgcDhariwalOdeMarks};
\addplot[C2, only marks, forget plot]    coordinates {\genSdePlotNfeImgcDhariwalPlainMarks};
\addplot[C1, only marks, forget plot]    coordinates {\genSdePlotNfeImgcDhariwalLambdaMarks};
\addplot[C0, only marks, forget plot]    coordinates {\genSdePlotNfeImgcDhariwalRangeMarks};
\addplot[C4, only marks, forget plot]    coordinates {\genSdePlotNfeImgcDhariwalOursMarks};
\addplot[black, only marks, forget plot] coordinates {\genSdePlotNfeImgcDhariwalReproMarks};
\node at (axis cs:1023,1.54) [anchor={north}] {\textcolor{C4}{$1.55$}};
\end{axis}
\end{tikzpicture}
}\\\hfill%
\makebox[0.33\linewidth]{\footnotesize (a) Uncond. CIFAR-10, VP}\hfill%
\makebox[0.33\linewidth]{\footnotesize (b) Uncond. CIFAR-10, VE}\hfill%
\makebox[0.33\linewidth]{\footnotesize (c) Class-cond. ImageNet-64}\hfill%
\caption{\label{fig:SdePlotNfe}%
Evaluation of our stochastic sampler (Algorithm~\ref{alg:stochastic}).
The purple curve corresponds to optimal choices for $\{\Schurn, \Stmin, \Stmax, \Snoise\}$; orange, blue, and green correspond to disabling the effects of $\StminStmax$ and/or $\Snoise$.
The red curves show reference results for our deterministic sampler (Algorithm~\ref{alg:heun}), equivalent to setting $\Schurn = 0$.
The dashed black curves correspond to the original stochastic samplers from previous work: Euler--Maruyama~\cite{Song2021sde} for VP, predictor-corrector~\cite{Song2021sde} for VE, and iDDPM~\cite{Nichol2021a} for ImageNet-64.
The dots indicate lowest observed FID.
}
\vspace*{-2mm}
\end{figure}
}

\newcommand{\tabTrainingTable}{%
\forcenewcolumntype{x}{>{\centering\arraybackslash\hspace{0pt}}p{9mm}}%
\forcenewcolumntype{y}{>{\centering\arraybackslash\hspace{0pt}}p{13mm}}%
\newcommand{\smast}{\smash{\rlap{$^*$}}}%
\renewcommand{\cs}{\hspace{0.5mm}}%
\tabulinesep=0.7mm%
\tabulinestyle{0.17mm}%
\begin{table}[t]%
\centering%
\footnotesize%
\caption{\label{tab:TrainingTable}%
Evaluation of our training improvements.
The starting point (config \textsc{a}) is VP \& VE using our \textbf{deterministic} sampler.
At the end (configs \textsc{e,f}), VP \& VE only differ in the architecture of $F_\theta$.
}
\vspace{2mm}%
\resizebox{\textwidth}{!}{%
\begin{tabu}{|@{\ \ }l@{\ \ }|x@{\cs}x|x@{\cs}x|y@{\cs}y|y@{\cs}y|}
\tabucline{2-}
\genTrainingTable
\tabucline{-}
\end{tabu}%
}%
\vspace*{-2mm}%
\end{table}%
}

\newcommand{\figTrainingPlots}{%
\renewcommand{\hh}{85mm}\renewcommand{\vv}{68mm}\renewcommand{\hhh}{-1mm}%
\begin{figure}[t]
\centering%
\resizebox{\textwidth}{!}{%
\begin{tikzpicture}
\begin{axis}[
  width=\hh, height=\vv,
  xmin={0.002}, xmax={100}, xmode={log}, xtick={0.005, 0.02, 0.1, 0.5, 1, 2, 5, 10, 20, 50}, xticklabels={\tickSigma{0.005}, $0.02$, $0.1$, $0.5$, $1$, $2$, $5$, $10$, $20$, $50$},
  ymin={-0.1}, ymax={1.6}, ymode={linear}, ytick={0.0, 0.2, 0.4, 0.6, 0.8, 1.0, 1.2, 1.4, 1.6}, yticklabels={$0.0$, $0.2$, $0.4$, $0.6$, $0.8$, $1.0$, $1.2$, $1.4$, \tickLoss},
  grid={major}, legend pos={north west}, legend cell align={left}, legend columns={2}, legend style={font=\normalsize, /tikz/every even column/.append style={column sep=1.6mm}},
]
\fillbetween[C2, opacity=0.15, forget plot]{coordinates {\genTrainingPlotLossCifaruDdpmppInitLo}}{coordinates {\genTrainingPlotLossCifaruDdpmppInitHi}};
\fillbetween[C0, opacity=0.15, forget plot]{coordinates {\genTrainingPlotLossCifaruDdpmppPrecLo}}{coordinates {\genTrainingPlotLossCifaruDdpmppPrecHi}};
\fillbetween[C1, opacity=0.15, forget plot]{coordinates {\genTrainingPlotLossFfhqDdpmppPrecLo}}{coordinates {\genTrainingPlotLossFfhqDdpmppPrecHi}};
\addplot[C2] coordinates {\genTrainingPlotLossCifaruDdpmppInit};
\addplot[C0] coordinates {\genTrainingPlotLossCifaruDdpmppPrec};
\addplot[C3, dashed, domain=0.002:100, samples=256] {exp(-0.5 * ((ln(x) - (-1.2)) / (1.2))^2)};
\addplot[C1] coordinates {\genTrainingPlotLossFfhqDdpmppPrec};
\legend{
  {Loss after init},
  {CIFAR-10},
  {Distribution of $\sigma$},
  {FFHQ-64},
}
\end{axis}
\end{tikzpicture}
\hspace*{\hhh}
\begin{tikzpicture}
\begin{axis}[
  width=\hh, height=\vv,
  xmin={0}, xmax={100}, xmode={linear}, xtick={0, 10, 20, 30, 40, 50, 60, 70, 80, 90, 100}, xticklabels={\tickSchurnB{0}, $10$, $20$, $30$, $40$, $50$, $60$, $70$, $80$, $90$, $100$},
  ymin={1.8}, ymax={4.3}, ymode={linear}, ytick={1.0, 1.5, 2.0, 2.5, 3.0, 3.5, 4.0, 4.3}, yticklabels={$1.0$, $1.5$, $2.0$, $2.5$, $3.0$, $3.5$, $4.0$, \tickFID},
  grid={major}, legend pos={north east}, legend cell align={left}, legend columns={2}, legend style={font=\normalsize, /tikz/every even column/.append style={column sep=1.6mm}},
]
\fillbetween[C0, opacity=0.15, forget plot]{coordinates {\genTrainingPlotChurnCifaruDdpmppOrigLo}}{coordinates {\genTrainingPlotChurnCifaruDdpmppOrigHi}};
\fillbetween[C1, opacity=0.15, forget plot]{coordinates {\genTrainingPlotChurnCifaruDdpmppAugmLo}}{coordinates {\genTrainingPlotChurnCifaruDdpmppAugmHi}};
\fillbetween[C2, opacity=0.15, forget plot]{coordinates {\genTrainingPlotChurnCifaruNcsnppOrigLo}}{coordinates {\genTrainingPlotChurnCifaruNcsnppOrigHi}};
\fillbetween[C3, opacity=0.15, forget plot]{coordinates {\genTrainingPlotChurnCifaruNcsnppAugmLo}}{coordinates {\genTrainingPlotChurnCifaruNcsnppAugmHi}};
\addplot[C0] coordinates {\genTrainingPlotChurnCifaruDdpmppOrig};
\addplot[C1] coordinates {\genTrainingPlotChurnCifaruDdpmppAugm};
\addplot[C2] coordinates {\genTrainingPlotChurnCifaruNcsnppOrig};
\addplot[C3] coordinates {\genTrainingPlotChurnCifaruNcsnppAugm};
\legend{
  {VP, original},
  {VP, our model},
  {VE, original},
  {VE, our model},
}
\end{axis}
\end{tikzpicture}
\hspace*{\hhh}
\begin{tikzpicture}
\begin{axis}[
  width=\hh, height=\vv,
  xmin={0}, xmax={100}, xmode={linear}, xtick={0, 10, 20, 30, 40, 50, 60, 70, 80, 90, 100}, xticklabels={\tickSchurnB{0}, $10$, $20$, $30$, $40$, $50$, $60$, $70$, $80$, $90$, $100$},
  ymin={1.1}, ymax={2.8}, ymode={linear}, ytick={1.2, 1.4, 1.6, 1.8, 2.0, 2.2, 2.4, 2.6, 2.8}, yticklabels={$1.2$, $1.4$, $1.6$, $1.8$, $2.0$, $2.2$, $2.4$, $2.6$, \tickFID},
  grid={major}, legend pos={north east}, legend cell align={left}, legend columns={-1}, legend style={font=\normalsize, /tikz/every even column/.append style={column sep=1.6mm}},
]
\fillbetween[C0, opacity=0.15, forget plot]{coordinates {\genTrainingPlotImgcOrigLo}}{coordinates {\genTrainingPlotImgcOrigHi}};
\fillbetween[C1, opacity=0.15, forget plot]{coordinates {\genTrainingPlotImgcOursLo}}{coordinates {\genTrainingPlotImgcOursHi}};
\addplot[C0] coordinates {\genTrainingPlotImgcOrig};
\addplot[C1] coordinates {\genTrainingPlotImgcOurs};
\addplot[C0, only marks, forget plot] coordinates {(75, 1.57)}; \node at (axis cs:75,1.57) [anchor={north}] {\textcolor{C0}{$1.57$}};
\addplot[C1, only marks, forget plot] coordinates {(40, 1.36)}; \node at (axis cs:40,1.36) [anchor={north}] {\textcolor{C1}{$1.36$}};
\addplot[C0, only marks, forget plot] coordinates {(0, 2.66)}; \node at (axis cs:0,2.66) [anchor={west}] {\textcolor{C0}{$2.66$}};
\addplot[C1, only marks, forget plot] coordinates {(0, 2.22)}; \node at (axis cs:0,2.22) [anchor={west}] {\textcolor{C1}{$2.22$}};
\legend{
  {Original},
  {Our model},
}
\end{axis}
\end{tikzpicture}
}\\\hfill%
\makebox[0.33\linewidth]{\footnotesize (a) Loss \& noise distribution}\hfill%
\makebox[0.33\linewidth]{\footnotesize (b) Stochasticity on CIFAR-10}\hfill%
\makebox[0.33\linewidth]{\footnotesize{(c) Stochasticity on ImageNet-64}}\hfill%
\caption{\label{fig:TrainingPlots}%
\textbf{(a)}
Observed initial (green) and final loss per noise level, representative of the the 32$\times$32 (blue) and 64$\times$64 (orange) models considered in this paper. The shaded regions represent the standard deviation over 10k random samples. Our proposed training sample density is shown by the dashed red curve.
\textbf{(b)}
{Effect of $\Schurn$ on unconditional CIFAR-10 with 256 steps (NFE $=$ 511).}
For the original training setup of Song~et~al.~\cite{Song2021sde}, stochastic sampling is highly beneficial (blue, green), while deterministic sampling ($\Schurn = 0$) leads to relatively poor FID.
For our training setup, the situation is reversed (orange, red); stochastic sampling is not only unnecessary but harmful.
{\textbf{(c)}
Effect of $\Schurn$ on class-conditional ImageNet-64 with 256 steps (NFE $=$ 511).
In this more challenging scenario, stochastic sampling turns out to be useful again.
Our training setup improves the results for both deterministic and stochastic sampling.
}
}
\vspace*{-2mm}
\end{figure}
}

\title{Elucidating the Design Space of Diffusion-Based Generative Models}

\author{
  \ifemail
    Tero Karras   \\ NVIDIA \\ \texttt{tkarras@nvidia.com}  \And
    Miika Aittala \\ NVIDIA \\ \texttt{maittala@nvidia.com} \AND
    Timo Aila     \\ NVIDIA \\ \texttt{taila@nvidia.com}    \And
    Samuli Laine  \\ NVIDIA \\ \texttt{slaine@nvidia.com}
  \else
    Tero Karras   \\ NVIDIA \And
    Miika Aittala \\ NVIDIA \And
    Timo Aila     \\ NVIDIA \And
    Samuli Laine  \\ NVIDIA
  \fi
}

\begin{document}
\maketitle
\confnotice

\begin{abstract}
We argue that the theory and practice of diffusion-based generative models are currently unnecessarily convoluted and seek to remedy the situation by presenting a design space that clearly separates the concrete design choices.
This lets us identify several changes to both the sampling and training processes, as well as preconditioning of the score networks.
Together, our improvements yield new state-of-the-art FID of 1.79 for CIFAR-10 in a class-conditional setting and 1.97 in an unconditional setting, with much faster sampling (35 network evaluations per image) than prior designs.
To further demonstrate their modular nature, we show that our design changes dramatically improve both the efficiency and quality obtainable with pre-trained score networks from previous work, including improving the FID of {a previously trained} ImageNet-64 model from 2.07 to near-SOTA 1.55,
{and after re-training with our proposed improvements to a new SOTA of 1.36.}
\end{abstract}

\section{Introduction}

Diffusion-based generative models~{\cite{SohlDickstein2015}} have emerged as a powerful new framework for neural image synthesis, in both unconditional \cite{Ho2020,Nichol2021a,Song2021sde} and conditional \cite{Ho2021cascaded,Nichol2021b,Nichol2021a,Preechakul2021diffusion,Ramesh2022,Rombach2021highresolution,Saharia2021,Song2021sde} settings,
  even surpassing the quality of GANs~\cite{Goodfellow2014} in certain situations~\cite{Dhariwal2021}.
They are also rapidly finding use in other domains such as audio~\cite{Kong2021,Popov2021} and video~\cite{Ho2022} generation, image segmentation~\cite{Baranchuk2022,Wolleb2022} and language translation~\cite{Nachmani2021}.
As such, there is great interest in applying these models and improving them further in terms of image/distribution quality, training cost, and generation speed.

The literature on these models is dense on theory, and %
derivations of sampling schedule, training dynamics, noise level parameterization, etc., tend to be based as directly as possible on theoretical frameworks, which ensures that the models are on a solid theoretical footing.
However, this approach has a danger of obscuring the available design space\,---\,a proposed model may appear as a tightly coupled package where no individual component can be modified without breaking the entire system.

As our first contribution, we take a look at the theory behind these models from a practical standpoint, focusing more on the ``tangible'' objects and algorithms that appear in the training and sampling phases, and less on the statistical processes from which they might be derived.
The goal is to obtain better insights into how these components are linked together and what degrees of freedom are available in the design of the overall system.
We focus on the broad class of models where a neural network is used to model the score~\cite{Hyvarinen05} of a noise level dependent marginal distribution of the training data corrupted by Gaussian noise.
Thus, our work is in the context of \emph{denoising score matching}~\cite{Vincent11}.

Our second set of contributions concerns the sampling processes used to synthesize images using diffusion models.
We identify the best-performing time discretization for sampling, apply a higher-order Runge--Kutta method for the sampling process, evaluate different sampler schedules, and analyze the usefulness of stochasticity in the sampling process.
The result of these improvements is a significant drop in the number of sampling steps required during synthesis, and the improved sampler can be used as a drop-in replacement with several widely used diffusions models \cite{Nichol2021a,Song2021sde}.

The third set of contributions focuses on the training of the score-modeling neural network.
While we continue to rely on the commonly used network architectures (DDPM~\cite{Ho2020}, NCSN~\cite{Song2019gradients}), we provide the first principled analysis of the preconditioning of the networks' inputs, outputs, and loss functions in a diffusion model setting and derive best practices for improving the training dynamics.
We also suggest an improved distribution of noise levels during training, and note that non-leaking augmentation~\cite{Karras2020ada}\,---\,typically used with GANs\,---\,is beneficial for diffusion models as well.

Taken together, our contributions enable significant improvements in result quality, e.g., leading to record FID{s} of 1.79 for CIFAR-10~\cite{Krizhevsky2009cifar} {and 1.36 for ImageNet~\cite{Deng2009imagenet} in 64$\times$64 resolution}.
With all key ingredients of the design space explicitly tabulated, we believe that our approach will allow easier innovation on the individual components, and thus enable more extensive and targeted exploration of the design space of diffusion models.
{Our implementation and pre-trained models are available at \url{https://github.com/NVlabs/edm}}

\vspace*{-1mm}
\section{Expressing diffusion models in a common framework}

\label{sec:samplingbackground}

Let us denote the data distribution by $\pdata(\xx)$, with standard deviation $\sdata$, and consider the family of mollified distributions $p(\xx; \sigma)$ obtained by adding i.i.d.\ Gaussian noise of standard deviation $\sigma$ to the data.
For $\smax\gg\sdata$, $p(\xx; \smax)$ is practically indistinguishable from pure Gaussian noise. %
The idea of diffusion models is to randomly sample a noise image $\xx_0 \sim \mathcal{N}(\boldzero, \smax^2 \boldi)$,
and sequentially denoise it into images $\xx_i$ with noise levels $\sigma_0 = \sigmamax > \sigma_1 > \dots > \sigma_N = 0$ so that
at each noise level $\xx_i \sim p(\xx_i; \sigma_i)$.
The endpoint $\xx_N$ of this process is thus distributed according to the data.

Song et al.~\cite{Song2021sde} present a stochastic differential equation (SDE) that maintains the desired distribution $p$ as sample $\xx$ evolves over time.
This allows the above process to be implemented using a stochastic solver that both removes and adds noise at each iteration.
They also give a corresponding ``probability flow'' ordinary differential equation (ODE) where the only source of randomness is the initial noise image $\xx_0$.
Contrary to the usual order of treatment, we begin by examining the ODE, as it offers a fruitful setting for analyzing sampling trajectories and their discretizations. The insights carry over to stochastic sampling, which we reintroduce as a generalization in Section~\ref{sec:stochasticity}. %

\figDenoising

\vparagraph{ODE formulation.}
A probability flow ODE \cite{Song2021sde} continuously increases or reduces noise level of the image when moving forward or backward in time, respectively.
To specify the ODE, we must first choose a schedule $\sigma(\odetime)$ that defines the desired noise level at time $\odetime$.
For example, setting \low{$\sigma(t)\propto\sqrt{t}$} is mathematically natural, as it corresponds to constant-speed heat diffusion~\cite{Fourier1822}. %
However, we will show in Section~\ref{sec:deterministic} that the choice of schedule has major practical implications and should not be made on the basis of theoretical convenience.

The defining characteristic of the probability flow ODE is that evolving a sample \smash{$\xx_a \sim p \big( \xx_a; \sigma(t_a) \big)$} from time $t_a$ to $t_b$ (either forward or backward in time) yields a sample \smash{$\xx_b \sim p \big( \xx_b; \sigma(t_b) \big)$}.
Following previous work~\cite{Song2021sde}, this requirement is satisfied ({see Appendix~\ref{app:originalode} and~\ref{app:ourode}}) by
\begin{equation}
\label{eq:ode}
\diff \xx = -\dot\sigma(t) ~\sigma(t) ~\nablaxx \log p \big( \xx; \sigma(t) \big) ~\diff\odetime \text{,}
\end{equation}
where the dot denotes a time derivative.
$\nablaxx \log p(\xx; \sigma)$ is the \emph{score function}~\cite{Hyvarinen05}, a vector field that points towards higher density of data at a given noise level.
Intuitively, an infinitesimal forward step of this ODE nudges the sample away from the data, at a rate that depends on the change in noise level.
Equivalently, a backward step nudges the sample towards the data distribution.

\vparagraph{Denoising score matching.}
The score function has the remarkable property that it does not depend on the generally intractable normalization constant of the underlying density function $p(\xx; \sigma)$~\cite{Hyvarinen05}, and thus can be much easier to evaluate.
Specifically, if $D(\xx;\sigma)$ is a denoiser function that minimizes the expected $L_2$ denoising error for samples drawn from $\pdata$ separately for every $\sigma$, i.e.,
\newtagform{emptyTag}{}{}
\begin{equation*}
\mathbb{E}_{\signal \sim \pdata} \mathbb{E}_{\noise \sim \mathcal{N}(\boldzero, \sigma^2 \boldi)} \lVert D(\signal + \noise; \sigma) - \signal \rVert^2_2
\text{,}\hspace*{2mm}\text{then}\hspace*{2mm}
\nablaxx \log p(\xx; \sigma) = \big( D(\xx; \sigma) - \xx \big) / \sigma^2 \text{,}
\tag{
{\begin{minipage}[b][0pt][b]{.55em}\usetagform{emptyTag}\begin{equation}\label{eq:score}\end{equation}\usetagform{default}\end{minipage}},\,%
{\begin{minipage}[b][0pt][b]{.50em}\usetagform{emptyTag}\begin{equation}\label{eq:scoredenoiser}\end{equation}\usetagform{default}\end{minipage}}
}
\end{equation*}
where $\signal$ is a training image and $\noise$ is noise.
In this light, the score function isolates the noise component from the signal in $\xx$, and Eq.~\ref{eq:ode} amplifies (or diminishes) it over time.
Figure~\ref{fig:denoising} illustrates the behavior of ideal $D$ in practice.
The key observation in diffusion models is that $D(\xx;\sigma)$ can be implemented as a neural network $D_\theta(\xx;\sigma)$ trained according to Eq.~\ref{eq:score}.
Note that $D_\theta$ may include additional pre- and post-processing steps, such as scaling $\xx$ to an appropriate dynamic range; we will return to such \emph{preconditioning} in Section~\ref{sec:training}.

\tabSpecifics

\vparagraph{Time-dependent signal scaling.}
Some methods (see Appendix~{\ref{app:reframingvp}}) introduce an additional scale schedule $s(t)$ and consider $\xx = s(t) \hat\xx$ to be a scaled version of the original, non-scaled variable~$\hat\xx$.
This changes the time-dependent probability density, and consequently also the ODE solution trajectories.
The resulting ODE is a generalization of Eq.~\ref{eq:ode}:
\begin{equation}
\label{eq:odescale}
\diff \xx = \left[ \frac{\dot \scale(\odetime)}{\scale(\odetime)} ~\xx -\scale(\odetime)^2 ~\dot\sigma(\odetime) ~\sigma(\odetime) ~\nablaxx \log p\left(\frac{\xx}{\scale(\odetime)}; \sigma(\odetime)\right) \right] ~\diff\odetime \text{.}
\end{equation}
Note that we explicitly undo the scaling of $\xx$ when evaluating the score function to keep the definition of $p(\xx; \sigma)$ independent of $s(t)$.

\vparagraph{Solution by discretization.}
The ODE to be solved is obtained by substituting Eq.~\ref{eq:scoredenoiser} into Eq.~\ref{eq:odescale} to define the point-wise gradient,
  and the solution can be found by numerical integration, i.e., taking finite steps over discrete time intervals.
This requires choosing both the integration scheme (e.g., Euler or a variant of Runge--Kutta), as well as the discrete sampling times $\{t_0, t_1, \dots, t_N\}$.
Many prior works rely on Euler's method, but we show in Section~\ref{sec:deterministic} that a 2\textsuperscript{nd} order solver offers a better computational tradeoff.
For brevity, we do not provide a separate pseudocode for Euler's method applied to our ODE here, but it can be extracted from Algorithm~\ref{alg:heun} by omitting lines 6--8.

\vparagraph{Putting it together.}
Table~\ref{tab:specifics} presents formulas for reproducing deterministic variants of three earlier methods in our framework.
These methods were chosen because they are widely used and achieve state-of-the-art performance, but also because they were derived from different theoretical foundations.
Some of our formulas appear quite different from the original papers as indirection and recursion have been removed; see Appendix~\ref{app:reframing} for details.
The main purpose of this reframing is to bring into light all the independent components that often appear tangled together in previous work.
In our framework, there are no implicit dependencies between the components\,---\,any choices (within reason) for the individual formulas will, in principle, lead to a functioning model.
{In other words, changing one component does not necessitate changes elsewhere in order to, e.g., maintain the property that the model converges to the data in the limit.
In practice, some choices and combinations will of course work better than others.}

\vspace*{-1mm}
\section{Improvements to deterministic sampling}
\vspace*{-.5mm}
\label{sec:deterministic}

{Improving the output quality and/or decreasing the computational cost of sampling are common topics in diffusion model research (e.g.,~\cite{Dockhorn2022damped,Jolicoeur2021,Liu2022pseudo,Lu2022dpm,Luhman2021speed,Nichol2021a,Salimans2022,Vahdat2021,Watson2022fastsample,Watson2021efficientsample,Zhang2022exp}).}
Our hypothesis is that the choices related to the sampling process are largely independent of the other components, such as network architecture and training details.
In other words, the training procedure of $D_\theta$ should not dictate $\sigma(t)$, $s(t)$, and $\{t_i\}$, nor vice versa; from the viewpoint of the sampler, $D_\theta$ is simply a black box~\cite{Watson2022fastsample,Watson2021efficientsample}.
We test this by evaluating different samplers on three \emph{pre-trained} models, each representing a different theoretical framework and model family.
We first measure baseline results for these models using their original sampler implementations, and then bring these samplers into our unified framework using the formulas in Table~\ref{tab:specifics}, followed by our improvements.
{This allows us to evaluate different practical choices and propose general improvements to the sampling process that are applicable to all models.}

We evaluate the ``\ddpmpp{} cont. (VP)'' and ``\ncsnpp{} cont. (VE)'' models by Song~et~al.~\cite{Song2021sde} trained on unconditional CIFAR-10~\cite{Krizhevsky2009cifar} at 32$\times$32, corresponding to the variance preserving (VP) and variance exploding (VE) formulations~\cite{Song2021sde}, originally inspired by DDPM~\cite{Ho2020} and SMLD~\cite{Song2019gradients}.
We also evaluate the ``ADM (dropout)'' model by Dhariwal~and~Nichol~\cite{Dhariwal2021} trained on class-conditional ImageNet \cite{Deng2009imagenet} at 64$\times$64, corresponding to the improved DDPM (iDDPM) formulation~\cite{Nichol2021a}.
This model was trained using a {discrete} set of $\origN=1000$ noise levels.
\mbox{Further details are given in Appendix~\ref{app:reframing}.} %

\figOdePlotNfe

We evaluate the result quality in terms of Fr\'echet inception distance (FID) \cite{Heusel2017} computed between 50,000 generated images and all available real images.
Figure~\ref{fig:OdePlotNfe} shows FID as a function of neural function evaluations (NFE), i.e., how many times $D_\theta$ is evaluated to produce a single image.
Given that the sampling process is dominated entirely by the cost of $D_\theta$, improvements in NFE translate directly to sampling speed.
The original deterministic samplers are shown in blue, and the reimplementations of these methods in our unified framework (orange) yield similar but consistently better results.
The differences are explained by certain oversights in the original implementations as well as our more careful treatment of discrete noise levels in the case of DDIM; see Appendix~\ref{app:reframing}.
Note that our reimplementations are fully specified by Algorithm~\ref{alg:heun} and Table~\ref{tab:specifics}, even though the original codebases are structured very differently from each other.

\vparagraph{Discretization and higher-order integrators.}
Solving an ODE numerically is necessarily an approximation of following the true solution trajectory.
At each step, the solver introduces \emph{truncation error} that accumulates over the course of $N$ steps.
The local error generally scales superlinearly with respect to step size, and thus increasing $N$ improves the accuracy of the solution.

The commonly used Euler's method is a first order ODE solver with $\mathcal{O}(h^2)$ local error with respect to step size $h$.
Higher-order Runge--Kutta methods~\cite{Suli2003} scale more favorably but require multiple evaluations of $D_\theta$ per step.
{Linear multistep methods have also been recently proposed for sampling diffusion models~\cite{Liu2022pseudo,Zhang2022exp}.}
Through extensive tests, we have found Heun's 2\textsuperscript{nd} order method~\cite{Ascher1998} (a.k.a.~improved Euler, trapezoidal rule)\,---\,previously explored in the context of diffusion models by Jolicoeur-Martineau~et~al.~\cite{Jolicoeur2021}\,---\,to provide an excellent tradeoff between truncation error and NFE.
As illustrated in Algorithm~\ref{alg:heun}, it introduces an additional correction step for $\xx_{i+1}$ to account for change in $\diff\xx / \diff\odetime$ between $\odetime_i$ and $\odetime_{i+1}$. %
This correction leads to $\mathcal{O}(h^3)$ local error at the cost of one additional evaluation of $D_\theta$ per step.
Note that stepping to $\sigma=0$ would result in a division by zero, so we revert to Euler's method in this case.
\mbox{We discuss the general family of 2\textsuperscript{nd} order solvers in Appendix~{\ref{app:alpha}}.} %

\algHeun

The time steps $\{t_i\}$ determine how the step sizes and thus truncation errors are distributed between different noise levels.
We provide a detailed analysis in Appendix~{\ref{app:truncationerror}}, concluding that the step size should decrease monotonically with decreasing $\sigma$ and it does not need to vary on a per-sample basis.
We adopt a parameterized scheme where the time steps are defined according to a sequence of noise levels $\{\sigma_i\}$, i.e., $t_i=\sigma^{-1}(\sigma_i)$.
We set $\sigma_{i<N} = (Ai + B)^\rho$ and select the constants $A$ and $B$ so that $\sigma_0 = \smax$ and $\sigma_{N-1} = \smin$, which gives
\begin{equation}
\label{eq:discretization}
\sigma_{i<N} = \big( {\smax}^\frac{1}{\rho} + {\textstyle\frac{i}{N-1}} ( {\smin}^\frac{1}{\rho} - {\smax}^\frac{1}{\rho} ) \big)^\rho \hspace*{3mm}\text{and}\hspace*{3mm}\sigma_N = 0 \text{.}
\end{equation}
Here $\rho$ controls how much the steps near $\smin$ are shortened at the expense of longer steps near $\smax$.
Our analysis in Appendix~{\ref{app:truncationerror}} shows that setting $\rho=3$ nearly equalizes the truncation error at each step,
  but that $\rho$ in range of 5 to 10 performs much better for sampling images.
This suggests that errors near $\smin$ have a large impact.
We set $\rho=7$ for the remainder of this paper.

Results for Heun's method and Eq.~\ref{eq:discretization} are shown as the green curves in Figure~\ref{fig:OdePlotNfe}. %
We observe consistent improvement in all cases: Heun's method reaches the same FID as Euler's method with considerably lower NFE.

\ifemail
  \figTrajectories
\fi

\vparagraph{Trajectory curvature and noise schedule.}
The shape of the ODE solution trajectories is defined by functions $\sigma(t)$ and $s(t)$.
The choice of these functions offers a way to reduce the truncation errors discussed above, as their magnitude can be expected to scale proportional to the curvature of $\diff\xx / \diff\odetime$.
We argue that the best choice for these functions is $\sigma(t)=t$ and $s(t)=1$, which is also the choice made in DDIM~\cite{Song2020ddim}.
With this choice, the ODE of Eq.~\ref{eq:odescale} simplifies to $\diff\xx / \diff\odetime = \big( \xx - D(\xx; \odetime) \big) / \odetime$
{and $\sigma$ and $t$ become interchangeable}.

An immediate consequence is that at any $\xx$ and $t$, a single Euler step to $t=0$ yields the denoised image $D_\theta(\xx; t)$.
The tangent of the solution trajectory therefore always points towards the denoiser output.
This can be expected to change only slowly with the noise level, which corresponds to largely linear solution trajectories.
The 1D ODE sketch of Figure~\ref{fig:trajectories}c supports this intuition; the solution trajectories approach linear at both large and small noise levels, and have substantial curvature in only a small region in between.
The same effect can be seen with real data in Figure~\ref{fig:denoising}b, where the change between different denoiser targets occurs in a relatively narrow $\sigma$ range.
With the advocated schedule, this corresponds to high ODE curvature being limited to this same range.

The effect of setting $\sigma(t)=t$ and $s(t)=1$ is shown as the red curves in Figure~\ref{fig:OdePlotNfe}.
As DDIM already employs these same choices, the red curve is identical to the green one for ImageNet-64.
However, VP and VE benefit considerably from switching away from their original schedules.

\ifemail
\else
  \figTrajectories
\fi

\vparagraph{Discussion.}
The choices that we made in this section to improve deterministic sampling are summarized in the \emph{Sampling} part of Table~\ref{tab:specifics}.
Together, they reduce the NFE needed to reach high-quality results by a large factor: 7.3$\times$ for VP, 300$\times$ for VE, and 3.2$\times$ for DDIM, corresponding to the highlighted NFE values in Figure~\ref{fig:OdePlotNfe}.
In practice, we can generate 26.3 high-quality CIFAR-10 images per second on a single NVIDIA V100. %
The consistency of improvements corroborates our hypothesis that the sampling process is orthogonal to how each model was originally trained.
As further validation, 
we show results for the adaptive RK45 method~\cite{Dormand1980} using our schedule as the dashed black curves in Figure~\ref{fig:OdePlotNfe}; the cost of this sophisticated ODE solver outweighs its benefits.

\section{Stochastic sampling}
\label{sec:stochasticity}

Deterministic sampling offers many benefits, e.g., the ability to turn real images into their corresponding latent representations by inverting the ODE.
However, it tends to lead to worse output quality~\cite{Song2020ddim,Song2021sde} than stochastic sampling that injects fresh noise into the image in each step.
Given that ODEs and SDEs recover the same distributions in theory, what exactly is the role of stochasticity?

\vparagraph{Background.}
\label{sec:stochasticbackground}
The %
SDEs of Song et al.~\cite{Song2021sde} can be generalized~\cite{Huang2021,Zhang2021} as a sum of the probability flow ODE of Eq.~\ref{eq:ode} and a {time-varying} \emph{Langevin diffusion} SDE~\cite{Grenander1994} {(see Appendix~\ref{app:oursde})}:
\begin{equation}
  \label{eq:sde}
  \diff \xx_{\pm} =
    \underbrace{-\dot\sigma(\odetime) \sigma(\odetime) \nablaxx \log p\big( \xx; \sigma(\odetime) \big) \,\diff\odetime}_{\text{probability flow ODE (Eq.~\ref{eq:ode})}}\,\pm\,
    \underbrace{
      \underbrace{\churn(\odetime) \sigma(\odetime)^2 \nablaxx \log p\big( \xx; \sigma(\odetime) \big) \,\diff\odetime}_{\text{deterministic noise decay}} +
      \underbrace{\sqrt{2 \churn(\odetime)} \sigma(\odetime) \,\diff\wproc}_{\text{noise injection}}
    }_{\text{Langevin diffusion SDE}} ,
\end{equation}
where {$\wproc$} is the standard Wiener process.
$\diff \xx_+$ and $\diff \xx_-$ are now separate SDEs for moving forward and backward in time, related by the time reversal formula of Anderson~\cite{Anderson1982}.
The Langevin term can further be seen as a combination of %
  a deterministic score-based denoising term and a stochastic noise injection term, whose net noise level contributions cancel out.
As such, $\churn(t)$ effectively expresses the relative rate at which existing noise is replaced with new noise.
The SDEs of Song et al.~\cite{Song2021sde} are recovered with the choice $\churn(\odetime) = {\dot \sigma(\odetime)}/{\sigma(\odetime)}$, whereby the score vanishes from the forward SDE.

\algStochastic

This perspective reveals why stochasticity is helpful in practice:
  The implicit Langevin diffusion drives the sample towards the desired marginal distribution at a given time, actively correcting for any errors made in earlier sampling steps.
On the other hand, approximating the Langevin term with discrete SDE solver steps introduces error in itself.
Previous results~\cite{Bao2022analytic,Jolicoeur2021,Song2020ddim,Song2021sde} suggest that non-zero $\churn(t)$ is helpful, but
  as far as we can tell, the implicit choice for $\churn(t)$ in Song et al.~\cite{Song2021sde} enjoys no special properties.
Hence, the optimal amount of stochasticity should be determined empirically.

\vparagraph{Our stochastic sampler.}
We propose a stochastic sampler that combines our 2\textsuperscript{nd} order deterministic ODE integrator with explicit Langevin-like ``churn'' of adding and removing noise.
A pseudocode is given in Algorithm~\ref{alg:stochastic}.
At each step $i$, given the sample $\xx_i$ at noise level $t_i$ ($=\sigma(t_i)$), we perform two sub-steps.
First, we add noise to the sample according to a factor $\gamma_i\ge0$ to reach a higher noise level \smash{$\hat t_i = t_i + \gamma_i t_i$}.
Second, from the increased-noise sample \smash{$\xxh_i$}, we %
solve the ODE backward from \smash{$\hat t_i$} to $t_{i+1}$ with a single step.
This yields a sample $\xx_{i+1}$ with noise level $t_{i+1}$, and the iteration continues. We stress that this is not a general-purpose SDE solver, but a sampling procedure tailored for the specific problem. Its correctness stems from the alternation of two sub-steps that each maintain the correct distribution (up to truncation error in the ODE step).
{The predictor-corrector sampler of Song et al.~\cite{Song2021sde} has a conceptually similar structure to ours.}

{
To analyze the main difference between our method and Euler--Maruyama, we first note a subtle discrepancy in the latter when discretizing Eq.~\ref{eq:sde}.
One can interpret Euler--Maruyama as first adding noise and then performing an ODE step, not from the intermediate state after noise injection, but assuming that $\xx$ and $\sigma$ remained at the initial state at the beginning of the iteration step.
In our method, the parameters used to evaluate $D_\theta$ on line~7 of Algorithm~\ref{alg:stochastic} correspond to the state after noise injection, whereas an Euler--Maruyama -like method would use \smash{$\xx_i;\odetime_i$} instead of \smash{$\xxh_i;\hat\odetime_i$}.
In the limit of $\Delta_t$ approaching zero there may be no difference between these choices, but the distinction appears to become significant when pursuing low NFE with large steps.
}

\vparagraph{Practical considerations.}

{Increasing the amount of stochasticity is effective in correcting errors made by earlier sampling steps, but it has its own drawbacks.}
We have observed (see Appendix~{\ref{app:degradation}}) that excessive Langevin-like addition and removal of noise results in gradual loss of detail in the generated images with all datasets and denoiser networks.
There is also a drift toward oversaturated colors at very low and high noise levels.
We suspect that practical denoisers induce a slightly non-conservative vector field in Eq.~\ref{eq:scoredenoiser}, violating the premises of Langevin diffusion and causing these detrimental effects.
Notably, our experiments with analytical denoisers (such as the one in Figure~\ref{fig:denoising}b) have not shown such degradation.

If the degradation is caused by flaws in $D_\theta(\xx; \sigma)$, they can only be remedied using heuristic means during sampling.
We address the drift toward oversaturated colors by only enabling stochasticity within a specific range of noise levels $\odetime_i \in [\Stmin, \Stmax]$.
For these noise levels, we define $\gamma_i = \Schurn / N$, where $\Schurn$ controls the overall amount of stochasticity. %
We further clamp $\gamma_i$ to never introduce more new noise than what is already present in the image.
Finally, we have found that the loss of detail can be partially counteracted by setting $\Snoise$ slightly above $1$ to inflate the standard deviation for the newly added noise.
This suggests that a major component of the hypothesized non-conservativity of $D_\theta(\xx;\sigma)$ is a tendency to remove slightly too much noise\,---\,most likely due to {regression toward the mean} that can be expected to happen with any $L_2$-trained denoiser~\cite{Lehtinen2018}.

\figSdePlotNfe

\vparagraph{Evaluation.}

Figure~\ref{fig:SdePlotNfe} shows that our stochastic sampler outperforms previous samplers \cite{Jolicoeur2021,Nichol2021a,Song2021sde} by a significant margin, especially at low step counts.
{Jolicoeur-Martineau et al.~\cite{Jolicoeur2021} use a standard higher-order adaptive SDE solver~\cite{Roberts2012sde} and its performance is a good baseline for such solvers in general.
Our sampler has been tailored to the use case by, e.g., performing noise injection and ODE step sequentially, and it is not adaptive.
It is an open question if adaptive solvers can be a net win over a well-tuned fixed schedule in sampling diffusion models.}

Through sampler improvements alone, we are able to bring the ImageNet-64 model that originally achieved FID 2.07~\cite{Dhariwal2021} to 1.55 that is very close to the state-of-the-art; previously, FID 1.48 has been reported for cascaded diffusion~\cite{Ho2021cascaded}, 1.55 for classifier-free guidance~\cite{Ho2021classifierfree}, and 1.52 for StyleGAN-XL~\cite{Sauer2022}.
While our results showcase the potential gains achievable through sampler improvements, they also highlight the main shortcoming of stochasticity: For best results, one must make several heuristic choices\,---\,either implicit or explicit\,---\,that depend on the specific model.
Indeed, we had to find the optimal values of $\{\Schurn, \Stmin, \Stmax, \Snoise\}$ on a case-by-case basis using grid search (Appendix~{\ref{app:stochasticparams}}).
This raises a general concern that using stochastic sampling as the primary means of evaluating model improvements may inadvertently end up influencing the design choices related to model architecture and training.

\vspace*{-1.5mm}
\section{Preconditioning and training}
\label{sec:training}
\vspace*{-0.5mm}

There are various known good practices for training neural networks in a supervised fashion.
For example, it is advisable to keep input and output signal magnitudes fixed to, e.g., unit variance, and to
  avoid large variation in gradient magnitudes on a per-sample basis~\cite{Bishop1995book,Huang2020normalize}.
Training a neural network to model $D$ directly would be far from ideal\,---\,%
  for example, as the input $\xx=\signal+\noise$ is a combination of clean signal $\signal$ and noise $\noise\sim\mathcal{N}(\boldzero,\sigma^2 \boldi)$,
  its magnitude varies immensely depending on noise level $\sigma$.
For this reason, the common practice is to not represent $D_\theta$ as a neural network directly, but instead train a different network $F_\theta$ from which $D_\theta$ is derived.

Previous methods~\cite{Nichol2021a,Song2020ddim,Song2021sde} address the input scaling via a $\sigma$-dependent normalization factor and
  attempt to precondition the output by training $F_\theta$ to predict $\noise$ scaled to unit variance,
  from which the signal is then reconstructed via $D_\theta(\xx;\sigma) = \xx-\sigma F_\theta(\cdot)$.
This has the drawback that at large $\sigma$, the network needs to fine-tune its output carefully to cancel out the existing noise $\noise$ exactly and give the output at the correct scale;
  note that any errors made by the network are amplified by a factor of $\sigma$.
In this situation, it would seem much easier to predict the expected output $D(\xx; \sigma)$ directly.
{In the same spirit as previous parameterizations that adaptively mix signal and noise (e.g., \cite{Dockhorn2022damped,Salimans2022,Vahdat2021}), }%
we propose to precondition the neural network with a $\sigma$-dependent skip connection that allows it to estimate either $\signal$ or $\noise$, or something in between.
We thus write $D_\theta$ in the following form:
\begin{equation}
D_\theta(\xx; \sigma) = \cskip(\sigma) ~\xx + \cout(\sigma) ~F_\theta \big( \cin(\sigma) ~\xx; ~\cnoise(\sigma) \big) \text{,}
\label{eq:preconditioning}
\end{equation}
where $F_\theta$ is the neural network to be trained,
  $\cskip(\sigma)$ modulates the skip connection,
  $\cin(\sigma)$ and $\cout(\sigma)$ scale the input and output magnitudes, and
  $\cnoise(\sigma)$ maps noise level $\sigma$ into a conditioning input for $F_\theta$.
Taking a weighted expectation of Eq.~\ref{eq:score} over the noise levels gives \mbox{the overall training} loss
  \smash{$\mathbb{E}_{\sigma, \signal, \noise} \left[ \lambda(\sigma) ~ \lVert D(\signal + \noise; \sigma) - \signal \rVert^2_2 \right]$},
  where \smash{$\sigma \sim \ptrain$}, \smash{$\signal \sim \pdata$}, and \smash{$\noise \sim \mathcal{N}(\boldzero, \sigma^2 \boldi)$}.
The probability of sampling a given noise level $\sigma$ is given by $\ptrain(\sigma)$ and the corresponding weight is given by $\lambda(\sigma)$.
We can equivalently express this loss with respect to the raw network output $F_\theta$ in Eq.~\ref{eq:preconditioning}:
\begin{equation}
  \mathbb{E}_{\sigma, \signal, \noise} \Big[
    \underbrace{\lambda(\sigma) ~ \cout(\sigma)^2}_{\text{effective weight}}
    \big\Vert
      \underbrace{F_\theta \big( \cin(\sigma) \cdot (\signal + \noise); \cnoise(\sigma) \big)}_{\text{network output}} -
      \underbrace{\tfrac{1}{\cout(\sigma)} \big(\signal - \cskip(\sigma) \cdot (\signal + \noise) \big)}_{\text{effective training target}}
    \big\Vert^2_2 \Big] \text{.}
  \label{eq:precloss}
\end{equation}
This form reveals the effective training target of $F_\theta$, allowing us to determine suitable choices for the preconditioning functions from first principles.
As detailed in Appendix~{\ref{app:ourprecond}}, we derive our choices shown in Table~\ref{tab:specifics} by
  requiring network inputs and training targets to have unit variance ($\cin$, $\cout$),
  and amplifying errors in $F_\theta$ as little as possible ($\cskip$).
The formula for $\cnoise$ is chosen empirically.

\tabTrainingTable

Table~\ref{tab:TrainingTable} shows FID for a series of training setups, evaluated using our deterministic sampler from Section~\ref{sec:deterministic}.
We start with the baseline training setup of Song~et~al.~\cite{Song2021sde}, which differs considerably between the VP and VE cases; we provide separate results for each (config \textsc{a}).
To obtain a more meaningful point of comparison, we re-adjust the basic hyperparameters (config \textsc{b}) and improve the expressive power of the model (config \textsc{c}) by removing the lowest-resolution layers and doubling the capacity of the highest-resolution layers instead; see Appendix~{\ref{app:training}} for further details.
We then replace the original choices of $\{\cin, \cout, \cnoise, \cskip\}$ with our preconditioning (config \textsc{d}), which keeps the results largely unchanged\,---\,except for VE that improves considerably at 64$\times$64 resolution.
Instead of improving FID per se, the main benefit of our preconditioning is that it makes the training more robust, enabling us to turn our focus on redesigning the loss function without adverse effects.

\vparagraph{Loss weighting and sampling.}

Eq.~\ref{eq:precloss} shows that training $F_\theta$ as preconditioned in Eq.~\ref{eq:preconditioning} incurs an effective per-sample loss weight of $\lambda(\sigma)\cout(\sigma)^2$.
To balance the effective loss weights, we set $\lambda(\sigma)=1/\cout(\sigma)^2$,
  which also equalizes the initial training loss over the entire $\sigma$ range as shown in Figure~\ref{fig:TrainingPlots}a (green curve).
Finally, we need to select $\ptrain(\sigma)$, i.e., how to choose noise levels during training.
Inspecting the per-$\sigma$ loss after training (blue and orange curves) reveals that a significant reduction is possible only at intermediate noise levels;
  at very low levels, it is both difficult and irrelevant to discern the vanishingly small noise component, 
  whereas at high levels the training targets are always dissimilar from the correct answer that approaches dataset average.
Therefore, we target the training efforts to the relevant range using a simple log-normal distribution for $\ptrain(\sigma)$ as detailed in Table~\ref{tab:specifics}
  and illustrated in Figure~\ref{fig:TrainingPlots}a (red curve).

Table~\ref{tab:TrainingTable} shows that our proposed $\ptrain$ and $\lambda$ (config \textsc{e}) lead to a dramatic improvement in FID in all cases when used in conjunction with our preconditioning (config \textsc{d}).
In concurrent work, Choi~et~al.~\cite{Choi2022} propose a similar scheme to prioritize noise levels that are most relevant w.r.t.~forming the perceptually recognizable content of the image.
However, they only consider the choice of $\lambda$ in isolation, which results in a smaller overall improvement.

\vparagraph{Augmentation regularization.}

To prevent potential overfitting that often plagues diffusion models with smaller datasets, we borrow an augmentation pipeline from the GAN literature~\cite{Karras2020ada}.
The pipeline consists of various geometric transformations (see Appendix~{\ref{app:augmentdetails}}) that we apply to a training image prior to adding noise.
To prevent the augmentations from leaking to the generated images, we provide the augmentation parameters as a conditioning input to $F_\theta$;
  during inference we set the them to zero to guarantee that only non-augmented images are generated.
Table~\ref{tab:TrainingTable} shows that data augmentation provides a consistent improvement (config \textsc{f}) that yields new state-of-the-art
  FIDs of 1.79 and 1.97 for conditional and unconditional CIFAR-10, beating the previous records of 1.85~\cite{Sauer2022} and 2.10~\cite{Vahdat2021}.

\vparagraph{Stochastic sampling revisited.}

\figTrainingPlots

{
Interestingly, the relevance of stochastic sampling appears to diminish as the model itself improves, as shown in Figure~\ref{fig:TrainingPlots}b,c.
When using our training setup in CIFAR-10 (Figure~\ref{fig:TrainingPlots}b), the best results were obtained with deterministic sampling, and any amount of stochastic sampling was detrimental.%
}

\vparagraph{ImageNet-64.}

{
As a final experiment, we trained a class-conditional ImageNet-64 model from scratch using our proposed training improvements.
This model achieved a new state-of-the-art FID of 1.36 compared to the previous record of 1.48~\cite{Ho2021cascaded}.
We used the ADM architecture~\cite{Dhariwal2021} with no changes, and trained it using our config~\textsc{e} with minimal tuning; see Appendix~\ref{app:training} for details.
We did not find overfitting to be a concern, and thus chose to not employ augmentation regularization.
As shown in Figure~\ref{fig:TrainingPlots}c, the optimal amount of stochastic sampling was much lower than with the pre-trained model, but unlike with CIFAR-10, stochastic sampling was clearly better than deterministic sampling.
This suggests that more diverse datasets continue to benefit from stochastic sampling.%
}

\vspace*{-1mm}
\section{Conclusions}
\label{sec:conclusions}

Our approach of putting diffusion models to a common framework exposes a modular design. 
This allows a targeted investigation of individual components, potentially helping to better cover the viable design space.
In our tests this let us simply replace the samplers in various earlier models, drastically improving the results.
For example, in ImageNet-64 our sampler turned an average model (FID 2.07) to a challenger (1.55) for the {previous} SOTA model (1.48)~\cite{Ho2021cascaded},
{and with training improvements achieved SOTA FID of 1.36}.
We also obtained new state-of-the-art results on CIFAR-10 while using only 35 model evaluations, deterministic sampling, and a small network.
The current high-resolution diffusion models rely either on separate super-resolution steps~\cite{Ho2021cascaded,Nichol2021b,Ramesh2022}, subspace projection~\cite{Jing2022}, very large networks~\cite{Dhariwal2021,Song2021sde}, or hybrid approaches~\cite{Preechakul2021diffusion,Rombach2021highresolution,Vahdat2021}\,---\,we believe that our contributions are orthogonal to these extensions.
That said, many of our parameter values may need to be re-adjusted for higher resolution datasets.
{Furthermore, we feel that the precise interaction between stochastic sampling and the training objective remains an interesting question for future work.}

\vparagraph{Societal impact.}
Our advances in sample quality can potentially amplify negative societal effects when used in a large-scale system like DALL$\cdot$E~2, including types of  disinformation or emphasizing sterotypes and harmful biases~\cite{Mishkin2022risks}.
The training and sampling of diffusion models needs a lot of electricity; our project consumed $\sim$250MWh on an in-house cluster of NVIDIA V100s.

{
\vparagraph{Acknowledgments.}
We thank Jaakko Lehtinen, Ming-Yu Liu, Tuomas Kynk\"a\"anniemi, Axel Sauer, Arash Vahdat, and Janne Hellsten for discussions and comments, and Tero Kuosmanen, Samuel Klenberg, and Janne Hellsten for maintaining our compute infrastructure.
}

{\small
  \bibliographystyle{ieee}
  \bibliography{paper}
}

\ifappendix
  \newpage
  \appendix
  {\LARGE\bf Appendices}
	\newcommand{\refpaper}[1]{\ref{#1}}
  \newif\iffigures\figurestrue
  \newcommand{\genOdeTable}{%
\multicolumn{1}{l|}{}                                   & \multicolumn{4}{c|}{Unconditional CIFAR-10 at 32$\times$32}                   & \multicolumn{2}{c|}{Class-conditional}  \\
\multicolumn{1}{l|}{}                                   & \multicolumn{2}{c|}{VP}               & \multicolumn{2}{c|}{VE}               & \multicolumn{2}{c|}{ImageNet-64}        \\
\tabucline{-}
{\bf Sampling method}                                   & FID~$\downarrow$  & NFE~$\downarrow$  & FID~$\downarrow$  & NFE~$\downarrow$  & FID~$\downarrow$  & NFE~$\downarrow$    \\
\tabucline{-}
\low{Original sampler~\cite{Song2021sde,Dhariwal2021}}  & 2.85              & 256               & 5.45              & 8192              & 2.85              & 250                 \\
Our Algorithm~\refpaper{alg:heun}                       & {\bf2.79}         & 512               & 4.78              & 8192              & 2.73              & 384                 \\
+ Heun \& our $t_i$                                     & 2.88              & 255               & 4.23              & \s191             & {\bf2.64}         & {\bf\s79}           \\
+ Our $\sigma(t)$ \& $s(t)$                             & 2.93              & {\bf\s35}         & 3.73              & {\bf\s\s27}       & \s\s--            & \s--                \\
\tabucline{-}
Black-box RK45                                          & 2.94              & 115               & {\bf3.69}         & \s\s93            & 2.66              & 131                 \\
}

\newcommand{\genOdePlotExpCifaruNcsnppErrEulerA}{%
(0.02, 0.558655)
(0.0206232, 0.561613)
(0.0212657, 0.559859)
(0.0219283, 0.558745)
(0.0226116, 0.558902)
(0.0233161, 0.560507)
(0.0240426, 0.560152)
(0.0247917, 0.559604)
(0.0255642, 0.560152)
(0.0263607, 0.559175)
(0.0271821, 0.560133)
(0.028029, 0.560453)
(0.0289023, 0.5611)
(0.0298029, 0.561208)
(0.0307315, 0.560457)
(0.031689, 0.560423)
(0.0326764, 0.561024)
(0.0336945, 0.560595)
(0.0347444, 0.560836)
(0.035827, 0.561179)
(0.0369432, 0.560846)
(0.0380943, 0.561993)
(0.0392813, 0.561875)
(0.0405052, 0.562152)
(0.0417673, 0.562733)
(0.0430687, 0.561715)
(0.0444106, 0.561251)
(0.0457943, 0.560956)
(0.0472212, 0.559959)
(0.0486925, 0.561317)
(0.0502097, 0.561818)
(0.0517741, 0.560882)
(0.0533873, 0.560992)
(0.0550508, 0.559823)
(0.056766, 0.55934)
(0.0585348, 0.559507)
(0.0603586, 0.559387)
(0.0622393, 0.559332)
(0.0641785, 0.559212)
(0.0661782, 0.5595)
(0.0682402, 0.558675)
(0.0703664, 0.559209)
(0.0725589, 0.558975)
(0.0748197, 0.558844)
(0.0771509, 0.556664)
(0.0795548, 0.557912)
(0.0820336, 0.557672)
(0.0845896, 0.554681)
(0.0872253, 0.554838)
(0.089943, 0.556618)
(0.0927455, 0.553974)
(0.0956352, 0.553571)
(0.0986151, 0.552802)
(0.101688, 0.552853)
(0.104856, 0.550666)
(0.108123, 0.550447)
(0.111492, 0.549947)
(0.114966, 0.548215)
(0.118548, 0.547022)
(0.122242, 0.547116)
(0.126051, 0.54527)
(0.129978, 0.543856)
(0.134028, 0.544394)
(0.138204, 0.543138)
(0.14251, 0.541274)
(0.146951, 0.540352)
(0.151529, 0.539053)
(0.156251, 0.53716)
(0.161119, 0.536198)
(0.166139, 0.533698)
(0.171316, 0.532853)
(0.176654, 0.531639)
(0.182158, 0.53033)
(0.187834, 0.528344)
(0.193686, 0.526408)
(0.199721, 0.524327)
(0.205944, 0.522867)
(0.212361, 0.52068)
(0.218978, 0.519063)
(0.225801, 0.515913)
(0.232836, 0.515092)
(0.240091, 0.512833)
(0.247572, 0.509669)
(0.255286, 0.507691)
(0.26324, 0.505601)
(0.271442, 0.50303)
(0.279899, 0.501162)
(0.288621, 0.499025)
(0.297613, 0.496119)
(0.306886, 0.49352)
(0.316448, 0.489678)
(0.326308, 0.487552)
(0.336476, 0.484825)
(0.346959, 0.481992)
(0.35777, 0.478649)
(0.368917, 0.47465)
(0.380412, 0.473055)
(0.392265, 0.469874)
(0.404487, 0.465955)
(0.41709, 0.462977)
(0.430086, 0.458998)
(0.443487, 0.456895)
(0.457305, 0.453086)
(0.471554, 0.449309)
(0.486247, 0.445725)
(0.501397, 0.442656)
(0.51702, 0.439222)
(0.533129, 0.434265)
(0.54974, 0.431517)
(0.566869, 0.427918)
(0.584532, 0.423341)
(0.602745, 0.420549)
(0.621525, 0.415941)
(0.64089, 0.412103)
(0.660859, 0.408319)
(0.68145, 0.40299)
(0.702683, 0.399129)
(0.724577, 0.394692)
(0.747154, 0.390618)
(0.770434, 0.386524)
(0.794439, 0.381637)
(0.819192, 0.377896)
(0.844717, 0.373676)
(0.871036, 0.36903)
(0.898176, 0.3636)
(0.926162, 0.360206)
(0.955019, 0.354411)
(0.984776, 0.349933)
(1.01546, 0.345389)
(1.0471, 0.341626)
(1.07972, 0.336322)
(1.11337, 0.331825)
(1.14806, 0.32739)
(1.18383, 0.322824)
(1.22071, 0.318335)
(1.25875, 0.313367)
(1.29797, 0.308778)
(1.33841, 0.304195)
(1.38011, 0.299049)
(1.42312, 0.294145)
(1.46746, 0.289833)
(1.51318, 0.285107)
(1.56033, 0.280143)
(1.60895, 0.275958)
(1.65908, 0.270824)
(1.71077, 0.266217)
(1.76408, 0.262068)
(1.81904, 0.257463)
(1.87572, 0.252216)
(1.93416, 0.247842)
(1.99443, 0.243678)
(2.05657, 0.23889)
(2.12065, 0.234326)
(2.18672, 0.230137)
(2.25486, 0.225515)
(2.32512, 0.22206)
(2.39756, 0.216616)
(2.47227, 0.212737)
(2.5493, 0.208604)
(2.62873, 0.203929)
(2.71063, 0.200023)
(2.79509, 0.195448)
(2.88218, 0.191905)
(2.97198, 0.187751)
(3.06459, 0.184147)
(3.16007, 0.179908)
(3.25853, 0.175938)
(3.36006, 0.172183)
(3.46476, 0.168631)
(3.57271, 0.165095)
(3.68403, 0.161048)
(3.79882, 0.157439)
(3.91718, 0.153725)
(4.03923, 0.150529)
(4.16509, 0.147336)
(4.29487, 0.14338)
(4.42869, 0.140293)
(4.56668, 0.137304)
(4.70896, 0.133977)
(4.85569, 0.130948)
(5.00698, 0.127762)
(5.16299, 0.12415)
(5.32386, 0.120911)
(5.48974, 0.118373)
(5.66079, 0.115624)
(5.83717, 0.11293)
(6.01904, 0.109879)
(6.20658, 0.107979)
(6.39997, 0.105319)
(6.59938, 0.10261)
(6.805, 0.0998765)
(7.01704, 0.0970993)
(7.23567, 0.0942533)
(7.46112, 0.0917961)
(7.6936, 0.0891668)
(7.93332, 0.0869712)
(8.1805, 0.0846815)
(8.43539, 0.0824444)
(8.69822, 0.0805714)
(8.96924, 0.0778091)
(9.24871, 0.0760172)
(9.53688, 0.0736472)
(9.83403, 0.0719696)
(10.1404, 0.0705019)
(10.4564, 0.0682142)
(10.7822, 0.0663394)
(11.1181, 0.064956)
(11.4646, 0.0628071)
(11.8218, 0.0610241)
(12.1901, 0.0594766)
(12.5699, 0.0581671)
(12.9616, 0.0562542)
(13.3655, 0.0549383)
(13.7819, 0.0533563)
(14.2113, 0.0516123)
(14.6541, 0.0504591)
(15.1107, 0.0487845)
(15.5815, 0.047781)
(16.067, 0.0462336)
(16.5676, 0.0447647)
(17.0839, 0.043069)
(17.6162, 0.0423792)
(18.165, 0.0416531)
(18.731, 0.0396529)
(19.3147, 0.0388143)
(19.9165, 0.0379408)
(20.537, 0.0364979)
(21.1769, 0.0354981)
(21.8368, 0.0346866)
(22.5171, 0.0336317)
(23.2187, 0.0326307)
(23.9422, 0.0318035)
(24.6882, 0.0309467)
(25.4574, 0.0296384)
(26.2506, 0.0288907)
(27.0685, 0.0286705)
(27.9119, 0.0273634)
(28.7816, 0.0269756)
(29.6784, 0.0260228)
(30.6031, 0.0249076)
(31.5567, 0.0247606)
(32.5399, 0.0236178)
(33.5538, 0.0227893)
(34.5993, 0.0224866)
(35.6773, 0.0216599)
(36.789, 0.0209729)
(37.9352, 0.020473)
(39.1172, 0.0201902)
(40.336, 0.0195326)
(41.5928, 0.0187142)
(42.8888, 0.0184692)
(44.2251, 0.0174983)
(45.6031, 0.016842)
(47.024, 0.0164666)
(48.4892, 0.0163824)
(50, 0.0153828)
}

\newcommand{\genOdePlotExpCifaruNcsnppErrEulerALo}{%
(0.02, 0.548261)
(0.0206232, 0.550346)
(0.0212657, 0.548574)
(0.0219283, 0.547493)
(0.0226116, 0.548118)
(0.0233161, 0.549456)
(0.0240426, 0.549671)
(0.0247917, 0.548529)
(0.0255642, 0.548525)
(0.0263607, 0.548834)
(0.0271821, 0.548764)
(0.028029, 0.549542)
(0.0289023, 0.551173)
(0.0298029, 0.549697)
(0.0307315, 0.549599)
(0.031689, 0.550277)
(0.0326764, 0.550096)
(0.0336945, 0.549962)
(0.0347444, 0.549419)
(0.035827, 0.551133)
(0.0369432, 0.550393)
(0.0380943, 0.551556)
(0.0392813, 0.551796)
(0.0405052, 0.551473)
(0.0417673, 0.551104)
(0.0430687, 0.551374)
(0.0444106, 0.550672)
(0.0457943, 0.550706)
(0.0472212, 0.549624)
(0.0486925, 0.549666)
(0.0502097, 0.551412)
(0.0517741, 0.550711)
(0.0533873, 0.550724)
(0.0550508, 0.550064)
(0.056766, 0.548884)
(0.0585348, 0.549057)
(0.0603586, 0.549294)
(0.0622393, 0.549415)
(0.0641785, 0.548586)
(0.0661782, 0.549206)
(0.0682402, 0.547999)
(0.0703664, 0.549052)
(0.0725589, 0.549016)
(0.0748197, 0.548142)
(0.0771509, 0.546497)
(0.0795548, 0.548075)
(0.0820336, 0.547419)
(0.0845896, 0.544767)
(0.0872253, 0.54555)
(0.089943, 0.546948)
(0.0927455, 0.54455)
(0.0956352, 0.543423)
(0.0986151, 0.542009)
(0.101688, 0.542955)
(0.104856, 0.540676)
(0.108123, 0.540378)
(0.111492, 0.539615)
(0.114966, 0.538157)
(0.118548, 0.536867)
(0.122242, 0.53767)
(0.126051, 0.535421)
(0.129978, 0.534118)
(0.134028, 0.534459)
(0.138204, 0.533893)
(0.14251, 0.532583)
(0.146951, 0.531851)
(0.151529, 0.529832)
(0.156251, 0.528852)
(0.161119, 0.527002)
(0.166139, 0.524704)
(0.171316, 0.524295)
(0.176654, 0.522992)
(0.182158, 0.521212)
(0.187834, 0.519838)
(0.193686, 0.518185)
(0.199721, 0.515895)
(0.205944, 0.515329)
(0.212361, 0.512518)
(0.218978, 0.511171)
(0.225801, 0.508484)
(0.232836, 0.507512)
(0.240091, 0.504317)
(0.247572, 0.501378)
(0.255286, 0.50004)
(0.26324, 0.497901)
(0.271442, 0.494953)
(0.279899, 0.494043)
(0.288621, 0.492101)
(0.297613, 0.488867)
(0.306886, 0.486326)
(0.316448, 0.481985)
(0.326308, 0.48098)
(0.336476, 0.477729)
(0.346959, 0.474983)
(0.35777, 0.471374)
(0.368917, 0.467881)
(0.380412, 0.46661)
(0.392265, 0.463804)
(0.404487, 0.459381)
(0.41709, 0.456609)
(0.430086, 0.452516)
(0.443487, 0.450746)
(0.457305, 0.446589)
(0.471554, 0.443158)
(0.486247, 0.439391)
(0.501397, 0.43634)
(0.51702, 0.43294)
(0.533129, 0.428049)
(0.54974, 0.426007)
(0.566869, 0.422165)
(0.584532, 0.417786)
(0.602745, 0.414863)
(0.621525, 0.410022)
(0.64089, 0.40643)
(0.660859, 0.403327)
(0.68145, 0.39743)
(0.702683, 0.393716)
(0.724577, 0.389427)
(0.747154, 0.38556)
(0.770434, 0.381892)
(0.794439, 0.376975)
(0.819192, 0.373108)
(0.844717, 0.369151)
(0.871036, 0.364294)
(0.898176, 0.359164)
(0.926162, 0.355377)
(0.955019, 0.349995)
(0.984776, 0.3455)
(1.01546, 0.340835)
(1.0471, 0.337443)
(1.07972, 0.332202)
(1.11337, 0.327459)
(1.14806, 0.323513)
(1.18383, 0.318898)
(1.22071, 0.314693)
(1.25875, 0.309475)
(1.29797, 0.305108)
(1.33841, 0.300033)
(1.38011, 0.29549)
(1.42312, 0.290767)
(1.46746, 0.286208)
(1.51318, 0.281855)
(1.56033, 0.276466)
(1.60895, 0.272576)
(1.65908, 0.267717)
(1.71077, 0.262681)
(1.76408, 0.258801)
(1.81904, 0.253949)
(1.87572, 0.248846)
(1.93416, 0.244711)
(1.99443, 0.240662)
(2.05657, 0.235885)
(2.12065, 0.231389)
(2.18672, 0.227427)
(2.25486, 0.223006)
(2.32512, 0.219525)
(2.39756, 0.213914)
(2.47227, 0.210176)
(2.5493, 0.205868)
(2.62873, 0.201598)
(2.71063, 0.197786)
(2.79509, 0.192868)
(2.88218, 0.189315)
(2.97198, 0.185359)
(3.06459, 0.182115)
(3.16007, 0.177574)
(3.25853, 0.173664)
(3.36006, 0.169839)
(3.46476, 0.166271)
(3.57271, 0.162989)
(3.68403, 0.159254)
(3.79882, 0.155743)
(3.91718, 0.151774)
(4.03923, 0.148728)
(4.16509, 0.145711)
(4.29487, 0.141688)
(4.42869, 0.138742)
(4.56668, 0.135593)
(4.70896, 0.13219)
(4.85569, 0.129137)
(5.00698, 0.125845)
(5.16299, 0.122254)
(5.32386, 0.118576)
(5.48974, 0.116223)
(5.66079, 0.113378)
(5.83717, 0.110659)
(6.01904, 0.107768)
(6.20658, 0.107198)
(6.39997, 0.104717)
(6.59938, 0.101847)
(6.805, 0.0990317)
(7.01704, 0.0961086)
(7.23567, 0.0931552)
(7.46112, 0.0906017)
(7.6936, 0.0879516)
(7.93332, 0.0857365)
(8.1805, 0.0834992)
(8.43539, 0.0813293)
(8.69822, 0.0796058)
(8.96924, 0.076839)
(9.24871, 0.07519)
(9.53688, 0.0727126)
(9.83403, 0.0710916)
(10.1404, 0.0696173)
(10.4564, 0.0672747)
(10.7822, 0.0654985)
(11.1181, 0.0641135)
(11.4646, 0.0619284)
(11.8218, 0.0603843)
(12.1901, 0.0587999)
(12.5699, 0.0573067)
(12.9616, 0.0557885)
(13.3655, 0.0542105)
(13.7819, 0.0525164)
(14.2113, 0.0507444)
(14.6541, 0.0498905)
(15.1107, 0.0483217)
(15.5815, 0.0471437)
(16.067, 0.0457079)
(16.5676, 0.0442968)
(17.0839, 0.042712)
(17.6162, 0.041637)
(18.165, 0.0410932)
(18.731, 0.0390536)
(19.3147, 0.0383665)
(19.9165, 0.0373147)
(20.537, 0.0360746)
(21.1769, 0.0349476)
(21.8368, 0.0340959)
(22.5171, 0.033247)
(23.2187, 0.0321001)
(23.9422, 0.0313357)
(24.6882, 0.0306278)
(25.4574, 0.0291878)
(26.2506, 0.0283832)
(27.0685, 0.0283943)
(27.9119, 0.0269153)
(28.7816, 0.0266604)
(29.6784, 0.0255621)
(30.6031, 0.0246492)
(31.5567, 0.0243857)
(32.5399, 0.0232452)
(33.5538, 0.0225402)
(34.5993, 0.0221401)
(35.6773, 0.0214487)
(36.789, 0.0206863)
(37.9352, 0.0201705)
(39.1172, 0.0198659)
(40.336, 0.0192213)
(41.5928, 0.0184995)
(42.8888, 0.0183147)
(44.2251, 0.0171814)
(45.6031, 0.0166032)
(47.024, 0.0162699)
(48.4892, 0.0161155)
(50, 0.0151198)
}

\newcommand{\genOdePlotExpCifaruNcsnppErrEulerAHi}{%
(0.02, 0.569049)
(0.0206232, 0.57288)
(0.0212657, 0.571144)
(0.0219283, 0.569997)
(0.0226116, 0.569686)
(0.0233161, 0.571558)
(0.0240426, 0.570632)
(0.0247917, 0.570679)
(0.0255642, 0.57178)
(0.0263607, 0.569517)
(0.0271821, 0.571501)
(0.028029, 0.571364)
(0.0289023, 0.571027)
(0.0298029, 0.572719)
(0.0307315, 0.571314)
(0.031689, 0.570568)
(0.0326764, 0.571952)
(0.0336945, 0.571229)
(0.0347444, 0.572254)
(0.035827, 0.571226)
(0.0369432, 0.571299)
(0.0380943, 0.572431)
(0.0392813, 0.571954)
(0.0405052, 0.572831)
(0.0417673, 0.574362)
(0.0430687, 0.572055)
(0.0444106, 0.57183)
(0.0457943, 0.571207)
(0.0472212, 0.570294)
(0.0486925, 0.572968)
(0.0502097, 0.572225)
(0.0517741, 0.571053)
(0.0533873, 0.571261)
(0.0550508, 0.569582)
(0.056766, 0.569796)
(0.0585348, 0.569957)
(0.0603586, 0.569481)
(0.0622393, 0.569249)
(0.0641785, 0.569838)
(0.0661782, 0.569793)
(0.0682402, 0.569351)
(0.0703664, 0.569367)
(0.0725589, 0.568934)
(0.0748197, 0.569546)
(0.0771509, 0.566831)
(0.0795548, 0.567749)
(0.0820336, 0.567925)
(0.0845896, 0.564596)
(0.0872253, 0.564126)
(0.089943, 0.566287)
(0.0927455, 0.563398)
(0.0956352, 0.563719)
(0.0986151, 0.563595)
(0.101688, 0.56275)
(0.104856, 0.560655)
(0.108123, 0.560516)
(0.111492, 0.56028)
(0.114966, 0.558274)
(0.118548, 0.557177)
(0.122242, 0.556561)
(0.126051, 0.55512)
(0.129978, 0.553595)
(0.134028, 0.554328)
(0.138204, 0.552383)
(0.14251, 0.549964)
(0.146951, 0.548853)
(0.151529, 0.548274)
(0.156251, 0.545468)
(0.161119, 0.545394)
(0.166139, 0.542691)
(0.171316, 0.541411)
(0.176654, 0.540286)
(0.182158, 0.539448)
(0.187834, 0.53685)
(0.193686, 0.53463)
(0.199721, 0.53276)
(0.205944, 0.530405)
(0.212361, 0.528842)
(0.218978, 0.526955)
(0.225801, 0.523342)
(0.232836, 0.522672)
(0.240091, 0.521349)
(0.247572, 0.517959)
(0.255286, 0.515341)
(0.26324, 0.513301)
(0.271442, 0.511107)
(0.279899, 0.50828)
(0.288621, 0.505949)
(0.297613, 0.50337)
(0.306886, 0.500714)
(0.316448, 0.497372)
(0.326308, 0.494125)
(0.336476, 0.491922)
(0.346959, 0.489001)
(0.35777, 0.485924)
(0.368917, 0.481418)
(0.380412, 0.4795)
(0.392265, 0.475943)
(0.404487, 0.47253)
(0.41709, 0.469345)
(0.430086, 0.46548)
(0.443487, 0.463044)
(0.457305, 0.459582)
(0.471554, 0.45546)
(0.486247, 0.452058)
(0.501397, 0.448972)
(0.51702, 0.445504)
(0.533129, 0.44048)
(0.54974, 0.437027)
(0.566869, 0.433671)
(0.584532, 0.428895)
(0.602745, 0.426235)
(0.621525, 0.42186)
(0.64089, 0.417776)
(0.660859, 0.41331)
(0.68145, 0.40855)
(0.702683, 0.404542)
(0.724577, 0.399957)
(0.747154, 0.395676)
(0.770434, 0.391156)
(0.794439, 0.386298)
(0.819192, 0.382684)
(0.844717, 0.378202)
(0.871036, 0.373767)
(0.898176, 0.368036)
(0.926162, 0.365035)
(0.955019, 0.358828)
(0.984776, 0.354366)
(1.01546, 0.349943)
(1.0471, 0.345809)
(1.07972, 0.340442)
(1.11337, 0.336192)
(1.14806, 0.331267)
(1.18383, 0.32675)
(1.22071, 0.321978)
(1.25875, 0.317259)
(1.29797, 0.312447)
(1.33841, 0.308357)
(1.38011, 0.302607)
(1.42312, 0.297523)
(1.46746, 0.293458)
(1.51318, 0.288359)
(1.56033, 0.283821)
(1.60895, 0.279341)
(1.65908, 0.27393)
(1.71077, 0.269753)
(1.76408, 0.265335)
(1.81904, 0.260977)
(1.87572, 0.255587)
(1.93416, 0.250974)
(1.99443, 0.246694)
(2.05657, 0.241895)
(2.12065, 0.237264)
(2.18672, 0.232848)
(2.25486, 0.228024)
(2.32512, 0.224596)
(2.39756, 0.219319)
(2.47227, 0.215298)
(2.5493, 0.21134)
(2.62873, 0.206261)
(2.71063, 0.202261)
(2.79509, 0.198028)
(2.88218, 0.194495)
(2.97198, 0.190144)
(3.06459, 0.186179)
(3.16007, 0.182242)
(3.25853, 0.178211)
(3.36006, 0.174528)
(3.46476, 0.170991)
(3.57271, 0.1672)
(3.68403, 0.162842)
(3.79882, 0.159136)
(3.91718, 0.155676)
(4.03923, 0.15233)
(4.16509, 0.148961)
(4.29487, 0.145072)
(4.42869, 0.141845)
(4.56668, 0.139014)
(4.70896, 0.135764)
(4.85569, 0.132759)
(5.00698, 0.129678)
(5.16299, 0.126046)
(5.32386, 0.123246)
(5.48974, 0.120523)
(5.66079, 0.11787)
(5.83717, 0.115201)
(6.01904, 0.11199)
(6.20658, 0.108761)
(6.39997, 0.10592)
(6.59938, 0.103373)
(6.805, 0.100721)
(7.01704, 0.0980901)
(7.23567, 0.0953514)
(7.46112, 0.0929906)
(7.6936, 0.0903821)
(7.93332, 0.0882058)
(8.1805, 0.0858637)
(8.43539, 0.0835596)
(8.69822, 0.0815371)
(8.96924, 0.0787792)
(9.24871, 0.0768445)
(9.53688, 0.0745817)
(9.83403, 0.0728475)
(10.1404, 0.0713865)
(10.4564, 0.0691538)
(10.7822, 0.0671804)
(11.1181, 0.0657984)
(11.4646, 0.0636857)
(11.8218, 0.061664)
(12.1901, 0.0601532)
(12.5699, 0.0590276)
(12.9616, 0.05672)
(13.3655, 0.0556661)
(13.7819, 0.0541963)
(14.2113, 0.0524801)
(14.6541, 0.0510276)
(15.1107, 0.0492472)
(15.5815, 0.0484183)
(16.067, 0.0467593)
(16.5676, 0.0452327)
(17.0839, 0.0434259)
(17.6162, 0.0431214)
(18.165, 0.0422129)
(18.731, 0.0402523)
(19.3147, 0.0392622)
(19.9165, 0.0385669)
(20.537, 0.0369212)
(21.1769, 0.0360486)
(21.8368, 0.0352774)
(22.5171, 0.0340165)
(23.2187, 0.0331612)
(23.9422, 0.0322714)
(24.6882, 0.0312656)
(25.4574, 0.0300891)
(26.2506, 0.0293981)
(27.0685, 0.0289467)
(27.9119, 0.0278115)
(28.7816, 0.0272907)
(29.6784, 0.0264835)
(30.6031, 0.025166)
(31.5567, 0.0251356)
(32.5399, 0.0239904)
(33.5538, 0.0230385)
(34.5993, 0.0228332)
(35.6773, 0.021871)
(36.789, 0.0212596)
(37.9352, 0.0207755)
(39.1172, 0.0205146)
(40.336, 0.019844)
(41.5928, 0.018929)
(42.8888, 0.0186237)
(44.2251, 0.0178152)
(45.6031, 0.0170807)
(47.024, 0.0166632)
(48.4892, 0.0166493)
(50, 0.0156458)
}

\newcommand{\genOdePlotExpCifaruNcsnppErrEulerB}{%
(0.02, 0.0791215)
(0.0206232, 0.0790947)
(0.0212657, 0.0792898)
(0.0219283, 0.079094)
(0.0226116, 0.0797705)
(0.0233161, 0.0800898)
(0.0240426, 0.0799071)
(0.0247917, 0.0804416)
(0.0255642, 0.0805373)
(0.0263607, 0.0810136)
(0.0271821, 0.0816531)
(0.028029, 0.0817523)
(0.0289023, 0.0815471)
(0.0298029, 0.0817765)
(0.0307315, 0.0817545)
(0.031689, 0.0817286)
(0.0326764, 0.0823593)
(0.0336945, 0.0825742)
(0.0347444, 0.0822869)
(0.035827, 0.0832378)
(0.0369432, 0.0832771)
(0.0380943, 0.0838334)
(0.0392813, 0.0837868)
(0.0405052, 0.0838053)
(0.0417673, 0.0839516)
(0.0430687, 0.0846045)
(0.0444106, 0.0844139)
(0.0457943, 0.0852114)
(0.0472212, 0.0844882)
(0.0486925, 0.0850024)
(0.0502097, 0.0854964)
(0.0517741, 0.0855361)
(0.0533873, 0.0857298)
(0.0550508, 0.0856566)
(0.056766, 0.0856638)
(0.0585348, 0.0859213)
(0.0603586, 0.0863895)
(0.0622393, 0.086494)
(0.0641785, 0.0865116)
(0.0661782, 0.0865994)
(0.0682402, 0.0864773)
(0.0703664, 0.0872011)
(0.0725589, 0.0868621)
(0.0748197, 0.0871455)
(0.0771509, 0.0871414)
(0.0795548, 0.0876634)
(0.0820336, 0.0876093)
(0.0845896, 0.0873968)
(0.0872253, 0.0877805)
(0.089943, 0.0879078)
(0.0927455, 0.0879736)
(0.0956352, 0.0881549)
(0.0986151, 0.0882195)
(0.101688, 0.0878135)
(0.104856, 0.0882579)
(0.108123, 0.0883514)
(0.111492, 0.0880517)
(0.114966, 0.0882119)
(0.118548, 0.0881739)
(0.122242, 0.088318)
(0.126051, 0.0885378)
(0.129978, 0.0884243)
(0.134028, 0.0885484)
(0.138204, 0.088515)
(0.14251, 0.088275)
(0.146951, 0.0885432)
(0.151529, 0.0885177)
(0.156251, 0.0882515)
(0.161119, 0.0883858)
(0.166139, 0.088348)
(0.171316, 0.0883857)
(0.176654, 0.0881593)
(0.182158, 0.0883255)
(0.187834, 0.0879516)
(0.193686, 0.0882505)
(0.199721, 0.0882903)
(0.205944, 0.0879512)
(0.212361, 0.0878003)
(0.218978, 0.088172)
(0.225801, 0.0874142)
(0.232836, 0.0875295)
(0.240091, 0.0873487)
(0.247572, 0.0872524)
(0.255286, 0.0871355)
(0.26324, 0.0866285)
(0.271442, 0.0866885)
(0.279899, 0.0865029)
(0.288621, 0.0866162)
(0.297613, 0.0862395)
(0.306886, 0.0861408)
(0.316448, 0.0857874)
(0.326308, 0.0855168)
(0.336476, 0.0853869)
(0.346959, 0.085355)
(0.35777, 0.0850876)
(0.368917, 0.0847335)
(0.380412, 0.0845963)
(0.392265, 0.0841995)
(0.404487, 0.0843111)
(0.41709, 0.0837696)
(0.430086, 0.0837665)
(0.443487, 0.0831262)
(0.457305, 0.0832207)
(0.471554, 0.0826193)
(0.486247, 0.0822305)
(0.501397, 0.0820852)
(0.51702, 0.0819221)
(0.533129, 0.0815639)
(0.54974, 0.0811298)
(0.566869, 0.0813429)
(0.584532, 0.0805828)
(0.602745, 0.0803979)
(0.621525, 0.079811)
(0.64089, 0.079753)
(0.660859, 0.0792987)
(0.68145, 0.0791016)
(0.702683, 0.0785665)
(0.724577, 0.0781597)
(0.747154, 0.0776428)
(0.770434, 0.0774063)
(0.794439, 0.0771361)
(0.819192, 0.0763985)
(0.844717, 0.0762556)
(0.871036, 0.0759516)
(0.898176, 0.0757056)
(0.926162, 0.0751594)
(0.955019, 0.0748911)
(0.984776, 0.0745397)
(1.01546, 0.0739596)
(1.0471, 0.0735481)
(1.07972, 0.0732301)
(1.11337, 0.072822)
(1.14806, 0.0719035)
(1.18383, 0.0718915)
(1.22071, 0.0715861)
(1.25875, 0.071233)
(1.29797, 0.0705894)
(1.33841, 0.0703491)
(1.38011, 0.069734)
(1.42312, 0.0693489)
(1.46746, 0.0689217)
(1.51318, 0.0685189)
(1.56033, 0.0681742)
(1.60895, 0.0678019)
(1.65908, 0.0670145)
(1.71077, 0.0665977)
(1.76408, 0.0662063)
(1.81904, 0.0657246)
(1.87572, 0.0655652)
(1.93416, 0.0652335)
(1.99443, 0.0646221)
(2.05657, 0.0641062)
(2.12065, 0.0636903)
(2.18672, 0.0632307)
(2.25486, 0.0630104)
(2.32512, 0.0623503)
(2.39756, 0.0619005)
(2.47227, 0.0618084)
(2.5493, 0.0611658)
(2.62873, 0.0606686)
(2.71063, 0.0600093)
(2.79509, 0.0596178)
(2.88218, 0.0590777)
(2.97198, 0.0587809)
(3.06459, 0.0582994)
(3.16007, 0.058034)
(3.25853, 0.0576719)
(3.36006, 0.0569223)
(3.46476, 0.0565289)
(3.57271, 0.0561304)
(3.68403, 0.05591)
(3.79882, 0.0554993)
(3.91718, 0.0546334)
(4.03923, 0.0544045)
(4.16509, 0.0536334)
(4.29487, 0.0531498)
(4.42869, 0.0530463)
(4.56668, 0.0527723)
(4.70896, 0.0523389)
(4.85569, 0.0516894)
(5.00698, 0.0513434)
(5.16299, 0.0509065)
(5.32386, 0.0506561)
(5.48974, 0.05036)
(5.66079, 0.0499158)
(5.83717, 0.0491624)
(6.01904, 0.0489209)
(6.20658, 0.0485084)
(6.39997, 0.0480187)
(6.59938, 0.0484284)
(6.805, 0.0477302)
(7.01704, 0.0472335)
(7.23567, 0.0466602)
(7.46112, 0.0462471)
(7.6936, 0.0455269)
(7.93332, 0.0453792)
(8.1805, 0.0447318)
(8.43539, 0.0445713)
(8.69822, 0.0440761)
(8.96924, 0.0438693)
(9.24871, 0.0429733)
(9.53688, 0.0427512)
(9.83403, 0.0421118)
(10.1404, 0.0421619)
(10.4564, 0.0420516)
(10.7822, 0.0412875)
(11.1181, 0.0410131)
(11.4646, 0.0408836)
(11.8218, 0.0399348)
(12.1901, 0.0401379)
(12.5699, 0.0390469)
(12.9616, 0.0391822)
(13.3655, 0.0382959)
(13.7819, 0.038236)
(14.2113, 0.0381133)
(14.6541, 0.0375546)
(15.1107, 0.0368367)
(15.5815, 0.0369929)
(16.067, 0.0366788)
(16.5676, 0.0364841)
(17.0839, 0.0358895)
(17.6162, 0.035467)
(18.165, 0.035686)
(18.731, 0.0350091)
(19.3147, 0.0347757)
(19.9165, 0.0343189)
(20.537, 0.0339288)
(21.1769, 0.0337161)
(21.8368, 0.0333688)
(22.5171, 0.0330233)
(23.2187, 0.0326816)
(23.9422, 0.0324975)
(24.6882, 0.032238)
(25.4574, 0.0315529)
(26.2506, 0.0312793)
(27.0685, 0.0317131)
(27.9119, 0.0310833)
(28.7816, 0.0307947)
(29.6784, 0.0306067)
(30.6031, 0.0301975)
(31.5567, 0.0300863)
(32.5399, 0.0294387)
(33.5538, 0.028986)
(34.5993, 0.0290797)
(35.6773, 0.0284861)
(36.789, 0.0281476)
(37.9352, 0.0283749)
(39.1172, 0.0274731)
(40.336, 0.0277519)
(41.5928, 0.0276036)
(42.8888, 0.026971)
(44.2251, 0.0270283)
(45.6031, 0.0267754)
(47.024, 0.0260261)
(48.4892, 0.0261568)
(50, 0.0256204)
}

\newcommand{\genOdePlotExpCifaruNcsnppErrEulerBLo}{%
(0.02, 0.0753372)
(0.0206232, 0.0750695)
(0.0212657, 0.075451)
(0.0219283, 0.0754584)
(0.0226116, 0.0759389)
(0.0233161, 0.0765674)
(0.0240426, 0.075974)
(0.0247917, 0.0766123)
(0.0255642, 0.0770082)
(0.0263607, 0.0768769)
(0.0271821, 0.0781077)
(0.028029, 0.0779032)
(0.0289023, 0.077672)
(0.0298029, 0.0779355)
(0.0307315, 0.0780501)
(0.031689, 0.0775597)
(0.0326764, 0.0786957)
(0.0336945, 0.0788123)
(0.0347444, 0.0787419)
(0.035827, 0.0797564)
(0.0369432, 0.0796998)
(0.0380943, 0.0800436)
(0.0392813, 0.0799475)
(0.0405052, 0.0801846)
(0.0417673, 0.080255)
(0.0430687, 0.0811108)
(0.0444106, 0.0810612)
(0.0457943, 0.0818633)
(0.0472212, 0.080883)
(0.0486925, 0.0816613)
(0.0502097, 0.0819445)
(0.0517741, 0.0823362)
(0.0533873, 0.0825293)
(0.0550508, 0.0825093)
(0.056766, 0.0823159)
(0.0585348, 0.0828985)
(0.0603586, 0.0833812)
(0.0622393, 0.083148)
(0.0641785, 0.0833354)
(0.0661782, 0.0834977)
(0.0682402, 0.0837197)
(0.0703664, 0.0841667)
(0.0725589, 0.084048)
(0.0748197, 0.0843277)
(0.0771509, 0.0843595)
(0.0795548, 0.0849029)
(0.0820336, 0.0851574)
(0.0845896, 0.0846761)
(0.0872253, 0.0851974)
(0.089943, 0.0851999)
(0.0927455, 0.0852059)
(0.0956352, 0.0852166)
(0.0986151, 0.0855761)
(0.101688, 0.0851556)
(0.104856, 0.0856994)
(0.108123, 0.0855029)
(0.111492, 0.0853297)
(0.114966, 0.0855504)
(0.118548, 0.0853159)
(0.122242, 0.0859628)
(0.126051, 0.086074)
(0.129978, 0.0859481)
(0.134028, 0.086366)
(0.138204, 0.0865458)
(0.14251, 0.0859512)
(0.146951, 0.0863349)
(0.151529, 0.0863676)
(0.156251, 0.0858398)
(0.161119, 0.0859966)
(0.166139, 0.0860775)
(0.171316, 0.0861685)
(0.176654, 0.0862014)
(0.182158, 0.0861335)
(0.187834, 0.085961)
(0.193686, 0.0862921)
(0.199721, 0.0862294)
(0.205944, 0.0858361)
(0.212361, 0.0859327)
(0.218978, 0.0861442)
(0.225801, 0.0854502)
(0.232836, 0.0854723)
(0.240091, 0.0854262)
(0.247572, 0.0854957)
(0.255286, 0.0854452)
(0.26324, 0.0846745)
(0.271442, 0.0848006)
(0.279899, 0.0848603)
(0.288621, 0.0850168)
(0.297613, 0.0846182)
(0.306886, 0.0844842)
(0.316448, 0.0840922)
(0.326308, 0.0838974)
(0.336476, 0.0837788)
(0.346959, 0.0837295)
(0.35777, 0.0836327)
(0.368917, 0.0832666)
(0.380412, 0.0831701)
(0.392265, 0.0827019)
(0.404487, 0.0829241)
(0.41709, 0.0822573)
(0.430086, 0.0824163)
(0.443487, 0.081663)
(0.457305, 0.0819494)
(0.471554, 0.0812743)
(0.486247, 0.0809835)
(0.501397, 0.0807555)
(0.51702, 0.0807066)
(0.533129, 0.0802904)
(0.54974, 0.0798565)
(0.566869, 0.0801341)
(0.584532, 0.0793184)
(0.602745, 0.0792377)
(0.621525, 0.0786205)
(0.64089, 0.0785939)
(0.660859, 0.0781111)
(0.68145, 0.0779434)
(0.702683, 0.0774573)
(0.724577, 0.0769146)
(0.747154, 0.0765424)
(0.770434, 0.0763402)
(0.794439, 0.0759149)
(0.819192, 0.0752964)
(0.844717, 0.0750654)
(0.871036, 0.0748476)
(0.898176, 0.0747562)
(0.926162, 0.0740939)
(0.955019, 0.0738376)
(0.984776, 0.0735647)
(1.01546, 0.0729696)
(1.0471, 0.0725133)
(1.07972, 0.0723123)
(1.11337, 0.0718158)
(1.14806, 0.0709901)
(1.18383, 0.0708993)
(1.22071, 0.0706846)
(1.25875, 0.0702657)
(1.29797, 0.0697165)
(1.33841, 0.0694073)
(1.38011, 0.0689348)
(1.42312, 0.0684961)
(1.46746, 0.0681015)
(1.51318, 0.067638)
(1.56033, 0.0673927)
(1.60895, 0.0669589)
(1.65908, 0.0662402)
(1.71077, 0.0658066)
(1.76408, 0.0654481)
(1.81904, 0.0649314)
(1.87572, 0.0648153)
(1.93416, 0.0645009)
(1.99443, 0.0638272)
(2.05657, 0.06325)
(2.12065, 0.0628555)
(2.18672, 0.0623844)
(2.25486, 0.0622209)
(2.32512, 0.0615617)
(2.39756, 0.0611117)
(2.47227, 0.0610414)
(2.5493, 0.0604253)
(2.62873, 0.0599296)
(2.71063, 0.0592909)
(2.79509, 0.0589371)
(2.88218, 0.0583064)
(2.97198, 0.0581349)
(3.06459, 0.0574695)
(3.16007, 0.0572929)
(3.25853, 0.0570889)
(3.36006, 0.056238)
(3.46476, 0.0558155)
(3.57271, 0.0553417)
(3.68403, 0.055116)
(3.79882, 0.0548492)
(3.91718, 0.0539207)
(4.03923, 0.0536642)
(4.16509, 0.0530877)
(4.29487, 0.0525008)
(4.42869, 0.0524529)
(4.56668, 0.052164)
(4.70896, 0.0517717)
(4.85569, 0.0510313)
(5.00698, 0.0505621)
(5.16299, 0.0501757)
(5.32386, 0.0499478)
(5.48974, 0.0495141)
(5.66079, 0.0491849)
(5.83717, 0.0482926)
(6.01904, 0.048015)
(6.20658, 0.0475843)
(6.39997, 0.0469815)
(6.59938, 0.0480479)
(6.805, 0.0474373)
(7.01704, 0.0469087)
(7.23567, 0.046171)
(7.46112, 0.0455406)
(7.6936, 0.0449434)
(7.93332, 0.0448349)
(8.1805, 0.0441826)
(8.43539, 0.044038)
(8.69822, 0.0433638)
(8.96924, 0.0433251)
(9.24871, 0.042394)
(9.53688, 0.0423506)
(9.83403, 0.0415021)
(10.1404, 0.0416649)
(10.4564, 0.0415117)
(10.7822, 0.0406135)
(11.1181, 0.0403774)
(11.4646, 0.0404267)
(11.8218, 0.0393824)
(12.1901, 0.0396415)
(12.5699, 0.0386409)
(12.9616, 0.0385783)
(13.3655, 0.0378518)
(13.7819, 0.037661)
(14.2113, 0.0374307)
(14.6541, 0.0369607)
(15.1107, 0.0363778)
(15.5815, 0.0366739)
(16.067, 0.0362641)
(16.5676, 0.0360638)
(17.0839, 0.0355952)
(17.6162, 0.0348146)
(18.165, 0.0351609)
(18.731, 0.0344424)
(19.3147, 0.0343587)
(19.9165, 0.0337017)
(20.537, 0.0335626)
(21.1769, 0.0331978)
(21.8368, 0.0328081)
(22.5171, 0.0326394)
(23.2187, 0.0322764)
(23.9422, 0.0319918)
(24.6882, 0.0319463)
(25.4574, 0.0311041)
(26.2506, 0.0306861)
(27.0685, 0.0313827)
(27.9119, 0.0305326)
(28.7816, 0.030417)
(29.6784, 0.0301291)
(30.6031, 0.0298614)
(31.5567, 0.0295542)
(32.5399, 0.0290492)
(33.5538, 0.028722)
(34.5993, 0.0285764)
(35.6773, 0.0281973)
(36.789, 0.0277782)
(37.9352, 0.0279902)
(39.1172, 0.0270123)
(40.336, 0.0273946)
(41.5928, 0.0272767)
(42.8888, 0.0267295)
(44.2251, 0.0266285)
(45.6031, 0.0264141)
(47.024, 0.0256527)
(48.4892, 0.0257905)
(50, 0.025242)
}

\newcommand{\genOdePlotExpCifaruNcsnppErrEulerBHi}{%
(0.02, 0.0829057)
(0.0206232, 0.0831198)
(0.0212657, 0.0831287)
(0.0219283, 0.0827297)
(0.0226116, 0.0836022)
(0.0233161, 0.0836122)
(0.0240426, 0.0838402)
(0.0247917, 0.0842709)
(0.0255642, 0.0840665)
(0.0263607, 0.0851504)
(0.0271821, 0.0851985)
(0.028029, 0.0856015)
(0.0289023, 0.0854222)
(0.0298029, 0.0856174)
(0.0307315, 0.0854588)
(0.031689, 0.0858975)
(0.0326764, 0.0860228)
(0.0336945, 0.0863361)
(0.0347444, 0.0858319)
(0.035827, 0.0867193)
(0.0369432, 0.0868544)
(0.0380943, 0.0876233)
(0.0392813, 0.0876262)
(0.0405052, 0.0874259)
(0.0417673, 0.0876482)
(0.0430687, 0.0880981)
(0.0444106, 0.0877666)
(0.0457943, 0.0885595)
(0.0472212, 0.0880933)
(0.0486925, 0.0883435)
(0.0502097, 0.0890483)
(0.0517741, 0.0887359)
(0.0533873, 0.0889302)
(0.0550508, 0.088804)
(0.056766, 0.0890117)
(0.0585348, 0.0889441)
(0.0603586, 0.0893977)
(0.0622393, 0.08984)
(0.0641785, 0.0896878)
(0.0661782, 0.0897011)
(0.0682402, 0.0892349)
(0.0703664, 0.0902355)
(0.0725589, 0.0896762)
(0.0748197, 0.0899633)
(0.0771509, 0.0899232)
(0.0795548, 0.0904239)
(0.0820336, 0.0900613)
(0.0845896, 0.0901176)
(0.0872253, 0.0903637)
(0.089943, 0.0906158)
(0.0927455, 0.0907414)
(0.0956352, 0.0910932)
(0.0986151, 0.090863)
(0.101688, 0.0904714)
(0.104856, 0.0908165)
(0.108123, 0.0911999)
(0.111492, 0.0907737)
(0.114966, 0.0908733)
(0.118548, 0.0910319)
(0.122242, 0.0906731)
(0.126051, 0.0910016)
(0.129978, 0.0909005)
(0.134028, 0.0907309)
(0.138204, 0.0904843)
(0.14251, 0.0905988)
(0.146951, 0.0907515)
(0.151529, 0.0906678)
(0.156251, 0.0906633)
(0.161119, 0.090775)
(0.166139, 0.0906186)
(0.171316, 0.0906028)
(0.176654, 0.0901172)
(0.182158, 0.0905174)
(0.187834, 0.0899421)
(0.193686, 0.0902089)
(0.199721, 0.0903512)
(0.205944, 0.0900663)
(0.212361, 0.0896678)
(0.218978, 0.0901998)
(0.225801, 0.0893782)
(0.232836, 0.0895866)
(0.240091, 0.0892712)
(0.247572, 0.0890092)
(0.255286, 0.0888259)
(0.26324, 0.0885826)
(0.271442, 0.0885764)
(0.279899, 0.0881455)
(0.288621, 0.0882156)
(0.297613, 0.0878608)
(0.306886, 0.0877974)
(0.316448, 0.0874826)
(0.326308, 0.0871362)
(0.336476, 0.086995)
(0.346959, 0.0869806)
(0.35777, 0.0865425)
(0.368917, 0.0862003)
(0.380412, 0.0860226)
(0.392265, 0.085697)
(0.404487, 0.085698)
(0.41709, 0.085282)
(0.430086, 0.0851166)
(0.443487, 0.0845894)
(0.457305, 0.084492)
(0.471554, 0.0839643)
(0.486247, 0.0834776)
(0.501397, 0.083415)
(0.51702, 0.0831376)
(0.533129, 0.0828373)
(0.54974, 0.0824031)
(0.566869, 0.0825518)
(0.584532, 0.0818473)
(0.602745, 0.081558)
(0.621525, 0.0810015)
(0.64089, 0.080912)
(0.660859, 0.0804863)
(0.68145, 0.0802598)
(0.702683, 0.0796758)
(0.724577, 0.0794048)
(0.747154, 0.0787433)
(0.770434, 0.0784724)
(0.794439, 0.0783573)
(0.819192, 0.0775006)
(0.844717, 0.0774459)
(0.871036, 0.0770556)
(0.898176, 0.076655)
(0.926162, 0.0762248)
(0.955019, 0.0759446)
(0.984776, 0.0755146)
(1.01546, 0.0749495)
(1.0471, 0.0745829)
(1.07972, 0.0741478)
(1.11337, 0.0738282)
(1.14806, 0.0728169)
(1.18383, 0.0728836)
(1.22071, 0.0724877)
(1.25875, 0.0722003)
(1.29797, 0.0714623)
(1.33841, 0.0712909)
(1.38011, 0.0705333)
(1.42312, 0.0702017)
(1.46746, 0.0697419)
(1.51318, 0.0693999)
(1.56033, 0.0689556)
(1.60895, 0.0686449)
(1.65908, 0.0677888)
(1.71077, 0.0673889)
(1.76408, 0.0669645)
(1.81904, 0.0665179)
(1.87572, 0.0663152)
(1.93416, 0.065966)
(1.99443, 0.0654171)
(2.05657, 0.0649624)
(2.12065, 0.0645252)
(2.18672, 0.064077)
(2.25486, 0.0637999)
(2.32512, 0.0631388)
(2.39756, 0.0626894)
(2.47227, 0.0625754)
(2.5493, 0.0619064)
(2.62873, 0.0614075)
(2.71063, 0.0607277)
(2.79509, 0.0602984)
(2.88218, 0.0598489)
(2.97198, 0.059427)
(3.06459, 0.0591293)
(3.16007, 0.0587751)
(3.25853, 0.0582549)
(3.36006, 0.0576065)
(3.46476, 0.0572423)
(3.57271, 0.0569191)
(3.68403, 0.0567039)
(3.79882, 0.0561494)
(3.91718, 0.0553461)
(4.03923, 0.0551447)
(4.16509, 0.0541791)
(4.29487, 0.0537987)
(4.42869, 0.0536398)
(4.56668, 0.0533806)
(4.70896, 0.0529061)
(4.85569, 0.0523475)
(5.00698, 0.0521247)
(5.16299, 0.0516373)
(5.32386, 0.0513645)
(5.48974, 0.0512058)
(5.66079, 0.0506468)
(5.83717, 0.0500321)
(6.01904, 0.0498268)
(6.20658, 0.0494324)
(6.39997, 0.0490558)
(6.59938, 0.0488089)
(6.805, 0.0480231)
(7.01704, 0.0475584)
(7.23567, 0.0471494)
(7.46112, 0.0469536)
(7.6936, 0.0461103)
(7.93332, 0.0459236)
(8.1805, 0.0452809)
(8.43539, 0.0451045)
(8.69822, 0.0447885)
(8.96924, 0.0444135)
(9.24871, 0.0435526)
(9.53688, 0.0431517)
(9.83403, 0.0427214)
(10.1404, 0.0426589)
(10.4564, 0.0425914)
(10.7822, 0.0419615)
(11.1181, 0.0416489)
(11.4646, 0.0413405)
(11.8218, 0.0404873)
(12.1901, 0.0406343)
(12.5699, 0.039453)
(12.9616, 0.0397861)
(13.3655, 0.03874)
(13.7819, 0.0388109)
(14.2113, 0.0387959)
(14.6541, 0.0381485)
(15.1107, 0.0372956)
(15.5815, 0.0373118)
(16.067, 0.0370936)
(16.5676, 0.0369045)
(17.0839, 0.0361838)
(17.6162, 0.0361194)
(18.165, 0.0362111)
(18.731, 0.0355758)
(19.3147, 0.0351926)
(19.9165, 0.0349361)
(20.537, 0.034295)
(21.1769, 0.0342343)
(21.8368, 0.0339296)
(22.5171, 0.0334072)
(23.2187, 0.0330868)
(23.9422, 0.0330031)
(24.6882, 0.0325297)
(25.4574, 0.0320016)
(26.2506, 0.0318725)
(27.0685, 0.0320434)
(27.9119, 0.031634)
(28.7816, 0.0311723)
(29.6784, 0.0310844)
(30.6031, 0.0305336)
(31.5567, 0.0306184)
(32.5399, 0.0298282)
(33.5538, 0.0292501)
(34.5993, 0.0295829)
(35.6773, 0.0287749)
(36.789, 0.0285171)
(37.9352, 0.0287597)
(39.1172, 0.027934)
(40.336, 0.0281093)
(41.5928, 0.0279305)
(42.8888, 0.0272124)
(44.2251, 0.0274281)
(45.6031, 0.0271368)
(47.024, 0.0263994)
(48.4892, 0.0265232)
(50, 0.0259988)
}

\newcommand{\genOdePlotExpCifaruNcsnppErrEulerC}{%
(0.02, 0.0179517)
(0.0206232, 0.01792)
(0.0212657, 0.0180692)
(0.0219283, 0.0182005)
(0.0226116, 0.0185827)
(0.0233161, 0.0183229)
(0.0240426, 0.0186053)
(0.0247917, 0.0188405)
(0.0255642, 0.0189543)
(0.0263607, 0.0191399)
(0.0271821, 0.0192719)
(0.028029, 0.0195352)
(0.0289023, 0.0196952)
(0.0298029, 0.0196674)
(0.0307315, 0.0198081)
(0.031689, 0.0198886)
(0.0326764, 0.0202275)
(0.0336945, 0.0203134)
(0.0347444, 0.0204805)
(0.035827, 0.0204292)
(0.0369432, 0.02073)
(0.0380943, 0.0208894)
(0.0392813, 0.0211142)
(0.0405052, 0.0212467)
(0.0417673, 0.0213319)
(0.0430687, 0.0216725)
(0.0444106, 0.0216812)
(0.0457943, 0.0218157)
(0.0472212, 0.021848)
(0.0486925, 0.0220556)
(0.0502097, 0.0221989)
(0.0517741, 0.0225253)
(0.0533873, 0.0225573)
(0.0550508, 0.0227333)
(0.056766, 0.0230067)
(0.0585348, 0.02307)
(0.0603586, 0.0232686)
(0.0622393, 0.0234639)
(0.0641785, 0.0234611)
(0.0661782, 0.0237251)
(0.0682402, 0.0238931)
(0.0703664, 0.0240991)
(0.0725589, 0.0240699)
(0.0748197, 0.0243371)
(0.0771509, 0.0244571)
(0.0795548, 0.0245808)
(0.0820336, 0.0248086)
(0.0845896, 0.0249134)
(0.0872253, 0.0248278)
(0.089943, 0.025285)
(0.0927455, 0.0252933)
(0.0956352, 0.0253194)
(0.0986151, 0.0255375)
(0.101688, 0.0256772)
(0.104856, 0.0259727)
(0.108123, 0.0260391)
(0.111492, 0.0262102)
(0.114966, 0.0264323)
(0.118548, 0.0265782)
(0.122242, 0.0265927)
(0.126051, 0.0269033)
(0.129978, 0.0269499)
(0.134028, 0.0271183)
(0.138204, 0.0272512)
(0.14251, 0.0273639)
(0.146951, 0.0276318)
(0.151529, 0.0276496)
(0.156251, 0.0278711)
(0.161119, 0.0279295)
(0.166139, 0.0279952)
(0.171316, 0.0282063)
(0.176654, 0.0282879)
(0.182158, 0.0284791)
(0.187834, 0.0285416)
(0.193686, 0.0288466)
(0.199721, 0.0288617)
(0.205944, 0.0289872)
(0.212361, 0.0292124)
(0.218978, 0.0292318)
(0.225801, 0.0293279)
(0.232836, 0.029602)
(0.240091, 0.029736)
(0.247572, 0.0296172)
(0.255286, 0.0298748)
(0.26324, 0.0300823)
(0.271442, 0.0301031)
(0.279899, 0.0302217)
(0.288621, 0.0302407)
(0.297613, 0.0304488)
(0.306886, 0.0307236)
(0.316448, 0.0307434)
(0.326308, 0.0308014)
(0.336476, 0.0308859)
(0.346959, 0.0311286)
(0.35777, 0.0312382)
(0.368917, 0.0313638)
(0.380412, 0.031419)
(0.392265, 0.0315076)
(0.404487, 0.0316114)
(0.41709, 0.0316852)
(0.430086, 0.0318066)
(0.443487, 0.0318497)
(0.457305, 0.0318176)
(0.471554, 0.0321007)
(0.486247, 0.0321923)
(0.501397, 0.0321555)
(0.51702, 0.0324971)
(0.533129, 0.0325615)
(0.54974, 0.032486)
(0.566869, 0.0327119)
(0.584532, 0.0328298)
(0.602745, 0.0329954)
(0.621525, 0.032862)
(0.64089, 0.0330787)
(0.660859, 0.0331376)
(0.68145, 0.0332595)
(0.702683, 0.0335536)
(0.724577, 0.033397)
(0.747154, 0.0335849)
(0.770434, 0.0336045)
(0.794439, 0.033684)
(0.819192, 0.0337159)
(0.844717, 0.0339413)
(0.871036, 0.033951)
(0.898176, 0.0339357)
(0.926162, 0.0342324)
(0.955019, 0.0341401)
(0.984776, 0.0343855)
(1.01546, 0.0344711)
(1.0471, 0.0346108)
(1.07972, 0.0344299)
(1.11337, 0.0345579)
(1.14806, 0.0345104)
(1.18383, 0.034694)
(1.22071, 0.0348203)
(1.25875, 0.0347985)
(1.29797, 0.0346124)
(1.33841, 0.0350148)
(1.38011, 0.0350003)
(1.42312, 0.0352818)
(1.46746, 0.0352775)
(1.51318, 0.0353935)
(1.56033, 0.0354638)
(1.60895, 0.035478)
(1.65908, 0.0354456)
(1.71077, 0.0355223)
(1.76408, 0.0355912)
(1.81904, 0.0354401)
(1.87572, 0.0357027)
(1.93416, 0.0356914)
(1.99443, 0.0357168)
(2.05657, 0.035854)
(2.12065, 0.036064)
(2.18672, 0.0361182)
(2.25486, 0.0359275)
(2.32512, 0.0360701)
(2.39756, 0.0361933)
(2.47227, 0.0361402)
(2.5493, 0.0363675)
(2.62873, 0.0364251)
(2.71063, 0.0362627)
(2.79509, 0.0363488)
(2.88218, 0.0364248)
(2.97198, 0.0364447)
(3.06459, 0.0367125)
(3.16007, 0.036594)
(3.25853, 0.0365732)
(3.36006, 0.0365471)
(3.46476, 0.036808)
(3.57271, 0.0368151)
(3.68403, 0.0366457)
(3.79882, 0.0370293)
(3.91718, 0.0371048)
(4.03923, 0.0368773)
(4.16509, 0.0369017)
(4.29487, 0.0367566)
(4.42869, 0.0369461)
(4.56668, 0.037143)
(4.70896, 0.0373133)
(4.85569, 0.0372329)
(5.00698, 0.0371149)
(5.16299, 0.0373487)
(5.32386, 0.0376688)
(5.48974, 0.0375901)
(5.66079, 0.0374039)
(5.83717, 0.0375044)
(6.01904, 0.0372666)
(6.20658, 0.0375767)
(6.39997, 0.0375245)
(6.59938, 0.0377854)
(6.805, 0.0381946)
(7.01704, 0.0380514)
(7.23567, 0.0378494)
(7.46112, 0.0375327)
(7.6936, 0.0378752)
(7.93332, 0.0377607)
(8.1805, 0.0377413)
(8.43539, 0.0379284)
(8.69822, 0.0379229)
(8.96924, 0.0379287)
(9.24871, 0.0376662)
(9.53688, 0.0376323)
(9.83403, 0.0376684)
(10.1404, 0.0377695)
(10.4564, 0.038137)
(10.7822, 0.0380021)
(11.1181, 0.0380515)
(11.4646, 0.0383378)
(11.8218, 0.0378941)
(12.1901, 0.0381867)
(12.5699, 0.0375582)
(12.9616, 0.0381843)
(13.3655, 0.0376511)
(13.7819, 0.038096)
(14.2113, 0.0383322)
(14.6541, 0.0380817)
(15.1107, 0.0376532)
(15.5815, 0.038195)
(16.067, 0.0382251)
(16.5676, 0.0383644)
(17.0839, 0.0380068)
(17.6162, 0.0379922)
(18.165, 0.0388415)
(18.731, 0.0381497)
(19.3147, 0.0382934)
(19.9165, 0.0386714)
(20.537, 0.0382415)
(21.1769, 0.0381902)
(21.8368, 0.0385053)
(22.5171, 0.0385928)
(23.2187, 0.0384934)
(23.9422, 0.0387367)
(24.6882, 0.0385911)
(25.4574, 0.0383454)
(26.2506, 0.038742)
(27.0685, 0.0385129)
(27.9119, 0.0393844)
(28.7816, 0.0381208)
(29.6784, 0.0387582)
(30.6031, 0.0386189)
(31.5567, 0.0383091)
(32.5399, 0.0386887)
(33.5538, 0.0392307)
(34.5993, 0.0385163)
(35.6773, 0.038837)
(36.789, 0.0381709)
(37.9352, 0.0388872)
(39.1172, 0.0387955)
(40.336, 0.0384482)
(41.5928, 0.0385491)
(42.8888, 0.0392782)
(44.2251, 0.0383546)
(45.6031, 0.0392836)
(47.024, 0.0380221)
(48.4892, 0.0394265)
(50, 0.0395386)
}

\newcommand{\genOdePlotExpCifaruNcsnppErrEulerCLo}{%
(0.02, 0.0164839)
(0.0206232, 0.0164613)
(0.0212657, 0.0166302)
(0.0219283, 0.0168078)
(0.0226116, 0.0174475)
(0.0233161, 0.0168455)
(0.0240426, 0.0172184)
(0.0247917, 0.0175951)
(0.0255642, 0.0176182)
(0.0263607, 0.0178862)
(0.0271821, 0.0178568)
(0.028029, 0.0182439)
(0.0289023, 0.0183438)
(0.0298029, 0.018369)
(0.0307315, 0.0185903)
(0.031689, 0.0184975)
(0.0326764, 0.0189453)
(0.0336945, 0.0190335)
(0.0347444, 0.0192001)
(0.035827, 0.0191423)
(0.0369432, 0.0194876)
(0.0380943, 0.0196089)
(0.0392813, 0.0198198)
(0.0405052, 0.0200004)
(0.0417673, 0.0199725)
(0.0430687, 0.0204444)
(0.0444106, 0.0204874)
(0.0457943, 0.0206876)
(0.0472212, 0.0206582)
(0.0486925, 0.0207971)
(0.0502097, 0.0209301)
(0.0517741, 0.0214177)
(0.0533873, 0.0213576)
(0.0550508, 0.0214459)
(0.056766, 0.0218889)
(0.0585348, 0.0219388)
(0.0603586, 0.0221086)
(0.0622393, 0.02232)
(0.0641785, 0.0222564)
(0.0661782, 0.0225931)
(0.0682402, 0.0227328)
(0.0703664, 0.0229964)
(0.0725589, 0.0229819)
(0.0748197, 0.023435)
(0.0771509, 0.0233303)
(0.0795548, 0.0235171)
(0.0820336, 0.0237596)
(0.0845896, 0.0238712)
(0.0872253, 0.0238195)
(0.089943, 0.0243366)
(0.0927455, 0.0242997)
(0.0956352, 0.024271)
(0.0986151, 0.0245187)
(0.101688, 0.0246714)
(0.104856, 0.0250405)
(0.108123, 0.0250074)
(0.111492, 0.0252637)
(0.114966, 0.0254899)
(0.118548, 0.0256679)
(0.122242, 0.025555)
(0.126051, 0.0260382)
(0.129978, 0.0261147)
(0.134028, 0.0262471)
(0.138204, 0.0263674)
(0.14251, 0.0265303)
(0.146951, 0.0268232)
(0.151529, 0.0268631)
(0.156251, 0.0270114)
(0.161119, 0.0271123)
(0.166139, 0.0272028)
(0.171316, 0.0274358)
(0.176654, 0.0274816)
(0.182158, 0.02767)
(0.187834, 0.0277817)
(0.193686, 0.0281225)
(0.199721, 0.0280115)
(0.205944, 0.0282988)
(0.212361, 0.0285079)
(0.218978, 0.0285078)
(0.225801, 0.028606)
(0.232836, 0.0288801)
(0.240091, 0.0290904)
(0.247572, 0.0288961)
(0.255286, 0.0291752)
(0.26324, 0.029418)
(0.271442, 0.0293948)
(0.279899, 0.0295434)
(0.288621, 0.0295377)
(0.297613, 0.029823)
(0.306886, 0.0300224)
(0.316448, 0.0300699)
(0.326308, 0.0301991)
(0.336476, 0.0302425)
(0.346959, 0.0305143)
(0.35777, 0.0306056)
(0.368917, 0.0307748)
(0.380412, 0.0308055)
(0.392265, 0.0309023)
(0.404487, 0.0310635)
(0.41709, 0.0311221)
(0.430086, 0.0312092)
(0.443487, 0.0312966)
(0.457305, 0.0312479)
(0.471554, 0.0315887)
(0.486247, 0.0316822)
(0.501397, 0.0315793)
(0.51702, 0.0319485)
(0.533129, 0.0320717)
(0.54974, 0.0319277)
(0.566869, 0.032186)
(0.584532, 0.032322)
(0.602745, 0.0324257)
(0.621525, 0.0323921)
(0.64089, 0.0326111)
(0.660859, 0.0326865)
(0.68145, 0.0327832)
(0.702683, 0.0330116)
(0.724577, 0.0328918)
(0.747154, 0.0331062)
(0.770434, 0.0331454)
(0.794439, 0.0332148)
(0.819192, 0.0332341)
(0.844717, 0.0334293)
(0.871036, 0.0334514)
(0.898176, 0.0334273)
(0.926162, 0.0337311)
(0.955019, 0.0335781)
(0.984776, 0.033866)
(1.01546, 0.0340192)
(1.0471, 0.034173)
(1.07972, 0.0339906)
(1.11337, 0.034091)
(1.14806, 0.0340869)
(1.18383, 0.034176)
(1.22071, 0.0343805)
(1.25875, 0.0343256)
(1.29797, 0.0341526)
(1.33841, 0.0345389)
(1.38011, 0.034597)
(1.42312, 0.034848)
(1.46746, 0.034885)
(1.51318, 0.034983)
(1.56033, 0.0349934)
(1.60895, 0.0350606)
(1.65908, 0.0350087)
(1.71077, 0.0350955)
(1.76408, 0.0351913)
(1.81904, 0.0350088)
(1.87572, 0.0352429)
(1.93416, 0.0352824)
(1.99443, 0.035326)
(2.05657, 0.0354617)
(2.12065, 0.0356812)
(2.18672, 0.0355972)
(2.25486, 0.0354321)
(2.32512, 0.0355839)
(2.39756, 0.0356462)
(2.47227, 0.0356662)
(2.5493, 0.0358822)
(2.62873, 0.0359787)
(2.71063, 0.0358253)
(2.79509, 0.0359057)
(2.88218, 0.036039)
(2.97198, 0.0359952)
(3.06459, 0.0363169)
(3.16007, 0.0360917)
(3.25853, 0.0361263)
(3.36006, 0.0361359)
(3.46476, 0.0363965)
(3.57271, 0.0363114)
(3.68403, 0.0361517)
(3.79882, 0.0365233)
(3.91718, 0.0365896)
(4.03923, 0.0364018)
(4.16509, 0.0364448)
(4.29487, 0.036395)
(4.42869, 0.0365046)
(4.56668, 0.036681)
(4.70896, 0.0369227)
(4.85569, 0.0367659)
(5.00698, 0.0366637)
(5.16299, 0.0368422)
(5.32386, 0.0371606)
(5.48974, 0.0369814)
(5.66079, 0.0367207)
(5.83717, 0.0368936)
(6.01904, 0.0365705)
(6.20658, 0.0368783)
(6.39997, 0.0367611)
(6.59938, 0.0372699)
(6.805, 0.0379137)
(7.01704, 0.0378019)
(7.23567, 0.0375131)
(7.46112, 0.0368576)
(7.6936, 0.0374044)
(7.93332, 0.0373141)
(8.1805, 0.0372757)
(8.43539, 0.0373991)
(8.69822, 0.0372555)
(8.96924, 0.0374461)
(9.24871, 0.0372271)
(9.53688, 0.0372232)
(9.83403, 0.0371095)
(10.1404, 0.0372846)
(10.4564, 0.037675)
(10.7822, 0.0374249)
(11.1181, 0.0375181)
(11.4646, 0.0379154)
(11.8218, 0.0374081)
(12.1901, 0.0377641)
(12.5699, 0.0371095)
(12.9616, 0.0374774)
(13.3655, 0.0371698)
(13.7819, 0.0375082)
(14.2113, 0.0375867)
(14.6541, 0.0374345)
(15.1107, 0.0372364)
(15.5815, 0.0378614)
(16.067, 0.0377932)
(16.5676, 0.037947)
(17.0839, 0.0376836)
(17.6162, 0.0372624)
(18.165, 0.0382946)
(18.731, 0.0375577)
(19.3147, 0.0379083)
(19.9165, 0.0380789)
(20.537, 0.0378666)
(21.1769, 0.0376379)
(21.8368, 0.0378535)
(22.5171, 0.0380993)
(23.2187, 0.0379154)
(23.9422, 0.0381291)
(24.6882, 0.0382032)
(25.4574, 0.0378963)
(26.2506, 0.0379813)
(27.0685, 0.0380821)
(27.9119, 0.0387164)
(28.7816, 0.037673)
(29.6784, 0.0383088)
(30.6031, 0.0381333)
(31.5567, 0.0375856)
(32.5399, 0.0381648)
(33.5538, 0.0388603)
(34.5993, 0.0379135)
(35.6773, 0.038384)
(36.789, 0.0376119)
(37.9352, 0.0383012)
(39.1172, 0.0382135)
(40.336, 0.0379431)
(41.5928, 0.0379106)
(42.8888, 0.0388895)
(44.2251, 0.0377781)
(45.6031, 0.038694)
(47.024, 0.0374889)
(48.4892, 0.0388818)
(50, 0.0389525)
}

\newcommand{\genOdePlotExpCifaruNcsnppErrEulerCHi}{%
(0.02, 0.0194195)
(0.0206232, 0.0193788)
(0.0212657, 0.0195081)
(0.0219283, 0.0195933)
(0.0226116, 0.019718)
(0.0233161, 0.0198004)
(0.0240426, 0.0199922)
(0.0247917, 0.0200859)
(0.0255642, 0.0202903)
(0.0263607, 0.0203936)
(0.0271821, 0.020687)
(0.028029, 0.0208265)
(0.0289023, 0.0210467)
(0.0298029, 0.0209657)
(0.0307315, 0.021026)
(0.031689, 0.0212797)
(0.0326764, 0.0215097)
(0.0336945, 0.0215932)
(0.0347444, 0.021761)
(0.035827, 0.021716)
(0.0369432, 0.0219723)
(0.0380943, 0.0221699)
(0.0392813, 0.0224086)
(0.0405052, 0.022493)
(0.0417673, 0.0226914)
(0.0430687, 0.0229006)
(0.0444106, 0.0228749)
(0.0457943, 0.0229438)
(0.0472212, 0.0230377)
(0.0486925, 0.0233141)
(0.0502097, 0.0234677)
(0.0517741, 0.0236329)
(0.0533873, 0.023757)
(0.0550508, 0.0240208)
(0.056766, 0.0241244)
(0.0585348, 0.0242012)
(0.0603586, 0.0244286)
(0.0622393, 0.0246077)
(0.0641785, 0.0246659)
(0.0661782, 0.0248571)
(0.0682402, 0.0250535)
(0.0703664, 0.0252018)
(0.0725589, 0.0251579)
(0.0748197, 0.0252392)
(0.0771509, 0.0255839)
(0.0795548, 0.0256445)
(0.0820336, 0.0258576)
(0.0845896, 0.0259556)
(0.0872253, 0.0258361)
(0.089943, 0.0262334)
(0.0927455, 0.0262869)
(0.0956352, 0.0263678)
(0.0986151, 0.0265562)
(0.101688, 0.0266831)
(0.104856, 0.026905)
(0.108123, 0.0270709)
(0.111492, 0.0271566)
(0.114966, 0.0273748)
(0.118548, 0.0274885)
(0.122242, 0.0276305)
(0.126051, 0.0277684)
(0.129978, 0.027785)
(0.134028, 0.0279895)
(0.138204, 0.028135)
(0.14251, 0.0281974)
(0.146951, 0.0284404)
(0.151529, 0.0284361)
(0.156251, 0.0287307)
(0.161119, 0.0287467)
(0.166139, 0.0287876)
(0.171316, 0.0289768)
(0.176654, 0.0290942)
(0.182158, 0.0292883)
(0.187834, 0.0293015)
(0.193686, 0.0295707)
(0.199721, 0.0297119)
(0.205944, 0.0296755)
(0.212361, 0.0299169)
(0.218978, 0.0299558)
(0.225801, 0.0300498)
(0.232836, 0.0303239)
(0.240091, 0.0303816)
(0.247572, 0.0303383)
(0.255286, 0.0305744)
(0.26324, 0.0307466)
(0.271442, 0.0308114)
(0.279899, 0.0309001)
(0.288621, 0.0309437)
(0.297613, 0.0310746)
(0.306886, 0.0314247)
(0.316448, 0.031417)
(0.326308, 0.0314037)
(0.336476, 0.0315293)
(0.346959, 0.0317428)
(0.35777, 0.0318709)
(0.368917, 0.0319528)
(0.380412, 0.0320325)
(0.392265, 0.032113)
(0.404487, 0.0321594)
(0.41709, 0.0322483)
(0.430086, 0.032404)
(0.443487, 0.0324029)
(0.457305, 0.0323873)
(0.471554, 0.0326127)
(0.486247, 0.0327024)
(0.501397, 0.0327317)
(0.51702, 0.0330456)
(0.533129, 0.0330512)
(0.54974, 0.0330443)
(0.566869, 0.0332377)
(0.584532, 0.0333376)
(0.602745, 0.0335651)
(0.621525, 0.0333319)
(0.64089, 0.0335464)
(0.660859, 0.0335886)
(0.68145, 0.0337357)
(0.702683, 0.0340957)
(0.724577, 0.0339022)
(0.747154, 0.0340637)
(0.770434, 0.0340636)
(0.794439, 0.0341533)
(0.819192, 0.0341976)
(0.844717, 0.0344533)
(0.871036, 0.0344506)
(0.898176, 0.034444)
(0.926162, 0.0347336)
(0.955019, 0.0347022)
(0.984776, 0.034905)
(1.01546, 0.034923)
(1.0471, 0.0350485)
(1.07972, 0.0348692)
(1.11337, 0.0350248)
(1.14806, 0.0349339)
(1.18383, 0.0352121)
(1.22071, 0.0352602)
(1.25875, 0.0352713)
(1.29797, 0.0350722)
(1.33841, 0.0354908)
(1.38011, 0.0354035)
(1.42312, 0.0357156)
(1.46746, 0.03567)
(1.51318, 0.0358041)
(1.56033, 0.0359341)
(1.60895, 0.0358953)
(1.65908, 0.0358824)
(1.71077, 0.0359491)
(1.76408, 0.0359911)
(1.81904, 0.0358714)
(1.87572, 0.0361626)
(1.93416, 0.0361003)
(1.99443, 0.0361076)
(2.05657, 0.0362463)
(2.12065, 0.0364468)
(2.18672, 0.0366391)
(2.25486, 0.0364229)
(2.32512, 0.0365562)
(2.39756, 0.0367404)
(2.47227, 0.0366142)
(2.5493, 0.0368527)
(2.62873, 0.0368715)
(2.71063, 0.0367)
(2.79509, 0.0367919)
(2.88218, 0.0368106)
(2.97198, 0.0368942)
(3.06459, 0.0371082)
(3.16007, 0.0370963)
(3.25853, 0.0370202)
(3.36006, 0.0369582)
(3.46476, 0.0372195)
(3.57271, 0.0373189)
(3.68403, 0.0371396)
(3.79882, 0.0375352)
(3.91718, 0.0376199)
(4.03923, 0.0373527)
(4.16509, 0.0373586)
(4.29487, 0.0371181)
(4.42869, 0.0373876)
(4.56668, 0.037605)
(4.70896, 0.037704)
(4.85569, 0.0376998)
(5.00698, 0.0375661)
(5.16299, 0.0378552)
(5.32386, 0.038177)
(5.48974, 0.0381987)
(5.66079, 0.0380871)
(5.83717, 0.0381151)
(6.01904, 0.0379627)
(6.20658, 0.0382751)
(6.39997, 0.0382878)
(6.59938, 0.0383008)
(6.805, 0.0384755)
(7.01704, 0.038301)
(7.23567, 0.0381857)
(7.46112, 0.0382078)
(7.6936, 0.0383459)
(7.93332, 0.0382073)
(8.1805, 0.0382069)
(8.43539, 0.0384576)
(8.69822, 0.0385902)
(8.96924, 0.0384113)
(9.24871, 0.0381053)
(9.53688, 0.0380413)
(9.83403, 0.0382273)
(10.1404, 0.0382544)
(10.4564, 0.038599)
(10.7822, 0.0385793)
(11.1181, 0.0385848)
(11.4646, 0.0387601)
(11.8218, 0.0383802)
(12.1901, 0.0386093)
(12.5699, 0.038007)
(12.9616, 0.0388912)
(13.3655, 0.0381324)
(13.7819, 0.0386839)
(14.2113, 0.0390776)
(14.6541, 0.0387289)
(15.1107, 0.0380701)
(15.5815, 0.0385286)
(16.067, 0.038657)
(16.5676, 0.0387818)
(17.0839, 0.03833)
(17.6162, 0.0387221)
(18.165, 0.0393883)
(18.731, 0.0387417)
(19.3147, 0.0386785)
(19.9165, 0.039264)
(20.537, 0.0386164)
(21.1769, 0.0387425)
(21.8368, 0.039157)
(22.5171, 0.0390864)
(23.2187, 0.0390714)
(23.9422, 0.0393444)
(24.6882, 0.0389791)
(25.4574, 0.0387945)
(26.2506, 0.0395026)
(27.0685, 0.0389437)
(27.9119, 0.0400524)
(28.7816, 0.0385687)
(29.6784, 0.0392077)
(30.6031, 0.0391045)
(31.5567, 0.0390326)
(32.5399, 0.0392126)
(33.5538, 0.0396011)
(34.5993, 0.039119)
(35.6773, 0.0392901)
(36.789, 0.0387299)
(37.9352, 0.0394731)
(39.1172, 0.0393775)
(40.336, 0.0389534)
(41.5928, 0.0391875)
(42.8888, 0.0396668)
(44.2251, 0.0389312)
(45.6031, 0.0398732)
(47.024, 0.0385552)
(48.4892, 0.0399712)
(50, 0.0401246)
}

\newcommand{\genOdePlotExpCifaruNcsnppErrEulerD}{%
(0.02, 0.00332297)
(0.0206232, 0.00340076)
(0.0212657, 0.00339866)
(0.0219283, 0.0034879)
(0.0226116, 0.00353713)
(0.0233161, 0.00358532)
(0.0240426, 0.00368616)
(0.0247917, 0.00371537)
(0.0255642, 0.0037773)
(0.0263607, 0.00384585)
(0.0271821, 0.00389113)
(0.028029, 0.00396787)
(0.0289023, 0.00401211)
(0.0298029, 0.004102)
(0.0307315, 0.00414822)
(0.031689, 0.0041993)
(0.0326764, 0.00423638)
(0.0336945, 0.00433633)
(0.0347444, 0.00440029)
(0.035827, 0.00448506)
(0.0369432, 0.00453635)
(0.0380943, 0.00463882)
(0.0392813, 0.00465345)
(0.0405052, 0.00473329)
(0.0417673, 0.00480398)
(0.0430687, 0.00488943)
(0.0444106, 0.00495499)
(0.0457943, 0.00504176)
(0.0472212, 0.00506288)
(0.0486925, 0.00519506)
(0.0502097, 0.00523658)
(0.0517741, 0.00530308)
(0.0533873, 0.00534809)
(0.0550508, 0.00547321)
(0.056766, 0.00552035)
(0.0585348, 0.00564352)
(0.0603586, 0.00563993)
(0.0622393, 0.00582944)
(0.0641785, 0.00586608)
(0.0661782, 0.00594398)
(0.0682402, 0.00608131)
(0.0703664, 0.00614954)
(0.0725589, 0.00626949)
(0.0748197, 0.00636312)
(0.0771509, 0.00640371)
(0.0795548, 0.00658304)
(0.0820336, 0.00661368)
(0.0845896, 0.00672976)
(0.0872253, 0.00679361)
(0.089943, 0.00689106)
(0.0927455, 0.00696537)
(0.0956352, 0.0070668)
(0.0986151, 0.00719116)
(0.101688, 0.00726161)
(0.104856, 0.00734515)
(0.108123, 0.00750239)
(0.111492, 0.00756777)
(0.114966, 0.00765838)
(0.118548, 0.00775657)
(0.122242, 0.00792898)
(0.126051, 0.00796568)
(0.129978, 0.0080732)
(0.134028, 0.0082133)
(0.138204, 0.00833424)
(0.14251, 0.00839086)
(0.146951, 0.00856945)
(0.151529, 0.0087082)
(0.156251, 0.00878835)
(0.161119, 0.00882826)
(0.166139, 0.00900878)
(0.171316, 0.00919056)
(0.176654, 0.00925828)
(0.182158, 0.00934144)
(0.187834, 0.00945373)
(0.193686, 0.0096028)
(0.199721, 0.0097483)
(0.205944, 0.00986288)
(0.212361, 0.00990237)
(0.218978, 0.0101125)
(0.225801, 0.0102361)
(0.232836, 0.0103634)
(0.240091, 0.0105287)
(0.247572, 0.0106998)
(0.255286, 0.0107507)
(0.26324, 0.0108543)
(0.271442, 0.0110547)
(0.279899, 0.0112012)
(0.288621, 0.0113326)
(0.297613, 0.011448)
(0.306886, 0.0115953)
(0.316448, 0.011721)
(0.326308, 0.0119713)
(0.336476, 0.0120623)
(0.346959, 0.0121892)
(0.35777, 0.0123073)
(0.368917, 0.0125442)
(0.380412, 0.0126292)
(0.392265, 0.0127893)
(0.404487, 0.0128908)
(0.41709, 0.0130241)
(0.430086, 0.0132004)
(0.443487, 0.0133886)
(0.457305, 0.0135838)
(0.471554, 0.0137206)
(0.486247, 0.0138197)
(0.501397, 0.0140509)
(0.51702, 0.0141749)
(0.533129, 0.0143863)
(0.54974, 0.0145983)
(0.566869, 0.014731)
(0.584532, 0.0150048)
(0.602745, 0.0149847)
(0.621525, 0.0151343)
(0.64089, 0.0154457)
(0.660859, 0.0156916)
(0.68145, 0.0158592)
(0.702683, 0.0159308)
(0.724577, 0.0162277)
(0.747154, 0.016338)
(0.770434, 0.0165059)
(0.794439, 0.0166622)
(0.819192, 0.0169156)
(0.844717, 0.0171256)
(0.871036, 0.0173113)
(0.898176, 0.0174824)
(0.926162, 0.0177212)
(0.955019, 0.0179617)
(0.984776, 0.0180949)
(1.01546, 0.0183908)
(1.0471, 0.0183762)
(1.07972, 0.0186737)
(1.11337, 0.0189242)
(1.14806, 0.0191748)
(1.18383, 0.0194615)
(1.22071, 0.019551)
(1.25875, 0.0199146)
(1.29797, 0.0200769)
(1.33841, 0.0202467)
(1.38011, 0.0202747)
(1.42312, 0.0207008)
(1.46746, 0.0208778)
(1.51318, 0.0212277)
(1.56033, 0.0214361)
(1.60895, 0.0215858)
(1.65908, 0.0218931)
(1.71077, 0.0222085)
(1.76408, 0.0223279)
(1.81904, 0.0226165)
(1.87572, 0.0228236)
(1.93416, 0.023038)
(1.99443, 0.0233751)
(2.05657, 0.0236077)
(2.12065, 0.0239912)
(2.18672, 0.0242355)
(2.25486, 0.0246133)
(2.32512, 0.0245892)
(2.39756, 0.0249478)
(2.47227, 0.0252557)
(2.5493, 0.0255568)
(2.62873, 0.0259466)
(2.71063, 0.0261207)
(2.79509, 0.0263171)
(2.88218, 0.0265768)
(2.97198, 0.026864)
(3.06459, 0.0271954)
(3.16007, 0.0277088)
(3.25853, 0.0278878)
(3.36006, 0.0282789)
(3.46476, 0.0285586)
(3.57271, 0.0285928)
(3.68403, 0.0290633)
(3.79882, 0.0295575)
(3.91718, 0.0296893)
(4.03923, 0.0299089)
(4.16509, 0.0304865)
(4.29487, 0.030875)
(4.42869, 0.0309826)
(4.56668, 0.0314803)
(4.70896, 0.0317814)
(4.85569, 0.0322379)
(5.00698, 0.0323226)
(5.16299, 0.0327774)
(5.32386, 0.0330599)
(5.48974, 0.0336521)
(5.66079, 0.0337308)
(5.83717, 0.0343375)
(6.01904, 0.0345067)
(6.20658, 0.0349973)
(6.39997, 0.0355533)
(6.59938, 0.0358827)
(6.805, 0.0365588)
(7.01704, 0.0369421)
(7.23567, 0.036979)
(7.46112, 0.0370555)
(7.6936, 0.0377388)
(7.93332, 0.0379721)
(8.1805, 0.0383563)
(8.43539, 0.0389976)
(8.69822, 0.0393821)
(8.96924, 0.0397786)
(9.24871, 0.039792)
(9.53688, 0.040236)
(9.83403, 0.0406554)
(10.1404, 0.0414331)
(10.4564, 0.0421187)
(10.7822, 0.0420125)
(11.1181, 0.0426806)
(11.4646, 0.0434259)
(11.8218, 0.0433966)
(12.1901, 0.0443931)
(12.5699, 0.0445027)
(12.9616, 0.0447241)
(13.3655, 0.0454927)
(13.7819, 0.0459095)
(14.2113, 0.04648)
(14.6541, 0.0470789)
(15.1107, 0.0472972)
(15.5815, 0.0481993)
(16.067, 0.0486392)
(16.5676, 0.0488529)
(17.0839, 0.0490688)
(17.6162, 0.0504182)
(18.165, 0.0504037)
(18.731, 0.050756)
(19.3147, 0.0523609)
(19.9165, 0.0521752)
(20.537, 0.0528801)
(21.1769, 0.0532395)
(21.8368, 0.0537734)
(22.5171, 0.0548599)
(23.2187, 0.0551456)
(23.9422, 0.0555483)
(24.6882, 0.0561192)
(25.4574, 0.05662)
(26.2506, 0.0572161)
(27.0685, 0.0581677)
(27.9119, 0.0584053)
(28.7816, 0.058921)
(29.6784, 0.0594346)
(30.6031, 0.0603371)
(31.5567, 0.0616103)
(32.5399, 0.0617598)
(33.5538, 0.0622715)
(34.5993, 0.0633038)
(35.6773, 0.0636006)
(36.789, 0.0646032)
(37.9352, 0.0661671)
(39.1172, 0.065509)
(40.336, 0.0664511)
(41.5928, 0.0673947)
(42.8888, 0.0669471)
(44.2251, 0.0680478)
(45.6031, 0.0694117)
(47.024, 0.0700728)
(48.4892, 0.0705661)
(50, 0.0715327)
}

\newcommand{\genOdePlotExpCifaruNcsnppErrEulerDLo}{%
(0.02, 0.0029894)
(0.0206232, 0.00305584)
(0.0212657, 0.00306019)
(0.0219283, 0.00316691)
(0.0226116, 0.00318933)
(0.0233161, 0.00323901)
(0.0240426, 0.00332851)
(0.0247917, 0.00334712)
(0.0255642, 0.00341805)
(0.0263607, 0.0035072)
(0.0271821, 0.00356001)
(0.028029, 0.00367025)
(0.0289023, 0.00364973)
(0.0298029, 0.00377527)
(0.0307315, 0.00378716)
(0.031689, 0.00383998)
(0.0326764, 0.00388516)
(0.0336945, 0.00398263)
(0.0347444, 0.00404968)
(0.035827, 0.00414198)
(0.0369432, 0.00418919)
(0.0380943, 0.00430722)
(0.0392813, 0.00431948)
(0.0405052, 0.00439732)
(0.0417673, 0.0044654)
(0.0430687, 0.00455836)
(0.0444106, 0.00463373)
(0.0457943, 0.00470476)
(0.0472212, 0.00471357)
(0.0486925, 0.00486084)
(0.0502097, 0.00492157)
(0.0517741, 0.00495431)
(0.0533873, 0.0050082)
(0.0550508, 0.00514505)
(0.056766, 0.00519075)
(0.0585348, 0.00530377)
(0.0603586, 0.00529046)
(0.0622393, 0.00552151)
(0.0641785, 0.00550527)
(0.0661782, 0.00557477)
(0.0682402, 0.00573297)
(0.0703664, 0.00586304)
(0.0725589, 0.00590942)
(0.0748197, 0.00601711)
(0.0771509, 0.0060845)
(0.0795548, 0.00624094)
(0.0820336, 0.00629322)
(0.0845896, 0.006408)
(0.0872253, 0.00649658)
(0.089943, 0.00657923)
(0.0927455, 0.00663554)
(0.0956352, 0.00673733)
(0.0986151, 0.00688573)
(0.101688, 0.00696527)
(0.104856, 0.00703204)
(0.108123, 0.00715863)
(0.111492, 0.00726613)
(0.114966, 0.00735216)
(0.118548, 0.00743378)
(0.122242, 0.00763668)
(0.126051, 0.0076595)
(0.129978, 0.00776866)
(0.134028, 0.00790441)
(0.138204, 0.0080162)
(0.14251, 0.00809338)
(0.146951, 0.00828717)
(0.151529, 0.00839589)
(0.156251, 0.00848769)
(0.161119, 0.00853086)
(0.166139, 0.00869916)
(0.171316, 0.00890767)
(0.176654, 0.00897101)
(0.182158, 0.00905159)
(0.187834, 0.00914535)
(0.193686, 0.00931087)
(0.199721, 0.00944695)
(0.205944, 0.00960336)
(0.212361, 0.00961604)
(0.218978, 0.00984621)
(0.225801, 0.00994458)
(0.232836, 0.0100876)
(0.240091, 0.0102539)
(0.247572, 0.0104286)
(0.255286, 0.0105072)
(0.26324, 0.0105699)
(0.271442, 0.0107887)
(0.279899, 0.0109514)
(0.288621, 0.0110464)
(0.297613, 0.0111777)
(0.306886, 0.0113505)
(0.316448, 0.011464)
(0.326308, 0.0116951)
(0.336476, 0.011807)
(0.346959, 0.0119334)
(0.35777, 0.0120602)
(0.368917, 0.0122899)
(0.380412, 0.0123457)
(0.392265, 0.0125495)
(0.404487, 0.0126482)
(0.41709, 0.0127726)
(0.430086, 0.0129344)
(0.443487, 0.0131487)
(0.457305, 0.0133456)
(0.471554, 0.013476)
(0.486247, 0.0135783)
(0.501397, 0.013822)
(0.51702, 0.0138906)
(0.533129, 0.0141762)
(0.54974, 0.0143781)
(0.566869, 0.0145074)
(0.584532, 0.0147741)
(0.602745, 0.014724)
(0.621525, 0.0148686)
(0.64089, 0.0152019)
(0.660859, 0.0154399)
(0.68145, 0.0156368)
(0.702683, 0.0157148)
(0.724577, 0.0159761)
(0.747154, 0.0161035)
(0.770434, 0.0162533)
(0.794439, 0.0164108)
(0.819192, 0.0166406)
(0.844717, 0.0168797)
(0.871036, 0.0171067)
(0.898176, 0.0172154)
(0.926162, 0.0174631)
(0.955019, 0.0176954)
(0.984776, 0.0178315)
(1.01546, 0.0181154)
(1.0471, 0.0181129)
(1.07972, 0.0184217)
(1.11337, 0.0186649)
(1.14806, 0.0189033)
(1.18383, 0.019207)
(1.22071, 0.0193229)
(1.25875, 0.0196375)
(1.29797, 0.0198269)
(1.33841, 0.0199963)
(1.38011, 0.0200232)
(1.42312, 0.0204202)
(1.46746, 0.0206228)
(1.51318, 0.0209379)
(1.56033, 0.0211038)
(1.60895, 0.0213114)
(1.65908, 0.0216243)
(1.71077, 0.0218975)
(1.76408, 0.0220574)
(1.81904, 0.0223598)
(1.87572, 0.0225249)
(1.93416, 0.0227478)
(1.99443, 0.0231211)
(2.05657, 0.0233197)
(2.12065, 0.0237222)
(2.18672, 0.0239638)
(2.25486, 0.0242683)
(2.32512, 0.0242436)
(2.39756, 0.0245905)
(2.47227, 0.0248646)
(2.5493, 0.0252271)
(2.62873, 0.025609)
(2.71063, 0.0257854)
(2.79509, 0.0260467)
(2.88218, 0.0262485)
(2.97198, 0.0266048)
(3.06459, 0.0268089)
(3.16007, 0.0273935)
(3.25853, 0.0275055)
(3.36006, 0.0278875)
(3.46476, 0.02823)
(3.57271, 0.0282095)
(3.68403, 0.0287432)
(3.79882, 0.0291645)
(3.91718, 0.0292897)
(4.03923, 0.0295113)
(4.16509, 0.030079)
(4.29487, 0.0304926)
(4.42869, 0.0306349)
(4.56668, 0.031189)
(4.70896, 0.0314193)
(4.85569, 0.0318398)
(5.00698, 0.031906)
(5.16299, 0.0323187)
(5.32386, 0.0325886)
(5.48974, 0.033116)
(5.66079, 0.0331355)
(5.83717, 0.0337748)
(6.01904, 0.0338891)
(6.20658, 0.0343746)
(6.39997, 0.034855)
(6.59938, 0.0352985)
(6.805, 0.0361778)
(7.01704, 0.0366936)
(7.23567, 0.036618)
(7.46112, 0.0364568)
(7.6936, 0.037267)
(7.93332, 0.0374833)
(8.1805, 0.0378787)
(8.43539, 0.0384705)
(8.69822, 0.0387332)
(8.96924, 0.0393185)
(9.24871, 0.0392793)
(9.53688, 0.0397955)
(9.83403, 0.0400907)
(10.1404, 0.0409542)
(10.4564, 0.0416083)
(10.7822, 0.0413517)
(11.1181, 0.0420177)
(11.4646, 0.0429977)
(11.8218, 0.0428319)
(12.1901, 0.0437466)
(12.5699, 0.0438985)
(12.9616, 0.04416)
(13.3655, 0.0449165)
(13.7819, 0.0451763)
(14.2113, 0.0456911)
(14.6541, 0.0464948)
(15.1107, 0.0467831)
(15.5815, 0.047545)
(16.067, 0.0481008)
(16.5676, 0.0483821)
(17.0839, 0.0486036)
(17.6162, 0.0496347)
(18.165, 0.0497509)
(18.731, 0.0500486)
(19.3147, 0.0516033)
(19.9165, 0.051705)
(20.537, 0.0521828)
(21.1769, 0.0524353)
(21.8368, 0.0531905)
(22.5171, 0.0541871)
(23.2187, 0.0542014)
(23.9422, 0.0547025)
(24.6882, 0.055537)
(25.4574, 0.0557534)
(26.2506, 0.0561177)
(27.0685, 0.0574558)
(27.9119, 0.0578385)
(28.7816, 0.0581046)
(29.6784, 0.058689)
(30.6031, 0.0597136)
(31.5567, 0.0605797)
(32.5399, 0.0610721)
(33.5538, 0.0615812)
(34.5993, 0.0624947)
(35.6773, 0.0627456)
(36.789, 0.0637263)
(37.9352, 0.0652999)
(39.1172, 0.0644784)
(40.336, 0.0657324)
(41.5928, 0.0666397)
(42.8888, 0.0660822)
(44.2251, 0.0670353)
(45.6031, 0.0685862)
(47.024, 0.069173)
(48.4892, 0.0695893)
(50, 0.0705478)
}

\newcommand{\genOdePlotExpCifaruNcsnppErrEulerDHi}{%
(0.02, 0.00365654)
(0.0206232, 0.00374567)
(0.0212657, 0.00373713)
(0.0219283, 0.0038089)
(0.0226116, 0.00388493)
(0.0233161, 0.00393162)
(0.0240426, 0.00404382)
(0.0247917, 0.00408363)
(0.0255642, 0.00413655)
(0.0263607, 0.0041845)
(0.0271821, 0.00422224)
(0.028029, 0.00426548)
(0.0289023, 0.00437449)
(0.0298029, 0.00442873)
(0.0307315, 0.00450928)
(0.031689, 0.00455863)
(0.0326764, 0.0045876)
(0.0336945, 0.00469002)
(0.0347444, 0.0047509)
(0.035827, 0.00482815)
(0.0369432, 0.00488352)
(0.0380943, 0.00497042)
(0.0392813, 0.00498742)
(0.0405052, 0.00506925)
(0.0417673, 0.00514257)
(0.0430687, 0.0052205)
(0.0444106, 0.00527626)
(0.0457943, 0.00537877)
(0.0472212, 0.00541219)
(0.0486925, 0.00552928)
(0.0502097, 0.00555158)
(0.0517741, 0.00565185)
(0.0533873, 0.00568797)
(0.0550508, 0.00580137)
(0.056766, 0.00584994)
(0.0585348, 0.00598327)
(0.0603586, 0.00598941)
(0.0622393, 0.00613737)
(0.0641785, 0.0062269)
(0.0661782, 0.0063132)
(0.0682402, 0.00642966)
(0.0703664, 0.00643605)
(0.0725589, 0.00662955)
(0.0748197, 0.00670913)
(0.0771509, 0.00672293)
(0.0795548, 0.00692514)
(0.0820336, 0.00693415)
(0.0845896, 0.00705152)
(0.0872253, 0.00709063)
(0.089943, 0.00720289)
(0.0927455, 0.00729521)
(0.0956352, 0.00739627)
(0.0986151, 0.00749659)
(0.101688, 0.00755796)
(0.104856, 0.00765826)
(0.108123, 0.00784615)
(0.111492, 0.00786942)
(0.114966, 0.00796461)
(0.118548, 0.00807935)
(0.122242, 0.00822127)
(0.126051, 0.00827186)
(0.129978, 0.00837775)
(0.134028, 0.0085222)
(0.138204, 0.00865229)
(0.14251, 0.00868833)
(0.146951, 0.00885173)
(0.151529, 0.00902052)
(0.156251, 0.00908901)
(0.161119, 0.00912566)
(0.166139, 0.0093184)
(0.171316, 0.00947346)
(0.176654, 0.00954554)
(0.182158, 0.00963128)
(0.187834, 0.0097621)
(0.193686, 0.00989472)
(0.199721, 0.0100497)
(0.205944, 0.0101224)
(0.212361, 0.0101887)
(0.218978, 0.0103788)
(0.225801, 0.0105277)
(0.232836, 0.0106392)
(0.240091, 0.0108035)
(0.247572, 0.010971)
(0.255286, 0.0109942)
(0.26324, 0.0111388)
(0.271442, 0.0113208)
(0.279899, 0.0114509)
(0.288621, 0.0116189)
(0.297613, 0.0117183)
(0.306886, 0.0118401)
(0.316448, 0.0119781)
(0.326308, 0.0122475)
(0.336476, 0.0123176)
(0.346959, 0.0124451)
(0.35777, 0.0125543)
(0.368917, 0.0127984)
(0.380412, 0.0129127)
(0.392265, 0.0130291)
(0.404487, 0.0131335)
(0.41709, 0.0132757)
(0.430086, 0.0134663)
(0.443487, 0.0136286)
(0.457305, 0.013822)
(0.471554, 0.0139651)
(0.486247, 0.0140611)
(0.501397, 0.0142798)
(0.51702, 0.0144592)
(0.533129, 0.0145964)
(0.54974, 0.0148185)
(0.566869, 0.0149546)
(0.584532, 0.0152355)
(0.602745, 0.0152454)
(0.621525, 0.0154)
(0.64089, 0.0156894)
(0.660859, 0.0159433)
(0.68145, 0.0160815)
(0.702683, 0.0161468)
(0.724577, 0.0164792)
(0.747154, 0.0165725)
(0.770434, 0.0167584)
(0.794439, 0.0169136)
(0.819192, 0.0171905)
(0.844717, 0.0173715)
(0.871036, 0.017516)
(0.898176, 0.0177493)
(0.926162, 0.0179792)
(0.955019, 0.0182279)
(0.984776, 0.0183582)
(1.01546, 0.0186662)
(1.0471, 0.0186395)
(1.07972, 0.0189257)
(1.11337, 0.0191834)
(1.14806, 0.0194462)
(1.18383, 0.0197159)
(1.22071, 0.019779)
(1.25875, 0.0201918)
(1.29797, 0.0203269)
(1.33841, 0.020497)
(1.38011, 0.0205261)
(1.42312, 0.0209815)
(1.46746, 0.0211328)
(1.51318, 0.0215174)
(1.56033, 0.0217683)
(1.60895, 0.0218602)
(1.65908, 0.0221618)
(1.71077, 0.0225195)
(1.76408, 0.0225984)
(1.81904, 0.0228731)
(1.87572, 0.0231223)
(1.93416, 0.0233282)
(1.99443, 0.0236291)
(2.05657, 0.0238958)
(2.12065, 0.0242601)
(2.18672, 0.0245071)
(2.25486, 0.0249582)
(2.32512, 0.0249347)
(2.39756, 0.0253052)
(2.47227, 0.0256468)
(2.5493, 0.0258865)
(2.62873, 0.0262841)
(2.71063, 0.026456)
(2.79509, 0.0265874)
(2.88218, 0.0269051)
(2.97198, 0.0271233)
(3.06459, 0.027582)
(3.16007, 0.0280241)
(3.25853, 0.0282701)
(3.36006, 0.0286703)
(3.46476, 0.0288871)
(3.57271, 0.0289761)
(3.68403, 0.0293834)
(3.79882, 0.0299505)
(3.91718, 0.0300889)
(4.03923, 0.0303065)
(4.16509, 0.030894)
(4.29487, 0.0312574)
(4.42869, 0.0313302)
(4.56668, 0.0317715)
(4.70896, 0.0321435)
(4.85569, 0.0326361)
(5.00698, 0.0327392)
(5.16299, 0.0332362)
(5.32386, 0.0335312)
(5.48974, 0.0341883)
(5.66079, 0.034326)
(5.83717, 0.0349002)
(6.01904, 0.0351242)
(6.20658, 0.0356199)
(6.39997, 0.0362517)
(6.59938, 0.0364669)
(6.805, 0.0369398)
(7.01704, 0.0371905)
(7.23567, 0.03734)
(7.46112, 0.0376543)
(7.6936, 0.0382105)
(7.93332, 0.0384609)
(8.1805, 0.0388339)
(8.43539, 0.0395247)
(8.69822, 0.0400311)
(8.96924, 0.0402387)
(9.24871, 0.0403046)
(9.53688, 0.0406766)
(9.83403, 0.0412201)
(10.1404, 0.041912)
(10.4564, 0.0426292)
(10.7822, 0.0426733)
(11.1181, 0.0433435)
(11.4646, 0.0438542)
(11.8218, 0.0439613)
(12.1901, 0.0450396)
(12.5699, 0.0451068)
(12.9616, 0.0452882)
(13.3655, 0.0460689)
(13.7819, 0.0466427)
(14.2113, 0.0472688)
(14.6541, 0.047663)
(15.1107, 0.0478114)
(15.5815, 0.0488537)
(16.067, 0.0491775)
(16.5676, 0.0493238)
(17.0839, 0.0495339)
(17.6162, 0.0512017)
(18.165, 0.0510565)
(18.731, 0.0514635)
(19.3147, 0.0531184)
(19.9165, 0.0526454)
(20.537, 0.0535774)
(21.1769, 0.0540438)
(21.8368, 0.0543563)
(22.5171, 0.0555328)
(23.2187, 0.0560897)
(23.9422, 0.056394)
(24.6882, 0.0567014)
(25.4574, 0.0574867)
(26.2506, 0.0583145)
(27.0685, 0.0588796)
(27.9119, 0.0589721)
(28.7816, 0.0597375)
(29.6784, 0.0601801)
(30.6031, 0.0609606)
(31.5567, 0.062641)
(32.5399, 0.0624476)
(33.5538, 0.0629618)
(34.5993, 0.0641129)
(35.6773, 0.0644555)
(36.789, 0.06548)
(37.9352, 0.0670344)
(39.1172, 0.0665396)
(40.336, 0.0671698)
(41.5928, 0.0681497)
(42.8888, 0.067812)
(44.2251, 0.0690603)
(45.6031, 0.0702372)
(47.024, 0.0709725)
(48.4892, 0.071543)
(50, 0.0725177)
}

\newcommand{\genOdePlotExpCifaruNcsnppErrEulerE}{%
(0.02, 0.000556345)
(0.0206232, 0.000574639)
(0.0212657, 0.00059237)
(0.0219283, 0.000615838)
(0.0226116, 0.000627789)
(0.0233161, 0.000637239)
(0.0240426, 0.000656166)
(0.0247917, 0.000673664)
(0.0255642, 0.000685451)
(0.0263607, 0.000722767)
(0.0271821, 0.000731707)
(0.028029, 0.000744331)
(0.0289023, 0.000765524)
(0.0298029, 0.000784017)
(0.0307315, 0.000811125)
(0.031689, 0.000839147)
(0.0326764, 0.000859947)
(0.0336945, 0.00087706)
(0.0347444, 0.000901847)
(0.035827, 0.000923219)
(0.0369432, 0.000953231)
(0.0380943, 0.000979451)
(0.0392813, 0.000996332)
(0.0405052, 0.00102223)
(0.0417673, 0.00105067)
(0.0430687, 0.00108159)
(0.0444106, 0.0011044)
(0.0457943, 0.00114137)
(0.0472212, 0.00116634)
(0.0486925, 0.00119419)
(0.0502097, 0.00124301)
(0.0517741, 0.00126058)
(0.0533873, 0.00129939)
(0.0550508, 0.00133842)
(0.056766, 0.00137534)
(0.0585348, 0.00138922)
(0.0603586, 0.00145458)
(0.0622393, 0.00147802)
(0.0641785, 0.00151164)
(0.0661782, 0.00154809)
(0.0682402, 0.00157282)
(0.0703664, 0.0016187)
(0.0725589, 0.0016586)
(0.0748197, 0.00170572)
(0.0771509, 0.00174947)
(0.0795548, 0.0017899)
(0.0820336, 0.0018572)
(0.0845896, 0.00187182)
(0.0872253, 0.00192207)
(0.089943, 0.00198445)
(0.0927455, 0.00204383)
(0.0956352, 0.00209245)
(0.0986151, 0.00214705)
(0.101688, 0.0021849)
(0.104856, 0.00225474)
(0.108123, 0.00228555)
(0.111492, 0.00234233)
(0.114966, 0.00238262)
(0.118548, 0.00248662)
(0.122242, 0.00253642)
(0.126051, 0.0026188)
(0.129978, 0.0026828)
(0.134028, 0.00275447)
(0.138204, 0.00277931)
(0.14251, 0.00285439)
(0.146951, 0.00294029)
(0.151529, 0.00302209)
(0.156251, 0.00309709)
(0.161119, 0.00317795)
(0.166139, 0.00321551)
(0.171316, 0.00333111)
(0.176654, 0.00340837)
(0.182158, 0.00350569)
(0.187834, 0.00357139)
(0.193686, 0.00367443)
(0.199721, 0.00374056)
(0.205944, 0.00380986)
(0.212361, 0.0039182)
(0.218978, 0.00400483)
(0.225801, 0.0041309)
(0.232836, 0.00420562)
(0.240091, 0.00428106)
(0.247572, 0.00442531)
(0.255286, 0.00452106)
(0.26324, 0.00463239)
(0.271442, 0.00472725)
(0.279899, 0.00486506)
(0.288621, 0.00498886)
(0.297613, 0.00511091)
(0.306886, 0.0052158)
(0.316448, 0.00530504)
(0.326308, 0.00544888)
(0.336476, 0.00555758)
(0.346959, 0.00571458)
(0.35777, 0.00591158)
(0.368917, 0.00600505)
(0.380412, 0.00613209)
(0.392265, 0.00625368)
(0.404487, 0.00645543)
(0.41709, 0.00659691)
(0.430086, 0.00669018)
(0.443487, 0.0069122)
(0.457305, 0.00707353)
(0.471554, 0.00725579)
(0.486247, 0.00735)
(0.501397, 0.00749356)
(0.51702, 0.00763741)
(0.533129, 0.00788194)
(0.54974, 0.00803633)
(0.566869, 0.0082531)
(0.584532, 0.00846815)
(0.602745, 0.00861201)
(0.621525, 0.00887571)
(0.64089, 0.00903179)
(0.660859, 0.00921844)
(0.68145, 0.00952316)
(0.702683, 0.00966841)
(0.724577, 0.00984409)
(0.747154, 0.0101438)
(0.770434, 0.0103654)
(0.794439, 0.0106366)
(0.819192, 0.0109748)
(0.844717, 0.0111029)
(0.871036, 0.0112556)
(0.898176, 0.011608)
(0.926162, 0.0118648)
(0.955019, 0.0121638)
(0.984776, 0.0124377)
(1.01546, 0.0126848)
(1.0471, 0.0130757)
(1.07972, 0.0131548)
(1.11337, 0.0135449)
(1.14806, 0.0139071)
(1.18383, 0.0141934)
(1.22071, 0.0145808)
(1.25875, 0.0147903)
(1.29797, 0.015182)
(1.33841, 0.0155383)
(1.38011, 0.0159209)
(1.42312, 0.0162878)
(1.46746, 0.0166959)
(1.51318, 0.0169251)
(1.56033, 0.0173277)
(1.60895, 0.0179893)
(1.65908, 0.01834)
(1.71077, 0.0186165)
(1.76408, 0.0189521)
(1.81904, 0.0194382)
(1.87572, 0.0197921)
(1.93416, 0.0202178)
(1.99443, 0.0208423)
(2.05657, 0.021177)
(2.12065, 0.0217995)
(2.18672, 0.0222495)
(2.25486, 0.0228675)
(2.32512, 0.0230408)
(2.39756, 0.0237556)
(2.47227, 0.0243048)
(2.5493, 0.0248435)
(2.62873, 0.0254618)
(2.71063, 0.0259908)
(2.79509, 0.0264387)
(2.88218, 0.027018)
(2.97198, 0.0276575)
(3.06459, 0.028299)
(3.16007, 0.0290925)
(3.25853, 0.0296934)
(3.36006, 0.0303615)
(3.46476, 0.0310613)
(3.57271, 0.0316663)
(3.68403, 0.0323792)
(3.79882, 0.0331247)
(3.91718, 0.0338962)
(4.03923, 0.0346019)
(4.16509, 0.0353853)
(4.29487, 0.0359059)
(4.42869, 0.0367908)
(4.56668, 0.037862)
(4.70896, 0.0387914)
(4.85569, 0.0395586)
(5.00698, 0.0404756)
(5.16299, 0.0416609)
(5.32386, 0.0423733)
(5.48974, 0.043547)
(5.66079, 0.0440566)
(5.83717, 0.0448489)
(6.01904, 0.0461782)
(6.20658, 0.047374)
(6.39997, 0.0481876)
(6.59938, 0.0500844)
(6.805, 0.0511513)
(7.01704, 0.0519603)
(7.23567, 0.0526727)
(7.46112, 0.0540386)
(7.6936, 0.0551642)
(7.93332, 0.0560639)
(8.1805, 0.0576083)
(8.43539, 0.0590392)
(8.69822, 0.0602363)
(8.96924, 0.0612574)
(9.24871, 0.0626862)
(9.53688, 0.0641136)
(9.83403, 0.0653643)
(10.1404, 0.0674719)
(10.4564, 0.0685633)
(10.7822, 0.0702318)
(11.1181, 0.0721265)
(11.4646, 0.0729154)
(11.8218, 0.0753916)
(12.1901, 0.0761295)
(12.5699, 0.0782208)
(12.9616, 0.079799)
(13.3655, 0.0817807)
(13.7819, 0.0838177)
(14.2113, 0.0855613)
(14.6541, 0.0871151)
(15.1107, 0.0895036)
(15.5815, 0.0915326)
(16.067, 0.0932849)
(16.5676, 0.0948668)
(17.0839, 0.0979983)
(17.6162, 0.0995084)
(18.165, 0.101545)
(18.731, 0.105427)
(19.3147, 0.106701)
(19.9165, 0.109287)
(20.537, 0.111243)
(21.1769, 0.1136)
(21.8368, 0.116745)
(22.5171, 0.11954)
(23.2187, 0.121606)
(23.9422, 0.123855)
(24.6882, 0.127205)
(25.4574, 0.129696)
(26.2506, 0.132691)
(27.0685, 0.136721)
(27.9119, 0.138794)
(28.7816, 0.141027)
(29.6784, 0.145615)
(30.6031, 0.148433)
(31.5567, 0.151209)
(32.5399, 0.155386)
(33.5538, 0.15824)
(34.5993, 0.161848)
(35.6773, 0.165615)
(36.789, 0.169282)
(37.9352, 0.174115)
(39.1172, 0.177619)
(40.336, 0.180841)
(41.5928, 0.185656)
(42.8888, 0.188877)
(44.2251, 0.192958)
(45.6031, 0.197102)
(47.024, 0.201414)
(48.4892, 0.208286)
(50, 0.210962)
}

\newcommand{\genOdePlotExpCifaruNcsnppErrEulerELo}{%
(0.02, 0.000478419)
(0.0206232, 0.000502633)
(0.0212657, 0.000534452)
(0.0219283, 0.000545674)
(0.0226116, 0.000558534)
(0.0233161, 0.000560566)
(0.0240426, 0.000587976)
(0.0247917, 0.000608432)
(0.0255642, 0.000609463)
(0.0263607, 0.000650811)
(0.0271821, 0.000658646)
(0.028029, 0.000671978)
(0.0289023, 0.00068511)
(0.0298029, 0.000704423)
(0.0307315, 0.000735512)
(0.031689, 0.000765745)
(0.0326764, 0.000785229)
(0.0336945, 0.000799312)
(0.0347444, 0.000823673)
(0.035827, 0.000841331)
(0.0369432, 0.000875144)
(0.0380943, 0.000904826)
(0.0392813, 0.000913278)
(0.0405052, 0.000948403)
(0.0417673, 0.000966829)
(0.0430687, 0.00100416)
(0.0444106, 0.00102871)
(0.0457943, 0.00106602)
(0.0472212, 0.00107742)
(0.0486925, 0.00110971)
(0.0502097, 0.00116437)
(0.0517741, 0.00117105)
(0.0533873, 0.00121965)
(0.0550508, 0.00124696)
(0.056766, 0.00128996)
(0.0585348, 0.00130427)
(0.0603586, 0.00135976)
(0.0622393, 0.00139812)
(0.0641785, 0.00142055)
(0.0661782, 0.00145026)
(0.0682402, 0.00146639)
(0.0703664, 0.00153481)
(0.0725589, 0.00157018)
(0.0748197, 0.001602)
(0.0771509, 0.00164753)
(0.0795548, 0.00170485)
(0.0820336, 0.00175141)
(0.0845896, 0.00176511)
(0.0872253, 0.00182072)
(0.089943, 0.00188127)
(0.0927455, 0.00192714)
(0.0956352, 0.00199713)
(0.0986151, 0.00202845)
(0.101688, 0.00208103)
(0.104856, 0.00214715)
(0.108123, 0.00218998)
(0.111492, 0.00223968)
(0.114966, 0.00228649)
(0.118548, 0.00238221)
(0.122242, 0.00242972)
(0.126051, 0.00251415)
(0.129978, 0.00257931)
(0.134028, 0.00264114)
(0.138204, 0.00267064)
(0.14251, 0.00274564)
(0.146951, 0.00284082)
(0.151529, 0.00291666)
(0.156251, 0.00298894)
(0.161119, 0.0030665)
(0.166139, 0.0031058)
(0.171316, 0.00322412)
(0.176654, 0.00329242)
(0.182158, 0.003396)
(0.187834, 0.00346013)
(0.193686, 0.00355545)
(0.199721, 0.00362607)
(0.205944, 0.00369222)
(0.212361, 0.00380229)
(0.218978, 0.00388306)
(0.225801, 0.00400571)
(0.232836, 0.0040922)
(0.240091, 0.00417238)
(0.247572, 0.00430482)
(0.255286, 0.00440548)
(0.26324, 0.0045222)
(0.271442, 0.00461621)
(0.279899, 0.00475394)
(0.288621, 0.00486475)
(0.297613, 0.00497788)
(0.306886, 0.00508216)
(0.316448, 0.00518211)
(0.326308, 0.00532461)
(0.336476, 0.00543717)
(0.346959, 0.00559913)
(0.35777, 0.00580098)
(0.368917, 0.005872)
(0.380412, 0.00600576)
(0.392265, 0.00612136)
(0.404487, 0.00632596)
(0.41709, 0.00646855)
(0.430086, 0.00657326)
(0.443487, 0.00677746)
(0.457305, 0.00692305)
(0.471554, 0.00712581)
(0.486247, 0.00720605)
(0.501397, 0.00736063)
(0.51702, 0.00750491)
(0.533129, 0.00774199)
(0.54974, 0.0078912)
(0.566869, 0.00809795)
(0.584532, 0.00833453)
(0.602745, 0.00846944)
(0.621525, 0.0087416)
(0.64089, 0.00887948)
(0.660859, 0.00907538)
(0.68145, 0.0093646)
(0.702683, 0.00951351)
(0.724577, 0.00970509)
(0.747154, 0.0100035)
(0.770434, 0.0102365)
(0.794439, 0.0104664)
(0.819192, 0.0108003)
(0.844717, 0.0109217)
(0.871036, 0.0110949)
(0.898176, 0.0114357)
(0.926162, 0.0116832)
(0.955019, 0.0119987)
(0.984776, 0.0122383)
(1.01546, 0.0125006)
(1.0471, 0.0128919)
(1.07972, 0.012957)
(1.11337, 0.0133632)
(1.14806, 0.0137144)
(1.18383, 0.0139722)
(1.22071, 0.0144174)
(1.25875, 0.0146094)
(1.29797, 0.014956)
(1.33841, 0.0153506)
(1.38011, 0.0156753)
(1.42312, 0.0161074)
(1.46746, 0.0164714)
(1.51318, 0.0167103)
(1.56033, 0.0171301)
(1.60895, 0.0177875)
(1.65908, 0.0181065)
(1.71077, 0.0183965)
(1.76408, 0.0186899)
(1.81904, 0.0192285)
(1.87572, 0.0195774)
(1.93416, 0.0200167)
(1.99443, 0.0206143)
(2.05657, 0.0209326)
(2.12065, 0.021548)
(2.18672, 0.0219857)
(2.25486, 0.0225977)
(2.32512, 0.0227367)
(2.39756, 0.0234287)
(2.47227, 0.0239437)
(2.5493, 0.0245277)
(2.62873, 0.0251075)
(2.71063, 0.0256681)
(2.79509, 0.02616)
(2.88218, 0.026681)
(2.97198, 0.027363)
(3.06459, 0.0279489)
(3.16007, 0.0287699)
(3.25853, 0.0293411)
(3.36006, 0.0298992)
(3.46476, 0.0307231)
(3.57271, 0.0312518)
(3.68403, 0.0319767)
(3.79882, 0.0326626)
(3.91718, 0.0334477)
(4.03923, 0.0341495)
(4.16509, 0.0349555)
(4.29487, 0.035557)
(4.42869, 0.0363248)
(4.56668, 0.0373909)
(4.70896, 0.0383563)
(4.85569, 0.0391414)
(5.00698, 0.0399281)
(5.16299, 0.0411172)
(5.32386, 0.0417189)
(5.48974, 0.0428251)
(5.66079, 0.0433092)
(5.83717, 0.0440416)
(6.01904, 0.0452889)
(6.20658, 0.0465216)
(6.39997, 0.0471886)
(6.59938, 0.0496934)
(6.805, 0.0507737)
(7.01704, 0.0516215)
(7.23567, 0.0521084)
(7.46112, 0.0534002)
(7.6936, 0.0543619)
(7.93332, 0.0553429)
(8.1805, 0.0567664)
(8.43539, 0.058136)
(8.69822, 0.0593846)
(8.96924, 0.0606025)
(9.24871, 0.0620949)
(9.53688, 0.0632952)
(9.83403, 0.0645158)
(10.1404, 0.0666052)
(10.4564, 0.0676068)
(10.7822, 0.0693079)
(11.1181, 0.0712708)
(11.4646, 0.072003)
(11.8218, 0.0744608)
(12.1901, 0.0751657)
(12.5699, 0.0772661)
(12.9616, 0.0787964)
(13.3655, 0.0804272)
(13.7819, 0.0824858)
(14.2113, 0.0843396)
(14.6541, 0.0861879)
(15.1107, 0.0882544)
(15.5815, 0.0904278)
(16.067, 0.0923324)
(16.5676, 0.0939657)
(17.0839, 0.0967106)
(17.6162, 0.0981644)
(18.165, 0.100051)
(18.731, 0.103909)
(19.3147, 0.10562)
(19.9165, 0.107997)
(20.537, 0.109629)
(21.1769, 0.112383)
(21.8368, 0.11539)
(22.5171, 0.117872)
(23.2187, 0.119954)
(23.9422, 0.122348)
(24.6882, 0.125509)
(25.4574, 0.127788)
(26.2506, 0.131009)
(27.0685, 0.135427)
(27.9119, 0.137049)
(28.7816, 0.139456)
(29.6784, 0.143708)
(30.6031, 0.146342)
(31.5567, 0.149322)
(32.5399, 0.153427)
(33.5538, 0.156523)
(34.5993, 0.159708)
(35.6773, 0.163403)
(36.789, 0.166684)
(37.9352, 0.171741)
(39.1172, 0.17539)
(40.336, 0.178977)
(41.5928, 0.183351)
(42.8888, 0.186569)
(44.2251, 0.190569)
(45.6031, 0.194695)
(47.024, 0.198388)
(48.4892, 0.205415)
(50, 0.207672)
}

\newcommand{\genOdePlotExpCifaruNcsnppErrEulerEHi}{%
(0.02, 0.000634272)
(0.0206232, 0.000646646)
(0.0212657, 0.000650287)
(0.0219283, 0.000686002)
(0.0226116, 0.000697045)
(0.0233161, 0.000713912)
(0.0240426, 0.000724357)
(0.0247917, 0.000738897)
(0.0255642, 0.000761438)
(0.0263607, 0.000794722)
(0.0271821, 0.000804769)
(0.028029, 0.000816684)
(0.0289023, 0.000845938)
(0.0298029, 0.000863611)
(0.0307315, 0.000886739)
(0.031689, 0.000912548)
(0.0326764, 0.000934666)
(0.0336945, 0.000954807)
(0.0347444, 0.00098002)
(0.035827, 0.00100511)
(0.0369432, 0.00103132)
(0.0380943, 0.00105408)
(0.0392813, 0.00107938)
(0.0405052, 0.00109605)
(0.0417673, 0.0011345)
(0.0430687, 0.00115903)
(0.0444106, 0.00118009)
(0.0457943, 0.00121672)
(0.0472212, 0.00125526)
(0.0486925, 0.00127867)
(0.0502097, 0.00132166)
(0.0517741, 0.00135011)
(0.0533873, 0.00137913)
(0.0550508, 0.00142988)
(0.056766, 0.00146073)
(0.0585348, 0.00147418)
(0.0603586, 0.0015494)
(0.0622393, 0.00155792)
(0.0641785, 0.00160274)
(0.0661782, 0.00164592)
(0.0682402, 0.00167924)
(0.0703664, 0.00170258)
(0.0725589, 0.00174701)
(0.0748197, 0.00180944)
(0.0771509, 0.00185142)
(0.0795548, 0.00187496)
(0.0820336, 0.00196299)
(0.0845896, 0.00197853)
(0.0872253, 0.00202341)
(0.089943, 0.00208764)
(0.0927455, 0.00216053)
(0.0956352, 0.00218778)
(0.0986151, 0.00226566)
(0.101688, 0.00228876)
(0.104856, 0.00236232)
(0.108123, 0.00238112)
(0.111492, 0.00244498)
(0.114966, 0.00247875)
(0.118548, 0.00259103)
(0.122242, 0.00264311)
(0.126051, 0.00272345)
(0.129978, 0.00278629)
(0.134028, 0.00286781)
(0.138204, 0.00288797)
(0.14251, 0.00296314)
(0.146951, 0.00303977)
(0.151529, 0.00312752)
(0.156251, 0.00320525)
(0.161119, 0.0032894)
(0.166139, 0.00332521)
(0.171316, 0.00343811)
(0.176654, 0.00352432)
(0.182158, 0.00361539)
(0.187834, 0.00368264)
(0.193686, 0.00379342)
(0.199721, 0.00385505)
(0.205944, 0.0039275)
(0.212361, 0.00403411)
(0.218978, 0.0041266)
(0.225801, 0.00425609)
(0.232836, 0.00431903)
(0.240091, 0.00438975)
(0.247572, 0.00454581)
(0.255286, 0.00463665)
(0.26324, 0.00474258)
(0.271442, 0.00483829)
(0.279899, 0.00497619)
(0.288621, 0.00511296)
(0.297613, 0.00524393)
(0.306886, 0.00534943)
(0.316448, 0.00542798)
(0.326308, 0.00557315)
(0.336476, 0.005678)
(0.346959, 0.00583003)
(0.35777, 0.00602218)
(0.368917, 0.00613809)
(0.380412, 0.00625843)
(0.392265, 0.006386)
(0.404487, 0.0065849)
(0.41709, 0.00672528)
(0.430086, 0.0068071)
(0.443487, 0.00704694)
(0.457305, 0.00722401)
(0.471554, 0.00738577)
(0.486247, 0.00749396)
(0.501397, 0.00762649)
(0.51702, 0.00776992)
(0.533129, 0.00802188)
(0.54974, 0.00818147)
(0.566869, 0.00840825)
(0.584532, 0.00860176)
(0.602745, 0.00875457)
(0.621525, 0.00900982)
(0.64089, 0.00918411)
(0.660859, 0.00936149)
(0.68145, 0.00968172)
(0.702683, 0.0098233)
(0.724577, 0.00998309)
(0.747154, 0.0102841)
(0.770434, 0.0104943)
(0.794439, 0.0108068)
(0.819192, 0.0111493)
(0.844717, 0.0112841)
(0.871036, 0.0114163)
(0.898176, 0.0117802)
(0.926162, 0.0120464)
(0.955019, 0.0123289)
(0.984776, 0.0126371)
(1.01546, 0.012869)
(1.0471, 0.0132596)
(1.07972, 0.0133526)
(1.11337, 0.0137266)
(1.14806, 0.0140998)
(1.18383, 0.0144147)
(1.22071, 0.0147442)
(1.25875, 0.0149713)
(1.29797, 0.015408)
(1.33841, 0.015726)
(1.38011, 0.0161665)
(1.42312, 0.0164683)
(1.46746, 0.0169204)
(1.51318, 0.0171398)
(1.56033, 0.0175253)
(1.60895, 0.0181911)
(1.65908, 0.0185735)
(1.71077, 0.0188365)
(1.76408, 0.0192143)
(1.81904, 0.0196478)
(1.87572, 0.0200068)
(1.93416, 0.020419)
(1.99443, 0.0210703)
(2.05657, 0.0214213)
(2.12065, 0.022051)
(2.18672, 0.0225134)
(2.25486, 0.0231373)
(2.32512, 0.0233448)
(2.39756, 0.0240825)
(2.47227, 0.024666)
(2.5493, 0.0251594)
(2.62873, 0.0258161)
(2.71063, 0.0263136)
(2.79509, 0.0267174)
(2.88218, 0.0273549)
(2.97198, 0.027952)
(3.06459, 0.0286491)
(3.16007, 0.029415)
(3.25853, 0.0300458)
(3.36006, 0.0308238)
(3.46476, 0.0313994)
(3.57271, 0.0320809)
(3.68403, 0.0327818)
(3.79882, 0.0335869)
(3.91718, 0.0343447)
(4.03923, 0.0350543)
(4.16509, 0.0358152)
(4.29487, 0.0362548)
(4.42869, 0.0372568)
(4.56668, 0.038333)
(4.70896, 0.0392265)
(4.85569, 0.0399757)
(5.00698, 0.0410232)
(5.16299, 0.0422046)
(5.32386, 0.0430277)
(5.48974, 0.0442689)
(5.66079, 0.044804)
(5.83717, 0.0456563)
(6.01904, 0.0470674)
(6.20658, 0.0482263)
(6.39997, 0.0491865)
(6.59938, 0.0504754)
(6.805, 0.051529)
(7.01704, 0.0522991)
(7.23567, 0.0532371)
(7.46112, 0.054677)
(7.6936, 0.0559665)
(7.93332, 0.056785)
(8.1805, 0.0584502)
(8.43539, 0.0599423)
(8.69822, 0.061088)
(8.96924, 0.0619124)
(9.24871, 0.0632774)
(9.53688, 0.064932)
(9.83403, 0.0662129)
(10.1404, 0.0683387)
(10.4564, 0.0695199)
(10.7822, 0.0711556)
(11.1181, 0.0729822)
(11.4646, 0.0738278)
(11.8218, 0.0763225)
(12.1901, 0.0770932)
(12.5699, 0.0791755)
(12.9616, 0.0808017)
(13.3655, 0.0831342)
(13.7819, 0.0851496)
(14.2113, 0.086783)
(14.6541, 0.0880423)
(15.1107, 0.0907527)
(15.5815, 0.0926375)
(16.067, 0.0942374)
(16.5676, 0.0957679)
(17.0839, 0.0992859)
(17.6162, 0.100852)
(18.165, 0.103038)
(18.731, 0.106945)
(19.3147, 0.107782)
(19.9165, 0.110577)
(20.537, 0.112856)
(21.1769, 0.114817)
(21.8368, 0.118099)
(22.5171, 0.121208)
(23.2187, 0.123259)
(23.9422, 0.125363)
(24.6882, 0.128901)
(25.4574, 0.131603)
(26.2506, 0.134374)
(27.0685, 0.138014)
(27.9119, 0.140538)
(28.7816, 0.142598)
(29.6784, 0.147523)
(30.6031, 0.150523)
(31.5567, 0.153096)
(32.5399, 0.157345)
(33.5538, 0.159956)
(34.5993, 0.163989)
(35.6773, 0.167827)
(36.789, 0.171879)
(37.9352, 0.176489)
(39.1172, 0.179847)
(40.336, 0.182705)
(41.5928, 0.187962)
(42.8888, 0.191184)
(44.2251, 0.195346)
(45.6031, 0.19951)
(47.024, 0.204439)
(48.4892, 0.211157)
(50, 0.214252)
}

\newcommand{\genOdePlotExpCifaruNcsnppErrHeunA}{%
(0.02, 14.7153)
(0.0206232, 13.9958)
(0.0212657, 14.3543)
(0.0219283, 14.1665)
(0.0226116, 13.2785)
(0.0233161, 12.6515)
(0.0240426, 12.7504)
(0.0247917, 12.3579)
(0.0255642, 13.8928)
(0.0263607, 12.3307)
(0.0271821, 12.0556)
(0.028029, 11.9846)
(0.0289023, 11.3647)
(0.0298029, 11.3169)
(0.0307315, 11.5309)
(0.031689, 10.8822)
(0.0326764, 11.3681)
(0.0336945, 11.3201)
(0.0347444, 10.3747)
(0.035827, 10.341)
(0.0369432, 10.642)
(0.0380943, 10.41)
(0.0392813, 10.7698)
(0.0405052, 11.1762)
(0.0417673, 10.7179)
(0.0430687, 10.9988)
(0.0444106, 10.4037)
(0.0457943, 10.549)
(0.0472212, 10.5304)
(0.0486925, 10.1366)
(0.0502097, 9.80554)
(0.0517741, 9.1021)
(0.0533873, 9.23381)
(0.0550508, 8.99691)
(0.056766, 8.3778)
(0.0585348, 7.68808)
(0.0603586, 6.92147)
(0.0622393, 7.20095)
(0.0641785, 6.30628)
(0.0661782, 5.89918)
(0.0682402, 5.76632)
(0.0703664, 5.43338)
(0.0725589, 5.24325)
(0.0748197, 4.99659)
(0.0771509, 4.84574)
(0.0795548, 4.71324)
(0.0820336, 4.58434)
(0.0845896, 4.43979)
(0.0872253, 4.28375)
(0.089943, 4.16398)
(0.0927455, 4.02756)
(0.0956352, 3.88955)
(0.0986151, 3.7657)
(0.101688, 3.65794)
(0.104856, 3.53154)
(0.108123, 3.42236)
(0.111492, 3.30111)
(0.114966, 3.19702)
(0.118548, 3.11066)
(0.122242, 3.02545)
(0.126051, 2.91435)
(0.129978, 2.80986)
(0.134028, 2.73986)
(0.138204, 2.64161)
(0.14251, 2.56863)
(0.146951, 2.4993)
(0.151529, 2.40623)
(0.156251, 2.34)
(0.161119, 2.26399)
(0.166139, 2.18357)
(0.171316, 2.12296)
(0.176654, 2.05935)
(0.182158, 2.01378)
(0.187834, 1.93617)
(0.193686, 1.89594)
(0.199721, 1.82272)
(0.205944, 1.78035)
(0.212361, 1.73013)
(0.218978, 1.66259)
(0.225801, 1.62731)
(0.232836, 1.57181)
(0.240091, 1.51303)
(0.247572, 1.47665)
(0.255286, 1.4329)
(0.26324, 1.38342)
(0.271442, 1.3368)
(0.279899, 1.29483)
(0.288621, 1.25611)
(0.297613, 1.21815)
(0.306886, 1.17909)
(0.316448, 1.12855)
(0.326308, 1.10515)
(0.336476, 1.05835)
(0.346959, 1.02188)
(0.35777, 0.989489)
(0.368917, 0.953977)
(0.380412, 0.922617)
(0.392265, 0.893621)
(0.404487, 0.856028)
(0.41709, 0.828367)
(0.430086, 0.796376)
(0.443487, 0.770074)
(0.457305, 0.743634)
(0.471554, 0.717299)
(0.486247, 0.698365)
(0.501397, 0.670965)
(0.51702, 0.64564)
(0.533129, 0.616975)
(0.54974, 0.598488)
(0.566869, 0.576554)
(0.584532, 0.555868)
(0.602745, 0.53735)
(0.621525, 0.518629)
(0.64089, 0.499135)
(0.660859, 0.480897)
(0.68145, 0.464288)
(0.702683, 0.447211)
(0.724577, 0.428734)
(0.747154, 0.414887)
(0.770434, 0.400976)
(0.794439, 0.386732)
(0.819192, 0.370664)
(0.844717, 0.359177)
(0.871036, 0.347075)
(0.898176, 0.331987)
(0.926162, 0.323631)
(0.955019, 0.311039)
(0.984776, 0.300335)
(1.01546, 0.2895)
(1.0471, 0.279755)
(1.07972, 0.268823)
(1.11337, 0.26036)
(1.14806, 0.251077)
(1.18383, 0.240224)
(1.22071, 0.229793)
(1.25875, 0.221524)
(1.29797, 0.213626)
(1.33841, 0.20696)
(1.38011, 0.196125)
(1.42312, 0.189647)
(1.46746, 0.182862)
(1.51318, 0.17466)
(1.56033, 0.170409)
(1.60895, 0.161383)
(1.65908, 0.156057)
(1.71077, 0.148286)
(1.76408, 0.142308)
(1.81904, 0.137321)
(1.87572, 0.131108)
(1.93416, 0.127757)
(1.99443, 0.121237)
(2.05657, 0.117583)
(2.12065, 0.112191)
(2.18672, 0.107028)
(2.25486, 0.104107)
(2.32512, 0.101252)
(2.39756, 0.0969602)
(2.47227, 0.0931028)
(2.5493, 0.0896788)
(2.62873, 0.0855135)
(2.71063, 0.0819293)
(2.79509, 0.0796096)
(2.88218, 0.0759425)
(2.97198, 0.073327)
(3.06459, 0.0695099)
(3.16007, 0.067629)
(3.25853, 0.0640303)
(3.36006, 0.061857)
(3.46476, 0.0583011)
(3.57271, 0.0560918)
(3.68403, 0.053509)
(3.79882, 0.0513739)
(3.91718, 0.0484436)
(4.03923, 0.0466783)
(4.16509, 0.0443585)
(4.29487, 0.0417586)
(4.42869, 0.0400597)
(4.56668, 0.0380599)
(4.70896, 0.0359894)
(4.85569, 0.034108)
(5.00698, 0.0327165)
(5.16299, 0.0312636)
(5.32386, 0.0294903)
(5.48974, 0.0278709)
(5.66079, 0.0262376)
(5.83717, 0.0247208)
(6.01904, 0.02329)
(6.20658, 0.0217548)
(6.39997, 0.020857)
(6.59938, 0.0199258)
(6.805, 0.0185003)
(7.01704, 0.0173918)
(7.23567, 0.0168471)
(7.46112, 0.0163777)
(7.6936, 0.0155819)
(7.93332, 0.0144934)
(8.1805, 0.0140492)
(8.43539, 0.0128967)
(8.69822, 0.0121073)
(8.96924, 0.0119441)
(9.24871, 0.0109495)
(9.53688, 0.0103212)
(9.83403, 0.00959469)
(10.1404, 0.00905577)
(10.4564, 0.00888397)
(10.7822, 0.00836449)
(11.1181, 0.00755533)
(11.4646, 0.00712116)
(11.8218, 0.00696493)
(12.1901, 0.00669253)
(12.5699, 0.00582061)
(12.9616, 0.00587725)
(13.3655, 0.00539384)
(13.7819, 0.00472315)
(14.2113, 0.00493815)
(14.6541, 0.00441839)
(15.1107, 0.00418582)
(15.5815, 0.00400604)
(16.067, 0.00376747)
(16.5676, 0.00394987)
(17.0839, 0.00378026)
(17.6162, 0.0030289)
(18.165, 0.0026843)
(18.731, 0.00326781)
(19.3147, 0.00277578)
(19.9165, 0.00256956)
(20.537, 0.00271503)
(21.1769, 0.00242492)
(21.8368, 0.00231366)
(22.5171, 0.00220631)
(23.2187, 0.00193981)
(23.9422, 0.0018638)
(24.6882, 0.00182257)
(25.4574, 0.00164423)
(26.2506, 0.00151279)
(27.0685, 0.00141704)
(27.9119, 0.00123296)
(28.7816, 0.00129521)
(29.6784, 0.00119286)
(30.6031, 0.00132466)
(31.5567, 0.000872198)
(32.5399, 0.00113948)
(33.5538, 0.000976134)
(34.5993, 0.000882942)
(35.6773, 0.000946234)
(36.789, 0.000779625)
(37.9352, 0.000712097)
(39.1172, 0.000612112)
(40.336, 0.000295203)
(41.5928, 0.000563782)
(42.8888, 0.000414814)
(44.2251, 0.00051364)
(45.6031, 0.000506752)
(47.024, 0.000767715)
(48.4892, 0.000279555)
(50, 0.000674325)
}

\newcommand{\genOdePlotExpCifaruNcsnppErrHeunALo}{%
(0.02, 13.1878)
(0.0206232, 12.5453)
(0.0212657, 12.796)
(0.0219283, 12.3741)
(0.0226116, 11.773)
(0.0233161, 11.2818)
(0.0240426, 11.4015)
(0.0247917, 11.0283)
(0.0255642, 11.9624)
(0.0263607, 10.6357)
(0.0271821, 10.5333)
(0.028029, 10.3217)
(0.0289023, 9.86992)
(0.0298029, 9.76386)
(0.0307315, 10.1115)
(0.031689, 9.28393)
(0.0326764, 9.70861)
(0.0336945, 9.65467)
(0.0347444, 8.79914)
(0.035827, 8.68009)
(0.0369432, 9.07203)
(0.0380943, 8.85124)
(0.0392813, 9.21678)
(0.0405052, 9.74096)
(0.0417673, 9.22622)
(0.0430687, 9.8521)
(0.0444106, 9.13739)
(0.0457943, 9.4655)
(0.0472212, 9.71529)
(0.0486925, 9.28499)
(0.0502097, 8.97792)
(0.0517741, 8.22904)
(0.0533873, 8.51161)
(0.0550508, 8.27587)
(0.056766, 7.63098)
(0.0585348, 6.84181)
(0.0603586, 6.26614)
(0.0622393, 6.43741)
(0.0641785, 5.78337)
(0.0661782, 5.47616)
(0.0682402, 5.35792)
(0.0703664, 5.05963)
(0.0725589, 4.85798)
(0.0748197, 4.70958)
(0.0771509, 4.53346)
(0.0795548, 4.38584)
(0.0820336, 4.25227)
(0.0845896, 4.12063)
(0.0872253, 3.97755)
(0.089943, 3.8487)
(0.0927455, 3.74239)
(0.0956352, 3.62891)
(0.0986151, 3.52184)
(0.101688, 3.42198)
(0.104856, 3.30208)
(0.108123, 3.18077)
(0.111492, 3.07427)
(0.114966, 2.97567)
(0.118548, 2.89202)
(0.122242, 2.81487)
(0.126051, 2.71737)
(0.129978, 2.64387)
(0.134028, 2.54785)
(0.138204, 2.46619)
(0.14251, 2.39408)
(0.146951, 2.32947)
(0.151529, 2.25276)
(0.156251, 2.18679)
(0.161119, 2.10493)
(0.166139, 2.0367)
(0.171316, 1.99779)
(0.176654, 1.94623)
(0.182158, 1.88744)
(0.187834, 1.81458)
(0.193686, 1.76723)
(0.199721, 1.6982)
(0.205944, 1.65757)
(0.212361, 1.62566)
(0.218978, 1.55959)
(0.225801, 1.53186)
(0.232836, 1.46884)
(0.240091, 1.4274)
(0.247572, 1.39048)
(0.255286, 1.34963)
(0.26324, 1.30091)
(0.271442, 1.25529)
(0.279899, 1.2158)
(0.288621, 1.1794)
(0.297613, 1.14222)
(0.306886, 1.11112)
(0.316448, 1.06436)
(0.326308, 1.0456)
(0.336476, 0.998706)
(0.346959, 0.966318)
(0.35777, 0.941016)
(0.368917, 0.905321)
(0.380412, 0.874257)
(0.392265, 0.84502)
(0.404487, 0.812729)
(0.41709, 0.788173)
(0.430086, 0.757017)
(0.443487, 0.73122)
(0.457305, 0.705547)
(0.471554, 0.684423)
(0.486247, 0.666576)
(0.501397, 0.637955)
(0.51702, 0.613704)
(0.533129, 0.590014)
(0.54974, 0.573403)
(0.566869, 0.549295)
(0.584532, 0.529203)
(0.602745, 0.513493)
(0.621525, 0.494623)
(0.64089, 0.477258)
(0.660859, 0.45937)
(0.68145, 0.443432)
(0.702683, 0.427731)
(0.724577, 0.410089)
(0.747154, 0.395721)
(0.770434, 0.383557)
(0.794439, 0.367126)
(0.819192, 0.354337)
(0.844717, 0.343753)
(0.871036, 0.330877)
(0.898176, 0.316962)
(0.926162, 0.308272)
(0.955019, 0.295635)
(0.984776, 0.285154)
(1.01546, 0.276616)
(1.0471, 0.267697)
(1.07972, 0.256654)
(1.11337, 0.250369)
(1.14806, 0.24118)
(1.18383, 0.229636)
(1.22071, 0.220749)
(1.25875, 0.213331)
(1.29797, 0.204712)
(1.33841, 0.198048)
(1.38011, 0.187794)
(1.42312, 0.180803)
(1.46746, 0.174473)
(1.51318, 0.166551)
(1.56033, 0.162337)
(1.60895, 0.153683)
(1.65908, 0.147964)
(1.71077, 0.140299)
(1.76408, 0.134265)
(1.81904, 0.128959)
(1.87572, 0.122327)
(1.93416, 0.119512)
(1.99443, 0.113154)
(2.05657, 0.109677)
(2.12065, 0.104174)
(2.18672, 0.0993105)
(2.25486, 0.0963464)
(2.32512, 0.0929424)
(2.39756, 0.0894993)
(2.47227, 0.0852816)
(2.5493, 0.082206)
(2.62873, 0.0792237)
(2.71063, 0.0751608)
(2.79509, 0.0731311)
(2.88218, 0.0696259)
(2.97198, 0.0676623)
(3.06459, 0.063602)
(3.16007, 0.0620209)
(3.25853, 0.0586074)
(3.36006, 0.0568756)
(3.46476, 0.0532895)
(3.57271, 0.0515598)
(3.68403, 0.0496069)
(3.79882, 0.0473604)
(3.91718, 0.0441355)
(4.03923, 0.0432531)
(4.16509, 0.0412654)
(4.29487, 0.0389982)
(4.42869, 0.0373123)
(4.56668, 0.0351882)
(4.70896, 0.0336881)
(4.85569, 0.031685)
(5.00698, 0.0302624)
(5.16299, 0.0292067)
(5.32386, 0.0270798)
(5.48974, 0.0255852)
(5.66079, 0.0240097)
(5.83717, 0.0228207)
(6.01904, 0.0218487)
(6.20658, 0.0206068)
(6.39997, 0.0196423)
(6.59938, 0.0186135)
(6.805, 0.017108)
(7.01704, 0.0160046)
(7.23567, 0.0152612)
(7.46112, 0.0154905)
(7.6936, 0.0148047)
(7.93332, 0.0137171)
(8.1805, 0.0134952)
(8.43539, 0.0121531)
(8.69822, 0.0115811)
(8.96924, 0.0115481)
(9.24871, 0.0106081)
(9.53688, 0.0100492)
(9.83403, 0.00910603)
(10.1404, 0.00877643)
(10.4564, 0.0086817)
(10.7822, 0.00816328)
(11.1181, 0.00736636)
(11.4646, 0.00684613)
(11.8218, 0.00666308)
(12.1901, 0.00648675)
(12.5699, 0.00564174)
(12.9616, 0.00558997)
(13.3655, 0.0052001)
(13.7819, 0.00462227)
(14.2113, 0.00476562)
(14.6541, 0.00412225)
(15.1107, 0.00392952)
(15.5815, 0.00382209)
(16.067, 0.0036858)
(16.5676, 0.00381349)
(17.0839, 0.00371356)
(17.6162, 0.00285316)
(18.165, 0.00244048)
(18.731, 0.00315355)
(19.3147, 0.00265114)
(19.9165, 0.00246359)
(20.537, 0.002365)
(21.1769, 0.00213553)
(21.8368, 0.00219191)
(22.5171, 0.00209475)
(23.2187, 0.00175974)
(23.9422, 0.00162745)
(24.6882, 0.00156242)
(25.4574, 0.00158952)
(26.2506, 0.00137344)
(27.0685, 0.00115861)
(27.9119, 0.00102645)
(28.7816, 0.00103312)
(29.6784, 0.00105525)
(30.6031, 0.00127291)
(31.5567, 0.000764973)
(32.5399, 0.000942)
(33.5538, 0.000934272)
(34.5993, 0.000807266)
(35.6773, 0.000829332)
(36.789, 0.000746907)
(37.9352, 0.000681089)
(39.1172, 0.000551169)
(40.336, 0.000263623)
(41.5928, 0.000513384)
(42.8888, 0.000249501)
(44.2251, 0.000416135)
(45.6031, 0.000483389)
(47.024, 0.000712871)
(48.4892, 0.000227145)
(50, 0.000607769)
}

\newcommand{\genOdePlotExpCifaruNcsnppErrHeunAHi}{%
(0.02, 16.2428)
(0.0206232, 15.4462)
(0.0212657, 15.9126)
(0.0219283, 15.959)
(0.0226116, 14.784)
(0.0233161, 14.0212)
(0.0240426, 14.0992)
(0.0247917, 13.6875)
(0.0255642, 15.8232)
(0.0263607, 14.0257)
(0.0271821, 13.578)
(0.028029, 13.6475)
(0.0289023, 12.8594)
(0.0298029, 12.87)
(0.0307315, 12.9504)
(0.031689, 12.4804)
(0.0326764, 13.0276)
(0.0336945, 12.9854)
(0.0347444, 11.9502)
(0.035827, 12.002)
(0.0369432, 12.2119)
(0.0380943, 11.9688)
(0.0392813, 12.3229)
(0.0405052, 12.6114)
(0.0417673, 12.2096)
(0.0430687, 12.1455)
(0.0444106, 11.6701)
(0.0457943, 11.6324)
(0.0472212, 11.3455)
(0.0486925, 10.9882)
(0.0502097, 10.6332)
(0.0517741, 9.97516)
(0.0533873, 9.95602)
(0.0550508, 9.71794)
(0.056766, 9.12462)
(0.0585348, 8.53435)
(0.0603586, 7.57681)
(0.0622393, 7.9645)
(0.0641785, 6.82919)
(0.0661782, 6.3222)
(0.0682402, 6.17472)
(0.0703664, 5.80713)
(0.0725589, 5.62853)
(0.0748197, 5.2836)
(0.0771509, 5.15801)
(0.0795548, 5.04065)
(0.0820336, 4.9164)
(0.0845896, 4.75896)
(0.0872253, 4.58994)
(0.089943, 4.47927)
(0.0927455, 4.31273)
(0.0956352, 4.15018)
(0.0986151, 4.00956)
(0.101688, 3.89391)
(0.104856, 3.76099)
(0.108123, 3.66396)
(0.111492, 3.52794)
(0.114966, 3.41836)
(0.118548, 3.3293)
(0.122242, 3.23602)
(0.126051, 3.11133)
(0.129978, 2.97585)
(0.134028, 2.93187)
(0.138204, 2.81702)
(0.14251, 2.74318)
(0.146951, 2.66913)
(0.151529, 2.5597)
(0.156251, 2.4932)
(0.161119, 2.42306)
(0.166139, 2.33045)
(0.171316, 2.24814)
(0.176654, 2.17246)
(0.182158, 2.14013)
(0.187834, 2.05777)
(0.193686, 2.02465)
(0.199721, 1.94724)
(0.205944, 1.90313)
(0.212361, 1.8346)
(0.218978, 1.76559)
(0.225801, 1.72276)
(0.232836, 1.67478)
(0.240091, 1.59865)
(0.247572, 1.56281)
(0.255286, 1.51617)
(0.26324, 1.46593)
(0.271442, 1.41832)
(0.279899, 1.37387)
(0.288621, 1.33282)
(0.297613, 1.29408)
(0.306886, 1.24705)
(0.316448, 1.19274)
(0.326308, 1.16469)
(0.336476, 1.11799)
(0.346959, 1.07745)
(0.35777, 1.03796)
(0.368917, 1.00263)
(0.380412, 0.970978)
(0.392265, 0.942223)
(0.404487, 0.899327)
(0.41709, 0.868561)
(0.430086, 0.835736)
(0.443487, 0.808928)
(0.457305, 0.781721)
(0.471554, 0.750174)
(0.486247, 0.730155)
(0.501397, 0.703975)
(0.51702, 0.677576)
(0.533129, 0.643936)
(0.54974, 0.623572)
(0.566869, 0.603813)
(0.584532, 0.582533)
(0.602745, 0.561207)
(0.621525, 0.542635)
(0.64089, 0.521012)
(0.660859, 0.502425)
(0.68145, 0.485143)
(0.702683, 0.466691)
(0.724577, 0.44738)
(0.747154, 0.434053)
(0.770434, 0.418395)
(0.794439, 0.406339)
(0.819192, 0.386991)
(0.844717, 0.374601)
(0.871036, 0.363273)
(0.898176, 0.347012)
(0.926162, 0.338991)
(0.955019, 0.326442)
(0.984776, 0.315516)
(1.01546, 0.302384)
(1.0471, 0.291813)
(1.07972, 0.280991)
(1.11337, 0.270351)
(1.14806, 0.260973)
(1.18383, 0.250812)
(1.22071, 0.238838)
(1.25875, 0.229717)
(1.29797, 0.222541)
(1.33841, 0.215873)
(1.38011, 0.204456)
(1.42312, 0.198492)
(1.46746, 0.19125)
(1.51318, 0.182769)
(1.56033, 0.178481)
(1.60895, 0.169083)
(1.65908, 0.16415)
(1.71077, 0.156272)
(1.76408, 0.150351)
(1.81904, 0.145684)
(1.87572, 0.13989)
(1.93416, 0.136002)
(1.99443, 0.129321)
(2.05657, 0.125489)
(2.12065, 0.120208)
(2.18672, 0.114746)
(2.25486, 0.111867)
(2.32512, 0.109561)
(2.39756, 0.104421)
(2.47227, 0.100924)
(2.5493, 0.0971516)
(2.62873, 0.0918032)
(2.71063, 0.0886977)
(2.79509, 0.0860881)
(2.88218, 0.082259)
(2.97198, 0.0789917)
(3.06459, 0.0754177)
(3.16007, 0.073237)
(3.25853, 0.0694532)
(3.36006, 0.0668385)
(3.46476, 0.0633128)
(3.57271, 0.0606237)
(3.68403, 0.0574111)
(3.79882, 0.0553874)
(3.91718, 0.0527517)
(4.03923, 0.0501036)
(4.16509, 0.0474515)
(4.29487, 0.0445191)
(4.42869, 0.042807)
(4.56668, 0.0409315)
(4.70896, 0.0382906)
(4.85569, 0.036531)
(5.00698, 0.0351707)
(5.16299, 0.0333205)
(5.32386, 0.0319008)
(5.48974, 0.0301565)
(5.66079, 0.0284656)
(5.83717, 0.0266208)
(6.01904, 0.0247312)
(6.20658, 0.0229027)
(6.39997, 0.0220717)
(6.59938, 0.021238)
(6.805, 0.0198925)
(7.01704, 0.018779)
(7.23567, 0.018433)
(7.46112, 0.0172649)
(7.6936, 0.016359)
(7.93332, 0.0152696)
(8.1805, 0.0146031)
(8.43539, 0.0136404)
(8.69822, 0.0126335)
(8.96924, 0.0123401)
(9.24871, 0.0112908)
(9.53688, 0.0105932)
(9.83403, 0.0100833)
(10.1404, 0.00933511)
(10.4564, 0.00908624)
(10.7822, 0.00856569)
(11.1181, 0.0077443)
(11.4646, 0.00739619)
(11.8218, 0.00726678)
(12.1901, 0.00689831)
(12.5699, 0.00599948)
(12.9616, 0.00616453)
(13.3655, 0.00558759)
(13.7819, 0.00482403)
(14.2113, 0.00511068)
(14.6541, 0.00471452)
(15.1107, 0.00444213)
(15.5815, 0.00418999)
(16.067, 0.00384914)
(16.5676, 0.00408624)
(17.0839, 0.00384696)
(17.6162, 0.00320464)
(18.165, 0.00292813)
(18.731, 0.00338206)
(19.3147, 0.00290043)
(19.9165, 0.00267553)
(20.537, 0.00306505)
(21.1769, 0.00271432)
(21.8368, 0.00243541)
(22.5171, 0.00231787)
(23.2187, 0.00211987)
(23.9422, 0.00210014)
(24.6882, 0.00208272)
(25.4574, 0.00169893)
(26.2506, 0.00165213)
(27.0685, 0.00167547)
(27.9119, 0.00143947)
(28.7816, 0.00155729)
(29.6784, 0.00133047)
(30.6031, 0.0013764)
(31.5567, 0.000979422)
(32.5399, 0.00133696)
(33.5538, 0.001018)
(34.5993, 0.000958617)
(35.6773, 0.00106314)
(36.789, 0.000812343)
(37.9352, 0.000743104)
(39.1172, 0.000673054)
(40.336, 0.000326782)
(41.5928, 0.00061418)
(42.8888, 0.000580126)
(44.2251, 0.000611146)
(45.6031, 0.000530115)
(47.024, 0.00082256)
(48.4892, 0.000331966)
(50, 0.00074088)
}

\newcommand{\genOdePlotExpCifaruNcsnppErrHeunB}{%
(0.02, 0.289735)
(0.0206232, 0.283797)
(0.0212657, 0.266604)
(0.0219283, 0.261766)
(0.0226116, 0.250753)
(0.0233161, 0.259861)
(0.0240426, 0.246767)
(0.0247917, 0.244557)
(0.0255642, 0.241731)
(0.0263607, 0.234657)
(0.0271821, 0.229815)
(0.028029, 0.229918)
(0.0289023, 0.217322)
(0.0298029, 0.218655)
(0.0307315, 0.213991)
(0.031689, 0.216085)
(0.0326764, 0.208976)
(0.0336945, 0.2079)
(0.0347444, 0.201727)
(0.035827, 0.200763)
(0.0369432, 0.194005)
(0.0380943, 0.192801)
(0.0392813, 0.186571)
(0.0405052, 0.189502)
(0.0417673, 0.182161)
(0.0430687, 0.176924)
(0.0444106, 0.175404)
(0.0457943, 0.170307)
(0.0472212, 0.172796)
(0.0486925, 0.166433)
(0.0502097, 0.165275)
(0.0517741, 0.162771)
(0.0533873, 0.160529)
(0.0550508, 0.159379)
(0.056766, 0.153343)
(0.0585348, 0.154113)
(0.0603586, 0.155002)
(0.0622393, 0.148784)
(0.0641785, 0.146907)
(0.0661782, 0.146671)
(0.0682402, 0.143188)
(0.0703664, 0.14374)
(0.0725589, 0.144006)
(0.0748197, 0.13979)
(0.0771509, 0.137802)
(0.0795548, 0.137341)
(0.0820336, 0.134914)
(0.0845896, 0.135193)
(0.0872253, 0.13073)
(0.089943, 0.130235)
(0.0927455, 0.127253)
(0.0956352, 0.128593)
(0.0986151, 0.128171)
(0.101688, 0.123871)
(0.104856, 0.125156)
(0.108123, 0.121047)
(0.111492, 0.121811)
(0.114966, 0.11903)
(0.118548, 0.116041)
(0.122242, 0.116199)
(0.126051, 0.117073)
(0.129978, 0.111634)
(0.134028, 0.113716)
(0.138204, 0.11178)
(0.14251, 0.11046)
(0.146951, 0.109804)
(0.151529, 0.106898)
(0.156251, 0.104987)
(0.161119, 0.104177)
(0.166139, 0.102646)
(0.171316, 0.102733)
(0.176654, 0.101712)
(0.182158, 0.0991314)
(0.187834, 0.0978528)
(0.193686, 0.0966552)
(0.199721, 0.0947273)
(0.205944, 0.0948149)
(0.212361, 0.0925463)
(0.218978, 0.0908598)
(0.225801, 0.0891718)
(0.232836, 0.0882454)
(0.240091, 0.0852112)
(0.247572, 0.0854406)
(0.255286, 0.0839943)
(0.26324, 0.0830568)
(0.271442, 0.0815612)
(0.279899, 0.0795033)
(0.288621, 0.0789836)
(0.297613, 0.0771192)
(0.306886, 0.0749085)
(0.316448, 0.0743238)
(0.326308, 0.0723872)
(0.336476, 0.0701554)
(0.346959, 0.0696077)
(0.35777, 0.0675784)
(0.368917, 0.0662162)
(0.380412, 0.0651658)
(0.392265, 0.0634628)
(0.404487, 0.0616711)
(0.41709, 0.0602525)
(0.430086, 0.058925)
(0.443487, 0.0578351)
(0.457305, 0.0566095)
(0.471554, 0.0555166)
(0.486247, 0.053969)
(0.501397, 0.0520946)
(0.51702, 0.051334)
(0.533129, 0.0496825)
(0.54974, 0.0482972)
(0.566869, 0.047384)
(0.584532, 0.0463047)
(0.602745, 0.0450603)
(0.621525, 0.0439476)
(0.64089, 0.0428439)
(0.660859, 0.041688)
(0.68145, 0.0408602)
(0.702683, 0.039994)
(0.724577, 0.0391791)
(0.747154, 0.0381465)
(0.770434, 0.0371008)
(0.794439, 0.0363812)
(0.819192, 0.0355414)
(0.844717, 0.0348158)
(0.871036, 0.0341209)
(0.898176, 0.0331325)
(0.926162, 0.032647)
(0.955019, 0.0316235)
(0.984776, 0.0309451)
(1.01546, 0.0305538)
(1.0471, 0.0296473)
(1.07972, 0.0292139)
(1.11337, 0.0284146)
(1.14806, 0.027738)
(1.18383, 0.0269461)
(1.22071, 0.0262353)
(1.25875, 0.0257078)
(1.29797, 0.0252773)
(1.33841, 0.024682)
(1.38011, 0.0240822)
(1.42312, 0.0234377)
(1.46746, 0.0232823)
(1.51318, 0.0223222)
(1.56033, 0.0219634)
(1.60895, 0.0213786)
(1.65908, 0.0209034)
(1.71077, 0.0204791)
(1.76408, 0.0200066)
(1.81904, 0.0197227)
(1.87572, 0.0187287)
(1.93416, 0.0183688)
(1.99443, 0.0181346)
(2.05657, 0.0178039)
(2.12065, 0.0173038)
(2.18672, 0.0168465)
(2.25486, 0.0166197)
(2.32512, 0.0161227)
(2.39756, 0.0157758)
(2.47227, 0.0153363)
(2.5493, 0.0148643)
(2.62873, 0.0144768)
(2.71063, 0.0142103)
(2.79509, 0.0138721)
(2.88218, 0.0133913)
(2.97198, 0.0131133)
(3.06459, 0.0126242)
(3.16007, 0.0123397)
(3.25853, 0.0119882)
(3.36006, 0.0118488)
(3.46476, 0.0115342)
(3.57271, 0.0110798)
(3.68403, 0.0107412)
(3.79882, 0.0106099)
(3.91718, 0.0103391)
(4.03923, 0.0101013)
(4.16509, 0.00995198)
(4.29487, 0.00941902)
(4.42869, 0.00920402)
(4.56668, 0.00896731)
(4.70896, 0.00854633)
(4.85569, 0.00847855)
(5.00698, 0.00833612)
(5.16299, 0.00810545)
(5.32386, 0.00773926)
(5.48974, 0.0075367)
(5.66079, 0.00752437)
(5.83717, 0.00728202)
(6.01904, 0.00679154)
(6.20658, 0.00650263)
(6.39997, 0.00657975)
(6.59938, 0.00616904)
(6.805, 0.00601941)
(7.01704, 0.00579277)
(7.23567, 0.00589461)
(7.46112, 0.00566594)
(7.6936, 0.00553852)
(7.93332, 0.00533973)
(8.1805, 0.00538301)
(8.43539, 0.00495563)
(8.69822, 0.00490816)
(8.96924, 0.00483199)
(9.24871, 0.00479559)
(9.53688, 0.00449036)
(9.83403, 0.00442062)
(10.1404, 0.0042301)
(10.4564, 0.0040678)
(10.7822, 0.00415319)
(11.1181, 0.00382122)
(11.4646, 0.00342349)
(11.8218, 0.00381732)
(12.1901, 0.00359558)
(12.5699, 0.00359954)
(12.9616, 0.00347744)
(13.3655, 0.00346559)
(13.7819, 0.00293088)
(14.2113, 0.00303961)
(14.6541, 0.00299209)
(15.1107, 0.00304985)
(15.5815, 0.0028407)
(16.067, 0.00266375)
(16.5676, 0.00282096)
(17.0839, 0.00272268)
(17.6162, 0.00247384)
(18.165, 0.00213458)
(18.731, 0.00260691)
(19.3147, 0.00224218)
(19.9165, 0.00231027)
(20.537, 0.00240124)
(21.1769, 0.00220831)
(21.8368, 0.0022041)
(22.5171, 0.00215019)
(23.2187, 0.00195075)
(23.9422, 0.00189607)
(24.6882, 0.00192212)
(25.4574, 0.0017525)
(26.2506, 0.00173112)
(27.0685, 0.00164442)
(27.9119, 0.00135987)
(28.7816, 0.00176055)
(29.6784, 0.00150472)
(30.6031, 0.00153209)
(31.5567, 0.00133406)
(32.5399, 0.00154633)
(33.5538, 0.00135428)
(34.5993, 0.00133903)
(35.6773, 0.0014574)
(36.789, 0.00128676)
(37.9352, 0.00108024)
(39.1172, 0.0014667)
(40.336, 0.000775924)
(41.5928, 0.000942086)
(42.8888, 0.00110788)
(44.2251, 0.000846213)
(45.6031, 0.000713942)
(47.024, 0.00132915)
(48.4892, 0.000857046)
(50, 0.00111044)
}

\newcommand{\genOdePlotExpCifaruNcsnppErrHeunBLo}{%
(0.02, 0.223398)
(0.0206232, 0.213347)
(0.0212657, 0.214532)
(0.0219283, 0.208424)
(0.0226116, 0.202857)
(0.0233161, 0.19705)
(0.0240426, 0.19943)
(0.0247917, 0.189685)
(0.0255642, 0.189618)
(0.0263607, 0.180432)
(0.0271821, 0.178335)
(0.028029, 0.176048)
(0.0289023, 0.172192)
(0.0298029, 0.170577)
(0.0307315, 0.165077)
(0.031689, 0.167815)
(0.0326764, 0.158497)
(0.0336945, 0.156849)
(0.0347444, 0.16048)
(0.035827, 0.158988)
(0.0369432, 0.153581)
(0.0380943, 0.14795)
(0.0392813, 0.145192)
(0.0405052, 0.141936)
(0.0417673, 0.142971)
(0.0430687, 0.140711)
(0.0444106, 0.136688)
(0.0457943, 0.136669)
(0.0472212, 0.136205)
(0.0486925, 0.134683)
(0.0502097, 0.133775)
(0.0517741, 0.133792)
(0.0533873, 0.130867)
(0.0550508, 0.12735)
(0.056766, 0.123981)
(0.0585348, 0.122969)
(0.0603586, 0.124501)
(0.0622393, 0.122298)
(0.0641785, 0.120602)
(0.0661782, 0.11879)
(0.0682402, 0.118734)
(0.0703664, 0.117808)
(0.0725589, 0.118404)
(0.0748197, 0.113414)
(0.0771509, 0.112872)
(0.0795548, 0.114577)
(0.0820336, 0.111508)
(0.0845896, 0.113228)
(0.0872253, 0.109641)
(0.089943, 0.110123)
(0.0927455, 0.109032)
(0.0956352, 0.107214)
(0.0986151, 0.10776)
(0.101688, 0.106089)
(0.104856, 0.106537)
(0.108123, 0.101795)
(0.111492, 0.103206)
(0.114966, 0.102015)
(0.118548, 0.100233)
(0.122242, 0.100701)
(0.126051, 0.101464)
(0.129978, 0.0970945)
(0.134028, 0.0983827)
(0.138204, 0.0954412)
(0.14251, 0.0966084)
(0.146951, 0.0945877)
(0.151529, 0.0930552)
(0.156251, 0.0919048)
(0.161119, 0.0912613)
(0.166139, 0.0905025)
(0.171316, 0.0895601)
(0.176654, 0.0885329)
(0.182158, 0.0873381)
(0.187834, 0.0863418)
(0.193686, 0.0861497)
(0.199721, 0.0842421)
(0.205944, 0.0837043)
(0.212361, 0.0827239)
(0.218978, 0.0813695)
(0.225801, 0.0794835)
(0.232836, 0.0792952)
(0.240091, 0.076647)
(0.247572, 0.076593)
(0.255286, 0.075996)
(0.26324, 0.0759516)
(0.271442, 0.0746823)
(0.279899, 0.0716848)
(0.288621, 0.0724916)
(0.297613, 0.0710983)
(0.306886, 0.0689002)
(0.316448, 0.0691088)
(0.326308, 0.0673986)
(0.336476, 0.0648479)
(0.346959, 0.0649155)
(0.35777, 0.0629901)
(0.368917, 0.0620262)
(0.380412, 0.0616822)
(0.392265, 0.0598037)
(0.404487, 0.0585035)
(0.41709, 0.0570331)
(0.430086, 0.0559459)
(0.443487, 0.0550666)
(0.457305, 0.0541744)
(0.471554, 0.0530387)
(0.486247, 0.0513976)
(0.501397, 0.049556)
(0.51702, 0.0491812)
(0.533129, 0.0476285)
(0.54974, 0.0461455)
(0.566869, 0.0452652)
(0.584532, 0.0442428)
(0.602745, 0.0430399)
(0.621525, 0.0419631)
(0.64089, 0.040897)
(0.660859, 0.039655)
(0.68145, 0.0389979)
(0.702683, 0.0379979)
(0.724577, 0.0371928)
(0.747154, 0.0362024)
(0.770434, 0.0351326)
(0.794439, 0.0344047)
(0.819192, 0.0336947)
(0.844717, 0.0329263)
(0.871036, 0.0323415)
(0.898176, 0.0313363)
(0.926162, 0.030939)
(0.955019, 0.0298067)
(0.984776, 0.0290715)
(1.01546, 0.029014)
(1.0471, 0.0280439)
(1.07972, 0.0276567)
(1.11337, 0.0266793)
(1.14806, 0.0260966)
(1.18383, 0.0252817)
(1.22071, 0.0247643)
(1.25875, 0.0241456)
(1.29797, 0.0238266)
(1.33841, 0.023167)
(1.38011, 0.0226624)
(1.42312, 0.022045)
(1.46746, 0.0219448)
(1.51318, 0.0209111)
(1.56033, 0.0207655)
(1.60895, 0.020244)
(1.65908, 0.0197485)
(1.71077, 0.0194058)
(1.76408, 0.0188823)
(1.81904, 0.0188371)
(1.87572, 0.0177451)
(1.93416, 0.0173666)
(1.99443, 0.0172102)
(2.05657, 0.0168556)
(2.12065, 0.0163492)
(2.18672, 0.0159318)
(2.25486, 0.0158269)
(2.32512, 0.0153294)
(2.39756, 0.0150611)
(2.47227, 0.0146705)
(2.5493, 0.0142438)
(2.62873, 0.0138947)
(2.71063, 0.0136569)
(2.79509, 0.0133166)
(2.88218, 0.012818)
(2.97198, 0.0125874)
(3.06459, 0.0121515)
(3.16007, 0.0118638)
(3.25853, 0.0116359)
(3.36006, 0.0113974)
(3.46476, 0.0110509)
(3.57271, 0.0106812)
(3.68403, 0.010334)
(3.79882, 0.0102267)
(3.91718, 0.00988097)
(4.03923, 0.0096924)
(4.16509, 0.00952885)
(4.29487, 0.00900584)
(4.42869, 0.0088055)
(4.56668, 0.00860547)
(4.70896, 0.00816003)
(4.85569, 0.00812307)
(5.00698, 0.00793936)
(5.16299, 0.00769605)
(5.32386, 0.00736031)
(5.48974, 0.00714076)
(5.66079, 0.00718838)
(5.83717, 0.0069125)
(6.01904, 0.00648862)
(6.20658, 0.00620166)
(6.39997, 0.00616898)
(6.59938, 0.00570137)
(6.805, 0.00560866)
(7.01704, 0.00544476)
(7.23567, 0.00520234)
(7.46112, 0.00539721)
(7.6936, 0.00520981)
(7.93332, 0.00503779)
(8.1805, 0.0051112)
(8.43539, 0.00470648)
(8.69822, 0.00465961)
(8.96924, 0.00460132)
(9.24871, 0.00467529)
(9.53688, 0.0043633)
(9.83403, 0.00420151)
(10.1404, 0.00412751)
(10.4564, 0.00388104)
(10.7822, 0.00405836)
(11.1181, 0.00372564)
(11.4646, 0.00313455)
(11.8218, 0.00371573)
(12.1901, 0.00349907)
(12.5699, 0.00347407)
(12.9616, 0.00340046)
(13.3655, 0.00332854)
(13.7819, 0.00285323)
(14.2113, 0.00295549)
(14.6541, 0.00288265)
(15.1107, 0.00288494)
(15.5815, 0.00275594)
(16.067, 0.00260439)
(16.5676, 0.00274625)
(17.0839, 0.00261002)
(17.6162, 0.00225615)
(18.165, 0.00198282)
(18.731, 0.00250781)
(19.3147, 0.00213726)
(19.9165, 0.00217595)
(20.537, 0.00205179)
(21.1769, 0.00188702)
(21.8368, 0.00209255)
(22.5171, 0.00204271)
(23.2187, 0.00178546)
(23.9422, 0.00164715)
(24.6882, 0.00166626)
(25.4574, 0.00171433)
(26.2506, 0.00157313)
(27.0685, 0.00132192)
(27.9119, 0.00112533)
(28.7816, 0.00152979)
(29.6784, 0.00137904)
(30.6031, 0.00143726)
(31.5567, 0.00119694)
(32.5399, 0.00134545)
(33.5538, 0.00130816)
(34.5993, 0.00124832)
(35.6773, 0.00131154)
(36.789, 0.00124342)
(37.9352, 0.00103481)
(39.1172, 0.0013697)
(40.336, 0.000666503)
(41.5928, 0.00079769)
(42.8888, 0.000889731)
(44.2251, 0.000725292)
(45.6031, 0.00063023)
(47.024, 0.00127139)
(48.4892, 0.000810229)
(50, 0.00105084)
}

\newcommand{\genOdePlotExpCifaruNcsnppErrHeunBHi}{%
(0.02, 0.356071)
(0.0206232, 0.354246)
(0.0212657, 0.318677)
(0.0219283, 0.315108)
(0.0226116, 0.298649)
(0.0233161, 0.322673)
(0.0240426, 0.294104)
(0.0247917, 0.29943)
(0.0255642, 0.293844)
(0.0263607, 0.288881)
(0.0271821, 0.281294)
(0.028029, 0.283787)
(0.0289023, 0.262452)
(0.0298029, 0.266733)
(0.0307315, 0.262904)
(0.031689, 0.264355)
(0.0326764, 0.259455)
(0.0336945, 0.25895)
(0.0347444, 0.242975)
(0.035827, 0.242538)
(0.0369432, 0.234428)
(0.0380943, 0.237651)
(0.0392813, 0.227949)
(0.0405052, 0.237067)
(0.0417673, 0.221351)
(0.0430687, 0.213136)
(0.0444106, 0.214119)
(0.0457943, 0.203946)
(0.0472212, 0.209387)
(0.0486925, 0.198182)
(0.0502097, 0.196774)
(0.0517741, 0.191749)
(0.0533873, 0.19019)
(0.0550508, 0.191409)
(0.056766, 0.182706)
(0.0585348, 0.185257)
(0.0603586, 0.185503)
(0.0622393, 0.175271)
(0.0641785, 0.173212)
(0.0661782, 0.174553)
(0.0682402, 0.167641)
(0.0703664, 0.169671)
(0.0725589, 0.169608)
(0.0748197, 0.166166)
(0.0771509, 0.162731)
(0.0795548, 0.160106)
(0.0820336, 0.15832)
(0.0845896, 0.157158)
(0.0872253, 0.15182)
(0.089943, 0.150348)
(0.0927455, 0.145475)
(0.0956352, 0.149972)
(0.0986151, 0.148582)
(0.101688, 0.141653)
(0.104856, 0.143776)
(0.108123, 0.1403)
(0.111492, 0.140416)
(0.114966, 0.136045)
(0.118548, 0.131849)
(0.122242, 0.131697)
(0.126051, 0.132683)
(0.129978, 0.126174)
(0.134028, 0.12905)
(0.138204, 0.128119)
(0.14251, 0.124312)
(0.146951, 0.12502)
(0.151529, 0.120742)
(0.156251, 0.11807)
(0.161119, 0.117093)
(0.166139, 0.114789)
(0.171316, 0.115906)
(0.176654, 0.114892)
(0.182158, 0.110925)
(0.187834, 0.109364)
(0.193686, 0.107161)
(0.199721, 0.105213)
(0.205944, 0.105925)
(0.212361, 0.102369)
(0.218978, 0.10035)
(0.225801, 0.0988601)
(0.232836, 0.0971956)
(0.240091, 0.0937755)
(0.247572, 0.0942882)
(0.255286, 0.0919927)
(0.26324, 0.090162)
(0.271442, 0.0884401)
(0.279899, 0.0873217)
(0.288621, 0.0854756)
(0.297613, 0.0831402)
(0.306886, 0.0809169)
(0.316448, 0.0795389)
(0.326308, 0.0773758)
(0.336476, 0.075463)
(0.346959, 0.0742999)
(0.35777, 0.0721667)
(0.368917, 0.0704062)
(0.380412, 0.0686493)
(0.392265, 0.067122)
(0.404487, 0.0648387)
(0.41709, 0.063472)
(0.430086, 0.0619041)
(0.443487, 0.0606036)
(0.457305, 0.0590446)
(0.471554, 0.0579946)
(0.486247, 0.0565403)
(0.501397, 0.0546331)
(0.51702, 0.0534867)
(0.533129, 0.0517364)
(0.54974, 0.0504489)
(0.566869, 0.0495028)
(0.584532, 0.0483667)
(0.602745, 0.0470807)
(0.621525, 0.045932)
(0.64089, 0.0447908)
(0.660859, 0.0437211)
(0.68145, 0.0427226)
(0.702683, 0.0419902)
(0.724577, 0.0411654)
(0.747154, 0.0400906)
(0.770434, 0.0390691)
(0.794439, 0.0383576)
(0.819192, 0.0373882)
(0.844717, 0.0367053)
(0.871036, 0.0359003)
(0.898176, 0.0349287)
(0.926162, 0.034355)
(0.955019, 0.0334403)
(0.984776, 0.0328187)
(1.01546, 0.0320936)
(1.0471, 0.0312507)
(1.07972, 0.0307711)
(1.11337, 0.0301499)
(1.14806, 0.0293794)
(1.18383, 0.0286105)
(1.22071, 0.0277063)
(1.25875, 0.0272701)
(1.29797, 0.026728)
(1.33841, 0.026197)
(1.38011, 0.025502)
(1.42312, 0.0248305)
(1.46746, 0.0246198)
(1.51318, 0.0237332)
(1.56033, 0.0231612)
(1.60895, 0.0225131)
(1.65908, 0.0220584)
(1.71077, 0.0215524)
(1.76408, 0.0211309)
(1.81904, 0.0206083)
(1.87572, 0.0197124)
(1.93416, 0.019371)
(1.99443, 0.019059)
(2.05657, 0.0187522)
(2.12065, 0.0182583)
(2.18672, 0.0177612)
(2.25486, 0.0174124)
(2.32512, 0.0169161)
(2.39756, 0.0164905)
(2.47227, 0.0160021)
(2.5493, 0.0154847)
(2.62873, 0.0150589)
(2.71063, 0.0147637)
(2.79509, 0.0144275)
(2.88218, 0.0139647)
(2.97198, 0.0136393)
(3.06459, 0.0130968)
(3.16007, 0.0128156)
(3.25853, 0.0123405)
(3.36006, 0.0123002)
(3.46476, 0.0120174)
(3.57271, 0.0114785)
(3.68403, 0.0111483)
(3.79882, 0.0109931)
(3.91718, 0.0107972)
(4.03923, 0.0105101)
(4.16509, 0.0103751)
(4.29487, 0.0098322)
(4.42869, 0.00960254)
(4.56668, 0.00932915)
(4.70896, 0.00893263)
(4.85569, 0.00883403)
(5.00698, 0.00873287)
(5.16299, 0.00851484)
(5.32386, 0.0081182)
(5.48974, 0.00793265)
(5.66079, 0.00786037)
(5.83717, 0.00765154)
(6.01904, 0.00709445)
(6.20658, 0.00680359)
(6.39997, 0.00699052)
(6.59938, 0.00663671)
(6.805, 0.00643017)
(7.01704, 0.00614078)
(7.23567, 0.00658688)
(7.46112, 0.00593467)
(7.6936, 0.00586724)
(7.93332, 0.00564167)
(8.1805, 0.00565482)
(8.43539, 0.00520478)
(8.69822, 0.00515671)
(8.96924, 0.00506266)
(9.24871, 0.00491589)
(9.53688, 0.00461742)
(9.83403, 0.00463973)
(10.1404, 0.00433269)
(10.4564, 0.00425455)
(10.7822, 0.00424802)
(11.1181, 0.00391679)
(11.4646, 0.00371244)
(11.8218, 0.0039189)
(12.1901, 0.0036921)
(12.5699, 0.003725)
(12.9616, 0.00355442)
(13.3655, 0.00360264)
(13.7819, 0.00300853)
(14.2113, 0.00312373)
(14.6541, 0.00310153)
(15.1107, 0.00321475)
(15.5815, 0.00292546)
(16.067, 0.0027231)
(16.5676, 0.00289568)
(17.0839, 0.00283534)
(17.6162, 0.00269154)
(18.165, 0.00228634)
(18.731, 0.00270601)
(19.3147, 0.0023471)
(19.9165, 0.00244458)
(20.537, 0.00275069)
(21.1769, 0.00252959)
(21.8368, 0.00231565)
(22.5171, 0.00225768)
(23.2187, 0.00211603)
(23.9422, 0.00214499)
(24.6882, 0.00217797)
(25.4574, 0.00179068)
(26.2506, 0.00188911)
(27.0685, 0.00196693)
(27.9119, 0.0015944)
(28.7816, 0.00199132)
(29.6784, 0.00163041)
(30.6031, 0.00162693)
(31.5567, 0.00147118)
(32.5399, 0.00174722)
(33.5538, 0.00140039)
(34.5993, 0.00142974)
(35.6773, 0.00160326)
(36.789, 0.00133009)
(37.9352, 0.00112568)
(39.1172, 0.00156371)
(40.336, 0.000885345)
(41.5928, 0.00108648)
(42.8888, 0.00132603)
(44.2251, 0.000967134)
(45.6031, 0.000797654)
(47.024, 0.00138691)
(48.4892, 0.000903864)
(50, 0.00117004)
}

\newcommand{\genOdePlotExpCifaruNcsnppErrHeunC}{%
(0.02, 0.0313467)
(0.0206232, 0.0309874)
(0.0212657, 0.0307712)
(0.0219283, 0.0311797)
(0.0226116, 0.031203)
(0.0233161, 0.0310091)
(0.0240426, 0.0310308)
(0.0247917, 0.0311728)
(0.0255642, 0.0308195)
(0.0263607, 0.0301758)
(0.0271821, 0.0306385)
(0.028029, 0.0300633)
(0.0289023, 0.0300984)
(0.0298029, 0.0309949)
(0.0307315, 0.0305396)
(0.031689, 0.0294928)
(0.0326764, 0.029842)
(0.0336945, 0.0302257)
(0.0347444, 0.0296233)
(0.035827, 0.0297303)
(0.0369432, 0.029483)
(0.0380943, 0.0300522)
(0.0392813, 0.0298375)
(0.0405052, 0.029366)
(0.0417673, 0.0298706)
(0.0430687, 0.0292485)
(0.0444106, 0.0295877)
(0.0457943, 0.0289057)
(0.0472212, 0.0293476)
(0.0486925, 0.0290439)
(0.0502097, 0.0287534)
(0.0517741, 0.0289439)
(0.0533873, 0.028543)
(0.0550508, 0.0286486)
(0.056766, 0.0284302)
(0.0585348, 0.0283532)
(0.0603586, 0.0281427)
(0.0622393, 0.0280129)
(0.0641785, 0.0276804)
(0.0661782, 0.0277626)
(0.0682402, 0.027375)
(0.0703664, 0.0273898)
(0.0725589, 0.0272932)
(0.0748197, 0.0270346)
(0.0771509, 0.0272442)
(0.0795548, 0.0266281)
(0.0820336, 0.0263522)
(0.0845896, 0.0263147)
(0.0872253, 0.0259406)
(0.089943, 0.025878)
(0.0927455, 0.0254498)
(0.0956352, 0.0258239)
(0.0986151, 0.0251312)
(0.101688, 0.0248296)
(0.104856, 0.0246587)
(0.108123, 0.0244472)
(0.111492, 0.0243632)
(0.114966, 0.0237381)
(0.118548, 0.0236628)
(0.122242, 0.0232196)
(0.126051, 0.0231439)
(0.129978, 0.0229742)
(0.134028, 0.0226249)
(0.138204, 0.0219314)
(0.14251, 0.0219295)
(0.146951, 0.0218643)
(0.151529, 0.0213167)
(0.156251, 0.0210794)
(0.161119, 0.0209528)
(0.166139, 0.0204815)
(0.171316, 0.020177)
(0.176654, 0.0201111)
(0.182158, 0.0197781)
(0.187834, 0.0193498)
(0.193686, 0.0193203)
(0.199721, 0.0189372)
(0.205944, 0.0187549)
(0.212361, 0.0184541)
(0.218978, 0.0181206)
(0.225801, 0.0178612)
(0.232836, 0.0176928)
(0.240091, 0.0173443)
(0.247572, 0.0171979)
(0.255286, 0.0169375)
(0.26324, 0.0166562)
(0.271442, 0.0165001)
(0.279899, 0.0163005)
(0.288621, 0.0159993)
(0.297613, 0.015839)
(0.306886, 0.0155373)
(0.316448, 0.0154326)
(0.326308, 0.0151987)
(0.336476, 0.0149743)
(0.346959, 0.0147551)
(0.35777, 0.0146068)
(0.368917, 0.0143829)
(0.380412, 0.014312)
(0.392265, 0.014086)
(0.404487, 0.0138934)
(0.41709, 0.0138509)
(0.430086, 0.0135338)
(0.443487, 0.0133086)
(0.457305, 0.0133443)
(0.471554, 0.0130509)
(0.486247, 0.0128633)
(0.501397, 0.0126078)
(0.51702, 0.0126044)
(0.533129, 0.0123826)
(0.54974, 0.0121795)
(0.566869, 0.011999)
(0.584532, 0.0119643)
(0.602745, 0.0117875)
(0.621525, 0.0116904)
(0.64089, 0.0115609)
(0.660859, 0.0114164)
(0.68145, 0.0113175)
(0.702683, 0.0111563)
(0.724577, 0.0109899)
(0.747154, 0.0108796)
(0.770434, 0.010771)
(0.794439, 0.0106859)
(0.819192, 0.0105879)
(0.844717, 0.0103054)
(0.871036, 0.0104035)
(0.898176, 0.0102102)
(0.926162, 0.0100265)
(0.955019, 0.00988665)
(0.984776, 0.00971424)
(1.01546, 0.00968028)
(1.0471, 0.00951838)
(1.07972, 0.00948823)
(1.11337, 0.00926967)
(1.14806, 0.00933215)
(1.18383, 0.00913143)
(1.22071, 0.0089356)
(1.25875, 0.00894651)
(1.29797, 0.00899732)
(1.33841, 0.00868133)
(1.38011, 0.00852689)
(1.42312, 0.00830106)
(1.46746, 0.00843428)
(1.51318, 0.00817969)
(1.56033, 0.0082057)
(1.60895, 0.00802347)
(1.65908, 0.00797978)
(1.71077, 0.0078129)
(1.76408, 0.00783122)
(1.81904, 0.00785795)
(1.87572, 0.00745039)
(1.93416, 0.00745572)
(1.99443, 0.00742335)
(2.05657, 0.00738055)
(2.12065, 0.00723307)
(2.18672, 0.00712118)
(2.25486, 0.00720253)
(2.32512, 0.00707215)
(2.39756, 0.00699487)
(2.47227, 0.00689698)
(2.5493, 0.00667972)
(2.62873, 0.00656496)
(2.71063, 0.00659442)
(2.79509, 0.00656094)
(2.88218, 0.00644867)
(2.97198, 0.00637774)
(3.06459, 0.00604387)
(3.16007, 0.0061465)
(3.25853, 0.00604874)
(3.36006, 0.00605539)
(3.46476, 0.00592275)
(3.57271, 0.00575968)
(3.68403, 0.00581591)
(3.79882, 0.00566253)
(3.91718, 0.00556384)
(4.03923, 0.00565483)
(4.16509, 0.00558417)
(4.29487, 0.00540595)
(4.42869, 0.00536438)
(4.56668, 0.00533231)
(4.70896, 0.00502339)
(4.85569, 0.00504792)
(5.00698, 0.00520851)
(5.16299, 0.00497977)
(5.32386, 0.00474352)
(5.48974, 0.00478477)
(5.66079, 0.00495621)
(5.83717, 0.00474872)
(6.01904, 0.00457842)
(6.20658, 0.00430861)
(6.39997, 0.00451329)
(6.59938, 0.00442393)
(6.805, 0.00417607)
(7.01704, 0.00413177)
(7.23567, 0.00441784)
(7.46112, 0.00426762)
(7.6936, 0.00412099)
(7.93332, 0.00413604)
(8.1805, 0.00420964)
(8.43539, 0.00382704)
(8.69822, 0.00387643)
(8.96924, 0.00392021)
(9.24871, 0.00391819)
(9.53688, 0.00376569)
(9.83403, 0.00376387)
(10.1404, 0.00371031)
(10.4564, 0.0035756)
(10.7822, 0.00365703)
(11.1181, 0.00339945)
(11.4646, 0.00310197)
(11.8218, 0.0035121)
(12.1901, 0.00338749)
(12.5699, 0.00341261)
(12.9616, 0.00332853)
(13.3655, 0.00340922)
(13.7819, 0.00289767)
(14.2113, 0.00304697)
(14.6541, 0.00303969)
(15.1107, 0.00313379)
(15.5815, 0.00298836)
(16.067, 0.00285217)
(16.5676, 0.00306522)
(17.0839, 0.00305606)
(17.6162, 0.00270918)
(18.165, 0.00236295)
(18.731, 0.00305279)
(19.3147, 0.0027103)
(19.9165, 0.00263192)
(20.537, 0.0028519)
(21.1769, 0.0027889)
(21.8368, 0.00264622)
(22.5171, 0.00261067)
(23.2187, 0.00243942)
(23.9422, 0.00234558)
(24.6882, 0.00256094)
(25.4574, 0.00225914)
(26.2506, 0.00208678)
(27.0685, 0.00260313)
(27.9119, 0.00170482)
(28.7816, 0.00268753)
(29.6784, 0.00225493)
(30.6031, 0.00230658)
(31.5567, 0.00234995)
(32.5399, 0.00222423)
(33.5538, 0.00168073)
(34.5993, 0.00224129)
(35.6773, 0.00204268)
(36.789, 0.00212194)
(37.9352, 0.00192551)
(39.1172, 0.00199812)
(40.336, 0.00170506)
(41.5928, 0.00198313)
(42.8888, 0.00158704)
(44.2251, 0.00196351)
(45.6031, 0.00144227)
(47.024, 0.00238609)
(48.4892, 0.00147484)
(50, 0.0016092)
}

\newcommand{\genOdePlotExpCifaruNcsnppErrHeunCLo}{%
(0.02, 0.026682)
(0.0206232, 0.0262145)
(0.0212657, 0.0260217)
(0.0219283, 0.0262528)
(0.0226116, 0.0266523)
(0.0233161, 0.0265083)
(0.0240426, 0.0265283)
(0.0247917, 0.0265916)
(0.0255642, 0.0259345)
(0.0263607, 0.0257083)
(0.0271821, 0.0259168)
(0.028029, 0.0256298)
(0.0289023, 0.0261404)
(0.0298029, 0.0265119)
(0.0307315, 0.0262937)
(0.031689, 0.0249829)
(0.0326764, 0.0254689)
(0.0336945, 0.0258962)
(0.0347444, 0.0253552)
(0.035827, 0.0254478)
(0.0369432, 0.025462)
(0.0380943, 0.0259836)
(0.0392813, 0.0260267)
(0.0405052, 0.0253825)
(0.0417673, 0.0258444)
(0.0430687, 0.0255133)
(0.0444106, 0.025569)
(0.0457943, 0.0254628)
(0.0472212, 0.0258796)
(0.0486925, 0.0253464)
(0.0502097, 0.0253261)
(0.0517741, 0.0257621)
(0.0533873, 0.025223)
(0.0550508, 0.0251282)
(0.056766, 0.024939)
(0.0585348, 0.024713)
(0.0603586, 0.0250833)
(0.0622393, 0.0247745)
(0.0641785, 0.0245874)
(0.0661782, 0.0249724)
(0.0682402, 0.0243511)
(0.0703664, 0.0246091)
(0.0725589, 0.0246958)
(0.0748197, 0.0244379)
(0.0771509, 0.0247434)
(0.0795548, 0.0240626)
(0.0820336, 0.0239872)
(0.0845896, 0.0235053)
(0.0872253, 0.0235061)
(0.089943, 0.0237249)
(0.0927455, 0.0230688)
(0.0956352, 0.0239079)
(0.0986151, 0.0230136)
(0.101688, 0.0227783)
(0.104856, 0.0227466)
(0.108123, 0.0225136)
(0.111492, 0.0225839)
(0.114966, 0.0218273)
(0.118548, 0.0221231)
(0.122242, 0.0216531)
(0.126051, 0.0216028)
(0.129978, 0.0214444)
(0.134028, 0.0210026)
(0.138204, 0.0204002)
(0.14251, 0.020422)
(0.146951, 0.0206287)
(0.151529, 0.0199652)
(0.156251, 0.0196951)
(0.161119, 0.0197782)
(0.166139, 0.0192715)
(0.171316, 0.0190312)
(0.176654, 0.0190579)
(0.182158, 0.0187068)
(0.187834, 0.0182668)
(0.193686, 0.0183485)
(0.199721, 0.0179389)
(0.205944, 0.0178427)
(0.212361, 0.0174561)
(0.218978, 0.0172148)
(0.225801, 0.0170662)
(0.232836, 0.0168939)
(0.240091, 0.0165872)
(0.247572, 0.0164943)
(0.255286, 0.016192)
(0.26324, 0.015917)
(0.271442, 0.0158414)
(0.279899, 0.0156288)
(0.288621, 0.0153284)
(0.297613, 0.0152428)
(0.306886, 0.0148775)
(0.316448, 0.0147623)
(0.326308, 0.0145092)
(0.336476, 0.0143404)
(0.346959, 0.0141164)
(0.35777, 0.0139285)
(0.368917, 0.0137806)
(0.380412, 0.0137361)
(0.392265, 0.0134384)
(0.404487, 0.0132252)
(0.41709, 0.0132346)
(0.430086, 0.0129368)
(0.443487, 0.0126793)
(0.457305, 0.012739)
(0.471554, 0.0124777)
(0.486247, 0.0122135)
(0.501397, 0.0119435)
(0.51702, 0.0119385)
(0.533129, 0.0118054)
(0.54974, 0.0115374)
(0.566869, 0.0113539)
(0.584532, 0.0113378)
(0.602745, 0.0111684)
(0.621525, 0.0110724)
(0.64089, 0.0109644)
(0.660859, 0.0108107)
(0.68145, 0.0107246)
(0.702683, 0.0105408)
(0.724577, 0.0103966)
(0.747154, 0.0102784)
(0.770434, 0.0102486)
(0.794439, 0.0101008)
(0.819192, 0.00997462)
(0.844717, 0.00974676)
(0.871036, 0.00982863)
(0.898176, 0.00969838)
(0.926162, 0.00949102)
(0.955019, 0.00934495)
(0.984776, 0.00915574)
(1.01546, 0.00920902)
(1.0471, 0.00906416)
(1.07972, 0.00900209)
(1.11337, 0.00880562)
(1.14806, 0.00889493)
(1.18383, 0.00865934)
(1.22071, 0.00846401)
(1.25875, 0.00846402)
(1.29797, 0.00860295)
(1.33841, 0.00819885)
(1.38011, 0.0081285)
(1.42312, 0.00788232)
(1.46746, 0.00803933)
(1.51318, 0.0077963)
(1.56033, 0.00781793)
(1.60895, 0.00762251)
(1.65908, 0.00763891)
(1.71077, 0.00747038)
(1.76408, 0.00748335)
(1.81904, 0.00756803)
(1.87572, 0.00710656)
(1.93416, 0.00716689)
(1.99443, 0.00714059)
(2.05657, 0.00707361)
(2.12065, 0.00690732)
(2.18672, 0.00680467)
(2.25486, 0.00688387)
(2.32512, 0.00677262)
(2.39756, 0.00680117)
(2.47227, 0.00670287)
(2.5493, 0.00646259)
(2.62873, 0.00635751)
(2.71063, 0.00640976)
(2.79509, 0.00632807)
(2.88218, 0.00631954)
(2.97198, 0.00615728)
(3.06459, 0.00583202)
(3.16007, 0.00591143)
(3.25853, 0.00582294)
(3.36006, 0.00585116)
(3.46476, 0.00573849)
(3.57271, 0.00555177)
(3.68403, 0.005634)
(3.79882, 0.00543643)
(3.91718, 0.00540292)
(4.03923, 0.00547835)
(4.16509, 0.00541328)
(4.29487, 0.00519271)
(4.42869, 0.00512617)
(4.56668, 0.00515214)
(4.70896, 0.00483797)
(4.85569, 0.00485138)
(5.00698, 0.0049934)
(5.16299, 0.0047417)
(5.32386, 0.00451951)
(5.48974, 0.00454894)
(5.66079, 0.00466792)
(5.83717, 0.00451411)
(6.01904, 0.0043308)
(6.20658, 0.00411921)
(6.39997, 0.00423564)
(6.59938, 0.00414483)
(6.805, 0.00377541)
(7.01704, 0.00388943)
(7.23567, 0.00393552)
(7.46112, 0.0040839)
(7.6936, 0.00391784)
(7.93332, 0.00394051)
(8.1805, 0.00400054)
(8.43539, 0.00364661)
(8.69822, 0.00372532)
(8.96924, 0.00363783)
(9.24871, 0.00380777)
(9.53688, 0.00366586)
(9.83403, 0.00356675)
(10.1404, 0.0036032)
(10.4564, 0.00343983)
(10.7822, 0.00357632)
(11.1181, 0.00331026)
(11.4646, 0.00288349)
(11.8218, 0.00341779)
(12.1901, 0.00329274)
(12.5699, 0.00329675)
(12.9616, 0.0032443)
(13.3655, 0.00326921)
(13.7819, 0.00280503)
(14.2113, 0.0029545)
(14.6541, 0.0029192)
(15.1107, 0.00294162)
(15.5815, 0.00288953)
(16.067, 0.00278554)
(16.5676, 0.00297863)
(17.0839, 0.0029458)
(17.6162, 0.00250193)
(18.165, 0.00214651)
(18.731, 0.00295368)
(19.3147, 0.00258577)
(19.9165, 0.00252756)
(20.537, 0.00251193)
(21.1769, 0.00247204)
(21.8368, 0.00248592)
(22.5171, 0.00247981)
(23.2187, 0.00227171)
(23.9422, 0.00205882)
(24.6882, 0.00231109)
(25.4574, 0.00221779)
(26.2506, 0.00189367)
(27.0685, 0.00224583)
(27.9119, 0.00144366)
(28.7816, 0.00248501)
(29.6784, 0.00214526)
(30.6031, 0.00213428)
(31.5567, 0.00215798)
(32.5399, 0.00200574)
(33.5538, 0.00164554)
(34.5993, 0.00214253)
(35.6773, 0.00191168)
(36.789, 0.00200133)
(37.9352, 0.00182586)
(39.1172, 0.00191519)
(40.336, 0.00154688)
(41.5928, 0.00171434)
(42.8888, 0.00132247)
(44.2251, 0.00181806)
(45.6031, 0.00137724)
(47.024, 0.00229056)
(48.4892, 0.00138179)
(50, 0.00149434)
}

\newcommand{\genOdePlotExpCifaruNcsnppErrHeunCHi}{%
(0.02, 0.0360113)
(0.0206232, 0.0357603)
(0.0212657, 0.0355206)
(0.0219283, 0.0361065)
(0.0226116, 0.0357537)
(0.0233161, 0.0355099)
(0.0240426, 0.0355333)
(0.0247917, 0.0357539)
(0.0255642, 0.0357045)
(0.0263607, 0.0346434)
(0.0271821, 0.0353602)
(0.028029, 0.0344968)
(0.0289023, 0.0340564)
(0.0298029, 0.0354778)
(0.0307315, 0.0347855)
(0.031689, 0.0340028)
(0.0326764, 0.0342151)
(0.0336945, 0.0345553)
(0.0347444, 0.0338914)
(0.035827, 0.0340128)
(0.0369432, 0.0335039)
(0.0380943, 0.0341207)
(0.0392813, 0.0336483)
(0.0405052, 0.0333496)
(0.0417673, 0.0338967)
(0.0430687, 0.0329837)
(0.0444106, 0.0336064)
(0.0457943, 0.0323487)
(0.0472212, 0.0328156)
(0.0486925, 0.0327414)
(0.0502097, 0.0321806)
(0.0517741, 0.0321258)
(0.0533873, 0.0318631)
(0.0550508, 0.0321689)
(0.056766, 0.0319215)
(0.0585348, 0.0319935)
(0.0603586, 0.0312022)
(0.0622393, 0.0312514)
(0.0641785, 0.0307735)
(0.0661782, 0.0305528)
(0.0682402, 0.030399)
(0.0703664, 0.0301705)
(0.0725589, 0.0298905)
(0.0748197, 0.0296312)
(0.0771509, 0.029745)
(0.0795548, 0.0291935)
(0.0820336, 0.0287173)
(0.0845896, 0.0291242)
(0.0872253, 0.0283751)
(0.089943, 0.028031)
(0.0927455, 0.0278308)
(0.0956352, 0.02774)
(0.0986151, 0.0272489)
(0.101688, 0.0268809)
(0.104856, 0.0265707)
(0.108123, 0.0263808)
(0.111492, 0.0261424)
(0.114966, 0.0256489)
(0.118548, 0.0252026)
(0.122242, 0.024786)
(0.126051, 0.0246849)
(0.129978, 0.024504)
(0.134028, 0.0242472)
(0.138204, 0.0234626)
(0.14251, 0.0234369)
(0.146951, 0.0230998)
(0.151529, 0.0226682)
(0.156251, 0.0224637)
(0.161119, 0.0221273)
(0.166139, 0.0216916)
(0.171316, 0.0213228)
(0.176654, 0.0211642)
(0.182158, 0.0208493)
(0.187834, 0.0204328)
(0.193686, 0.0202922)
(0.199721, 0.0199354)
(0.205944, 0.0196672)
(0.212361, 0.0194522)
(0.218978, 0.0190264)
(0.225801, 0.0186562)
(0.232836, 0.0184916)
(0.240091, 0.0181013)
(0.247572, 0.0179016)
(0.255286, 0.017683)
(0.26324, 0.0173954)
(0.271442, 0.0171588)
(0.279899, 0.0169722)
(0.288621, 0.0166703)
(0.297613, 0.0164352)
(0.306886, 0.016197)
(0.316448, 0.016103)
(0.326308, 0.0158883)
(0.336476, 0.0156083)
(0.346959, 0.0153938)
(0.35777, 0.0152851)
(0.368917, 0.0149853)
(0.380412, 0.0148878)
(0.392265, 0.0147336)
(0.404487, 0.0145616)
(0.41709, 0.0144672)
(0.430086, 0.0141309)
(0.443487, 0.013938)
(0.457305, 0.0139497)
(0.471554, 0.0136241)
(0.486247, 0.0135132)
(0.501397, 0.013272)
(0.51702, 0.0132703)
(0.533129, 0.0129598)
(0.54974, 0.0128217)
(0.566869, 0.0126442)
(0.584532, 0.0125908)
(0.602745, 0.0124066)
(0.621525, 0.0123084)
(0.64089, 0.0121574)
(0.660859, 0.0120222)
(0.68145, 0.0119103)
(0.702683, 0.0117718)
(0.724577, 0.0115831)
(0.747154, 0.0114808)
(0.770434, 0.0112935)
(0.794439, 0.0112709)
(0.819192, 0.0112011)
(0.844717, 0.0108639)
(0.871036, 0.0109784)
(0.898176, 0.0107221)
(0.926162, 0.0105621)
(0.955019, 0.0104283)
(0.984776, 0.0102728)
(1.01546, 0.0101515)
(1.0471, 0.00997261)
(1.07972, 0.00997437)
(1.11337, 0.00973372)
(1.14806, 0.00976937)
(1.18383, 0.00960353)
(1.22071, 0.0094072)
(1.25875, 0.009429)
(1.29797, 0.0093917)
(1.33841, 0.00916381)
(1.38011, 0.00892528)
(1.42312, 0.0087198)
(1.46746, 0.00882923)
(1.51318, 0.00856308)
(1.56033, 0.00859347)
(1.60895, 0.00842442)
(1.65908, 0.00832065)
(1.71077, 0.00815542)
(1.76408, 0.00817908)
(1.81904, 0.00814787)
(1.87572, 0.00779421)
(1.93416, 0.00774456)
(1.99443, 0.0077061)
(2.05657, 0.00768749)
(2.12065, 0.00755882)
(2.18672, 0.00743769)
(2.25486, 0.0075212)
(2.32512, 0.00737167)
(2.39756, 0.00718857)
(2.47227, 0.00709108)
(2.5493, 0.00689685)
(2.62873, 0.00677242)
(2.71063, 0.00677909)
(2.79509, 0.0067938)
(2.88218, 0.00657781)
(2.97198, 0.0065982)
(3.06459, 0.00625573)
(3.16007, 0.00638156)
(3.25853, 0.00627453)
(3.36006, 0.00625963)
(3.46476, 0.00610702)
(3.57271, 0.0059676)
(3.68403, 0.00599782)
(3.79882, 0.00588862)
(3.91718, 0.00572477)
(4.03923, 0.00583132)
(4.16509, 0.00575506)
(4.29487, 0.00561918)
(4.42869, 0.00560259)
(4.56668, 0.00551248)
(4.70896, 0.0052088)
(4.85569, 0.00524445)
(5.00698, 0.00542362)
(5.16299, 0.00521784)
(5.32386, 0.00496753)
(5.48974, 0.00502059)
(5.66079, 0.0052445)
(5.83717, 0.00498333)
(6.01904, 0.00482603)
(6.20658, 0.00449801)
(6.39997, 0.00479093)
(6.59938, 0.00470302)
(6.805, 0.00457672)
(7.01704, 0.00437411)
(7.23567, 0.00490016)
(7.46112, 0.00445134)
(7.6936, 0.00432414)
(7.93332, 0.00433158)
(8.1805, 0.00441875)
(8.43539, 0.00400746)
(8.69822, 0.00402753)
(8.96924, 0.00420259)
(9.24871, 0.00402861)
(9.53688, 0.00386552)
(9.83403, 0.00396099)
(10.1404, 0.00381743)
(10.4564, 0.00371137)
(10.7822, 0.00373774)
(11.1181, 0.00348863)
(11.4646, 0.00332045)
(11.8218, 0.0036064)
(12.1901, 0.00348225)
(12.5699, 0.00352847)
(12.9616, 0.00341276)
(13.3655, 0.00354923)
(13.7819, 0.00299031)
(14.2113, 0.00313944)
(14.6541, 0.00316017)
(15.1107, 0.00332597)
(15.5815, 0.00308718)
(16.067, 0.00291879)
(16.5676, 0.0031518)
(17.0839, 0.00316632)
(17.6162, 0.00291643)
(18.165, 0.00257939)
(18.731, 0.0031519)
(19.3147, 0.00283482)
(19.9165, 0.00273629)
(20.537, 0.00319187)
(21.1769, 0.00310576)
(21.8368, 0.00280652)
(22.5171, 0.00274154)
(23.2187, 0.00260713)
(23.9422, 0.00263233)
(24.6882, 0.00281079)
(25.4574, 0.0023005)
(26.2506, 0.00227988)
(27.0685, 0.00296042)
(27.9119, 0.00196599)
(28.7816, 0.00289004)
(29.6784, 0.0023646)
(30.6031, 0.00247889)
(31.5567, 0.00254193)
(32.5399, 0.00244273)
(33.5538, 0.00171591)
(34.5993, 0.00234005)
(35.6773, 0.00217368)
(36.789, 0.00224255)
(37.9352, 0.00202515)
(39.1172, 0.00208105)
(40.336, 0.00186324)
(41.5928, 0.00225192)
(42.8888, 0.00185161)
(44.2251, 0.00210895)
(45.6031, 0.00150731)
(47.024, 0.00248161)
(48.4892, 0.00156789)
(50, 0.00172406)
}

\newcommand{\genOdePlotExpCifaruNcsnppErrHeunD}{%
(0.02, 0.00285489)
(0.0206232, 0.0028006)
(0.0212657, 0.00283452)
(0.0219283, 0.00285997)
(0.0226116, 0.00288741)
(0.0233161, 0.00288075)
(0.0240426, 0.00290684)
(0.0247917, 0.0029079)
(0.0255642, 0.00293233)
(0.0263607, 0.00290396)
(0.0271821, 0.00293129)
(0.028029, 0.00299698)
(0.0289023, 0.0029789)
(0.0298029, 0.00298078)
(0.0307315, 0.00300972)
(0.031689, 0.00300524)
(0.0326764, 0.00301898)
(0.0336945, 0.00303133)
(0.0347444, 0.00305088)
(0.035827, 0.00305001)
(0.0369432, 0.00306259)
(0.0380943, 0.00308778)
(0.0392813, 0.00310135)
(0.0405052, 0.00311082)
(0.0417673, 0.00308784)
(0.0430687, 0.00310371)
(0.0444106, 0.00315321)
(0.0457943, 0.00311411)
(0.0472212, 0.00311412)
(0.0486925, 0.00319379)
(0.0502097, 0.0031637)
(0.0517741, 0.00314551)
(0.0533873, 0.00313582)
(0.0550508, 0.00317929)
(0.056766, 0.00322055)
(0.0585348, 0.00323939)
(0.0603586, 0.00317337)
(0.0622393, 0.00324758)
(0.0641785, 0.00317557)
(0.0661782, 0.00321664)
(0.0682402, 0.00321453)
(0.0703664, 0.00324594)
(0.0725589, 0.00323071)
(0.0748197, 0.00325916)
(0.0771509, 0.00323688)
(0.0795548, 0.00322866)
(0.0820336, 0.00325629)
(0.0845896, 0.00322978)
(0.0872253, 0.00330952)
(0.089943, 0.00326683)
(0.0927455, 0.00328941)
(0.0956352, 0.00331469)
(0.0986151, 0.00330241)
(0.101688, 0.00329835)
(0.104856, 0.0032906)
(0.108123, 0.00326582)
(0.111492, 0.00334258)
(0.114966, 0.00328828)
(0.118548, 0.00326009)
(0.122242, 0.00328844)
(0.126051, 0.00332644)
(0.129978, 0.00332743)
(0.134028, 0.00335995)
(0.138204, 0.00331326)
(0.14251, 0.00338703)
(0.146951, 0.00334778)
(0.151529, 0.00334911)
(0.156251, 0.00334882)
(0.161119, 0.00332884)
(0.166139, 0.00328862)
(0.171316, 0.00337194)
(0.176654, 0.00338074)
(0.182158, 0.00339448)
(0.187834, 0.00338148)
(0.193686, 0.0033822)
(0.199721, 0.00335563)
(0.205944, 0.00337082)
(0.212361, 0.00342619)
(0.218978, 0.0033699)
(0.225801, 0.00337117)
(0.232836, 0.00339619)
(0.240091, 0.00337513)
(0.247572, 0.00337979)
(0.255286, 0.00342039)
(0.26324, 0.0034415)
(0.271442, 0.00341754)
(0.279899, 0.00342839)
(0.288621, 0.00341412)
(0.297613, 0.00345672)
(0.306886, 0.00345821)
(0.316448, 0.00347211)
(0.326308, 0.00347402)
(0.336476, 0.00344173)
(0.346959, 0.00346407)
(0.35777, 0.00347106)
(0.368917, 0.00345857)
(0.380412, 0.0035627)
(0.392265, 0.00350595)
(0.404487, 0.00354922)
(0.41709, 0.00358539)
(0.430086, 0.00354116)
(0.443487, 0.00349206)
(0.457305, 0.00357962)
(0.471554, 0.00357285)
(0.486247, 0.00356044)
(0.501397, 0.00355168)
(0.51702, 0.0036275)
(0.533129, 0.00353662)
(0.54974, 0.00349141)
(0.566869, 0.00357202)
(0.584532, 0.0035745)
(0.602745, 0.003626)
(0.621525, 0.00371902)
(0.64089, 0.00360369)
(0.660859, 0.00359181)
(0.68145, 0.00365088)
(0.702683, 0.0037109)
(0.724577, 0.00363281)
(0.747154, 0.00366145)
(0.770434, 0.00366862)
(0.794439, 0.00371988)
(0.819192, 0.00365602)
(0.844717, 0.00362703)
(0.871036, 0.0038035)
(0.898176, 0.00367569)
(0.926162, 0.00371432)
(0.955019, 0.0036809)
(0.984776, 0.0037174)
(1.01546, 0.00370408)
(1.0471, 0.00379785)
(1.07972, 0.00377338)
(1.11337, 0.00368146)
(1.14806, 0.00378358)
(1.18383, 0.00378061)
(1.22071, 0.00372801)
(1.25875, 0.0038171)
(1.29797, 0.00384813)
(1.33841, 0.00382978)
(1.38011, 0.00384584)
(1.42312, 0.00376332)
(1.46746, 0.00387705)
(1.51318, 0.00378639)
(1.56033, 0.003814)
(1.60895, 0.00383044)
(1.65908, 0.00381893)
(1.71077, 0.00380244)
(1.76408, 0.00388732)
(1.81904, 0.00391956)
(1.87572, 0.00376965)
(1.93416, 0.00386172)
(1.99443, 0.00389915)
(2.05657, 0.00393282)
(2.12065, 0.00390637)
(2.18672, 0.00384274)
(2.25486, 0.00395632)
(2.32512, 0.00403689)
(2.39756, 0.00404109)
(2.47227, 0.00397166)
(2.5493, 0.00391614)
(2.62873, 0.00386883)
(2.71063, 0.00393868)
(2.79509, 0.0040689)
(2.88218, 0.00400472)
(2.97198, 0.00406154)
(3.06459, 0.00390967)
(3.16007, 0.0039261)
(3.25853, 0.00394521)
(3.36006, 0.00395161)
(3.46476, 0.00398357)
(3.57271, 0.00405855)
(3.68403, 0.00401331)
(3.79882, 0.00402221)
(3.91718, 0.00410157)
(4.03923, 0.00414544)
(4.16509, 0.00407607)
(4.29487, 0.00391969)
(4.42869, 0.00408194)
(4.56668, 0.0040554)
(4.70896, 0.00393268)
(4.85569, 0.00393171)
(5.00698, 0.00425488)
(5.16299, 0.00411758)
(5.32386, 0.00408939)
(5.48974, 0.00403718)
(5.66079, 0.00423521)
(5.83717, 0.00409552)
(6.01904, 0.00402511)
(6.20658, 0.00389922)
(6.39997, 0.00406337)
(6.59938, 0.0041118)
(6.805, 0.00396408)
(7.01704, 0.00387358)
(7.23567, 0.00432688)
(7.46112, 0.00417969)
(7.6936, 0.00411542)
(7.93332, 0.00417296)
(8.1805, 0.00432456)
(8.43539, 0.00396594)
(8.69822, 0.00411427)
(8.96924, 0.00417852)
(9.24871, 0.00427142)
(9.53688, 0.00417615)
(9.83403, 0.00420973)
(10.1404, 0.0041266)
(10.4564, 0.00408695)
(10.7822, 0.0043129)
(11.1181, 0.00409346)
(11.4646, 0.00381729)
(11.8218, 0.00428677)
(12.1901, 0.00419511)
(12.5699, 0.00402787)
(12.9616, 0.00438291)
(13.3655, 0.0041639)
(13.7819, 0.00381534)
(14.2113, 0.0041838)
(14.6541, 0.00400512)
(15.1107, 0.00407822)
(15.5815, 0.00404631)
(16.067, 0.00401393)
(16.5676, 0.00448787)
(17.0839, 0.00442003)
(17.6162, 0.0038065)
(18.165, 0.00404006)
(18.731, 0.00453012)
(19.3147, 0.00384006)
(19.9165, 0.00427673)
(20.537, 0.00429935)
(21.1769, 0.0042759)
(21.8368, 0.00432693)
(22.5171, 0.00416728)
(23.2187, 0.00409533)
(23.9422, 0.00431116)
(24.6882, 0.00439427)
(25.4574, 0.0040085)
(26.2506, 0.00408365)
(27.0685, 0.00435618)
(27.9119, 0.00400255)
(28.7816, 0.00450186)
(29.6784, 0.00443854)
(30.6031, 0.00425787)
(31.5567, 0.00397631)
(32.5399, 0.00434311)
(33.5538, 0.00408202)
(34.5993, 0.0041539)
(35.6773, 0.00440232)
(36.789, 0.00393584)
(37.9352, 0.00369676)
(39.1172, 0.00436105)
(40.336, 0.00366859)
(41.5928, 0.00392411)
(42.8888, 0.00453337)
(44.2251, 0.00423799)
(45.6031, 0.0039499)
(47.024, 0.00434449)
(48.4892, 0.00426475)
(50, 0.0044729)
}

\newcommand{\genOdePlotExpCifaruNcsnppErrHeunDLo}{%
(0.02, 0.00246218)
(0.0206232, 0.00233796)
(0.0212657, 0.00244755)
(0.0219283, 0.00243918)
(0.0226116, 0.00254188)
(0.0233161, 0.00249311)
(0.0240426, 0.00251868)
(0.0247917, 0.00249941)
(0.0255642, 0.00256492)
(0.0263607, 0.0025062)
(0.0271821, 0.00258098)
(0.028029, 0.00265632)
(0.0289023, 0.00260583)
(0.0298029, 0.00262687)
(0.0307315, 0.00267336)
(0.031689, 0.00267734)
(0.0326764, 0.00267403)
(0.0336945, 0.00270178)
(0.0347444, 0.00270705)
(0.035827, 0.00273771)
(0.0369432, 0.002719)
(0.0380943, 0.00278228)
(0.0392813, 0.00275448)
(0.0405052, 0.00283008)
(0.0417673, 0.00278818)
(0.0430687, 0.00277755)
(0.0444106, 0.00287184)
(0.0457943, 0.00281148)
(0.0472212, 0.00282114)
(0.0486925, 0.00294558)
(0.0502097, 0.00287982)
(0.0517741, 0.0028531)
(0.0533873, 0.00283754)
(0.0550508, 0.00290651)
(0.056766, 0.00297206)
(0.0585348, 0.00300893)
(0.0603586, 0.00292085)
(0.0622393, 0.0030115)
(0.0641785, 0.00291112)
(0.0661782, 0.00296896)
(0.0682402, 0.00295343)
(0.0703664, 0.0029762)
(0.0725589, 0.00298902)
(0.0748197, 0.00302004)
(0.0771509, 0.00300312)
(0.0795548, 0.00301973)
(0.0820336, 0.00302353)
(0.0845896, 0.00300131)
(0.0872253, 0.00308775)
(0.089943, 0.00305391)
(0.0927455, 0.00306771)
(0.0956352, 0.00311759)
(0.0986151, 0.00309533)
(0.101688, 0.00311209)
(0.104856, 0.00307911)
(0.108123, 0.00303648)
(0.111492, 0.00312779)
(0.114966, 0.00312187)
(0.118548, 0.00306219)
(0.122242, 0.00311094)
(0.126051, 0.00313211)
(0.129978, 0.00314968)
(0.134028, 0.00316431)
(0.138204, 0.00314097)
(0.14251, 0.00319895)
(0.146951, 0.00318342)
(0.151529, 0.00317451)
(0.156251, 0.00317429)
(0.161119, 0.00316378)
(0.166139, 0.00312685)
(0.171316, 0.00320524)
(0.176654, 0.00321)
(0.182158, 0.00323711)
(0.187834, 0.00321496)
(0.193686, 0.00322521)
(0.199721, 0.0032011)
(0.205944, 0.00320732)
(0.212361, 0.00327112)
(0.218978, 0.0032245)
(0.225801, 0.00320481)
(0.232836, 0.00323645)
(0.240091, 0.00321947)
(0.247572, 0.00320761)
(0.255286, 0.00325794)
(0.26324, 0.00328435)
(0.271442, 0.00326665)
(0.279899, 0.00326669)
(0.288621, 0.00325458)
(0.297613, 0.00331178)
(0.306886, 0.00329092)
(0.316448, 0.00333273)
(0.326308, 0.00330414)
(0.336476, 0.00328291)
(0.346959, 0.00328237)
(0.35777, 0.00330753)
(0.368917, 0.00329892)
(0.380412, 0.00338668)
(0.392265, 0.0033282)
(0.404487, 0.00336892)
(0.41709, 0.00340641)
(0.430086, 0.00336086)
(0.443487, 0.00332459)
(0.457305, 0.00341513)
(0.471554, 0.00338625)
(0.486247, 0.0033857)
(0.501397, 0.00335976)
(0.51702, 0.00344526)
(0.533129, 0.00336871)
(0.54974, 0.00330305)
(0.566869, 0.00337833)
(0.584532, 0.00340029)
(0.602745, 0.00344113)
(0.621525, 0.00353423)
(0.64089, 0.00342602)
(0.660859, 0.00338856)
(0.68145, 0.0034974)
(0.702683, 0.00352559)
(0.724577, 0.00344241)
(0.747154, 0.00346449)
(0.770434, 0.00347601)
(0.794439, 0.0035162)
(0.819192, 0.00344613)
(0.844717, 0.00343726)
(0.871036, 0.00365585)
(0.898176, 0.00351137)
(0.926162, 0.00351827)
(0.955019, 0.00347276)
(0.984776, 0.00353176)
(1.01546, 0.00351114)
(1.0471, 0.00361686)
(1.07972, 0.00360999)
(1.11337, 0.00350612)
(1.14806, 0.00360564)
(1.18383, 0.00363013)
(1.22071, 0.00357053)
(1.25875, 0.00362663)
(1.29797, 0.00367647)
(1.33841, 0.00365024)
(1.38011, 0.0036615)
(1.42312, 0.00360619)
(1.46746, 0.00372929)
(1.51318, 0.00361315)
(1.56033, 0.00364266)
(1.60895, 0.00365633)
(1.65908, 0.00368664)
(1.71077, 0.00364285)
(1.76408, 0.00373312)
(1.81904, 0.00376499)
(1.87572, 0.00362306)
(1.93416, 0.00372264)
(1.99443, 0.00376368)
(2.05657, 0.00378558)
(2.12065, 0.00379302)
(2.18672, 0.00368249)
(2.25486, 0.0038065)
(2.32512, 0.00390622)
(2.39756, 0.00390631)
(2.47227, 0.00385873)
(2.5493, 0.00380462)
(2.62873, 0.00375758)
(2.71063, 0.0038191)
(2.79509, 0.00395296)
(2.88218, 0.00387489)
(2.97198, 0.00395927)
(3.06459, 0.00374816)
(3.16007, 0.00382075)
(3.25853, 0.00384249)
(3.36006, 0.00380625)
(3.46476, 0.00385617)
(3.57271, 0.00393508)
(3.68403, 0.00386783)
(3.79882, 0.00388288)
(3.91718, 0.00397377)
(4.03923, 0.00399208)
(4.16509, 0.0039632)
(4.29487, 0.00378437)
(4.42869, 0.00392268)
(4.56668, 0.00389528)
(4.70896, 0.0037829)
(4.85569, 0.00375476)
(5.00698, 0.00407707)
(5.16299, 0.0039551)
(5.32386, 0.00389327)
(5.48974, 0.00386903)
(5.66079, 0.00399199)
(5.83717, 0.0038941)
(6.01904, 0.00382557)
(6.20658, 0.00370558)
(6.39997, 0.00382181)
(6.59938, 0.00386797)
(6.805, 0.00359429)
(7.01704, 0.00362548)
(7.23567, 0.00393218)
(7.46112, 0.00395448)
(7.6936, 0.00392994)
(7.93332, 0.0039728)
(8.1805, 0.00409247)
(8.43539, 0.00375143)
(8.69822, 0.00392462)
(8.96924, 0.00387298)
(9.24871, 0.00416413)
(9.53688, 0.0040631)
(9.83403, 0.00401172)
(10.1404, 0.00402104)
(10.4564, 0.00390589)
(10.7822, 0.00421924)
(11.1181, 0.00398986)
(11.4646, 0.00353521)
(11.8218, 0.00413494)
(12.1901, 0.00408141)
(12.5699, 0.00387616)
(12.9616, 0.00419621)
(13.3655, 0.00401741)
(13.7819, 0.00372234)
(14.2113, 0.00408432)
(14.6541, 0.00375509)
(15.1107, 0.00383681)
(15.5815, 0.00386888)
(16.067, 0.00392572)
(16.5676, 0.00428182)
(17.0839, 0.0043441)
(17.6162, 0.00367639)
(18.165, 0.00373308)
(18.731, 0.00436961)
(19.3147, 0.00356766)
(19.9165, 0.0041131)
(20.537, 0.00392297)
(21.1769, 0.00391206)
(21.8368, 0.00402452)
(22.5171, 0.00401972)
(23.2187, 0.00396512)
(23.9422, 0.00394875)
(24.6882, 0.00411546)
(25.4574, 0.00390821)
(26.2506, 0.00386275)
(27.0685, 0.00406813)
(27.9119, 0.0039218)
(28.7816, 0.00436153)
(29.6784, 0.00431058)
(30.6031, 0.00410867)
(31.5567, 0.00386958)
(32.5399, 0.00403784)
(33.5538, 0.00397684)
(34.5993, 0.0040751)
(35.6773, 0.00428144)
(36.789, 0.00371616)
(37.9352, 0.00354123)
(39.1172, 0.00423044)
(40.336, 0.00343911)
(41.5928, 0.00375988)
(42.8888, 0.00425778)
(44.2251, 0.00389367)
(45.6031, 0.00380178)
(47.024, 0.00407106)
(48.4892, 0.00412402)
(50, 0.00425635)
}

\newcommand{\genOdePlotExpCifaruNcsnppErrHeunDHi}{%
(0.02, 0.00324761)
(0.0206232, 0.00326323)
(0.0212657, 0.00322149)
(0.0219283, 0.00328075)
(0.0226116, 0.00323293)
(0.0233161, 0.0032684)
(0.0240426, 0.003295)
(0.0247917, 0.00331639)
(0.0255642, 0.00329973)
(0.0263607, 0.00330173)
(0.0271821, 0.00328161)
(0.028029, 0.00333764)
(0.0289023, 0.00335198)
(0.0298029, 0.00333469)
(0.0307315, 0.00334607)
(0.031689, 0.00333314)
(0.0326764, 0.00336392)
(0.0336945, 0.00336087)
(0.0347444, 0.00339472)
(0.035827, 0.00336231)
(0.0369432, 0.00340619)
(0.0380943, 0.00339328)
(0.0392813, 0.00344822)
(0.0405052, 0.00339157)
(0.0417673, 0.0033875)
(0.0430687, 0.00342987)
(0.0444106, 0.00343459)
(0.0457943, 0.00341675)
(0.0472212, 0.00340709)
(0.0486925, 0.00344199)
(0.0502097, 0.00344758)
(0.0517741, 0.00343791)
(0.0533873, 0.0034341)
(0.0550508, 0.00345207)
(0.056766, 0.00346904)
(0.0585348, 0.00346986)
(0.0603586, 0.00342589)
(0.0622393, 0.00348367)
(0.0641785, 0.00344002)
(0.0661782, 0.00346431)
(0.0682402, 0.00347563)
(0.0703664, 0.00351568)
(0.0725589, 0.0034724)
(0.0748197, 0.00349828)
(0.0771509, 0.00347065)
(0.0795548, 0.0034376)
(0.0820336, 0.00348906)
(0.0845896, 0.00345825)
(0.0872253, 0.0035313)
(0.089943, 0.00347975)
(0.0927455, 0.00351111)
(0.0956352, 0.00351178)
(0.0986151, 0.00350948)
(0.101688, 0.00348461)
(0.104856, 0.00350209)
(0.108123, 0.00349517)
(0.111492, 0.00355737)
(0.114966, 0.0034547)
(0.118548, 0.00345799)
(0.122242, 0.00346593)
(0.126051, 0.00352076)
(0.129978, 0.00350517)
(0.134028, 0.0035556)
(0.138204, 0.00348556)
(0.14251, 0.00357511)
(0.146951, 0.00351213)
(0.151529, 0.00352371)
(0.156251, 0.00352335)
(0.161119, 0.0034939)
(0.166139, 0.00345039)
(0.171316, 0.00353865)
(0.176654, 0.00355147)
(0.182158, 0.00355185)
(0.187834, 0.00354801)
(0.193686, 0.0035392)
(0.199721, 0.00351015)
(0.205944, 0.00353432)
(0.212361, 0.00358126)
(0.218978, 0.0035153)
(0.225801, 0.00353753)
(0.232836, 0.00355593)
(0.240091, 0.00353079)
(0.247572, 0.00355196)
(0.255286, 0.00358283)
(0.26324, 0.00359865)
(0.271442, 0.00356843)
(0.279899, 0.00359009)
(0.288621, 0.00357367)
(0.297613, 0.00360167)
(0.306886, 0.0036255)
(0.316448, 0.0036115)
(0.326308, 0.00364389)
(0.336476, 0.00360054)
(0.346959, 0.00364577)
(0.35777, 0.0036346)
(0.368917, 0.00361821)
(0.380412, 0.00373871)
(0.392265, 0.00368371)
(0.404487, 0.00372951)
(0.41709, 0.00376437)
(0.430086, 0.00372145)
(0.443487, 0.00365953)
(0.457305, 0.00374411)
(0.471554, 0.00375945)
(0.486247, 0.00373517)
(0.501397, 0.00374359)
(0.51702, 0.00380973)
(0.533129, 0.00370452)
(0.54974, 0.00367976)
(0.566869, 0.0037657)
(0.584532, 0.00374871)
(0.602745, 0.00381086)
(0.621525, 0.00390381)
(0.64089, 0.00378136)
(0.660859, 0.00379506)
(0.68145, 0.00380435)
(0.702683, 0.0038962)
(0.724577, 0.00382321)
(0.747154, 0.00385841)
(0.770434, 0.00386124)
(0.794439, 0.00392355)
(0.819192, 0.00386592)
(0.844717, 0.00381679)
(0.871036, 0.00395116)
(0.898176, 0.00384)
(0.926162, 0.00391038)
(0.955019, 0.00388904)
(0.984776, 0.00390304)
(1.01546, 0.00389703)
(1.0471, 0.00397884)
(1.07972, 0.00393676)
(1.11337, 0.00385681)
(1.14806, 0.00396151)
(1.18383, 0.00393108)
(1.22071, 0.00388548)
(1.25875, 0.00400757)
(1.29797, 0.0040198)
(1.33841, 0.00400933)
(1.38011, 0.00403018)
(1.42312, 0.00392045)
(1.46746, 0.00402482)
(1.51318, 0.00395963)
(1.56033, 0.00398534)
(1.60895, 0.00400455)
(1.65908, 0.00395123)
(1.71077, 0.00396203)
(1.76408, 0.00404151)
(1.81904, 0.00407413)
(1.87572, 0.00391624)
(1.93416, 0.0040008)
(1.99443, 0.00403463)
(2.05657, 0.00408005)
(2.12065, 0.00401971)
(2.18672, 0.00400299)
(2.25486, 0.00410613)
(2.32512, 0.00416756)
(2.39756, 0.00417587)
(2.47227, 0.00408459)
(2.5493, 0.00402767)
(2.62873, 0.00398008)
(2.71063, 0.00405827)
(2.79509, 0.00418483)
(2.88218, 0.00413454)
(2.97198, 0.00416381)
(3.06459, 0.00407119)
(3.16007, 0.00403145)
(3.25853, 0.00404792)
(3.36006, 0.00409696)
(3.46476, 0.00411097)
(3.57271, 0.00418202)
(3.68403, 0.0041588)
(3.79882, 0.00416153)
(3.91718, 0.00422937)
(4.03923, 0.0042988)
(4.16509, 0.00418893)
(4.29487, 0.00405502)
(4.42869, 0.0042412)
(4.56668, 0.00421551)
(4.70896, 0.00408246)
(4.85569, 0.00410865)
(5.00698, 0.00443269)
(5.16299, 0.00428007)
(5.32386, 0.00428551)
(5.48974, 0.00420533)
(5.66079, 0.00447843)
(5.83717, 0.00429695)
(6.01904, 0.00422466)
(6.20658, 0.00409286)
(6.39997, 0.00430494)
(6.59938, 0.00435563)
(6.805, 0.00433387)
(7.01704, 0.00412168)
(7.23567, 0.00472157)
(7.46112, 0.0044049)
(7.6936, 0.0043009)
(7.93332, 0.00437312)
(8.1805, 0.00455666)
(8.43539, 0.00418046)
(8.69822, 0.00430393)
(8.96924, 0.00448405)
(9.24871, 0.0043787)
(9.53688, 0.00428919)
(9.83403, 0.00440774)
(10.1404, 0.00423216)
(10.4564, 0.00426801)
(10.7822, 0.00440656)
(11.1181, 0.00419705)
(11.4646, 0.00409937)
(11.8218, 0.00443859)
(12.1901, 0.00430882)
(12.5699, 0.00417958)
(12.9616, 0.00456961)
(13.3655, 0.00431039)
(13.7819, 0.00390834)
(14.2113, 0.00428328)
(14.6541, 0.00425515)
(15.1107, 0.00431962)
(15.5815, 0.00422374)
(16.067, 0.00410214)
(16.5676, 0.00469392)
(17.0839, 0.00449596)
(17.6162, 0.00393662)
(18.165, 0.00434703)
(18.731, 0.00469062)
(19.3147, 0.00411245)
(19.9165, 0.00444037)
(20.537, 0.00467573)
(21.1769, 0.00463973)
(21.8368, 0.00462933)
(22.5171, 0.00431484)
(23.2187, 0.00422555)
(23.9422, 0.00467357)
(24.6882, 0.00467307)
(25.4574, 0.00410879)
(26.2506, 0.00430454)
(27.0685, 0.00464424)
(27.9119, 0.00408329)
(28.7816, 0.00464219)
(29.6784, 0.00456649)
(30.6031, 0.00440707)
(31.5567, 0.00408304)
(32.5399, 0.00464838)
(33.5538, 0.00418721)
(34.5993, 0.00423269)
(35.6773, 0.00452319)
(36.789, 0.00415553)
(37.9352, 0.0038523)
(39.1172, 0.00449166)
(40.336, 0.00389806)
(41.5928, 0.00408833)
(42.8888, 0.00480896)
(44.2251, 0.00458231)
(45.6031, 0.00409802)
(47.024, 0.00461791)
(48.4892, 0.00440548)
(50, 0.00468945)
}

\newcommand{\genOdePlotExpCifaruNcsnppErrHeunE}{%
(0.02, 0.000156081)
(0.0206232, 0.000163535)
(0.0212657, 0.000164766)
(0.0219283, 0.000166505)
(0.0226116, 0.000170842)
(0.0233161, 0.000177894)
(0.0240426, 0.000178814)
(0.0247917, 0.00018867)
(0.0255642, 0.000188015)
(0.0263607, 0.000190992)
(0.0271821, 0.000200165)
(0.028029, 0.000206175)
(0.0289023, 0.000207032)
(0.0298029, 0.000214696)
(0.0307315, 0.000215017)
(0.031689, 0.000221152)
(0.0326764, 0.000229901)
(0.0336945, 0.000230059)
(0.0347444, 0.000235822)
(0.035827, 0.000237389)
(0.0369432, 0.000246233)
(0.0380943, 0.00025353)
(0.0392813, 0.000267899)
(0.0405052, 0.000274423)
(0.0417673, 0.00027185)
(0.0430687, 0.000282832)
(0.0444106, 0.000294626)
(0.0457943, 0.00029153)
(0.0472212, 0.000289283)
(0.0486925, 0.000300879)
(0.0502097, 0.000304858)
(0.0517741, 0.000308753)
(0.0533873, 0.000320486)
(0.0550508, 0.000321183)
(0.056766, 0.000339447)
(0.0585348, 0.000349264)
(0.0603586, 0.000344045)
(0.0622393, 0.000357987)
(0.0641785, 0.000362617)
(0.0661782, 0.000369786)
(0.0682402, 0.000382873)
(0.0703664, 0.000395588)
(0.0725589, 0.000399262)
(0.0748197, 0.000414183)
(0.0771509, 0.000404973)
(0.0795548, 0.000425598)
(0.0820336, 0.000418593)
(0.0845896, 0.000442541)
(0.0872253, 0.000460851)
(0.089943, 0.000447952)
(0.0927455, 0.000474006)
(0.0956352, 0.000465815)
(0.0986151, 0.000490087)
(0.101688, 0.00049129)
(0.104856, 0.00050801)
(0.108123, 0.000522218)
(0.111492, 0.000549049)
(0.114966, 0.000531905)
(0.118548, 0.00054585)
(0.122242, 0.000528064)
(0.126051, 0.000564208)
(0.129978, 0.000576355)
(0.134028, 0.000604732)
(0.138204, 0.000609145)
(0.14251, 0.000642155)
(0.146951, 0.000631232)
(0.151529, 0.000634926)
(0.156251, 0.00064261)
(0.161119, 0.000641107)
(0.166139, 0.000664166)
(0.171316, 0.000694477)
(0.176654, 0.000709387)
(0.182158, 0.000717497)
(0.187834, 0.000730611)
(0.193686, 0.000743777)
(0.199721, 0.000758093)
(0.205944, 0.000776939)
(0.212361, 0.000812352)
(0.218978, 0.00080097)
(0.225801, 0.000799731)
(0.232836, 0.000849456)
(0.240091, 0.000852585)
(0.247572, 0.000869285)
(0.255286, 0.000904859)
(0.26324, 0.00090298)
(0.271442, 0.00091535)
(0.279899, 0.000954254)
(0.288621, 0.000944757)
(0.297613, 0.000979341)
(0.306886, 0.00100332)
(0.316448, 0.00104658)
(0.326308, 0.00105456)
(0.336476, 0.00107167)
(0.346959, 0.00109397)
(0.35777, 0.00106721)
(0.368917, 0.00113043)
(0.380412, 0.00118492)
(0.392265, 0.00119405)
(0.404487, 0.00118311)
(0.41709, 0.00124171)
(0.430086, 0.00124712)
(0.443487, 0.00124245)
(0.457305, 0.0013008)
(0.471554, 0.00134073)
(0.486247, 0.00133486)
(0.501397, 0.00138466)
(0.51702, 0.0014678)
(0.533129, 0.00140261)
(0.54974, 0.00141994)
(0.566869, 0.00149385)
(0.584532, 0.0015125)
(0.602745, 0.00154494)
(0.621525, 0.00159137)
(0.64089, 0.00160777)
(0.660859, 0.00164494)
(0.68145, 0.00165945)
(0.702683, 0.00173164)
(0.724577, 0.00179523)
(0.747154, 0.00175304)
(0.770434, 0.00178121)
(0.794439, 0.00185117)
(0.819192, 0.0018222)
(0.844717, 0.0018705)
(0.871036, 0.00207695)
(0.898176, 0.0019741)
(0.926162, 0.00203234)
(0.955019, 0.00205088)
(0.984776, 0.00209986)
(1.01546, 0.00216456)
(1.0471, 0.00213231)
(1.07972, 0.00229875)
(1.11337, 0.00226026)
(1.14806, 0.00231025)
(1.18383, 0.00238488)
(1.22071, 0.00232284)
(1.25875, 0.00253304)
(1.29797, 0.00254739)
(1.33841, 0.00257464)
(1.38011, 0.00257438)
(1.42312, 0.00255098)
(1.46746, 0.00268456)
(1.51318, 0.00273281)
(1.56033, 0.00284566)
(1.60895, 0.00273463)
(1.65908, 0.00282327)
(1.71077, 0.00292816)
(1.76408, 0.00306692)
(1.81904, 0.00313008)
(1.87572, 0.00305166)
(1.93416, 0.00317802)
(1.99443, 0.00321607)
(2.05657, 0.00335789)
(2.12065, 0.00337246)
(2.18672, 0.0033911)
(2.25486, 0.00352259)
(2.32512, 0.00370304)
(2.39756, 0.00371357)
(2.47227, 0.00374908)
(2.5493, 0.00376116)
(2.62873, 0.00376301)
(2.71063, 0.00392515)
(2.79509, 0.00410279)
(2.88218, 0.00410897)
(2.97198, 0.00424413)
(3.06459, 0.00418896)
(3.16007, 0.00427197)
(3.25853, 0.00434653)
(3.36006, 0.00449965)
(3.46476, 0.0045796)
(3.57271, 0.00464234)
(3.68403, 0.00474423)
(3.79882, 0.0048683)
(3.91718, 0.00493392)
(4.03923, 0.00509493)
(4.16509, 0.00518138)
(4.29487, 0.00524911)
(4.42869, 0.00538385)
(4.56668, 0.00550846)
(4.70896, 0.00535652)
(4.85569, 0.00553652)
(5.00698, 0.00586476)
(5.16299, 0.00575121)
(5.32386, 0.00586928)
(5.48974, 0.00592131)
(5.66079, 0.00637454)
(5.83717, 0.00647987)
(6.01904, 0.00618427)
(6.20658, 0.00619163)
(6.39997, 0.00661331)
(6.59938, 0.00647168)
(6.805, 0.00659397)
(7.01704, 0.00669849)
(7.23567, 0.00731264)
(7.46112, 0.00720128)
(7.6936, 0.00737672)
(7.93332, 0.00759154)
(8.1805, 0.00782703)
(8.43539, 0.00757946)
(8.69822, 0.00788681)
(8.96924, 0.00824677)
(9.24871, 0.00831226)
(9.53688, 0.00828973)
(9.83403, 0.0085357)
(10.1404, 0.00848912)
(10.4564, 0.0089215)
(10.7822, 0.00902311)
(11.1181, 0.00888114)
(11.4646, 0.00918574)
(11.8218, 0.00932602)
(12.1901, 0.0100203)
(12.5699, 0.00969564)
(12.9616, 0.0101693)
(13.3655, 0.010181)
(13.7819, 0.00979368)
(14.2113, 0.0103792)
(14.6541, 0.0106421)
(15.1107, 0.0104336)
(15.5815, 0.0109722)
(16.067, 0.0112014)
(16.5676, 0.0120309)
(17.0839, 0.0114975)
(17.6162, 0.0116811)
(18.165, 0.0119752)
(18.731, 0.012157)
(19.3147, 0.0125502)
(19.9165, 0.0127538)
(20.537, 0.0133004)
(21.1769, 0.0133597)
(21.8368, 0.0134404)
(22.5171, 0.0137614)
(23.2187, 0.0139209)
(23.9422, 0.0145648)
(24.6882, 0.0144959)
(25.4574, 0.014499)
(26.2506, 0.0146967)
(27.0685, 0.0152949)
(27.9119, 0.0149969)
(28.7816, 0.0163808)
(29.6784, 0.0158098)
(30.6031, 0.0164118)
(31.5567, 0.0167726)
(32.5399, 0.016886)
(33.5538, 0.0169362)
(34.5993, 0.0176813)
(35.6773, 0.0179202)
(36.789, 0.0177672)
(37.9352, 0.01806)
(39.1172, 0.0185538)
(40.336, 0.0181545)
(41.5928, 0.0186437)
(42.8888, 0.0194715)
(44.2251, 0.0197576)
(45.6031, 0.0199993)
(47.024, 0.0213292)
(48.4892, 0.0205804)
(50, 0.0222069)
}

\newcommand{\genOdePlotExpCifaruNcsnppErrHeunELo}{%
(0.02, 0.000132197)
(0.0206232, 0.00014068)
(0.0212657, 0.000141159)
(0.0219283, 0.000140272)
(0.0226116, 0.000143828)
(0.0233161, 0.000153485)
(0.0240426, 0.000151227)
(0.0247917, 0.000165047)
(0.0255642, 0.000164959)
(0.0263607, 0.000164254)
(0.0271821, 0.000175281)
(0.028029, 0.000182416)
(0.0289023, 0.000181873)
(0.0298029, 0.000187219)
(0.0307315, 0.000191565)
(0.031689, 0.000197262)
(0.0326764, 0.000200906)
(0.0336945, 0.000200701)
(0.0347444, 0.000208177)
(0.035827, 0.000214899)
(0.0369432, 0.00021954)
(0.0380943, 0.000227733)
(0.0392813, 0.000245192)
(0.0405052, 0.000244243)
(0.0417673, 0.000243565)
(0.0430687, 0.000258669)
(0.0444106, 0.000268549)
(0.0457943, 0.000261004)
(0.0472212, 0.000263822)
(0.0486925, 0.000274295)
(0.0502097, 0.000276197)
(0.0517741, 0.000278865)
(0.0533873, 0.000297164)
(0.0550508, 0.000293604)
(0.056766, 0.000312096)
(0.0585348, 0.000319447)
(0.0603586, 0.000315551)
(0.0622393, 0.000330215)
(0.0641785, 0.000334709)
(0.0661782, 0.00034054)
(0.0682402, 0.000348842)
(0.0703664, 0.000363476)
(0.0725589, 0.000368834)
(0.0748197, 0.000385405)
(0.0771509, 0.000376144)
(0.0795548, 0.000391239)
(0.0820336, 0.000386593)
(0.0845896, 0.000412689)
(0.0872253, 0.000429212)
(0.089943, 0.000418201)
(0.0927455, 0.000442348)
(0.0956352, 0.000434016)
(0.0986151, 0.000459699)
(0.101688, 0.000459137)
(0.104856, 0.000478106)
(0.108123, 0.000486881)
(0.111492, 0.000515407)
(0.114966, 0.000495324)
(0.118548, 0.000511822)
(0.122242, 0.000492232)
(0.126051, 0.000528961)
(0.129978, 0.000543077)
(0.134028, 0.000571321)
(0.138204, 0.000578235)
(0.14251, 0.000605791)
(0.146951, 0.000595927)
(0.151529, 0.000598502)
(0.156251, 0.000608418)
(0.161119, 0.000602831)
(0.166139, 0.000625555)
(0.171316, 0.000657129)
(0.176654, 0.000672709)
(0.182158, 0.000684148)
(0.187834, 0.000691674)
(0.193686, 0.000708846)
(0.199721, 0.00072118)
(0.205944, 0.000741471)
(0.212361, 0.000776612)
(0.218978, 0.00075897)
(0.225801, 0.000764266)
(0.232836, 0.000802791)
(0.240091, 0.000814598)
(0.247572, 0.000827778)
(0.255286, 0.000862225)
(0.26324, 0.000859215)
(0.271442, 0.000874916)
(0.279899, 0.000908822)
(0.288621, 0.00090377)
(0.297613, 0.00093508)
(0.306886, 0.000952431)
(0.316448, 0.00099603)
(0.326308, 0.00100216)
(0.336476, 0.00101628)
(0.346959, 0.0010446)
(0.35777, 0.00101868)
(0.368917, 0.0010684)
(0.380412, 0.0011301)
(0.392265, 0.00112927)
(0.404487, 0.0011173)
(0.41709, 0.00118111)
(0.430086, 0.0011894)
(0.443487, 0.00117967)
(0.457305, 0.00123677)
(0.471554, 0.0012787)
(0.486247, 0.00127316)
(0.501397, 0.00131072)
(0.51702, 0.00139618)
(0.533129, 0.00133293)
(0.54974, 0.00132949)
(0.566869, 0.00142647)
(0.584532, 0.00144793)
(0.602745, 0.00146647)
(0.621525, 0.00150243)
(0.64089, 0.0015178)
(0.660859, 0.00155952)
(0.68145, 0.00158253)
(0.702683, 0.00165145)
(0.724577, 0.00171554)
(0.747154, 0.00166003)
(0.770434, 0.00170453)
(0.794439, 0.0017413)
(0.819192, 0.00172811)
(0.844717, 0.00177319)
(0.871036, 0.00198856)
(0.898176, 0.00189846)
(0.926162, 0.00193836)
(0.955019, 0.00194846)
(0.984776, 0.00197987)
(1.01546, 0.00206312)
(1.0471, 0.00203543)
(1.07972, 0.00219267)
(1.11337, 0.00216705)
(1.14806, 0.00221359)
(1.18383, 0.00227951)
(1.22071, 0.00222212)
(1.25875, 0.00240791)
(1.29797, 0.0024185)
(1.33841, 0.00246911)
(1.38011, 0.00244708)
(1.42312, 0.00245279)
(1.46746, 0.00257466)
(1.51318, 0.00261263)
(1.56033, 0.00275074)
(1.60895, 0.00262062)
(1.65908, 0.00271339)
(1.71077, 0.00280999)
(1.76408, 0.00295733)
(1.81904, 0.00303808)
(1.87572, 0.00293108)
(1.93416, 0.00307296)
(1.99443, 0.00308959)
(2.05657, 0.00323084)
(2.12065, 0.00326912)
(2.18672, 0.00326886)
(2.25486, 0.00338819)
(2.32512, 0.00358529)
(2.39756, 0.00358082)
(2.47227, 0.00363016)
(2.5493, 0.00365452)
(2.62873, 0.00365235)
(2.71063, 0.003837)
(2.79509, 0.00399423)
(2.88218, 0.00399446)
(2.97198, 0.00414167)
(3.06459, 0.00408475)
(3.16007, 0.00415234)
(3.25853, 0.00419554)
(3.36006, 0.0043726)
(3.46476, 0.00445441)
(3.57271, 0.00453237)
(3.68403, 0.00461414)
(3.79882, 0.00470387)
(3.91718, 0.00479168)
(4.03923, 0.00493443)
(4.16509, 0.0049973)
(4.29487, 0.00506167)
(4.42869, 0.00518311)
(4.56668, 0.00535702)
(4.70896, 0.00514223)
(4.85569, 0.00530379)
(5.00698, 0.00560162)
(5.16299, 0.00549)
(5.32386, 0.00560156)
(5.48974, 0.00560153)
(5.66079, 0.00602187)
(5.83717, 0.00618134)
(6.01904, 0.00581699)
(6.20658, 0.00586237)
(6.39997, 0.00620395)
(6.59938, 0.00589289)
(6.805, 0.00617345)
(7.01704, 0.00620324)
(7.23567, 0.00663455)
(7.46112, 0.00687245)
(7.6936, 0.00701922)
(7.93332, 0.00701044)
(8.1805, 0.00753156)
(8.43539, 0.00719681)
(8.69822, 0.00744371)
(8.96924, 0.00789908)
(9.24871, 0.00809295)
(9.53688, 0.0080658)
(9.83403, 0.008171)
(10.1404, 0.00818736)
(10.4564, 0.00866257)
(10.7822, 0.00882132)
(11.1181, 0.00855883)
(11.4646, 0.00893899)
(11.8218, 0.00911855)
(12.1901, 0.0097786)
(12.5699, 0.00943743)
(12.9616, 0.00980643)
(13.3655, 0.0099145)
(13.7819, 0.00960983)
(14.2113, 0.0098339)
(14.6541, 0.010113)
(15.1107, 0.010211)
(15.5815, 0.0107343)
(16.067, 0.0109452)
(16.5676, 0.0117899)
(17.0839, 0.0112397)
(17.6162, 0.0114515)
(18.165, 0.0115624)
(18.731, 0.0116513)
(19.3147, 0.0121414)
(19.9165, 0.0125089)
(20.537, 0.0129369)
(21.1769, 0.0129566)
(21.8368, 0.0129969)
(22.5171, 0.0134441)
(23.2187, 0.013565)
(23.9422, 0.0141685)
(24.6882, 0.0139661)
(25.4574, 0.0140035)
(26.2506, 0.0144027)
(27.0685, 0.014489)
(27.9119, 0.0145968)
(28.7816, 0.0159735)
(29.6784, 0.015569)
(30.6031, 0.0161194)
(31.5567, 0.0163709)
(32.5399, 0.0163896)
(33.5538, 0.0166727)
(34.5993, 0.017315)
(35.6773, 0.0175204)
(36.789, 0.0174112)
(37.9352, 0.0176684)
(39.1172, 0.0179295)
(40.336, 0.0174789)
(41.5928, 0.0183219)
(42.8888, 0.0188736)
(44.2251, 0.0190191)
(45.6031, 0.0195486)
(47.024, 0.0207638)
(48.4892, 0.0200552)
(50, 0.0215298)
}

\newcommand{\genOdePlotExpCifaruNcsnppErrHeunEHi}{%
(0.02, 0.000179965)
(0.0206232, 0.00018639)
(0.0212657, 0.000188373)
(0.0219283, 0.000192738)
(0.0226116, 0.000197855)
(0.0233161, 0.000202303)
(0.0240426, 0.000206401)
(0.0247917, 0.000212293)
(0.0255642, 0.000211072)
(0.0263607, 0.00021773)
(0.0271821, 0.00022505)
(0.028029, 0.000229934)
(0.0289023, 0.000232191)
(0.0298029, 0.000242172)
(0.0307315, 0.000238469)
(0.031689, 0.000245042)
(0.0326764, 0.000258895)
(0.0336945, 0.000259417)
(0.0347444, 0.000263467)
(0.035827, 0.00025988)
(0.0369432, 0.000272925)
(0.0380943, 0.000279326)
(0.0392813, 0.000290606)
(0.0405052, 0.000304603)
(0.0417673, 0.000300135)
(0.0430687, 0.000306995)
(0.0444106, 0.000320702)
(0.0457943, 0.000322056)
(0.0472212, 0.000314744)
(0.0486925, 0.000327463)
(0.0502097, 0.000333519)
(0.0517741, 0.000338641)
(0.0533873, 0.000343808)
(0.0550508, 0.000348763)
(0.056766, 0.000366798)
(0.0585348, 0.000379082)
(0.0603586, 0.000372538)
(0.0622393, 0.000385759)
(0.0641785, 0.000390525)
(0.0661782, 0.000399032)
(0.0682402, 0.000416905)
(0.0703664, 0.0004277)
(0.0725589, 0.00042969)
(0.0748197, 0.000442961)
(0.0771509, 0.000433802)
(0.0795548, 0.000459956)
(0.0820336, 0.000450592)
(0.0845896, 0.000472393)
(0.0872253, 0.00049249)
(0.089943, 0.000477704)
(0.0927455, 0.000505665)
(0.0956352, 0.000497614)
(0.0986151, 0.000520475)
(0.101688, 0.000523444)
(0.104856, 0.000537914)
(0.108123, 0.000557555)
(0.111492, 0.00058269)
(0.114966, 0.000568486)
(0.118548, 0.000579878)
(0.122242, 0.000563896)
(0.126051, 0.000599455)
(0.129978, 0.000609633)
(0.134028, 0.000638142)
(0.138204, 0.000640055)
(0.14251, 0.000678519)
(0.146951, 0.000666538)
(0.151529, 0.00067135)
(0.156251, 0.000676801)
(0.161119, 0.000679383)
(0.166139, 0.000702778)
(0.171316, 0.000731824)
(0.176654, 0.000746065)
(0.182158, 0.000750846)
(0.187834, 0.000769548)
(0.193686, 0.000778708)
(0.199721, 0.000795006)
(0.205944, 0.000812407)
(0.212361, 0.000848093)
(0.218978, 0.00084297)
(0.225801, 0.000835197)
(0.232836, 0.000896121)
(0.240091, 0.000890571)
(0.247572, 0.000910792)
(0.255286, 0.000947492)
(0.26324, 0.000946745)
(0.271442, 0.000955784)
(0.279899, 0.000999687)
(0.288621, 0.000985745)
(0.297613, 0.0010236)
(0.306886, 0.00105421)
(0.316448, 0.00109712)
(0.326308, 0.00110695)
(0.336476, 0.00112706)
(0.346959, 0.00114335)
(0.35777, 0.00111574)
(0.368917, 0.00119247)
(0.380412, 0.00123973)
(0.392265, 0.00125883)
(0.404487, 0.00124893)
(0.41709, 0.00130231)
(0.430086, 0.00130484)
(0.443487, 0.00130523)
(0.457305, 0.00136483)
(0.471554, 0.00140275)
(0.486247, 0.00139656)
(0.501397, 0.00145861)
(0.51702, 0.00153942)
(0.533129, 0.00147229)
(0.54974, 0.00151039)
(0.566869, 0.00156123)
(0.584532, 0.00157707)
(0.602745, 0.00162342)
(0.621525, 0.00168031)
(0.64089, 0.00169774)
(0.660859, 0.00173036)
(0.68145, 0.00173637)
(0.702683, 0.00181183)
(0.724577, 0.00187492)
(0.747154, 0.00184604)
(0.770434, 0.0018579)
(0.794439, 0.00196105)
(0.819192, 0.0019163)
(0.844717, 0.00196781)
(0.871036, 0.00216533)
(0.898176, 0.00204975)
(0.926162, 0.00212633)
(0.955019, 0.0021533)
(0.984776, 0.00221984)
(1.01546, 0.002266)
(1.0471, 0.00222919)
(1.07972, 0.00240482)
(1.11337, 0.00235347)
(1.14806, 0.00240691)
(1.18383, 0.00249025)
(1.22071, 0.00242356)
(1.25875, 0.00265816)
(1.29797, 0.00267627)
(1.33841, 0.00268018)
(1.38011, 0.00270168)
(1.42312, 0.00264916)
(1.46746, 0.00279446)
(1.51318, 0.002853)
(1.56033, 0.00294057)
(1.60895, 0.00284863)
(1.65908, 0.00293316)
(1.71077, 0.00304632)
(1.76408, 0.00317651)
(1.81904, 0.00322209)
(1.87572, 0.00317224)
(1.93416, 0.00328309)
(1.99443, 0.00334255)
(2.05657, 0.00348494)
(2.12065, 0.0034758)
(2.18672, 0.00351334)
(2.25486, 0.003657)
(2.32512, 0.0038208)
(2.39756, 0.00384633)
(2.47227, 0.00386799)
(2.5493, 0.0038678)
(2.62873, 0.00387366)
(2.71063, 0.00401329)
(2.79509, 0.00421135)
(2.88218, 0.00422348)
(2.97198, 0.0043466)
(3.06459, 0.00429317)
(3.16007, 0.0043916)
(3.25853, 0.00449752)
(3.36006, 0.00462671)
(3.46476, 0.00470478)
(3.57271, 0.00475232)
(3.68403, 0.00487431)
(3.79882, 0.00503273)
(3.91718, 0.00507616)
(4.03923, 0.00525543)
(4.16509, 0.00536546)
(4.29487, 0.00543655)
(4.42869, 0.0055846)
(4.56668, 0.0056599)
(4.70896, 0.0055708)
(4.85569, 0.00576924)
(5.00698, 0.0061279)
(5.16299, 0.00601241)
(5.32386, 0.006137)
(5.48974, 0.00624109)
(5.66079, 0.00672721)
(5.83717, 0.00677841)
(6.01904, 0.00655155)
(6.20658, 0.00652088)
(6.39997, 0.00702267)
(6.59938, 0.00705047)
(6.805, 0.0070145)
(7.01704, 0.00719373)
(7.23567, 0.00799074)
(7.46112, 0.00753012)
(7.6936, 0.00773421)
(7.93332, 0.00817264)
(8.1805, 0.00812251)
(8.43539, 0.00796211)
(8.69822, 0.00832991)
(8.96924, 0.00859445)
(9.24871, 0.00853157)
(9.53688, 0.00851366)
(9.83403, 0.0089004)
(10.1404, 0.00879088)
(10.4564, 0.00918043)
(10.7822, 0.0092249)
(11.1181, 0.00920344)
(11.4646, 0.0094325)
(11.8218, 0.00953349)
(12.1901, 0.010262)
(12.5699, 0.00995385)
(12.9616, 0.0105321)
(13.3655, 0.0104475)
(13.7819, 0.00997754)
(14.2113, 0.0109246)
(14.6541, 0.0111712)
(15.1107, 0.0106562)
(15.5815, 0.0112101)
(16.067, 0.0114576)
(16.5676, 0.0122719)
(17.0839, 0.0117554)
(17.6162, 0.0119108)
(18.165, 0.0123881)
(18.731, 0.0126627)
(19.3147, 0.0129589)
(19.9165, 0.0129986)
(20.537, 0.0136638)
(21.1769, 0.0137627)
(21.8368, 0.0138839)
(22.5171, 0.0140788)
(23.2187, 0.0142767)
(23.9422, 0.0149611)
(24.6882, 0.0150257)
(25.4574, 0.0149944)
(26.2506, 0.0149906)
(27.0685, 0.0161008)
(27.9119, 0.015397)
(28.7816, 0.016788)
(29.6784, 0.0160505)
(30.6031, 0.0167041)
(31.5567, 0.0171744)
(32.5399, 0.0173823)
(33.5538, 0.0171998)
(34.5993, 0.0180477)
(35.6773, 0.01832)
(36.789, 0.0181232)
(37.9352, 0.0184516)
(39.1172, 0.019178)
(40.336, 0.0188301)
(41.5928, 0.0189655)
(42.8888, 0.0200693)
(44.2251, 0.0204961)
(45.6031, 0.02045)
(47.024, 0.0218946)
(48.4892, 0.0211057)
(50, 0.0228839)
}

\newcommand{\genOdePlotExpCifaruNcsnppFidHeun}{%
(1, 455.803)
(1.5, 231.336)
(2, 25.2514)
(2.5, 12.325)
(3, 7.9415)
(3.5, 6.4068)
(4, 5.8359)
(4.5, 5.6592)
(5, 5.511)
(5.5, 5.4855)
(6, 5.5538)
(6.5, 5.5993)
(7, 5.5331)
(7.5, 5.6458)
(8, 5.706)
(8.5, 5.7929)
(9, 5.8209)
(9.5, 5.8622)
(10, 5.9274)
(11, 5.9916)
(12, 6.1237)
(13, 6.2261)
(14, 6.1688)
(15, 6.3499)
(16, 6.1815)
(17, 6.384)
(18, 6.363)
(19, 6.3824)
(20, 6.5839)
}

\newcommand{\genOdePlotExpCifaruNcsnppFidHeunLo}{%
(1, 455.803)
(1.5, 231.336)
(2, 25.2514)
(2.5, 12.325)
(3, 7.9415)
(3.5, 6.4068)
(4, 5.8359)
(4.5, 5.6592)
(5, 5.511)
(5.5, 5.4855)
(6, 5.5538)
(6.5, 5.5993)
(7, 5.5331)
(7.5, 5.6458)
(8, 5.706)
(8.5, 5.7929)
(9, 5.8209)
(9.5, 5.8622)
(10, 5.9274)
(11, 5.9916)
(12, 6.1237)
(13, 6.2261)
(14, 6.1688)
(15, 6.3499)
(16, 6.1815)
(17, 6.384)
(18, 6.363)
(19, 6.3824)
(20, 6.5839)
}

\newcommand{\genOdePlotExpCifaruNcsnppFidHeunHi}{%
(1, 456.053)
(1.5, 231.439)
(2, 25.3716)
(2.5, 12.4852)
(3, 8.0099)
(3.5, 6.5101)
(4, 5.8822)
(4.5, 5.7875)
(5, 5.623)
(5.5, 5.609)
(6, 5.6381)
(6.5, 5.7536)
(7, 5.6049)
(7.5, 5.715)
(8, 5.7975)
(8.5, 5.9216)
(9, 5.8587)
(9.5, 5.9896)
(10, 6.011)
(11, 6.1036)
(12, 6.2343)
(13, 6.2484)
(14, 6.3451)
(15, 6.3754)
(16, 6.3177)
(17, 6.5025)
(18, 6.436)
(19, 6.5611)
(20, 6.6146)
}

\newcommand{\genOdePlotExpCifaruNcsnppFidHeunMarks}{%
(7, 5.5331)
}

\newcommand{\genOdePlotExpCifaruDdpmppFidHeun}{%
(1, 453.936)
(1.5, 150.576)
(2, 93.6837)
(2.5, 32.7488)
(3, 4.1762)
(3.5, 3.6928)
(4, 3.5918)
(4.5, 3.5258)
(5, 3.4162)
(5.5, 3.3893)
(6, 3.3429)
(6.5, 3.351)
(7, 3.2798)
(7.5, 3.293)
(8, 3.2981)
(8.5, 3.2599)
(9, 3.2768)
(9.5, 3.3007)
(10, 3.3016)
}

\newcommand{\genOdePlotExpCifaruDdpmppFidHeunLo}{%
(1, 453.936)
(1.5, 150.576)
(2, 93.6837)
(2.5, 32.7488)
(3, 4.1762)
(3.5, 3.6928)
(4, 3.5918)
(4.5, 3.5258)
(5, 3.4162)
(5.5, 3.3893)
(6, 3.3429)
(6.5, 3.351)
(7, 3.2798)
(7.5, 3.293)
(8, 3.2981)
(8.5, 3.2599)
(9, 3.2768)
(9.5, 3.3007)
(10, 3.3016)
}

\newcommand{\genOdePlotExpCifaruDdpmppFidHeunHi}{%
(1, 454.092)
(1.5, 150.783)
(2, 94.2017)
(2.5, 32.8802)
(3, 4.2475)
(3.5, 3.7895)
(4, 3.6216)
(4.5, 3.5744)
(5, 3.4759)
(5.5, 3.4508)
(6, 3.3779)
(6.5, 3.3874)
(7, 3.3454)
(7.5, 3.35)
(8, 3.3553)
(8.5, 3.324)
(9, 3.3649)
(9.5, 3.3967)
(10, 3.3534)
}

\newcommand{\genOdePlotExpCifaruDdpmppFidHeunMarks}{%
(7, 3.2798)
}

\newcommand{\genOdePlotExpImgcDhariwalFidHeun}{%
(1, 398.501)
(1.5, 230.184)
(2, 66.3787)
(2.5, 21.2751)
(3, 8.292)
(3.5, 4.6431)
(4, 3.6426)
(4.5, 3.3116)
(5, 3.177)
(5.5, 3.1447)
(6, 3.1642)
(6.5, 3.1754)
(7, 3.1145)
(7.5, 3.1324)
(8, 3.144)
(8.5, 3.1809)
(9, 3.1499)
(9.5, 3.2079)
(10, 3.2077)
}

\newcommand{\genOdePlotExpImgcDhariwalFidHeunLo}{%
(1, 398.501)
(1.5, 230.184)
(2, 66.3787)
(2.5, 21.2751)
(3, 8.292)
(3.5, 4.6431)
(4, 3.6426)
(4.5, 3.3116)
(5, 3.177)
(5.5, 3.1447)
(6, 3.1642)
(6.5, 3.1754)
(7, 3.1145)
(7.5, 3.1324)
(8, 3.144)
(8.5, 3.1809)
(9, 3.1499)
(9.5, 3.2079)
(10, 3.2077)
}

\newcommand{\genOdePlotExpImgcDhariwalFidHeunHi}{%
(1, 398.69)
(1.5, 230.346)
(2, 66.6483)
(2.5, 21.3627)
(3, 8.3484)
(3.5, 4.671)
(4, 3.7624)
(4.5, 3.4098)
(5, 3.2636)
(5.5, 3.1938)
(6, 3.2015)
(6.5, 3.2236)
(7, 3.2036)
(7.5, 3.2374)
(8, 3.1971)
(8.5, 3.1923)
(9, 3.2655)
(9.5, 3.2398)
(10, 3.2723)
}

\newcommand{\genOdePlotExpImgcDhariwalFidHeunMarks}{%
(7, 3.1145)
}

\newcommand{\genSdePlotChurnCifaruDdpmppOde}{%
(0, 2.9325)
(100, 2.9325)
}

\newcommand{\genSdePlotChurnCifaruDdpmppPlain}{%
(0, 2.9325)
(5, 2.7562)
(10, 2.7115)
(15, 2.7837)
(20, 2.8301)
(25, 2.9396)
(30, 2.9777)
(35, 3.1896)
(40, 3.2702)
(45, 3.3546)
(50, 3.4931)
(55, 3.5662)
(60, 3.6771)
(65, 3.7932)
(70, 3.9502)
(75, 3.9849)
(80, 4.1461)
(85, 4.2446)
(90, 4.3144)
(95, 4.299)
(100, 4.3921)
}

\newcommand{\genSdePlotChurnCifaruDdpmppPlainLo}{%
(0, 2.9325)
(5, 2.7562)
(10, 2.7115)
(15, 2.7837)
(20, 2.8301)
(25, 2.9396)
(30, 2.9777)
(35, 3.1896)
(40, 3.2702)
(45, 3.3546)
(50, 3.4931)
(55, 3.5662)
(60, 3.6771)
(65, 3.7932)
(70, 3.9502)
(75, 3.9849)
(80, 4.1461)
(85, 4.2446)
(90, 4.3144)
(95, 4.299)
(100, 4.3921)
}

\newcommand{\genSdePlotChurnCifaruDdpmppPlainHi}{%
(0, 2.9889)
(5, 2.8115)
(10, 2.772)
(15, 2.8062)
(20, 2.8798)
(25, 2.9616)
(30, 3.0665)
(35, 3.2328)
(40, 3.314)
(45, 3.441)
(50, 3.5388)
(55, 3.6657)
(60, 3.8061)
(65, 3.8782)
(70, 3.9729)
(75, 4.1281)
(80, 4.2464)
(85, 4.3058)
(90, 4.3906)
(95, 4.5166)
(100, 4.5626)
}

\newcommand{\genSdePlotChurnCifaruDdpmppLambda}{%
(0, 2.9325)
(5, 2.6822)
(10, 2.6075)
(15, 2.5552)
(20, 2.5663)
(25, 2.6286)
(30, 2.6061)
(35, 2.7095)
(40, 2.8286)
(45, 2.9184)
(50, 2.9695)
(55, 3.1231)
(60, 3.1397)
(65, 3.264)
(70, 3.3403)
(75, 3.4404)
(80, 3.5589)
(85, 3.5243)
(90, 3.5669)
(95, 3.6944)
(100, 3.7211)
}

\newcommand{\genSdePlotChurnCifaruDdpmppLambdaLo}{%
(0, 2.9325)
(5, 2.6822)
(10, 2.6075)
(15, 2.5552)
(20, 2.5663)
(25, 2.6286)
(30, 2.6061)
(35, 2.7095)
(40, 2.8286)
(45, 2.9184)
(50, 2.9695)
(55, 3.1231)
(60, 3.1397)
(65, 3.264)
(70, 3.3403)
(75, 3.4404)
(80, 3.5589)
(85, 3.5243)
(90, 3.5669)
(95, 3.6944)
(100, 3.7211)
}

\newcommand{\genSdePlotChurnCifaruDdpmppLambdaHi}{%
(0, 2.9889)
(5, 2.7005)
(10, 2.6548)
(15, 2.5954)
(20, 2.6596)
(25, 2.6602)
(30, 2.7451)
(35, 2.8195)
(40, 2.8541)
(45, 2.9854)
(50, 3.0559)
(55, 3.1601)
(60, 3.2113)
(65, 3.3388)
(70, 3.4203)
(75, 3.496)
(80, 3.5729)
(85, 3.6225)
(90, 3.6554)
(95, 3.7431)
(100, 3.7675)
}

\newcommand{\genSdePlotChurnCifaruDdpmppRange}{%
(0, 2.9325)
(5, 2.731)
(10, 2.6408)
(15, 2.5932)
(20, 2.5764)
(25, 2.5449)
(30, 2.5355)
(35, 2.5724)
(40, 2.6334)
(45, 2.7212)
(50, 2.7025)
(55, 2.8057)
(60, 2.8647)
(65, 2.8955)
(70, 2.9902)
(75, 3.06)
(80, 3.1548)
(85, 3.1733)
(90, 3.2971)
(95, 3.3452)
(100, 3.3635)
}

\newcommand{\genSdePlotChurnCifaruDdpmppRangeLo}{%
(0, 2.9325)
(5, 2.731)
(10, 2.6408)
(15, 2.5932)
(20, 2.5764)
(25, 2.5449)
(30, 2.5355)
(35, 2.5724)
(40, 2.6334)
(45, 2.7212)
(50, 2.7025)
(55, 2.8057)
(60, 2.8647)
(65, 2.8955)
(70, 2.9902)
(75, 3.06)
(80, 3.1548)
(85, 3.1733)
(90, 3.2971)
(95, 3.3452)
(100, 3.3635)
}

\newcommand{\genSdePlotChurnCifaruDdpmppRangeHi}{%
(0, 2.9889)
(5, 2.8032)
(10, 2.7101)
(15, 2.6256)
(20, 2.5999)
(25, 2.5996)
(30, 2.6134)
(35, 2.6404)
(40, 2.6649)
(45, 2.7456)
(50, 2.7819)
(55, 2.857)
(60, 2.919)
(65, 2.9823)
(70, 3.0404)
(75, 3.1426)
(80, 3.2328)
(85, 3.2896)
(90, 3.3433)
(95, 3.3684)
(100, 3.4996)
}

\newcommand{\genSdePlotChurnCifaruDdpmppOurs}{%
(0, 2.9325)
(5, 2.6727)
(10, 2.5744)
(15, 2.4327)
(20, 2.3358)
(25, 2.3527)
(30, 2.2738)
(35, 2.3143)
(40, 2.3215)
(45, 2.3613)
(50, 2.325)
(55, 2.3848)
(60, 2.4057)
(65, 2.4339)
(70, 2.4797)
(75, 2.5291)
(80, 2.5801)
(85, 2.5831)
(90, 2.6149)
(95, 2.6859)
(100, 2.7303)
}

\newcommand{\genSdePlotChurnCifaruDdpmppOursLo}{%
(0, 2.9325)
(5, 2.6727)
(10, 2.5744)
(15, 2.4327)
(20, 2.3358)
(25, 2.3527)
(30, 2.2738)
(35, 2.3143)
(40, 2.3215)
(45, 2.3613)
(50, 2.325)
(55, 2.3848)
(60, 2.4057)
(65, 2.4339)
(70, 2.4797)
(75, 2.5291)
(80, 2.5801)
(85, 2.5831)
(90, 2.6149)
(95, 2.6859)
(100, 2.7303)
}

\newcommand{\genSdePlotChurnCifaruDdpmppOursHi}{%
(0, 2.9889)
(5, 2.7267)
(10, 2.5912)
(15, 2.5097)
(20, 2.3967)
(25, 2.3823)
(30, 2.3506)
(35, 2.3393)
(40, 2.4031)
(45, 2.3721)
(50, 2.3996)
(55, 2.4351)
(60, 2.4643)
(65, 2.496)
(70, 2.5358)
(75, 2.5371)
(80, 2.6014)
(85, 2.6438)
(90, 2.6469)
(95, 2.7123)
(100, 2.7678)
}

\newcommand{\genSdePlotChurnCifaruNcsnppOde}{%
(0, 3.7343)
(100, 3.7343)
}

\newcommand{\genSdePlotChurnCifaruNcsnppPlain}{%
(0, 3.7343)
(5, 3.4057)
(10, 3.184)
(15, 3.0818)
(20, 3.0249)
(25, 3.03)
(30, 3.0332)
(35, 3.0715)
(40, 3.0802)
(45, 3.1725)
(50, 3.2336)
(55, 3.2007)
(60, 3.3866)
(65, 3.4397)
(70, 3.4508)
(75, 3.5381)
(80, 3.6618)
(85, 3.6509)
(90, 3.7019)
(95, 3.738)
(100, 3.8445)
}

\newcommand{\genSdePlotChurnCifaruNcsnppPlainLo}{%
(0, 3.7343)
(5, 3.4057)
(10, 3.184)
(15, 3.0818)
(20, 3.0249)
(25, 3.03)
(30, 3.0332)
(35, 3.0715)
(40, 3.0802)
(45, 3.1725)
(50, 3.2336)
(55, 3.2007)
(60, 3.3866)
(65, 3.4397)
(70, 3.4508)
(75, 3.5381)
(80, 3.6618)
(85, 3.6509)
(90, 3.7019)
(95, 3.738)
(100, 3.8445)
}

\newcommand{\genSdePlotChurnCifaruNcsnppPlainHi}{%
(0, 3.8719)
(5, 3.5015)
(10, 3.2385)
(15, 3.1201)
(20, 3.0814)
(25, 3.0929)
(30, 3.1182)
(35, 3.1316)
(40, 3.1553)
(45, 3.1929)
(50, 3.3041)
(55, 3.3381)
(60, 3.4236)
(65, 3.4863)
(70, 3.4995)
(75, 3.5856)
(80, 3.6789)
(85, 3.74)
(90, 3.8074)
(95, 3.8481)
(100, 3.9051)
}

\newcommand{\genSdePlotChurnCifaruNcsnppLambda}{%
(0, 3.7343)
(5, 3.289)
(10, 3.0645)
(15, 2.855)
(20, 2.7632)
(25, 2.6829)
(30, 2.6336)
(35, 2.6258)
(40, 2.6203)
(45, 2.589)
(50, 2.607)
(55, 2.6187)
(60, 2.5828)
(65, 2.6421)
(70, 2.6265)
(75, 2.6657)
(80, 2.6747)
(85, 2.7169)
(90, 2.7075)
(95, 2.7454)
(100, 2.7221)
}

\newcommand{\genSdePlotChurnCifaruNcsnppLambdaLo}{%
(0, 3.7343)
(5, 3.289)
(10, 3.0645)
(15, 2.855)
(20, 2.7632)
(25, 2.6829)
(30, 2.6336)
(35, 2.6258)
(40, 2.6203)
(45, 2.589)
(50, 2.607)
(55, 2.6187)
(60, 2.5828)
(65, 2.6421)
(70, 2.6265)
(75, 2.6657)
(80, 2.6747)
(85, 2.7169)
(90, 2.7075)
(95, 2.7454)
(100, 2.7221)
}

\newcommand{\genSdePlotChurnCifaruNcsnppLambdaHi}{%
(0, 3.8719)
(5, 3.3529)
(10, 3.1027)
(15, 2.9191)
(20, 2.7838)
(25, 2.6956)
(30, 2.7033)
(35, 2.6499)
(40, 2.6461)
(45, 2.6303)
(50, 2.6295)
(55, 2.6576)
(60, 2.6693)
(65, 2.6597)
(70, 2.6543)
(75, 2.6997)
(80, 2.7008)
(85, 2.7316)
(90, 2.7722)
(95, 2.7786)
(100, 2.78)
}

\newcommand{\genSdePlotChurnCifaruNcsnppRange}{%
(0, 3.7343)
(5, 3.4405)
(10, 3.2862)
(15, 3.0857)
(20, 3.0334)
(25, 2.9204)
(30, 2.8965)
(35, 2.9227)
(40, 2.8987)
(45, 2.8683)
(50, 2.874)
(55, 2.9261)
(60, 2.9341)
(65, 2.9503)
(70, 2.984)
(75, 3.0081)
(80, 3.0368)
(85, 3.0608)
(90, 3.1431)
(95, 3.1295)
(100, 3.1471)
}

\newcommand{\genSdePlotChurnCifaruNcsnppRangeLo}{%
(0, 3.7343)
(5, 3.4405)
(10, 3.2862)
(15, 3.0857)
(20, 3.0334)
(25, 2.9204)
(30, 2.8965)
(35, 2.9227)
(40, 2.8987)
(45, 2.8683)
(50, 2.874)
(55, 2.9261)
(60, 2.9341)
(65, 2.9503)
(70, 2.984)
(75, 3.0081)
(80, 3.0368)
(85, 3.0608)
(90, 3.1431)
(95, 3.1295)
(100, 3.1471)
}

\newcommand{\genSdePlotChurnCifaruNcsnppRangeHi}{%
(0, 3.8719)
(5, 3.5371)
(10, 3.3119)
(15, 3.1719)
(20, 3.06)
(25, 3.0063)
(30, 2.9682)
(35, 2.9681)
(40, 2.9357)
(45, 2.9051)
(50, 2.894)
(55, 2.9549)
(60, 2.9623)
(65, 2.9952)
(70, 3.0446)
(75, 3.0753)
(80, 3.0636)
(85, 3.1808)
(90, 3.1458)
(95, 3.1908)
(100, 3.2046)
}

\newcommand{\genSdePlotChurnCifaruNcsnppOurs}{%
(0, 3.7343)
(5, 3.3556)
(10, 3.0772)
(15, 2.913)
(20, 2.7897)
(25, 2.6445)
(30, 2.5774)
(35, 2.5194)
(40, 2.4641)
(45, 2.408)
(50, 2.3798)
(55, 2.3687)
(60, 2.3485)
(65, 2.3592)
(70, 2.3166)
(75, 2.3072)
(80, 2.323)
(85, 2.3316)
(90, 2.3432)
(95, 2.2605)
(100, 2.3154)
}

\newcommand{\genSdePlotChurnCifaruNcsnppOursLo}{%
(0, 3.7343)
(5, 3.3556)
(10, 3.0772)
(15, 2.913)
(20, 2.7897)
(25, 2.6445)
(30, 2.5774)
(35, 2.5194)
(40, 2.4641)
(45, 2.408)
(50, 2.3798)
(55, 2.3687)
(60, 2.3485)
(65, 2.3592)
(70, 2.3166)
(75, 2.3072)
(80, 2.323)
(85, 2.3316)
(90, 2.3432)
(95, 2.2605)
(100, 2.3154)
}

\newcommand{\genSdePlotChurnCifaruNcsnppOursHi}{%
(0, 3.8719)
(5, 3.4672)
(10, 3.2023)
(15, 2.9532)
(20, 2.8392)
(25, 2.7233)
(30, 2.6342)
(35, 2.5613)
(40, 2.5025)
(45, 2.459)
(50, 2.4252)
(55, 2.4524)
(60, 2.3985)
(65, 2.4007)
(70, 2.3444)
(75, 2.3513)
(80, 2.3651)
(85, 2.3536)
(90, 2.349)
(95, 2.3328)
(100, 2.349)
}

\newcommand{\genSdePlotChurnImgcDhariwalOde}{%
(0, 2.6363)
(100, 2.6363)
}

\newcommand{\genSdePlotChurnImgcDhariwalPlain}{%
(0, 2.6363)
(5, 2.3085)
(10, 2.0985)
(15, 1.9795)
(20, 1.937)
(25, 1.956)
(30, 1.9198)
(35, 1.913)
(40, 1.8943)
(45, 1.9153)
(50, 1.894)
(55, 1.9121)
(60, 1.9572)
(65, 2.0337)
(70, 2.04)
(75, 2.0524)
(80, 2.075)
(85, 2.108)
(90, 2.1101)
(95, 2.1488)
(100, 2.1295)
}

\newcommand{\genSdePlotChurnImgcDhariwalPlainLo}{%
(0, 2.6363)
(5, 2.3085)
(10, 2.0985)
(15, 1.9795)
(20, 1.937)
(25, 1.956)
(30, 1.9198)
(35, 1.913)
(40, 1.8943)
(45, 1.9153)
(50, 1.894)
(55, 1.9121)
(60, 1.9572)
(65, 2.0337)
(70, 2.04)
(75, 2.0524)
(80, 2.075)
(85, 2.108)
(90, 2.1101)
(95, 2.1488)
(100, 2.1295)
}

\newcommand{\genSdePlotChurnImgcDhariwalPlainHi}{%
(0, 2.715)
(5, 2.3336)
(10, 2.1941)
(15, 2.081)
(20, 1.9991)
(25, 1.9871)
(30, 1.9779)
(35, 1.9438)
(40, 1.9339)
(45, 1.9618)
(50, 1.9499)
(55, 1.9716)
(60, 2.0366)
(65, 2.0437)
(70, 2.0735)
(75, 2.1114)
(80, 2.1185)
(85, 2.1304)
(90, 2.2056)
(95, 2.1631)
(100, 2.1799)
}

\newcommand{\genSdePlotChurnImgcDhariwalLambda}{%
(0, 2.6363)
(5, 2.2007)
(10, 2.0151)
(15, 1.922)
(20, 1.8673)
(25, 1.816)
(30, 1.7823)
(35, 1.7172)
(40, 1.6966)
(45, 1.6881)
(50, 1.7101)
(55, 1.6667)
(60, 1.7217)
(65, 1.7388)
(70, 1.7237)
(75, 1.7175)
(80, 1.6852)
(85, 1.6911)
(90, 1.6945)
(95, 1.7261)
(100, 1.6949)
}

\newcommand{\genSdePlotChurnImgcDhariwalLambdaLo}{%
(0, 2.6363)
(5, 2.2007)
(10, 2.0151)
(15, 1.922)
(20, 1.8673)
(25, 1.816)
(30, 1.7823)
(35, 1.7172)
(40, 1.6966)
(45, 1.6881)
(50, 1.7101)
(55, 1.6667)
(60, 1.7217)
(65, 1.7388)
(70, 1.7237)
(75, 1.7175)
(80, 1.6852)
(85, 1.6911)
(90, 1.6945)
(95, 1.7261)
(100, 1.6949)
}

\newcommand{\genSdePlotChurnImgcDhariwalLambdaHi}{%
(0, 2.715)
(5, 2.2649)
(10, 2.1099)
(15, 1.9743)
(20, 1.9094)
(25, 1.8463)
(30, 1.8124)
(35, 1.7832)
(40, 1.7382)
(45, 1.746)
(50, 1.7401)
(55, 1.7003)
(60, 1.7354)
(65, 1.7497)
(70, 1.7598)
(75, 1.751)
(80, 1.7534)
(85, 1.7658)
(90, 1.7978)
(95, 1.7532)
(100, 1.7554)
}

\newcommand{\genSdePlotChurnImgcDhariwalRange}{%
(0, 2.6363)
(5, 2.3381)
(10, 2.1104)
(15, 2.0041)
(20, 1.9355)
(25, 1.9148)
(30, 1.867)
(35, 1.8629)
(40, 1.8616)
(45, 1.8312)
(50, 1.8015)
(55, 1.8569)
(60, 1.8517)
(65, 1.8726)
(70, 1.8678)
(75, 1.8994)
(80, 1.8719)
(85, 1.9089)
(90, 1.9419)
(95, 1.966)
(100, 1.9542)
}

\newcommand{\genSdePlotChurnImgcDhariwalRangeLo}{%
(0, 2.6363)
(5, 2.3381)
(10, 2.1104)
(15, 2.0041)
(20, 1.9355)
(25, 1.9148)
(30, 1.867)
(35, 1.8629)
(40, 1.8616)
(45, 1.8312)
(50, 1.8015)
(55, 1.8569)
(60, 1.8517)
(65, 1.8726)
(70, 1.8678)
(75, 1.8994)
(80, 1.8719)
(85, 1.9089)
(90, 1.9419)
(95, 1.966)
(100, 1.9542)
}

\newcommand{\genSdePlotChurnImgcDhariwalRangeHi}{%
(0, 2.715)
(5, 2.3771)
(10, 2.1706)
(15, 2.0722)
(20, 1.9888)
(25, 1.9306)
(30, 1.9066)
(35, 1.9182)
(40, 1.8817)
(45, 1.854)
(50, 1.8589)
(55, 1.8646)
(60, 1.8954)
(65, 1.9185)
(70, 1.8912)
(75, 1.9381)
(80, 1.9392)
(85, 1.9719)
(90, 1.9774)
(95, 2.0023)
(100, 2.0593)
}

\newcommand{\genSdePlotChurnImgcDhariwalOurs}{%
(0, 2.6363)
(5, 2.2657)
(10, 2.0252)
(15, 1.9378)
(20, 1.8601)
(25, 1.7925)
(30, 1.7597)
(35, 1.7429)
(40, 1.6788)
(45, 1.6419)
(50, 1.6742)
(55, 1.6285)
(60, 1.6324)
(65, 1.6022)
(70, 1.6166)
(75, 1.5726)
(80, 1.5874)
(85, 1.62)
(90, 1.5889)
(95, 1.6179)
(100, 1.6075)
}

\newcommand{\genSdePlotChurnImgcDhariwalOursLo}{%
(0, 2.6363)
(5, 2.2657)
(10, 2.0252)
(15, 1.9378)
(20, 1.8601)
(25, 1.7925)
(30, 1.7597)
(35, 1.7429)
(40, 1.6788)
(45, 1.6419)
(50, 1.6742)
(55, 1.6285)
(60, 1.6324)
(65, 1.6022)
(70, 1.6166)
(75, 1.5726)
(80, 1.5874)
(85, 1.62)
(90, 1.5889)
(95, 1.6179)
(100, 1.6075)
}

\newcommand{\genSdePlotChurnImgcDhariwalOursHi}{%
(0, 2.715)
(5, 2.3206)
(10, 2.0709)
(15, 2.0056)
(20, 1.9126)
(25, 1.8176)
(30, 1.7766)
(35, 1.7964)
(40, 1.7226)
(45, 1.7034)
(50, 1.6926)
(55, 1.6769)
(60, 1.6668)
(65, 1.6381)
(70, 1.6736)
(75, 1.6274)
(80, 1.6226)
(85, 1.6433)
(90, 1.6198)
(95, 1.6314)
(100, 1.6554)
}

\newcommand{\genSdePlotChurnCifaruDdpmppPlainMarks}{%
(10, 2.7115)
}

\newcommand{\genSdePlotChurnCifaruDdpmppLambdaMarks}{%
(20, 2.5663)
}

\newcommand{\genSdePlotChurnCifaruDdpmppRangeMarks}{%
(30, 2.5355)
}

\newcommand{\genSdePlotChurnCifaruDdpmppOursMarks}{%
(30, 2.2738)
}

\newcommand{\genSdePlotChurnCifaruNcsnppPlainMarks}{%
(20, 3.0249)
}

\newcommand{\genSdePlotChurnCifaruNcsnppLambdaMarks}{%
(50, 2.607)
}

\newcommand{\genSdePlotChurnCifaruNcsnppRangeMarks}{%
(50, 2.874)
}

\newcommand{\genSdePlotChurnCifaruNcsnppOursMarks}{%
(80, 2.323)
}

\newcommand{\genSdePlotChurnImgcDhariwalPlainMarks}{%
(40, 1.8943)
}

\newcommand{\genSdePlotChurnImgcDhariwalLambdaMarks}{%
(60, 1.7217)
}

\newcommand{\genSdePlotChurnImgcDhariwalRangeMarks}{%
(50, 1.8015)
}

\newcommand{\genSdePlotChurnImgcDhariwalOursMarks}{%
(80, 1.5874)
}

\newcommand{\genSdeTable}{%
\multicolumn{1}{l|}{}                                                        & \multicolumn{4}{c|}{Unconditional CIFAR-10 at 32$\times$32}                   & \multicolumn{2}{c|}{Class-conditional}  \\
\multicolumn{1}{l|}{}                                                        & \multicolumn{2}{c|}{VP}               & \multicolumn{2}{c|}{VE}               & \multicolumn{2}{c|}{ImageNet-64}        \\
\tabucline{-}
{\bf Sampling method}                                                        & FID~$\downarrow$  & NFE~$\downarrow$  & FID~$\downarrow$  & NFE~$\downarrow$  & FID~$\downarrow$  & NFE~$\downarrow$    \\
\tabucline{-}
Deterministic baseline (Alg.~\refpaper{alg:heun})                            & 2.93              & {\bf\s35}         & 3.73              & {\bf\s\s27}       & 2.64              & {\bf\s79}           \\
Alg.~\refpaper{alg:stochastic}, $\StminStmax = [0, \infty]$, $\Snoise = 1$   & 2.69              & \s95              & 2.97              & \s383             & 1.86              & 383                 \\
Alg.~\refpaper{alg:stochastic}, $\StminStmax = [0, \infty]$                  & 2.54              & 127               & 2.51              & \s511             & 1.63              & 767                 \\
Alg.~\refpaper{alg:stochastic}, $\Snoise = 1$                                & 2.52              & \s95              & 2.84              & \s191             & 1.84              & 255                 \\
Alg.~\refpaper{alg:stochastic}, Optimal settings                             & {\bf2.27}         & 383               & {\bf2.23}         & \s767             & {\bf1.55}         & 511                 \\
\tabucline{-}
Previous work~\cite{Song2021sde,Dhariwal2021}                                & 2.55              & 768               & 2.46              & 1024              & 2.01              & 384                 \\
}

\newcommand{\tabOdeTable}{%
\forcenewcolumntype{x}{>{\centering\arraybackslash\hspace{0pt}}p{13mm}}%
\renewcommand{\cs}{\hspace{0.5mm}}%
\tabulinesep=0.7mm%
\tabulinestyle{0.17mm}%
\begin{table}[t]%
\centering%
\footnotesize%
\caption{\label{tab:OdeTable}%
Evaluation of our improvements to deterministic sampling.
The values correspond to the curves shown in Figure~\refpaper{fig:OdePlotNfe}.
We summarize each curve with two key values: the lowest observed FID for any NFE (``FID''), and the lowest NFE whose FID is within 3\% of the lowest FID (``NFE'').
The values marked with ``--'' are identical to the ones above them, because our sampler uses the same $\sigma(t)$ and $s(t)$ as DDIM.
}
\vspace{1.5mm}%
\begin{tabu}{|@{\ \ }l@{\ \ }|x@{\cs}x|x@{\cs}x|x@{\cs}x|}
\tabucline{2-}
\genOdeTable
\tabucline{-}
\end{tabu}%
\end{table}%
}

\newcommand{\figOdePlotExp}{%
\renewcommand{\hh}{85mm}\renewcommand{\vv}{75mm}\renewcommand{\hhh}{-3mm}%
\begin{figure}[t]
\centering%
\resizebox{\textwidth}{!}{%
\hspace*{-.5mm}%
\begin{tikzpicture}%
\begin{axis}[
  width=\hh, height=\vv,
  xmin={0.02}, xmax={50}, xmode={log}, xtick={0.02, 0.1, 0.5, 1, 2, 5, 10, 20, 50}, xticklabels={\tickSigma{0.02}, $0.1$, $0.5$, $1$, $2$, $5$, $10$, $20$, $50$},
  ymin={5e-5}, ymax={500}, ymode={log}, ytick={1e-4, 1e-3, 1e-2, 1e-1, 1e+0, 1e+1, 1e+2, 500}, yticklabels={$10^{\text{-}4}$, $10^{\text{-}3}$, $10^{\text{-}2}$, $10^{\text{-}1}$, $10^0$, $10^1$, $10^2$, \tickTau},
  grid={major}, legend pos={north east}, legend cell align={left}, legend columns={-1}, legend style={font=\normalsize, /tikz/every even column/.append style={column sep=1.35mm}},
]
\fillbetween[C0, opacity=0.2, forget plot]{coordinates {\genOdePlotExpCifaruNcsnppErrEulerALo}}{coordinates {\genOdePlotExpCifaruNcsnppErrEulerAHi}};
\fillbetween[C1, opacity=0.2, forget plot]{coordinates {\genOdePlotExpCifaruNcsnppErrEulerBLo}}{coordinates {\genOdePlotExpCifaruNcsnppErrEulerBHi}};
\fillbetween[C2, opacity=0.2, forget plot]{coordinates {\genOdePlotExpCifaruNcsnppErrEulerCLo}}{coordinates {\genOdePlotExpCifaruNcsnppErrEulerCHi}};
\fillbetween[C3, opacity=0.2, forget plot]{coordinates {\genOdePlotExpCifaruNcsnppErrEulerDLo}}{coordinates {\genOdePlotExpCifaruNcsnppErrEulerDHi}};
\fillbetween[C4, opacity=0.2, forget plot]{coordinates {\genOdePlotExpCifaruNcsnppErrEulerELo}}{coordinates {\genOdePlotExpCifaruNcsnppErrEulerEHi}};
\addplot[C0] coordinates {\genOdePlotExpCifaruNcsnppErrEulerA};
\addplot[C1] coordinates {\genOdePlotExpCifaruNcsnppErrEulerB};
\addplot[C2] coordinates {\genOdePlotExpCifaruNcsnppErrEulerC};
\addplot[C3] coordinates {\genOdePlotExpCifaruNcsnppErrEulerD};
\addplot[C4] coordinates {\genOdePlotExpCifaruNcsnppErrEulerE};
\legend{
  {$\rho=1.0$},
  {$1.5$},
  {$2.0$},
  {$3.0$},
  {$7.0$},
}
\end{axis}
\end{tikzpicture}
\hspace*{\hhh}
\begin{tikzpicture}
\begin{axis}[
  width=\hh, height=\vv,
  xmin={0.02}, xmax={50}, xmode={log}, xtick={0.02, 0.1, 0.5, 1, 2, 5, 10, 20, 50}, xticklabels={\tickSigma{0.02\hspace*{-1em}}, $0.1$, $0.5$, $1$, $2$, $5$, $10$, $20$, $50$},
  ymin={5e-5}, ymax={500}, ymode={log}, ytick={1e-4, 1e-3, 1e-2, 1e-1, 1e+0, 1e+1, 1e+2, 500}, yticklabels={$10^{\text{-}4}$, $10^{\text{-}3}$, $10^{\text{-}2}$, $10^{\text{-}1}$, $10^0$, $10^1$, $10^2$, \tickTau},
  grid={major}, legend pos={north east}, legend cell align={left}, legend columns={-1}, legend style={font=\normalsize, /tikz/every even column/.append style={column sep=1.35mm}},
]
\fillbetween[C0, opacity=0.2, forget plot]{coordinates {\genOdePlotExpCifaruNcsnppErrHeunALo}}{coordinates {\genOdePlotExpCifaruNcsnppErrHeunAHi}};
\fillbetween[C1, opacity=0.2, forget plot]{coordinates {\genOdePlotExpCifaruNcsnppErrHeunBLo}}{coordinates {\genOdePlotExpCifaruNcsnppErrHeunBHi}};
\fillbetween[C2, opacity=0.2, forget plot]{coordinates {\genOdePlotExpCifaruNcsnppErrHeunCLo}}{coordinates {\genOdePlotExpCifaruNcsnppErrHeunCHi}};
\fillbetween[C3, opacity=0.2, forget plot]{coordinates {\genOdePlotExpCifaruNcsnppErrHeunDLo}}{coordinates {\genOdePlotExpCifaruNcsnppErrHeunDHi}};
\fillbetween[C4, opacity=0.2, forget plot]{coordinates {\genOdePlotExpCifaruNcsnppErrHeunELo}}{coordinates {\genOdePlotExpCifaruNcsnppErrHeunEHi}};
\addplot[C0] coordinates {\genOdePlotExpCifaruNcsnppErrHeunA};
\addplot[C1] coordinates {\genOdePlotExpCifaruNcsnppErrHeunB};
\addplot[C2] coordinates {\genOdePlotExpCifaruNcsnppErrHeunC};
\addplot[C3] coordinates {\genOdePlotExpCifaruNcsnppErrHeunD};
\addplot[C4] coordinates {\genOdePlotExpCifaruNcsnppErrHeunE};
\legend{
  {$\rho=1.0$},
  {$1.5$},
  {$2.0$},
  {$3.0$},
  {$7.0$},
}
\end{axis}
\end{tikzpicture}
\hspace*{\hhh}
\begin{tikzpicture}
\begin{axis}[
  width=\hh, height=\vv,
  xmin={1}, xmax={10}, xmode={linear}, xtick={1, 2, 3, 4, 5, 6, 7, 8, 9, 10}, xticklabels={\tickRho{1}, $2$, $3$, $4$, $5$, $6$, $7$, $8$, $9$, $10$},
  ymin={2.5}, ymax={11.5}, ymode={log}, ytick={2, 3, 4, 5, 6, 7, 8, 9, 10, 11.5}, yticklabels={$2$, $3$, $4$, $5$, $6$, $7$, $8$, $9$, $10$, \tickFID},
  grid={major}, legend pos={north east}, legend cell align={left}, legend style={font=\normalsize},
]
\fillbetween[C0, opacity=0.2, forget plot]{coordinates {\genOdePlotExpCifaruDdpmppFidHeunLo}}{coordinates {\genOdePlotExpCifaruDdpmppFidHeunHi}};
\fillbetween[C1, opacity=0.2, forget plot]{coordinates {\genOdePlotExpCifaruNcsnppFidHeunLo}}{coordinates {\genOdePlotExpCifaruNcsnppFidHeunHi}};
\fillbetween[C2, opacity=0.2, forget plot]{coordinates {\genOdePlotExpImgcDhariwalFidHeunLo}}{coordinates {\genOdePlotExpImgcDhariwalFidHeunHi}};
\addplot[C0] coordinates {\genOdePlotExpCifaruDdpmppFidHeun};
\addplot[C1] coordinates {\genOdePlotExpCifaruNcsnppFidHeun};
\addplot[C2] coordinates {\genOdePlotExpImgcDhariwalFidHeun};
\addplot[C0, only marks, forget plot] coordinates {\genOdePlotExpCifaruDdpmppFidHeunMarks};
\addplot[C1, only marks, forget plot] coordinates {\genOdePlotExpCifaruNcsnppFidHeunMarks};
\addplot[C2, only marks, forget plot] coordinates {\genOdePlotExpImgcDhariwalFidHeunMarks};
\legend{
  {CIFAR-10, VP, $N=32$},
  {CIFAR-10, VE, $N=64$},
  {ImageNet-64, $N=12$},
}
\end{axis}
\end{tikzpicture}
}\\\hfill%
\makebox[0.33\linewidth]{\footnotesize (a) Truncation error, VE + Euler}\hfill%
\makebox[0.33\linewidth]{\footnotesize (b) Truncation error, VE + Heun}\hfill%
\makebox[0.33\linewidth]{\footnotesize (c) FID as a function of $\rho$}\hfill%
\caption{\label{fig:OdePlotExp}%
\textbf{(a)}
Local truncation error ($y$-axis) at different noise levels ($x$-axis) using Euler's method with the VE-based CIFAR-10 model.
Each curve corresponds to a different time step discretization, defined for $N=64$ and a specific choice for the polynomial exponent $\rho$.
The values represent the root mean square error (RMSE) between one Euler iteration and a sequence of multiple smaller Euler iterations, representing the ground truth.
The shaded regions, barely visible at low $\sigma$, represent standard deviation over different latents $\xx_0$.
\textbf{(b)}
Corresponding error curves for Heun's 2\textsuperscript{nd} order method (Algorithm~\refpaper{alg:heun}).
\textbf{(c)}
FID ($y$-axis) as a function of the polynomial exponent ($x$-axis) for different models, measured using Heun's 2\textsuperscript{nd} order method.
The shaded regions indicate the range of variation between the lowest and highest observed FID, and the dots indicate the value of $\rho$ that we use in all other experiments.
}
\end{figure}
}

\newcommand{\figSdePlotChurn}{%
\renewcommand{\hh}{85mm}\renewcommand{\vv}{70mm}\renewcommand{\hhh}{-3mm}%
\begin{figure}[t]
\centering%
\resizebox{\textwidth}{!}{%
\hspace*{-1mm}%
\begin{tikzpicture}%
\begin{axis}[
  width=\hh, height=\vv,
  xmin={0}, xmax={100}, xmode={linear}, xtick={0, 10, 20, 30, 40, 50, 60, 70, 80, 90, 100}, xticklabels={\tickSchurnB{0}, $10$, $20$, $30$, $40$, $50$, $60$, $70$, $80$, $90$, $100$},
  ymin={1.9}, ymax={5.4}, ymode={linear}, ytick={1.0, 1.5, 2.0, 2.5, 3.0, 3.5, 4.0, 4.5, 5.0, 5.4}, yticklabels={$1.0$, $1.5$, $2.0$, $2.5$, $3.0$, $3.5$, $4.0$, $4.5$, $5.0$, \tickFID},
  grid={major}, legend pos={north west}, legend cell align={left}, legend style={font=\small},
]
\fillbetween[C2, opacity=0.15, forget plot]{coordinates {\genSdePlotChurnCifaruDdpmppPlainLo}}{coordinates {\genSdePlotChurnCifaruDdpmppPlainHi}};
\fillbetween[C1, opacity=0.15, forget plot]{coordinates {\genSdePlotChurnCifaruDdpmppLambdaLo}}{coordinates {\genSdePlotChurnCifaruDdpmppLambdaHi}};
\fillbetween[C0, opacity=0.15, forget plot]{coordinates {\genSdePlotChurnCifaruDdpmppRangeLo}}{coordinates {\genSdePlotChurnCifaruDdpmppRangeHi}};
\fillbetween[C4, opacity=0.15, forget plot]{coordinates {\genSdePlotChurnCifaruDdpmppOursLo}}{coordinates {\genSdePlotChurnCifaruDdpmppOursHi}};
\addplot[C3, dashed]                  coordinates {\genSdePlotChurnCifaruDdpmppOde};
\addplot[C2]                          coordinates {\genSdePlotChurnCifaruDdpmppPlain};
\addplot[C1]                          coordinates {\genSdePlotChurnCifaruDdpmppLambda};
\addplot[C0]                          coordinates {\genSdePlotChurnCifaruDdpmppRange};
\addplot[C4]                          coordinates {\genSdePlotChurnCifaruDdpmppOurs};
\legend{
  {Deterministic},
  {$\StminStmax$ + $\Snoise = 1$},
  {$\StminStmax = [0, \infty]$},
  {$\Snoise = 1$},
  {Optimal settings},
}
\end{axis}
\end{tikzpicture}
\hspace*{\hhh}
\begin{tikzpicture}
\begin{axis}[
  width=\hh, height=\vv,
  xmin={0}, xmax={100}, xmode={linear}, xtick={0, 10, 20, 30, 40, 50, 60, 70, 80, 90, 100}, xticklabels={$0$, $10$, $20$, $30$, $40$, $50$, $60$, $70$, $80$, $90$, $100$},
  ymin={1.9}, ymax={4.3}, ymode={linear}, ytick={1.0, 1.5, 2.0, 2.5, 3.0, 3.5, 4.0, 4.3}, yticklabels={$1.0$, $1.5$, $2.0$, $2.5$, $3.0$, $3.5$, $4.0$, \tickFID},
  grid={major}, legend pos={north west}, legend cell align={left}, legend style={font=\small},
]
\fillbetween[C2, opacity=0.15, forget plot]{coordinates {\genSdePlotChurnCifaruNcsnppPlainLo}}{coordinates {\genSdePlotChurnCifaruNcsnppPlainHi}};
\fillbetween[C1, opacity=0.15, forget plot]{coordinates {\genSdePlotChurnCifaruNcsnppLambdaLo}}{coordinates {\genSdePlotChurnCifaruNcsnppLambdaHi}};
\fillbetween[C0, opacity=0.15, forget plot]{coordinates {\genSdePlotChurnCifaruNcsnppRangeLo}}{coordinates {\genSdePlotChurnCifaruNcsnppRangeHi}};
\fillbetween[C4, opacity=0.15, forget plot]{coordinates {\genSdePlotChurnCifaruNcsnppOursLo}}{coordinates {\genSdePlotChurnCifaruNcsnppOursHi}};
\addplot[C3, dashed]                  coordinates {\genSdePlotChurnCifaruNcsnppOde};
\addplot[C2]                          coordinates {\genSdePlotChurnCifaruNcsnppPlain};
\addplot[C1]                          coordinates {\genSdePlotChurnCifaruNcsnppLambda};
\addplot[C0]                          coordinates {\genSdePlotChurnCifaruNcsnppRange};
\addplot[C4]                          coordinates {\genSdePlotChurnCifaruNcsnppOurs};
\end{axis}
\end{tikzpicture}
\hspace*{\hhh}
\begin{tikzpicture}
\begin{axis}[
  width=\hh, height=\vv,
  xmin={0}, xmax={100}, xmode={linear}, xtick={0, 10, 20, 30, 40, 50, 60, 70, 80, 90, 100}, xticklabels={$0$, $10$, $20$, $30$, $40$, $50$, $60$, $70$, $80$, $90$, $100$},
  ymin={1.3}, ymax={3.0}, ymode={linear}, ytick={1.0, 1.2, 1.4, 1.6, 1.8, 2.0, 2.2, 2.4, 2.6, 2.8, 3.0}, yticklabels={$1.0$, $1.2$, $1.4$, $1.6$, $1.8$, $2.0$, $2.2$, $2.4$, $2.6$, $2.8$, \tickFID},
  grid={major}, legend pos={north west}, legend cell align={left}, legend style={font=\small},
]
\fillbetween[C2, opacity=0.15, forget plot]{coordinates {\genSdePlotChurnImgcDhariwalPlainLo}}{coordinates {\genSdePlotChurnImgcDhariwalPlainHi}};
\fillbetween[C1, opacity=0.15, forget plot]{coordinates {\genSdePlotChurnImgcDhariwalLambdaLo}}{coordinates {\genSdePlotChurnImgcDhariwalLambdaHi}};
\fillbetween[C0, opacity=0.15, forget plot]{coordinates {\genSdePlotChurnImgcDhariwalRangeLo}}{coordinates {\genSdePlotChurnImgcDhariwalRangeHi}};
\fillbetween[C4, opacity=0.15, forget plot]{coordinates {\genSdePlotChurnImgcDhariwalOursLo}}{coordinates {\genSdePlotChurnImgcDhariwalOursHi}};
\addplot[C3, dashed]                  coordinates {\genSdePlotChurnImgcDhariwalOde};
\addplot[C2]                          coordinates {\genSdePlotChurnImgcDhariwalPlain};
\addplot[C1]                          coordinates {\genSdePlotChurnImgcDhariwalLambda};
\addplot[C0]                          coordinates {\genSdePlotChurnImgcDhariwalRange};
\addplot[C4]                          coordinates {\genSdePlotChurnImgcDhariwalOurs};
\end{axis}
\end{tikzpicture}
}\\\hfill%
\makebox[0.33\linewidth]{\footnotesize (a) Uncond. CIFAR-10, VP}\hfill%
\makebox[0.33\linewidth]{\footnotesize (b) Uncond. CIFAR-10, VE}\hfill%
\makebox[0.33\linewidth]{\footnotesize (c) Class-cond. ImageNet-64}\hfill%
\caption{\label{fig:SdePlotChurn}%
Ablations of our stochastic sampler (Algorithm~\refpaper{alg:stochastic}) parameters using pre-trained networks of Song~et~al.~\cite{Song2021sde} and Dhariwal and Nichol~\cite{Dhariwal2021}.
Each curve shows FID ($y$-axis) as a function of $\Schurn$ ($x$-axis) for $N = 256$ steps (NFE = 511).
The dashed red lines correspond to our deterministic sampler (Algorithm~\refpaper{alg:heun}), equivalent to setting $\Schurn = 0$.
The purple curves correspond to optimal choices for $\{\Stmin, \Stmax, \Snoise\}$, found separately for each case using grid search.
Orange, blue, and green correspond to disabling the effects of $\StminStmax$ and/or $\Snoise$.
The shaded regions indicate the range of variation between the lowest and highest observed FID.
}
\end{figure}
}

\newcommand{\tabSdeTable}{%
\forcenewcolumntype{x}{>{\centering\arraybackslash\hspace{0pt}}p{12mm}}%
\renewcommand{\cs}{\hspace{0.5mm}}%
\tabulinesep=0.7mm%
\tabulinestyle{0.17mm}%
\begin{table}[t]%
\centering%
\footnotesize%
\caption{\label{tab:SdeTable}%
Evaluation and ablations of our improvements to stochastic sampling.
The values correspond to the curves shown in Figure~\refpaper{fig:SdePlotNfe}.
}
\vspace{1.5mm}%
\begin{tabu}{|@{\ \ }l@{\ \ }|x@{\cs}x|x@{\cs}x|x@{\cs}x|}
\tabucline{2-}
\genSdeTable
\tabucline{-}
\end{tabu}%
\end{table}%
}

\newcommand{\imgclasses}{\vlabel{\hhh}{\vv}{\makebox[\vv]{\scriptsize\hrlabel{Agaric}\hrlabel{Daisy}\hrlabel{Valley}\hrlabel{Pizza}\hrlabel{Balloon}\hrlabel{Beagle}\hrlabel{Ostrich}}}}

\newcommand{\figGridsImgcSampling}{%
\renewcommand{\hh}{0.478\linewidth}
\renewcommand{\vv}{\hh/\real{5}*\real{7}}%
\renewcommand{\hhh}{2.5mm}
\begin{figure}[p]%
\centering%
\hspace{\hhh}\makebox[\hh]{Deterministic, Original sampler (DDIM)}\hfill%
\hspace{\hhh}\makebox[\hh]{Deterministic, Our sampler (Alg.~\refpaper{alg:heun})}\\%
\imgclasses\includegraphics[width=\hh]{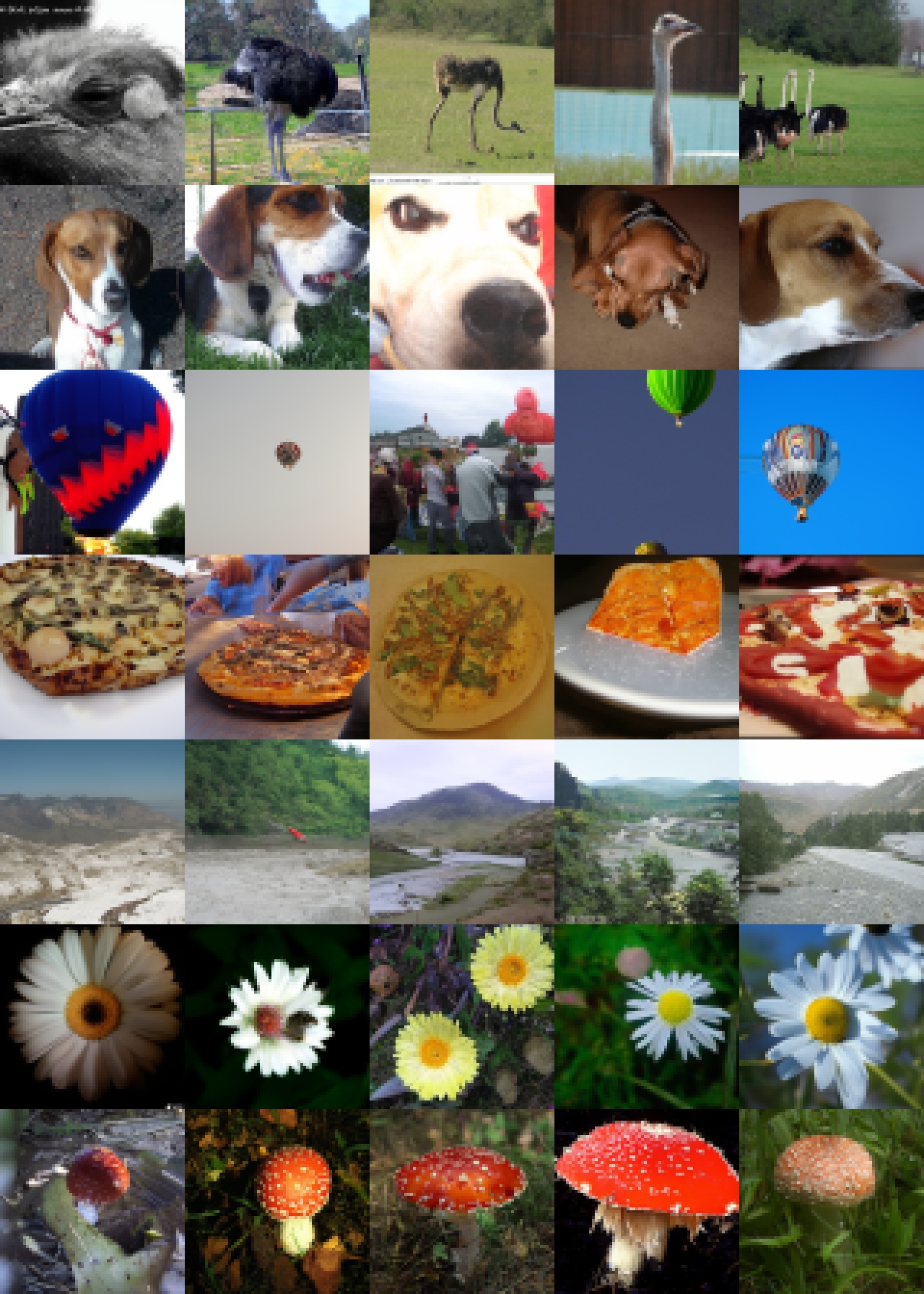}\hfill%
\imgclasses\includegraphics[width=\hh]{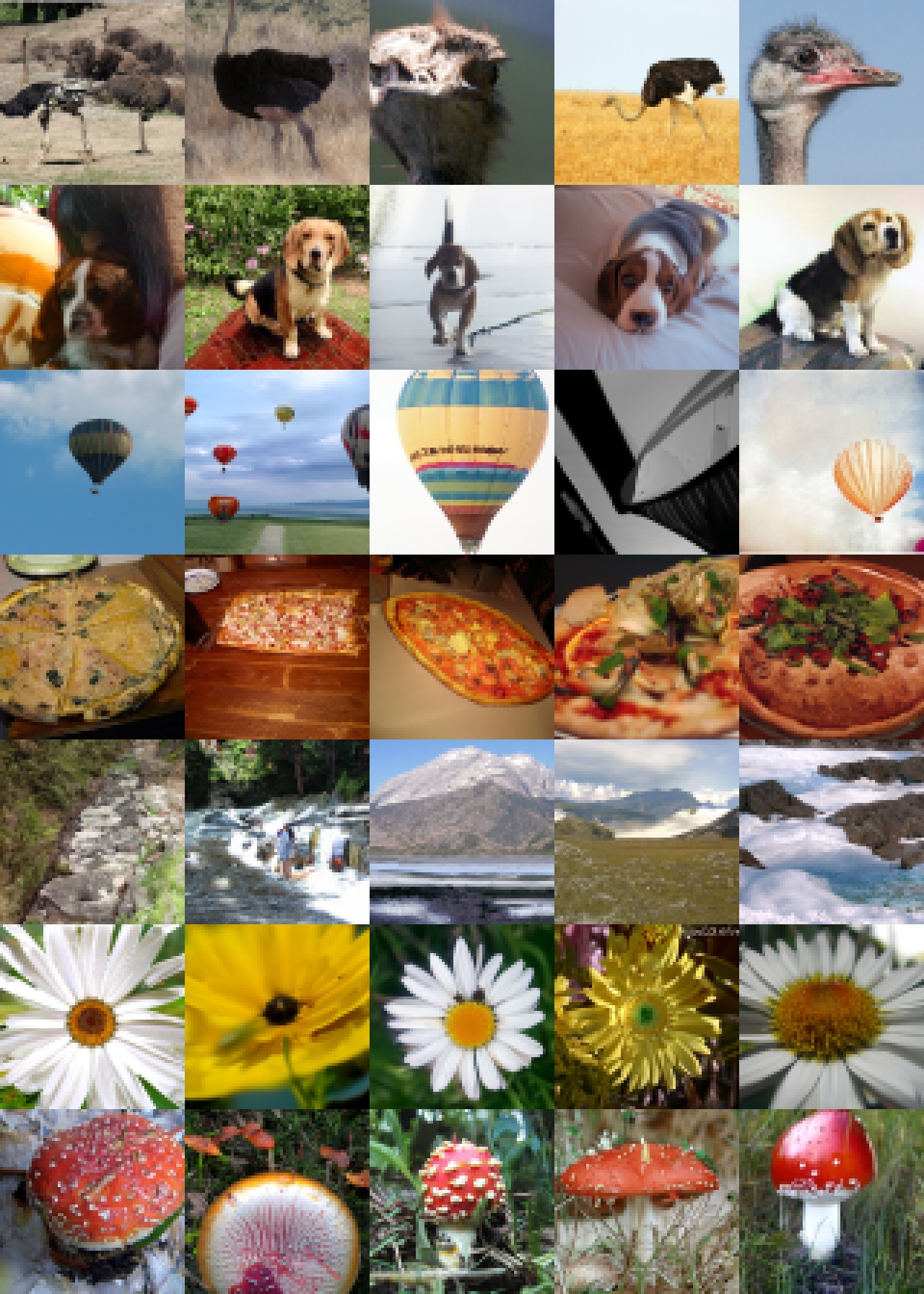}\\[-0.9mm]%
\hspace{\hhh}\makebox[\hh]{\footnotesize FID 2.91~~~~NFE 250}\hfill%
\hspace{\hhh}\makebox[\hh]{\footnotesize FID 2.66~~~~NFE {\bf 79}}\\[3mm]%
\hspace{\hhh}\makebox[\hh]{Stochastic, Original sampler (iDDPM)}\hfill%
\hspace{\hhh}\makebox[\hh]{Stochastic, Our sampler (Alg.~\refpaper{alg:stochastic})}\\%
\imgclasses\includegraphics[width=\hh]{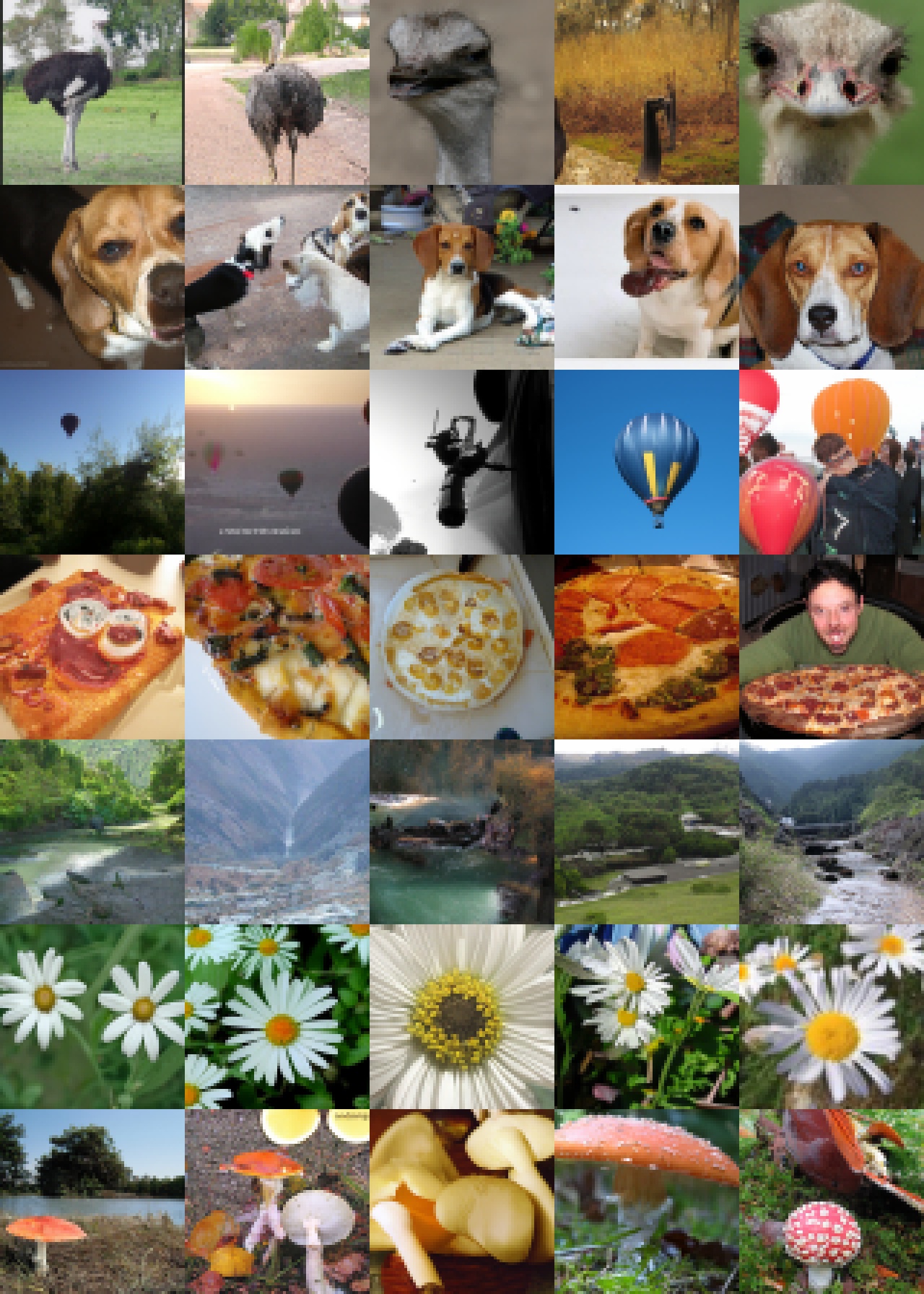}\hfill%
\imgclasses\includegraphics[width=\hh]{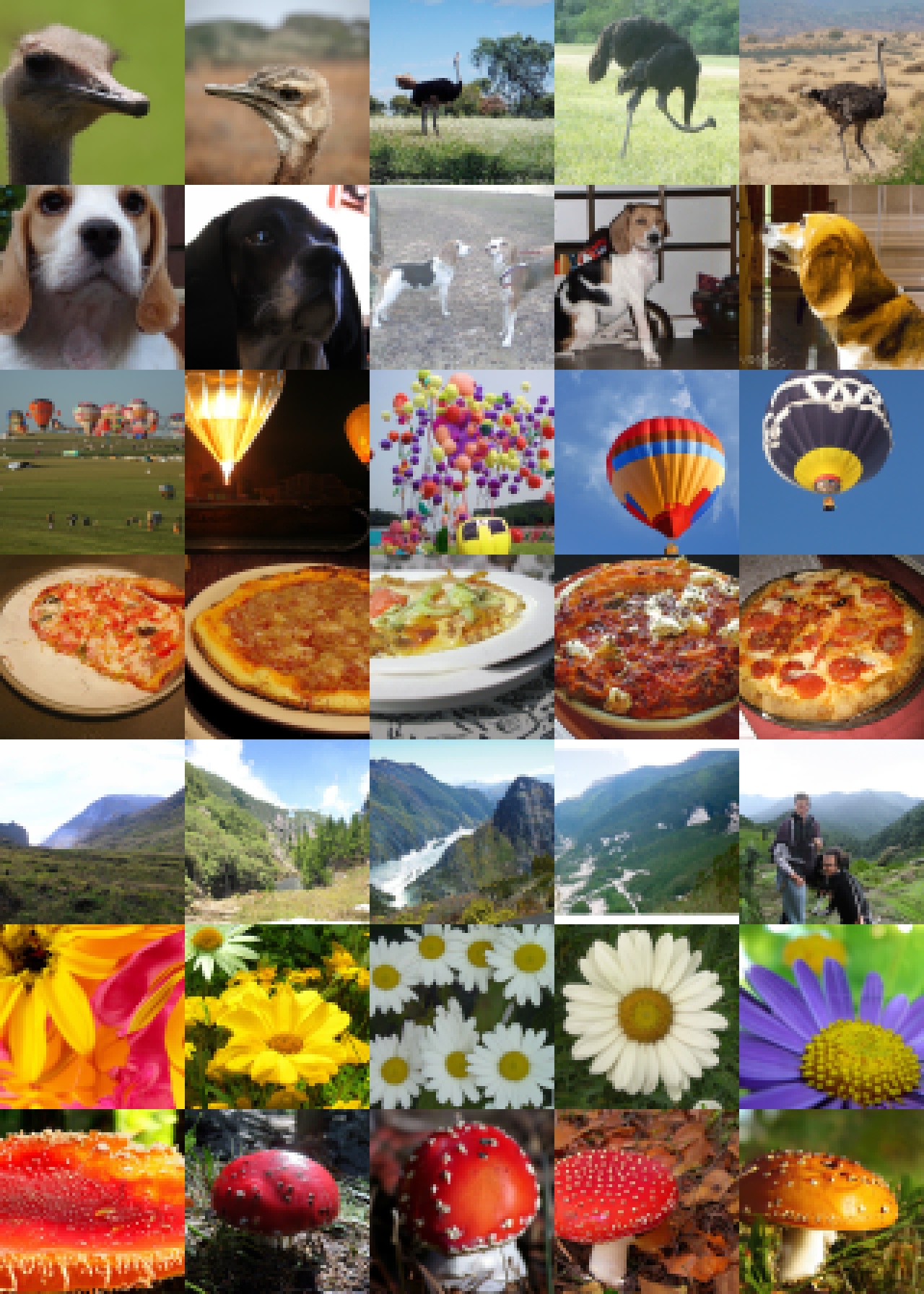}\\[-0.9mm]%
\hspace{\hhh}\makebox[\hh]{\footnotesize FID 2.01~~~~NFE 512}\hfill%
\hspace{\hhh}\makebox[\hh]{\footnotesize FID {\bf 1.55}~~~~NFE 1023}\\[3mm]%
\caption{\label{fig:GridsImgcSampling}%
Results for different samplers on class-conditional ImageNet~\cite{Deng2009imagenet} at 64$\times$64 resolution, using the pre-trained model by Dhariwal~and~Nichol~\cite{Dhariwal2021}.
The cases correspond to dots in Figures~\refpaper{fig:OdePlotNfe}c and~\refpaper{fig:SdePlotNfe}c.
}%
\end{figure}
}

\newcommand{\figGridsImgcTraining}{%
\renewcommand{\hh}{0.956\linewidth}
\renewcommand{\vv}{\hh/\real{10}*\real{7}}%
\renewcommand{\hhh}{2.5mm}
\begin{figure}[p]%
\centering%
\hspace{\hhh}\makebox[\hh]{{Deterministic, Our sampler \& training configuration}}\\%
\imgclasses\includegraphics[width=\hh]{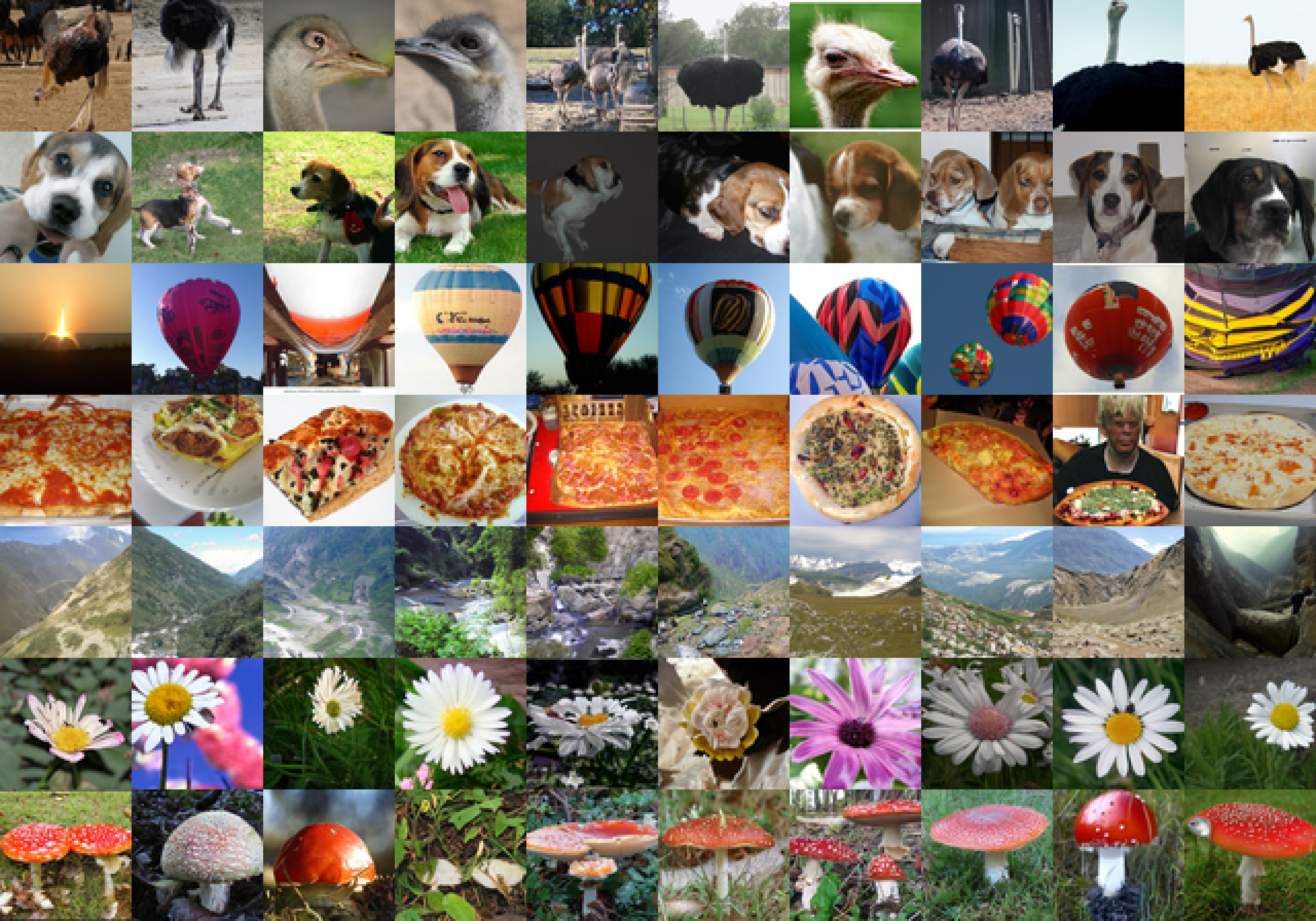}\\[-0.9mm]%
\hspace{\hhh}\makebox[\hh]{\footnotesize{FID 2.23~~~~NFE {\bf 79}}}\\[3mm]%
\hspace{\hhh}\makebox[\hh]{{Stochastic, Our sampler \& training configuration}}\\%
\imgclasses\includegraphics[width=\hh]{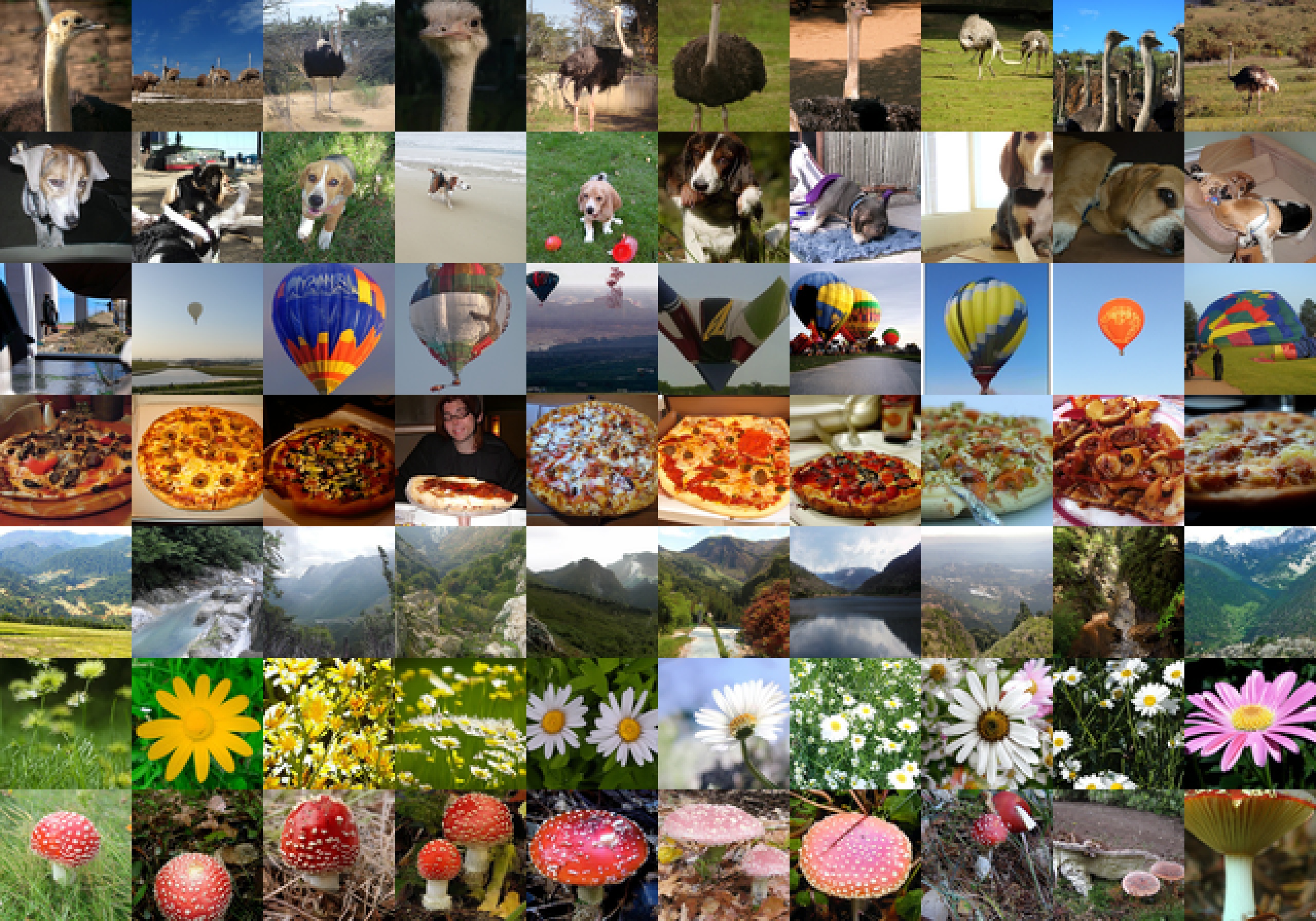}\\[-0.9mm]%
\hspace{\hhh}\makebox[\hh]{\footnotesize{FID {\bf 1.36}~~~~NFE 511}}\\[3mm]%
\caption{\label{fig:GridsImgcTraining}%
{Results for our training configuration on class-conditional ImageNet~\cite{Deng2009imagenet} at 64$\times$64 resolution, using our deterministic and stochastic samplers.}
}%
\end{figure}
}

\newcommand{\figGridsCifaruSampling}{%
\renewcommand{\hh}{0.478\linewidth}
\renewcommand{\hhh}{2.5mm}
\begin{figure}[p]%
\centering%
\hspace{\hhh}\makebox[\hh]{Deterministic, Original sampler (p.flow), VP}\hfill%
\hspace{\hhh}\makebox[\hh]{Deterministic, Original sampler (p.flow), VE}\\%
\hspace{\hhh}\includegraphics[width=\hh]{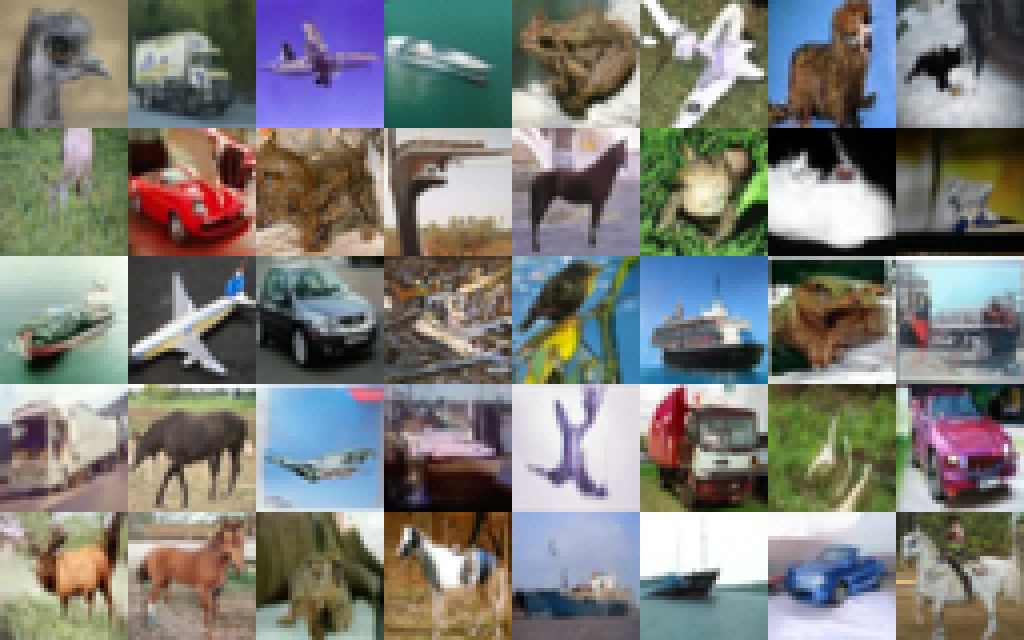}\hfill%
\hspace{\hhh}\includegraphics[width=\hh]{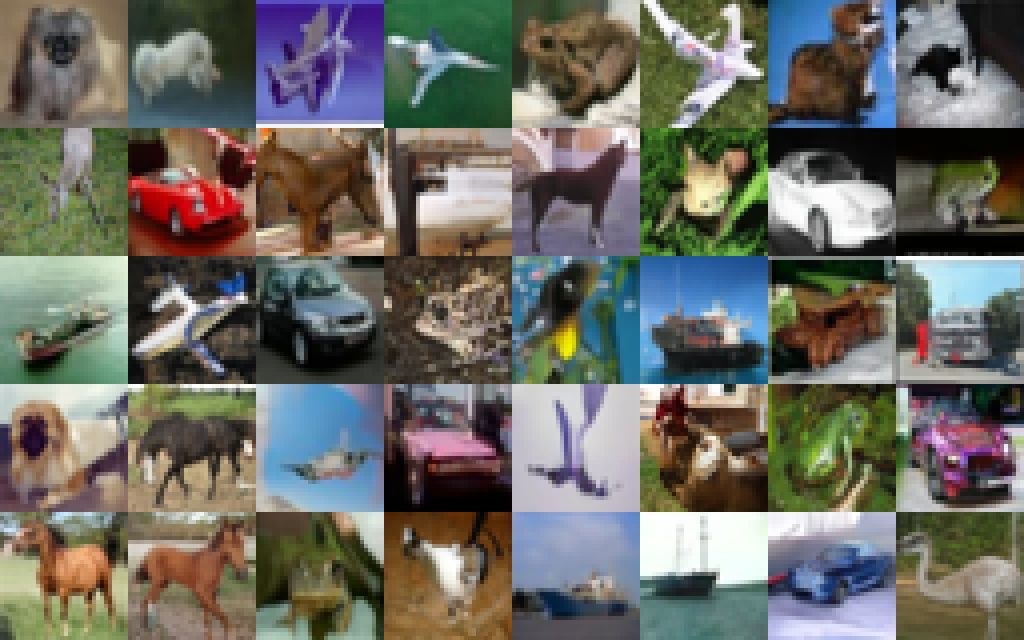}\\[-0.9mm]%
\hspace{\hhh}\makebox[\hh]{\footnotesize FID 2.94~~~~NFE 256}\hfill%
\hspace{\hhh}\makebox[\hh]{\footnotesize FID 5.45~~~~NFE 8192}\\[3mm]%
\hspace{\hhh}\makebox[\hh]{Deterministic, Our sampler (Alg.~\refpaper{alg:heun}), VP}\hfill%
\hspace{\hhh}\makebox[\hh]{Deterministic, Our sampler (Alg.~\refpaper{alg:heun}), VE}\\%
\hspace{\hhh}\includegraphics[width=\hh]{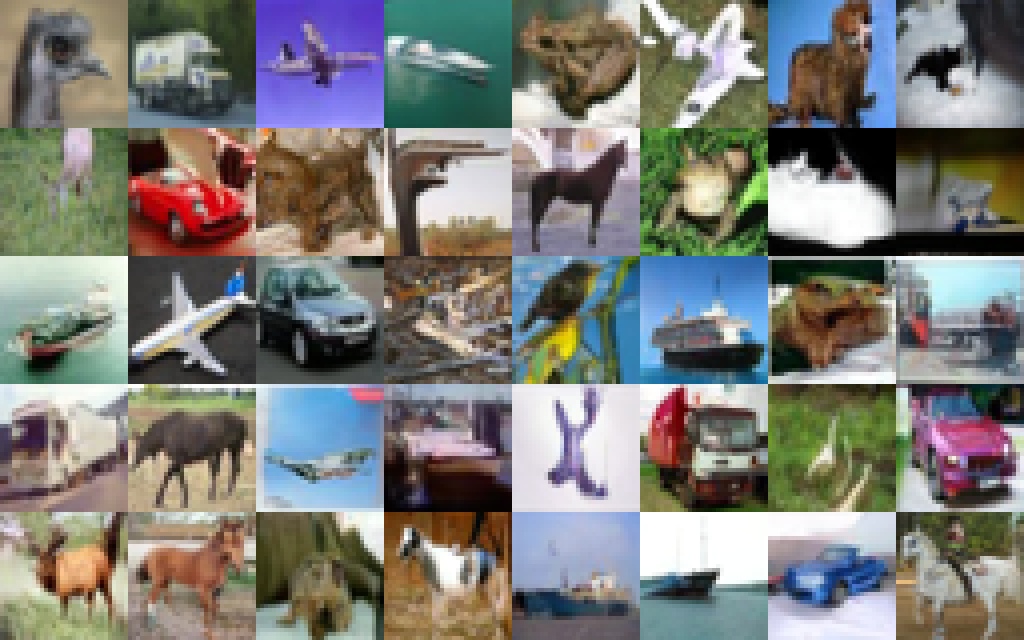}\hfill%
\hspace{\hhh}\includegraphics[width=\hh]{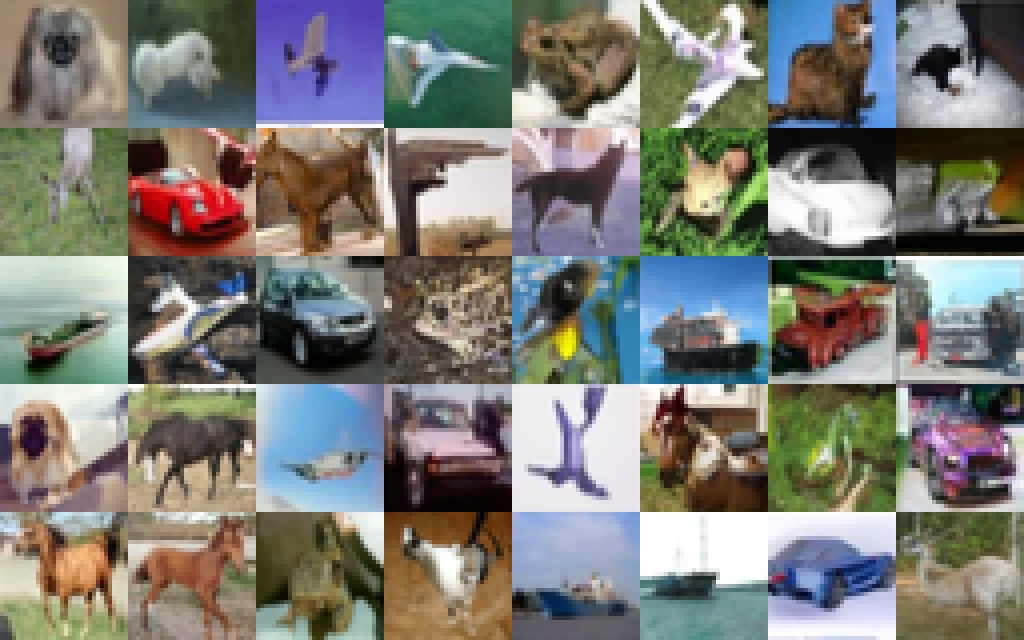}\\[-0.9mm]%
\hspace{\hhh}\makebox[\hh]{\footnotesize FID 3.01~~~~NFE 35}\hfill%
\hspace{\hhh}\makebox[\hh]{\footnotesize FID 3.82~~~~NFE {\bf 27}}\\[3mm]%
\hspace{\hhh}\makebox[\hh]{Stochastic, Original sampler (E--M), VP}\hfill%
\hspace{\hhh}\makebox[\hh]{Stochastic, Original sampler (P--C), VE}\\%
\hspace{\hhh}\includegraphics[width=\hh]{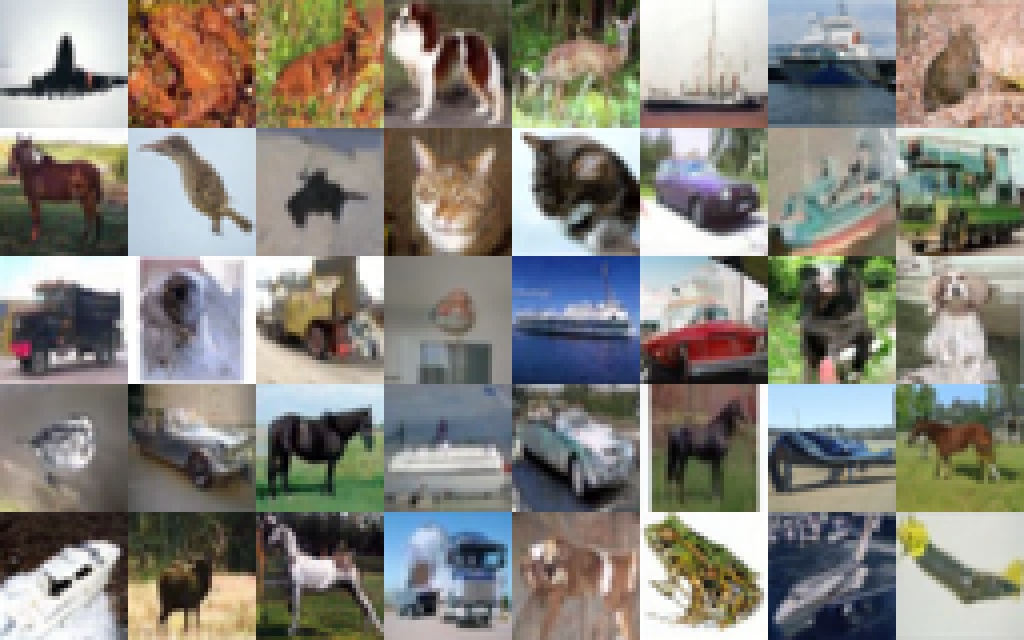}\hfill%
\hspace{\hhh}\includegraphics[width=\hh]{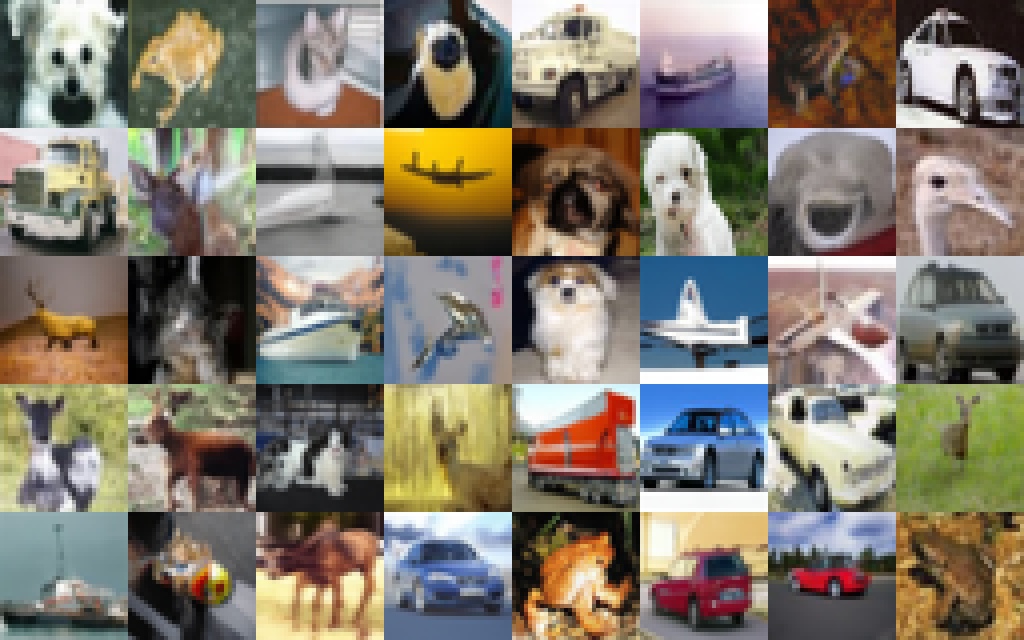}\\[-0.9mm]%
\hspace{\hhh}\makebox[\hh]{\footnotesize FID 2.55~~~~NFE 1024}\hfill%
\hspace{\hhh}\makebox[\hh]{\footnotesize FID 2.46~~~~NFE 2048}\\[3mm]%
\hspace{\hhh}\makebox[\hh]{Stochastic, Our sampler (Alg.~\refpaper{alg:stochastic}), VP}\hfill%
\hspace{\hhh}\makebox[\hh]{Stochastic, Our sampler (Alg.~\refpaper{alg:stochastic}), VE}\\%
\hspace{\hhh}\includegraphics[width=\hh]{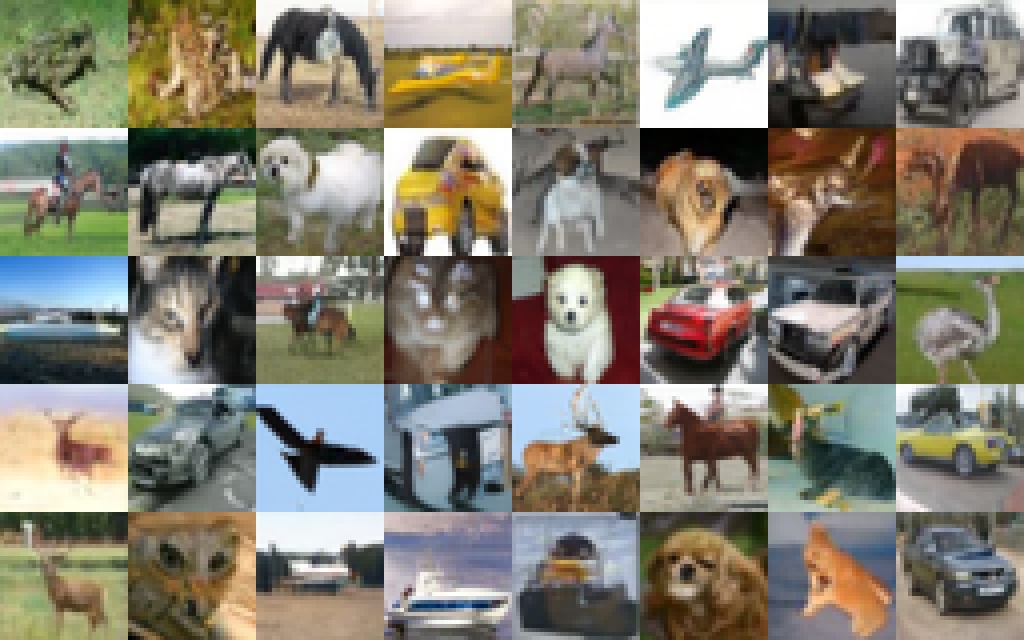}\hfill%
\hspace{\hhh}\includegraphics[width=\hh]{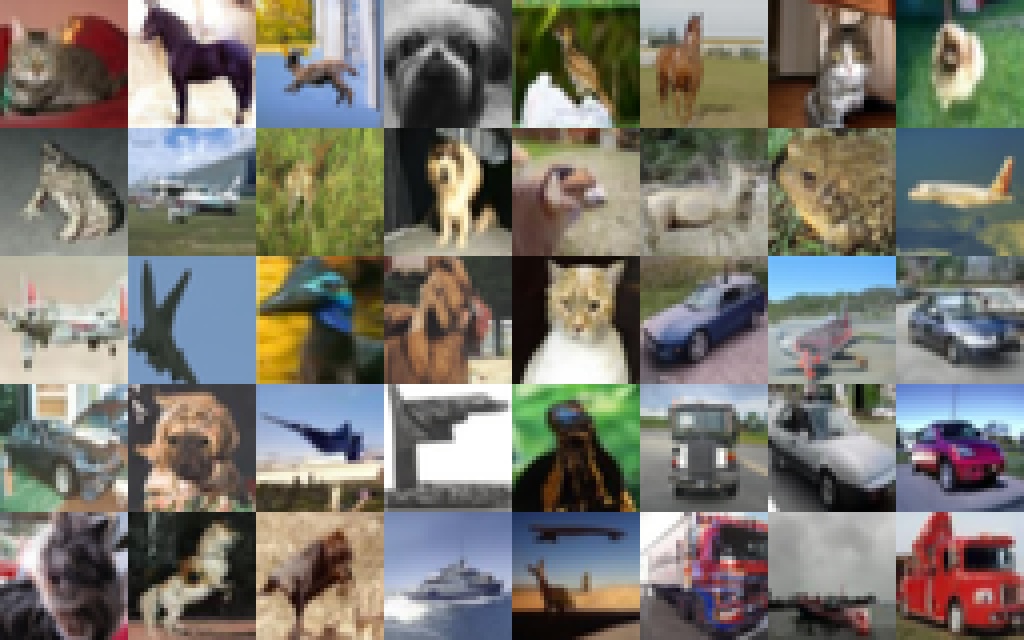}\\[-0.9mm]%
\hspace{\hhh}\makebox[\hh]{\footnotesize FID 2.27~~~~NFE 511}\hfill%
\hspace{\hhh}\makebox[\hh]{\footnotesize FID {\bf 2.23}~~~~NFE 2047}\\[3mm]%
\caption{\label{fig:GridsCifaruSampling}%
Results for different samplers on unconditional CIFAR-10~\cite{Krizhevsky2009cifar} at 32$\times$32 resolution, using the pre-trained models by Song~et~al.~\cite{Song2021sde}.
The cases correspond to dots in Figures~\refpaper{fig:OdePlotNfe}a,b and~\refpaper{fig:SdePlotNfe}a,b.
}%
\end{figure}
}

\newcommand{\figGridsCifaruTraining}{%
\renewcommand{\hh}{0.478\linewidth}
\renewcommand{\hhh}{2.5mm}
\begin{figure}[p]%
\centering%
\hspace{\hhh}\makebox[\hh]{Original training (config \textsc{a}), VP}\hfill%
\hspace{\hhh}\makebox[\hh]{Original training (config \textsc{a}), VE}\\%
\hspace{\hhh}\includegraphics[width=\hh]{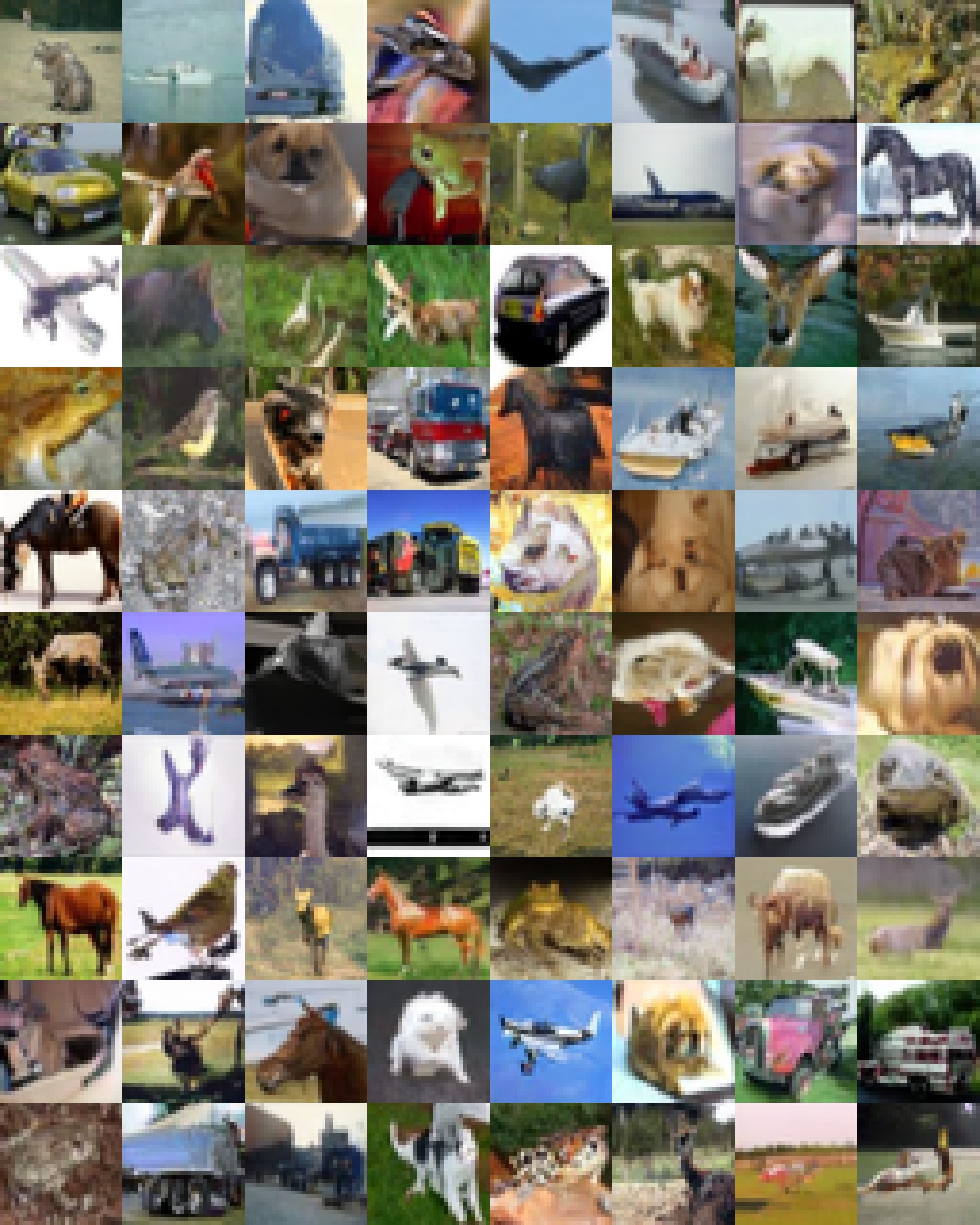}\hfill%
\hspace{\hhh}\includegraphics[width=\hh]{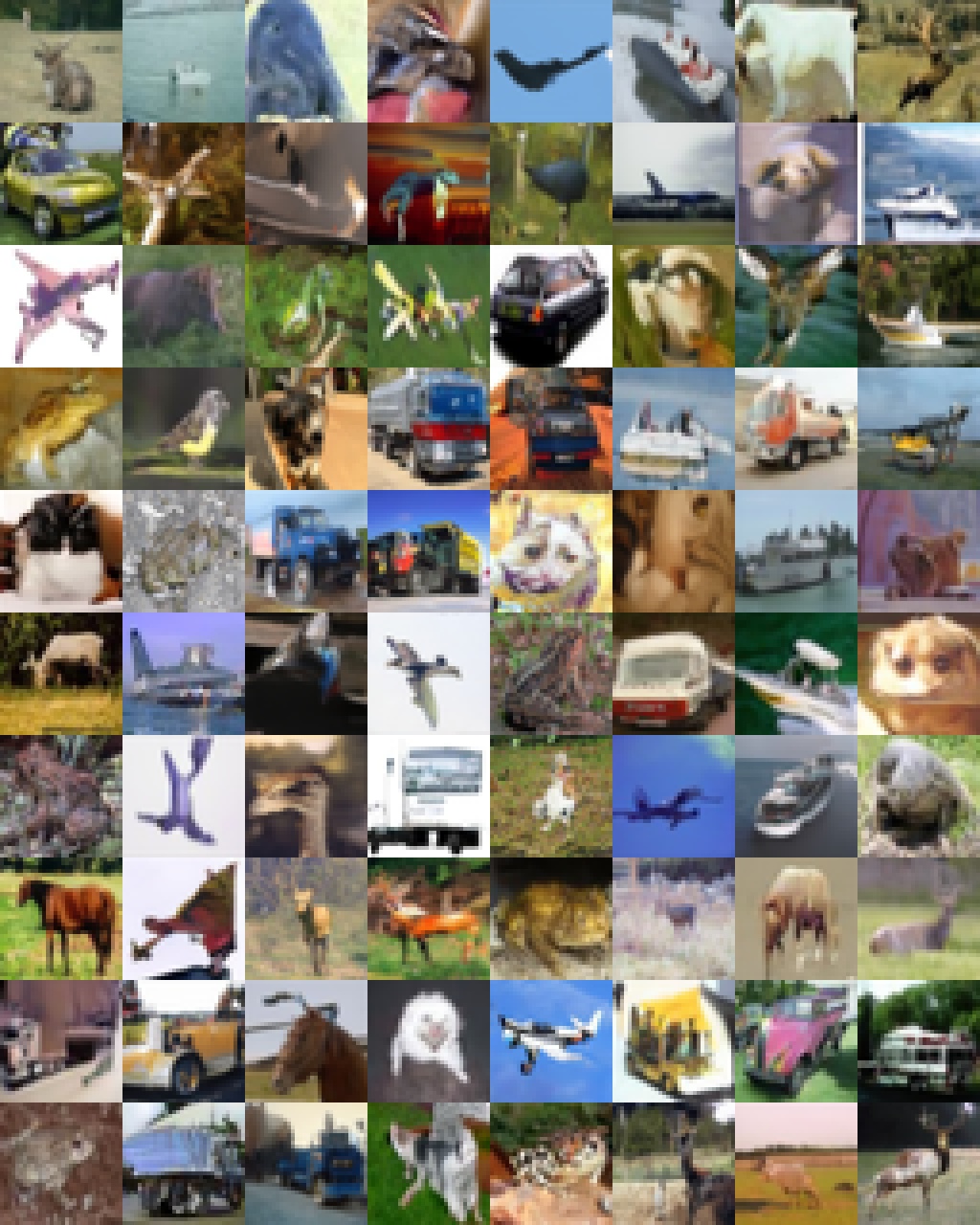}\\[-0.9mm]%
\hspace{\hhh}\makebox[\hh]{\footnotesize FID 3.01~~~~NFE 35}\hfill%
\hspace{\hhh}\makebox[\hh]{\footnotesize FID 3.77~~~~NFE 35}\\[3mm]%
\hspace{\hhh}\makebox[\hh]{Our training (config \textsc{f}), VP}\hfill%
\hspace{\hhh}\makebox[\hh]{Our training (config \textsc{f}), VE}\\%
\hspace{\hhh}\includegraphics[width=\hh]{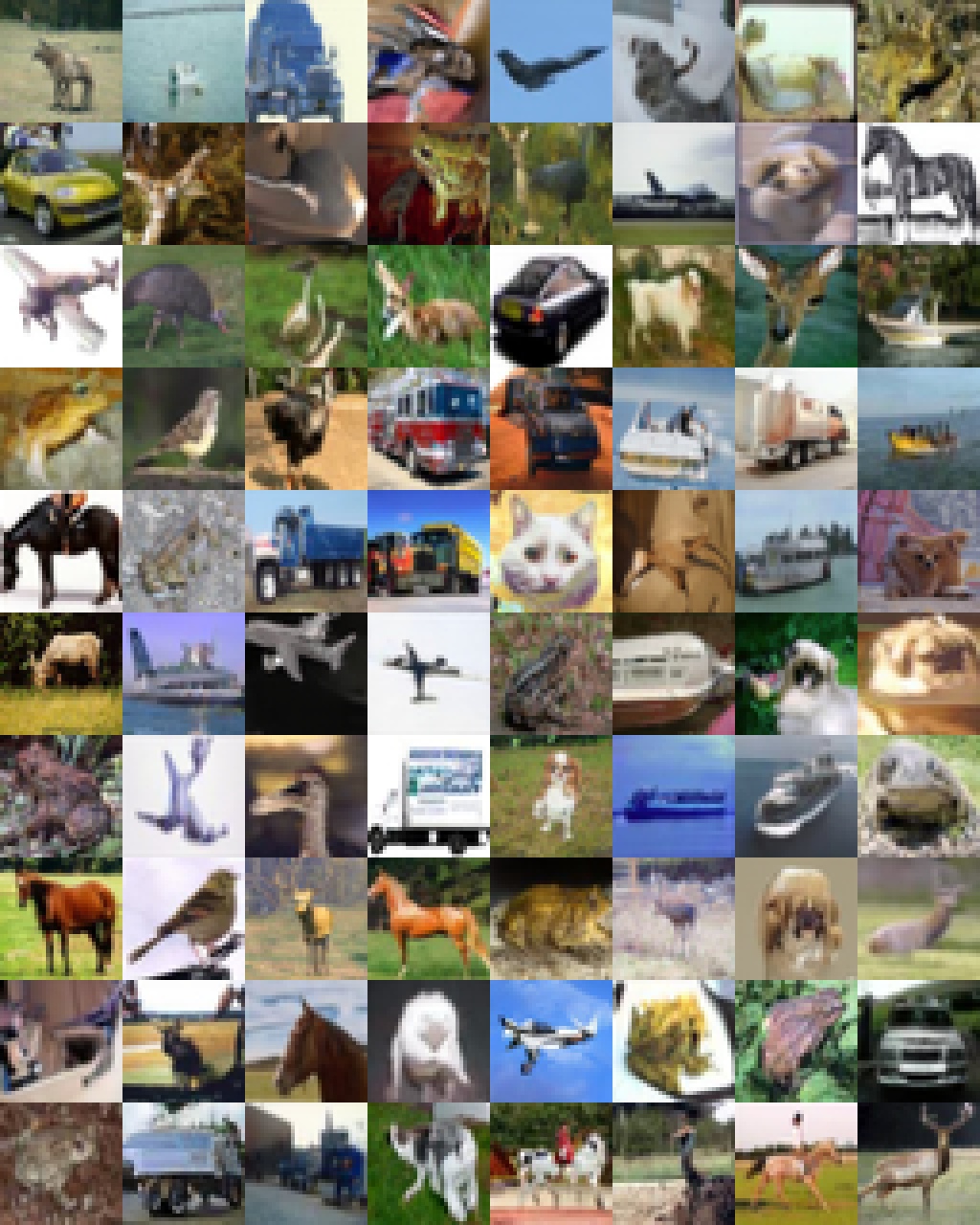}\hfill%
\hspace{\hhh}\includegraphics[width=\hh]{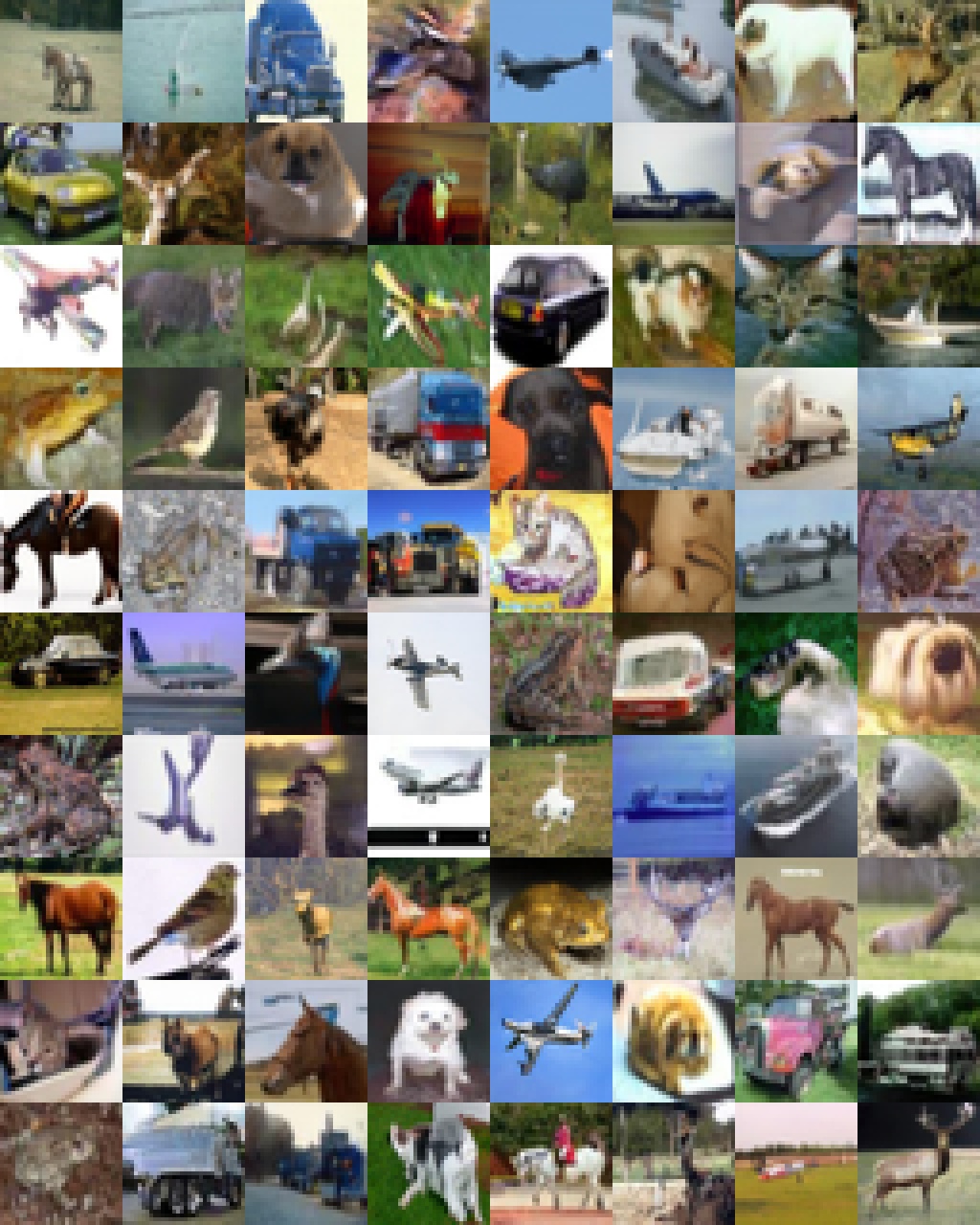}\\[-0.9mm]%
\hspace{\hhh}\makebox[\hh]{\footnotesize FID {\bf 1.97}~~~~NFE 35}\hfill%
\hspace{\hhh}\makebox[\hh]{\footnotesize FID 1.98~~~~NFE 35}\\[3mm]%
\caption{\label{fig:GridsCifaruTraining}%
Results for different training configurations on unconditional CIFAR-10~\cite{Krizhevsky2009cifar} at 32$\times$32 resolution, using our deterministic sampler with the same set of latent codes ($\xx_0$) in each case.
}%
\end{figure}
}

\newcommand{\cifarclasses}{\vlabel{\hhh}{\vv}{\makebox[\vv]{\scriptsize\hrlabel{Truck}\hrlabel{Ship}\hrlabel{Horse}\hrlabel{Frog}\hrlabel{Dog}\hrlabel{Deer}\hrlabel{Cat}\hrlabel{Bird}\hrlabel{Car}\hrlabel{Plane}}}}

\newcommand{\figGridsCifarcTraining}{%
\renewcommand{\hh}{0.478\linewidth}
\renewcommand{\vv}{\hh/\real{8}*\real{10}}%
\renewcommand{\hhh}{2.5mm}
\begin{figure}[p]%
\centering%
\hspace{\hhh}\makebox[\hh]{Original training (config \textsc{a}), VP}\hfill%
\hspace{\hhh}\makebox[\hh]{Original training (config \textsc{a}), VE}\\%
\cifarclasses\includegraphics[width=\hh]{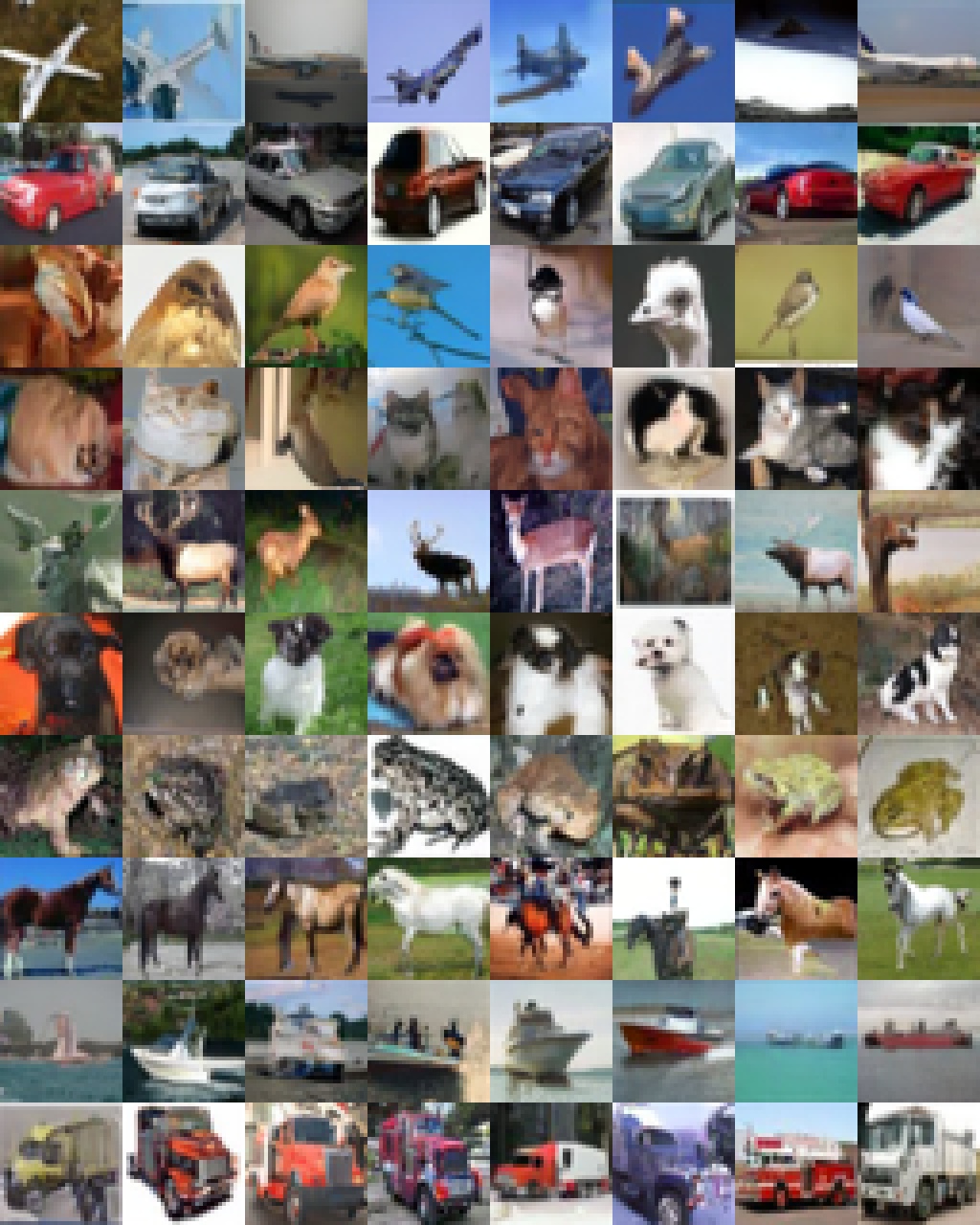}\hfill%
\cifarclasses\includegraphics[width=\hh]{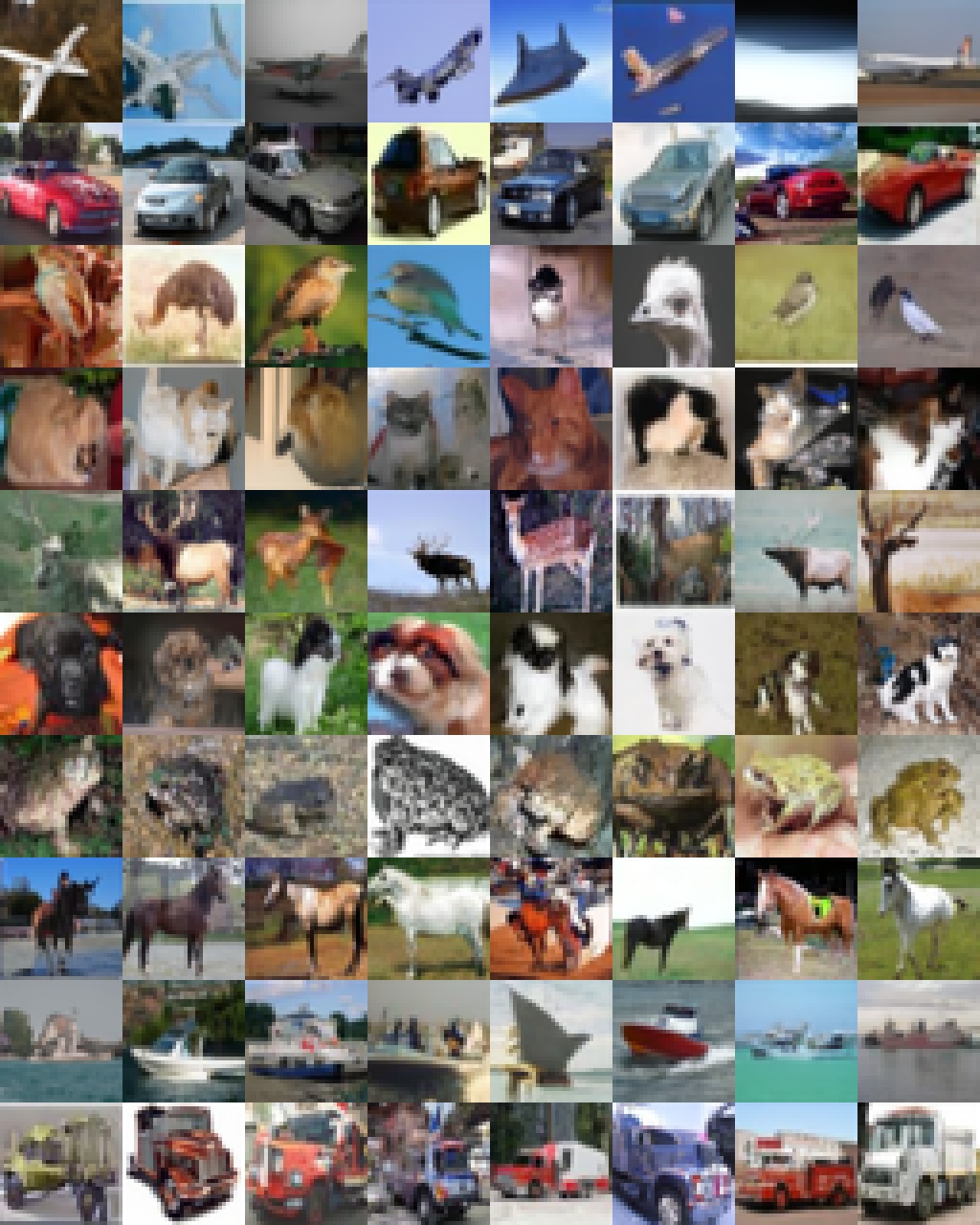}\\[-0.9mm]%
\hspace{\hhh}\makebox[\hh]{\footnotesize FID 2.48~~~~NFE 35}\hfill%
\hspace{\hhh}\makebox[\hh]{\footnotesize FID 3.11~~~~NFE 35}\\[3mm]%
\hspace{\hhh}\makebox[\hh]{Our training (config \textsc{f}), VP}\hfill%
\hspace{\hhh}\makebox[\hh]{Our training (config \textsc{f}), VE}\\%
\cifarclasses\includegraphics[width=\hh]{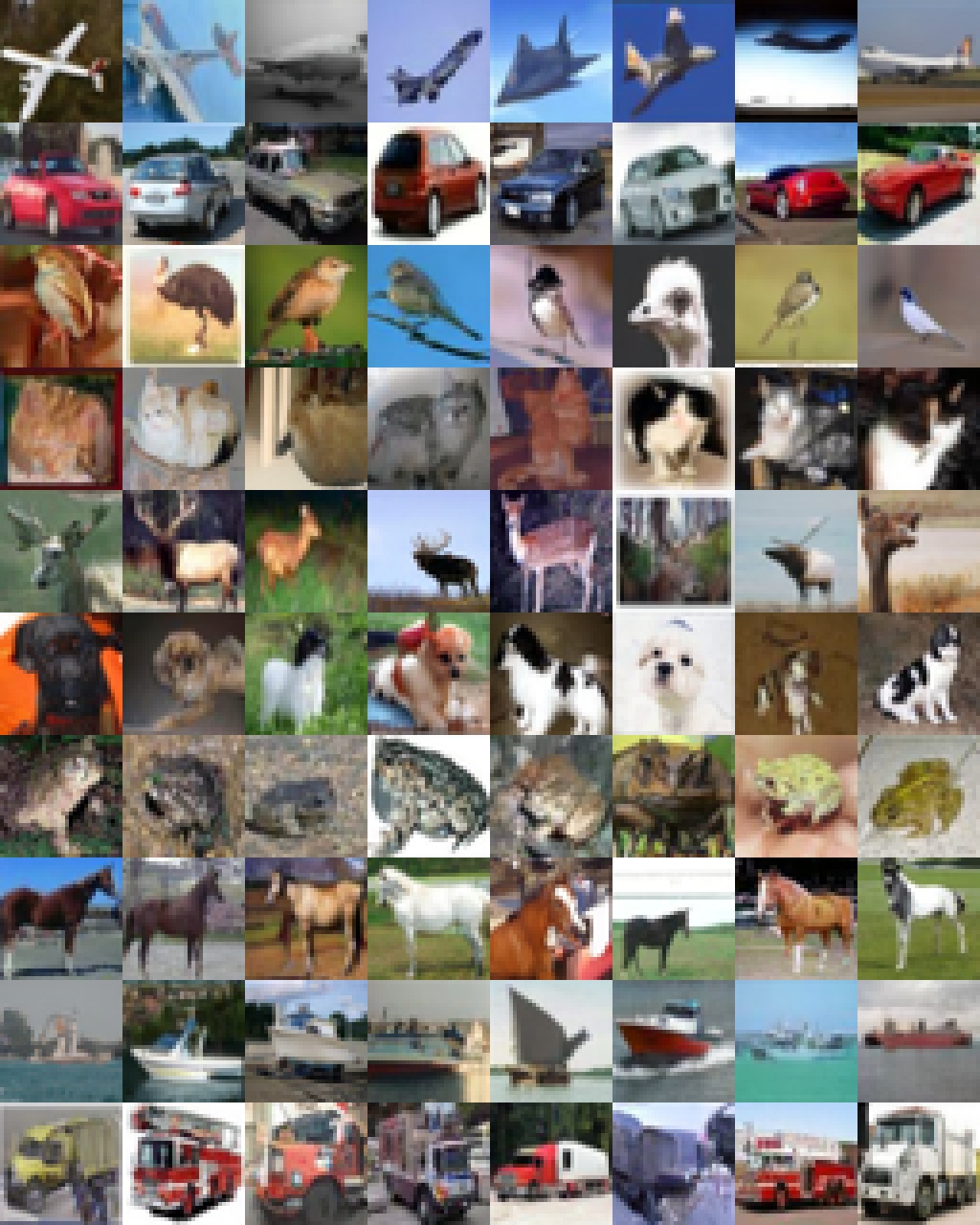}\hfill%
\cifarclasses\includegraphics[width=\hh]{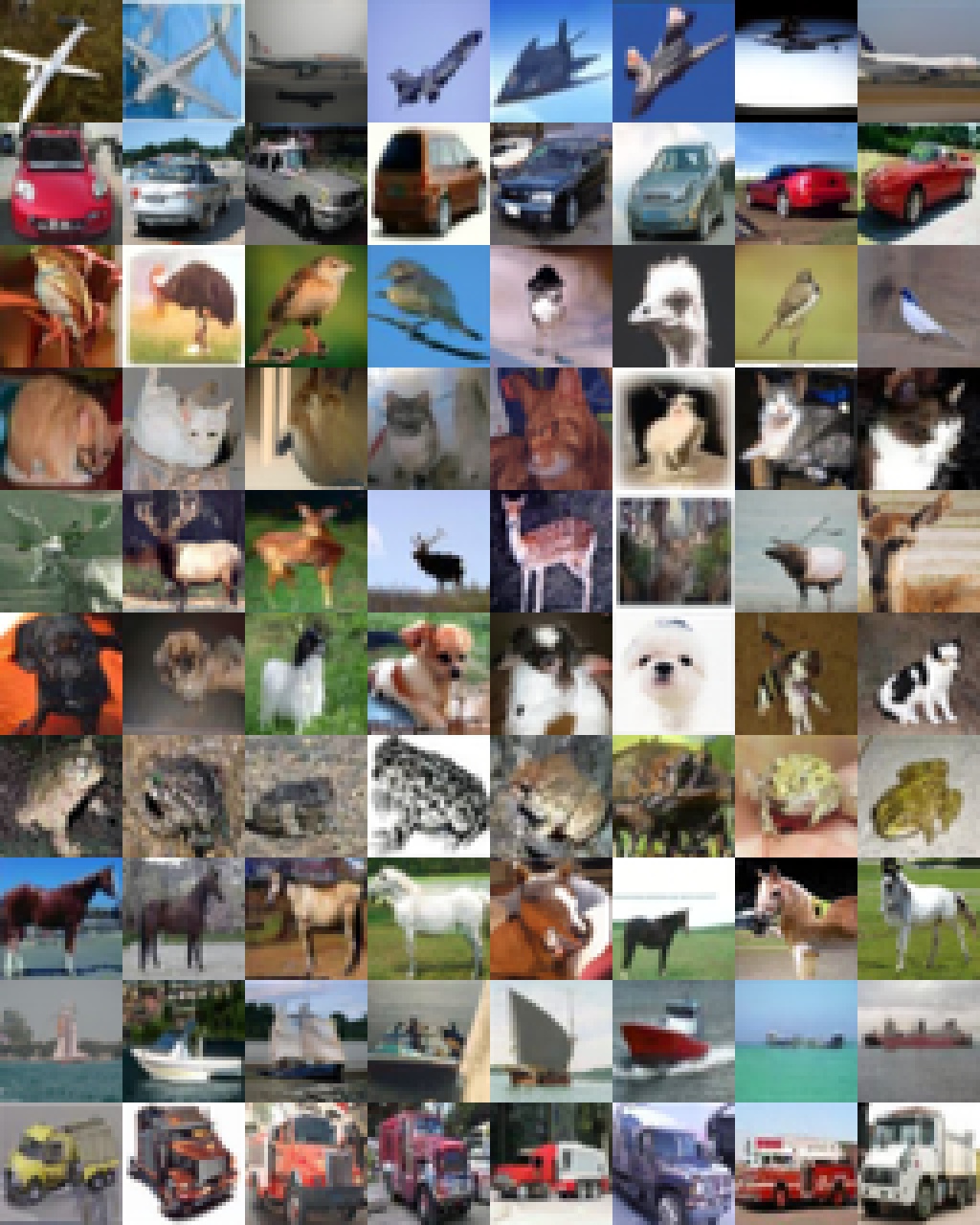}\\[-0.9mm]%
\hspace{\hhh}\makebox[\hh]{\footnotesize FID {\bf 1.79}~~~~NFE 35}\hfill%
\hspace{\hhh}\makebox[\hh]{\footnotesize FID {\bf 1.79}~~~~NFE 35}\\[3mm]%
\caption{\label{fig:GridsCifarcTraining}%
Results for different training configurations on class-conditional CIFAR-10~\cite{Krizhevsky2009cifar} at 32$\times$32 resolution, using our deterministic sampler with the same set of latent codes ($\xx_0$) in each case.
}%
\end{figure}
}

\newcommand{\figGridsFfhqAfhqTraining}{%
\renewcommand{\hh}{0.478\linewidth}
\renewcommand{\hhh}{2.5mm}
\begin{figure}[p]%
\centering%
\hspace{\hhh}\makebox[\hh]{FFHQ, Original training (config \textsc{a}), VP}\hfill%
\hspace{\hhh}\makebox[\hh]{FFHQ, Original training (config \textsc{a}), VE}\\%
\hspace{\hhh}\includegraphics[width=\hh]{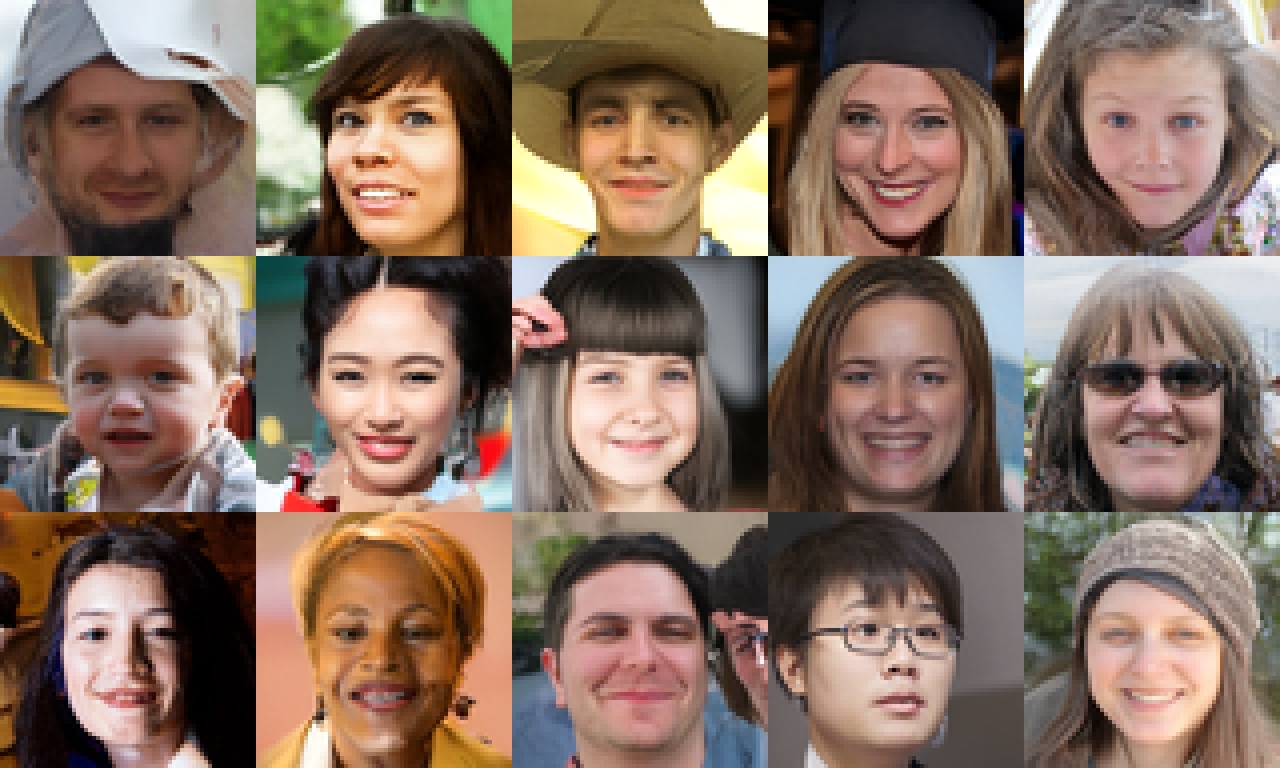}\hfill%
\hspace{\hhh}\includegraphics[width=\hh]{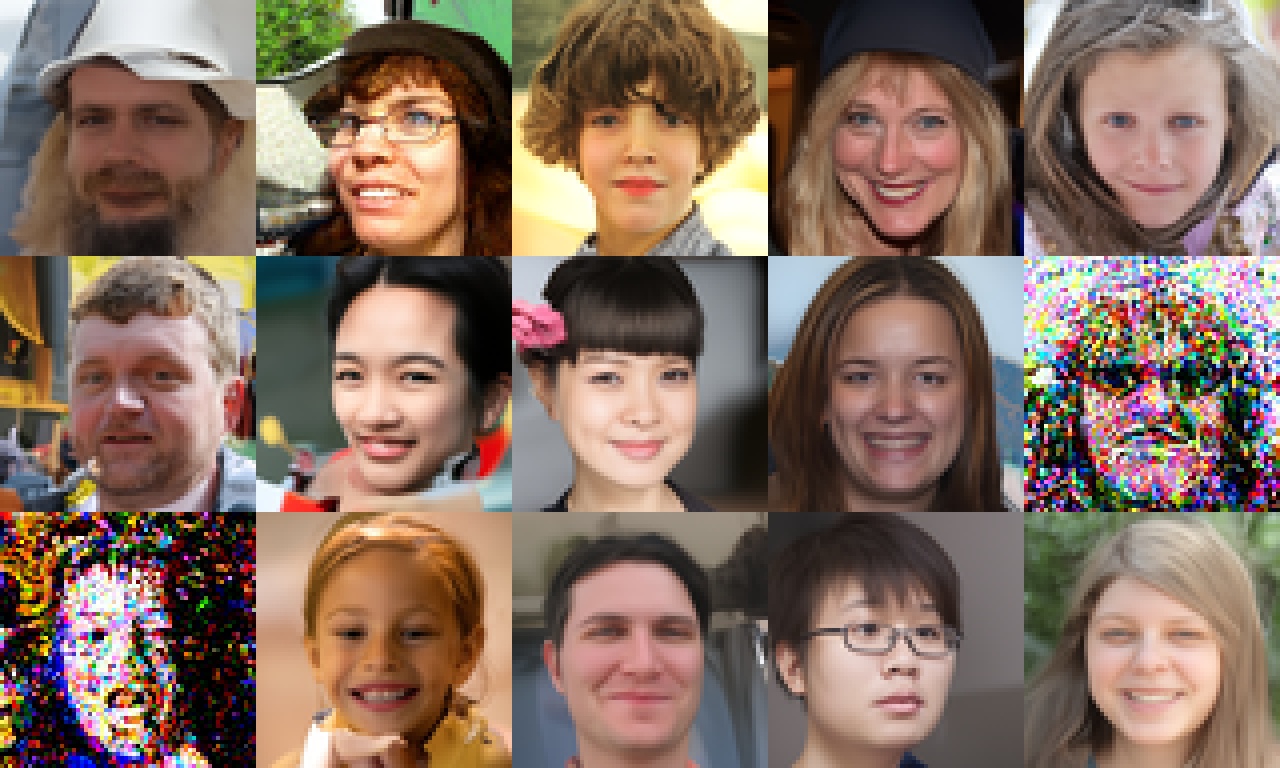}\\[-0.9mm]%
\hspace{\hhh}\makebox[\hh]{\footnotesize FID 3.39~~~~NFE 79}\hfill%
\hspace{\hhh}\makebox[\hh]{\footnotesize FID 25.95~~~~NFE 79}\\[3mm]%
\hspace{\hhh}\makebox[\hh]{FFHQ, Our training (config \textsc{f}), VP}\hfill%
\hspace{\hhh}\makebox[\hh]{FFHQ, Our training (config \textsc{f}), VE}\\%
\hspace{\hhh}\includegraphics[width=\hh]{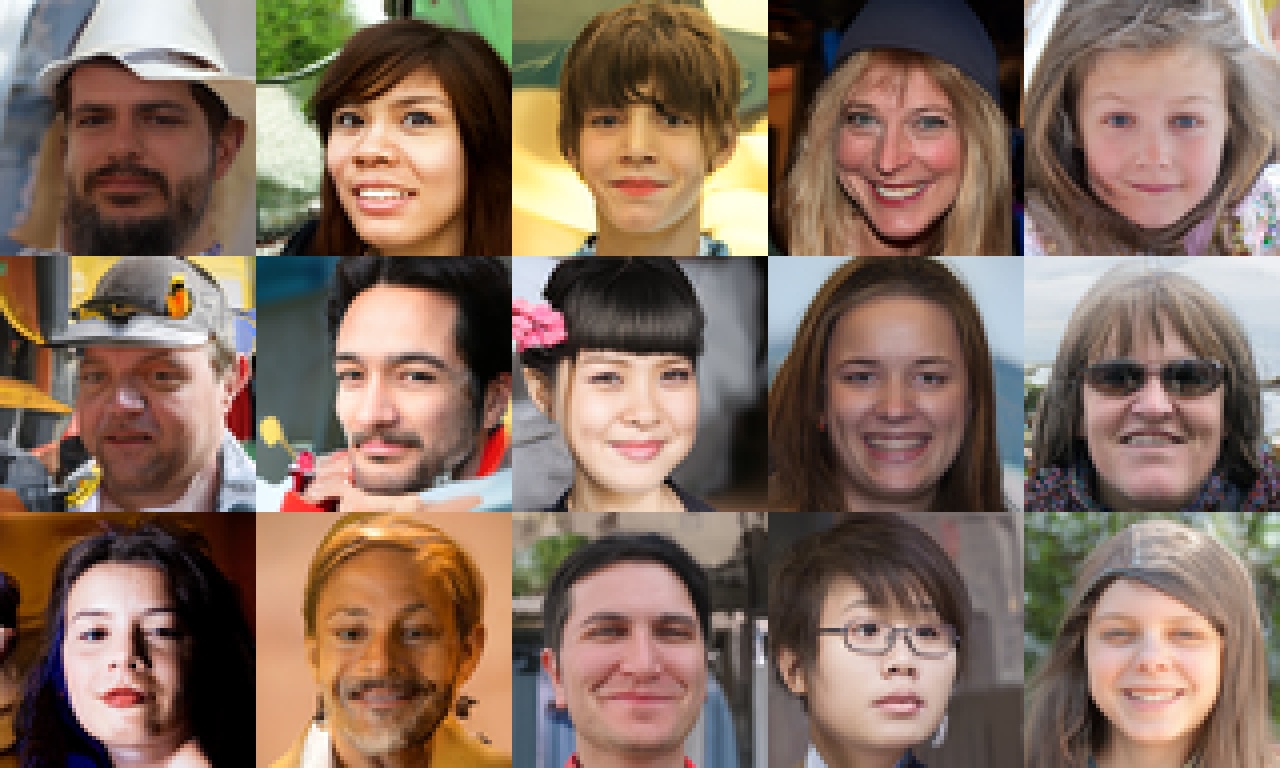}\hfill%
\hspace{\hhh}\includegraphics[width=\hh]{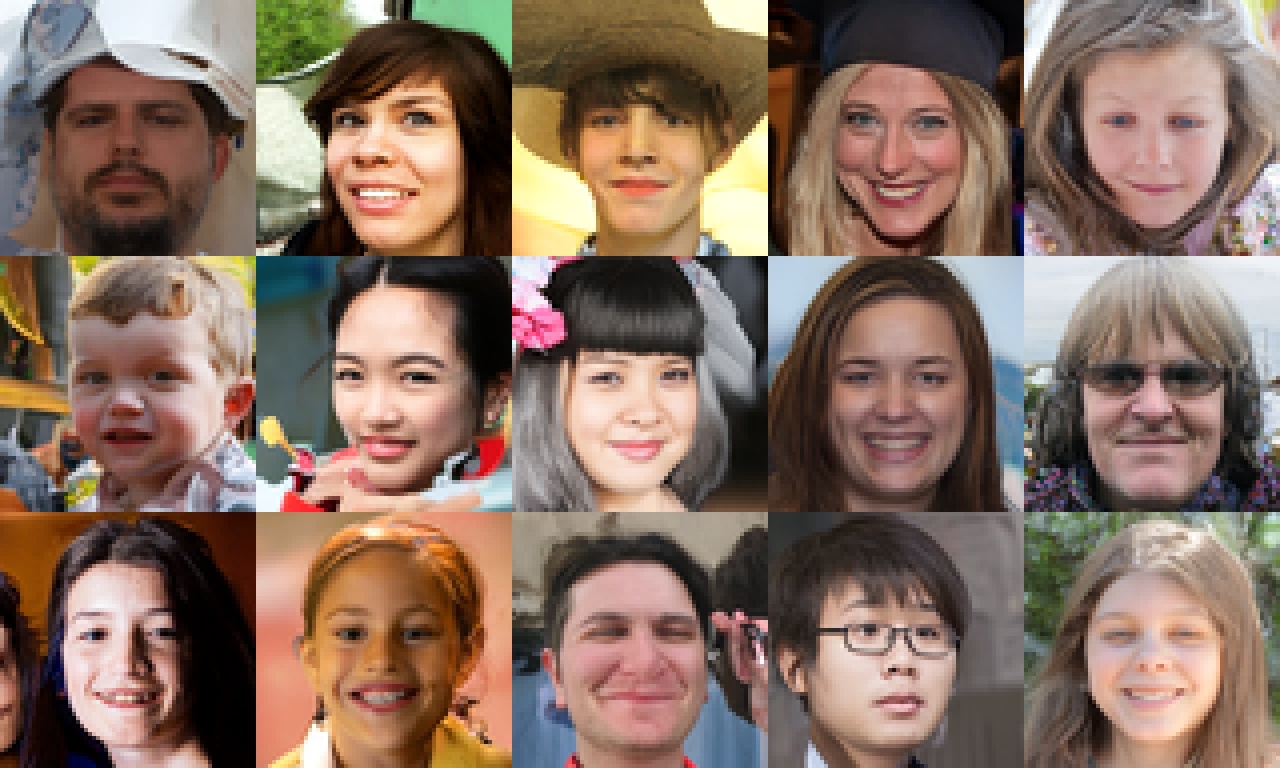}\\[-0.9mm]%
\hspace{\hhh}\makebox[\hh]{\footnotesize FID {\bf 2.39}~~~~NFE 79}\hfill%
\hspace{\hhh}\makebox[\hh]{\footnotesize FID 2.53~~~~NFE 79}\\[3mm]%
\hspace{\hhh}\makebox[\hh]{AFHQv2, Original training (config \textsc{a}), VP}\hfill%
\hspace{\hhh}\makebox[\hh]{AFHQv2, Original training (config \textsc{a}), VE}\\%
\hspace{\hhh}\includegraphics[width=\hh]{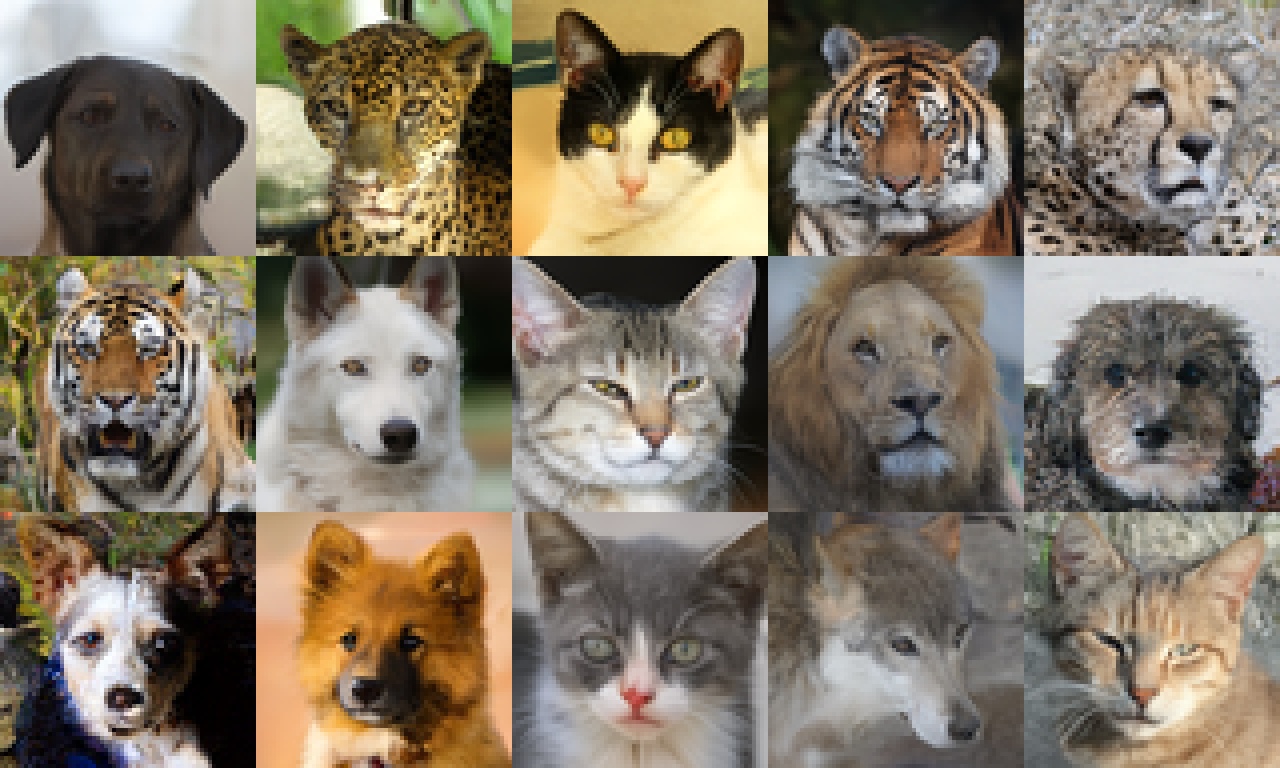}\hfill%
\hspace{\hhh}\includegraphics[width=\hh]{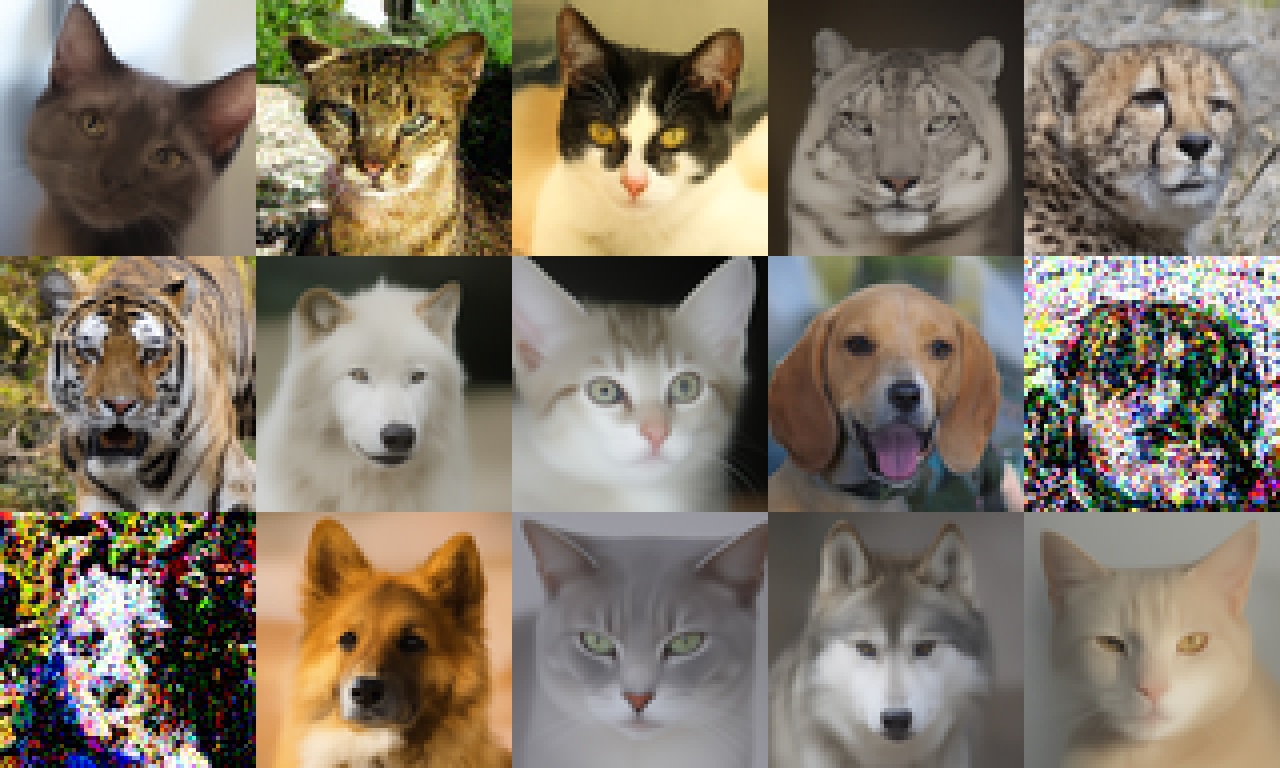}\\[-0.9mm]%
\hspace{\hhh}\makebox[\hh]{\footnotesize FID 2.58~~~~NFE 79}\hfill%
\hspace{\hhh}\makebox[\hh]{\footnotesize FID 18.52~~~~NFE 79}\\[3mm]%
\hspace{\hhh}\makebox[\hh]{AFHQv2, Our training (config \textsc{f}), VP}\hfill%
\hspace{\hhh}\makebox[\hh]{AFHQv2, Our training (config \textsc{f}), VE}\\%
\hspace{\hhh}\includegraphics[width=\hh]{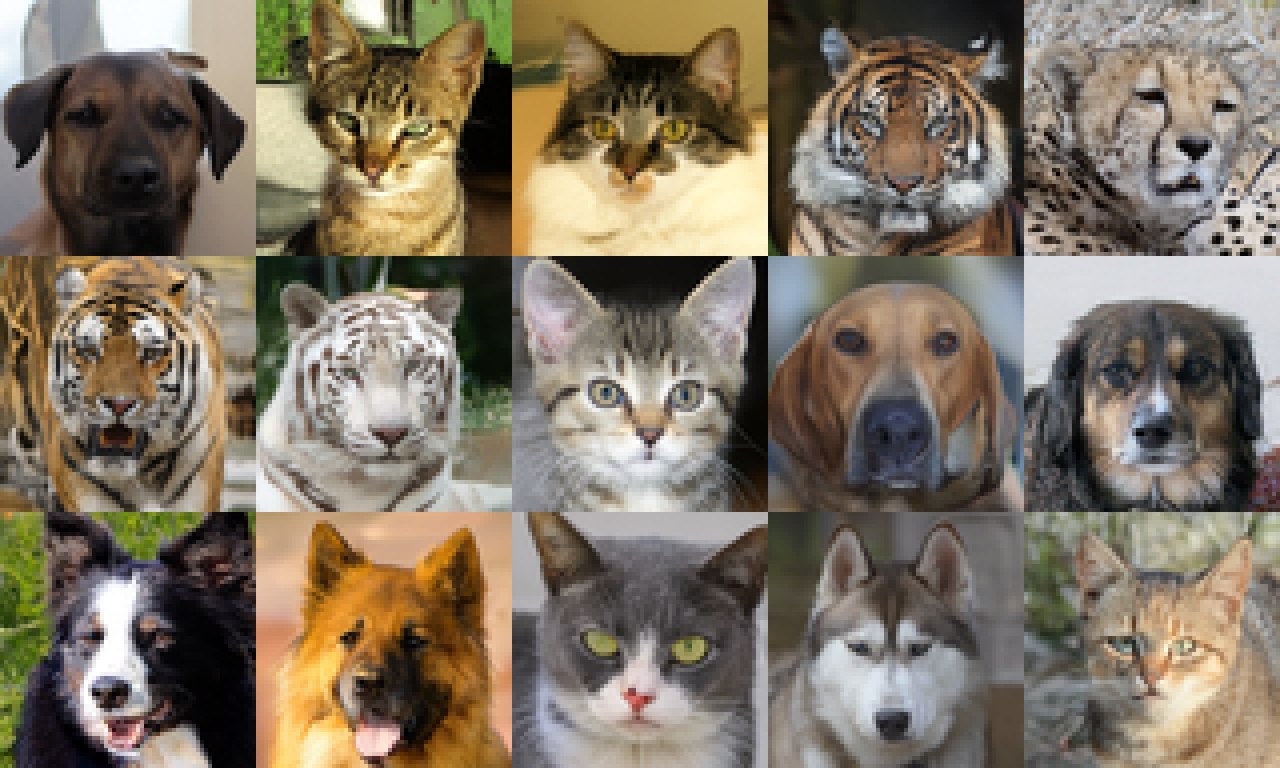}\hfill%
\hspace{\hhh}\includegraphics[width=\hh]{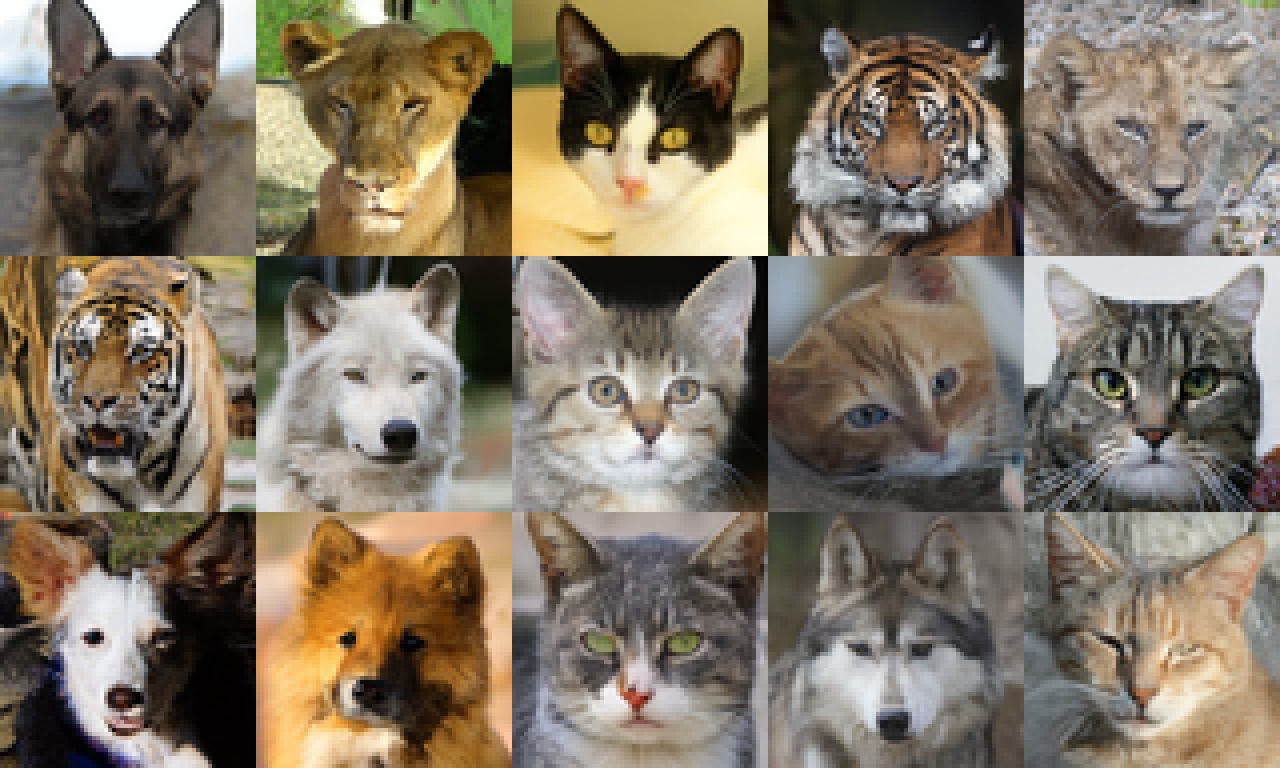}\\[-0.9mm]%
\hspace{\hhh}\makebox[\hh]{\footnotesize FID {\bf 1.96}~~~~NFE 79}\hfill%
\hspace{\hhh}\makebox[\hh]{\footnotesize FID 2.16~~~~NFE 79}\\[3mm]%
\caption{\label{fig:GridsFfhqAfhqTraining}%
Results for different training configurations on FFHQ~\cite{Karras2018stylegan} and AFHQv2~\cite{Choi2020afhq} at 64$\times$64 resolution, using our deterministic sampler with the same set of latent codes ($\xx_0$) in each case.
}%
\end{figure}
}

\newcommand{\figNfeSweep}{%
\renewcommand{\hh}{0.49\linewidth}
\begin{figure}[p]%
\centering%
\makebox[\hh]{Class-conditional ImageNet-64, Pre-trained}\hfill%
\makebox[\hh]{Class-conditional CIFAR-10, Our training, VP}\\%
\includegraphics[width=\hh]{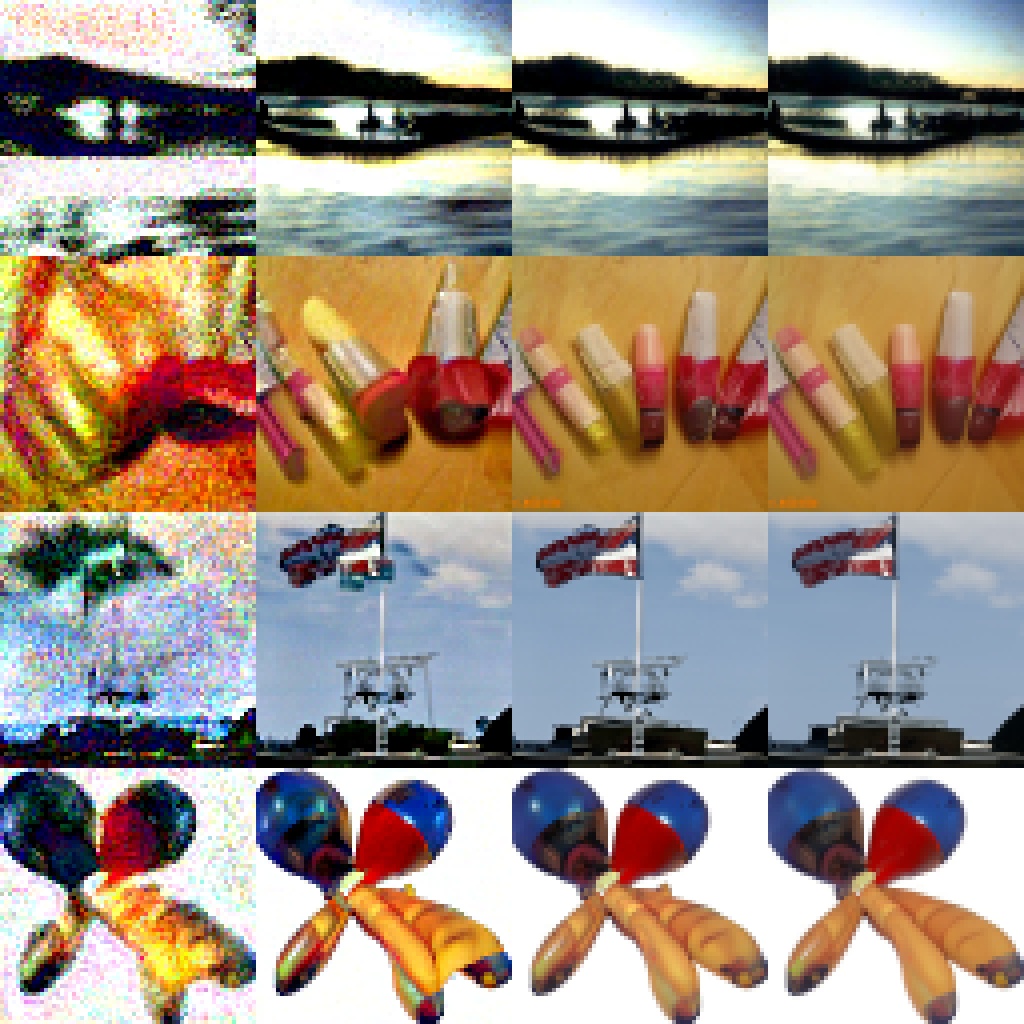}\hfill%
\includegraphics[width=\hh]{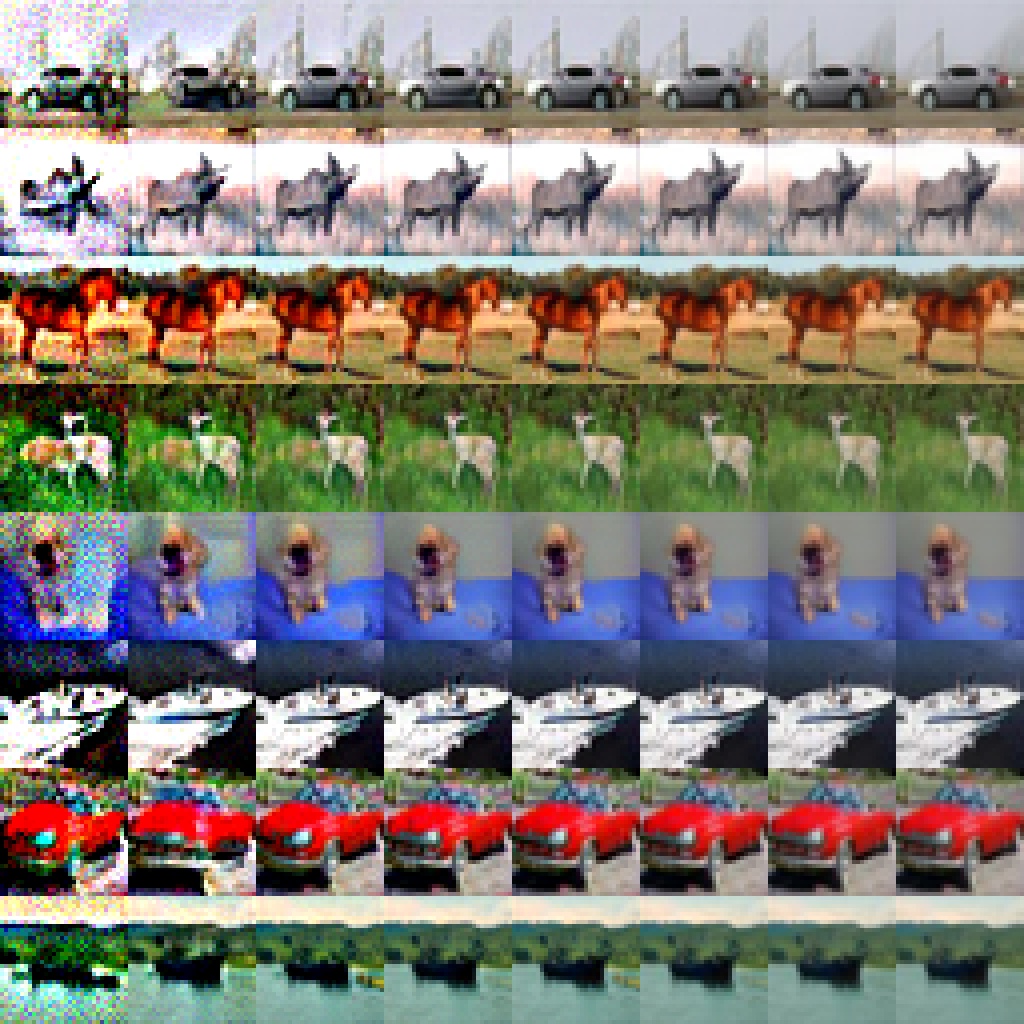}\\[-1.4mm]%
\makebox[\hh]{\scriptsize\makebox[0mm][l]{FID}\hrlabel{87.16}\hrlabel{12.39}\hrlabel{3.56}\hrlabel{\bf 2.66}}\hfill%
\makebox[\hh]{\scriptsize\hrlabel{85.46}\hrlabel{35.47}\hrlabel{14.32}\hrlabel{6.72}\hrlabel{4.22}\hrlabel{2.48}\hrlabel{1.86}\hrlabel{\bf 1.79}}\\[-1.2mm]%
\makebox[\hh]{\scriptsize\makebox[0mm][l]{NFE}\hrlabel{7}\hrlabel{11}\hrlabel{19}\hrlabel{79}}\hfill%
\makebox[\hh]{\scriptsize\hrlabel{7}\hrlabel{9}\hrlabel{11}\hrlabel{13}\hrlabel{15}\hrlabel{19}\hrlabel{27}\hrlabel{35}}\\[3mm]%
\makebox[\hh]{Unconditional FFHQ, Our training, VP}\hfill%
\makebox[\hh]{Unconditional AFHQv2, Our Training, VP}\\%
\includegraphics[width=\hh]{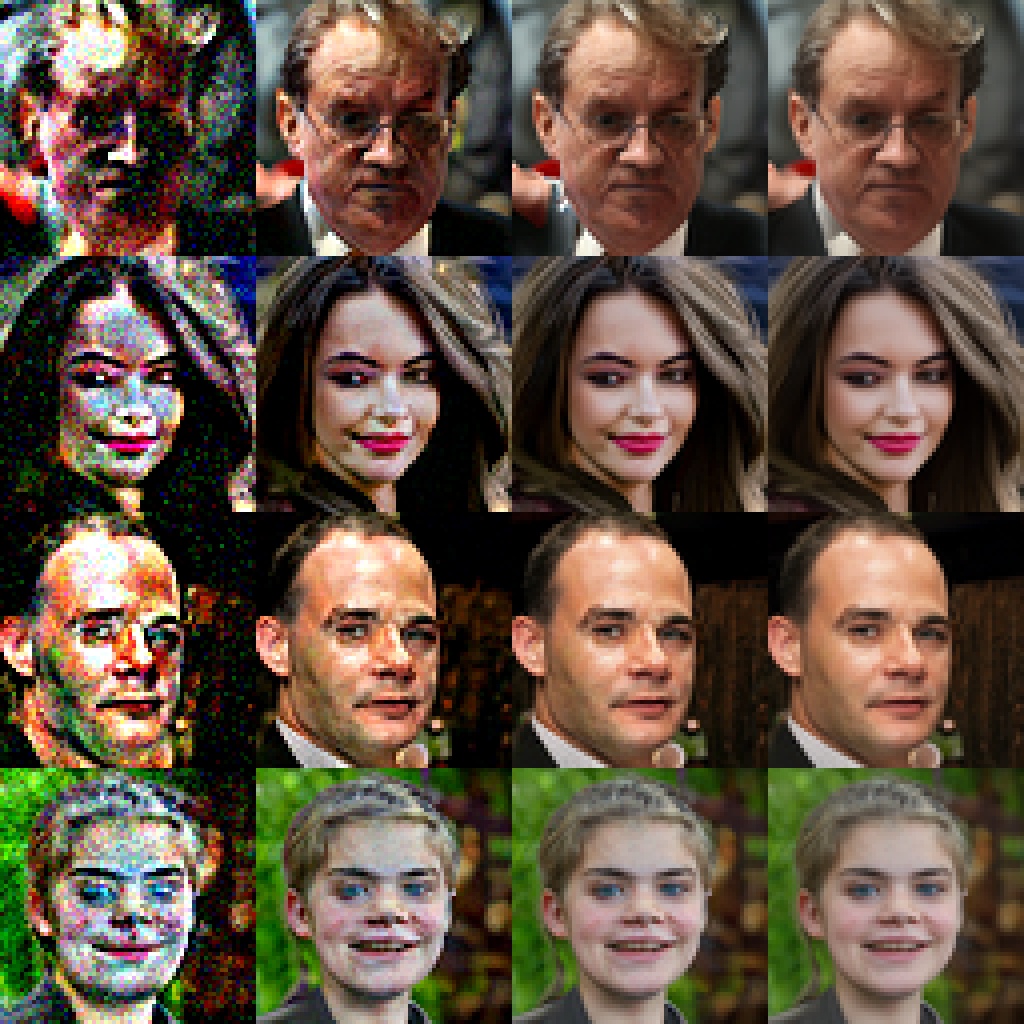}\hfill%
\includegraphics[width=\hh]{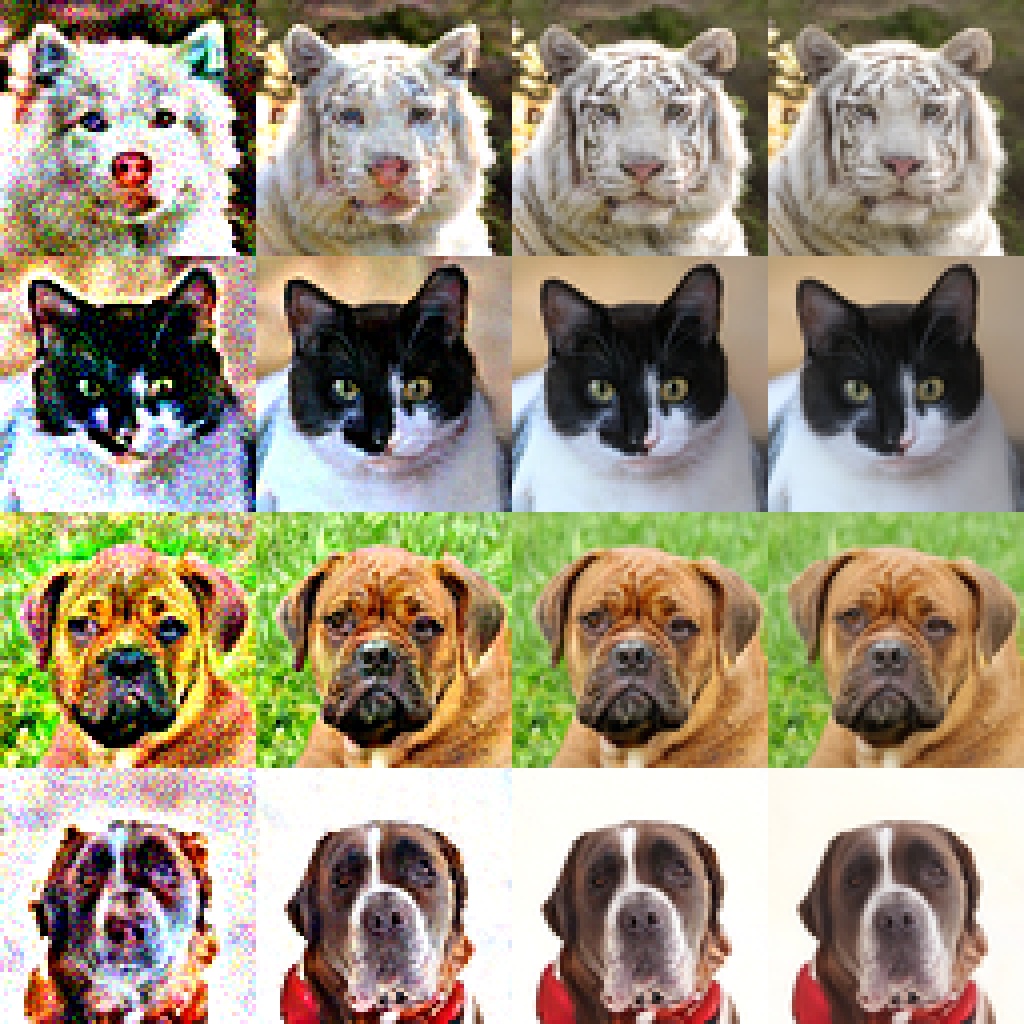}\\[-1.4mm]%
\makebox[\hh]{\scriptsize\makebox[0mm][l]{FID}\hrlabel{142.34}\hrlabel{29.22}\hrlabel{5.13}\hrlabel{\bf 2.39}}\hfill%
\makebox[\hh]{\scriptsize\hrlabel{61.57}\hrlabel{13.68}\hrlabel{3.00}\hrlabel{\bf 1.96}}\\[-1.2mm]%
\makebox[\hh]{\scriptsize\makebox[0mm][l]{NFE}\hrlabel{7}\hrlabel{11}\hrlabel{19}\hrlabel{79}}\hfill%
\makebox[\hh]{\scriptsize\hrlabel{7}\hrlabel{11}\hrlabel{19}\hrlabel{79}}\\[3mm]%
\caption{\label{fig:NfeSweep}%
Image quality and FID as a function of NFE using our deterministic sampler.
At 32$\times$32 resolution, reasonable image quality is reached around NFE $=$ 13, but FID keeps improving until NFE $=$ 35.
At 64$\times$64 resolution, reasonable image quality is reached around NFE $=$ 19, but FID keeps improving until NFE $=$ 79.
}%
\end{figure}
}

\newcommand{\figDegradation}{%
\renewcommand{\hh}{0.475\linewidth}
\begin{figure}[p]%
\centering%
\footnotesize%
\makebox[\hh]{Uncond. CIFAR-10, Pre-trained, VP, $\Snoise = 1.000$}\hfill%
\makebox[\hh]{Uncond. CIFAR-10, Pre-trained, VP, $\Snoise = 1.007$}\\%
\includegraphics[width=\hh]{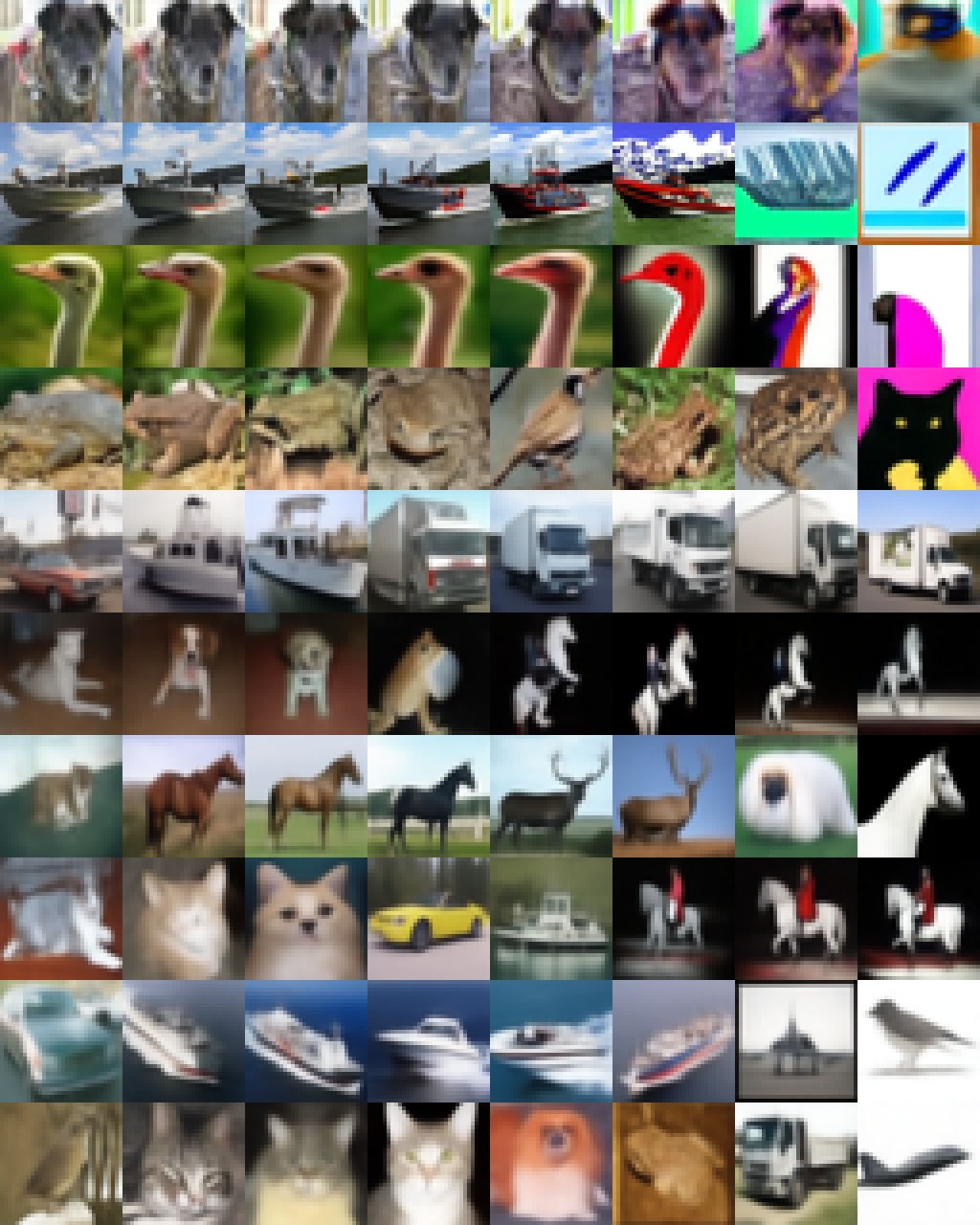}\hfill
\parbox[b][\hh/\real{8}*\real{10}]{0mm}{\scriptsize\vrlabel{0.02}\vrlabel{0.05}\vrlabel{0.10}\vrlabel{0.20}\vrlabel{0.30}\vrlabel{0.40}\vrlabel{0.50}\vrlabel{0.60}\vrlabel{0.70}\vrlabel{0.80}}\hfill%
\includegraphics[width=\hh]{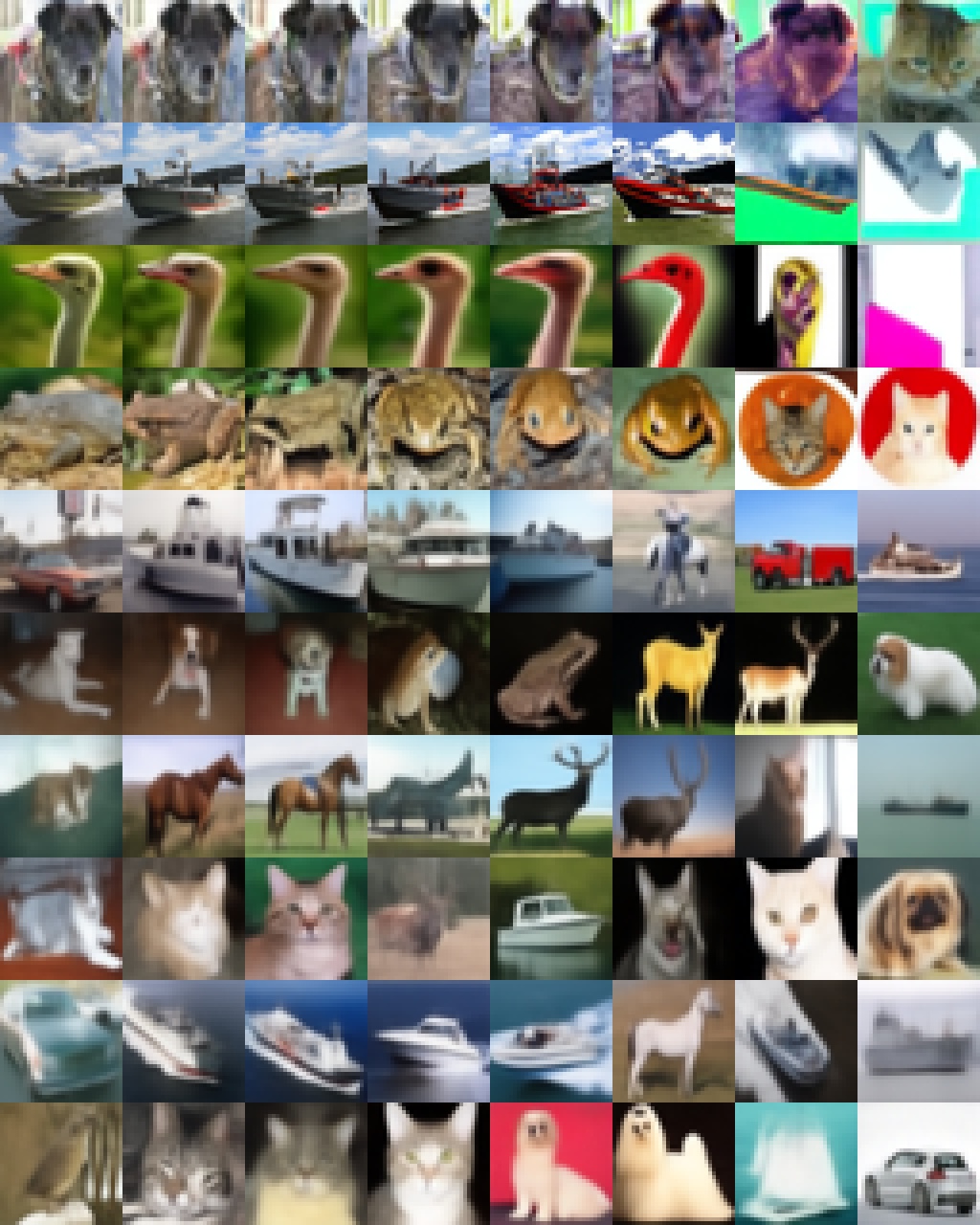}\\[-1mm]%
\makebox[\hh]{\scriptsize\hrlabel{Step 0\hspace{2.4mm}}\hrlabel{100}\hrlabel{200}\hrlabel{500}\hrlabel{1,000}\hrlabel{2,000}\hrlabel{5,000}\hrlabel{10,000}}\hfill%
\makebox[0mm]{$\sigma$}\hfill%
\makebox[\hh]{\scriptsize\hrlabel{Step 0\hspace{2.4mm}}\hrlabel{100}\hrlabel{200}\hrlabel{500}\hrlabel{1,000}\hrlabel{2,000}\hrlabel{5,000}\hrlabel{10,000}}\\[3mm]%
\makebox[\hh]{Cond. ImageNet-64, Pre-trained, $\Snoise = 1.000$}\hfill%
\makebox[\hh]{Cond. ImageNet-64, Pre-trained, $\Snoise = 1.003$}\\%
\includegraphics[width=\hh]{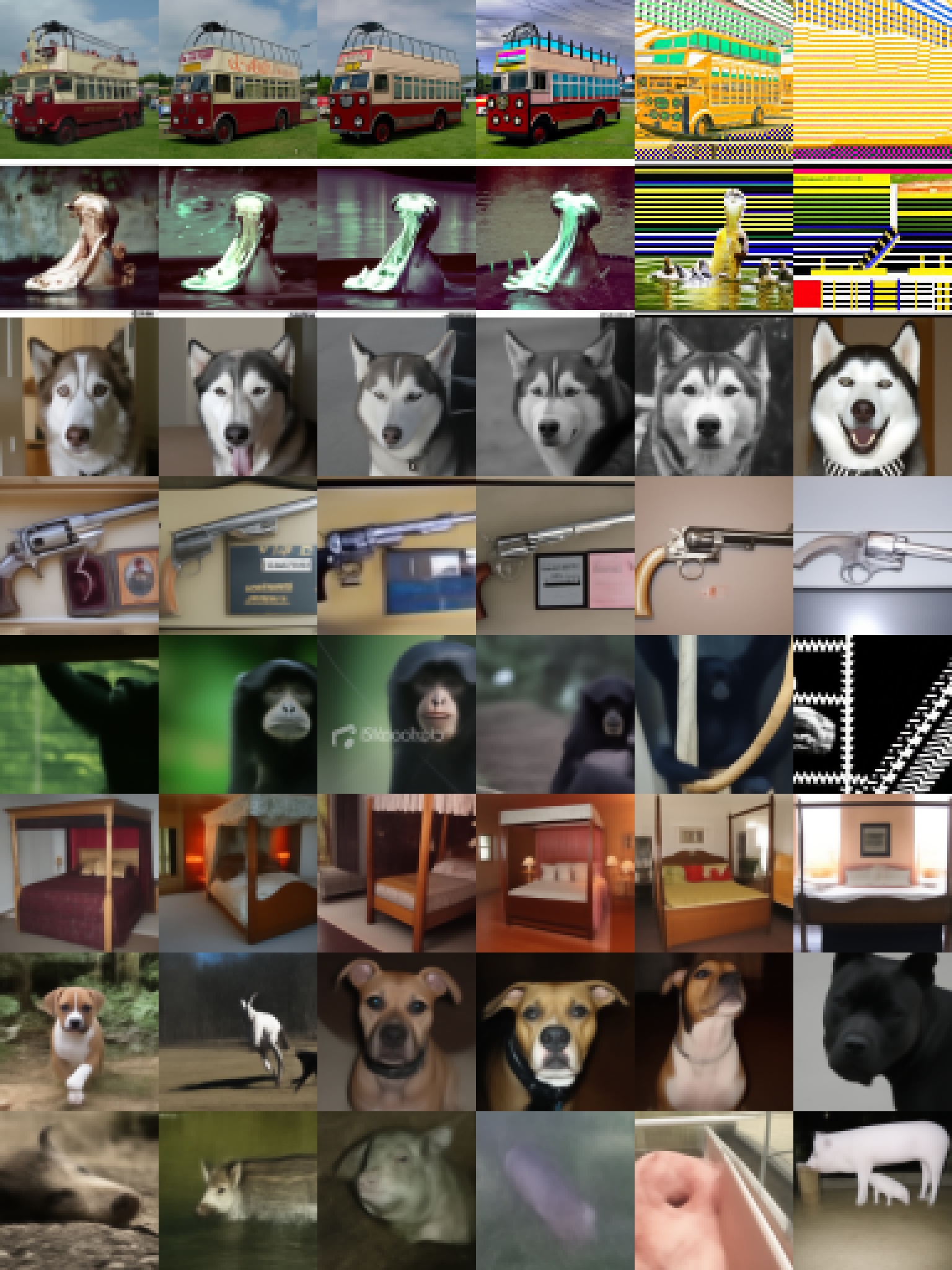}\hfill
\parbox[b][\hh/\real{6}*\real{8}]{0mm}{\scriptsize\vrlabel{0.05}\vrlabel{0.10}\vrlabel{0.20}\vrlabel{0.30}\vrlabel{0.40}\vrlabel{0.50}\vrlabel{0.60}\vrlabel{0.70}}\hfill%
\includegraphics[width=\hh]{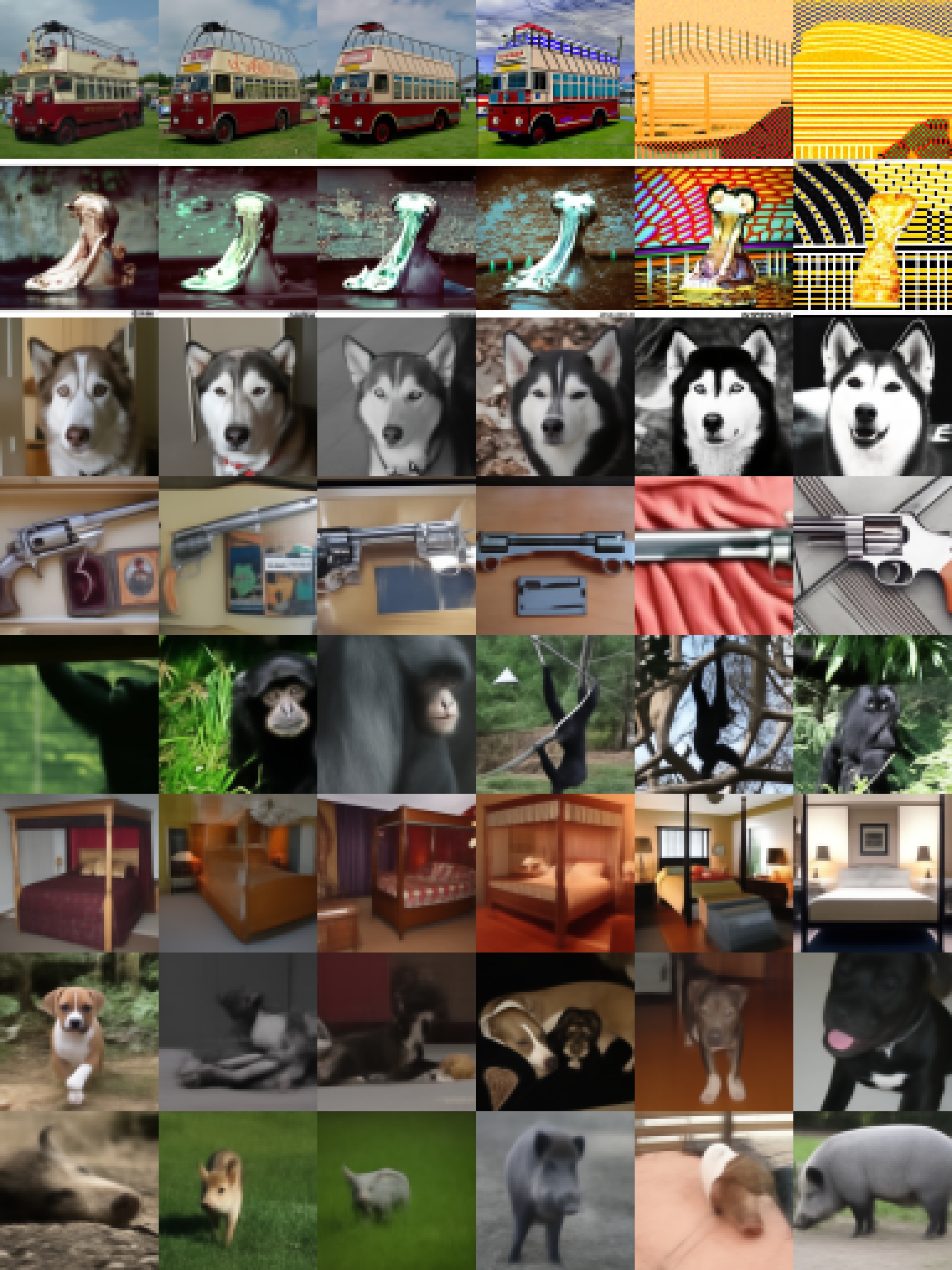}\\[-1mm]%
\makebox[\hh]{\scriptsize\hrlabel{Step 0\hspace{3.5mm}}\hrlabel{500}\hrlabel{1,000}\hrlabel{2,000}\hrlabel{5,000}\hrlabel{10,000}}\hfill%
\makebox[0mm]{$\sigma$}\hfill%
\makebox[\hh]{\scriptsize\hrlabel{Step 0\hspace{3.5mm}}\hrlabel{500}\hrlabel{1,000}\hrlabel{2,000}\hrlabel{5,000}\hrlabel{10,000}}\\[3mm]%
\caption{\label{fig:Degradation}%
Gradual image degradation with repeated addition and removal of noise.
We start with a random image drawn from $p(\xx; \sigma)$ (first column) and run Algorithm~\refpaper{alg:stochastic} for a certain number of steps (remaining columns) with fixed \smash{$\gamma_i = \sqrt{2}-1$}.
Each row corresponds to a specific choice of $\sigma$ (indicated in the middle) that we keep fixed throughout the entire process.
We visualize the results after running them through the denoiser, i.e., $D_\theta(\xx_i; \sigma)$.
}%
\end{figure}
}

\newcommand{\tabStochasticParams}{%
\forcenewcolumntype{x}{>{\centering\arraybackslash\hspace{0pt}}p{14mm}}%
\forcenewcolumntype{y}{>{\centering\arraybackslash\hspace{0pt}}p{19mm}}%
\forcenewcolumntype{z}{>{\centering\arraybackslash\hspace{0pt}}p{50mm}}%
\tabulinesep=0.7mm%
\tabulinestyle{0.17mm}%
\begin{table}[t]%
\centering%
\footnotesize%
\caption{\label{tab:StochasticParams}%
Parameters used for the stochastic sampling experiments in Section~\refpaper{sec:stochasticity}.
}
\vspace{1.5mm}%
\begin{tabu}{|c|@{}x@{}x@{}|@{}y@{}y@{}|z|}
\tabucline{-}
\multirow{2}{*}{Parameter}    & \multicolumn{2}{c|}{CIFAR-10} & \multicolumn{2}{c|}{ImageNet}     & \multirow{2}{*}{Grid search} \\
                              & VP      & VE                  & Pre-trained & {Our model}   & \\
\tabucline{-}
\makebox[7mm][l]{$\Schurn$}   & 30      & 80                  & 80          & {40}          & 0,\hfill 10,\hfill 20,\hfill 30,\hfill $\dots$,\hfill 70,\hfill 80,\hfill 90,\hfill 100 \\
\makebox[7mm][l]{$\Stmin$}    & 0.01    & 0.05                & 0.05        & {0.05}        & 0,\hfill 0.005,\hfill 0.01,\hfill 0.02,\hfill $\dots$,\hfill 1,\hfill 2,\hfill 5,\hfill 10 \\
\makebox[7mm][l]{$\Stmax$}    & 1       & 1                   & 50          & {50}          & 0.2,\hfill 0.5,\hfill 1,\hfill 2,\hfill $\dots$,\hfill 10,\hfill 20,\hfill 50,\hfill 80 \\
\makebox[7mm][l]{$\Snoise$}   & 1.007   & 1.007               & 1.003       & {1.003}       & 1.000,\hfill 1.001,\hfill $\dots$,\hfill 1.009,\hfill 1.010 \\
\tabucline{-}
\end{tabu}%
\end{table}%
}

\newcommand{\Translate}{\raisebox{0.1mm}{\scalebox{0.8}{\textsc{Translate2D}}}}
\newcommand{\Rotate}{\raisebox{0.1mm}{\scalebox{0.8}{\textsc{Rotate2D}}}}
\newcommand{\Scale}{\raisebox{0.1mm}{\scalebox{0.8}{\textsc{Scale2D}}}}

\newcommand{\tabAugmentPipe}{%
\tabulinesep=0.7mm%
\tabulinestyle{0.17mm}%
\begin{table}[t]%
\centering%
\footnotesize%
\caption{\label{tab:AugmentPipe}%
Our augmentation pipeline.
Each training image undergoes a combined geometric transformation based on 8 random parameters that receive non-zero values with a certain probability.
The model is conditioned with an additional 9-dimensional input vector derived from these parameters.
}
\vspace{1.5mm}%
\begin{tabu}{|l|l|l|c|l|l|}
\tabucline{-}
Augmentation  & Transformation                                & Parameters                          & Prob.             & Conditioning    & Constants                                     \\
\tabucline{-}
$x$-flip      & $\Scale\big(1 - 2 a_0, ~1\big)$               & $a_0 \sim \mathcal{U}\{0, 1\}$      & 100\%             & $a_0$           & \makebox[7mm]{$A_\text{prob}$}  $= 12$\%      \\
$y$-flip      & $\Scale\big(1, ~1 - 2 a_1\big)$               & $a_1 \sim \mathcal{U}\{0, 1\}$      & $A_\text{prob}$   & $a_1$           & \hspace{7mm}                    or $15$\%     \\
\tabucline{-}
Scaling       & $\Scale\big(  (A_\text{scale})^{a_2},$        & $a_2 \sim \mathcal{N}(0, 1)$        & $A_\text{prob}$   & $a_2$           & \makebox[7mm]{$A_\text{scale}$} $= 2^{0.2}$   \\
              & \hspace{12mm}$(A_\text{scale})^{a_2}\big)$    &                                     &                   &                 &                                               \\
\tabucline{-}
Rotation      & $\Rotate\big({-}a_3\big)$                     & $a_3 \sim \mathcal{U}(-\pi, \pi)$   & $A_\text{prob}$   & $\cos a_3 - 1$  &                                               \\
              &                                               &                                     &                   & $\sin a_3$      &                                               \\
\tabucline{-}
Anisotropy    & $\Rotate\big(a_4\big)$                        & $a_4 \sim \mathcal{U}(-\pi, \pi)$   & $A_\text{prob}$   & $a_5 ~\cos a_4$ & \makebox[7mm]{$A_\text{aniso}$} $= 2^{0.2}$   \\
              & $\Scale\big(  (A_\text{aniso})^{a_5},$        & $a_5 \sim \mathcal{N}(0, 1)$        &                   & $a_5 ~\sin a_4$ &                                               \\
              & \hspace{12mm}$1/(A_\text{aniso})^{a_5}\big)$  &                                     &                   &                 &                                               \\
              & $\Rotate\big({-}a_4\big)$                     &                                     &                   &                 &                                               \\
\tabucline{-}
Translation   & $\Translate\big((A_\text{trans}) a_6,$        & $a_6 \sim \mathcal{N}(0, 1)$        & $A_\text{prob}$   & $a_6$           & \makebox[7mm]{$A_\text{trans}$} $= 1/8$       \\
              & \hspace{17mm}  $(A_\text{trans}) a_7\big)$    & $a_7 \sim \mathcal{N}(0, 1)$        &                   & $a_7$           &                                               \\
\tabucline{-}
\end{tabu}%
\end{table}%
}

\newcommand{\SB}[1]{\scalebox{0.88}{#1}}
\newcommand{\YES}{\checkmark}
\newcommand{\NO}{--}

\newcommand{\tabTrainingParams}{%
\forcenewcolumntype{x}{>{\centering\arraybackslash\hspace{0pt}}p{20.9mm}}%
\tabulinesep=0.7mm%
\tabulinestyle{0.17mm}%
\begin{table}[t]%
\centering%
\footnotesize%
\caption{\label{tab:TrainingParams}%
Hyperparameters used for the training runs in Section~\refpaper{sec:training}.
}
\vspace{1.5mm}%
\begin{tabu}{|l|@{}x@{}x@{}|@{}x@{}x@{}|@{}x@{}|}
\tabucline{-}
\multirow{2}{*}{Hyperparameter}     & \multicolumn{2}{c|}{CIFAR-10} & \multicolumn{2}{c|}{FFHQ \& AFHQv2}   & {ImagetNet}   \\
                                    & \SB{Baseline}   & \SB{Ours}   & \SB{Baseline}   & \SB{Ours}           & {\SB{Ours}}   \\
\tabucline{-}
Number of GPUs                      & 4               & 8           & 4               & 8                   & {32}          \\
Duration \SB{(Mimg)}                & 200             & 200         & 200             & 200                 & {2500}        \\
Minibatch size                      & 128             & 512         & 128             & 256                 & {4096}        \\
Gradient clipping                   & \YES            & \NO         & \YES            & \NO                 & {\NO}         \\
{Mixed-precision \SB{(FP16)}} & {\NO}     & {\NO} & {\NO}     & {\NO}         & {\YES}        \\
\tabucline{-}
Learning rate \SB{${\times}10^4$}   & 2               & 10          & 2               & 2                   & {1}           \\
LR ramp-up \SB{(Mimg)}              & 0.64            & 10          & 0.64            & 10                  & {10}          \\
EMA half-life \SB{(Mimg)}           & 0.89 / 0.9      & 0.5         & 0.89 / 0.9      & 0.5                 & {50}          \\
                                    & \SB{(VP / VE)}  &             & \SB{(VP / VE)}  &                     &                     \\
Dropout probability                 & 10\%            & 13\%        & 10\%            & 5\% / 25\%          & {10\%}        \\
                                    &                 &             &                 & \SB{(FFHQ / AFHQ)}  &                     \\
\tabucline{-}
Channel multiplier                  & 128             & 128         & 128             & 128                 & {192}         \\
Channels per resolution             & 1-2-2-2         & 2-2-2       & 1-1-2-2-2       & 1-2-2-2             & {1-2-3-4}     \\
Dataset $x$-flips                   & \YES            & \NO         & \YES            & \NO                 & {\NO}         \\
Augment probability                 & \NO             & 12\%        & \NO             & 15\%                & {\NO}         \\
\tabucline{-}
\end{tabu}%
\end{table}%
}

\newcommand{\tabNetworkDetails}{%
\forcenewcolumntype{x}{>{\centering\arraybackslash\hspace{0pt}}p{23mm}}%
\tabulinesep=0.7mm%
\tabulinestyle{0.17mm}%
\begin{table}[t]%
\centering%
\footnotesize%
\caption{\label{tab:NetworkDetails}%
{Details of the network architectures used in this paper.}
}
\vspace{1.5mm}%
\begin{tabu}{|l|@{}x@{}x@{}x@{}|}
\tabucline{-}
\multirow{2}{*}{Parameter}              & \ddpmpp{}       & \ncsnpp{}       & {ADM}               \\
                                        & \SB{(VP)}       & \SB{(VE)}       & {\SB{(ImageNet)}}   \\
\tabucline{-}
Resampling filter                       & Box             & Bilinear        & {Box}               \\
Noise embedding                         & Positional      & Fourier         & {Positional}        \\
Skip connections in encoder             & \NO             & Residual        & {\NO}               \\
Skip connections in decoder             & \NO             & \NO             & {\NO}               \\
\tabucline{-}
{Residual blocks per resolution}  & {4}       & {4}       & {3}                 \\
{Attention resolutions}           & {\{16\}}  & {\{16\}}  & {\{32, 16, 8\}}     \\
{Attention heads}                 & {1}       & {1}       & {6-9-12}            \\
{Attention blocks in encoder}     & {4}       & {4}       & {9}                 \\
{Attention blocks in decoder}     & {2}       & {2}       & {13}                \\
\tabucline{-}
\end{tabu}%
\end{table}%
}

\newcommand{\algAlpha}{%
\begin{algorithm}[t]
\footnotesize
\captionof{algorithm}[alpha]{\atphantom\ \ Deterministic sampling using general 2\textsuperscript{nd} order Runge--Kutta, $\sigma(t)=t$ and $s(t)=1$.}
\begin{spacing}{1.1}
\begin{algorithmic}[1]
  \AProcedure{AlphaSampler}{$D_\theta(\xx;\sigma), ~t_{i \in \{0, \dots, N\}}, ~\alpha$}
    \AState{{\bf sample} $\xx_0 \sim \mathcal{N} \big( \boldzero, ~\odetime_0^2 ~\boldi \big)$}
    \AFor{$i \in \{0, \dots, N-1\}$}
      \AState{$h_i \gets \odetime_{i+1} - \odetime_i$}
        \AComment{Step length}
      \AState{$\dd_i \gets \big(\xx_i - D_\theta(\xx_i; \odetime_i)\big) / \odetime_i$}
        \AComment{Evaluate $\diff\xx / \diff\odetime$ at $(\xx,\,\odetime_i)$}
      \AState{$(\xx'_i,\,\odetime'_i) \gets (\xx_i + \alpha h \dd_i,\,\odetime_i + \alpha h)$}
        \AComment{Additional evaluation point}
      \AIf{$\odetime'_i \ne 0$}
        \AState{$\ddp_i \gets \big(\xx'_i - D_\theta(\xx'_i; \odetime'_i)\big) / \odetime'_i$}
          \AComment{Evaluate $\diff\xx / \diff\odetime$ at $(\xx'_i,\,\odetime'_i)$}
        \DState{$\displaystyle\xx_{i+1} \gets \xx_i + h\Big[\Big(1-\tfrac{1}{2\alpha}\Big)\dd_i + \tfrac{1}{2\alpha}\ddp_i\Big]$}
          \AComment{Second order step from $\odetime_i$ to $\odetime_{i+1}$}
      \Else
        \AState{$\xx_{i+1} \gets \xx_i + h \dd_i$}
          \AComment{Euler step from $\odetime_i$ to $\odetime_{i+1}$}
      \EndIf
    \EndFor
    \AState{\textbf{return} $\xx_N$}
  \EndProcedure
\end{algorithmic}
\end{spacing}
\label{alg:alpha}
\end{algorithm}
}

  \setcounter{figure}{5}    %
\setcounter{equation}{8}  %
\setcounter{table}{2}     %
\setcounter{algorithm}{2} %

\section{Additional results}
\label{app:results}

\figGridsImgcSampling
\figGridsImgcTraining
\figGridsCifaruSampling
\figGridsCifaruTraining
\figGridsCifarcTraining
\figGridsFfhqAfhqTraining
\figNfeSweep
\afterpage{\tabOdeTable}
\afterpage{\tabSdeTable}

{
Figure~\ref{fig:GridsImgcSampling} presents generated images for class-conditional ImageNet-64~\cite{Deng2009imagenet} using the pre-trained ADM model by Dhariwal and Nichol~\cite{Dhariwal2021}.
The original DDIM~\cite{Song2020ddim} and iDDPM~\cite{Nichol2021a} samplers are compared to ours in both deterministic and stochastic settings (Sections~\refpaper{sec:deterministic} and~\refpaper{sec:stochasticity}).
Figure~\ref{fig:GridsImgcTraining} shows the corresponding results that we obtain by training the model from scratch using our improved training configuration (Section~\refpaper{sec:training}).
}

{
The original samplers and training configurations by Song~et~al.~\cite{Song2021sde} are compared to ours in
  Figures~\ref{fig:GridsCifaruSampling} and~\ref{fig:GridsCifaruTraining} (unconditional CIFAR-10~\cite{Krizhevsky2009cifar}),
  Figure~\ref{fig:GridsCifarcTraining} (class-conditional CIFAR-10),
  and Figure~\ref{fig:GridsFfhqAfhqTraining} (FFHQ~\cite{Karras2018stylegan} and AFHQv2~\cite{Choi2020afhq}).
}
For ease of comparison, the same latent codes $\xx_0$ are used for each dataset/scenario across different training configurations and ODE choices.
Figure~\ref{fig:NfeSweep} shows generated image quality with various NFE when using deterministic sampling.

Tables~\ref{tab:OdeTable} and~\ref{tab:SdeTable} summarize the numerical results on deterministic and stochastic sampling methods in various datasets, previously shown as functions of NFE in Figures~\refpaper{fig:OdePlotNfe} and~\refpaper{fig:SdePlotNfe}.

\section{Derivation of formulas}
\label{app:formulas}

\subsection{Original ODE~/~SDE formulation from previous work}
\label{app:originalode}

Song~et~al.~\cite{Song2021sde} define their forward SDE (Eq.~5 in \cite{Song2021sde}) as
\begin{equation}
  \diff\xx = \boldsymbol{f}(\xx, t) ~\diff t + g(t) ~\diff \omega_t
  \text{,}
\end{equation}
where $\omega_t$ is the standard Wiener process and $\boldsymbol{f}(\cdot, \odetime): \mathbb{R}^d \rightarrow \mathbb{R}^d$ and $g(\cdot): \mathbb{R} \rightarrow \mathbb{R}$ are the drift and diffusion coefficients, respectively, where $d$ is the dimensionality of the dataset.
These coefficients are selected differently for the variance preserving (VP) and variance exploding (VE) formulations, and $\boldsymbol{f}(\cdot)$ is always of the form $\boldsymbol{f}(\xx, \odetime) = f(\odetime) ~\xx$, where $f(\cdot): \mathbb{R} \rightarrow \mathbb{R}$.
Thus, the SDE can be equivalently written as
\begin{equation}
  \label{eq:songsde}
  \diff\xx = f(t) ~\xx ~\diff t + g(t) ~\diff \omega_t
  \text{.}
\end{equation}

The perturbation kernels of this SDE (Eq.~29 in \cite{Song2021sde}) have the general form
\begin{equation}
  \label{eq:songperturbation}
  p_{0t}\big( \xx(t) ~|~ \xx(0) \big) = \mathcal{N} \big( \xx(t); ~s(t) ~\xx(0), ~s(t)^2 ~\sigma(t)^2 ~\boldi \big)
  \text{,}
\end{equation}
where {$\mathcal{N}(\xx; \boldsymbol{\mu}, \boldsymbol{\Sigma})$ denotes the probability density function of $\mathcal{N}(\boldsymbol{\mu}, \boldsymbol{\Sigma})$ evaluated at $\xx$,}
\begin{equation}
  \label{eq:songscale}
  s(t) = \exp\left( \int_0^t f(\xi) ~\diff\xi \right)
  \text{,}
  \hspace{4mm}\text{and}\hspace{4mm}
  \sigma(t) = \sqrt{\int_0^t \frac{g(\xi)^2}{s(\xi)^2} ~\diff\xi}
  \text{.}
\end{equation}

The marginal distribution $p_t(\xx)$ is obtained by integrating the perturbation kernels over $\xx(0)$:
\begin{equation}
  \label{eq:songmarginal}
  p_t(\xx) = \int_{\mathbb{R}^d} p_{0t}(\xx ~|~ \xx_0) ~\pdata(\xx_0) ~\diff \xx_0
  \text{.}
\end{equation}

Song~et~al.~\cite{Song2021sde} define the probability flow ODE (Eq.~13 in \cite{Song2021sde}) so that it obeys this same $p_t(\xx)$:
\begin{equation}
  \label{eq:songode}
  \diff\xx = \left[ f(t) ~\xx - \tfrac{1}{2} ~g(t)^2 ~\nablaxx \log p_t(\xx) \right] ~\diff t
  \text{.}
\end{equation}

\subsection{Our ODE formulation (Eq.~\refpaper{eq:ode} and Eq.~\refpaper{eq:odescale})}
\label{app:ourode}

The original ODE formulation (Eq.~\ref{eq:songode}) is built around the functions $f$ and $g$ that correspond directly to specific terms that appear in the formula; the properties of the marginal distribution (Eq.~\ref{eq:songscale}) can only be derived indirectly based on these functions.
However, $f$ and $g$ are of little practical interest in themselves, whereas the marginal distributions are of utmost importance in terms of training the model in the first place, bootstrapping the sampling process, and understanding how the ODE behaves in practice.
Given that the idea of the probability flow ODE is to match a particular set of marginal distributions, it makes sense to treat the marginal distributions as first-class citizens and define the ODE directly based on $\sigma(t)$ and $s(t)$, eliminating the need for $f(t)$ and $g(t)$.

Let us start by expressing the marginal distribution of Eq.~\ref{eq:songmarginal} in closed form:
\begin{eqnarray}
  p_t(\xx) &=& \int_{\mathbb{R}^d} p_{0t}(\xx ~|~ \xx_0) ~\pdata(\xx_0) ~\diff \xx_0 \\
  &=& \int_{\mathbb{R}^d} \pdata(\xx_0) ~\Big[ \mathcal{N} \big( \xx; ~s(t) ~\xx_0, ~s(t)^2 ~\sigma(t)^2 ~\boldi \big) \Big] ~\diff \xx_0 \\
  &=& \int_{\mathbb{R}^d} \pdata(\xx_0) ~\Big[ s(t)^{-d} ~\mathcal{N} \big( \xx / s(t); ~\xx_0, ~\sigma(t)^2 ~\boldi \big) \Big] ~\diff \xx_0 \\
  &=& s(t)^{-d} \int_{\mathbb{R}^d} \pdata(\xx_0) ~\mathcal{N} \big( \xx / s(t); ~\xx_0, ~\sigma(t)^2 ~\boldi \big) ~\diff \xx_0 \\
  &=& s(t)^{-d} ~\Big[ \pdata \ast \mathcal{N} \big( \boldzero, ~\sigma(t)^2 ~\boldi \big) \Big] \big( \xx / s(t) \big)
  \text{,}
\end{eqnarray}
where $p_a \ast p_b$ denotes the convolution of probability density functions $p_a$ and $p_b$.
The expression inside the brackets corresponds to a mollified version of $\pdata$ obtained by adding i.i.d.\ Gaussian noise to the samples.
Let us denote this distribution by $p(\xx; \sigma)$:
\begin{equation}
  \label{eq:psigma}
  p(\xx; \sigma) = \pdata \ast \mathcal{N} \big( \boldzero, ~\sigma(t)^2 ~\boldi \big)
  \hspace{5mm}\text{and}\hspace{5mm}
  p_t(\xx) = s(t)^{-d} ~p\big( \xx / s(t); \sigma(t) \big)
  \text{.}
\end{equation}

We can now express the probability flow ODE (Eq.~\ref{eq:songode}) using $p(\xx; \sigma)$ instead of $p_t(\xx)$:
\begin{eqnarray}
  \diff\xx &=& \left[ f(t) \xx - \tfrac{1}{2} ~g(t)^2 ~\nablaxx \log \big[ p_t(\xx) \big] \right] ~\diff t \\
  &=& \left[ f(t) \xx - \tfrac{1}{2} ~g(t)^2 ~\nablaxx \log \big[ s(t)^{-d} ~p\big( \xx / s(t); \sigma(t) \big) \big] \right] ~\diff t \\
  &=& \left[ f(t) \xx - \tfrac{1}{2} ~g(t)^2 ~\big[ \nablaxx \log s(t)^{-d} + \nablaxx \log p\big( \xx / s(t); \sigma(t) \big) \big] \right] ~\diff t \\
  &=& \left[ f(t) \xx - \tfrac{1}{2} ~g(t)^2 ~\nablaxx \log p\big( \xx / s(t); \sigma(t) \big) \right] ~\diff t
  \label{eq:tempode}
  \text{.}
\end{eqnarray}

Next, let us rewrite $f(t)$ in terms of $s(t)$ based on Eq.~\ref{eq:songscale}:
\begin{eqnarray}
  \exp\left( \int_0^t f(\xi) ~\diff\xi \right) &=& s(t) \\
  \int_0^t f(\xi) ~\diff\xi &=& \log s(t) \\
  \diff \bigg[ \int_0^t f(\xi) ~\diff\xi \bigg] \big/ \diff t &=& \diff \big[ \log s(t) \big] / \diff t \\
  f(t) &=& \dot s(t) / s(t)
  \label{eq:tempf}
  \text{.}
\end{eqnarray}

Similarly, we can also rewrite $g(t)$ in terms of $\sigma(t)$:
\begin{eqnarray}
  \sqrt{\int_0^t \frac{g(\xi)^2}{s(\xi)^2} ~\diff\xi} &=& \sigma(t) \\
  \int_0^t \frac{g(\xi)^2}{s(\xi)^2} ~\diff\xi &=& \sigma(t)^2 \\
  \diff \bigg[ \int_0^t \frac{g(\xi)^2}{s(\xi)^2} ~\diff\xi \bigg] \big/ \diff t &=& \diff \big[ \sigma(t)^2 \big] / \diff t \\
  g(t)^2 / s(t)^2 &=& 2 ~\dot\sigma(t) ~\sigma(t) \\
  g(t) / s(t) &=& \sqrt{2 ~\dot\sigma(t) ~\sigma(t)} \\
  g(t) &=& s(t) ~\sqrt{2 ~\dot\sigma(t) ~\sigma(t)}
  \label{eq:tempg}
  \text{.}
\end{eqnarray}

Finally, substitute $f$ (Eq.~\ref{eq:tempf}) and $g$ (Eq.~\ref{eq:tempg}) into the ODE of Eq.~\ref{eq:tempode}:
\begin{eqnarray}
  \diff\xx &=& \left[ [f(t)] ~\xx - \tfrac{1}{2} ~[g(t)]^2 ~\nablaxx \log p\big( \xx / s(t); \sigma(t) \big) \right] ~\diff t \\
  &=& \bigg[ \big[ \dot s(t) / s(t) \big] ~\xx - \tfrac{1}{2} ~\Big[ s(t) \sqrt{2 ~\dot\sigma(t) ~\sigma(t)} \Big]^2 ~\nablaxx \log p\big( \xx / s(t); \sigma(t) \big) \bigg] ~\diff t \\
  &=& \bigg[ \big[ \dot s(t) / s(t) \big] ~\xx - \tfrac{1}{2} ~\Big[ 2 ~s(t)^2 ~\dot\sigma(t) ~\sigma(t) \Big] ~\nablaxx \log p\big( \xx / s(t); \sigma(t) \big) \bigg] ~\diff t \\
  &=& \left[ \frac{\dot s(t)}{s(t)} ~\xx -s(t)^2 ~\dot\sigma(t) ~\sigma(t) ~\nablaxx \log p\left(\frac{\xx}{s(t)}; \sigma(t)\right) \right] ~\diff t
  \text{.}
\end{eqnarray}

Thus we have obtained Eq.~\refpaper{eq:odescale} in the main paper, and Eq.~\refpaper{eq:ode} is recovered by setting $s(t) = 1$:
\begin{equation}
  \diff\xx = -\dot\sigma(t) ~\sigma(t) ~\nablaxx \log p\big( \xx; \sigma(t) \big) ~\diff t
  \text{.}
\end{equation}

Our formulation (Eq.~\refpaper{eq:odescale}) highlights the fact that every realization of the probability flow ODE is simply a reparameterization of the same canonical ODE; changing $\sigma(t)$ corresponds to reparameterizing $t$, whereas changing $s(t)$ corresponds to reparameterizing $\xx$.

\subsection{Denoising score matching (Eq.~\refpaper{eq:score} and Eq.~\refpaper{eq:scoredenoiser})}
\label{app:scorematching}

For the sake of completeness, we derive the connection between score matching and denoising for a finite dataset. For a more general treatment and further background on the topic, see Hyv\"arinen~\cite{Hyvarinen05} and Vincent~\cite{Vincent11}.

Let us assume that our training set consists of a finite number of samples $\{\signal_1, \dots, \signal_Y\}$.
This implies $\pdata(\xx)$ is represented by a mixture of Dirac delta distributions:
\begin{equation}
  \pdata(\xx) = \frac{1}{Y} \sum_{i=1}^Y \delta \big( \xx - \signal_i \big)
  \text{,}
\end{equation}
which allows us to also express $p(\xx; \sigma)$ in closed form based on Eq.~\ref{eq:psigma}:
\begin{eqnarray}
  p(\xx; \sigma) &=& \pdata \ast \mathcal{N} \big( \boldzero, ~\sigma(t)^2 ~\boldi \big) \\
  &=& \int_{\mathbb{R}^d} \pdata(\xx_0) ~\mathcal{N} \big( \xx; ~\xx_0, ~\sigma^2 ~\boldi \big) ~\diff\xx_0 \\
  &=& \int_{\mathbb{R}^d} \Bigg[ \frac{1}{Y} \sum_{i=1}^Y \delta \big( \xx_0 - \signal_i \big) \Bigg] \mathcal{N} \big( \xx; ~\xx_0, ~\sigma^2 ~\boldi \big) ~\diff\xx_0 \\
  &=& \frac{1}{Y} \sum_{i=1}^Y \int_{\mathbb{R}^d} \mathcal{N} \big( \xx; ~\xx_0, ~\sigma^2 ~\boldi \big) ~\delta \big( \xx_0 - \signal_i \big) ~\diff\xx_0 \\
  &=& \frac{1}{Y} \sum_{i=1}^Y \mathcal{N} \big( \xx; ~\signal_i, ~\sigma^2 ~\boldi \big)
  \label{eq:diracpsigma}
  \text{.}
\end{eqnarray}

Let us now consider the denoising score matching loss of Eq.~\refpaper{eq:score}.
By expanding the expectations, we can rewrite the formula as an integral over the noisy samples $\xx$:
\begin{eqnarray}
  \mathcal{L}(D; \sigma) &=& \mathbb{E}_{\signal \sim \pdata} ~\mathbb{E}_{\noise \sim \mathcal{N}(\boldzero, \sigma^2 \boldi)} ~\big\lVert D(\signal + \noise; \sigma) - \signal \big\rVert^2_2 \\
  &=& \mathbb{E}_{\signal \sim \pdata} ~\mathbb{E}_{\xx \sim \mathcal{N}(\signal, \sigma^2 \boldi)} ~\big\lVert D(\xx; \sigma) - \signal \big\rVert^2_2 \\
  &=& \mathbb{E}_{\signal \sim \pdata} \int_{\mathbb{R}^d} \mathcal{N}(\xx; ~\signal, ~\sigma^2 ~\boldi) ~\big\lVert D(\xx; \sigma) - \signal \big\rVert^2_2 ~\diff\xx \\
  &=& \frac{1}{Y} \sum_{i=1}^Y \int_{\mathbb{R}^d} \mathcal{N}(\xx; ~\signal_i, ~\sigma^2 ~\boldi) ~\big\lVert D(\xx; \sigma) - \signal_i \big\rVert^2_2 ~\diff\xx \\
  &=& \int_{\mathbb{R}^d} \underbrace{\frac{1}{Y} \sum_{i=1}^Y \mathcal{N}(\xx; ~\signal_i, ~\sigma^2 ~\boldi) ~\big\lVert D(\xx; \sigma) - \signal_i \big\rVert^2_2}_{=: ~ \mathcal{L}(D; \xx, \sigma)}  ~\diff\xx
  \label{eq:slicedloss}
  \text{.}
\end{eqnarray}

Eq.~\ref{eq:slicedloss} means that we can minimize $\mathcal{L}(D; \sigma)$ by minimizing $\mathcal{L}(D; \xx, \sigma)$ independently for each $\xx$:
\begin{equation}
  D(\xx; \sigma) = \argmin_{D(\xx; \sigma)} \mathcal{L}(D; \xx, \sigma)
  \text{.}
\end{equation}
This is a convex optimization problem; its solution is uniquely identified by setting the gradient w.r.t. $D(\xx; \sigma)$ to zero:
\begin{eqnarray}
  \boldzero &=& \nabla_{D(\xx; \sigma)} \Big[ \mathcal{L}(D; \xx, \sigma) \Big] \\
  \boldzero &=& \nabla_{D(\xx; \sigma)} \Bigg[ \frac{1}{Y} \sum_{i=1}^Y \mathcal{N}(\xx; ~\signal_i, ~\sigma^2 ~\boldi) ~\big\lVert D(\xx; \sigma) - \signal_i \big\rVert^2_2 \Bigg] \\
  \boldzero &=& \sum_{i=1}^Y \mathcal{N}(\xx; ~\signal_i, ~\sigma^2 ~\boldi) ~\nabla_{D(\xx; \sigma)} \Big[ \big\lVert D(\xx; \sigma) - \signal_i \big\rVert^2_2 \Big] \\
  \boldzero &=& \sum_{i=1}^Y \mathcal{N}(\xx; ~\signal_i, ~\sigma^2 ~\boldi) ~\Big[ 2 ~D(\xx; \sigma) - 2~\signal_i \Big] \\
  \boldzero &=& \Bigg[ \sum_{i=1}^Y \mathcal{N}(\xx; ~\signal_i, ~\sigma^2 ~\boldi) \Bigg] D(\xx; \sigma) - \sum_{i=1}^Y \mathcal{N}(\xx; ~\signal_i, ~\sigma^2 ~\boldi) ~\signal_i \\
  D(\xx; \sigma) &=& \frac{ \sum_i \mathcal{N}(\xx; ~\signal_i, ~\sigma^2 ~\boldi) ~\signal_i }{ \sum_i \mathcal{N}(\xx; ~\signal_i, ~\sigma^2 ~\boldi) }
  \label{eq:idealdenoiser}
  \text{,}
\end{eqnarray}
which gives a closed-form solution for the ideal denoiser $D(\xx; \sigma)$.
Note that Eq.~\ref{eq:idealdenoiser} is feasible to compute in practice for small datasets\,---\,we show the results for CIFAR-10 in Figure~\refpaper{fig:denoising}b.

Next, let us consider the score of the distribution $p(\xx; \sigma)$ defined in Eq.~\ref{eq:diracpsigma}:
\begin{eqnarray}
  \nablaxx \log p(\xx; \sigma) &=& \frac{\nablaxx p(\xx; \sigma)}{p(\xx; \sigma)} \\
  &=& \frac{ \nablaxx \Big[ \frac{1}{Y} \sum_i \mathcal{N} \big( \xx; ~\signal_i, ~\sigma^2 ~\boldi \big) \Big] }{ \Big[ \frac{1}{Y} \sum_i \mathcal{N} \big( \xx; ~\signal_i, ~\sigma^2 ~\boldi \big) \Big] } \\
  &=& \frac{ \sum_i \nablaxx \mathcal{N} \big( \xx; ~\signal_i, ~\sigma^2 ~\boldi \big) }{ \sum_i \mathcal{N} \big( \xx; ~\signal_i, ~\sigma^2 ~\boldi \big) }
  \label{eq:diracscoretemp}
  \text{.}
\end{eqnarray}

We can simplify the numerator of Eq.~\ref{eq:diracscoretemp} further:
\begin{eqnarray}
  \nablaxx \mathcal{N} \big( \xx; ~\signal_i, ~\sigma^2 ~\boldi \big) &=& \nablaxx \Bigg[ \big( 2 \pi \sigma^2 \big)^{-\frac{d}{2}} ~\exp \frac{\lVert \xx - \signal_i \rVert_2^2}{-2 ~\sigma^2} \Bigg] \\
  &=& \big( 2 \pi \sigma^2 \big)^{-\frac{d}{2}} ~\nablaxx \Bigg[ \exp \frac{\lVert \xx - \signal_i \rVert_2^2}{-2 ~\sigma^2} \Bigg] \\
  &=& \Bigg[\big( 2 \pi \sigma^2 \big)^{-\frac{d}{2}} \exp \frac{\lVert \xx - \signal_i \rVert_2^2}{-2 ~\sigma^2} \Bigg] ~\nablaxx \Bigg[ \frac{\lVert \xx - \signal_i \rVert_2^2}{-2 ~\sigma^2} \Bigg] \\
  &=& \mathcal{N} \big( \xx; ~\signal_i, ~\sigma^2 ~\boldi \big) ~\nablaxx \Bigg[ \frac{\lVert \xx - \signal_i \rVert_2^2}{-2 ~\sigma^2} \Bigg] \\
  &=& \mathcal{N} \big( \xx; ~\signal_i, ~\sigma^2 ~\boldi \big) \bigg[ \frac{\signal_i - \xx}{\sigma^2} \bigg]
  \text{.}
\end{eqnarray}

Let us substitute the result back to Eq.~\ref{eq:diracscoretemp}:
\begin{eqnarray}
  \nablaxx \log p(\xx; \sigma) &=& \frac{ \sum_i \nablaxx \mathcal{N} \big( \xx; ~\signal_i, ~\sigma^2 ~\boldi \big) }{ \sum_i \mathcal{N} \big( \xx; ~\signal_i, ~\sigma^2 ~\boldi \big) } \\
  &=& \frac{ \sum_i \mathcal{N} \big( \xx; ~\signal_i, ~\sigma^2 ~\boldi \big) \Big[ \frac{\signal_i - \xx}{\sigma^2} \Big] }{ \sum_i \mathcal{N} \big( \xx; ~\signal_i, ~\sigma^2 ~\boldi \big) } \\
  &=& \Bigg( \frac{ \sum_i \mathcal{N} \big( \xx; ~\signal_i, ~\sigma^2 ~\boldi \big) \signal_i }{ \sum_i \mathcal{N} \big( \xx; ~\signal_i, ~\sigma^2 ~\boldi \big) } - \xx \Bigg) \big/ \sigma^2
  \label{eq:diracscore}
  \text{.}
\end{eqnarray}

Notice that the fraction in Eq.~\ref{eq:diracscore} is identical to Eq.~\ref{eq:idealdenoiser}.
We can thus equivalently write Eq.~\ref{eq:diracscore} as
\begin{equation}
  \nablaxx \log p(\xx; \sigma) = \big( D(\xx; ~\sigma) - \xx \big) / \sigma^2
  \text{,}
\end{equation}
which matches Eq.~\refpaper{eq:scoredenoiser} in the main paper.

\subsection{Evaluating our ODE in practice (Algorithm~\refpaper{alg:heun})}
\label{app:odeinpractice}

Let us consider $\xx$ to be a scaled version of an original, non-scaled variable $\hat\xx$ and substitute $\xx = s(t) ~\hat\xx$ into the score term that appears in our scaled ODE (Eq.~\refpaper{eq:odescale}):
\begin{eqnarray}
  && \nablaxx \log p\big( \xx / s(t); \sigma(t) \big) \\
  &=& \nabla_{[ s(t) \hat \xx ]} \log p\big( [s(t) ~\hat\xx] / s(t); \sigma(t) \big) \\
  &=& \nabla_{s(t) \hat \xx} \log p\big( \hat\xx; \sigma(t) \big) \\
  &=& \tfrac{1}{s(t)} \nabla_{\hat\xx} \log p\big( \hat\xx; \sigma(t) \big)
  \text{.}
\end{eqnarray}

We can further rewrite this with respect to $D(\cdot)$ using Eq.~\refpaper{eq:scoredenoiser}:
\begin{equation}
  \nablaxx \log p\big( \xx / s(t); \sigma(t) \big) ~=~ \tfrac{1}{s(t) \sigma(t)^2} \Big( D\big( \hat\xx; \sigma(t) \big) - \hat\xx \Big)
  \label{eq:scaledscore}
  \text{.}
\end{equation}

Let us now substitute Eq.~\ref{eq:scaledscore} into Eq.~\refpaper{eq:odescale}, approximating the ideal denoiser $D(\cdot)$ with our trained model $D_\theta(\cdot)$:
\begin{eqnarray}
  \diff\xx &=& \left[ \dot s(t) ~\xx / s(t) - s(t)^2 ~\dot\sigma(t) ~\sigma(t) ~\Big[ \tfrac{1}{s(t) \sigma(t)^2} \Big( D_\theta \big( \hat\xx; \sigma(t) \big) - \hat\xx \Big) \Big] \right] ~\diff t \\
  &=& \left[ \tfrac{\dot s(t)}{s(t)} ~\xx - \tfrac{\dot\sigma(t) s(t)}{\sigma(t)} \Big( D_\theta \big( \hat\xx; \sigma(t) \big) - \hat\xx \Big) \right] ~\diff t
  \text{.}
\end{eqnarray}

Finally, backsubstitute $\hat\xx = \xx / s(t)$:
\begin{eqnarray}
  \diff\xx &=& \left[ \tfrac{\dot s(t)}{s(t)} ~\xx - \tfrac{\dot\sigma(t) s(t)}{\sigma(t)} \Big( D_\theta \big( [\hat\xx]; \sigma(t) \big) - [\hat\xx] \Big) \right] ~\diff t \\
  &=& \left[ \tfrac{\dot s(t)}{s(t)} ~\xx - \tfrac{\dot\sigma(t) s(t)}{\sigma(t)} \Big( D_\theta \big( [\xx / s(t)]; \sigma(t) \big) - [\xx / s(t)] \Big) \right] ~\diff t \\
  &=& \left[ \tfrac{\dot s(t)}{s(t)} ~\xx - \tfrac{\dot\sigma(t) s(t)}{\sigma(t)} D_\theta \big( \xx / s(t); \sigma(t) \big) + \tfrac{\dot\sigma(t)}{\sigma(t)} ~\xx \right] ~\diff t \\
  &=& \left[ \left( \tfrac{\dot\sigma(t)}{\sigma(t)} + \tfrac{\dot s(t)}{s(t)} \right) \xx - \tfrac{\dot\sigma(t) s(t)}{\sigma(t)} D_\theta \big( \xx / s(t); \sigma(t) \big) \right] ~\diff t
  \label{eq:practicalode}
  \text{.}
\end{eqnarray}

We can equivalenty write Eq.~\ref{eq:practicalode} as
\begin{equation}
  \diff\xx / \diff t = \bigg( \frac{\dot\sigma(t)}{\sigma(t)} + \frac{\dot s(t)}{s(t)} \bigg) \xx - \frac{\dot\sigma(t) s(t)}{\sigma(t)} D_\theta \bigg( \frac{\xx}{s(t)}; \sigma(t) \bigg)
  \text{,}
\end{equation}
matching lines 4 and 7 of Algorithm~\refpaper{alg:heun}.

\subsection{Our SDE formulation (Eq.~\refpaper{eq:sde})}
\label{app:oursde}

We derive the SDE of Eq.~\refpaper{eq:sde} by the following strategy:
\begin{itemize}
\item
The desired marginal densities $p\big( \xx; \sigma(t) \big)$ are convolutions of the data density $\pdata$ and an isotropic Gaussian density with standard deviation $\sigma(t)$ (see Eq.~\ref{eq:psigma}). Hence, considered as a function of the time $t$, the density evolves according to a heat diffusion PDE with time-varying diffusivity. As a first step, we find this PDE.%
\item
We then use the Fokker--Planck equation to recover a family of SDEs for which the density evolves according to this PDE. Eq.~\refpaper{eq:sde} is obtained from a suitable parametrization of this family.
\end{itemize}

\subsubsection{Generating the marginals by heat diffusion}

We consider the time evolution of a probability density $\pd(\xx, t)$. Our goal is to find a PDE whose solution with the initial value $\pd(\xx, 0) := \pdata(\xx)$ is $\pd(\xx, t) = p\big( \xx, \sigma(t) \big)$. That is, the PDE should reproduce the marginals we postulate in Eq.~\ref{eq:psigma}.

The desired marginals are convolutions of $\pdata$ with isotropic normal distributions of time-varying standard deviation $\sigma(t)$, and as such, can be generated by the heat equation with time-varying diffusivity $\condu(t)$. The situation is most conveniently analyzed in the Fourier domain, where the marginal densities are simply pointwise products of a Gaussian function and the transformed data density. To find the diffusivity that induces the correct standard deviations, we first write down the heat equation PDE:
\begin{equation}
  \frac{\partial \pd(\xx, t)}{\partial t} = \condu(t) \Deltaxx \pd(\xx, t)
  \label{eq:pde}
  \text{.}
\end{equation}

The Fourier transformed counterpart of Eq.~\ref{eq:pde}, where the transform is taken along the $\xx$-dimension, is given by
\begin{equation}
  \frac{\partial \hat \pd(\freq, t)}{\partial t} = - \condu(t) |\freq|^2 \hat \pd(\freq, t)
  \label{eq:pdeft}
  \text{.}
\end{equation}

The target solution $\pd(\xx, t)$ and its Fourier transform $\hat \pd(\freq, t)$ are given by Eq.~\ref{eq:psigma}:
\begin{eqnarray}
  \pd(\xx, t) &=& p\big( \xx; \sigma(t) \big) = \pdata(\xx) \ast \mathcal{N}\big( \boldzero, ~\sigma(t)^2 ~\boldi \big) \\
  \hat \pd(\freq, t) &=& \pdatahat(\freq) ~\exp\Big( {-}\tfrac{1}{2} ~|\freq|^2 ~\sigma(t)^2 \Big)
  \text{.}
\end{eqnarray}

Differentiating the target solution along the time axis, we have
\begin{eqnarray}
  \frac{\partial \hat \pd(\freq, t)}{\partial t} &=& - \dot\sigma(t) \sigma(t) ~|\freq|^2 ~ \pdatahat(\freq) ~\exp\Big( {-}\tfrac{1}{2} ~|\freq|^2 ~\sigma(t)^2 \Big) \\
  &=& - \dot \sigma(t) \sigma(t) ~|\freq|^2 ~\hat\pd(\freq,t)
  \label{eq:pdesolutionftd}
  \text{.}
\end{eqnarray}

Eqs.~\ref{eq:pdeft} and \ref{eq:pdesolutionftd} share the same left hand side. Equating them allows us to solve for $\condu(t)$ that generates the desired evolution:
\begin{eqnarray}
  - \condu(t) |\freq|^2 \hat \pd(\freq, t) &=& - \dot \sigma(t) \sigma(t) ~ |\freq|^2 ~ \hat \pd(\freq,t) \\
  \condu(t) &=& \dot \sigma(t) \sigma(t)
  \text{.}
\end{eqnarray}

To summarize, the desired marginal densities corresponding to noise levels $\sigma(t)$ are generated by the PDE
\begin{equation}
  \frac{\partial \pd(\xx, t)}{\partial t} = \dot \sigma(t) \sigma(t) \Deltaxx \pd(\xx, t)
  \label{eq:marginalpde}
\end{equation}
from the initial density $\pd(\xx, 0) = \pdata(\xx)$.

\subsubsection{Derivation of our SDE}

Given an SDE
\begin{equation}
  \diff \xx = \ff(\xx, t) ~ \diff t ~ + ~ \gb(\xx, t) ~ \diff\wproc
  \label{eq:sdegeneral}
  \text{,}
\end{equation}
the Fokker--Planck PDE describes the time evolution of its solution probability density $\fp(\xx, t)$ as
\begin{equation}
  \frac{\partial \fp(\xx, t)}{\partial t} = -\nablaxx \cdot \big( \ff(\xx,t) ~\fp(\xx,t) \big) + \tfrac{1}{2} \nablaxx \nablaxx : \big( \mathbf{D}(\xx, t) ~\fp(\xx, t) \big)
  \text{,}
\end{equation}
where $\mathbf{D}_{ij} = \sum_k \gb_{ik} \gb_{jk}$ is the \emph{diffusion tensor}. %
We consider the special case $\gb(\xx, t) = g(t) ~\boldi$ of $\xx$-independent white noise addition, whereby the equation simplifies to
\begin{equation}
  \frac{\partial \fp(\xx, t)}{\partial t} = -\nablaxx \cdot \big( \ff(\xx,t) ~\fp(\xx,t) \big) + \tfrac{1}{2} ~g(t)^2 ~\Deltaxx \fp(\xx, t)
  \label{eq:fokkerplanck}
  \text{.}
\end{equation}

We are seeking an SDE whose solution density is described by the PDE in Eq.~\ref{eq:marginalpde}. Setting $\fp(\xx, t) = q(\xx, t)$ and equating Eqs.~\ref{eq:fokkerplanck} and \ref{eq:marginalpde}, we find the sufficient condition that the SDE must satisfy
\begin{eqnarray}
  -\nablaxx \cdot \big( \ff(\xx,t) ~\pd(\xx,t) \big) + \tfrac{1}{2} ~g(t)^2 ~\Deltaxx \pd(\xx, t) &=& \dot\sigma(t) ~\sigma(t) ~\Deltaxx \pd(\xx, t) \\
  \nablaxx \cdot \big( \ff(\xx,t) ~\pd(\xx,t) \big) &=& \Big( \tfrac{1}{2} ~g(t)^2 - \dot\sigma(t) ~\sigma(t) \Big) ~\Deltaxx \pd(\xx, t)
  \text{.}
\end{eqnarray}

Any choice of functions $\ff(\xx,t)$ and $g(t)$ satisfying this equation constitute a sought after SDE. Let us now find a specific family of such solutions. The key idea is given by the identity $\nablaxx \cdot \nablaxx = \Deltaxx$. Indeed, if we set $\ff(\xx,t) ~\pd(\xx,t) = \upsilon(t) ~\nablaxx \pd(\xx,t)$ for any choice of $\upsilon(t)$, the term $\Deltaxx \pd(\xx,t)$ appears on both sides and cancels out:
\begin{eqnarray}
  \nablaxx \cdot \big( \upsilon(t) ~\nablaxx \pd(\xx,t) \big) &=& \Big( \tfrac{1}{2} ~g(t)^2 - \dot\sigma(t) ~\sigma(t) \Big) ~\Deltaxx \pd(\xx, t) \\
  \upsilon(t) ~\Deltaxx \pd(\xx,t) &=& \Big( \tfrac{1}{2} ~g(t)^2 - \dot\sigma(t) ~\sigma(t) \Big) ~\Deltaxx \pd(\xx, t) \\
  \upsilon(t) &=& \tfrac{1}{2} ~g(t)^2 - \dot\sigma(t) ~\sigma(t)
  \text{.}
\end{eqnarray}

The stated $\ff(\xx,t)$ is in fact proportional to the score function, as the formula matches the gradient of the logarithm of the density:
\begin{eqnarray}
  \ff(\xx, t) &=& \upsilon(t) ~\frac{\nablaxx q(\xx,t)}{q(\xx,t)} \\
  &=& \upsilon(t) ~\nablaxx \log q(\xx, t) \\
  &=& \Big( \tfrac{1}{2} ~g(t)^2 - \dot\sigma(t) ~\sigma(t) \Big) ~\nablaxx \log q(\xx, t)
  \text{.}
\end{eqnarray}

Substituting this back into Eq.~\ref{eq:sdegeneral} and writing $p(\xx; \sigma(t))$ in place of $q(\xx,t)$, we recover a family of SDEs whose solution densities have the desired marginals with noise levels $\sigma(t)$ for any choice of $g(t)$:
\begin{equation}
  \diff \xx = \Big( \tfrac{1}{2} ~g(t)^2 - \dot\sigma(t) ~\sigma(t) \Big) ~\nablaxx \log p\big( \xx; \sigma(t) \big) ~\diff t ~+~ g(t) ~\diff\wproc
  \text{.}
\end{equation}

The free parameter $g(t)$ effectively specifies the rate of noise replacement at any given time instance. The special case choice of $g(t) = 0$ corresponds to the probability flow ODE. The parametrization by $g(t)$ is not particularly intuitive, however. To obtain a more interpretable parametrization, we set $g(t) = \sqrt{2 ~\churn(\odetime)} ~\sigma(\odetime)$, which yields the (forward) SDE of Eq.~\refpaper{eq:sde} in the main paper:
\begin{equation}
  \diff \xx_{+} =
    -\dot\sigma(\odetime) \sigma(\odetime) \nablaxx \log p\big( \xx; \sigma(\odetime) \big) \,\diff\odetime\, + \,
      \churn(\odetime) \sigma(\odetime)^2 \nablaxx \log p\big( \xx; \sigma(\odetime) \big) \,\diff\odetime +
      \sqrt{2 \churn(\odetime)} \sigma(\odetime) \,\diff\wproc
  \label{eq:sdeforward}
  \text{.}
\end{equation}

The noise replacement is now proportional to the standard deviation $\sigma(t)$ of the noise, with the proportionality factor $\churn(t)$. Indeed, expanding the score function in the middle term according to Eq.~\refpaper{eq:scoredenoiser} yields $\beta(t) ~\big[ D\big( \xx;\sigma(t) \big) - \xx \big] ~\diff t$, which changes $\xx$ proportionally to the negative noise component; the stochastic term injects new noise at the same rate. Intuitively, scaling the magnitude of Langevin exploration according to the current noise standard deviation is a reasonable baseline, as the data manifold is effectively ``spread out'' by this amount due to the blurring of the density.

The \emph{reverse} SDE used in denoising diffusion is simply obtained by applying the time reversal formula of Anderson~\cite{Anderson1982} (as stated in Eq. 6 of Song et al.~\cite{Song2021sde}) on Eq.~\ref{eq:sdeforward}; the entire effect of the reversal is a change of sign in the middle term. 

The scaled generalization of the SDE can be derived using a similar approach as with the ODE previously.
As such, the derivation is omitted here.

\subsection{Our preconditioning and training (Eq.~\refpaper{eq:precloss})}
\label{app:ourprecond}

Following Eq.~\refpaper{eq:score}, the denoising score matching loss for a given denoiser $D_\theta$ on a given noise level $\sigma$ is given by
\begin{equation}
  \mathcal{L}(D_\theta; \sigma) = \mathbb{E}_{\signal \sim \pdata} ~\mathbb{E}_{\noise \sim \mathcal{N}(\boldzero, \sigma^2 \boldi)} ~\big\lVert D_\theta(\signal + \noise; \sigma) - \signal \big\rVert^2_2
  \text{.}
\end{equation}

We obtain overall training loss by taking a weighted expectation of $\mathcal{L}(D_\theta; \sigma)$ over the noise levels:
\begin{eqnarray}
  \mathcal{L}(D_\theta) &=& \mathbb{E}_{\sigma \sim \ptrain} \big[ \lambda(\sigma) ~\mathcal{L}(D_\theta; \sigma) \big] \\
  &=& \mathbb{E}_{\sigma \sim \ptrain} ~\Big[ \lambda(\sigma) ~\mathbb{E}_{\signal \sim \pdata} ~\mathbb{E}_{\noise \sim \mathcal{N}(\boldzero, \sigma^2 \boldi)} ~\big\lVert D_\theta(\signal + \noise; \sigma) - \signal \big\rVert^2_2 \Big] \\
  &=& \mathbb{E}_{\sigma \sim \ptrain} ~\mathbb{E}_{\signal \sim \pdata} ~\mathbb{E}_{\noise \sim \mathcal{N}(\boldzero, \sigma^2 \boldi)}  ~\Big[ \lambda(\sigma) ~\big\lVert D_\theta(\signal + \noise; \sigma) - \signal \big\rVert^2_2 \Big] \\
  &=& \mathbb{E}_{\sigma, \signal, \noise} \Big[ \lambda(\sigma) ~\big\lVert D_\theta(\signal + \noise; \sigma) - \signal \big\rVert^2_2 \Big]
  \label{eq:totalloss}
  \text{,}
\end{eqnarray}
where the noise levels are distributed according to $\sigma \sim \ptrain$ and weighted by $\lambda(\sigma)$.

Using our definition of $D_\theta(\cdot)$ from Eq.~\refpaper{eq:preconditioning}, we can further rewrite $\mathcal{L}(D_\theta)$ as
\begin{eqnarray}
  && \mathbb{E}_{\sigma, \signal, \noise} \Big[ \lambda(\sigma) \big\lVert \cskip(\sigma) (\signal {+} \noise) + \cout(\sigma) F_\theta\big(\Fargs\big) - \signal \big\rVert^2_2 \Big] \label{eq:lossexpanded} \\
  &=& \mathbb{E}_{\sigma, \signal, \noise} \Big[ \lambda(\sigma) \big\lVert \cout(\sigma) F_\theta\big(\Fargs\big) - \big( \signal - \cskip(\sigma) (\signal + \noise) \big) \big\rVert^2_2 \Big] \\
  &=& \mathbb{E}_{\sigma, \signal, \noise} \Big[ \lambda(\sigma) \cout(\sigma)^2 \big\lVert F_\theta\big(\Fargs\big) - \tfrac{1}{\cout(\sigma)} \big( \signal - \cskip(\sigma) (\signal {+} \noise) \big) \big\rVert^2_2 \Big] \\
  &=& \mathbb{E}_{\sigma, \signal, \noise} \Big[ w(\sigma) ~\big\lVert F_\theta\big(\Fargs\big) - F_\text{target}(\signal, \noise; \sigma) \big\rVert^2_2 \Big]
  \text{,}
\end{eqnarray}
which matches Eq.~\refpaper{eq:precloss} and corresponds to traditional supervised training of $F_\theta$ using standard $L_2$ loss with effective weight $w(\cdot)$ and target $F_\text{target}(\cdot)$ given by
\begin{equation}
  w(\sigma) = \lambda(\sigma) ~\cout(\sigma)^2
  \hspace{4mm}\text{and}\hspace{4mm}
  F_\text{target}(\signal, \noise; \sigma) = \tfrac{1}{\cout(\sigma)} \big( \signal - \cskip(\sigma) (\signal + \noise) \big)
  \text{,}
\end{equation}

We can now derive formulas for $\cin(\sigma)$, $\cout(\sigma)$, $\cskip(\sigma)$, and $\lambda(\sigma)$ from first principles, shown in the ``Ours'' column of Table~\refpaper{tab:specifics}.

First, we require the training inputs of $F_\theta(\cdot)$ to have unit variance:
\begin{eqnarray}
  \Var_{\signal, \noise} \big[ \cin(\sigma) (\signal + \noise) \big] &=& 1 \\
  \cin(\sigma)^2 ~\Var_{\signal, \noise} \big[ \signal + \noise \big] &=& 1 \\
  \cin(\sigma)^2 \big( \sdata^2 + \sigma^2 \big) &=& 1 \\
  \cin(\sigma) &=& 1 \big/ \sqrt{\sigma^2 + \sdata^2}
  \text{.}
\end{eqnarray}

Second, we require the effective training target $F_\text{target}$ to have unit variance:
\begin{eqnarray}
  \Var_{\signal, \noise} \big[ F_\text{target}(\signal, \noise; \sigma) \big] &=& 1 \\
  \Var_{\signal, \noise} \Big[ \tfrac{1}{\cout(\sigma)} \big( \signal - \cskip(\sigma) (\signal + \noise) \big) \Big] &=& 1 \\
  \tfrac{1}{\cout(\sigma)^2} \Var_{\signal, \noise} \big[ \signal - \cskip(\sigma) (\signal + \noise) \big] &=& 1 \\
  \cout(\sigma)^2 &=& \Var_{\signal, \noise} \big[ \signal - \cskip(\sigma) (\signal + \noise) \big] \\
  \cout(\sigma)^2 &=& \Var_{\signal, \noise} \Big[ \big( 1 - \cskip(\sigma) \big) ~\signal + \cskip(\sigma) ~\noise \Big] \\
  \cout(\sigma)^2 &=& \big( 1 - \cskip(\sigma) \big)^2 ~\sdata^2 + \cskip(\sigma)^2 ~\sigma^2
  \label{eq:coutpre}
  \text{.}
\end{eqnarray}

Third, we select $\cskip(\sigma)$ to minimize $\cout(\sigma)$, so that the errors of $F_\theta$ are amplified as little as possible:
\begin{equation}
  \cskip(\sigma) = \argmin_{\cskip(\sigma)} \cout(\sigma)
  \text{.}
\end{equation}
Since $\cout(\sigma) \ge 0$, we can equivalently write
\begin{equation}
  \cskip(\sigma) = \argmin_{\cskip(\sigma)} \cout(\sigma)^2
  \text{.}
\end{equation}
This is a convex optimization problem; its solution is uniquely identified by setting the derivative w.r.t. $\cskip(\sigma)$ to zero:
\begin{eqnarray}
  0 &=& \diff\big[ \cout(\sigma)^2 \big] / \diff\cskip(\sigma) \\
  0 &=& \diff\Big[ \big(1 - \cskip(\sigma)\big)^2 ~\sdata^2 + \cskip(\sigma)^2 ~\sigma^2 \Big] / \diff\cskip(\sigma) \\
  0 &=& \sdata^2 ~\diff \Big[ \big(1 - \cskip(\sigma)\big)^2 \Big] / \diff\cskip(\sigma) + \sigma^2 ~\diff\big[\cskip(\sigma)^2 \big] / \diff \cskip(\sigma) \\
  0 &=& \sdata^2 ~\big[ 2 ~\cskip(\sigma) - 2 \big] + \sigma^2 ~\big[ 2 ~\cskip(\sigma) \big] \\
  0 &=& \big( \sigma^2 + \sdata^2 \big) ~\cskip(\sigma) - \sdata^2 \\
  \cskip(\sigma) &=& \sdata^2 / \big( \sigma^2 + \sdata^2 \big)
  \label{eq:cskip}
  \text{.}
\end{eqnarray}

We can now substitute Eq.~\ref{eq:cskip} into Eq.~\ref{eq:coutpre} to complete the formula for $\cout(\sigma)$:
\begin{eqnarray}
  \cout(\sigma)^2 &=& \big( 1 - \big[ \cskip(\sigma) \big] \big)^2 ~\sdata^2 + \big[ \cskip(\sigma) \big]^2 ~\sigma^2 \\
  \cout(\sigma)^2 &=& \bigg( 1 - \bigg[ \frac{\sdata^2}{\sigma^2 + \sdata^2} \bigg] \bigg)^2 ~\sdata^2 + \bigg[ \frac{\sdata^2}{\sigma^2 + \sdata^2} \bigg]^2 ~\sigma^2 \\
  \cout(\sigma)^2 &=& \bigg[ \frac{\sigma^2 ~\sdata}{\sigma^2 + \sdata^2} \bigg]^2 + \bigg[ \frac{\sdata^2 ~\sigma}{\sigma^2 + \sdata^2} \bigg]^2 \\
  \cout(\sigma)^2 &=& \frac{\big( \sigma^2 ~\sdata \big)^2 + \big( \sdata^2 ~\sigma \big)^2}{\big( \sigma^2 + \sdata^2 \big)^2} \\
  \cout(\sigma)^2 &=& \frac{(\sigma \cdot \sdata)^2 ~\big( \sigma^2 + \sdata^2 \big)}{\big( \sigma^2 + \sdata^2 \big)^2} \\
  \cout(\sigma)^2 &=& \frac{(\sigma \cdot \sdata)^2}{\sigma^2 + \sdata^2} \\
  \cout(\sigma) &=& \sigma \cdot \sdata \big/ \sqrt{\sigma^2 + \sdata^2}
  \text{.}
\end{eqnarray}

Fourth, we require the effective weight $w(\sigma)$ to be uniform across noise levels:
\begin{eqnarray}
  w(\sigma) &=& 1 \\
  \lambda(\sigma) ~\cout(\sigma)^2 &=& 1 \\
  \lambda(\sigma) &=& 1 / \cout(\sigma)^2 \\
  \lambda(\sigma) &=& 1 \big/ \bigg[ \frac{\sigma \cdot \sdata}{\sqrt{\sigma^2 + \sdata^2}} \bigg]^2 \\
  \lambda(\sigma) &=& 1 \big/ \bigg[ \frac{(\sigma \cdot \sdata)^2}{\sigma^2 + \sdata^2} \bigg] \\
  \lambda(\sigma) &=& \big( \sigma^2 + \sdata^2 \big) / (\sigma \cdot \sdata)^2
  \text{.}
\end{eqnarray}

We follow previous work and initialize the output layer weights to zero. Consequently, upon initialization $F_\theta(\cdot) = 0$ and the expected value of the loss at each noise level is $1$. This can be seen by substituting the choices of $\lambda(\sigma)$ and $\cskip(\sigma)$ into Eq.~\ref{eq:lossexpanded}, considered at a fixed $\sigma$:
\begin{eqnarray}
  && \mathbb{E}_{\signal, \noise} \Big[ \lambda(\sigma) \big\lVert \cskip(\sigma) (\signal {+} \noise) + \cout(\sigma) F_\theta\big(\Fargs\big) - \signal \big\rVert^2_2 \Big] \\
  &=& \mathbb{E}_{\signal, \noise} \bigg[ \frac{\sigma^2 + \sdata^2}{(\sigma \cdot \sdata)^2} \bigg\lVert \frac{\sdata^2}{\sigma^2 + \sdata^2} (\signal {+} \noise) - \signal \bigg\rVert^2_2 \bigg] \\
  &=& \mathbb{E}_{\signal, \noise} \bigg[ \frac{\sigma^2 + \sdata^2}{(\sigma \cdot \sdata)^2} \bigg\lVert \frac{\sdata^2 \noise - \sigma^2 \signal}{\sigma^2 + \sdata^2} \bigg\rVert^2_2 \bigg] \\
  &=& \mathbb{E}_{\signal, \noise} \bigg[ \frac{1}{\sigma^2 + \sdata^2} \bigg\lVert \frac{\sdata}{\sigma} \noise - \frac{\sigma}{\sdata} \signal \bigg\rVert^2_2 \bigg] \\
  &=& \frac{1}{\sigma^2 + \sdata^2} \mathbb{E}_{\signal, \noise} \bigg[   \frac{\sdata^2}{\sigma^2} \langle \noise, \noise \rangle + \frac{\sigma^2}{\sdata^2} \langle \signal, \signal \rangle - 2 \langle \signal, \noise \rangle \bigg] \\
  &=& \frac{1}{\sigma^2 + \sdata^2} \bigg[ \frac{\sdata^2}{\sigma^2} \underbrace{\Var(\noise)}_{= \sigma^2} + \frac{\sigma^2}{\sdata^2} \underbrace{\Var(\signal)}_{= \sdata^2} - 2 \underbrace{\Cov(\signal, \noise)}_{=0} \bigg] \\
  &=& 1
\end{eqnarray}

\section{Reframing previous methods in our framework}
\label{app:reframing}

In this section, we derive the formulas shown in Table~\refpaper{tab:specifics} for previous methods, discuss the corresponding original samplers and pre-trained models, and detail the practical considerations associated with using them in our framework.

In practice, the original implementations of these methods differ considerably in terms of the definitions of model inputs and outputs, dynamic range of image data, scaling of $\xx$, and interpretation of $\sigma$.
We eliminate this variation by standardizing on a unified setup where the model always matches our definition of $F_\theta$, image data is always represented in the continuous range $[-1, 1]$, and the details of $\xx$ and $\sigma$ are always in agreement with Eq.~\refpaper{eq:odescale}.

We minimize the accumulation of floating point round-off errors by always executing Algorithms~\refpaper{alg:heun} and~\refpaper{alg:stochastic} at double precision (\texttt{float64}).
However, we still execute the network $F_\theta(\cdot)$ at single precision (\texttt{float32}) to minimize runtime and remain faithful to previous work in terms of network architecture.

\subsection{Variance preserving formulation}
\label{app:reframingvp}

\subsubsection{VP sampling}

Song~et~al.~\cite{Song2021sde} define the VP SDE (Eq.~32 in \cite{Song2021sde}) as
\begin{equation}
  \diff\xx = -\tfrac{1}{2} ~\Big( \bmin + t ~\big( \bmax - \bmin \big) \Big) ~\xx ~\diff t + \sqrt{ \bmin + t ~\big( \bmax - \bmin \big) } ~\diff \omega_t
  \text{,}
\end{equation}
which matches Eq.~\ref{eq:songsde} with the following choices for $f$ and $g$:
\begin{equation}
  \label{eq:vpfg}
  f(t) = -\tfrac{1}{2} ~\beta(t)
  \text{,}\hspace{4mm}
  g(t) = \sqrt{\beta(t)}
  \text{,}\hspace{4mm}\text{and}\hspace{4mm}
  \beta(t) = \big( \bmax - \bmin \big) ~t + \bmin
  \text{.}
\end{equation}

Let $\alpha(t)$ denote the integral of $\beta(t)$:
\begin{eqnarray}
  \alpha(t) &=& \int_0^t \beta(\xi) ~\diff\xi \\
  &=& \int_0^t \Big[ \big( \bmax - \bmin \big) ~\xi + \bmin \Big] ~\diff\xi \\
  &=& \tfrac{1}{2} ~\big( \bmax - \bmin \big) ~t^2 + \bmin ~t \\
  &=& \tfrac{1}{2} ~\bdyn ~t^2 + \bmin ~t
  \text{,}
\end{eqnarray}
where $\bdyn = \bmax - \bmin$.
We can now obtain the formula for $\sigma(t)$ by substituting Eq.~\ref{eq:vpfg} into Eq.~\ref{eq:songscale}:
\begin{eqnarray}
  \sigma(t) &=& \sqrt{\int_0^t \frac{\big[ g(\xi) \big]^2}{\big[ s(\xi) \big]^2} ~\diff\xi} \\
  &=& \sqrt{\int_0^t \frac{\big[ \sqrt{\beta(\xi)} \big]^2}{\big[ 1 / \sqrt{e^{\alpha(\xi)}} \big]^2} ~\diff\xi} \\
  &=& \sqrt{\int_0^t \frac{\beta(\xi)}{1 / e^{\alpha(\xi)}} ~\diff\xi} \\
  &=& \sqrt{\int_0^t \dot\alpha(\xi) ~e^{\alpha(\xi)} ~\diff\xi} \\
  &=& \sqrt{e^{\alpha(t)} - e^{\alpha(0)}} \\
  &=& \sqrt{e^{\frac{1}{2} \bdyn t^2 + \bmin t} - 1}
  \label{eq:vpsigma}
  \text{,}
\end{eqnarray}
which matches the ``Schedule'' row of Table~\refpaper{tab:specifics}.
Similarly for $s(t)$:
\begin{eqnarray}
  s(t) &=& \exp\left( \int_0^t \big[ f(\xi) \big] ~\diff\xi \right) \\
  &=& \exp\left( \int_0^t \big[ -\tfrac{1}{2} ~\beta(\xi) \big] ~\diff\xi \right) \\
  &=& \exp\left( -\tfrac{1}{2} \left[ \int_0^t \beta(\xi) ~\diff\xi \right] \right) \\
  &=& \exp\left( -\tfrac{1}{2} ~\alpha(t) \right) \\
  &=& 1 / \sqrt{e^{\alpha(t)}} \\
  &=& 1 / \sqrt{e^{\frac{1}{2} \bdyn t^2 + \bmin t}}
  \label{eq:vpscale}
  \text{,}
\end{eqnarray}
which matches the ``Scaling'' row of Table~\refpaper{tab:specifics}.
We can equivalently write Eq.~\ref{eq:vpscale} in a slightly simpler form by utilizing Eq.~\ref{eq:vpsigma}:
\begin{equation}
  s(t) = 1 / \sqrt{\sigma(t)^2 + 1}
  \label{eq:vpscalesimple}
  \text{.}
\end{equation}

Song~et~al.~\cite{Song2021sde} choose to distribute the sampling time steps $\{t_0, \dots, t_{N-1}\}$ at uniform intervals within $[\epsilon_\text{s}, 1]$.
This corresponds to setting
\begin{equation}
  t_{i<N} = 1 + \tfrac{i}{N-1}(\epsilon_\text{s} - 1)
  \text{,}
\end{equation}
which matches the ``Time steps'' row of Table~\refpaper{tab:specifics}.

Finally, Song~et~al.~\cite{Song2021sde} set $\bmin = 0.1$, $\bmax = 20$, and $\epsilon_\text{s} = 10^{-3}$ (Appendix~C in \cite{Song2021sde}), and choose to represent images in the range $[-1, 1]$.
These choices are readily compatible with our formulation and are reflected by the ``Parameters'' section of Table~\refpaper{tab:specifics}.

\subsubsection{VP preconditioning}

In the VP case, Song~et~al.~\cite{Song2021sde} approximate the score of $p_t(\xx)$ of Eq.~\ref{eq:songmarginal} as%
\footnote{\tiny\url{https://github.com/yang-song/score\_sde\_pytorch/blob/1618ddea340f3e4a2ed7852a0694a809775cf8d0/models/utils.py\#L144}}
\begin{equation}
  \nablaxx \log p_t(\xx) ~\approx~ \underbrace{{-}\tfrac{1}{\bar\sigma(t)} ~F_\theta\big( \xx; ~(M{-}1)t \big)}_{\score(\xx; F_\theta, t)}
  \label{eq:vpprecondorig}
  \text{,}
\end{equation}
where $M = 1000$, $F_\theta$ denotes the network, and $\bar\sigma(t)$ corresponds to the standard deviation of the perturbation kernel of Eq.~\ref{eq:songperturbation}.

Let us expand the definitions of $p_t(\xx)$ and $\bar\sigma(t)$ from Eqs.~\ref{eq:psigma} and~\ref{eq:songperturbation}, respectively, and substitute $\xx = s(t) \hat\xx$ to obtain the corresponding formula with respect to the non-scaled variable $\hat\xx$:
\begin{eqnarray}
  \nabla_{\xx} \log \big[ p\big( \xx / s(t); \sigma(t) \big) \big] &\approx& {-}\tfrac{1}{[s(t) \sigma(t)]} ~F_\theta\big( \xx; ~(M{-}1)t \big) \\
  \nabla_{[s(t) \hat\xx]} \log p\big( [s(t) ~\hat\xx] / s(t); \sigma(t) \big) &\approx& {-}\tfrac{1}{s(t) \sigma(t)} ~F_\theta\big( [s(t) ~\hat\xx]; ~(M{-}1)t \big) \\
  \tfrac{1}{s(t)} \nabla_{\hat\xx} \log p\big( \hat\xx; \sigma(t) \big) &\approx& {-}\tfrac{1}{s(t) \sigma(t)} ~F_\theta\big( s(t) ~\hat\xx; ~(M{-}1)t \big) \\
  \nabla_{\hat\xx} \log p\big( \hat\xx; \sigma(t) \big) &\approx& {-}\tfrac{1}{\sigma(t)} ~F_\theta\big( s(t) ~\hat\xx; ~(M{-}1)t \big)
  \text{.}
\end{eqnarray}

We can now replace the left-hand side with Eq.~\refpaper{eq:scoredenoiser} and expand the definition of $s(t)$ from Eq.~\ref{eq:vpscalesimple}:
\begin{eqnarray}
  \Big[ \Big( D\big( \hat\xx; \sigma(t) \big) - \hat\xx \Big) / \sigma(t)^2 \Big] &\approx& {-}\tfrac{1}{\sigma(t)} ~F_\theta\big( s(t) ~\hat\xx; ~(M{-}1)t \big) \\
  D\big( \hat\xx; \sigma(t) \big) &\approx& \hat\xx - \sigma(t) ~F_\theta\big( s(t) ~\hat\xx; ~(M{-}1)t \big) \\
  D\big( \hat\xx; \sigma(t) \big) &\approx& \hat\xx - \sigma(t) ~F_\theta\bigg( \bigg[ \tfrac{1}{\sqrt{\sigma(t)^2 + 1}} \bigg] ~\hat\xx; ~(M{-}1)t \bigg)
  \text{,}
\end{eqnarray}
which can be further expressed in terms of $\sigma$ by replacing $\sigma(t) \rightarrow \sigma$ and $t \rightarrow \sigma^{-1}(\sigma)$:
\begin{equation}
  D(\hat\xx; \sigma) ~\approx~ \hat\xx - \sigma ~F_\theta\Big( \tfrac{1}{\sqrt{\sigma^2 + 1}} ~\hat\xx; ~(M{-}1) ~\sigma^{-1}(\sigma) \Big)
  \label{eq:vpprecondtemp}
  \text{.}
\end{equation}

We adopt the right-hand side of Eq.~\ref{eq:vpprecondtemp} as the definition of $D_\theta$, obtaining
\begin{equation}
  D_\theta(\hat\xx; \sigma) = \underbrace{1~\cdot}_{\cskip}\hat\xx ~\underbrace{-~\sigma}_{\cout} \,\cdot ~F_\theta\Big( \underbrace{\tfrac{1}{\sqrt{\sigma^2 + 1}}}_{\cin} \,\cdot~\hat\xx; ~\underbrace{(M{-}1)~\sigma^{-1}(\sigma)}_{\cnoise} \Big)
  \label{eq:vpprecond}
  \text{,}
\end{equation}
where $\cskip$, $\cout$, $\cin$, and $\cnoise$ match the ``Network and preconditioning'' section of Table~\refpaper{tab:specifics}.

\subsubsection{VP training}

Song~et~al.~\cite{Song2021sde} define their training loss as%
\footnote{\tiny\url{https://github.com/yang-song/score\_sde\_pytorch/blob/1618ddea340f3e4a2ed7852a0694a809775cf8d0/losses.py\#L73}}
\begin{equation}
  \mathbb{E}_{t \sim \mathcal{U}(\epsilon_\text{t}, 1), \signal \sim \pdata, \bar\noise \sim \mathcal{N}(\boldzero, \boldi)} \Big[ \big\lVert \bar\sigma(t) ~\score\big( s(t) ~\signal + \bar\sigma(t) ~\bar\noise; ~F_\theta, t \big) + \bar\noise \big\rVert^2_2 \Big]
  \text{,}
\end{equation}
where the definition of $\score(\cdot)$ is the same as in Eq.~\ref{eq:vpprecondorig}.
Let us simplify the formula by substituting $\bar\sigma(t) = s(t) \sigma(t)$ and $\bar\noise = \noise / \sigma(t)$, where $\noise \sim \mathcal{N}(\boldzero, \sigma(t)^2 \boldi)$:
\begin{eqnarray}
  && \mathbb{E}_{t, \signal, \bar\noise} \Big[ \big\lVert s(t) \sigma(t) ~\score\big( s(t) ~\signal + [s(t)\sigma(t)] ~\bar\noise; ~F_\theta, t \big) + \bar\noise \big\rVert^2_2 \Big] \\
  &=& \mathbb{E}_{t, \signal, \noise} \Big[ \big\lVert s(t) \sigma(t) ~\score\big( s(t) ~\signal + s(t)\sigma(t) ~[\noise / \sigma(t)]; ~F_\theta, t \big) + [\noise / \sigma(t)] \big\rVert^2_2 \Big] \\
  &=& \mathbb{E}_{t, \signal, \noise} \Big[ \big\lVert s(t) \sigma(t) ~\score\big( s(t) ~(\signal + \noise); ~F_\theta, t \big) + \noise / \sigma(t) \big\rVert^2_2 \Big]
  \label{eq:vplosstemp}
  \text{.}
\end{eqnarray}

We can express $\score(\cdot)$ in terms of $D_\theta(\cdot)$ by combining Eqs.~\ref{eq:vpprecondorig}, \ref{eq:vpscalesimple}, and \ref{eq:scaledscore}:
\begin{equation}
  \score\big( s(t) ~\xx; F_\theta, t \big) ~=~ \tfrac{1}{s(t) \sigma(t)^2} \Big( D_\theta \big( \xx; \sigma(t) \big) - \xx \Big)
  \text{.}
\end{equation}

Substituting this back into Eq.~\ref{eq:vplosstemp} gives
\begin{eqnarray}
  && \mathbb{E}_{t, \signal, \noise} \Big[ \big\lVert s(t) \sigma(t) ~\Big[ \tfrac{1}{s(t) \sigma(t)^2} \Big( D_\theta \big( \signal + \noise; \sigma(t) \big) - (\signal + \noise) \Big) \Big] + \tfrac{1}{\sigma(t)} ~\noise \big\rVert^2_2 \Big] \\
  &=& \mathbb{E}_{t, \signal, \noise} \Big[ \big\lVert \tfrac{1}{\sigma(t)} \Big( D_\theta \big( \signal + \noise; \sigma(t) \big) - (\signal + \noise) \Big) + \tfrac{1}{\sigma(t)} ~\noise \big\rVert^2_2 \Big] \\
  &=& \mathbb{E}_{t, \signal, \noise} \Big[ \tfrac{1}{\sigma(t)^2} ~\big\lVert D_\theta \big( \signal + \noise; \sigma(t) \big) - \signal \big\rVert^2_2 \Big]
  \text{.}
\end{eqnarray}

We can further express this in terms of $\sigma$ by replacing $\sigma(t) \rightarrow \sigma$ and $t \rightarrow \sigma^{-1}(\sigma)$:
\begin{equation}
  \underbrace{\mathbb{E}_{\sigma^{-1}(\sigma) \sim \mathcal{U}(\epsilon_\text{t}, 1)}}_{\ptrain} \mathbb{E}_{\signal, \noise} \Big[ \underbrace{\tfrac{1}{\sigma^2}}_{\lambda} \big\lVert D_\theta \big( \signal + \noise; \sigma \big) - \signal \big\rVert^2_2 \Big]
  \label{eq:vploss}
  \text{,}
\end{equation}
which matches Eq.~\ref{eq:totalloss} with the choices for $\ptrain$ and $\lambda$ shown in the ``Training'' section of Table~\refpaper{tab:specifics}.

\subsubsection{VP practical considerations}

The pre-trained VP model that we use on CIFAR-10 corresponds to the ``DDPM++ cont.\ (VP)'' checkpoint%
\footnote{\tiny\texttt{vp/cifar10\_ddpmpp\_continuous/checkpoint\_8.pth}, \url{https://drive.google.com/drive/folders/1xYjVMx10N9ivQQBIsEoXEeu9nvSGTBrC}}
provided by Song~et~al.~\cite{Song2021sde}.
It contains a total of 62 million trainable parameters and supports a continuous range of noise levels $\sigma \in \big[ \sigma(\epsilon_\text{t}), \sigma(1) \big] \approx [0.001, 152]$, i.e., wider than our preferred sampling range $[0.002, 80]$.
We import the model directly as $F_\theta(\cdot)$ and run Algorithms~\refpaper{alg:heun} and~\refpaper{alg:stochastic} using the definitions in Table~\refpaper{tab:specifics}.

In Figure~\refpaper{fig:OdePlotNfe}a, the differences between the original sampler (blue) and our reimplementation (orange) are explained by oversights in the implementation of Song~et~al.~\cite{Song2021sde}, also noted by Jolicoeur-Martineau~et~al.~\cite{Jolicoeur2021} (Appendix~D in \cite{Jolicoeur2021}).
First, the original sampler employs an incorrect multiplier%
\footnote{\tiny\url{https://github.com/yang-song/score\_sde\_pytorch/blob/1618ddea340f3e4a2ed7852a0694a809775cf8d0/sampling.py\#L182}}
in the Euler step: it multiplies $\diff\xx / \diff t$ by $-1 / N$ instead of $(\epsilon_\text{s} - 1) / (N - 1)$.
Second, it either overshoots or undershoots on the last step by going from $t_{N-1} = \epsilon_\text{s}$ to $t_N = \epsilon_\text{s} - 1 / N$, where $t_N < 0$ when $N < 1000$.
In practice, this means that the generated images contain noticeable noise that becomes quite severe with, e.g., $N = 128$.
Our formulation avoids these issues, because the step sizes in Algorithm~\refpaper{alg:heun} are computed consistently from $\{t_i\}$ and $t_N = 0$.

\subsection{Variance exploding formulation}
\label{app:reframingve}

\subsubsection{VE sampling in theory}

Song~et~al.~\cite{Song2021sde} define the VE SDE (Eq.~30 in \cite{Song2021sde}) as
\begin{equation}
  \diff\xx = \smin \bigg( \frac{\smax}{\smin} \bigg)^t \sqrt{2 \log \frac{\smax}{\smin}} ~\diff \omega_t
  \text{,}
\end{equation}
which matches Eq.~\ref{eq:songsde} with
\begin{equation}
  \label{eq:vefg}
  f(t) = 0
  \text{,}\hspace{4mm}
  g(t) = \smin \sqrt{2\log\sdyn} ~\sdyn^t
  \text{,}\hspace{4mm}\text{and}\hspace{4mm}
  \sdyn = \smax / \smin
  \text{.}
\end{equation}

The VE formulation does not employ scaling, which can be easily seen from Eq.~\ref{eq:songscale}:
\begin{equation}
  s(t) = \exp\left( \int_0^t \big[ f(\xi) \big] ~\diff\xi \right) = \exp\left( \int_0^t \big[ 0 \big] ~\diff\xi \right) = \exp(0) = 1
  \text{.}
\end{equation}

Substituting Eq.~\ref{eq:vefg} into Eq.~\ref{eq:songscale} suggests the following form for $\sigma(t)$:
\begin{eqnarray}
  \sigma(t) &=& \sqrt{\int_0^t \frac{\big[ g(\xi) \big]^2}{\big[ s(\xi) \big]^2} ~\diff\xi} \\
  &=& \sqrt{\int_0^t \frac{\big[ \smin \sqrt{2\log\sdyn} ~\sdyn^\xi \big]^2}{\big[ 1 \big]^2} ~\diff\xi} \\
  &=& \sqrt{\int_0^t \smin^2 ~\big[ 2\log\sdyn \big] ~\big[ \sdyn^{2\xi} \big] ~\diff\xi} \\
  &=& \smin \sqrt{\int_0^t \Big[ \log \big( \sdyn^2 \big) \Big] ~\Big[ \big( \sdyn^2 \big)^\xi \Big] ~\diff\xi} \\
  &=& \smin \sqrt{\big( \sdyn^2 \big)^t - \big( \sdyn^2 \big)^0} \\
  &=& \smin \sqrt{\sdyn^{2t} - 1}
  \label{eq:fakevesigma}
  \text{.}
\end{eqnarray}

Eq.~\ref{eq:fakevesigma} is consistent with the perturbation kernel reported by Song~et~al. (Eq.~29 in \cite{Song2021sde}).
However, we note that this does not fulfill their intended definition of $\sigma(t) = \smin ~\big( \tfrac{\smax}{\smin} \big)^t$ (Appendix~C in~\cite{Song2021sde}).

\subsubsection{VE sampling in practice}

The original implementation%
\footnote{\small\url{https://github.com/yang-song/score\_sde\_pytorch}}
of Song~et~al.~\cite{Song2021sde} uses reverse diffusion predictor%
\footnote{\tiny\url{https://github.com/yang-song/score\_sde\_pytorch/blob/1618ddea340f3e4a2ed7852a0694a809775cf8d0/sampling.py\#L191}}
to integrate discretized reverse probability flow%
\footnote{\tiny\url{https://github.com/yang-song/score\_sde\_pytorch/blob/1618ddea340f3e4a2ed7852a0694a809775cf8d0/sde\_lib.py\#L102}}
of discretized VE SDE%
\footnote{\tiny\url{https://github.com/yang-song/score\_sde\_pytorch/blob/1618ddea340f3e4a2ed7852a0694a809775cf8d0/sde\_lib.py\#L246}}.
Put together, these yield the following update rule for $\xx_{i+1}$:
\begin{equation}
  \label{eq:vestep}
  \xx_{i+1} = \xx_i + \tfrac{1}{2} ~\big( \bar\sigma_i^2 - \bar\sigma_{i+1}^2 \big) ~\nablaxx \log \bar p_i (\xx)
  \text{,}
\end{equation}
where
\begin{equation}
  \bar\sigma_{i<N} = \smin ~\bigg( \frac{\smax}{\smin} \bigg)^{1 - i / (N-1)}
  \hspace{4mm}\text{and}\hspace{4mm}
  \bar\sigma_N = 0
  \text{.}
\end{equation}

Interestingly, Eq.~\ref{eq:vestep} is identical to the Euler iteration of our ODE with the following choices:
\begin{equation}
  s(t) = 1
  \text{,}\hspace{4mm}
  \sigma(t) = \sqrt{t}
  \text{,}\hspace{4mm}\text{and}\hspace{4mm}
  t_i = \bar\sigma_i^2
  \text{.}
\end{equation}

These formulas match the ``Sampling'' section of Table~\refpaper{tab:specifics}, and their correctness can be verified by substituting them into line 5 of Algorithm~\refpaper{alg:heun}:
\begin{eqnarray}
  \xx_{i+1} &=& \xx_i + (t_{i+1} - t_i) ~\boldsymbol{d}_i \\
  &=& \xx_i + (t_{i+1} - t_i) \bigg[ \bigg( \frac{\dot\sigma(t)}{\sigma(t)} + \frac{\dot s(t)}{s(t)} \bigg) \xx - \frac{\dot\sigma(t) s(t)}{\sigma(t)} D \bigg( \frac{\xx}{s(t)}; \sigma(t) \bigg) \bigg] \\
  &=& \xx_i + (t_{i+1} - t_i) \bigg[ \frac{\dot\sigma(t)}{\sigma(t)} ~\xx - \frac{\dot\sigma(t)}{\sigma(t)} ~D \big( \xx; \sigma(t) \big) \bigg] \\
  &=& \xx_i - (t_{i+1} - t_i) ~\dot\sigma(t) ~\sigma(t) \bigg[ \Big( D \big( \xx; \sigma(t) \big) - \xx \Big) \big/ \sigma(t)^2 \bigg] \\
  &=& \xx_i - (t_{i+1} - t_i) ~\dot\sigma(t) ~\sigma(t) ~\nablaxx \log p\big( \xx; \sigma(t) \big) \\
  &=& \xx_i - (t_{i+1} - t_i) \Big[ \tfrac{1}{2\sqrt{t}} \Big] \Big[ \sqrt{t} \Big] ~\nablaxx \log p\big( \xx; \sigma(t) \big) \\
  &=& \xx_i + \tfrac{1}{2} ~(t_i - t_{i+1}) ~\nablaxx \log p\big( \xx; \sigma(t) \big) \\
  &=& \xx_i + \tfrac{1}{2} ~\big( \bar\sigma_i^2 - \bar\sigma_{i+1}^2 \big) ~\nablaxx \log p\big( \xx; \sigma(t) \big)
  \text{,}
\end{eqnarray}
which is made identical to Eq.~\ref{eq:vestep} by the choice $\bar p_i(\xx) = p\big( \xx; \sigma(t_i) \big)$.

Finally, Song~et~al.~\cite{Song2021sde} set $\smin = 0.01$ and $\smax = 50$ for CIFAR-10 (Appendix~C in \cite{Song2021sde}), and choose to represent their images in the range $[0, 1]$ to match previous SMLD models.
Since our standardized range $[-1, 1]$ is twice as large, we must multiply $\smin$ and $\smax$ by 2$\times$ to compensate.
The ``Parameters'' section of Table~\refpaper{tab:specifics} reflects these adjusted values.

\subsubsection{VE preconditioning}

In the VE case, Song~et~al.~\cite{Song2021sde} approximate the score of $p_t(\xx)$ of Eq.~\ref{eq:songmarginal} directly as%
\footnote{\tiny\url{https://github.com/yang-song/score\_sde\_pytorch/blob/1618ddea340f3e4a2ed7852a0694a809775cf8d0/models/utils.py\#L163}}
\begin{equation}
  \nablaxx \log p_t(\xx) ~\approx~ \bar{F}_\theta\big( \xx; \sigma(t) \big)
  \label{eq:vescoreorig}
  \text{,}
\end{equation}
where the network $\bar{F}_\theta$ is designed to include additional pre-%
\footnote{\tiny\url{https://github.com/yang-song/score\_sde\_pytorch/blob/1618ddea340f3e4a2ed7852a0694a809775cf8d0/models/ncsnpp.py\#L239}}
and%
\footnote{\tiny\url{https://github.com/yang-song/score\_sde\_pytorch/blob/1618ddea340f3e4a2ed7852a0694a809775cf8d0/models/ncsnpp.py\#L261}}
postprocessing%
\footnote{\tiny\url{https://github.com/yang-song/score\_sde\_pytorch/blob/1618ddea340f3e4a2ed7852a0694a809775cf8d0/models/ncsnpp.py\#L379}}
steps:
\begin{equation}
  \bar{F}_\theta\big( \xx; \sigma \big) ~=~ \tfrac{1}{\sigma} ~F_\theta\big( 2 \xx {-} 1; \log(\sigma) \big)
  \label{eq:veprecondorig}
  \text{.}
\end{equation}
For consistency, we handle the pre- and postprocessing using $\{\cskip, \cout, \cin, \cnoise\}$ as opposed to baking them into the network itself.

We cannot use Eqs.~\ref{eq:vescoreorig} and~\ref{eq:veprecondorig} directly in our framework, however, because they assume that the images are represented in range $[0, 1]$.
In order to use $[-1, 1]$ instead, we replace $p_t(\xx) \rightarrow p_t(2 \xx {-} 1)$, $\xx \rightarrow \tfrac{1}{2} \xx + \tfrac{1}{2}$ and $\sigma \rightarrow \tfrac{1}{2} \sigma$:
\begin{eqnarray}
  \nabla_{[\frac{1}{2} \xx + \frac{1}{2}]} \log p_t \big( 2 \big[ \tfrac{1}{2} \xx + \tfrac{1}{2} \big] {-} 1 \big) &\approx& \tfrac{1}{[\frac{1}{2} \sigma]} ~F_\theta\big( 2 \big[ \tfrac{1}{2} \xx + \tfrac{1}{2} \big] {-} 1; \log \big[ \tfrac{1}{2} \sigma \big] \big) \\
  2 ~\nabla_{\xx} \log p_t(\xx) &\approx& \tfrac{2}{\sigma} ~F_\theta\Big( \xx; \log \big( \tfrac{1}{2} \sigma \big) \Big) \\
  \nabla_{\xx} \log p(\xx; \sigma) &\approx& \tfrac{1}{\sigma} ~F_\theta\Big( \xx; \log \big( \tfrac{1}{2} \sigma \big) \Big)
  \label{eq:veprecondtemp}
  \text{.}
\end{eqnarray}

We can now express the model in terms of $D_\theta(\cdot)$ by replacing the left-hand side of Eq.~\ref{eq:veprecondtemp} with Eq.~\refpaper{eq:scoredenoiser}:
\begin{eqnarray}
  \Big( D_\theta \big( \xx; \sigma \big) - \xx \Big) / \sigma^2 &=& \tfrac{1}{\sigma} ~F_\theta\Big( \xx; \log\big( \tfrac{1}{2} \sigma \big) \Big) \label{eq:vedenoisertemp} \\
  D_\theta \big( \xx; \sigma \big) &=& \underbrace{1~\cdot}_{\cskip} \xx + \underbrace{\sigma~\cdot}_{\cout} F_\theta\Big(\underbrace{1~\cdot}_{\cin} \xx; ~\underbrace{\log \big( \tfrac{1}{2} \sigma \big)}_{\cnoise} \Big)
  \label{eq:veprecond}
  \text{,}
\end{eqnarray}
where $\cskip$, $\cout$, $\cin$, and $\cnoise$ match the ``Network and preconditioning'' section of Table~\refpaper{tab:specifics}.

\subsubsection{VE training}

Song~et~al.~\cite{Song2021sde} define their training loss similarly for VP and VE, so we can reuse Eq.~\ref{eq:vplosstemp} by borrowing the definition of $\score(\cdot)$ from Eq.~\ref{eq:vedenoisertemp}:
\begin{eqnarray}
  && \mathbb{E}_{t, \signal, \noise} \Big[ \big\lVert s(t) \sigma(t) ~\score\big( s(t) ~(\signal + \noise); ~F_\theta, t \big) + \noise / \sigma(t) \big\rVert^2_2 \Big] \\
  &=& \mathbb{E}_{t, \signal, \noise} \Big[ \big\lVert \sigma(t) ~\score\big( \signal + \noise; ~F_\theta, t \big) + \noise / \sigma(t) \big\rVert^2_2 \Big] \\
  &=& \mathbb{E}_{t, \signal, \noise} \Big[ \big\lVert \sigma(t) ~\Big[ \Big( D_\theta \big( \signal + \noise; \sigma(t) \big) - (\signal + \noise) \Big) / \sigma(t)^2 \Big] + \noise / \sigma(t) \big\rVert^2_2 \Big] \\
  &=& \mathbb{E}_{t, \signal, \noise} \Big[ \tfrac{1}{\sigma(t)^2} ~\big\lVert D_\theta \big( \signal + \noise; \sigma(t) \big) - \signal \big\rVert^2_2 \Big]
  \label{eq:velosstemp}
  \text{.}
\end{eqnarray}

For VE training, the original implementation%
\footnote{\tiny\url{https://github.com/yang-song/score\_sde\_pytorch/blob/1618ddea340f3e4a2ed7852a0694a809775cf8d0/sde_lib.py\#L234}}
defines $\sigma(t) = \smin ~\big( \tfrac{\smax}{\smin} \big)^t$.
We can thus rewrite Eq.~\ref{eq:velosstemp} as
\begin{equation}
  \underbrace{\mathbb{E}_{\ln(\sigma) \sim \mathcal{U}( \ln(\smin), \ln(\smax))}}_{\ptrain} \mathbb{E}_{\signal, \noise} \Big[ \underbrace{\tfrac{1}{\sigma^2}}_{\lambda} \big\lVert D_\theta \big( \signal + \noise; \sigma \big) - \signal \big\rVert^2_2 \Big]
  \text{,}
\end{equation}
which matches Eq.~\ref{eq:totalloss} with the choices for $\ptrain$ and $\lambda$ shown in the ``Training'' section of Table~\refpaper{tab:specifics}.

\subsubsection{VE practical considerations}

The pre-trained VE model that we use on CIFAR-10 corresponds to the ``NCSN++ cont. (VE)'' checkpoint%
\footnote{\tiny\texttt{ve/cifar10\_ncsnpp\_continuous/checkpoint\_24.pth}, \url{https://drive.google.com/drive/folders/1b0gy\_LLgO\_DaQBgoWXwlVnL\_rcAUgREh}}
provided by Song~et~al.~\cite{Song2021sde}.
It contains a total of 63 million trainable parameters and supports a continuous range of noise levels $\sigma \in \big[ \sigma(\epsilon_\text{t}), \sigma(1) \big] \approx [0.02, 100]$.
This is narrower than our preferred sampling range $[0.002, 80]$, so we set $\smin = 0.02$ in all related experiments.
Note that this limitation is lifted by our training improvements in config~\textsc{e}, so we revert back to using $\smin = 0.002$ with configs~\textsc{e} and~\textsc{f} in Table~\refpaper{tab:TrainingTable}.
When importing the model, we remove the pre- and postprocessing steps shown in Eq.~\ref{eq:veprecondorig} to stay consistent with the definition of $F_\theta(\cdot)$ in Eq.~\ref{eq:veprecond}.
With these changes, we can run Algorithms~\refpaper{alg:heun} and~\refpaper{alg:stochastic} using the definitions in Table~\refpaper{tab:specifics}.

In Figure~\refpaper{fig:OdePlotNfe}b, the differences between the original sampler (blue) and our reimplementation (orange) are explained by floating point round-off errors that the original implementation suffers from at high step counts.
Our results are more accurate in these cases because we represent $\xx_i$ at double precision in Algorithm~\refpaper{alg:heun}.

\subsection{Improved DDPM and DDIM}
\label{app:reframingiddpm}

\subsubsection{DDIM ODE formulation}
\label{app:ddim}

Song~et~al.~\cite{Song2020ddim} make the observation that their deterministic DDIM sampler can be expressed as Euler integration of the following ODE (Eq.~14 in \cite{Song2020ddim}):
\begin{equation}
  \label{eq:ddimode}
  \diff\xx(t) = \epsilon_\theta^{(t)} \left( \frac{\xx(t)}{\sqrt{\sigma(t)^2 + 1}} \right) ~\diff\sigma(t)
  \text{,}
\end{equation}
where $\xx(t)$ is a scaled version of the iterate that appears in their discrete update formula (Eq.~10 in \cite{Song2020ddim}) and $\epsilon_\theta$ is a model trained to predict the normalized noise vector, i.e., $\epsilon_\theta^{(t)}\big( \xx(t) / \sqrt{\sigma(t)^2 + 1} \big) \approx \noise(t) / \sigma(t)$ for $\xx(t) = \signal(t) + \noise(t)$.
In our formulation, $D_\theta$ is trained to approximate the clean signal, i.e., $D_\theta\big( \xx(t); \sigma(t) \big) \approx \signal$, so we can reinterpret $\epsilon_\theta$ in terms of $D_\theta$ as follows:
\begin{eqnarray}
  \noise(t) &=& \xx(t) - \signal(t) \\
  \big[ \noise(t) / \sigma(t) \big] &=& \big( \xx(t) - \big[ \signal(t) \big] \big) / \sigma(t) \\
  \epsilon_\theta^{(t)} \big( \xx(t) / \sqrt{\sigma(t)^2 + 1} \big) &=& \big( \xx(t) - D_\theta \big( \xx(t); \sigma(t) \big) \big) / \sigma(t)
  \text{.}
\end{eqnarray}

Assuming ideal $\epsilon(\cdot)$ and $D(\cdot)$ in $L_2$ sense, we can further simplify the above formula using Eq.~\refpaper{eq:scoredenoiser}:
\begin{eqnarray}
  \epsilon^{(t)} \big( \xx(t) / \sqrt{\sigma(t)^2 + 1} \big) &=& \big( \xx(t) - D \big( \xx(t); \sigma(t) \big) \big) / \sigma(t) \label{eq:ddimdenoiser} \\
  &=& -\sigma(t) ~\Big[ \Big( D \big( \xx(t); \sigma(t) \big) - \xx(t) \Big) / \sigma(t)^2 \Big] \\
  &=& -\sigma(t) ~\nabla_{\xx(t)} \log p\big( \xx(t); \sigma(t) \big)
  \label{eq:ddimeps}
  \text{.}
\end{eqnarray}

Substituting Eq.~\ref{eq:ddimeps} back into Eq.~\ref{eq:ddimode} gives
\begin{equation}
  \diff\xx(t) = -\sigma(t) ~\nabla_{\xx(t)} \log p\big( \xx(t); \sigma(t) \big) ~\diff\sigma(t)
  \text{,}
\end{equation}
which we can further simplify by setting $\sigma(t) = t$:
\begin{equation}
  \diff\xx = -t ~\nablaxx \log p\big( \xx; \sigma(t) \big) ~\diff t
  \text{.}
\end{equation}
This matches our Eq.~\refpaper{eq:odescale} with $s(t) = 1$ and $\sigma(t) = t$, reflected by the ``Sampling'' section of Table~\refpaper{tab:specifics}.

\subsubsection{iDDPM time step discretization}

The original DDPM formulation of Ho~et~al.~\cite{Ho2020} defines the forward process (Eq.~2 in \cite{Ho2020}) as a Markov chain that gradually adds Gaussian noise to $\bar\xx_0 \sim \pdata$ according to a discrete variance schedule $\{\beta_1, \dots, \beta_T\}$:
\begin{equation}
  q(\bar\xx_t ~|~ \bar\xx_{t-1}) = \mathcal{N}\big( \bar\xx_t; ~\sqrt{1 - \beta_t} ~\bar\xx_{t-1}, ~\beta_t ~\boldi \big)
  \text{.}
\end{equation}

The corresponding transition probability from $\bar\xx_0$ to $\bar\xx_t$ (Eq.~4 in \cite{Ho2020}) is given by
\begin{equation}
  \label{eq:ddpmtrans}
  q(\bar\xx_t ~|~ \bar\xx_0) = \mathcal{N}\big( \bar\xx_t; ~\sqrt{\bar\alpha_t} ~\bar\xx_0, ~(1 - \bar\alpha_t) ~\boldi \big)
  \text{,}\hspace{4mm}\text{where}\hspace{4mm}
  \bar\alpha_t = \prod_{s=1}^t ~(1 - \beta_s)
  \text{.}
\end{equation}

Ho~et~al.~\cite{Ho2020} define $\{\beta_t\}$ based on a linear schedule and then calculate the corresponding $\{\bar\alpha_t\}$ from Eq.~\ref{eq:ddpmtrans}.
Alternatively, one can also define $\{\bar\alpha_t\}$ first and then solve for $\{\beta_t\}$:
\begin{eqnarray}
  \label{eq:ddpmbeta}
  \bar\alpha_t &=& \prod_{s=1}^t ~(1 - \beta_s) \\
  \bar\alpha_t &=& \bar\alpha_{t-1} ~(1 - \beta_t) \\
  \beta_t &=& 1 - \frac{\bar\alpha_t}{\bar\alpha_{t-1}}
  \text{.}
\end{eqnarray}

The improved DDPM formulation of Nichol~and~Dhariwal~\cite{Nichol2021a} employs a cosine schedule for $\bar\alpha_t$ (Eq.~17 in \cite{Nichol2021a}), defined as
\begin{equation}
  \bar\alpha_t = \frac{f(t)}{f(0)}
  \text{,}\hspace{4mm}\text{where}\hspace{4mm}
  f(t) = \cos^2 \bigg( \frac{t/T + s}{1 + s} \cdot \frac{\pi}{2} \bigg)
  \text{,}
\end{equation}
where $s = 0.008$.
In their implementation%
\footnote{\small\url{https://github.com/openai/improved-diffusion}},
however, Nichol~et~al. leave out the division by $f(0)$ and simply define%
\footnote{\tiny\url{https://github.com/openai/improved-diffusion/blob/783b6740edb79fdb7d063250db2c51cc9545dcd1/improved_diffusion/gaussian_diffusion.py\#L39}}
\begin{equation}
  \label{eq:ddpmalpha}
  \bar\alpha_t = \cos^2 \bigg( \frac{t/T + s}{1 + s} \cdot \frac{\pi}{2} \bigg)
  \text{.}
\end{equation}

To prevent singularities near $t = T$, they also clamp $\beta_t$ to $0.999$.
We can express the clamping in terms of $\bar\alpha_t$ by utilizing Eq.~\ref{eq:ddpmtrans} and Eq.~\ref{eq:ddpmbeta}:
\begin{eqnarray}
  \label{eq:ddpmclamp}
  \bar\alpha'_t &=& \prod_{s=1}^t ~\big( 1 - [\beta'_s] \big) \\
  &=& \prod_{s=1}^t ~\Big( 1 - \min\big( [\beta_s], ~0.999) \Big) \\
  &=& \prod_{s=1}^t ~\bigg( 1 - \min\bigg( 1 - \frac{\bar\alpha_s}{\bar\alpha_{s-1}}, ~0.999 \bigg) \bigg) \\
  &=& \prod_{s=1}^t ~\max\bigg( \frac{\bar\alpha_s}{\bar\alpha_{s-1}}, ~0.001 \bigg)
  \text{.}
\end{eqnarray}

Let us now reinterpret the above formulas in our unified framework.
Recall from Table~\refpaper{tab:specifics} that we denote the original iDDPM sampling steps by $\{u_j\}$ in the order of descending noise level $\sigma(u_j)$, where $j \in \{0, \dots, M\}$.
To harmonize the notation of Eq.~\ref{eq:ddpmtrans}, Eq.~\ref{eq:ddpmalpha}, and Eq.~\ref{eq:ddpmclamp}, we thus have to replace $T \longrightarrow M$ and $t \longrightarrow M-j$:
\begin{eqnarray}
  q(\bar\xx_j ~|~ \bar\xx_M) &=& \mathcal{N}\big( \bar\xx_j; ~\sqrt{\bar\alpha'_j} ~\bar\xx_M, ~(1 - \bar\alpha'_j) ~\boldi \big) \label{eq:ddpmtransnew} \text{,} \\[2mm]
  \bar\alpha_j &=& \cos^2 \bigg( \frac{(M - j) / M + C_2}{1 + C_2} \cdot \frac{\pi}{2} \bigg) \label{eq:ddpmalphanew} \text{,}\hspace{4mm}\text{and} \\
  \bar\alpha'_j &=& \prod_{s=M-1}^j ~\max\bigg( \frac{\bar\alpha_j}{\bar\alpha_{j+1}}, ~C_1 \bigg) ~=~ \bar\alpha'_{j+1} ~\max\bigg( \frac{\bar\alpha_j}{\bar\alpha_{j+1}}, ~C_1 \bigg) \label{eq:ddpmclampnew}
  \text{,}
\end{eqnarray}
where the constants are $C_1 = 0.001$ and $C_2 = 0.008$.

We can further simplify Eq.~\ref{eq:ddpmalphanew}:
\begin{eqnarray}
  \bar\alpha_j &=& \cos^2 \bigg( \frac{(M - j) / M + C_2}{1 + C_2} \cdot \frac{\pi}{2} \bigg) \\
  &=& \cos^2 \bigg( \frac{\pi}{2} ~\frac{(1 + C_2) - j / M}{1 + C_2} \bigg)\\
  &=& \cos^2 \bigg( \frac{\pi}{2} - \frac{\pi}{2} ~\frac{j}{M (1 + C_2)} \bigg)\\
  &=& \sin^2 \bigg( \frac{\pi}{2} ~\frac{j}{M (1 + C_2)} \bigg)
  \text{,}
\end{eqnarray}
giving the formula shown in the ``Parameters'' section of Table~\refpaper{tab:specifics}.

To harmonize the definitions of $\xx$ and $\bar\xx$, we must match the perturbation kernel of Eq.~\ref{eq:songperturbation} with the transition probability of Eq.~\ref{eq:ddpmtransnew} for each time step $t = u_j$:
\begin{eqnarray}
  p_{0t}\big( \xx(u_j) ~|~ \xx(0) \big) &=& q(\bar\xx_j ~|~ \bar\xx_M) \\
  \mathcal{N} \big( \xx(u_j); ~s(t) ~\xx(0), ~s(u_j)^2 ~\sigma(u_j)^2 ~\boldi \big) &=& \mathcal{N}\left( \bar\xx_j; ~\sqrt{\bar\alpha'_j} ~\bar\xx_M, ~\big( 1 - \bar\alpha'_j \big) ~\boldi \right)
  \text{.}
\end{eqnarray}

Substituting $s(t) = 1$ and $\sigma(t) = t$ from Appendix~\ref{app:ddim}, as well as $\bar\xx_M = \xx(0)$:
\begin{equation}
  \mathcal{N} \big( \xx(u_j); ~\xx(0), ~u_j^2 ~\boldi \big) = \mathcal{N}\left( \bar\xx_j; ~\sqrt{\bar\alpha'_j} ~\xx(0), ~\big( 1 - \bar\alpha'_j \big) ~\boldi \right)
  \text{.}
\end{equation}

We can match the means of these two distributions by defining $\bar\xx_j = \sqrt{\bar\alpha'_j} ~\xx(u_j)$:
\begin{eqnarray}
  \mathcal{N} \big( \xx(u_j); ~\xx(0), ~u_j^2 ~\boldi \big) &=& \mathcal{N}\left( \sqrt{\bar\alpha'_j} ~\xx(u_j); ~\sqrt{\bar\alpha'_j} ~\xx(0), ~\big( 1 - \bar\alpha'_j \big) ~\boldi \right) \\
  &=& \mathcal{N}\bigg( \xx(u_j); ~\xx(0), ~\frac{1 - \bar\alpha'_j}{\bar\alpha'_j} ~\boldi \bigg)
  \text{.}
\end{eqnarray}

Matching the variances and solving for $\bar\alpha'_j$ gives
\begin{eqnarray}
  u_j^2 &=& (1 - \bar\alpha'_j) ~/~ \bar\alpha'_j \\
  u_j^2 ~\bar\alpha'_j &=& 1 - \bar\alpha'_j \\
  u_j^2 ~\bar\alpha'_j + \bar\alpha'_j &=& 1 \\
  (u_j^2 + 1) ~\bar\alpha'_j &=& 1 \\
  \bar\alpha'_j &=& 1 ~/~ (u_j^2 + 1)
  \text{.}
\end{eqnarray}

Finally, we can expand the left-hand side using Eq.~\ref{eq:ddpmclampnew} and solve for $u_{j-1}$:
\begin{eqnarray}
  \bar\alpha'_{j+1} ~\max(\bar\alpha_j / \bar\alpha_{j+1}, ~C_1) &=& 1 ~/~ (u_j^2 + 1) \\
  \bar\alpha'_j ~\max(\bar\alpha_{j-1} / \bar\alpha_j, ~C_1) &=& 1 ~/~ (u_{j-1}^2 + 1) \\
  \big[ 1 ~/~ (u_j^2 + 1) \big] ~\max(\bar\alpha_{j-1} / \bar\alpha_j, ~C_1) &=& 1 ~/~ (u_{j-1}^2 + 1) \\
  \max(\bar\alpha_{j-1} / \bar\alpha_j, ~C_1) ~(u_{j-1}^2 + 1) &=& u_j^2 + 1 \\
  u_{j-1}^2 + 1 &=& (u_j^2 + 1) ~/~ \max(\bar\alpha_{j-1} / \bar\alpha_j, ~C_1) \\
  u_{j-1} &=& \sqrt{\frac{u_j^2 + 1}{\max(\bar\alpha_{j-1} / \bar\alpha_j, ~C_1)} - 1}
  \text{,}
\end{eqnarray}
giving a recurrence formula for $\{u_j\}$, bootstrapped by $u_M = 0$, that matches the ``Time steps'' row of Table~\refpaper{tab:specifics}.

\subsubsection{iDDPM preconditioning and training}

We can solve $D_\theta(\cdot)$ from Eq.~\ref{eq:ddimdenoiser} by substituting $\sigma(t) = t$ from Appendix~\ref{app:ddim}:
\begin{eqnarray}
  \epsilon_\theta^{(j)} \left( \xx / \sqrt{\sigma^2 + 1} \right) &=& \big( \xx - D_\theta(\xx; \sigma) \big) / \sigma \\
  D_\theta(\xx; \sigma) &=& \xx - \sigma ~\epsilon_\theta^{(j)} \left( \xx / \sqrt{\sigma^2 + 1} \right)
  \text{.}
\end{eqnarray}

We choose to define $F_\theta(\cdot; j) = \epsilon_\theta^{(j)}(\cdot)$ and solve $j$ from $\sigma$ by finding the nearest $u_j$:
\begin{equation}
  D_\theta(\xx; \sigma) = \underbrace{1~\cdot}_{\cskip} \xx ~\underbrace{-~\sigma}_{\cout} \,\cdot ~F_\theta\Big( \underbrace{\tfrac{1}{\sqrt{\sigma^2 + 1}}}_{\cin} \,\cdot~\xx; ~\underbrace{\argmin_j |u_j - \sigma|}_{\cnoise} \Big)
  \label{eq:iddpmprecond}
  \text{,}
\end{equation}
where $\cskip$, $\cout$, $\cin$, and $\cnoise$ match the ``Network and preconditioning'' section of Table~\refpaper{tab:specifics}.

Note that Eq.~\ref{eq:iddpmprecond} is identical to the VP preconditioning formula in Eq.~\ref{eq:vpprecond}.
Furthermore, Nichol~and~Dhariwal~\cite{Nichol2021a} define their main training loss $L_\text{simple}$ (Eq.~14 in \cite{Nichol2021a}) the same way as Song~et~al.~\cite{Song2021sde}, with $\sigma$ drawn uniformly from $\{u_j\}$.
Thus, we can reuse Eq.~\ref{eq:vploss} with $\sigma = u_j$, $j \sim \mathcal{U}(0, M-1)$, and $\lambda(\sigma) = 1 / \sigma^2$, matching the ``Training'' section of Table~\refpaper{tab:specifics}.
In addition to $L_\text{simple}$, Nichol~and~Dhariwal~\cite{Nichol2021a} also employ a secondary loss term $L_\text{vlb}$; we refer the reader to Section~3.1 in \cite{Nichol2021a} for details.

\subsubsection{iDDPM practical considerations}

The pre-trained iDDPM model that we use on ImageNet-64 corresponds to the ``ADM (dropout)'' checkpoint%
\footnote{\small\texttt{https://openaipublic.blob.core.windows.net/diffusion/jul-2021/64x64\_diffusion.pt}}
provided by Dhariwal~and~Nichol~\cite{Dhariwal2021}.
It contains 296 million trainable parameters and supports a discrete set of $M = 1000$ noise levels $\sigma \in \{u_j\} \approx \{$20291, 642, 321, 214, 160, 128, 106, 92, 80, 71, $\dots$, 0.0064$\}$.
The fact that we can only evaluate $F_\theta$ these specific choices of $\sigma$ presents three practical challenges:
\begin{enumerate}
\item
In the context of DDIM, we must choose how to resample $\{u_j\}$ to yield $\{t_i\}$ for $N \ne M$.
Song~et~al.~\cite{Song2020ddim} employ a simple resampling scheme where $t_i = u_{k \cdot i}$ for resampling factor $k \in \mathbb{Z}^+$.
This scheme, however, requires that $1000 \equiv 0 \pmod{N}$, which limits the possible choices for $N$ considerably.
Nichol~and~Dhariwal~\cite{Nichol2021a}, on the other hand, employ a more flexible scheme where $t_i = u_j$ with $j = \lfloor (M - 1) / (N - 1) \cdot i \rfloor$.
We note, however, that in practice the values of $u_{j<8}$ are considerably larger than our preferred $\smax = 80$.
We choose to skip these values by defining $j = \lfloor j_0 + (M - 1 - j_0) / (N - 1) \cdot i \rfloor$ with $j_0 = 8$, matching the ``Time steps'' row in Table~\refpaper{tab:specifics}.
In Figure~\refpaper{fig:OdePlotNfe}c, the differences between the original sampler (blue) and our reimplementation (orange) are explained by this choice.
\item
In the context of our time step discretization (Eq.~\refpaper{eq:discretization}), we must ensure that $\sigma_i \in \{u_j\}$.
We accomplish this by rounding each $\sigma_i$ to its nearest supported counterpart, i.e., $\sigma_i \gets u_{\argmin_j |u_j - \sigma_i|}$, and setting $\smin = 0.0064 ~\approx~ u_{N-1}$.
This is sufficient, because Algorithm~\refpaper{alg:heun} only evaluates $D_\theta(\cdot; \sigma)$ with $\sigma \in \{\sigma_{i<N}\}$.
\item
In the context of our stochastic sampler, we must ensure that $\hat\odetime_i \in \{u_j\}$.
We accomplish this by replacing line~5 of Algorithm~\refpaper{alg:stochastic} with $\hat\odetime_i \gets u_{\argmin_j |u_j - (\odetime_i + \gamma_i \odetime_i)|}$.
\end{enumerate}

With these changes, we are able to import the pre-trained model directly as $F_\theta(\cdot)$ and run Algorithms~\refpaper{alg:heun} and~\refpaper{alg:stochastic} using the definitions in Table~\refpaper{tab:specifics}.
Note that the model outputs both \smash{$\epsilon_\theta(\cdot)$} and $\Sigma_\theta(\cdot)$, as described in Section~3.1 of \cite{Nichol2021a}; we use only the former and ignore the latter.

\section{Further analysis of deterministic sampling}
\label{app:deterministic}

\subsection{Truncation error analysis and choice of discretization parameters}
\label{app:truncationerror}

As discussed in Section~\refpaper{sec:deterministic},
the fundamental reason why diffusion models tend to require a large number of sampling steps is that any numerical ODE solver is necessarily an approximation; the larger the steps, the farther away we drift from the true solution at each step.
Specifically, given the value of $\xx_{i-1}$ at time step $i-1$, the solver approximates the true $\xx^*_i$ as $\xx_i$, resulting in local truncation error $\lte_i = \xx^*_i - \xx_i$.
The local errors get accumulated over the $N$ steps, ultimately leading to global truncation error $\gte_N$.

Euler's method is a first order ODE solver, meaning that $\lte_i = \mathcal{O}\left(h_i^2\right)$ for any sufficiently smooth $\xx(t)$, where $h_i = |t_i - t_{i-1}|$ is the local step size~\cite{Suli2003}.
In other words, there exist some $C$ and $H$ such that $||\lte_i|| < C h_i^2$ for every $h_i < H$, i.e., halving $h_i$ reduces $\lte_i$ by 4$\times$.
Furthermore, if we assume that $D_\theta$ is Lipschitz continuous\,---\,which is true for all network architectures considered in this paper\,---\,the global truncation error is bounded by $||\gte_N|| \le E \max_i ||\lte_i||$, where the value of $E$ depends on $N$, $t_0$, $t_N$, and the Lipschitz constant~\cite{Suli2003}.
Thus, reducing the global error for given $N$, which in turn enables reducing $N$ itself, boils down to choosing the solver and $\{t_i\}$ so that $\max_i ||\lte_i||$ is minimized.

\figOdePlotExp

To gain insight on how the local truncation error behaves in practice, we measure the values of $\lte_i$ over different noise levels using the VE-based CIFAR-10 model.
For a given noise level, we set $t_i = \sigma^{-1}(\sigma_i)$ and choose some $t_{i-1} > t_i$ depending on the case.
We then sample $\xx_{i-1}$ from $p(\xx; \sigma_{i-1})$ and estimate the true $\xx^*_i$ by performing 200 Euler steps over uniformly selected subintervals between $t_{i-1}$ and $t$.
Finally, we plot the mean and standard deviation of the root mean square error (RMSE), i.e., $||\lte_i|| / \scriptstyle\sqrt{\dim\lte}$, as a function of $\sigma_i$, averaged over 200 random samples of $\xx_{i-1}$.
Results for Euler's method are shown in Figure~\ref{fig:OdePlotExp}a, where the blue curve corresponds to uniform step size $h_\sigma = 1.25$ with respect to $\sigma$, i.e., $\sigma_{i-1} = \sigma_i + h_\sigma$ and $t_{i-1} = \sigma^{-1}(\sigma_{i-1})$.
We see that the error is very large ($\text{RMSE} \approx 0.56$) for low noise levels ($\sigma_i \le 0.5$) and considerably smaller for high noise levels.
This is in line with the common intuition that, in order to reduce $\gte_N$, the step size should be decreased monotonically with decreasing $\sigma$.
Each curve is surrounded by a shaded region that indicates standard deviation, barely visible at low values of $\sigma$.
This indicates that $\lte_i$ is nearly constant with respect to $\xx_{i-1}$, and thus there would be no benefit in varying $\{t_i\}$ schedule on a per-sample basis.

A convenient way to vary the local step size depending on the noise level is to define $\{\sigma_i\}$ as a linear resampling of some monotonically increasing, unbounded warp function $w(z)$.
In other words, $\sigma_{i<N} = w(A i + B)$ and $\sigma_N = 0$, where constants $A$ and $B$ are selected so that $\sigma_0 = \smax$ and $\sigma_{N-1} = \smin$. %
In practice, we set $\smin = \max(\sigma_\text{lo}, 0.002)$ and $\smax = \min(\sigma_\text{hi}, 80)$, where $\sigma_\text{lo}$ and $\sigma_\text{hi}$ are the lowest and highest noise levels supported by a given model, respectively; we have found these choices to perform reasonably well in practice.
Now, to balance $\lte_i$ between low and high noise levels, we can, for example, use a polynomial warp function $w(z) = z^\rho$ parameterized by the exponent $\rho$.
This choice leads to the following formula for $\{\sigma_i\}$:
\begin{equation}
\label{eq:discretizationII}
\sigma_{i<N} = \left( {\smax}^\frac{1}{\rho} + \frac{i}{N-1} \left( {\smin}^\frac{1}{\rho} - {\smax}^\frac{1}{\rho} \right) \right)^\rho, \sigma_N = 0,
\end{equation}
which reduces to uniform discretization when $\rho=1$ and gives more and more emphasis to low noise levels as $\rho$ increases.\footnote{
  In the limit, Eq.~\ref{eq:discretizationII} reduces to the same geometric sequence employed by original VE ODE when \mbox{$\rho \rightarrow \infty$}.
  Thus, our discretization can be seen as a parametric generalization of the one proposed by Song~et~al.~\cite{Song2021sde}.
}

Based on the value of $\sigma_i$, we can now compute $\sigma_{i-1} = \big( \sigma_i^{1 / \rho} - A \big)^\rho$, which enables us to visualize $\lte_i$ for different choices of $\rho$ in Figure~\ref{fig:OdePlotExp}a.
We see that increasing $\rho$ reduces the error for low noise levels ($\sigma < 10$) while increasing it for high noise levels ($\sigma > 10$).
Approximate balance is achieved at $\rho=2$, but RMSE remains relatively high ($\sim0.03$), meaning that Euler's method drifts away from the correct result by several ULPs at each step.
While the error could be reduced by increasing $N$, we would ideally like the RMSE to be well below 0.01 even with low step counts.

Heun's method introduces an additional correction step for $\xx_{i+1}$ to account for the fact that $\diff\xx / \diff\odetime$ may change between $t_i$ and $t_{i+1}$; Euler's method assumes it to be constant.
The correction leads to cubic convergence of the local truncation error, i.e., $\lte_i = \mathcal{O}\left(h_i^3\right)$, at the cost of one additional evaluation of $D_\theta$ per step.
We discuss the general family of Heun-like schemes later in Appendix~\ref{app:alpha}.
Figure~\ref{fig:OdePlotExp}b shows local truncation error for Heun's method using the same setup as Figure~\ref{fig:OdePlotExp}a.
We see that the differences in $||\lte_i||$ are generally more pronounced, which is to be expected given the quadratic vs.~cubic convergence of the two methods.
Cases where Euler's method has low RMSE tend to have even lower RMSE with Heun's method, and vice versa for cases with high RMSE.
Most remarkably, the red curve shows almost constant $\text{RMSE} \in [0.0030, 0.0045]$.
This means that the combination of Eq.~\ref{eq:discretizationII} and Heun's method is, in fact, very close to optimal with $\rho=3$.

Thus far, we have only considered the raw numerical error, i.e., component-wise deviation from the true result in RGB space.
The raw numerical error is relevant for certain use cases, e.g., image manipulation where the ODE is first evaluated in the direction of increasing $t$ and then back to $t=0$ again\,---\,in this case, $||\gte_N||$ directly tells us how much the original image degrades in the process and we can use $\rho=3$ to minimize it.
Considering the generation of novel images from scratch, however, it is reasonable to expect different noise levels to introduce different kinds of errors that may not necessarily be on equal footing considering their perceptual importance.
We investigate this in Figure~\ref{fig:OdePlotExp}c, where we plot FID as a function of $\rho$ for different models and different choices of $N$.
Note that the ImageNet-64 model was only trained for a discrete set of noise levels; in order to use it with Eq.~\ref{eq:discretizationII}, we round each $t_i$ to its nearest supported counterpart, i.e., $t'_i = u_{\argmin_j |u_j - t_i|}$.

From the plot, we can see that even though $\rho=3$ leads to relatively good FID, it can be reduced further by choosing $\rho > 3$.
This corresponds to intentionally introducing error at high noise levels to reduce it at low noise levels, which makes intuitive sense because the value of $\smax$ is somewhat arbitrary to begin with\,---\,increasing $\smax$ can have a large impact on $||\gte_N||$, but it does not affect the resulting image distribution nearly as much.
In general, we have found $\rho=7$ to perform reasonably well in all cases, and use this value in all other experiments.

\subsection{\texorpdfstring{General family of 2\textsuperscript{nd} order Runge--Kutta variants}{General family of 2nd order Runge-Kutta variants}}
\label{app:alpha}

Heun's method illustrated in Algorithm~\refpaper{alg:heun} belongs to a family of explicit two-stage 2\textsuperscript{nd} order Runge--Kutta methods, each having the same computational cost.
A common parameterization~\cite{Suli2003} of this family is,
\begin{equation}
\dd_i = f(\xx_i;t_i)\ \ \ \textrm{;} \ \ \ \xx_{i+1} = \xx_i + h\Big[\Big(1-{\tfrac{1}{2\alpha}}\Big)\dd_i+{\tfrac{1}{2\alpha}}f(\xx_i + \alpha h \dd_i;t_i+\alpha h)\Big]\textrm{,}
\end{equation}
where $h=t_{i+1}-t_i$ and $\alpha$ is a parameter that controls where the additional gradient is evaluated and how much it influences the step taken.
Setting $\alpha=1$ corresponds to Heun's method, and $\alpha=\tfrac{1}{2}$ and $\alpha=\tfrac{2}{3}$ yield so-called midpoint and Ralston methods, respectively.
All these variants differ in the kind of approximation error they incur due to the geometry of the underlying function $f$.

To establish the optimal $\alpha$ in our use case, we ran a separate series of experiments.
According to the results, it appears that $\alpha=1$ is very close to being optimal.
Nonetheless, the experimentally best choice was $\alpha=1.1$ that performed slightly better, even though values greater than one are theoretically hard to justify as they overshoot the target $t_{i+1}$.
As we have no good explanation for this observation and cannot tell if it holds in general, we chose not to make $\alpha$ a new hyperparameter and instead fixed it to $1$, corresponding exactly to Heun's method.
Further analysis is left as future work, including the possibility of having $\alpha$ vary during sampling.

An additional benefit of setting $\alpha=1$ is that it makes it possible to use pre-trained neural networks $D_\theta(\xx;\sigma)$ that have been trained only for specific values of $\sigma$.
This is because a Heun step evaluates the additional gradient at exactly $t_{i+1}$ unlike the other 2\textsuperscript{nd} order variants.
Hence it is sufficient to ensure that each $t_i$ corresponds to a value of $\sigma$ that the network was trained for.

\algAlpha

Algorithm~\ref{alg:alpha} shows the pseudocode for a general 2\textsuperscript{nd} order solver parameterized by $\alpha$.
For clarity, the pseudocode assumes the specific choices of $\sigma(t)=t$ and $s(t)=1$ that we advocate in Section~\refpaper{sec:deterministic}.
Note that the fallback to Euler step (line~11) can occur only when $\alpha \ge 1$.

\section{Further results with stochastic sampling}
\label{app:stochastic}

\subsection{Image degradation due to excessive stochastic iteration}
\label{app:degradation}

\figDegradation

Figure~\ref{fig:Degradation} illustrates the image degradation caused by excessive Langevin iteration (Section~\refpaper{sec:stochasticity}, ``Practical considerations'').
These images are generated by doing a specified number of iterations at a fixed noise level $\sigma$ so that at each iteration an equal amount of noise is added and removed.
In theory, Langevin dynamics should bring the distribution towards the ideal distribution $p(\xx;\sigma)$ but as noted in Section~\refpaper{sec:stochasticity}, 
  this holds only if the denoiser $D_\theta(\xx;\sigma)$ induces a conservative vector field in Eq.~\refpaper{eq:scoredenoiser}.

As seen in the figure, it is clear that the image distribution suffers from repeated iteration in all cases, although the exact failure mode depends on dataset and noise level.
For low noise levels (below $0.2$ or so), the images tend to oversaturate starting at 2k iterations and become fully corrupted after that.
Our heuristic of setting $\Stmin > 0$ is designed to prevent stochastic sampling altogether at very low noise levels to avoid this effect.

For high noise levels, we can see that iterating without the standard deviation correction, i.e., when $\Snoise=1.000$, the images tend to become more abstract and devoid of color at high iteration counts;
  this is especially visible in the 10k column of CIFAR-10 where the images become mostly black and white with no discernible backgrounds.
Our heuristic inflation of standard deviation by setting $\Snoise > 1$ counteracts this tendency efficiently, as seen in the corresponding images on the right hand side of the figure.
Notably, this still does not fix the oversaturation and corruption at low noise levels, suggesting multiple sources for the detrimental effects of excessive iteration.
Further research will be required to better understand the root causes of these observed effects.

\figSdePlotChurn

Figure~\ref{fig:SdePlotChurn} presents the output quality of our stochastic sampler in terms of FID as a function of $\Schurn$ at fixed NFE,
  using pre-trained networks of Song~et~al.~\cite{Song2021sde} and Dhariwal and Nichol~\cite{Dhariwal2021}.
Generally, for each case and combination of our heuristic corrections, there is an optimal amount of stochasticity after which the results start to degrade.
It can also be seen that regardless of the value of $\Schurn$, the best results are obtained by enabling all corrections, although whether $\Snoise$ or $\StminStmax$ is more important depends on the case.

\subsection{\texorpdfstring{{Stochastic sampling parameters}}{Stochastic sampling parameters}}
\label{app:stochasticparams}

\tabStochasticParams

Table~\ref{tab:StochasticParams} lists {the values for $\Schurn$, $\Stmin$, $\Stmax$, and $\Snoise$ that we used in our stochastic sampling experiments}.
These were determined with a grid search over the combinations listed in the rightmost column.
It can be seen that the optimal parameters depend on the case;
  better understanding of the degradation phenomena will hopefully give rise to more direct ways of handling the problem in the future.

\section{Implementation details}
\label{app:implementation}

We implemented our techniques in a newly written codebase, based loosely on the original implementations by
Song~et~al.\footnote{\small\url{https://github.com/yang-song/score\_sde\_pytorch}}~\cite{Song2021sde},
Dhariwal~and~Nichol\footnote{\small\url{https://github.com/openai/guided-diffusion}}~\cite{Dhariwal2021},
and Karras~et~al.\footnote{\small\url{https://github.com/NVlabs/stylegan3}}~\cite{Karras2021alias}.
We performed extensive testing to verify that our implementation produced exactly the same results as previous work, including samplers, pre-trained models, network architectures, training configurations, and evaluation.
We ran all experiments using PyTorch 1.10.0, CUDA 11.4, and CuDNN 8.2.0 on NVIDIA DGX-1's with 8 Tesla V100 GPUs each.

{Our implementation and pre-trained models are available at \url{https://github.com/NVlabs/edm}}

\subsection{FID calculation}
\label{app:fid}

We calculate FID~\cite{Heusel2017} between 50,000 generated images and all available real images, without any augmentation such as $x$-flips.
We use the pre-trained Inception-v3 model provided with StyleGAN3%
\footnote{\tiny\texttt{https://api.ngc.nvidia.com/v2/models/nvidia/research/stylegan3/versions/1/files/metrics/inception-2015-12-05.pkl}}~\cite{Karras2021alias}
that is, in turn, a direct PyTorch translation of the original TensorFlow-based model%
\footnote{\tiny\texttt{http://download.tensorflow.org/models/image/imagenet/inception-2015-12-05.tgz}}.
We have verified that our FID implementation produces identical results compared to Dhariwal~and~Nichol~\cite{Dhariwal2021} and Karras~et~al.~\cite{Karras2021alias}.
To reduce the impact of random variation, typically in the order of $\pm$2\%, we compute FID three times in each experiment and report the minimum.
We also highlight the difference between the highest and lowest achieved FID in Figures \refpaper{fig:SdePlotNfe}, \refpaper{fig:TrainingPlots}b, \ref{fig:OdePlotExp}c, and \ref{fig:SdePlotChurn}.

\subsection{Augmentation regularization}
\label{app:augmentdetails}

In Section~\refpaper{sec:training}, we propose to combat overfitting of $D_\theta$ using conditional augmentation.
We build our augmentation pipeline around the same concepts that were originally proposed by Karras~et~al.~\cite{Karras2020ada} in the context of GANs.
In practice, we employ a set of 6 geometric transformations; we have found other types of augmentations, such as color corruption and image-space filtering, to be consistently harmful for diffusion-based models.

The details of our augmentation pipeline are shown in Table~\ref{tab:AugmentPipe}.
We apply the augmentations independently to each training image $\signal \sim \pdata$ prior to adding the noise $\noise \sim \mathcal{N}(\boldzero, \sigma^2 \boldi)$.
First, we determine whether to enable or disable each augmentation based on a weighted coin toss.
The probability of enabling a given augmentation (``Prob.'' column) is fixed to 12\% for CIFAR-10 and 15\% for FFHQ and AFHQv2, except for $x$-flips that are always enabled.
We then draw 8 random parameters from their corresponding distributions (``Parameters'' column); if a given augmentation is disabled, we override the associated parameters with zero.
Based on these, we construct a homogeneous 2D transformation matrix based on the parameters (``Transformation'' column).
This transformation is applied to the image using the implementation of~\cite{Karras2020ada} that employs 2$\times$ supersampled high-quality Wavelet filters.
Finally, we construct a 9-dimensional conditioning input vector (``Conditioning'' column) and feed it to the denoiser network, in addition to the image and noise level inputs.

\tabAugmentPipe

The role of the conditioning input is to present the network with a set of auxiliary tasks; in addition to the main task of modeling $p(\xx; \sigma)$, we effectively ask the network to also model an infinite set of distributions $p(\xx; \sigma, \boldsymbol{a})$ for each possible choice of the augmentation parameters $\boldsymbol{a}$.
These auxiliary tasks provide the network with a large variety of unique training samples, preventing it from overfitting to any individual sample.
Still, the auxiliary tasks appear to be beneficial for the main task; we speculate that this is because the denoising operation itself is similar for every choice of $\boldsymbol{a}$.

We have designed the conditioning input so that zero corresponds to the case where no augmentations were applied.
During sampling, we simply set $\boldsymbol{a} = \boldzero$ to obtain results consistent with the main task.
We have not observed any leakage between the auxiliary tasks and the main task; the generated images exhibit no traces of out-of-domain geometric transformations even with $A_\text{prob} = 100$\%.
In practice, this means that we are free to choose the constants $\{A_\text{prob}, A_\text{scale}, A_\text{aniso}, A_\text{trans}\}$ any way we like as long as the results improve.
Horizontal flips serve as an interesting example.
Most of the prior work augments the training set with random $x$-flips, which is beneficial for most datasets but has the downside that any text or logos may appear mirrored in the generated images.
With our non-leaky augmentations, we get the same benefits without the downsides by executing the $x$-flip augmentation with 100\% probability.
Thus, we rely exclusively on our augmentation scheme and disable dataset $x$-flips to ensure that the generated images stay true to the original distribution.

\subsection{Training configurations}
\label{app:training}

\tabTrainingParams

Table~\ref{tab:TrainingParams} shows the exact set of hyperparameters that we used in our training experiments reported in {Section~\refpaper{sec:training}}.
{We will first detail the configurations used with CIFAR-10, FFHQ, and AFHQv2, and then discuss the training of our improved ImageNet model.}

Config~\textsc{a} {of Table~\refpaper{tab:TrainingTable} (``Baseline'')} corresponds to the original setup of Song~et~al.~\cite{Song2021sde} for the two {cases} (VP and VE), and config~\textsc{f} {(``Ours'')} corresponds to our improved setup.
We trained each model until a total of 200 million images had been drawn from the training set, abbreviated as ``200 Mimg'' in Table~\ref{tab:TrainingParams}; this corresponds to a total of $\sim$400,000 training iterations using a batch size of 512.
We saved a snapshot of the model every 2.5 million images and reported results for the snapshot that achieved the lowest FID according to our deterministic sampler with NFE $=$ 35 or NFE $=$ 79, depending on the resolution.

In config~\textsc{b}, we re-adjust the basic hyperparameters to enable faster training and obtain a more meaningful point of comparison.
Specifically, we increase the parallelism from 4 to 8 GPUs and batch size from 128 to 512 or 256, depending on the resolution.
We also disable gradient clipping, i.e., forcing $\lVert \diff\mathcal{L}(D_\theta) / \diff\theta \rVert_2 \le 1$, that we found to provide no benefit in practice.
Furthermore, we increase the learning rate from 0.0002 to 0.001 for CIFAR-10, ramping it up during the first 10 million images, and standardize the half-life of the exponential moving average of $\theta$ to 0.5 million images.
Finally, we adjust the dropout probability for each dataset as shown in Table~\ref{tab:TrainingParams} via a full grid search at 1\% increments.
Our total training time is approximately 2 days {for CIFAR-10} at 32$\times$32 resolution and 4 days {for FFHQ and AFHQv2} at 64$\times$64 resolution.

In config~\textsc{c}, we improve the expressive power of the model by removing the 4$\times$4 layers and doubling the capacity of the 16$\times$16 layers instead; we found the former to mainly contribute to overfitting, whereas the latter were critical for obtaining high-quality results.
The original models of Song~et~al.~\cite{Song2021sde} employ 128 {channels} at 64$\times$64 (where applicable) and 32$\times$32, and 256 {channels} at 16$\times$16, 8$\times$8, and 4$\times$4.
We change these numbers to 128 {channels} at 64$\times$64 (where applicable), and 256 {channels} at 32$\times$32, 16$\times$16, and 8$\times$8.
We abbreviate these counts in Table~\ref{tab:TrainingParams} as multiples of 128, listed from the highest resolution to the lowest.
In practice, this rebalancing reduces the total number of trainable parameters slightly, resulting in $\sim$56 million parameters for each model at 32$\times$32 resolution and $\sim$62 million parameters at 64$\times$64 resolution.

In {config~\textsc{d}}, we replace the original preconditioning with our improved formulas (``Network and preconditioning'' section in Table~\refpaper{tab:specifics}).
{In config~\textsc{e}}, we do the same for the noise distribution and loss weighting (``Training'' section in Table~\refpaper{tab:specifics}).
{Finally, in config~\textsc{f}}, we enable augmentation regularization as discussed in Appendix~\ref{app:augmentdetails}.
The other hyperparameters remain the same as in config~\textsc{c}.

{
With ImageNet-64, it is necessary to train considerably longer compared to the other datasets in order to reach state-of-the-art results.
To reduce the training time, we employed 32 NVIDIA Ampere GPUs (4 nodes) with a batch size of 4096 (128 per GPU) and utilized the high-performance Tensor Cores via mixed-precision FP16/FP32 training.
In practice, we store the trainable parameters as FP32 but cast them to FP16 when evaluating $F_\theta$, except for the embedding and self-attention layers, where we found the limited exponent range of FP16 to occasionally lead to stability issues.
We trained the model for two weeks, corresponding to $\sim$2500 million images drawn from the training set and $\sim$600,000 training iterations, using learning rate 0.0001, exponential moving average of 50 million images, and the same model architecture and dropout probability as Dhariwal and Nichol~\cite{Dhariwal2021}.
We did not find overfitting to be a concern, and thus chose to not employ augmentation regularization.
}

\subsection{Network architectures}
\label{app:architectures}

\tabNetworkDetails

As a result of our {training} improvements, the VP and VE cases become otherwise identical in config~\textsc{f} except for the network architecture; VP employs the \ddpmpp{} architecture while VE employs \ncsnpp{}, both of which were originally proposed by Song~et~al.~\cite{Song2021sde}.
These architectures correspond to relatively straightforward variations of the same U-net backbone with {three differences, as illustrated in Table~\ref{tab:NetworkDetails}}.
First, \ddpmpp{} employs box filter $[1, 1]$ for the upsampling and downsampling layers whereas \ncsnpp{} employs bilinear filter $[1, 3, 3, 1]$.
Second, \ddpmpp{} inherits its positional encoding scheme for the noise level directly from DDPM~\cite{Ho2020} whereas \ncsnpp{} replaces it with random Fourier features~\cite{Tancik2020fourier}.
Third, \ncsnpp{} incorporates additional residual skip connections from the input image to each block in the encoder, as explained in Appendix~H of~\cite{Song2021sde} (``progressive growing architectures'').

For class conditioning and augmentation regularization, we extend the original \ddpmpp{} and \ncsnpp{} arhictectures by introducing two optional conditioning inputs alongside the noise level input.
We represent class labels as one-hot encoded vectors that we first scale by \smash{$\sqrt{C}$}, where $C$ is the total number of classes, and then feed through a fully-connected layer.
For the augmentation parameters, we feed the conditioning inputs of Appendix~\ref{app:augmentdetails} through a fully-connected layer as-is.
We then combine the resulting feature vectors with the original noise level conditioning vector through elementwise addition.

{
For class-conditional ImageNet-64, we use the ADM architecture of Dhariwal and Nichol~\cite{Dhariwal2021} with no changes.
The model has a total of $\sim$296 million trainable parameters.
As detailed in Tables~\ref{tab:TrainingParams} and~\ref{tab:NetworkDetails}, the most notable differences to \ddpmpp{} include the use of a slightly shallower model (3 residual blocks per resolution instead of 4) with considerably more channels (e.g., 768 in the lowest resolution instead of 256), more self-attention layers interspersed throughout the network (22 instead of 6), and the use of multi-head attention (e.g., 12 heads in the lowest resolution).
We feel that the precise impact of architectural choices remains an interesting question for future work.
}

\subsection{Licenses}
\label{app:licenses}

Datasets:
\begin{itemize}
\item \makebox[24mm][l]{CIFAR-10~\cite{Krizhevsky2009cifar}:}  MIT license
\item \makebox[24mm][l]{FFHQ~\cite{Karras2018stylegan}:}       Creative Commons BY-NC-SA 4.0 license
\item \makebox[24mm][l]{AFHQv2~\cite{Choi2020afhq}:}           Creative Commons BY-NC 4.0 license
\item \makebox[24mm][l]{ImageNet~\cite{Deng2009imagenet}:}     The license status is unclear
\end{itemize}

Pre-trained models:
\begin{itemize}
\item \makebox[72mm][l]{CIFAR-10 models by Song~et~al.~\cite{Song2021sde}:}                 Apache V2.0 license
\item \makebox[72mm][l]{ImageNet-64 model by Dhariwal~and~Nichol~\cite{Dhariwal2021}:}      MIT license
\item \makebox[72mm][l]{Inception-v3 model by Szegedy~et~al.~\cite{Szegedy2016inception}:}  Apache V2.0 license
\end{itemize}

\fi

\end{document}